\documentclass[journal,12pt,onecolumn]{article}
\usepackage[english]{babel}
\usepackage{csvsimple}
\usepackage{filecontents} 
\usepackage{qrcode}
\usepackage{pdfplots}
\usepackage{amsthm}
\usepackage{amsmath,graphicx}
\usepackage{booktabs} 
\usepackage{algorithm, algpseudocode}
\usepackage{imakeidx}

\usepackage{hyperref}      
\usepackage[automake,stylemods={tree}]{glossaries-extra}
\newglossary*{basic}{Basic Terms}
\newglossary*{math}{Tools}
\newglossary*{systems}{ML Systems}
\newglossary*{mlsystems}{ML Systems}
\newglossary*{advanced}{Advanced Terms}
\usepackage[utf8]{inputenc} 
\usepackage[T1]{fontenc}    
\usepackage{url}            
\usepackage{booktabs}       
\usepackage{amsfonts}       
\usepackage{bm}
\usepackage{nicefrac}       
\usepackage{microtype}      
\usepackage{xcolor}         
\usepackage{tikz}
\usetikzlibrary{matrix,positioning}
\usepackage{cite}
\usepackage{IEEEtrantools}
\usepackage{pgfkeys,pgfcalendar}
\usetikzlibrary{shapes}
\usetikzlibrary{shadows}
\usepackage{helvet}
\usepackage{float}
\usepackage{enumitem}
\usepackage{caption}
\usepackage{tikz}
\usetikzlibrary{fit, positioning, arrows.meta} 
\usepgfplotslibrary{fillbetween}
\usetikzlibrary{intersections}

\usepackage{xspace}

\usepackage{tikz}
\usetikzlibrary{positioning,trees}
\usetikzlibrary{shapes.geometric}
\usetikzlibrary{arrows.meta}
\usetikzlibrary{calc}
\usetikzlibrary{shapes.misc}
\usetikzlibrary{trees, arrows.meta}
\tikzstyle{mybox} = [draw=red,very thick,
rectangle, rounded corners, inner sep=10pt, inner ysep=20pt]
\tikzstyle{fancytitle} =[fill=red, text=black]

\tikzset{
	treenode/.style = {shape=rectangle, rounded corners,
		draw, align=center,
		top color=white, bottom color=blue!20},
	root/.style     = {treenode, font=\Large, bottom color=red!30},
	env/.style      = {treenode, font=\ttfamily\normalsize},
	dummy/.style    = {circle,draw}
}

\definecolor{aaltoBlue}{RGB}{0, 51, 102} 

\definecolor{pinegreen}{cmyk}{0.92,0,0.59,0.25}
\definecolor{royalblue}{cmyk}{1,0.50,0,0}
\definecolor{lavander}{cmyk}{0,0.48,0,0}
\definecolor{violet}{cmyk}{0.79,0.88,0,0}

\tikzstyle{ncyan}=[circle, draw=cyan!70, thin, fill=white, scale=0.8, font=\fontsize{11}{0}\selectfont]
\tikzstyle{ngreen}=[circle,  draw=green!70, thin, fill=white, scale=0.8, font=\fontsize{11}{0}\selectfont]
\tikzstyle{nred}=[circle, draw=red!70, thin, fill=white, scale=0.8, font=\fontsize{11}{0}\selectfont]
\tikzstyle{ngray}=[circle, draw=gray!70, thin, fill=white, scale=0.55, font=\fontsize{14}{0}\selectfont]
\tikzstyle{nyellow}=[circle, draw=yellow!70, thin, fill=white, scale=0.55, font=\fontsize{14}{0}\selectfont]
\tikzstyle{norange}=[circle,  draw=orange!70, thin, fill=white, scale=0.55, font=\fontsize{10}{0}\selectfont]
\tikzstyle{npurple}=[circle,draw=purple!70, thin, fill=white, scale=0.55, font=\fontsize{10}{0}\selectfont]
\tikzstyle{nblue}=[circle, draw=blue!70, thin, fill=white, scale=0.55, font=\fontsize{10}{0}\selectfont]
\tikzstyle{nteal}=[circle,draw=teal!70, thin, fill=white, scale=0.55, font=\fontsize{10}{0}\selectfont]
\tikzstyle{nviolet}=[circle, draw=violet!70, thin, fill=white, scale=0.55, font=\fontsize{10}{0}\selectfont]
\tikzstyle{qgre}=[rectangle, draw, thin,fill=green!20, scale=0.8]
\tikzstyle{rpath}=[ultra thick, red, opacity=0.4]
\tikzstyle{legend_isps}=[rectangle, rounded corners, thin,fill=gray!20, text=blue, draw]
\usetikzlibrary{positioning}
\newtheorem{theorem}{Theorem}
\newtheorem{definition}[theorem]{Definition}

\definecolor{lightblue}{RGB}{173, 216, 230}

\newcounter{exercise}[section]

\renewcommand{\theexercise}{\arabic{section}.\arabic{exercise}}

\usepackage{listings}
\usepackage{xspace}

\usepackage{tikz}
\usetikzlibrary{positioning,trees}
\usetikzlibrary{shapes.geometric}
\usetikzlibrary{calc}
\usetikzlibrary{shapes.misc}
\usetikzlibrary{trees}
\tikzstyle{mybox} = [draw=red,very thick,
rectangle, rounded corners, inner sep=10pt, inner ysep=20pt]
\tikzstyle{fancytitle} =[fill=red, text=black]

\tikzset{
	treenode/.style = {shape=rectangle, rounded corners,
		draw, align=center,
		top color=white, bottom color=blue!20},
	root/.style     = {treenode, font=\Large, bottom color=red!30},
	env/.style      = {treenode, font=\ttfamily\normalsize},
	dummy/.style    = {circle,draw}
}

\definecolor{aaltoBlue}{RGB}{0, 51, 102} 

\definecolor{pinegreen}{cmyk}{0.92,0,0.59,0.25}
\definecolor{royalblue}{cmyk}{1,0.50,0,0}
\definecolor{lavander}{cmyk}{0,0.48,0,0}
\definecolor{violet}{cmyk}{0.79,0.88,0,0}

\tikzstyle{ncyan}=[circle, draw=cyan!70, thin, fill=white, scale=0.8, font=\fontsize{11}{0}\selectfont]
\tikzstyle{ngreen}=[circle,  draw=green!70, thin, fill=white, scale=0.8, font=\fontsize{11}{0}\selectfont]
\tikzstyle{nred}=[circle, draw=red!70, thin, fill=white, scale=0.8, font=\fontsize{11}{0}\selectfont]
\tikzstyle{ngray}=[circle, draw=gray!70, thin, fill=white, scale=0.55, font=\fontsize{14}{0}\selectfont]
\tikzstyle{nyellow}=[circle, draw=yellow!70, thin, fill=white, scale=0.55, font=\fontsize{14}{0}\selectfont]
\tikzstyle{norange}=[circle,  draw=orange!70, thin, fill=white, scale=0.55, font=\fontsize{10}{0}\selectfont]
\tikzstyle{npurple}=[circle,draw=purple!70, thin, fill=white, scale=0.55, font=\fontsize{10}{0}\selectfont]
\tikzstyle{nblue}=[circle, draw=blue!70, thin, fill=white, scale=0.55, font=\fontsize{10}{0}\selectfont]
\tikzstyle{nteal}=[circle,draw=teal!70, thin, fill=white, scale=0.55, font=\fontsize{10}{0}\selectfont]
\tikzstyle{nviolet}=[circle, draw=violet!70, thin, fill=white, scale=0.55, font=\fontsize{10}{0}\selectfont]
\tikzstyle{qgre}=[rectangle, draw, thin,fill=green!20, scale=0.8]
\tikzstyle{rpath}=[ultra thick, red, opacity=0.4]
\tikzstyle{legend_isps}=[rectangle, rounded corners, thin,fill=gray!20, text=blue, draw]
\usetikzlibrary{positioning}
\definecolor{lightblue}{RGB}{173, 216, 230}

\definecolor{codegreen}{rgb}{0,0.6,0}
\definecolor{codegray}{rgb}{0.5,0.5,0.5}
\definecolor{codepurple}{rgb}{0.58,0,0.82}
\definecolor{backcolour}{rgb}{0.95,0.95,0.92}

\newtheorem{prop}{Proposition}
\numberwithin{prop}{section} 
\counterwithin{figure}{section}

 \usepackage{mleftright}
 
\newcommand{\hypothesis}{h} 
\newcommand{\algomap}{\mathcal{A}}
 
\newcommand{\estlocalparamsiter}[2]{\widehat{\weights}^{(#1,#2)}}

\newcommand{\estlocalhypositer}[2]{\widehat{\hypothesis}^{(#1)}_{#2}}
\newcommand{\estparams}[1]{\widehat{\weights}_{#1}}
\newcommand{\localmodel}[1]{\hypospace^{(#1)}} 
\newcommand{\localhypothesis}[1]{\hypothesis^{(#1)}} 
\newcommand{\learntlocalhypothesis}[1]{\widehat{\hypothesis}^{(#1)}} 

\newcommand{\discrepancy}[2]{d^{(#1,#2)}}  
\newcommand{\sphere}[1]{\mathbb{S}^{(#1)}}  
\newcommand{\firstnatural}[1]{[#1]}    

\newcommand{\metric}[2]{d(#1,#2)}  
\newcommand{\innerprod}[2]{\langle #1, #2 \rangle}  
\newcommand{\hilbertspace}{\mathcal{H}}  
\newcommand{\reals}{\mathbb{R}}

\newcommand{\defeq}{:=}

\newcommand{\vx}[0]{{\bf x}}
\newcommand{\vv}[0]{{\bf v}}
\newcommand{\vu}[0]{{\bf u}}
\newcommand{\vs}[0]{{\bf s}}
\newcommand{\vm}[0]{{\bf m}}

\newcommand{\vq}[0]{{\bf q}}
\newcommand{\mX}[0]{{\bf X}}
\newcommand{\mC}[0]{{\bf C}}
\newcommand{\mA}[0]{{\bf A}}
\newcommand{\mD}[0]{{\bf D}}
\newcommand{\mR}[0]{{\bf R}}
\newcommand{\mB}[0]{{\bf B}}
\newcommand{\mL}[0]{{\bf L}}

\newcommand{\mW}[0]{{\bf W}}

\newcommand{\vw}[0]{{\bf w}}

\newcommand{\mP}{\mathbf{P}}
\newcommand{\mI}{\mathbf{I}}
\newcommand{\mF}{\mathbf{F}}
\newcommand{\mG}{\mathbf{G}}
\newcommand{\mT}{\mathbf{T}}

\newcommand{\mQ}{\mathbf{Q}}
\newcommand{\mU}{\mathbf{U}}
\newcommand{\mV}{\mathbf{V}}

\newcommand{\vy}[0]{{\bf y}}
\newcommand{\va}[0]{{\bf a}}
\newcommand{\ve}[0]{{\bf e}}
\newcommand{\vf}[0]{{\bf f}}
\newcommand{\vn}[0]{{\bf n}}
\newcommand{\vb}[0]{{\bf b}}

\newcommand{\vz}[0]{{\bf z}}
\newcommand{\N}[0]{{\mathcal{N}}}
\newcommand{\vc}[0]{{\bf c}}
\newcommand{\vg}[0]{{\bf g}}
\newcommand{\vk}[0]{{\bf k}}

\newcommand{\linspan}[1]{{\rm span}\left(#1 \right)}  
\newcommand{\nullspace}[1]{{\rm null}\left(#1 \right)}  

\newcommand{\indicatorfunc}[1]{\mathbb{I}_{#1}}

\newcommand{\simplex}[1]{\Delta^{#1}}
\newcommand{\meanvecgeneric}{{\bm \mu}} 
\newcommand{\samplemeanvecgeneric}{\widehat{\bm \mu}} 
\newcommand{\covmtxgeneric}{\mC} 
\newcommand{\indexset}{\mathcal{I}}
\newcommand{\samplecovmtxgeneric}{\widehat{\mC}} 
 
\newcommand{\probmeasure}{\mathbb{P}} 
\newcommand{\prob}[1]{\mathbb{P}\left(#1\right)} 
\newcommand{\cdf}[2]{F^{(#1)}\left(#2\right)} 
\newcommand{\probdist}{\mathbb{P}}
\newcommand{\entropy}[1]{H\left(#1\right)}   
\newcommand{\pdf}[2]{p^{(#1)}\left(#2\right)}   
\newcommand{\pmf}[2]{p^{(#1)}\left(#2\right)}   
\newcommand{\mgf}[2]{M_{#1}\left(#2\right)}   
\newcommand{\mgfder}[3]{M_{#1}^{(#2)}\left(#3\right)}   
\newcommand{\meanvec}[1]{{\bm \mu}^{(#1)}}  
\newcommand{\covmtx}[1]{\mathbf{C}^{(#1)}}  
\newcommand{\expect}{\mathbb{E} }    
\newcommand{\cov}[2]{{\rm cov}\left(#1,#2\right)}    
\newcommand{\covfunc}[3]{{\rm cov}^{(#1)}\left(#2,#3\right)}  
\newcommand{\sigmaalgebra}{\Sigma}    
\newcommand{\samplespace}{\Omega}
\newcommand{\eventspace}{\mathcal{F}}
\newcommand{\outcome}{\omega}

\newcommand{\numlearners}{M}
\newcommand{\vcdim}[1]{\operatorname{VCdim}\left(#1\right)}
\newcommand{\paramspace}{\mathcal{W}}
\newcommand{\biasterm}{B}

\newcommand{\neighbourhood}[1]{\mathcal{N}^{(#1)}}

\newcommand{\med}{\mathrm{med}}

\newcommand{\nrbootstraps}{B}
\newcommand{\sampleweight}[1]{q^{(#1)}}
\newcommand{\nrcategories}{K} 

\newcommand{\maxnredges}{E_{\rm max}} 
\newcommand{\privattr}{s}
\newcommand{\sensattr}{s} 
\newcommand{\kernel}{k} 
\newcommand{\kernelmap}[2]{k\big(#1,#2\big)} 

\newcommand{\norm}[1]{\Vert  {#1} \Vert}
\newcommand{\normgeneric}[2]{\mleft\lVert #1 \mright\rVert_{#2}}

\newcommand{\bmx}[0]{\begin{bmatrix}}
\newcommand{\emx}[0]{\end{bmatrix}}
\newcommand{\T}[0]{\text{T}}
\DeclareMathOperator*{\rank}{rank}

\newcommand{\eigvecCov}{\mathbf{u}} 

\newcommand{\featuredim}{d}
\newcommand{\nrfeatures}{\featuredim}   

\newcommand{\featurelen}{\featuredim}     

\newcommand{\samplesize}{m}
\newcommand{\sampleidx}{r} 
 
\newcommand{\datapoint}{\vz} 
 
\newcommand{\clusteridx}{c} 
\newcommand{\nrcluster}{k}   
  
\newcommand{\featureidx}{j}  
 
\newcommand{\clustercentroid}[1]{{\vw}^{(#1)}} 
\newcommand{\clustercentroiditer}[2]{{\vw}^{(#1,#2)}} 

\newcommand{\truelabel}{y}

\newcommand{\labelvec}{\vy}
\newcommand{\featurevec}{\vx}

\newcommand{\feature}{x}
\newcommand{\predictedlabel}{\hat{\truelabel}}
\newcommand{\dataset}{\mathcal{D}}
\newcommand{\trainset}{\dataset^{(\rm t)}}
\newcommand{\valset}{\dataset^{(\rm v)}}

\newcommand{\effdim}[1]{d_{\rm eff} \left( #1 \right)}
\newcommand{\inspace}{\mathcal{X}}

\newcommand{\learnthypothesis}{\hat{\hypothesis}}

\newcommand{\hypospace}{\mathcal{H}}

\newcommand{\emperror}{\widehat{L}}
\newcommand{\emprisk}[2]{\widehat{L}\big(#1|#2\big)}
\newcommand{\risk}[1]{\bar{L} \big( #1 \big) } 
\newcommand{\featurespace}{\mathcal{X}}
\newcommand{\labelspace}{\mathcal{Y}}

\newcommand{\user}{u}
\newcommand{\actfun}{\sigma}

\newcommand{\eigval}[1]{\lambda_{#1}}
\newcommand{\eigvalgen}{\lambda}
\newcommand{\regparam}{\alpha}
\newcommand{\maxeigvallocalQ}{\lambda_{\rm max}} 
\newcommand{\avgmineigvallocalQ}{\bar{\lambda}_{\rm min}} 
\newcommand{\lrate}{\eta}
\newcommand{\upperboundeigval}{U}
\newcommand{\lowerboundeigval}{L}

\newcommand{\tr}[1]{{\rm tr} \left(#1\right)}

\DeclareMathOperator*{\argmin}{arg\,min}
\newcommand{\itercntr}{k}
\newcommand{\state}{s} 
\newcommand{\statespace}{\mathcal{S}} 
\newcommand{\timeidx}{t}

\newcommand{\reward}{r}

\newcommand{\arm}{a}    
\newcommand{\actionset}{\actionspace}
\newcommand{\actionspace}{\mathcal{A}}

\newcommand{\statevaluefunc}[1]{v_{#1}} 
\newcommand{\valuefunc}{v} 

\newcommand{\policy}{\pi}

\newcommand{\iteridx}{t}
\newcommand{\iteridxinner}{r}
\newcommand{\discountfac}{\gamma}

\newcommand{\valerror}{E_{v}}
\newcommand{\trainerror}{E_{t}}
\newcommand{\foldidx}{b}
\newcommand{\testset}{\dataset^{(\rm test)} }

\newcommand{\benchmarkerror}{E^{(\rm ref)}}
\newcommand{\loss}{L}
\newcommand{\lossfun}{L}
\newcommand{\lossfunczo}[2]{L^{(0/1)}\left(#1,#2 \right)}
\newcommand{\lossfunc}[2]{L\left(#1,#2 \right)}
 
\newcommand{\cluster}{\mathcal{C}}

\newcommand{\featuremtx}{\mX}
\newcommand{\weight}{w}
\newcommand{\weights}{\vw}
\newcommand{\regularizer}[1]{\mathcal{R}\big\{ #1 \big\}}
\newcommand{\decreg}[1]{\mathcal{R}_{#1}}

\newcommand{\featuremapvec}{{\bf \Phi}}
\newcommand{\featuremap}{\phi}
\newcommand{\maxdelay}{B}
\newcommand{\batchsize}{B}
\newcommand{\batch}{\mathcal{B}}

\newcommand{\nriter}{R}
\newcommand{\opttol}{\varepsilon^{(\rm tol)}}

\newcommand{\bd}[1]{\partial #1}

\newcommand{\poisoningrate}{\eta}
\newcommand{\thresholdfct}{\tau}
\newcommand{\complexconjugate}[1]{\overline{#1}}

\newcommand{\closedball}[2]{\overline{\mathcal{B}}_{#2}(#1)}

\newcommand{\proximityop}[2]{{\rm\bf prox}_{#1}(#2)}   

\newcommand{\fixedpointop}{\mathcal{F}} 
\newcommand{\contractfac}{\kappa}
\newcommand{\identityop}{\mathrm{Id}}

\newcommand{\localregularizer}[2]{\mathcal{R}^{(#1)}\big\{ #2 \big\}}

\newcommand{\locallossfunc}[2]{L_{#1}\left(#2 \right)}
\newcommand{\perturbedlocallossfunc}[2]{\tilde{L}_{#1}\left(#2 \right)}
\newcommand{\conjlocallossfunc}[2]{L^{*}_{#1}\left(#2 \right)}

\newcommand{\localdataset}[1]{\mathcal{D}^{(#1)}}
\newcommand{\localtestset}[1]{\mathcal{D}_{t}^{(#1)}}
\newcommand{\edges}{\mathcal{E}}

\newcommand{\edgeweight}{A}
\newcommand{\edgeweights}{\mA}

\newcommand{\edgeidx}{e}

\newcommand{\graph}{\mathcal{G}}
\newcommand{\nodes}{\mathcal{V}}
\newcommand{\LapMat}[1]{\mL^{(#1)}}
\newcommand{\LapMatEntry}[3]{L^{(#1)}_{#2,#3}}
\newcommand{\indsubgraph}[2]{#1^{(#2)}}
\newcommand{\nodedegree}[1]{d^{(#1)}}

\newcommand{\maxnodedegree}{d_{\rm max}}
\newcommand{\avgnodedegree}{\overline{d}}
\newcommand{\nodeidx}{i}
\newcommand{\nrnodes}{n}

\newcommand{\edge}[2]{\{#1,#2\}}

\newcommand{\mvnormal}[2]{\mathcal{N}\left(#1,#2\right)}
\newcommand{\gtvpenalty}{\phi}

\newcommand{\localsamplesize}[1]{m_{#1}}

\newcommand{\dimlocalmodel}{d}
\newcommand{\pair}[2]{\left( #1,#2 \right)}
\newcommand{\localparams}[1]{\mathbf{w}^{(#1)}}
\newcommand{\localparamsiter}[2]{\mathbf{w}^{(#1,#2)}}
\newcommand{\localflowvec}[1]{\mathbf{u}^{(#1)}}

\newcommand{\estlocalparams}[1]{\widehat{\mathbf{w}}^{(#1)}}
\newcommand{\localobj}[2]{f^{(#1)}\left( #2 \right)}

\newcommand{\netparams}{\mathbf{w}}
\newcommand{\nodeweight}[1]{\rho_{#1}}
\newcommand{\mutualinformation}[2]{I \left( #1;#2\right)}

\newcommand{\linmodel}[1]{\hypospace^{(#1)}}
\newcommand{\bregmandiv}[3]{D_{#3}\big( #1,#2 \big)}

\newcommand{\kld}[2]{D^{(\rm KL)}\big( #1,#2 \big)}

\newcommand{\projection}[2]{P_{#1}\big( #2\big) }
\newcommand{\perturbation}[1]{{\bm \varepsilon}^{(#1)}}
\newcommand{\gdcontract}[2]{\kappa^{(#1)} \left(#2\right)}

\newcommand{\upperboundnormestpar}{R}

\newcommand{\asyncactiveset}[1]{\mathcal{A}^{(#1)}} 
\newcommand{\updatetimes}[1]{T^{(#1)}} 
\newcommand{\trimmedset}{\mathcal{T}}

\newglossary[rlg]{reinflearning}{rls}{rlo}{Reinforcement Learning}
\newglossary[rgg]{regulation}{rgs}{rgo}{Regulation}

\newglossaryentry{imagesegmentation}
{name={image segmentation}, 
	description={Image segmentation\index{image segmentation} refers to the task of 
    		\gls{clustering} the pixels of an image into a few segments \cite{JungLocalGraphClustering}, \cite{Wren1997}. 
		Each segment is a subset (or \gls{cluster}) of pixels that are similar to each other in 
		terms of color, texture, or other visual properties. 
		\\ 
 		See also: \gls{clustering}.}, 
 	first={image segmentation}, 
 	text={image segmentation}
}

\newglossaryentry{stratification}
{name={stratification},
	description={The process of splitting a \gls{dataset} into subsets, so called \glspl{stratum},
 		according to some key attribute is called stratification\index{stratification} 
		\cite{Everitt2010}, \cite{OxfordStatisticsDictionary}, \cite{oecd2008glossary}. 
 		The goal is to ensure that an \gls{ml} method performs well for each \gls{stratum} 
		defined by these attributes. For example, in a medical \gls{dataset},
  		we may want to stratify a patient \gls{dataset} by age groups to ensure that an
  		\gls{ml} \gls{model} performs well across all age groups. \\
  		When splitting a \gls{dataset} into a \gls{trainset} and a \gls{valset}, 
  		stratification ensures that both sets have similar distributions of the key attribute. 
  		Without stratification, using a small \gls{valset} may underrepresent or 
  		even completely miss \glspl{datapoint} with a rare attribute, leading to misleading 
  		performance estimates. See Fig.\ \ref{fig_stratification_dict} for a visual illustration.
 		\begin{figure}[H]
 			\centering
			\begin{tikzpicture}[font=\small,x=1.7cm]
			\def\cellw{6.0}     
			\def\cellh{0.5}     
			\def\gap{1}       
			\def\tfrac{0.8}     
			\def\pA{0.9}        
			\node[anchor=south] at (0.5*\cellw, \cellh+0.2) {\gls{dataset}};
			\node[anchor=south] at ({0.5*\cellw}, {-(\cellh+\gap)+\cellh+0.2}) {\gls{trainset}/\gls{valset} split without stratification};
			\node[anchor=south] at (0.5*\cellw, {-2*(\cellh+\gap)+\cellh+0.2}) {with stratification};
			\pgfmathsetmacro{\yDataset}{0}
			\pgfmathsetmacro{\yNoStrat}{-(\cellh+\gap)}
			\pgfmathsetmacro{\yStrat}{-2*(\cellh+\gap)}
			\pgfmathsetmacro{\xAB}{\pA*\cellw}
			\pgfmathsetmacro{\wA}{\pA*\cellw}
			\pgfmathsetmacro{\wB}{(1-\pA)*\cellw}
			\draw (0,\yDataset) rectangle ++(\cellw,\cellh);
			\draw[dashed,  line width=0.5mm] (\xAB,\yDataset) -- ++(0,\cellh);
			\node at (0.5*\wA, \yDataset+0.5*\cellh) {$\sensattr\!=\!1$};
			\node at ({\xAB+0.5*\wB}, {\yDataset+0.5*\cellh}) {$\sensattr\!=\!2$};
			\draw (0,\yNoStrat) rectangle ++(\cellw,\cellh);
			\pgfmathsetmacro{\xValStart}{\tfrac*\cellw}
			\pgfmathsetmacro{\valw}{(1-\tfrac)*\cellw}
			\fill[gray!35] (\xValStart,\yNoStrat) rectangle ++(\valw,\cellh);
			\draw[dashed,  line width=0.5mm] (\xAB,\yNoStrat) -- ++(0,\cellh);
			\draw (0,\yStrat) rectangle ++(\cellw,\cellh);
			\pgfmathsetmacro{\xAB}{\pA*\cellw}
			\draw[dashed, line width=1mm] (\xAB,\yStrat) -- ++(0,\cellh);
			\pgfmathsetmacro{\xAVal}{\tfrac*\wA}              
			\pgfmathsetmacro{\wAVal}{(1-\tfrac)*\wA}          
			\fill[gray!35] (\xAVal,\yStrat) rectangle ++(\wAVal,\cellh);
			\pgfmathsetmacro{\xBVal}{\xAB + \tfrac*\wB}       
			\pgfmathsetmacro{\wBVal}{(1-\tfrac)*\wB}          
			\fill[gray!35] (\xBVal,\yStrat) rectangle ++(\wBVal,\cellh);
			\end{tikzpicture}
 		\caption{Stratification ensures that both the \gls{trainset} and the \gls{valset} 
			(shaded grey) have similar distributions of a binary key attribute $\sensattr$. 
			In other words, with stratification, both \glspl{stratum}  
			(i.e., $\sensattr\!=\!1$ and $\sensattr\!=\!2$ based on the key attribute) 
			allocate the same \gls{validation} proportion—20\% of their own width. \label{fig_stratification_dict}}
 		\end{figure}
  		See also: \gls{stratum}, \gls{validation}, \gls{kfoldcv}.},
  	text={stratification},
  	first={stratification},
  	plural={stratifications}
}

\newglossaryentry{stratum}
{name={stratum},
	description={A\index{stratum} stratum is a subset of \glspl{datapoint} that all 
		share a common property (which could be a \gls{feature} or a \gls{label}). 
		For example, in a weather \gls{dataset}, all measurements from the same 
		\gls{fmi} weather station form one stratum.
		\begin{center}
			Example (CSV snippet):\\
  			{\ttfamily
  			time, station, value, unit\\
  			2023-06-01 12:00, Helsinki, 18.2, degree Celsius\\
  			2023-06-01 13:00, Helsinki, 18.5, degree Celsius\\
  			2023-06-01 14:00, Helsinki, 19.0, degree Celsius\\
  			2023-06-01 12:00, Oulu, 12.1, degree Celsius\\
  			2023-06-01 13:00, Oulu, 12.4, degree Celsius\\
  			2023-06-01 14:00, Oulu, 12.7, degree Celsius\\
  			2023-06-01 12:00, Tampere, 15.3, degree Celsius\\
  			2023-06-01 13:00, Tampere, 15.6, degree Celsius\\
  			2023-06-01 14:00, Tampere, 16.0, degree Celsius
  			}
		\end{center} 
		Here, the rows for each station (i.e., \texttt{Helsinki}, \texttt{Oulu}, \texttt{Tampere}) 
		represent different strata.
		\\ 
		See also: \gls{datapoint}, \gls{dataset}, \gls{stratification}.},
  	first={stratum},
  	firstplural={strata},
 	plural={strata}, 
 	text={stratum}
}

\newglossaryentry{attention}
{name={attention}, 
	description={Some \gls{ml} applications\index{attention} involve \glspl{datapoint} 
		composed of smaller units, referred to as \glspl{token}. For example, a sentence 
		consists of words, an image of pixel patches, and a network of nodes. In general, 
		the \glspl{token} that constitute a single \gls{datapoint} are not independent 
		of one another. Instead, each \gls{token} of a \gls{datapoint} depends on (or pays attention to) 
		specific other \glspl{token}. \Glspl{probmodel} provide a principled framework for 
		representing and analyzing such dependencies \cite{Blei2003}. Attention mechanisms use 
		a more direct approach without explicit reference to a \gls{probmodel}. The idea is to 
		represent the relationship between two \glspl{token} $\nodeidx$ and $\nodeidx'$ 
		using a parameterized \gls{function} $f^{(\weights)}(\nodeidx,\nodeidx')$, 
		where the \glspl{parameter} $\weights$ are learned via a variant of \gls{erm}. 
		Practical attention mechanisms differ in their precise choice of attention \gls{model} $f^{(\weights)}(\nodeidx,\nodeidx')$ 
		as well as in the precise \gls{erm} variant used to learn the \glspl{parameter} $\weights$. 
		One widely used family of attention mechanisms 
		defines the \glspl{parameter} $\weights$ in terms of two \glspl{vector} associated with each \gls{token} $\nodeidx$, i.e.,
		a query \gls{vector} $\vq^{(\nodeidx)}$ and a key \gls{vector} $\vk^{(\nodeidx')}$. For a given \gls{token} $\nodeidx$ 
		with query $\vq^{(\nodeidx)}$ and another \gls{token} $\nodeidx'$ with key $\vk^{(\nodeidx')}$, the quantity 
		$\big( \vq^{(\nodeidx)} \big)^{\top} \vk^{(\nodeidx')}$ indicates the extent to which \gls{token} $\nodeidx$ attends 
		to (or depends on) \gls{token} $\nodeidx'$ (see Fig.\ \ref{fig_attention_dict}).
		\begin{figure}[H]
			\centering
			\begin{tikzpicture}[>=stealth, node distance=0.2cm and 0.2cm,
			every node/.style={inner sep=2pt, font=\large}, baseline]
			\node (w1) [draw, fill=gray!10, rounded corners] {All};
			\node (w2) [draw, fill=gray!10, right=of w1, rounded corners] {human};
			\node (w3) [draw, fill=gray!10, right=of w2, rounded corners] {beings};
			\node (w4) [draw, fill=gray!10, right=of w3, rounded corners] {are};
			\node (w5) [draw, fill=gray!10, right=of w4, rounded corners] {born};
			\node (w6) [draw, fill=gray!10, right=of w5, rounded corners] {free};
			\node (w7) [draw, fill=gray!10, right=of w6, rounded corners] {and};
			\node (w8) [draw, fill=blue!20, right=of w7, rounded corners] {equal};
   	   		\node[font=\footnotesize, below=0.15cm of w8, align=center] (labeli) {$\nodeidx$ \\ $\vq^{(\nodeidx)},\ \vk^{(\nodeidx)}$};
			\node[font=\footnotesize, below=0.15cm of w1, align=center] (labelii) {$\nodeidx'$ \\ $\vq^{(\nodeidx')},\ \vk^{(\nodeidx')}$};
	  		\node (eqTop) [above=1.8cm of w8] {};
	 		\draw[->, thick, opacity=0.3] (w8.north) .. controls +(up:1.0cm) and +(up:1.0cm) .. (w6.north); 
	 		\draw[->, thick, opacity=0.3] (w8.north) .. controls +(up:1.2cm) and +(up:1.0cm) .. (w5.north); 
			\draw[->, thick, opacity=1]  (w8.north)  .. controls +(up:1.8cm) and +(up:1.0cm) ..  node[midway, text=black, above] {$f^{(\weights)}(\nodeidx,\nodeidx')$}  (w1.north); 
			\end{tikzpicture}
		\caption{Attention mechanisms learn a parameterized \gls{function} $f^{(\weights)}(\nodeidx,\nodeidx')$ to measure 
			how much \gls{token} $\nodeidx$ attends to \gls{token} $\nodeidx'$. 
			One widely used construction of $f^{(\weights)}(\nodeidx,\nodeidx')$ 
			uses query and key \glspl{vector}, denoted by $\vq^{(\nodeidx)}$ and $\vk^{(\nodeidx)}$, assigned to each 
			\gls{token} $\nodeidx$ \cite{vaswani2017attention}. \label{fig_attention_dict}}
		\end{figure}
		See also: \gls{token}, \gls{function}, \gls{nlp}.},
	first={attention},
	text={attention}
}

\newglossaryentry{transformer}
{name={transformer},
	description={In the context of \gls{ml}, the term transformer\index{transformer} 
 		refers to an \gls{ann} that uses some form of \gls{attention} mechanism 
 		to capture dependencies among \glspl{token} \cite{vaswani2017attention}. 
 		The \gls{attention} mechanism is what sets transformers apart from 
		previous \glspl{model} used for sequential \gls{data} such as \glspl{rnn}. 
		A transformer \gls{ann} often combines several \gls{attention} \glspl{layer} 
		via more traditional \gls{layer} architectures. 
		\\ 
		See also: \gls{attention}, \gls{nlp}. }, 
 	first={transformer}, 
 	text={transformer},
 	firstplural={transformers}, 
 	plural={transformers}
}

\newglossaryentry{rnn}
{name={recurrent neural network (RNN)},
	description={An\index{recurrent neural network (RNN)} RNN 
		is a specific type of \gls{ann} that is designed for processing 
		\gls{data} that consist of a \gls{sequence} of \glspl{token}. 
		An RNN maintains an internal hidden \gls{state} that is updated recurrently 
		as new \glspl{token} are processed. This recurrent dependence allows 
		information to propagate across time steps, making RNNs suitable for 
		tasks such as speech recognition, language modeling, or time series \gls{prediction}. 
    		However, their inherently sequential computation limits parallelization and 
		is challenging for \glspl{gdmethod}. Variants like the long short-term memory (LSTM) 
		and gated recurrent unit (GRU) mitigate these problems.
		\\
		See also: \gls{ann}, \gls{token}. }, 
 	first={recurrent neural network (RNN)},
 	text={RNN},
 	firstplural={recurrent neural networks (RNNs)}, 
 	plural={RNNs}
}

\newglossaryentry{foundationmodel}
{name={foundation model},
	description={A\index{foundation model} foundation \gls{model} is a general-purpose
		\gls{featuremap} $\featuremap \colon \inspace \to \mathbb{R}^{d}$ learned via
		\gls{selfsupervisedlearning} on diverse \gls{data} \cite{Bommasani2021}.
		The learned \gls{featuremap} can be reused across many downstream tasks
		through \gls{transferlearning}, e.g., by appending a small task-specific
		\gls{hypothesis} on top of $\featuremap$ and fine-tuning on a
		task-specific \gls{dataset}.
		Examples include \glspl{llm} for text and analogous \glspl{model}
		for images, audio, and code.
		The EU AI Act refers to foundation \glspl{model} as general-purpose
		AI \glspl{model} \cite{EUAIAct2024} (see \gls{provider}).
		\\
		See also: \gls{featuremap}, \gls{selfsupervisedlearning}, \gls{transferlearning}, 
		\gls{llm}, \gls{provider}.},
	first={foundation model},
	plural={foundation models},
	firstplural={foundation models},
	text={foundation model}
}

\newglossaryentry{llm}
{name={large language model (LLM)},
	description={An\index{large language model (LLM)} LLM is an umbrella term for \gls{ml} 
		methods that use high-dimensional \gls{ml} \glspl{model} (with billions 
		of \glspl{modelparam}) trained on large collections of text \gls{data}. 
 		LLMs are used to analyze or generate \glspl{sequence} of \glspl{token} that 
               	constitute text \gls{data}. Many current LLMs use some variant of 
		a \gls{transformer} that is trained via \gls{selfsupervisedlearning}, i.e.,
		the \gls{training} is based on the task of predicting a few words that 
		are intentionally removed from a large text corpus. Thus, we can 
		construct \glspl{labeled datapoint} simply by selecting some words 
		from a given text as \glspl{label} and the remaining words as \glspl{feature} 
		of \glspl{datapoint}. This construction requires 
		very little human supervision and allows for generating sufficiently 
		large \glspl{trainset} for LLMs.
		\\ 
		See also: \gls{token}, \gls{transformer}, \gls{nlp}.}, 
  	first={large language model (LLM)}, 
  	firstplural={LLMs}, 
  	text={LLM}, 
  	plural={LLMs} 
}

\newglossaryentry{selfsupervisedlearning}
{name={self-supervised learning},
 	description={Self-supervised learning\index{self-supervised learning} uses some of 
  		the \glspl{feature} of a \gls{datapoint} as its \gls{label}. For 
		example, if a \gls{datapoint} consists of a sentence within a text document, 
		we can use the last word of the sentence as the \gls{label} that is 
		to be predicted from all the previous words, which form the \glspl{feature} 
		of the \gls{datapoint}. A main application of self-supervised learning 
		is in \gls{nlp} for the \gls{training} of \glspl{llm} from large collections 
		of text \gls{data}. 
		\\ 
      		See also: \gls{feature}, \gls{label}, \gls{llm}.}, 
   	first={self-supervised learning}, 
   	text={self-supervised learning}
}

\newglossaryentry{attack}
{name={attack},  
	description={An attack\index{attack} on an \gls{mlsystem} refers to an intentional \gls{action}—either 
		active or passive—that compromises the system's integrity, availability, or confidentiality. 
		Active attacks involve perturbing components such as \glspl{dataset} (via \gls{datapoisoning}) 
		or communication links between \glspl{device} within an \gls{ml} application. Passive attacks, 
		such as \glspl{privattack}, aim to infer \glspl{sensattr} without modifying the system. 
		Depending on their goal, we distinguish among \glspl{dosattack}, \gls{backdoor} attacks, and \glspl{privattack}.
		\\
		See also: \gls{datapoisoning}, \gls{privattack}, \gls{sensattr}, \gls{dosattack}, \gls{backdoor}.},
	plural={attacks}, 
	first={attack},
	firstplural={attacks},
	text={attack}
}

\newglossaryentry{privattack}
{name={privacy attack},
	description={A privacy \gls{attack}\index{privacy attack} on an \gls{mlsystem} aims to infer 
		\glspl{sensattr} of individuals by exploiting partial access to a trained \gls{ml} \gls{model}. 
		One form of a privacy \gls{attack} is \gls{modelinversion}.
		\\
		See also: \gls{attack}, \gls{sensattr}, \gls{modelinversion}, \gls{trustAI}, \gls{gdpr}.},
	plural={privacy attacks}, 
	first={privacy attack},
	firstplural={privacy attacks}, 
	text={privacy attack}
}

\newglossaryentry{discrepancy}
{name={discrepancy},
	description={Consider\index{discrepancy} an \gls{fl} application with \gls{netdata} 
		represented by an \gls{flnetwork}. \gls{fl} methods use a discrepancy \gls{measure} 
		to compare \gls{hypothesis} \glspl{map} from \glspl{localmodel} at nodes $\nodeidx,\nodeidx'$, 
		\gls{connected} by an edge in the \gls{flnetwork}.
					\\ 
		See also: \gls{fl}, \gls{flnetwork}, \gls{localmodel}.},
	first={discrepancy},
	firstplural={discrepancies}, 
  	plural={discrepancies}, 
	text={discrepancy}
}

\newglossaryentry{FedRelax}
{name={FedRelax},
	description={FedRelax refers to a \gls{gtvmin}-based \gls{fl} method\index{FedRelax} 
		for training local \glspl{model} at the \glspl{device} of an \gls{flnetwork} 
		\cite{JungFLBook}. It is a \gls{distributedalgorithm} that implements 
		a nonlinear variant of the \gls{jacobimethod} to solve \gls{gtvmin}. 
		\\ 
		See also: \gls{fl}, \gls{distributedalgorithm}.},
	first={FedRelax},
	text={FedRelax}
}

\newglossaryentry{fedavg}
{name={federated averaging (FedAvg)},
	description={FedAvg\index{federated averaging (FedAvg)} refers to a family of iterative \gls{fl} \glspl{algorithm}. 
		It uses a server-client setting and alternates between clientwise \glspl{localmodel} 
		retraining, followed by the aggregation of updated \glspl{modelparam} at the server 
		\cite{pmlr-v54-mcmahan17a}, \cite{JungFLBook}. The local update at client $\nodeidx=1, \,\ldots, \,\nrnodes$ 
		at time $\iteridx$ starts from the current \glspl{modelparam} $\weights^{(\iteridx)}$ provided 
		by the server and typically amounts to executing few \glspl{iteration} of \gls{stochGD}. 
		After completing the local updates, they are aggregated 
		by the server (e.g., by averaging them). Fig.\ \ref{fig_single_iteration_fedavg_dict} illustrates 
		the execution of a single \gls{iteration} of FedAvg. 
		\begin{figure}[H]
			\begin{center}
			\begin{tikzpicture}[>=Stealth, node distance=1cm and 1.5cm, every node/.style={font=\small}]
			\tikzstyle{server} = [circle, fill=black, minimum size=6pt, inner sep=0pt]
			\tikzstyle{client} = [circle, draw=black, minimum size=6pt, inner sep=0pt]
			\node (label1) at (0,3.5) {broadcast};
			\node[right=2.5cm of label1] (label2) {local update};
			\node[right=2.5cm of label2] (label3) {aggregate};
			\node[server] (s1) at (label1 |- 0,2.5) {};
			\node[client] (c1l) at ($(s1) + (-1cm,-1cm)$) {};
			\node[client] (c1r) at ($(s1) + (1cm,-1cm)$) {};
			\node[] (dots1) at ($(s1) + (0cm,-1cm)$) {\ldots};
			\draw[->] (s1) -- (c1l) node[midway,left] {$\weights^{(\iteridx)}$};
			\draw[->] (s1) -- (c1r) node[midway,right] {$\weights^{(\iteridx)}$};
			\draw[->] (s1) -- (dots1);
			\node[server] (s2) at (label2 |- 0,2.5) {};
			\node[client] (c2l) at ($(s2) + (-1cm,-1cm)$) {};
			\node[client] (c2r) at ($(s2) + (1cm,-1cm)$) {};
			\node[] (dots2) at ($(s2) + (0cm,-1cm)$) {\ldots};
			\node[below=0.2cm of c2l] {$\localparamsiter{\iteridx}{1}$};
			\node[below=0.2cm of c2r] {$\localparamsiter{\iteridx}{\nrnodes}$};
			\node[server] (s3) at (label3 |- 0,2.5) {};
			\node[above=0.01cm of s3, yshift=-4pt] {$\weights^{(\iteridx+1)}$};
			\node[client] (c3l) at ($(s3) + (-1cm,-1cm)$) {};
			\node[client] (c3r) at ($(s3) + (1cm,-1cm)$) {};
			\node[] (dots3) at ($(s3) + (0cm,-1cm)$) {\ldots};
			\draw[->] (c3l) -- (s3) node[midway,left] {$\localparamsiter{\iteridx}{1}$};
			\draw[->] (c3r) -- (s3)  node[midway,right] {$\localparamsiter{\iteridx}{\nrnodes}$};
			\draw[->] (dots3) -- (s3);
			\end{tikzpicture}
			\end{center}
		\caption{Illustration of a single \gls{iteration} of FedAvg, which consists of broadcasting \glspl{modelparam} by the 
			server, performing local updates at clients, and aggregating the updates by the server. 
			\label{fig_single_iteration_fedavg_dict}} 
		\end{figure} 
		See also: \gls{fl}, \gls{algorithm}, \gls{localmodel}, \gls{stochGD}.},
	first={federated averaging (FedAvg)},
	text={FedAvg}
} 

\newglossaryentry{FedGD}
{name={federated gradient descent (FedGD)},
	description={FedGD refers to  an\index{federated gradient descent (FedGD)} \gls{fl} \gls{distributedalgorithm} that 
		can be implemented as \gls{messagepassing} across an \gls{flnetwork} \cite{JungFLBook}. 
		\\ 
		See also: \gls{fl}, \gls{distributedalgorithm}, \gls{flnetwork}, \gls{gradstep}, \gls{gdmethod}.},
	first={federated gradient descent (FedGD)},
	text={FedGD}
} 

\newglossaryentry{FedSGD}
{name={federated stochastic gradient descent (FedSGD)},
	description={FedSGD refers to an\index{federated stochastic gradient descent (FedSGD)} \gls{fl} \gls{distributedalgorithm} that 
		can be implemented as \gls{messagepassing} across an \gls{flnetwork} \cite{JungFLBook}. 
		\\ 
		See also: \gls{fl}, \gls{distributedalgorithm}, \gls{flnetwork}, \gls{gradstep}, \gls{gdmethod}, \gls{stochGD}.},
	first={federated stochastic gradient descent (FedSGD)},
	text={FedSGD}
} 

\newglossaryentry{hfl}
{name={horizontal federated learning (HFL)},
	description={HFL\index{horizontal federated learning (HFL)} uses \glspl{localdataset} constitut\-ed by different
	   	\glspl{datapoint} but uses the same \glspl{feature} to characterize them \cite{HFLChapter2020}.
		For example, weather forecasting uses a network of spatially distributed
		weather (observation) stations. Each weather station measures the
		same quantities, such as daily temperature, air pressure, and precipitation.
		However, different weather stations measure the characteristics or
		\glspl{feature} of different spatiotemporal regions. Each spatiotemporal region 
		represents an individual \gls{datapoint}, each characterized by the same \glspl{feature} 
		(e.g., daily temperature or air pressure).
		\\
		See also: \gls{ssl}, \gls{fl}, \gls{vfl}.},
	first={horizontal federated learning (HFL)},
	text={HFL}
} 

\newglossaryentry{dimred}
{name={dimensionality reduction},
	description={Dimensionality reduction\index{dimensionality reduction} refers 
		to methods that learn a transformation 
		$\hypothesis: \mathbb{R}^{\nrfeatures} \rightarrow \mathbb{R}^{\nrfeatures'}$ 
		of a (typically large) set of raw \glspl{feature} $\feature_{1}, \,\ldots, \,\feature_{\nrfeatures}$ 
		into a smaller set of informative \glspl{feature} $z_{1}, \,\ldots, \,z_{\nrfeatures'}$. 
		Using a smaller set of \glspl{feature} is beneficial in several ways: 
		\begin{itemize} 
			\item {Statistical benefit:} It typically reduces the \gls{risk} of \gls{overfitting}, as 
			reducing the number of \glspl{feature} often reduces the \gls{effdim} of a \gls{model}. 
			\item {Computational benefit:} Using fewer \glspl{feature} means less computation 
			for the \gls{training} of \gls{ml} \glspl{model}. As a case in point, \gls{linreg} methods 
			need to invert a \gls{matrix} whose size is determined by the number of \glspl{feature}. 
			\item {Visualization:} Dimensionality reduction is also instrumental for \gls{data} visualization. 
			For example, we can learn a transformation that delivers two \glspl{feature} $z_{1},z_{2}$, 
			which we can use, in turn, as the coordinates of a \gls{scatterplot}. Fig.\ \ref{fig:dimred-scatter_dict} 
			depicts the \gls{scatterplot} of handwritten digits that are placed 
			using transformed \glspl{feature}. Here, the \glspl{datapoint} are 
			naturally represented by a large number of greyscale values (one value for each pixel).
		\end{itemize} 
		 \begin{figure}[H]
		 \centering
		 \begin{tikzpicture}[scale=1]	
		 	\draw[->] (-0.5,0) -- (5.5,0) node[right] {$z_1$};
		 	\draw[->] (0,-0.5) -- (0,4.5) node[above] {$z_2$};
		 	\foreach \x/\y/\label in {
  		 		1.2/0.5/3,
  		 		0.8/2.0/8,
  		 		2.5/1.8/1,
  		 		3.8/3.5/6,
  		 		4.2/0.7/9,
  		 		2.8/3.0/7,
  		 		1.5/3.8/2
		 	}{
  		 		\node[draw, minimum size=0.6cm, inner sep=0pt] at (\x,\y)
    	 		{\label};
		 	}
		 	\end{tikzpicture}
		 	\caption{Example of dimensionality reduction: High-dimensional image \gls{data} 
			(e.g., high-resolution images of handwritten digits) embedded into 2-D using 
			learned \glspl{feature} $(z_1, z_2)$ and visualized in a \gls{scatterplot}.}
		 	\label{fig:dimred-scatter_dict}
		 \end{figure}
		See also: \gls{overfitting}, \gls{autoencoder}, \gls{randomprojection}, \gls{pca}, \gls{johnsonlindenstrausslemma}.}, 
	first={dimensionality reduction},
	text={dimensionality reduction}
} 

\newglossaryentry{diagnosis}
{name={diagnosis},
	description={Consider\index{diagnosis} an \gls{erm}-based method that resulted in a trained \gls{model} 
		(or learned \gls{hypothesis}) $\learnthypothesis \in \hypospace$. We can diagnose the 
		method by comparing the \gls{trainerr} $\trainerror$ with the \gls{valerr} $\valerror$ 
		incurred by $\learnthypothesis$ on the \gls{trainset} and the \gls{valset}. 
		\begin{figure}[H]
			\begin{center}
			\begin{tikzpicture}[ycomb]
				\draw[color=green,line width=26pt]
				plot coordinates{(0,3)};
				\node [below] at (0,0) {\gls{trainerr}} ; 
				\draw[color=red,line width=26pt]
				plot coordinates{(5,5)};
				\node [below] at (5,0) {\gls{valerr}} ; 
				\draw[dashed,line width=2] (-1,2) -- (7,2) node[right,text width=5cm]{\gls{baseline} \\ (e.g., \gls{bayesrisk}, existing \gls{ml} methods, or human performance)};
			\end{tikzpicture}
			\end{center}
		\caption{We can diagnose an \gls{erm}-based \gls{ml} method by comparing its \gls{trainerr} 
			with its \gls{valerr}. Ideally, both are on the same level as a \gls{baseline}.\label{fig_diagnosis_dict}}
		\end{figure}
		See also: \gls{baseline}, \gls{validation}, \gls{kfoldcv}, \gls{generalization}.},
	first={diagnosis},
	text={diagnosis}
} 

\newglossaryentry{ml}
{name={machine learning (ML)},
	description={ML\index{machine learning (ML)} methods aim to learn (or find) 
		a useful \gls{hypothesis} \gls{map} $\learnthypothesis\!\in\!\hypospace$ out of 
		a \gls{model} $\hypospace$. The learned $\learnthypothesis$ is used to compute 
		a \gls{prediction} $\predictedlabel=\learnthypothesis(\featurevec)$ for 
		the \gls{label} $\truelabel$ of a \gls{datapoint}. The learning process is guided 
		by a quantitative \gls{measure} of the \gls{loss} incurred when  
		the \glspl{prediction} obtained from the learned \gls{hypothesis} differ from  
		the actual \gls{label} $\truelabel$. Different ML methods use different 
		design choices for this quantitative \gls{measure} (or \gls{lossfunc}) as well as 
		different choices for the \gls{model} and the \glspl{datapoint} (i.e., 
		their \glspl{feature} and \glspl{label}) \cite[Ch. 3]{MLBasics}. 
		\begin{figure}[H]
			\centering
			\hspace*{5mm}
			\begin{tikzpicture}[
				node distance=1cm,
				auto,
				punkt/.style={
				rectangle,
				rounded corners,
				draw=black,
				very thick,
				text width=3cm,
				minimum height=1cm,
				align=center
				}
				]
				\coordinate (OR) at (0.00, 1.50);
				\node[circle,inner sep=0,minimum size={6cm}](a) at (OR) {};
				\node[circle,inner sep=0,minimum size={8cm}](b) at (OR) {};
				\node[circle,inner sep=0,minimum size={8cm}](c) at (OR) {};
				\draw[red,line width=1,dashed] (a.-150) arc (-150:{-150+120}:3cm);
				\draw[blue,line width=1,dashed] (b.160) arc (160:{160+50}:4cm);
				\draw[black!30!green,line width=2,dashed] (c.20) arc (20:{160}:4cm);
				\draw[black!30!blue,line width=2,dashed] ([shift={(-30:6cm)}]OR) arc (-30:{20}:6cm);
				\draw[black!30!blue,line width=2,dashed] (OR) --([shift={(-30:6cm)}]OR);
				\draw[black!30!blue,line width=2,dashed] (OR) --([shift={(20:6cm)}]OR);
				\draw[black!30!green,line width=2,dashed] (OR) -- (c.20);
				\draw[black!30!green,line width=2,dashed] (OR) -- (c.160);
				\draw[blue,thin,dashed] (OR) -- (b.160);
				\draw[blue,thin,dashed] (OR) -- (b.210);
				\draw[red,thin,dashed] (OR) -- (a.-150);
				\draw[red,thin,dashed] (OR) -- (a.-30);   
				\node[punkt] (data) {observations};
				\node[red,below=0.5cm of data] (data1) {\gls{data}};
				\node[above=of data] (dummy) {};
				\node[punkt,above=1cm of dummy] (hypothesis) {\gls{hypothesis}};
				\node[right=1.4cm of dummy] (t) {make a \gls{prediction}} ; 
				\node[left=1.4cm of dummy] (g) {assess and adapt} ;
				\node[blue,below=0.5cm of g,anchor=east] (g1) {\gls{loss}} ; 
				\node[black!30!blue,below=1cm of t,anchor=west] (g6) {\gls{inference}} ; 
				\node[black!30!green,above=1cm of hypothesis,anchor=south] (g3) {\gls{model}} ; 
				\draw [->,line width=0.5mm] (hypothesis.east) to [out=0,in=90] (t.north);
				\draw [->,line width=0.5mm] (t.south) to [out=270,in=0] (data.east);
				\draw [->,line width=0.5mm] (data.west) to [out=180,in=270] (g.south);
				\draw [->,line width=0.5mm] (g.north) to [out=90,in=180] (hypothesis.west);
			\end{tikzpicture}
			\vspace*{-9mm}
		\caption{ML learns a \gls{hypothesis} out of a \gls{model} (or \gls{hypospace}) 
			by trying to minimize the \gls{loss} incurred by the \glspl{prediction} 
			for the \glspl{label} of \glspl{datapoint}. The \glspl{prediction} are 
			computed solely from the \glspl{feature} of the \glspl{datapoint}.}
			\label{fig_AlexMLBP_dict}
		\end{figure}
		Another distinction between ML methods is how they access \glspl{datapoint} during 
		learning. For example, some methods have access to a complete \gls{dataset} 
		during \gls{training}, which allows them to use \gls{erm} \cite{MLBasics}, \cite{HastieWainwrightBook}. 
		In contrast, \gls{onlinelearning} methods access \gls{data} 
		sequentially and, in turn, update the learned \gls{hypothesis} 
		whenever a new \gls{datapoint} arrives \cite{PredictionLearningGames}, \cite{HazanOCO}, \cite{Bottou99}.
	 			\\ 
		See also: \gls{model}, \gls{loss}, \gls{data}, \gls{erm}, \gls{onlinelearning}.},
	first={machine learning (ML)},
	text={ML}
}

\newglossaryentry{featlearn}
{name={feature learning},
	description={Consider an \gls{ml} application with \glspl{datapoint} characterized by 
		raw \glspl{feature} $\featurevec \in \featurespace$. \Gls{feature} learning\index{feature learning} 
		refers to the task of learning a \gls{map} 
		$$\featuremapvec: \featurespace \rightarrow \featurespace': \featurevec \mapsto \featurevec'$$ 
		that reads in the \glspl{feature} $\featurevec \in \featurespace$ of a \gls{datapoint} and delivers new 
		\glspl{feature} $\featurevec' \in \featurespace'$ from a new \gls{featurespace} $\featurespace'$. 
		Different \gls{feature} learning methods are obtained for different design 
		choices of $\featurespace,\featurespace'$, for a \gls{hypospace} $\hypospace$ 
		of potential \glspl{map} $\featuremapvec$, and for a quantitative \gls{measure} of the usefulness of 
		a specific $\featuremapvec \in \hypospace$. For example, \gls{pca} 
		uses $\featurespace \defeq \mathbb{R}^{\dimlocalmodel}$, $\featurespace' \defeq \mathbb{R}^{\dimlocalmodel'}$ 
		with $\dimlocalmodel' < \dimlocalmodel$, and a \gls{hypospace} 
		$$\hypospace\defeq \big\{ \featuremapvec: \mathbb{R}^{\dimlocalmodel}
		\!\rightarrow\! \mathbb{R}^{\dimlocalmodel'}\!:\!\featurevec'\!\defeq\!\mF \featurevec \mbox{ with some } \mF \!\in\! \mathbb{R}^{\dimlocalmodel' \!\times \dimlocalmodel} \big\}.$$ 
		\Gls{pca} measures the usefulness of a specific \gls{map} $\featuremapvec(\featurevec)= \mF \featurevec$ 
		by the \gls{minimum} linear reconstruction error incurred on a \gls{dataset} such that 
		$$\min_{\mG \in \mathbb{R}^{\dimlocalmodel \!\!\!\times \dimlocalmodel'}} \sum_{\sampleidx=1}^{\samplesize} \normgeneric{\mG \mF \featurevec^{(\sampleidx)} - \featurevec^{(\sampleidx)}}{2}^{2}.$$ 
			\\ 
		See also: \gls{feature}, \gls{featurespace}, \gls{hypospace}, \gls{pca}.}, 
	first={feature learning},
	text={feature learning}
}

\newglossaryentry{encoder} 
{name={encoder}, 
	description={See \gls{autoencoder}\index{encoder}.}, 
 	first={encoder}, 
 	plural={encoders}, 
 	text={encoder}, 
 	firstplural={encoders}
}

\newglossaryentry{autoencoder}
{name={autoencoder},
	description={An autoencoder\index{autoencoder} is an \gls{ml} method that simultaneously 
		learns an \gls{encoder} \gls{map} $\hypothesis \in \hypospace$ and a decoder \gls{map} 
		$\hypothesis^{*} \in \hypospace^{*}$. Different autoencoders use different 
		\glspl{model} $\hypospace, \hypospace^{*}$, e.g., \glspl{ann} with different architectures. 
        		The special case of an autoencoder using (\gls{vector}-valued) \glspl{linmodel} for 
		$\hypospace, \hypospace^{*}$ results in \gls{pca}. 
		\begin{figure}[H]
			\centering
			\begin{tikzpicture}[>=Latex, thick, node distance=1.6cm]
			\node (x) {$\vx$};
			\node[draw, rounded corners, right=of x, inner sep=4pt] (enc) {$\text{\gls{encoder} } \hypothesis$};
			\node[right=of enc] (z) {$\vz$};
			\node[draw, rounded corners, right=of z, inner sep=4pt] (dec) {$\text{decoder } \hypothesis^{*}$};
			\node[right=of dec] (xhat) {$\hat{\featurevec}$};
			\draw[->] (x) -- (enc);
			\draw[->] (enc) -- node[above] {$\vz=\hypothesis(\featurevec)$} (z);
			\draw[->] (z) -- (dec);
			\draw[->] (dec) -- node[above] {$\hat{\featurevec}=\hypothesis^{*}(\vz)$} (xhat);
			\end{tikzpicture}
		\caption{Autoencoder with an \gls{encoder} $\hypothesis$ mapping $\featurevec \mapsto \vz$ and 
			a decoder $\hypothesis^*$ mapping $\vz \mapsto \hat{\featurevec}$.}
		\end{figure}
		The \gls{training} of the \gls{encoder} and decoder can be implemented via \gls{erm} using a \gls{loss} that measures 
		the deviation of the reconstructed \gls{featurevec} $\hypothesis^{*}\big(\hypothesis\big(\featurevec \big) \big)$
		from the original \gls{featurevec} $\featurevec$.
					\\ 
		See also: \gls{encoder}, \gls{featlearn}, \gls{dimred}.},
	first={autoencoder},
	text={autoencoder}
} 

\newglossaryentry{modelparallelism}
{name={model parallelism},
sort={modelparallelism},
	description={\Gls{model} parallelism\index{model parallelism} refers to a particular 
		class of distributed \glspl{optmethod} used to train an \gls{ml} \gls{model}.  
		Here, different \glspl{device} store and process disjoint subsets of the 
		\glspl{modelparam}. This approach contrasts with \gls{dataparallelism}, where each 
		\gls{device} maintains a full replica of the \glspl{modelparam} while 
		processing disjoint subsets of the global \gls{dataset}. 
		\Gls{model} parallelism allows us to train an \gls{ml} \gls{model} whose 
		\glspl{parameter} cannot fit into the memory of a single \gls{device}. 
		One key application of \gls{model} parallelism is the \gls{training} of 
		extremely large \glspl{ann}, such as \gls{transformer} \glspl{model} with 
		billions of \glspl{modelparam}. 
             	\\
             	See also: \gls{vfl}.},
 	first={model parallelism},
 	text={model parallelism}
}

\newglossaryentry{dataparallelism}
{name={data parallelism},
	description={\Gls{data} parallelism\index{data parallelism} refers to a widely used 
 		class of distributed \glspl{optmethod} for \gls{training} an \gls{ml} 
             	\gls{model}. Here, each participating \gls{device} stores a full 
             	replica of the \glspl{modelparam} but processes a disjoint subset of 
             	the global \gls{dataset} \cite{DistrOptStatistLearningADMM}. 
             	Compared to \gls{modelparallelism}, which distributes the 
             	\glspl{modelparam} across \glspl{device}, \gls{data} parallelism distributes 
             	the computational workload associated with large \glspl{dataset}. 
             	It is especially effective when the \gls{model} fits into the memory 
             	of a single \gls{device}, but the \gls{dataset} or \gls{training} complexity 
		would be prohibitive without parallel processing. 
             	\\
             	See also: \gls{modelparallelism}, \gls{hfl}.},
 	first={data parallelism},
 	text={data parallelism}
}

\newglossaryentry{perplexity}
{name={perplexity},
	description={The\index{perplexity} perplexity of an \gls{rv} $x$ is defined as $e^{H_{x}}$.
         	\\
             	See also: \gls{entropy}.},
 	first={perplexity},
 	text={perplexity}
}

\newglossaryentry{vfl}
{name={vertical federated learning (VFL)},
	description={In VFL,\index{vertical federated learning (VFL)} different \glspl{device} have 
		access to different \glspl{feature} of the same set of \glspl{datapoint} \cite{VFLChapter}. 
		Formally, the underlying global \gls{dataset} is
		\[
		\dataset^{(\mathrm{global})} \defeq \left\{ \left(\featurevec^{(1)}, \truelabel^{(1)}\right), \,\ldots, \,\left(\featurevec^{(\samplesize)}, \truelabel^{(\samplesize)}\right) \right\}.
		\]
		We denote by $\featurevec^{(\sampleidx)} = \big( \feature^{(\sampleidx)}_{1}, \,\ldots, \,\feature^{(\sampleidx)}_{\nrfeatures'} \big)\,^{T}$, for $\sampleidx=1, \,\ldots, \,\samplesize$, 
	     	the complete \glspl{featurevec} for the \glspl{datapoint}. Each \gls{device} $\nodeidx \in \nodes$ 
		observes only a subset $\mathcal{F}^{(\nodeidx)} \subseteq \{1, \,\ldots, \,\nrfeatures'\}$ of \glspl{feature}, resulting 
		in a \gls{localdataset} $\localdataset{\nodeidx}$ with \glspl{featurevec}
		\[
		\featurevec^{(\nodeidx,\sampleidx)} = \big( \feature^{(\sampleidx)}_{\featureidx_{1}}, \,\ldots, \,\feature^{(\sampleidx)}_{\featureidx_{\nrfeatures}} \big)\,^{T}.
		\]
		Some of the \glspl{device} may also have access to the \glspl{label} $\truelabel^{(\sampleidx)}$, for $\sampleidx=1, \,\ldots, \,\samplesize$, 
		of the global \gls{dataset} (see Fig.\ \ref{fig_vertical_FL_dict}). 
		\begin{figure}[H]
			\begin{center}
			\begin{tikzpicture}[every node/.style={anchor=base}]

				\def\colZ{3.2}
				
				\def\colLabel{6.4}

				\foreach \i/\label in {1/1, 2/2, 4/\samplesize} {
					\pgfmathsetmacro{\y}{-1.2*(\i-1)}
					\node (x\i1) at (0,\y) {$x^{(\label)}_{1}$};
					\node (x\i2) at (1.6,\y) {$x^{(\label)}_{2}$};
					\node (dots\i) at (3.2,\y) {$\cdots$};
					\node (x\i3) at (4.8,\y) {$x^{(\label)}_{\dimlocalmodel}$};
					\node (y\i) at (6.4,\y) {$\truelabel^{(\label)}$};
				}
				\draw[dashed, rounded corners, thick]
				(-0.6,0.6) rectangle (6.9,-4.2);
				\node at (3.1,0.9) {$\dataset^{(\mathrm{global})} $};
				\draw[dashed, rounded corners, thick]
				(-0.9,0.9) rectangle (2.1,-4.0);
				\node at (0.25,1.0) {$\localdataset{1}$};
				\draw[dashed, rounded corners, thick]
				($( \colZ + 1,,0.9 )$) rectangle
				($( \colLabel + 0.4, -4.5)$);
				\node at ($( \colZ + 0.9,-5 )$) {$\localdataset{\nodeidx}$};
			\end{tikzpicture}
			\end{center}
		\caption{VFL uses \glspl{localdataset} that are derived from the \glspl{datapoint} of a common global \gls{dataset}. 
			The \glspl{localdataset} differ in the choice of \glspl{feature} used to characterize 
			the \glspl{datapoint}.\label{fig_vertical_FL_dict}}
		\end{figure}
		One potential application of VFL is to enable collaboration between 
		different health-care providers. Each provider collects distinct types of measurements—such as blood 
		values, electrocardiography, and lung X-rays—for the same patients. Another 
		application is a national social insurance system, where health records, financial 
		indicators, consumer behavior, and mobility \gls{data} are collected by different 
		institutions. VFL enables joint learning across these parties while allowing 
		well-defined levels of \gls{privprot}. We can view VFL as a specific form of
		\gls{modelparallelism}.
		\\
		See also: \gls{privprot}, \gls{fl}, \gls{modelparallelism}.},
	first={vertical federated learning (VFL)},
	text={VFL}
}

\newglossaryentry{multitask learning}
{name={multitask learning},
	description={Multitask learning\index{multitask learning} aims to leverage relations between 
	 	different \glspl{learningtask}. Consider two \glspl{learningtask} obtained from the 
	 	same \gls{dataset} of webcam snapshots. The first task is to predict the presence 
	 	of a human, while the second task is to predict the presence of a car. It may be useful 
	 	to use the same \gls{deepnet} structure for both tasks and only allow the \glspl{weight} of 
	 	the final \gls{output} \gls{layer} to be different.
	 			\\ 
		See also: \gls{learningtask}, \gls{dataset}, \gls{deepnet}, \gls{weight}, \gls{layer}.},
	first={multitask learning},
	text={multitask learning}
}

\newglossaryentry{learningtask}
{name={learning task},
	description={Consider\index{learning task} a \gls{dataset} $\dataset$ consisting of 
		multiple \glspl{datapoint} $\datapoint^{(1)},\,\ldots,\,\datapoint^{(\samplesize)}$. 
		For example, $\dataset$ can be a collection of images in an image database. 
		A learning task is defined by specifying those properties (or attributes) of a \gls{datapoint} 
		that are used as its \glspl{feature} and \glspl{label}. Given a choice of \gls{model} $\hypospace$ and 
		\gls{lossfunc}, a learning task leads to an instance of \gls{erm} and can thus be 
		represented by the associated \gls{objfunc} $\emprisk{\hypothesis}{\dataset}$ for $\hypothesis \in \hypospace$. 
		Importantly, multiple distinct learning tasks can be constructed from the same \gls{dataset} 
		by selecting different sets of \glspl{feature} and \glspl{label} (see Fig.\ \ref{fig:learning_tasks_cows_dict}).
    		\begin{figure}[H]
			\centering
			\begin{minipage}[t]{0.95\textwidth}
    			\centering
    			\includegraphics[width=\textwidth]{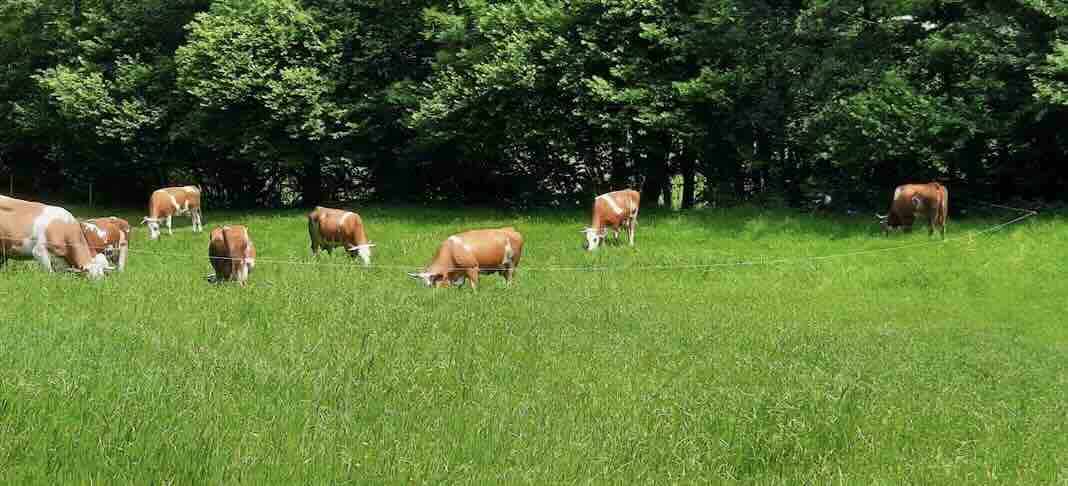}
    			\caption*{An image showing cows grazing in the Austrian countryside.}
			\vspace{5mm}
			\end{minipage}
			\vspace{5mm}
			\begin{minipage}[t]{0.45\textwidth}
    			Task 1 (\gls{regression}): \\
        			\Glspl{feature} are the RGB values of all image pixels,
        			and the \gls{label} is the number of cows depicted.
			\end{minipage}
			\hfill
			\begin{minipage}[t]{0.45\textwidth}
    			Task 2 (\gls{classification}): \\
			\Glspl{feature} include the average green intensity of the image, 
        			and the \gls{label} indicates whether cows should be moved to another location (i.e., yes/no).
			\end{minipage}
			\caption{Two learning tasks constructed from a single image \gls{dataset}. 
			These tasks differ in \gls{feature} selection and choice of \gls{label} (i.e., the objective), 
			but are both derived from the same \gls{dataset}.}
			\label{fig:learning_tasks_cows_dict}
		\end{figure}
		Different learning tasks arising from the same underlying \gls{dataset} are often coupled. 
		For example, when a \gls{probmodel} is used to generate \glspl{datapoint}, statistical 
		dependencies among different \glspl{label} induce dependencies among the corresponding 
		learning tasks. In general, solving learning tasks jointly, e.g., using \gls{multitask learning} 
		methods, tends to be more effective than solving them independently (thereby ignoring 
		dependencies among learning tasks) \cite{Caruana:1997wk}, \cite{JungGaphLassoSPL}, \cite{CSGraphSelJournal}.
	 			\\ 
		See also: \gls{multitask learning}, \gls{labelspace}.},
	first={learning task},
	firstplural={learning tasks},
	plural={learning tasks}, 
	text={learning task}
}

\newglossaryentry{lime}
{name={local interpretable model-agnostic explanations (LIME)},
	description={Consider\index{local interpretable model-agnostic explanations (LIME)} 
		a trained \gls{model} (or learned \gls{hypothesis}) $\widehat{\hypothesis} \in \hypospace$, 
		which maps the \gls{featurevec} of a \gls{datapoint} to the \gls{prediction} $\widehat{\truelabel}= \widehat{\hypothesis}$. 
		LIME is a technique for explaining the behavior of $\widehat{\hypothesis}$ 
		locally around a \gls{datapoint} with \gls{featurevec} $\featurevec^{(0)}$ \cite{Ribeiro2016}. 
		The \gls{explanation} is given in the form of a local approximation $g \in \hypospace'$ of $\widehat{\hypothesis}$ 
		(see Fig.\ \ref{fig_lime_dict}). This approximation can be obtained by an instance of \gls{erm} 
		with a carefully designed \gls{trainset}. In particular, the \gls{trainset} consists of \glspl{datapoint} with 
		\glspl{featurevec} centered around $\featurevec^{(0)}$ and the (pseudo-)\gls{label} $\widehat{\hypothesis}(\featurevec)$. 
		Note that we can use a different \gls{model} $\hypospace'$ for the approximation from 
		the original \gls{model} $\hypospace$. For example, we can use a \gls{decisiontree} 
		to locally approximate a \gls{deepnet}. Another widely used choice for $\hypospace'$ is 
		the \gls{linmodel}. 
		\begin{figure}[H]
		\begin{center}
		\begin{tikzpicture}
			\begin{axis}[
				axis lines=middle,
				xlabel={$\featurevec$},
				ylabel={$\truelabel$},
				xtick=\empty,
				ytick=\empty,
				xmin=0, xmax=6,
				ymin=0, ymax=6,
				domain=0:6,
				samples=100,
				width=10cm,
				height=6cm,
				clip=false
			]
  			\addplot[blue, thick, domain=0:6] {2 + sin(deg(x))} node[pos=0.85, above right,yshift=3pt] {$\widehat{\hypothesis}(\featurevec)$};
  			\addplot[dashed, gray] coordinates {(3,0) (3,6)};
  			\addplot[red, thick, domain=2.5:3.5] {2 + sin(deg(3))} node[pos=0.9, above] {$g(\featurevec)$};
  			\addplot[mark=*] coordinates {(3, {2 + sin(deg(3))})};
			\node at (axis cs:3,-0.3) {$\featurevec^{(0)}$};
			\end{axis}
		\end{tikzpicture}
		\end{center}
		\caption{To explain a trained \gls{model} $\widehat{\hypothesis} \in \hypospace$, around a 
			given \gls{featurevec} $\featurevec^{(0)}$, we can use a local approximation $g \in \hypospace'$. }
			\label{fig_lime_dict}
		\end{figure}
		See also: \gls{model}, \gls{explanation}, \gls{erm}, \gls{trainset}, \gls{label}, \gls{decisiontree}, \gls{deepnet}, \gls{linmodel}.},
	first={LIME},
	text={LIME}
}

\newglossaryentry{linmodel}
{name={linear model}, 
	description={Consider\index{linear model} an \gls{ml} application involving \glspl{datapoint}, each represented 
		by a numeric \gls{featurevec} $\featurevec \in \mathbb{R}^{\nrfeatures}$. A linear \gls{model} defines 
		a \gls{hypospace} consisting of all real-valued \glspl{linearmap} from $\mathbb{R}^{\nrfeatures}$ to $\mathbb{R}$ such that
		\begin{equation}
			\nonumber
			\label{equ_def_lin_model_hypspace_dict}
			\linmodel{\nrfeatures} \defeq \left\{ \hypothesis: \mathbb{R}^{\nrfeatures} \rightarrow \mathbb{R} \mid \hypothesis(\featurevec) = \weights^{\top} \featurevec \text{ for some } \weights \in \mathbb{R}^{\nrfeatures} \right\}.
		\end{equation}
		Each value of $\nrfeatures$ defines a different \gls{hypospace}, corresponding to the number of 
		\glspl{feature} used to compute the \gls{prediction} $\hypothesis(\featurevec)$. The choice of 
		$\nrfeatures$ is often guided not only by \glspl{compasp} (e.g., fewer \glspl{feature} reduce computation) and 
		\glspl{statasp} (e.g., more \glspl{feature} typically reduce \gls{bias} and \gls{risk}), but also by \gls{interpretability}. 
		A linear \gls{model} using a small number of well-chosen \glspl{feature} is generally considered 
		more interpretable \cite{rudin2019stop}, \cite{Ribeiro2016}.
		The linear \gls{model} is attractive because it can typically be trained using scalable \gls{convex} 
		\glspl{optmethod} \cite{BertsekasNonLinProgr}, \cite{hastie01statisticallearning}. 
		Moreover, linear \glspl{model} often permit rigorous 
		statistical analysis, including fundamental limits on the \gls{minimum} achievable \gls{risk} \cite{Wain2019}. 
		They are also useful for analyzing more complex nonlinear \glspl{model} such as \glspl{ann}. For instance, 
		a \gls{deepnet} can be viewed as the composition of a \gls{featuremap}—implemented by the input and 
		hidden \glspl{layer}—and a linear \gls{model} in the \gls{output} \gls{layer}. Similarly, a \gls{decisiontree} can be interpreted 
		as applying a one-hot-encoded \gls{featuremap} based on \glspl{decisionregion}, followed by a linear 
		\gls{model} that assigns a \gls{prediction} to each region.
		More generally, any trained \gls{model} $\learnthypothesis \in \hypospace$ that is 
		\gls{differentiable} at some $\featurevec'$ can be locally approximated by a \gls{linearmap} 
		$g(\featurevec)$. Fig.\ \ref{fig_linapprox_dict} illustrates such a local linear approximation, 
		defined by the \gls{gradient} $\nabla \learnthypothesis(\featurevec')$. Note that the \gls{gradient} 
		is only defined where $\learnthypothesis$ is \gls{differentiable}.
		To ensure \gls{robustness} in the context of \gls{trustAI}, one may prefer \glspl{model} whose 
		associated \gls{map} $\learnthypothesis$ is Lipschitz \gls{continuous}. A classic result in mathematical 
		analysis—Rademacher’s theorem—states that if $\learnthypothesis$ is Lipschitz \gls{continuous} with 
		some constant $L$ over an open set $\samplespace \subseteq \mathbb{R}^{\nrfeatures}$, then $\learnthypothesis$ 
		is \gls{differentiable} almost everywhere in $\samplespace$ \cite[Th.~3.1]{heinonen2005lectures}.
		\begin{figure}[H]
			\begin{center}
			\begin{tikzpicture}[x=0.5cm]
			\begin{axis}[
			hide axis,
			xmin=-3, xmax=6,
			ymin=0, ymax=6,
			domain=0:6,
			samples=100,
			width=10cm,
			height=6cm,
			clip=false
			]
			\addplot[blue, thick, domain=-2:6] {2 + sin(deg(x))} 
			node[pos=0.5, above right, yshift=3pt] {$\learnthypothesis(\featurevec)$};
			\addplot[red, thick, domain=4.5:6.5] 
			{2 + sin(deg(6)) + cos(deg(6))*(x - 6)}
			node[pos=0.95, above right] {$g(\featurevec)$};
			\addplot[mark=*] coordinates {(6, {2 + sin(deg(6))})};
			\addplot[dashed, gray] coordinates {(6,0) (6,2.4)};
			\node at (axis cs:6, -0.2) {$\featurevec'$};
			\pgfmathsetmacro{\xA}{-1.5}
			\pgfmathsetmacro{\xB}{3}
			\pgfmathsetmacro{\yA}{2 + sin(deg(\xA))}
			\pgfmathsetmacro{\yB}{2 + sin(deg(\xB))}
			\addplot[mark=*, only marks] coordinates {(\xA, \yA) (\xB, \yB)};
			\draw[dashed, gray] (axis cs:\xA,\yA) -- (axis cs:\xA,0);
			\draw[dashed, gray] (axis cs:\xB,\yB) -- (axis cs:\xB,0);
			\draw[dashed, gray] (axis cs:\xA,\yA) -- (axis cs:0,\yA);
			\draw[dashed, gray] (axis cs:\xB,\yB) -- (axis cs:0,\yB);
			\draw[<->, thick] (axis cs:\xA,-0.4) -- node[below] {$\normgeneric{\Delta \featurevec}{2}$} (axis cs:\xB,-0.4);
			\draw[<->, thick] (axis cs:-2.4,\yA) -- node[left] {$\leq L \normgeneric{\Delta \featurevec}{2}$} (axis cs:-2.4,\yB);
			\end{axis}
			\vspace*{-10mm}
			\end{tikzpicture}
			\vspace*{-5mm}
			\end{center}
		\caption{Trained \gls{model} $\learnthypothesis(\featurevec)$ that is \gls{differentiable} at a point $\featurevec'$, 
			locally approximated by a \gls{linearmap} $g \in \linmodel{\nrfeatures}$. This local approximation 
			is determined by the \gls{gradient} $\nabla \learnthypothesis(\featurevec')$.}
			\label{fig_linapprox_dict}
		\end{figure}
		See also: \gls{model}, \gls{hypospace}, \gls{linearmap}, \gls{interpretability}, \gls{lime}.}, 
   first={linear model},
   plural={linear models},
   firstplural={linear models}, 
   text={linear model}
}

\newglossaryentry{earlystopping}
{name={early stopping}, 
	description={Consider an \gls{erm}-based method that uses an 
		iterative \gls{optmethod} (such as \gls{gd}) to learn \glspl{modelparam} 
		by minimizing the \gls{emprisk} on a given \gls{trainset}.
		Early stopping\index{early stopping} refers to terminating the \glspl{iteration} 
		even if they still substantially decrease the \gls{emprisk} on the 
		\gls{trainset}. Instead of monitoring the \gls{objfunc} (which is the 
		\gls{emprisk} on the \gls{trainset}), early stopping monitors the \gls{valerr} 
		incurred by the \glspl{modelparam} in each \gls{iteration}. Early stopping can be 
		interpreted as an implementation of \gls{regularization} via \gls{model} 
		pruning. Indeed, terminating an iterative \gls{optmethod} after a small number 
		of \glspl{iteration} restricts the set of \glspl{modelparam} that can be reached from the 
		initialization (see Fig.\ \ref{fig_early_stopping_dict}).
		\begin{figure}[H]
		\centering
			\begin{tikzpicture}[>=Stealth, scale=2]
			\fill (0,0) circle (0.6pt) node[above] {\small $\weights^{(0)}$};
			\node at (-0.4,0) {\small $\hypospace^{(1)}$};
			\node at (-1.2,0) {\small $\hypospace^{(2)}$};
			\node at (-2,0) {\small $\hypospace \ldots$};
			\draw[densely dotted] (0,0) ellipse (0.8 and 0.4);
			\draw[dashed] (0,0) ellipse (1.6 and 0.8);
			\draw[->] (0,0) -- (0.8,0.) node [pos=0.6,above] {\tiny $1$ step};
			\draw[->] (0.0,0.0) -- (0,-0.8) node [pos=0.9,right] {\tiny $2$ steps};
			\end{tikzpicture}
		\caption{\Gls{gdmethod} for \gls{erm} using a \gls{hypospace} $\hypospace$ 
		    	defines a nested \gls{sequence} of effective \glspl{hypospace} 
			$\hypospace^{(1)} \subseteq \hypospace^{(2)} \subseteq \ldots \subseteq \hypospace$. 
			The effective \gls{hypospace} $\hypospace^{(\iteridx)}$ is 
		    	determined by all \glspl{modelparam} that can be reached from the initialization 
			$\weights^{(0)}$ within $\iteridx$ \glspl{gradstep}. \label{fig_early_stopping_dict}}
		\end{figure}
		See also: \gls{regularization}, \gls{gdmethod}, \gls{overfitting}.},
	first={early stopping},
	text={early stopping}
}

\newglossaryentry{statasp}
{name={statistical aspect}, 
	description={By statistical aspects\index{statistical aspect} 
		of an \gls{ml} method, we refer to (properties of) the \gls{probdist} of its \gls{output} 
		under a \gls{probmodel} for the \gls{data} fed into the method.
					\\ 
		See also: \gls{ml}, \gls{probdist}, \gls{probmodel}, \gls{data}.},
	firstplural={statistical aspects},
	plural={statistical aspects},
	first={statistical aspect},
	text={statistical aspect}
}

\newglossaryentry{compasp}
{name={computational aspect}, 
	description={By computational 
		aspects\index{computational aspect} of an \gls{ml} method, we mainly refer to the computational 
		resources required for its implementation. For example, if an \gls{ml} method uses iterative 
		\gls{optimization} techniques to solve \gls{erm}, then its computational aspects include: 1) how 
		many arithmetic operations are needed to implement a single \gls{iteration} (i.e., a \gls{gradstep}); 
		and 2) how many \glspl{iteration} are needed to obtain useful \glspl{modelparam}. One important 
		example of an iterative \gls{optimization} technique is \gls{gd}.
					\\ 
		See also: \gls{ml}, \gls{erm}, \gls{gradstep}, \gls{modelparam}, \gls{gd}.}, 
	first={computational aspect},
	text={computational aspect},
	firstplural={computational aspects},
	plural={computational aspects}
}

\newglossaryentry{zerooneloss}
{name={$\bf 0/1$ loss},
sort={zerooneloss},
	description={The $0/1$ \gls{loss}\index{$0/1$ loss} $\lossfunczo{\pair{\featurevec}{\truelabel}}{\hypothesis}$ 
		measures the quality of a \gls{classifier} $\hypothesis(\featurevec)$ that delivers a 
		\gls{prediction} $\predictedlabel$ (e.g., via thresholding \eqref{equ_def_threshold_bin_classifier_dict}) 
		for the \gls{label} $\truelabel$ of a \gls{datapoint} with \glspl{feature} $\featurevec$. It is equal to $0$ if 
		the \gls{prediction} is correct, i.e., 
		$\lossfunczo{\pair{\featurevec}{\truelabel}}{\hypothesis}=0$ when $\predictedlabel=\truelabel$. It is 
		equal to $1$ if the \gls{prediction} is wrong, i.e., $\lossfunczo{\pair{\featurevec}{\truelabel}}{\hypothesis}=1$ 
		when $\predictedlabel\neq\truelabel$.
				\\ 
		See also: \gls{loss}, \gls{classifier}, \gls{prediction}, \gls{acc}, \gls{label}, \gls{datapoint}, \gls{feature}.},
    	first={$0/1$ loss},
	text={$0/1$ loss}
}

\newglossaryentry{probability}
{name={probability}, 
	description={We\index{probability} assign a probability value, typically chosen in the 
		interval $[0,1]$, to each \gls{event} that can occur in a \gls{randomexperiment}  
		\cite{BillingsleyProbMeasure}, \cite{BertsekasProb}, \cite{KallenbergBook}, \cite{HalmosMeasure}.
		\\
		See also: \gls{event}, \gls{randomexperiment}. },
	first={probability},
	firstplural={probabilities},
	plural={probabilities},
	text={probability}
}
	
\newglossaryentry{underfitting}
{name={underfitting},
	description={Consider\index{underfitting} an \gls{ml} method applying \gls{erm} 
		to learn a \gls{hypothesis} that minimizes the \gls{emprisk} 
		on a given \gls{trainset}. The method is said to underfit if it fails to achieve a sufficiently low \gls{emprisk} on the \gls{trainset}. 
		Underfitting typically occurs when the chosen \gls{model} is too simple 
		to capture the underlying relationship between \glspl{feature} and \glspl{label}. 
		\begin{figure}[H]
		\begin{center}
			\begin{tikzpicture}[scale=1.0]
				\def\slope{0.35}
				\def\intercept{2.0}
				\draw[thick,dashed,domain=0:6.5,variable=\x]
				plot({\x},{\slope*\x+\intercept});
				\node[above right] at (6.5,{\slope*6.5+\intercept}) {$\hypothesis(\feature)$};
				\foreach \i/\x/\y in {1/0.6/2.4, 2/1.2/2.1, 3/1.8/2.0, 4/2.4/2.3,
					5/3.0/3.1, 6/3.6/4.0, 7/4.2/5.2, 8/4.8/6.0,
					9/5.4/6.3, 10/6.0/6.1} {
					\node[circle,draw,fill=blue!60,minimum size=5pt,inner sep=0pt] (p\i) at (\x,\y) {};
					\draw[<->,color=blue!70] (\x,{\slope*\x+\intercept}) -- (p\i);
				}
			\end{tikzpicture}
		\caption{No linear \gls{hypothesis} $\hypothesis$ can capture the relationship 
			between \glspl{feature} and \glspl{label} for 
			the depicted \gls{trainset}. Thus, any method that uses a 
			\gls{linmodel} will underfit this \gls{trainset}.}
			\label{fig_underfitting_min_dict}
		\end{center}
		\end{figure} 
		For example, an \gls{ml} method using a \gls{linmodel} 
		on \gls{data} with a highly nonlinear relationship between \glspl{feature} 
		and \glspl{label} will not be able to learn a \gls{hypothesis} with small 
		average \gls{loss} on the \gls{trainset}, let alone a low \gls{risk}. 
					\\ 
		See also: \gls{trainset}, \gls{model}, \gls{risk}, \gls{overfitting}.},
	first={underfitting},
	text={underfitting}
}

\newglossaryentry{overfitting}
{name={overfitting},
	description={Consider\index{overfitting} an 
		\gls{ml} method that uses \gls{erm} to learn a \gls{hypothesis} with the 
		\gls{minimum} \gls{emprisk} on a given \gls{trainset}. Such a method is 
		overfitting the \gls{trainset} if it learns a \gls{hypothesis} with an  
		\gls{emprisk} on the \gls{trainset} that is significantly smaller than the 
		average \gls{loss} outside the \gls{trainset}. In other words, if an \gls{ml} 
		method overfits, it has a large \gls{gengap}. 
					\\ 
		See also: \gls{erm}, \gls{gengap}, \gls{generalization}, \gls{validation}.},
	first={overfitting},
	text={overfitting}
}

\newglossaryentry{diffusionmethod}
{name={diffusion method},
	description={A diffusion method\index{diffusion method} is a generative \gls{ml} method that
		trains a \gls{model} to approximate the reversal of a gradual \gls{stochastic}
		corruption process \cite{DeepLearningBishop}, \cite{Prince2023UnderstandingDL}. 
		Diffusion methods train \glspl{model} using a combination of a forward (diffusion) 
		process and a learned reverse \gls{map}.
		In the forward process, random noise is incrementally added to clean
		representations of a \gls{datapoint}, such as an image, an audio recording,
		or other types of \gls{data}. Similar to a \gls{denautoencoder}, the 
		trained \gls{model} can be applied to a noisy representation of a \gls{datapoint}. 
		The resulting \gls{prediction} is a denoised representation. What sets 
		diffusion methods apart from generic \glspl{denautoencoder} is the gradual 
		nature of the corruption process used for \gls{model} \gls{training}.
		\\
		See also: \gls{denautoencoder}, \gls{training}. },
	first={diffusion method},
    	plural={diffusion methods},
    	firstplural={diffusion methods},
    	text={diffusion method}
}

\newglossaryentry{sqerrloss}
{name={squared error loss},
	description={The squared 
		error\index{squared error loss} \gls{loss} measures the \gls{prediction} error of a 
		\gls{hypothesis} $\hypothesis$ when predicting a numeric \gls{label} $\truelabel \in \mathbb{R}$ 
		from the \glspl{feature} $\featurevec$ of a \gls{datapoint}. It is defined as 
		\begin{equation} 
			\nonumber
			\lossfunc{(\featurevec,\truelabel)}{\hypothesis} \defeq \big(\truelabel - \underbrace{\hypothesis(\featurevec)}_{=\predictedlabel} \big)^{2}. 
		\end{equation} 
			\\ 
		See also: \gls{loss}, \gls{prediction}, \gls{hypothesis}, \gls{label}, \gls{feature}, \gls{datapoint}.},
	first={squared error loss},
	text={squared error loss}
}

\newglossaryentry{diffpriv}
{name={differential privacy (DP)},
	description={Consider\index{differential privacy (DP)} some \gls{ml} method $\algomap$ 
  		that reads in a \gls{dataset} (e.g., the \gls{trainset} 
  		used for \gls{erm}) and delivers some \gls{output} $\algomap(\dataset)$. The \gls{output} 
  		could be either the learned \glspl{modelparam} or the \glspl{prediction} for specific \glspl{datapoint}. 
  		DP is a precise \gls{measure} of \gls{privleakage} incurred by revealing the 
  		\gls{output}. Roughly speaking, an \gls{ml} method is differentially private if the \gls{probdist} 
  		of the \gls{output} $\algomap(\dataset)$ remains largely unchanged when the \gls{sensattr} 
  		of one \gls{datapoint} in the \gls{trainset} is changed. Note that DP 
  		builds on a \gls{probmodel} for an \gls{ml} method, i.e., we interpret its \gls{output} $\algomap(\dataset)$ 
  		as the \gls{realization} of an \gls{rv}. The randomness in the \gls{output} can be ensured 
  		by intentionally adding the \gls{realization} of an auxiliary \gls{rv} (i.e., adding noise) to 
  		the \gls{output} of the \gls{ml} method.
				\\ 
		See also: \gls{output}, \gls{privleakage}, \gls{sensattr}, \gls{privattack}, \gls{privfunnel}.}, 
  	first={DP}, 
  	text={DP} 
}

\newglossaryentry{robustness}
{name={robustness},
	description={Robustness\index{robustness} is a key requirement for \gls{trustAI}. It
		refers to the property of an \gls{mlsystem} to maintain acceptable performance even when 
		subjected to different forms of perturbations. These perturbations may affect the \glspl{feature} 
		of a \gls{datapoint} in order to manipulate the \gls{prediction} delivered by a trained \gls{ml} \gls{model}. 
		Robustness also includes the \gls{stability} of \gls{erm}-based methods against perturbations 
		of the \gls{trainset}. Such perturbations can occur within \gls{datapoisoning} \glspl{attack}. 
		\\ 
		See also: \gls{trustAI}, \gls{stability}, \gls{datapoisoning}, \gls{attack}.}, 
	first={robustness}, 
	text={robustness} 
}

\newglossaryentry{stability}
{name={stability},
	description={Mathematically, an \gls{ml} method is a \gls{map} $\algomap$ from a given \gls{dataset} $\dataset$ 
		to an \gls{output} $\algomap(\dataset)$. As a case in point, consider an \gls{erm}-based 
		\gls{ml} method that maps a \gls{trainset} $\dataset$ to the learned \glspl{modelparam} 
		$\algomap(\dataset) = \widehat{\weights}$, which achieve the \gls{minimum} average \gls{loss} 
		on the \gls{trainset}. Instead of the learned \glspl{modelparam}, the 
		\gls{output} $\algomap(\dataset)$ could also be the \glspl{prediction} obtained from 
		the trained \gls{model}. Stability\index{stability} refers to the desirable property 
		of $\algomap$ that small changes in the input \gls{dataset} $\dataset$ result in small 
		changes in the \gls{output} $\algomap(\dataset)$. The notion of stability is intimately related 
		to the notion of \gls{generalization}. In particular, there are formal notions of stability  
		that allow us to bound the \gls{gengap} (see \cite[Ch.~13]{ShalevMLBook}).
		To build intuition, consider the three \glspl{dataset} depicted in Fig.\ \ref{fig_three_data_stability_dict}, each 
		of which is equally likely under the same \gls{data}-generating \gls{probdist}. Since the 
		optimal \glspl{modelparam} are determined by this underlying \gls{probdist}, an accurate 
		\gls{ml} method $\algomap$ should return the same (or very similar) \gls{output} $\algomap(\dataset)$ 
		for all three \glspl{dataset}. In other words, any useful $\algomap$ must be robust to 
		variability in \gls{sample} \glspl{realization} from the same \gls{probdist}, i.e., it must be stable. 
		\begin{figure}[H]
			\centering
			\begin{tikzpicture}
				\begin{axis}[
					axis lines=none,
					xlabel={$\sampleidx$},
					ylabel={},
					legend pos=north west,
					ymin=0, ymax=10,
					xtick={1,2,3,4,5},
					grid style=dashed,
					every axis plot/.append style={very thick}
					]
					\addplot+[only marks,mark=*] coordinates {
						(1,2) (2,4) (3,3) (4,5) (5,7)
					};
					\addplot+[only marks,mark=square*] coordinates {
						(1,3) (2,2) (3,6) (4,4) (5,5)
					};
					\addplot+[only marks,mark=triangle*] coordinates {
						(1,5) (2,7) (3,4) (4,6) (5,3)
					};
				\end{axis}
			\end{tikzpicture}
		\caption{Three \glspl{dataset} $\dataset^{(*)}$, $\dataset^{(\square)}$, 
			and $\dataset^{(\triangle)}$, each sampled independently from the same 
			\gls{data}-generating \gls{probdist}. A stable \gls{ml} method should return 
			similar \glspl{output} when trained on any of these \glspl{dataset}. \label{fig_three_data_stability_dict}}
		\end{figure}
		See also: \gls{generalization}, \gls{robustness}.}, 
	first={stability}, 
	text={stability} 
}

\newglossaryentry{privprot}
{name={privacy protection},
     description={Consider\index{privacy protection} some \gls{ml} method $\algomap$ that reads 
	 	in a \gls{dataset} $\dataset$ and delivers some \gls{output} $\algomap(\dataset)$. The \gls{output} 
	 	could be the learned \glspl{modelparam} $\widehat{\weights}$ or the \gls{prediction} 
	 	$\learnthypothesis(\featurevec)$ obtained for a specific \gls{datapoint} with \glspl{feature} 
	 	$\featurevec$. Many important \gls{ml} applications involve \glspl{datapoint} 
		representing humans. Each \gls{datapoint} is characterized by \glspl{feature} $\featurevec$, 
		potentially a \gls{label} $\truelabel$, and a \gls{sensattr} $\sensattr$ (e.g., a recent medical diagnosis). 
		Roughly speaking, privacy protection means that it should be impossible to infer, from the \gls{output} $\algomap(\dataset)$, 
		any of the \glspl{sensattr} of \glspl{datapoint} in $\dataset$. Mathematically, privacy protection requires non-invertibility 
		of the \gls{map} $\algomap(\dataset)$. In general, just making $\algomap(\dataset)$ non-invertible 
		is typically insufficient for privacy protection. We need to make $\algomap(\dataset)$ sufficiently non-invertible. 
					\\ 
		See also: \gls{ml}, \gls{dataset}, \gls{modelparam}, \gls{prediction}, \gls{datapoint}, \gls{feature}, \gls{label}, \gls{sensattr}, \gls{map}.}, 
	first={privacy protection}, 
	text={privacy protection} 
}

\newglossaryentry{privleakage}
{name={privacy leakage},
	description={Consider\index{privacy leakage} an \gls{ml} application that processes a 
		\gls{dataset} $\dataset$ and delivers some \gls{output}, such as the \glspl{prediction} 
		obtained for new \glspl{datapoint}. Privacy leakage arises 
		if the \gls{output} carries information about a private (or sensitive) \gls{feature} of 
		a \gls{datapoint} of $\dataset$ (such as a human). Based on a \gls{probmodel} 
		for the \gls{data} generation, we can measure the privacy leakage via the \gls{mutualinformation} 
		between the \gls{output} and the sensitive \gls{feature}. Another quantitative \gls{measure} of privacy leakage 
		is \gls{diffpriv}. The relations between different \glspl{measure} of privacy leakage have been 
		studied in the literature (see \cite{InfThDiffPriv}). 
				\\ 
		See also: \gls{mutualinformation}, \gls{diffpriv}, \gls{privattack}, \gls{gdpr}. }, 
	first={privacy leakage}, 
	text={privacy leakage} 
}

\newglossaryentry{crossentropy}
{name={cross-entropy},
	description={Consider a multi-class \gls{classification} problem with a 
		\gls{featurespace} $\featurespace$ and a finite \gls{labelspace} 
              	$\labelspace = \{1,\,\ldots,\,\nrcluster\}$.  A \gls{datapoint} with a 
              	\gls{featurevec} $\featurevec$ is represented as a \gls{pmf} 
              	$\labelvec = (\truelabel_{1},\,\ldots,\,\truelabel_{\nrcluster})^{T}$ 
              	over~$\labelspace$, where $\truelabel_{\clusteridx}$ denotes the 
              	\gls{probability} that a randomly chosen \gls{datapoint} with 
		\gls{featurevec} $\featurevec$ has \gls{label} $\clusteridx$.  
		A \gls{hypothesis} $\hypothesis(\featurevec)$ outputs a predicted 
              	\gls{pmf} $\hat{\labelvec} = (\predictedlabel_{1},\,\ldots,
              	\,\predictedlabel_{\nrcluster})^{T}$. The associated 
              	\index{cross-entropy}cross-\gls{entropy} \gls{loss} is \cite{coverthomas} 
             	\begin{equation}
                		\nonumber
                		\lossfunc{(\featurevec,\hat{\labelvec})}{\hypothesis}
                		\defeq - \sum_{\clusteridx=1}^{\nrcluster} 
                		\truelabel_{\clusteridx}\,\log \predictedlabel_{\clusteridx}.
              	\end{equation}
		The cross-\gls{entropy} \gls{loss} quantifies the dissimilarity between the true 
		\gls{pmf} $\labelvec$ and the predicted \gls{pmf} $\hat{\labelvec}$. It is 
		also a \gls{measure} of the expected number of bits required to encode \glspl{label} 
		drawn from the true \gls{pmf} $\labelvec$ when using a coding scheme optimized for 
		the predicted \gls{pmf} $\hat{\labelvec}$ \cite{coverthomas}. \\
		Note that, for binary \gls{classification} (with $\nrcluster = 2$), the 
              	cross-\gls{entropy} \gls{loss} reduces to the \gls{logloss} when 
              	employing a \gls{parammodel} with \glspl{modelparam} $\weights$ 
              	such that $\predictedlabel_{2}/\predictedlabel_{1}
              	= \exp\,(\weights^{T}\featurevec)$. Moreover, the representation 
              	\eqref{equ_log_loss_gls_dict} of \gls{logloss} requires encoding 
              	the \gls{labelspace} $\{1,2\}$ using the values $-1$ and $1$. 
			  \\
		See also: \gls{classification}, \gls{pmf}, \gls{logloss}. }, 
 	first={cross-entropy},
 	text={cross-entropy}
}

\newglossaryentry{bce}
{name={binary cross-entropy (BCE)},
	description={The\index{binary cross-entropy (BCE)} BCE \gls{loss} is a 
   		special case of \gls{crossentropy} \gls{loss} for a binary \gls{classification} problem.
		\\
		See also: \gls{crossentropy}, \gls{classification}. }, 
 	first={binary cross-entropy (BCE)},
 	text={BCE}
}

\newglossaryentry{softlabel}
{name={soft label},
	description={Consider\index{soft label} a \gls{classification} problem where \glspl{datapoint} are characterized 
 		by \glspl{feature} $\featurevec$ and \glspl{label} from a finite \gls{labelspace} 
		$\labelspace = \{1,\,\ldots,\,\nrcluster\}$. Some \gls{ml} applications involve \glspl{datapoint} 
		that have almost identical \glspl{feature} but different \glspl{label}. In such 
		cases, instead of assigning to each \gls{datapoint} a single \gls{label} $\truelabel \in \labelspace$, 
		it can be more useful to assign an entire \gls{probdist} over the \gls{labelspace}. 
		This \gls{probdist} can be represented as a \gls{pmf} 
		$\labelvec = \big( \truelabel_{1},\,\ldots,\,\truelabel_{\nrcluster}\big)^{T} \in \simplex{\nrcluster}$. 
		We can view the entries $\truelabel_{\clusteridx}$, for $\clusteridx=1,\,\ldots,\,\nrcluster$, 
		as soft \glspl{label} of a \gls{datapoint}. Mathematically, the soft \gls{label} $\truelabel_{\clusteridx}$  
		is the \gls{probability} that a randomly chosen \gls{datapoint} with \gls{featurevec} $\featurevec$ 
		has \gls{label} $\clusteridx$.
		\\
              	See also: \gls{classification}, \gls{pmf}, \gls{crossentropy}.},
 	first={soft label},
 	plural={soft labels},
 	firstplural={soft labels},
 	text={soft label}
}

\newglossaryentry{nn}
{name={nearest neighbor (NN)},
	description={NN\index{nearest neighbor (NN)} methods learn a \gls{hypothesis} 
		$\hypothesis: \featurespace \rightarrow \labelspace$ whose \gls{function} value $\hypothesis(\featurevec)$ 
		is solely determined by the NNs within a given \gls{dataset}. Different 
		methods use different \glspl{metric} for determining the NNs. If \glspl{datapoint} 
		are characterized by numeric \glspl{featurevec}, we can use their \glspl{eucliddist} as 
		the \gls{metric}.
					\\ 
		See also: \gls{metric}, \gls{neighbor}.},
	first={nearest neighbor (NN)},
	plural={NNs},
 	firstplural={nearest neighbors (NNs)},
	text={NN} 
}

\newglossaryentry{neighbor}
{name={neighbor},
	description={A\index{neighbor} neighbor of a node $\nodeidx \in \nodes$ 
		within an \gls{undirectedgraph} is a node $\nodeidx' \in \nodes \setminus \{ \nodeidx\}$ 
		that is \gls{connected} via an edge to node $\nodeidx$.
				\\ 
		See also: \gls{undirectedgraph}, \gls{connected}.},
	first={neighbor},
	firstplural={neighbors}, 
	plural={neighbors}, 
	text={neighbor} 
}

\newglossaryentry{bias}
{name={bias},
	description={Consider\index{bias} an \gls{erm}-based \gls{ml} method that  
		learns a \gls{hypothesis} $\learnthypothesis \in \hypospace$ 
		from a given \gls{trainset}. The analysis of the \gls{ml} method 
		is often based on a \gls{probmodel} (such as the \gls{iidasspt}) 
		for the \gls{data} generation. Here, \glspl{datapoint} and, in turn, 
		the learned \gls{hypothesis} $\learnthypothesis$ are viewed as 
		(\glspl{realization} of) \glspl{rv}. Any property 
		$\theta\big(\learnthypothesis\big)$ of $\learnthypothesis$, 
		such as specific \glspl{modelparam} in a \gls{parammodel} 
		or the \gls{prediction} error $\truelabel - \learnthypothesis(\featurevec)$ 
		for a fixed \gls{datapoint}, then also becomes an \gls{rv}. 
		The squared bias of a numeric property $\theta\big(\learnthypothesis\big) \in \mathbb{R}^{r}$ 
		is \cite{BishopBook}, \cite{hastie01statisticallearning}
		$$\biasterm^{2} \defeq \big\| \expect \big\{ \theta\big(\learnthypothesis\big) \big\}\!-\!\theta\big(\bar{\hypothesis}\big)  \big\|_{2}^{2}.$$
		Here, $\bar{\hypothesis}$ is a reference \gls{hypothesis}, which 
		could be defined by $\bar{\hypothesis}(\featurevec)=\truelabel$ 
		for a fixed test \gls{datapoint} with \gls{featurevec} $\featurevec$ 
		and \gls{label} $\truelabel$. 
					\\ 
		See also: \gls{probmodel}, \gls{iid}, \gls{rv}, \gls{esterr}.},
	first={bias},
	text={bias},
	user1={biased}
}

\newglossaryentry{classification}
{name={classification},
	description={Classification\index{classification} is the task of determining a 
 		discrete-valued \gls{label} $\truelabel$ for a given \gls{datapoint}, based solely on its 
 		\glspl{feature} $\featurevec$. The \gls{label} $\truelabel$ belongs to a finite set, such as 
 		$\truelabel \in \{-1,1\}$ or $\truelabel \in \{1, \,\ldots, \,19\}$, and represents the 
 		category to which the corresponding \gls{datapoint} belongs.
				\\ 
		See also: \gls{label}, \gls{datapoint}, \gls{feature}.}, 
	first={classification},
	text={classification} 
}

\newglossaryentry{privfunnel}
{name={privacy funnel},
	description={The\index{privacy funnel} privacy funnel is a method for learning 
		a \gls{featuremap} that provides privacy-friendly \glspl{feature} for a 
		\gls{datapoint} \cite{PrivacyFunnel}.
				\\ 
		See also: \gls{feature}, \gls{datapoint}, \gls{gdpr}, \gls{diffpriv}.},
 	first={privacy funnel},
	text={privacy funnel} 
}

\newglossaryentry{classifier}
{name={classifier},
	description={A classifier\index{classifier} is a \gls{hypothesis} (i.e., a \gls{map}) $\hypothesis(\featurevec)$ 
		used to predict a \gls{label} taking on values from a finite \gls{labelspace}. We might use the 
		\gls{function} value $\hypothesis(\featurevec)$ itself as a \gls{prediction} $\predictedlabel$ for 
		the \gls{label}. However, it is customary to use a \gls{map} $\hypothesis(\cdot)$ that delivers 
		a numeric quantity. The \gls{prediction} is then obtained by a simple thresholding step. 
		For example, in a binary \gls{classification} problem with a \gls{labelspace} $\labelspace \in  \{ -1,1\}$, 
		we might use a real-valued \gls{hypothesis} \gls{map} $\hypothesis(\featurevec) \in \mathbb{R}$ 
		as a classifier. A \gls{prediction} $\predictedlabel$ can then be obtained via thresholding:
		 \begin{equation} 
		 	\label{equ_def_threshold_bin_classifier_dict}
		 	\predictedlabel =1 \quad  \mbox{for } \hypothesis(\featurevec)\!\geq\!0 \quad \mbox{and} \quad	\predictedlabel =-1  \quad \mbox{otherwise.}
	 	\end{equation}
 		We can characterize a classifier by its \glspl{decisionregion} $\decreg{a}$ for 
 		every possible \gls{label} value $a \in \labelspace$.
					\\ 
		See also: \gls{hypothesis}, \gls{classification}, \gls{decisionregion}. },
	first={classifier},
	text={classifier} 
}

\newglossaryentry{emprisk}
{name={empirical risk},
	description={The empirical \gls{risk}\index{empirical risk} $\emprisk{\hypothesis}{\dataset}$ 
  		of a \gls{hypothesis} on a \gls{dataset} $\dataset$ is the average \gls{loss} incurred 
  		by $\hypothesis$ when applied to the \glspl{datapoint} in $\dataset$.
				\\ 
		See also: \gls{risk}, \gls{hypothesis}, \gls{dataset}, \gls{loss}, \gls{datapoint}.},
  	first={empirical risk},
  	text={empirical risk} 
}

\newglossaryentry{token}
{name={token},
	description={A token\index{token} is a basic unit of information 
  		obtained by splitting a \gls{sequence} of symbols, such as a text string, 
  		into smaller parts. In \gls{nlp}, tokens often correspond to words, subwords, 
  		or characters that form the \glspl{feature} of a \gls{datapoint}. 
  		Tokenization transforms raw text (e.g., ``The cat sleeps'') into a 
  		\gls{sequence} of tokens (e.g., [``The'', ``cat'', ``sleeps'']), which can
  		then be mapped to numerical \glspl{featurevec}.
		\\
		See also: \gls{sequence}, \gls{featurevec}. }, 
  	first={token}, 
  	firstplural={tokens}, 
  	plural={tokens}, 
  	text={token}
}

\newglossaryentry{nlp}
{name={natural language processing (NLP)},
	description={NLP\index{natural language processing (NLP)} 
 		studies \gls{ml} methods for the the analysis and generation of human language. 
  		Typical NLP tasks include text \gls{classification}, machine translation, 
  		sentiment analysis, and question answering. Modern NLP systems represent 
  		language as \glspl{sequence} of \glspl{token} and train \glspl{model} that 
  		capture contextual dependencies, such as \gls{attention}-based methods.
		\\ 
  		See also: \gls{token}, \gls{attention}.}, 
	first={natural language processing (NLP)}, 
	text={NLP}
}

\newglossaryentry{riskstratification}
{name={risk stratification},
	description={\Gls{risk} \gls{stratification}\index{risk stratification} assigns 
		\glspl{datapoint} to \gls{risk} groups (or \glspl{stratum}) by 
		quantizing the \gls{prediction} $\learnthypothesis(\featurevec)$ 
		obtained from a trained \gls{risk} prognosis \gls{model}. Typical choices for these 
		\glspl{stratum} include low, intermediate, and high \gls{risk} 
             	\cite{Collins:2012aa}, \cite{Steyerberg2019}.
		\\
		See also: \gls{stratification}, \gls{prediction}, \gls{clustering}.},
	first={risk stratification},
	text={risk stratification}
}

\newglossaryentry{uncertainty}
{name={uncertainty},
	description={In the context of \gls{ml}, uncertainty\index{uncertainty} refers to the presence of multiple 
		plausible \glspl{outcome} or \glspl{explanation} based on available \gls{data}. For example, the 
		\gls{prediction} $\hat{\hypothesis}(\featurevec)$ produced by a trained \gls{ml} \gls{model} $\hat{\hypothesis}$
	 	often reflects a range of possible values for the true \gls{label} of a given \gls{datapoint}. 
	 	The broader this range, the greater the associated uncertainty. \Gls{probability} theory 
	 	allows us to represent, quantify, and reason about uncertainty in a 
	 	mathematically rigorous manner.
					\\ 
		See also: \gls{probmodel}, \gls{risk}, \gls{entropy}, \gls{variance}. },
	first={uncertainty},
	text={uncertainty}
}

\newglossaryentry{ucb}
{name={upper confidence bound (UCB)},
	description={Consider\index{upper confidence bound (UCB)} an \gls{ml} 
		application that requires selecting, at each time step $\iteridx$, an \gls{action} $\arm_{\iteridx}$ 
		from a finite set of alternatives $\actionset$. The utility of selecting \gls{action} $\arm_{\iteridx}$ 
		is quantified by a numeric \gls{reward} signal $\reward^{(\arm_{\iteridx})}$. 
		A widely used \gls{probmodel} for this type of sequential decision-making problem 
		is the \gls{stochmab} setting \cite{Bubeck2012}. In this \gls{model}, 
		the \gls{reward} $\reward^{(\arm)}$ is viewed as the \gls{realization} of an \gls{rv} 
		with unknown \gls{mean} $\mu^{(\arm)}$. Ideally, we would always choose the 
		\gls{action} with the largest expected \gls{reward} $\mu^{(\arm)}$, but these 
		\glspl{mean} are unknown and must be estimated from observed \gls{data}. Simply 
		choosing the \gls{action} with the largest estimate $\widehat{\mu}^{(\arm)}$ can 
		lead to suboptimal \glspl{outcome} due to estimation \gls{uncertainty}. The UCB strategy 
		addresses this by selecting \glspl{action} not only based on their estimated \glspl{mean} but 
		also by incorporating a term that reflects the \gls{uncertainty} in these estimates—favoring 
		\glspl{action} with a high-potential \gls{reward} and high \gls{uncertainty}. Theoretical guarantees 
		for the performance of UCB strategies, including logarithmic \gls{regret} bounds, are established in \cite{Bubeck2012}.
					\\ 
		See also: \gls{action}, \gls{reward}, \gls{stochmab}, \gls{uncertainty}, \gls{regret}, \gls{optimism in the face of uncertainty}.},
	first={upper confidence bound (UCB)},
	text={UCB} 
}

\newglossaryentry{optimism in the face of uncertainty}
{name={optimism in the face of uncertainty},
	description={\gls{ml}\index{optimism in the face of uncertainty} methods learn \glspl{modelparam} $\weights$ 
		according to some performance criterion $\bar{f}(\weights)$. However, they usually 
		cannot access $\bar{f}(\weights)$ directly but rely on an estimate (or approximation) 
		$f(\weights)$ of $\bar{f}(\weights)$. As a case in point, \gls{erm}-based methods use 
		the average \gls{loss} on a given \gls{dataset} (i.e., the \gls{trainset}) as an estimate 
		for the \gls{risk} of a \gls{hypothesis}. Using a \gls{probmodel}, one can construct 
		a confidence interval $\big[ l^{(\weights)},  u^{(\weights)} \big]$ for each choice $\weights$ for the \glspl{modelparam}.
		One simple construction is $l^{(\weights)} \defeq f(\weights) - \sigma/2$, $u^{(\weights)} \defeq f(\weights)+ \sigma/2$, 
	    	with $\sigma$ being a \gls{measure} of the (expected) deviation of $f(\weights)$ from $\bar{f}(\weights)$.
		We can also use other constructions for this interval as long as they ensure that $\bar{f}(\weights) \in\big[ l^{(\weights)},  u^{(\weights)} \big]$ 
		with a sufficiently high \gls{probability}. An optimist chooses the \glspl{modelparam} 
		according to the most favorable—yet still plausible—value $\tilde{f}(\weights) \defeq  l^{(\weights)}$ 
		of the performance criterion (see Fig.\ \ref{fig_optimism_dict}). Two examples of \gls{ml} methods that use such an optimistic 
		construction of an \gls{objfunc} are \gls{srm} \cite[Ch. 11]{ShalevMLBook} and \gls{ucb} methods 
		for sequential decision making \cite[Sec. 2.2]{Bubeck2012}. 
		\begin{figure}[H]
			\begin{center}
			\begin{tikzpicture}[x=3cm, y=1cm]
			\fill[blue!10] 
			(-1, 5) -- plot[domain=-2:1, samples=100] ({\x+1}, {\x*\x + 1}) -- 
			plot[domain=1:-2, samples=100] ({\x+1}, {\x*\x - 0.5}) -- cycle;
  			\node[anchor=west] at (2, 4) {$f(\weights)$};
  			\draw[line width=1, domain=-2:1, samples=100,dashed] plot  ({\x+1}, {\x*\x -0.5}) node[right] {$\tilde{f}(\weights)$};
   			\draw[line width=1, domain=-1:2, samples=100] plot ({\x}, {\x*\x});
  			\draw[<->, thick] (1, -0.5) -- (1, 1) node[midway, right] {$\big[ l^{(\weights)}\!,\!u^{(\weights)} \big]$};
			\end{tikzpicture}
		\caption{\gls{ml} methods learn \glspl{modelparam} $\weights$ by using some estimate of $f(\weights)$ for 
			the ultimate performance criterion $\bar{f}(\weights)$. Using a \gls{probmodel}, one can use $f(\weights)$ to 
			construct confidence intervals $\big[ l^{(\weights)},  u^{(\weights)} \big]$, which contain $\bar{f}(\weights)$  
			with high \gls{probability}. The best plausible performance \gls{measure} for a specific choice $\weights$ of \glspl{modelparam} 
			is $\tilde{f}(\weights) \defeq l^{(\weights)}$. \label{fig_optimism_dict}} 
			\end{center}
		\end{figure}
		See also: \gls{objfunc}, \gls{ucb}, \gls{optmethod}, \gls{gdmethod}.},
	first={optimism in the face of uncertainty},
	text={optimism in the face of uncertainty} 
}

\newglossaryentry{flnetwork}
{name={federated learning network (FL network)},
	description={An \gls{fl} network\index{federated learning network (FL network)} is a 
		mathematical model for a \gls{flsystem} that consists of interconnected 
		\glspl{device}. These \glspl{device} are represented by the nodes $\nodes$ 
		of an undirected weighted \gls{graph}
		\[
			\graph = (\nodes,\edges,\mA).
		\]
		An edge $\{\nodeidx,\nodeidx'\} \in \edges$ indicates collaborations 
		between two \glspl{device} $\nodeidx, \nodeidx' \in \nodes$. 
		The \glspl{edgeweight} $\edgeweight_{\nodeidx,\nodeidx'} > 0$ 
		quantify the extent of collaborations, which may be related to a 
		communication link capacity, statistical similarity between 
		\glspl{localdataset}, or both. Each \gls{device} $\nodeidx \in \nodes$ 
		has potentially access to a \gls{localdataset} $\localdataset{\nodeidx}$ and 
		might train a \gls{localmodel} $\localmodel{\nodeidx}$, e.g., using 
		\gls{erm}-based methods. Many popular \gls{fl} methods are obtained by 
		coupling the \gls{training} of these \glspl{localmodel} across the edges of the 
		\gls{fl} network \cite{JungFLBook}. This coupling can be implemented in different ways, 
		e.g., using a \gls{penaltyterm} that enforces similarity between the \glspl{modelparam} 
		of neighboring \glspl{device}, or using the \glspl{prediction} of neighboring 
		\glspl{device} to augment the \glspl{localdataset}.
		Fig.\ \ref{fig_fl_network_dict} illustrates an \gls{fl} network with four \glspl{device}.
		\begin{figure}[H]
		\centering
		\begin{tikzpicture}[
			scale=1,
			node/.style={circle, draw=black, fill=black, inner sep=2pt},
			edge/.style={line width=0.9pt},
			wlab/.style={fill=white, inner sep=1pt, font=\scriptsize}
			]
			\node[node] (v1) at (0,0) {};
			\node[node] (v2) at (2,0) {};
			\node[node] (v3) at (1,1.5) {};
			\node[node] (v4) at (3,1.5) {};
			\node[anchor=north] at (v1) {$\nodeidx_1$};
			\node[anchor=north] at (v2) {$\nodeidx_2$};
			\node[anchor=south] at (v3) {$\nodeidx_3$};
			\node[anchor=south] at (v4) {$\nodeidx_4$};
			\draw[edge] (v1) -- node[wlab, pos=0.5, below, yshift=-2pt] {$\edgeweight_{1,2}$} (v2);
			\draw[edge] (v2) -- node[wlab, pos=0.55, left,  xshift=-2pt] {$\edgeweight_{2,3}$} (v3);
			\draw[edge] (v3) -- node[wlab, pos=0.45, left,  xshift=-2pt] {$\edgeweight_{3,1}$} (v1);
			\draw[edge] (v2) -- node[wlab, pos=0.55, right, xshift=2pt] {$\edgeweight_{2,4}$} (v4);
			\end{tikzpicture}
		\caption{\Gls{fl} network with nodes $\nodes=\{\nodeidx_1,\nodeidx_2,\nodeidx_3,\nodeidx_4\}$ 
			representing four \glspl{device} of an \gls{flsystem}. \label{fig_fl_network_dict}}
		\end{figure}
		The \gls{fl} network specifies which \glspl{device} can interact and to what extent. 
	    	\\
		See also: \gls{fl}, \gls{device}, \gls{graph}, \gls{gtvmin}.},
	first={federated learning network (FL network)},
	text={FL network},
	plural={FL networks}, 
	firstplural={federated learning networks (FL networks)}
}

\newglossaryentry{explanation}
{name={explanation},
	description={One approach to enhance the \gls{transparency} of an \gls{ml} method for its human user 
		is to provide an explanation\index{explanation} alongside the \glspl{prediction} delivered 
		by the method. Explanations can take different forms. For instance, they may 
        		consist of human-readable text or quantitative indicators, such as \gls{feature} 
        		importance scores for the individual \glspl{feature} of a given \gls{datapoint}~\cite{Molnar2019}. 
	   	Alternatively, explanations can be visual—for example, intensity \glspl{map} that highlight 
	   	image regions that drive the \gls{prediction} \cite{GradCamPaper}. 
       		Fig.\ \ref{fig_explanation_dict} illustrates two types of explanations. The first 
       		is a local linear approximation $g(\featurevec)$ of a nonlinear trained \gls{model} 
       		$\learnthypothesis(\featurevec)$ around a specific \gls{featurevec} $\featurevec'$, 
       		as used in the method \gls{lime}. 
        		The second form of explanation depicted in the figure is a sparse set of \glspl{prediction} 
       		$\learnthypothesis(\featurevec^{(1)}), \learnthypothesis(\featurevec^{(2)}), \learnthypothesis(\featurevec^{(3)})$ 
       		at selected \glspl{featurevec}, offering concrete reference points for the user. 
	 	\begin{figure}[H]
	   		\begin{center}
	 		\begin{tikzpicture}[x=0.5cm]
	 		\begin{axis}[
	 			hide axis,
	 			xmin=-3, xmax=6,
	 			ymin=0, ymax=6,
	 			domain=0:6,
	 			samples=100,
	 			width=10cm,
	 			height=6cm,
	 			clip=false
	 		   ]
			\addplot[thick, domain=-2:6] {2 + sin(deg(x))} 
	   		node[pos=0.9, above right, yshift=10pt] {$\learnthypothesis(\featurevec)$};
	 		\addplot[blue, thick, domain=0.5:2.5] 
	 		{2 + sin(deg(1.5)) + cos(deg(1.5))*(x - 1.5)}
	 		node[pos=0.2, above] {$g(\featurevec)$};
	 		\addplot[mark=*] coordinates {(1.5, {2 + sin(deg(1.5))})};
			\addplot[dashed, gray] coordinates {(1.5,0) (1.5,2.4)};
	 		\node at (axis cs:1.5, -0.2) {$\featurevec'$};
	       		\addplot[mark=*,blue] coordinates {(-1, {2 + sin(deg(-1))})};
	       		\addplot[dashed, gray] coordinates {(-1,0) (-1,{2 + sin(deg(-1))})};
	       		\node at (axis cs:-1, -0.2) {$\featurevec^{(1)}$};
	 	   	\addplot[mark=*,blue] coordinates {(0, {2 + sin(deg(0))})};
	 	   	\addplot[dashed, gray] coordinates {(0,0) (0,{2 + sin(deg(0))})};
	 	   	\node at (axis cs:0, -0.2) {$\featurevec^{(2)}$};
	 	   	\addplot[mark=*,blue] coordinates {(5, {2 + sin(deg(5))})};
	 	   	\addplot[dashed, gray] coordinates {(5,0) (5,{2 + sin(deg(5))})};
	 	   	\node at (axis cs:5, -0.2) {$\featurevec^{(3)}$};
	 		\end{axis}
			\end{tikzpicture}
	 	\end{center}
	 	\caption{Trained \gls{model} $\learnthypothesis(\featurevec)$ explained 
	     		locally at some point $\featurevec'$ by a linear approximation $g(\featurevec)$. 
	     		For a \gls{differentiable} $\learnthypothesis(\featurevec)$, this approximation is 
	     		determined by the \gls{gradient} $\nabla \learnthypothesis(\featurevec')$. Another 
	     		form of explanation could be the \gls{function} values $\learnthypothesis\big(\featurevec^{(\sampleidx)} \big)$ 
	     		for $\sampleidx=1, \,2, \,3$. 
			\label{fig_explanation_dict}}
	 	\end{figure} 
		See also: \gls{ml}, \gls{prediction}, \gls{feature}, \gls{datapoint}, \gls{classification}.},
	first={explanation},
	plural={explanations},
	text={explanation} 
}

\newglossaryentry{risk}
{name={risk},
	description={Consider\index{risk} a \gls{hypothesis} $\hypothesis$ used to predict the \gls{label} 
		$\truelabel$ of a \gls{datapoint} based on its \glspl{feature} $\featurevec$. We measure 
		the quality of a particular \gls{prediction} using a \gls{lossfunc} $\lossfunc{(\featurevec,\truelabel)}{\hypothesis}$. 
		If we interpret \glspl{datapoint} as the \glspl{realization} of \gls{iid} \glspl{rv}, 
		the $\lossfunc{(\featurevec,\truelabel)}{\hypothesis}$ also becomes the \gls{realization} 
		of an \gls{rv}. The \gls{iidasspt} allows us to define the risk of a \gls{hypothesis} 
		as the expected \gls{loss} $\expect \big\{\lossfunc{(\featurevec,\truelabel)}{\hypothesis} \big\}$. 
		Note that the risk of $\hypothesis$ depends on both the specific choice for the \gls{lossfunc} and the 
		\gls{probdist} of the \glspl{datapoint}.
					\\ 
		See also: \gls{iid} \gls{rv}, \gls{iidasspt}, \gls{loss}, \gls{probdist}.},
	first={risk},
	text={risk} 
}

\newglossaryentry{actfun}
{name={activation function},
	description={Each\index{activation function} \gls{artificialneuron} within an \gls{ann} is 
		assigned an \gls{activation} \gls{function} $\actfun(\cdot)$ that maps a weighted 
		combination of the \gls{neuron} inputs $\feature_{1}, \,\ldots, \,\feature_{\nrfeatures}$ 
		to a single \gls{output} value $a = \actfun\big(\weight_{1} \feature_{1}+\ldots+\weight_{\nrfeatures} \feature_{\nrfeatures} \big)$. 
		Note that each \gls{neuron} is parameterized by the \glspl{weight} $\weight_{1}, \,\ldots, \,\weight_{\nrfeatures}$.
					\\ 
		See also: \gls{ann}, \gls{activation}, \gls{relu}.},
	first={activation function},
	text={activation function} 
}

\newglossaryentry{distributedalgorithm}
{name={distributed algorithm},
	description={A\index{distributed algorithm} distributed \gls{algorithm} is an \gls{algorithm} designed for 
		a special type of computer, i.e., a collection of interconnected computing \glspl{device}.
		These \glspl{device} communicate and coordinate their local computations by exchanging
		messages over a network \cite{ParallelDistrBook}, \cite{IntroDistAlg}. Unlike a classical \gls{algorithm},
		which is implemented on a single \gls{device}, a distributed \gls{algorithm} is
		executed concurrently on multiple \glspl{device} with computational capabilities.
		Similar to a classical \gls{algorithm}, a distributed \gls{algorithm} can be modeled as a
		set of potential executions. However, each execution in the distributed setting involves
		both local computations and message-passing \glspl{event}. A generic execution might look as
		follows:
		\[
		\begin{array}{l}
			\text{\Gls{device} } 1\text{: } {\rm input}^{(1)}, \,s_1^{(1)}, \,s_2^{(1)}, \,\ldots, \,s_{T_1}^{(1)}, \,{\rm \gls{output}}^{(1)}; \\
			\text{\Gls{device} } 2\text{: } {\rm input}^{(2)}, \,s_1^{(2)}, \,s_2^{(2)}, \,\ldots, \,s_{T_2}^{(2)}, \,{\rm \gls{output}}^{(2)}; \\
			\quad \vdots \\
			\text{\Gls{device} } \nrnodes\text{: } {\rm input}^{(\nrnodes)}, \,s_1^{(\nrnodes)}, \,s_2^{(\nrnodes)}, \,\ldots, \,s_{T_{\nrnodes}}^{(\nrnodes)}, \,{\rm \gls{output}}^{(\nrnodes)}.
		\end{array}
		\]
		Each \gls{device} $\nodeidx \in \{1,\,\ldots,\,\nrnodes\}$ starts from its own local input and performs a \gls{sequence} of 
		intermediate computations $s_{\iteridx}^{(\nodeidx)}$ at discrete-time instants $\iteridx = 1, \,\dots, \,T_{\nodeidx}$. 
		These computations may depend on both the previous local computations at the \gls{device} 
		and the messages received from other \glspl{device}. One important application of distributed 
		\glspl{algorithm} is in \gls{fl} where a network of \glspl{device} collaboratively trains a personal \gls{model} 
		for each \gls{device}. 
					\\ 
		See also: \gls{algorithm}, \gls{device}, \gls{event}, \gls{fl}, \gls{model}.},
	first={distributed algorithm}, 
	text={distributed algorithm}
}

\newglossaryentry{algorithm}
{name={algorithm},
 	description={An\index{algorithm} algorithm is a precise, step-by-step specification for 
  		producing an \gls{output} from a given input within a finite number of well-defined 
		computational steps \cite{Cormen:2022aa}. For example, a \gls{gdmethod} for \gls{linreg} 
		is an algorithm that explicitly describes how to map a given \gls{trainset} 
		into \glspl{modelparam} through a sequence of \glspl{gradstep}. The precise form of 
		an algorithm depends on the available computational infrastructure. For example, if 
		this infrastructure allows us to compute an \gls{inverse}, then we can 
		define a \gls{linreg} algorithm using the \gls{normalequations}. In contrast, if 
		the computational infrastructure only allows basic arithmetic operations (i.e., multiplication and addition), 
		the \gls{normalequations} need to be somehow translated into a sequence of arithmetic 
		operations (e.g., as in \glspl{gdmethod}). 
		To study algorithms rigorously, we can represent (or approximate) them by different 
		mathematical structures \cite{Sipser2013}. One approach is to represent an algorithm 
		as a collection of possible executions. Each individual execution is then a 
		\gls{sequence} of the following form: $${\rm input}, \,s_1, \,s_2, \,\ldots, \,s_T, \,{\rm \gls{output}}.$$ 
		This sequence starts from an input and progresses via intermediate steps until an 
		\gls{output} is delivered. Crucially, an algorithm encompasses more than just a mapping 
		from input to \gls{output}; it also includes intermediate computational 
     		steps $s_1, \,\ldots, \,s_T$.
				\\ 
		See also: \gls{output}, \gls{linreg}, \gls{trainset}, \gls{modelparam}, \gls{gradstep}, \gls{model}, \gls{stochastic}.},
	first={algorithm},
	plural={algorithms},
	text={algorithm} 
}

\newglossaryentry{stochalgorithm}
{name={stochastic algorithm},
	description={A\index{stochastic algorithm} \gls{stochastic} \gls{algorithm} uses a random mechanism 
		during its execution. For example, \gls{stochGD} uses a randomly selected subset of \glspl{datapoint} 
		to compute an approximation for the \gls{gradient} of an \gls{objfunc}. We can represent a 
		\gls{stochastic} \gls{algorithm} by a \glspl{stochproc}, i.e., the possible execution \gls{sequence} is the possible \glspl{outcome} of 
		a \gls{randomexperiment} \cite{BertsekasProb}, \cite{RandomizedAlgos}, \cite{Gallager13}.		
		\\ 
		See also: \gls{stochastic}, \gls{algorithm}, \gls{stochGD}, \gls{datapoint}, \gls{gradient}, \gls{objfunc}, \gls{stochproc}, 
		\gls{randomexperiment}, \gls{optmethod}, \gls{gdmethod}. },
	first={stochastic algorithm},
	text={stochastic algorithm},
	plural={stochastic algorithms},
	firstplural={stochastic algorithms}
}

\newglossaryentry{batchlearning}
{name={batch learning},
	description={In \gls{batch} learning\index{batch learning} (also known as offline learning), the \gls{ml} \gls{model} 
		is trained on the entire \gls{dataset} in a single \gls{training} iteration, instead of updating it incrementally as \gls{data} arrive. 
		All available \gls{data} are inputted into a learning \gls{algorithm}, resulting in a \gls{model} that can make \glspl{prediction}. 
		Since these \glspl{dataset} tend to be large, \gls{training} is computationally expensive and time-consuming, 
		so it is typically performed offline. After learning, the \gls{model} will be static and will not adapt to new \gls{data} automatically. 
		Updating the \gls{model} with new information requires retraining the \gls{model} entirely. Once the \gls{model} has been trained, 
		it is launched into production where it cannot be updated. \Gls{training} a \gls{model} can take many hours, so many \glspl{model} in production 
		settings are updated cyclically on a periodic schedule when the \gls{data} distribution is stable. For example, a retail analytics team 
		could retrain their demand forecast \gls{model} every Sunday using the previous week's sales \gls{data} to predict next week's demand. 
		If a system needs to be constantly updated to rapidly changing \gls{data}, such as in stock price \gls{prediction}, a more adaptable solution 
		such as \gls{onlinelearning} is necessary.
		\\
		See also: \gls{batch}, \gls{model}, \gls{dataset}, \gls{onlinelearning}. },
	first={batch learning}, 
	text={batch learning}
}

\newglossaryentry{onlinelearning}
{name={online learning},
	description={Some \gls{ml} methods\index{online learning} are designed to 
	    	process \gls{data} in a sequential manner, updating their \glspl{modelparam} 
		one at a time, as new \glspl{datapoint} become available. A typical example 
		is time-series \gls{data}, such as daily \gls{minimum} and \gls{maximum} 
		temperatures recorded by an \gls{fmi} weather station. These values form 
		a chronological sequence of observations. During each time step $\timeidx$, 
		online learning methods update (or refine) the current \gls{hypothesis} 
		$\hypothesis^{(\timeidx)}$ (or \glspl{modelparam} $\weights^{(\timeidx)}$) 
		based on the newly observed \gls{datapoint} $\datapoint^{(\timeidx)}$. 
		\\ 
		See also: \gls{onlineGD}, \gls{onlinealgorithm}. },
	first={online learning},
	text={online learning} 
}

\newglossaryentry{onlinealgorithm}
{name={online algorithm},
	description={An\index{online algorithm} online \gls{algorithm} processes input \gls{data} incrementally, 
		receiving \glspl{datapoint} sequentially and making decisions or producing \glspl{output} (or decisions) immediately 
		without having access to the entire input in advance \cite{PredictionLearningGames}, \cite{HazanOCO}. 
		Unlike an offline \gls{algorithm}, which has the entire input available from the start, an online \gls{algorithm} 
		must handle \gls{uncertainty} about future inputs and cannot revise past decisions. Similar to an 
		offline \gls{algorithm}, we represent an online \gls{algorithm} formally as a collection of possible 
		executions. However, the execution \gls{sequence} for an online \gls{algorithm} has a distinct structure as follows:
		$${\rm in}_{1}, \,s_1, \,{\rm out}_{1}, \,{\rm in}_{2}, \,s_2, \,{\rm out}_{2}, \,\ldots, \,{\rm in}_{T}, \,s_T, \,{\rm out}_{T}.$$ 
		Each execution begins from an initial state (i.e., \(\text{in}_{1}\)) and proceeds through alternating 
		computational steps, \glspl{output} (or decisions), and inputs. Specifically, at step \(\iteridx\), 
		the \gls{algorithm} performs a computational step \(s_{\iteridx}\), generates an \gls{output} \(\text{out}_{\iteridx}\), 
		and then subsequently receives the next input (\gls{datapoint}) \(\text{in}_{\iteridx+1}\). A 
		notable example of an online \gls{algorithm} in \gls{ml} is \gls{onlineGD}, which incrementally 
		updates \glspl{modelparam} as new \glspl{datapoint} arrive. 
					\\ 
		See also: \gls{algorithm}, \gls{data}, \gls{datapoint}, \gls{uncertainty}, \gls{ml}, \gls{onlineGD}, \gls{modelparam}, \gls{onlinelearning}.},
	first={online algorithm},
	text={online algorithm} 
}

\newglossaryentry{sensattr}
{name={sensitive attribute},
	description={\gls{ml}\index{sensitive attribute} revolves around learning a \gls{hypothesis} \gls{map} that allows 
		us to predict the \gls{label} of a \gls{datapoint} from its \glspl{feature}. In some 
		applications, we must ensure that the \gls{output} delivered by an \gls{mlsystem} does 
		not allow us to infer sensitive attributes of a \gls{datapoint}. Which part 
		of a \gls{datapoint} is considered a sensitive attribute is a design 
		choice that varies across different application domains.
					\\ 
		See also: \gls{ml}, \gls{hypothesis}, \gls{map}, \gls{label}, \gls{datapoint}, \gls{feature}.},
	first={sensitive attribute},
	plural={sensitive attributes},
	firstplural={sensitive attributes},
	text={sensitive attribute} 
}

\newglossaryentry{sbm}
{name={stochastic block model (SBM)},
	description={The\index{stochastic block model (SBM)} SBM \cite{holland1983stochastic} is a 
		probabilistic generative \gls{model} for an \gls{undirectedgraph} $\graph = \big( \nodes, \edges \big)$ 
		with a given set of nodes $\nodes$ \cite{AbbeSBM2018}. In its most basic variant, 
		the SBM generates a \gls{graph} by first randomly assigning each node $\nodeidx \in \nodes$ to 
		a \gls{cluster} index $\clusteridx_{\nodeidx} \in \{1, \,\ldots, \,\nrcluster\}$. A pair of different nodes in the 
		\gls{graph} is \gls{connected} by an edge with \gls{probability} $p_{\nodeidx,\nodeidx'}$ that depends 
		solely on the \glspl{label} $\clusteridx_{\nodeidx}, \clusteridx_{\nodeidx'}$. 
		The presence of edges between different pairs of 
		nodes is statistically independent.
					\\ 
		See also: \gls{model}, \gls{graph}, \gls{cluster}, \gls{probability}, \gls{label}. },
	first={stochastic block model (SBM)},
	text={SBM} 
}

\newglossaryentry{deepnet}
{name={deep net},
	description={A\index{deep net} deep net is an \gls{ann} with a 
	    	(relatively) large number of hidden \glspl{layer}. \Gls{deeplearning} 
		is an umbrella term for \gls{ml} methods that use a deep 
		net as their \gls{model} \cite{Goodfellow-et-al-2016}.
				\\ 
		See also: \gls{ann}, \gls{layer}, \gls{deeplearning}, \gls{model}, \gls{llm}.},
	first={deep net},
	plural={deep nets},
	firstplural={deep nets},
	text={deep net} 
}

\newglossaryentry{baseline}
{name={baseline},
	description={Consider\index{baseline} some \gls{ml} method that produces a learned 
    		\gls{hypothesis} (or trained \gls{model}) $\learnthypothesis \in \hypospace$. We evaluate the quality of a trained \gls{model} 
    		by computing the average \gls{loss} on a \gls{testset}. But how can we assess 
    		whether the resulting \gls{testset} performance is sufficiently good? How can we 
    		determine if the trained \gls{model} performs close to optimal such that there is little point 
   		in investing more resources (for \gls{data} collection or computation) to improve it? 
    		To this end, it is useful to have a reference (or baseline) level against which 
    		we can compare the performance of the trained \gls{model}. \\
		Such a reference value might be obtained from human performance, e.g., the misclassification rate of dermatologists 
    		who diagnose cancer from visual inspection of skin \cite{SkinHumanAI}. Another source for a baseline is an existing, 
    		but for some reason unsuitable, \gls{ml} method. For example, the existing \gls{ml} method 
    		might be computationally too expensive for the intended \gls{ml} application. 
    		Nevertheless, its \gls{testset} error can still serve as a baseline. Another, somewhat more principled, 
    		approach to constructing a baseline is via a \gls{probmodel}. In many cases, given a \gls{probmodel} $p(\featurevec,\truelabel)$,  
    		we can precisely determine the \gls{minimum} achievable \gls{risk} among any \glspl{hypothesis}
    		(not even required to belong to the \gls{hypospace} $\hypospace$) \cite{LC}. \\
    		This \gls{minimum} achievable \gls{risk} (referred to as the \gls{bayesrisk}) is the \gls{risk} 
    		of the \gls{bayesestimator} for the \gls{label} $\truelabel$ of a \gls{datapoint}, given
    		its \glspl{feature} $\featurevec$. Note that, for a given choice of \gls{lossfunc}, the 
    		\gls{bayesestimator} (if it exists) is completely determined by the \gls{probdist} 
		$\probdist$ \cite[Ch. 4]{LC}. However, computing the \gls{bayesestimator} 
		and \gls{bayesrisk} presents two main challenges. First, the \gls{probdist} 
		$\probdist$ is unknown and must be estimated from observed \gls{data}. Second, even if $\probdist$ 
		were known, computing the \gls{bayesrisk} exactly may be computationally 
		infeasible \cite{cooper1990computational}. 
		A widely used \gls{probmodel} is the \gls{mvndist} $\pair{\featurevec}{\truelabel} \sim \mathcal{N}({\bm \mu},{\bm \Sigma})$ 
		for \glspl{datapoint} characterized by numeric \glspl{feature} and \glspl{label}.
		Here, for the \gls{sqerrloss}, the \gls{bayesestimator} is given by the \gls{posterior} 
		\gls{mean} $\mu_{\truelabel|\featurevec}$ of the \gls{label} $\truelabel$, given the 
		\glspl{feature} $\featurevec$ \cite{GrayProbBook}, \cite{LC}. The corresponding \gls{bayesrisk} 
		is given by the \gls{posterior} \gls{variance} 
		$\sigma^{2}_{\truelabel|\featurevec}$ (see Fig.\ \ref{fig_post_baseline_dict}).
		\begin{figure}[H]
		\begin{center}
		\begin{tikzpicture}
			\draw[->] (-1,0) -- (7,0) node[right] {$\truelabel$}; 
			\draw[thick,domain=-1:7,smooth,variable=\x] 
			plot ({\x}, {2*exp(-0.5*((\x-3)^2))});
			\draw[dashed] (3,0) -- (3,2.5);
			\node[anchor=south] at ([yshift=-5pt] 3,2.5) {\small $\mu_{\truelabel|\featurevec}$};
			\draw[<->,thick] (3-1,1) -- (3+1,1.0);
			\node[anchor=west] at ([yshift=2pt] 3,1.2) {\small $\sigma_{\truelabel|\featurevec}$};
  			\foreach \x in {0.5} {
			\node[red] at (\x, 0) {\bf \large $\times$};
 			}
  			\node[anchor=north] at (0.5,-0.2) {\small $\learnthypothesis(\featurevec)$};
		\end{tikzpicture}
		\end{center}
		\caption{If the \glspl{feature} and the \gls{label} of a \gls{datapoint} are drawn 
			from a \gls{mvndist}, we can achieve the \gls{minimum} \gls{risk} (under \gls{sqerrloss}) 
			by using the \gls{bayesestimator} $\mu_{\truelabel|\featurevec}$ 
			to predict the \gls{label} $\truelabel$ of a \gls{datapoint} with \glspl{feature} $\featurevec$. The corresponding 
			\gls{minimum} \gls{risk} is given by the \gls{posterior} \gls{variance} $\sigma^{2}_{\truelabel|\featurevec}$. We can use 
			this quantity as a baseline for the average \gls{loss} of a trained \gls{model} $\learnthypothesis$. \label{fig_post_baseline_dict}}
		\end{figure}
		See also: \gls{bayesrisk}, \gls{bayesestimator}.},
    first={baseline},
    text={baseline}
}

\newglossaryentry{kfoldcv}
{name={$k$-fold cross-validation ($k$-fold CV)},
sort={kfold CV},
	description={A $k$-fold CV is a method\index{$k$-fold cross-validation ($k$-fold CV)} for evaluating the 
 		\gls{gengap} of an \gls{erm}-based \gls{ml} method. The idea is to divide a \gls{dataset} 
 		$\dataset$ evenly into $k$ subsets (or folds) $\dataset^{(1)},\,\ldots,\,\dataset^{(k)}$.
		\begin{figure}[H]
		\centering
			\begin{tikzpicture}[font=\small]
				\def\k{5}          
				\def\cellw{1.6}    
				\def\cellh{0.40}   
				\def\gap{0.18}     
				\foreach \i in {1,...,\k} {
					\pgfmathsetmacro{\y}{-(\i-1)*(\cellh+\gap)}
					\node[anchor=east] at (-0.25,\y+0.5*\cellh) {fold \i};
					\foreach \j in {1,...,\k} {
						\pgfmathsetmacro{\x}{(\j-1)*\cellw}
						\ifnum\i=\j
						\fill[gray!35] (\x,\y) rectangle ++(\cellw,\cellh);
						\draw (\x,\y) rectangle ++(\cellw,\cellh);
						\else
						\draw (\x,\y) rectangle ++(\cellw,\cellh);
						\fi
					}
				}
				\foreach \j in {1,...,\k} {
					\pgfmathsetmacro{\x}{(\j-1)*\cellw + 0.5*\cellw}
					\node[anchor=south] at (\x,\cellh+0.2) {$\dataset^{(\j)}$};
				}
			\end{tikzpicture}
		\caption{In $k$-fold CV, the available \gls{dataset} $\dataset$ is 
			evenly divided into $k$ folds $\dataset^{(1)},\,\ldots,\,\dataset^{(k)}$. Each fold is used once as a 
			\gls{valset}, while the remaining $k-1$ folds form the \gls{trainset}.		
			\label{fig_kfoldcv_dict}}
		\end{figure} 
		For each fold $\foldidx = 1, \,\ldots, \,k$, we train the \gls{model} on the 
		union of all folds except $\dataset^{(\foldidx)}$ and validate it on 
		$\dataset^{(\foldidx)}$. The overall performance is obtained by averaging 
		the \gls{validation} results across all $k$ folds.
		\\
		See also: \gls{validation}, \gls{valerr}.},
	first={$k$-fold cross-validation ($k$-fold CV)},
	text={$k$-fold CV} 
}

\newglossaryentry{loo}
{name={leave-one-out cross-validation (LOO-CV)},
	description={LOO-CV is a special case\index{leave-one-out cross-validation (LOO-CV)} 
		of \gls{kfoldcv} where the \gls{valset} is of size one, i.e., a single \gls{datapoint}. 
		\\
		See also: \gls{kfoldcv}, \gls{validation}, \gls{valerr}.},
	first={leave-one-out cross-validation (LOO-CV)},
	text={LOO-CV}
}

\newglossaryentry{nestedcv}
{name={nested cross-validation (nested CV)},
	description={Nested CV\index{nested cross-validation (nested CV)} is a method 
		of extending the \gls{kfoldcv} from training and \glspl{valset} to also cover the \gls{testset}. 
		Instead of simply choosing a single \gls{testset} from the \gls{data}, two loops are run, i.e., 
		the outer and the inner loop. The outer loop uses \gls{kfoldcv} to separate a \gls{testset} 
		from the \gls{data}, and the inner loop again uses \gls{kfoldcv} to separate the remaining data 
		into training and \glspl{valset}. Doing this decreases the \gls{variance} and \gls{bias} in the results. 
		\\
		See also: \gls{kfoldcv}, \gls{loo}, \gls{validation}, \gls{valerr}.},
	first={nested cross-validation (nested CV)},
	text={nested CV},
	firstplural={nested cross-validations (nested CVs)},
	plural={nested CVs}
}

\newglossaryentry{spectrogram}
{name={spectrogram},
	description={A\index{spectrogram} spectrogram represents the time-frequency distribution of the energy of a time signal $x(t)$.  
		Intuitively, it quantifies the amount of signal energy present within a specific time segment 
		$[t_{1},t_{2}] \subseteq \mathbb{R}$ and frequency interval $[f_{1},f_{2}]\subseteq \mathbb{R}$. 
		Formally, the spectrogram of a signal is defined as the squared magnitude of its 
		short-time Fourier transform (STFT) \cite{cohen1995time}.
        		Fig.\ \ref{fig:spectrogram_dict} depicts a time signal along with its spectrogram. 
		\begin{figure}[H]
			\centering
			\includegraphics[width=0.8\textwidth]{assets/spectrogram.png}
			\begin{minipage}{\textwidth}
				\vspace{3ex}
				\centering
				{\selectfont (a) \hspace{10em} (b)}
			\end{minipage}
		\caption{(a) Time signal consisting of two modulated \gls{gaussian} pulses. (b) Intensity 
			plot of the spectrogram.
			\label{fig:spectrogram_dict}}
		\end{figure}
        		The intensity plot of the spectrogram can serve as an image of the signal. A 
		simple recipe for audio signal \gls{classification} is to feed this signal image 
		into \glspl{deepnet} originally developed for image \gls{classification} and object detection \cite{Li:2022aa}. 
		It is worth noting that, beyond the spectrogram, several alternative representations exist 
		for the time-frequency distribution of signal energy \cite{TimeFrequencyAnalysisBoashash}, \cite{MallatBook}.
					\\ 
		See also: \gls{classification}, \gls{deepnet}.}, 
	first={spectrogram},
	text={spectrogram} 
}

\newglossaryentry{graphclustering}
{name={graph clustering},
	description={\Gls{graph} \gls{clustering}\index{graph clustering} aims to 
		cluster \glspl{datapoint} that are represented as the nodes 
		of a \gls{graph} $\graph$. The edges of $\graph$ represent 
		pairwise similarities between \glspl{datapoint}. We can sometimes
		quantify the extent of these similarities by an \gls{edgeweight} \cite{FlowSpecClustering2021}, \cite{Luxburg2007}.
					\\ 
		See also: \gls{graph}, \gls{clustering}, \gls{datapoint}, \gls{edgeweight}. }, 
	first={graph clustering},
	text={graph clustering} 
}

\newglossaryentry{specclustering}
{name={spectral clustering},
	description={Spectral \gls{clustering}\index{spectral clustering} is a particular instance of 
		\gls{graphclustering}, i.e., it clusters \glspl{datapoint} 
		represented as the nodes $\nodeidx=1, \,\ldots, \,\nrnodes$ of a \gls{graph} $\graph$. 
		Spectral \gls{clustering} uses the \glspl{eigenvector} of the \gls{LapMat} $\LapMat{\graph}$ 
		to construct \glspl{featurevec} $\featurevec^{(\nodeidx)} \in \mathbb{R}^{\nrfeatures}$ 
		for each node (i.e., for each \gls{datapoint}) $\nodeidx=1, \,\ldots, \,\nrnodes$. We can feed these \glspl{featurevec} 
		into \gls{euclidspace}-based \gls{clustering} methods, such as \gls{kmeans} 
		or \gls{softclustering} via \gls{gmm}. Roughly speaking, the \glspl{featurevec} of nodes 
		belonging to a well-connected subset (or \gls{cluster}) of nodes in $\graph$ are located 
		nearby in the \gls{euclidspace} $\mathbb{R}^{\nrfeatures}$ (see Fig.\ \ref{fig_lap_mtx_specclustering_dict}). 
		\begin{figure}[H]
			\begin{center}
				\begin{minipage}{0.4\textwidth}
			\begin{tikzpicture}
				\begin{scope}[every node/.style={circle, fill=black, inner sep=0pt, minimum size=0.3cm}]
					\node (1) at (0,0) {};
					\node (2) [below left=of 1, xshift=-0.2cm, yshift=-1cm] {};
					\node (3) [below right=of 1, xshift=0.2cm, yshift=-1cm] {};
					\node (4) [below=of 1, yshift=0.5cm] {}; 
				\end{scope}
				\draw (1) -- (2);
				\draw (1) -- (3);
				\node[above=0.2cm] at (1) {$\nodeidx=1$};
				\node[left=0.3cm] at (2) {$2$};
				\node[right=0.3cm] at (3) {$3$};
				\node[below=0.2cm] at (4) {$4$};
				\node at (0,-4) {(a)};
			\end{tikzpicture}
				\end{minipage} 
				\hspace*{5mm}
				\begin{minipage}{0.4\textwidth}
					\begin{equation} 
						\LapMat{\graph}\!=\!
						\begin{pmatrix} 
							2 & -1 & -1 & 0 \\ 
							-1 & 1 & 0 & 0 \\  
							-1 & 0 & 1 & 0 \\ 
							0 & 0 & 0 & 0 
						\end{pmatrix}\!=\!\mathbf{V} {\bm \Lambda} \mathbf{V}\,^{T}  
						\nonumber
					\end{equation} 
					\begin{minipage}{\textwidth}
						\vspace{3ex}
						\centering
						{\selectfont (b)}
					\end{minipage}
				\end{minipage}
				\vspace*{20mm}\\
				\begin{minipage}{0.4\textwidth}
			\begin{tikzpicture}[scale=3]
					\draw[->] (-0.2, 0) -- (1.2, 0) node[right] {$v^{(1)}_{\nodeidx}$};
					\draw[->] (0, -0.2) -- (0, 1.2) node[above] {$v^{(2)}_{\nodeidx}$};
					\filldraw[blue] (0.577, 0) circle (0.03cm) node[above right] {$\nodeidx=1,2,3$};
					\filldraw[blue] (0.577, 0) circle (0.03cm); 
					\filldraw[blue] (0.577, 0) circle (0.03cm); 
					\filldraw[red] (0, 1) circle (0.03cm) node[above right] {$4$};
					\node at (0.5,-0.5) {(c)};
			\end{tikzpicture}
				\end{minipage} 
    				\begin{minipage}{0.4\textwidth}
					\begin{align}
					& \mathbf{V} = \big( \vv^{(1)},\vv^{(2)},\vv^{(3)},\vv^{(4)} \big) \nonumber \\
					&	\mathbf{v}^{(1)}\!=\!\frac{1}{\sqrt{3}} \begin{pmatrix} 1 \\ 1 \\ 1 \\ 0 \end{pmatrix}, \,
					\mathbf{v}^{(2)}\!=\!\begin{pmatrix} 0 \\ 0 \\ 0 \\ 1 \end{pmatrix} \nonumber 
					\end{align}
					\begin{minipage}{\textwidth}
						\vspace{3ex}
						\centering
						{\selectfont (d)}
					\end{minipage}
				\end{minipage} 
		\caption{\label{fig_lap_mtx_specclustering_dict} (a) \Gls{undirectedgraph} 
			$\graph$ with four nodes $\nodeidx=1,\,2,\,3,\,4$, each representing a \gls{datapoint}. (b) \gls{LapMat} 
			$\LapMat{\graph} \in \mathbb{R}^{4 \times 4}$ and its \gls{evd}. 
			(c) \Gls{scatterplot} of \glspl{datapoint} using the \glspl{featurevec} 
			$\featurevec^{(\nodeidx)} = \big( v^{(1)}_{\nodeidx},v^{(2)}_{\nodeidx} \big)\,^{T}$. 
			(d) Two \glspl{eigenvector} $\vv^{(1)},\vv^{(2)} \in \mathbb{R}^{\nrfeatures}$ 
			corresponding to the \gls{eigenvalue} $\lambda=0$ of the \gls{LapMat} $\LapMat{\graph}$. } 
			\end{center}
		\end{figure}
		See also: \gls{clustering}, \gls{graphclustering}, \gls{LapMat}, \gls{eigenvalue}.}, 
	first={spectral clustering},
	text={spectral clustering} 
}

\newglossaryentry{flowbasedclustering}
{name={flow-based clustering},
	description={Flow-based \gls{clustering}\index{flow-based clustering} groups the nodes 
		of an \gls{undirectedgraph} by applying \gls{kmeans} \gls{clustering} to nodewise 
		\glspl{featurevec}. These \glspl{featurevec} are built from network flows between 
		carefully selected sources and destination nodes \cite{FlowSpecClustering2021}. 
					\\ 
		See also: \gls{clustering}, \gls{graph}, \gls{kmeans}, \gls{featurevec}.}, 
	first={flow-based clustering},
	text={flow-based clustering} 
}

\newglossaryentry{esterr}
{name={estimation error},
	description={Consider\index{estimation error} \glspl{datapoint}, each with \gls{featurevec} $\featurevec$ and \gls{label} 
		$\truelabel$. In some applications, we can model the relation between the \gls{featurevec} and the \gls{label}
		of a \gls{datapoint} as $\truelabel = \bar{\hypothesis}(\featurevec) + \varepsilon$. Here, we 
		use some true underlying \gls{hypothesis} $\bar{\hypothesis}$ and a noise term $\varepsilon$, 
		which summarizes any modeling or labeling errors. The estimation error incurred by an \gls{ml} 
		method that learns a \gls{hypothesis} $\widehat{\hypothesis}$, e.g., using \gls{erm}, is defined as 
		$\widehat{\hypothesis}(\featurevec) - \bar{\hypothesis}(\featurevec)$ for some \gls{featurevec}. 
		For a parametric \gls{hypospace}, which consists of \gls{hypothesis} \glspl{map} determined by 
		\glspl{modelparam} $\weights$, we can define the estimation error as 
		$\Delta \weights = \widehat{\weights} - \overline{\weights}$ \cite{kay}, \cite{hastie01statisticallearning}.
					\\ 
		See also: \gls{datapoint}, \gls{featurevec}, \gls{label}, \gls{hypothesis}, \gls{ml}, \gls{erm}, \gls{hypospace}, \gls{map}, \gls{modelparam}.},
	first={estimation error},
	text={estimation error} 
}

\newglossaryentry{dob}
{name={degree of belonging}, 
	description={Degree of belonging\index{degree of belonging} is a number that indicates the extent to which a \gls{datapoint} 
		belongs to a \gls{cluster} \cite[Ch. 8]{MLBasics}. The degree of belonging can be 
		interpreted as a soft \gls{cluster} assignment. \Gls{softclustering} methods can 
		encode the degree of belonging with a real number in the interval $[0,1]$. 
		\Gls{hardclustering} is obtained as the extreme case when the degree of belonging 
		only takes on values $0$ or $1$.
					\\ 
		See also: \gls{datapoint}, \gls{cluster}, \gls{softclustering}, \gls{hardclustering}.}, 
	first={degree of belonging},
	firstplural={degrees of belonging},
	plural={degrees of belonging},
	text={degree of belonging} 
}

\newglossaryentry{msee}
{name={mean squared estimation error (MSEE)},
	description={Consider\index{mean squared estimation error (MSEE)} an \gls{ml} method that 
		learns \glspl{modelparam} $\widehat{\weights}$ based on some \gls{dataset} $\dataset$. 
		If we interpret the \glspl{datapoint} in $\dataset$ as \gls{iid} \glspl{realization} of an \gls{rv} $\datapoint$, 
		we define the \gls{esterr} $\Delta \weights \defeq \widehat{\weight} - \overline{\weights}$. 
		Here, $\overline{\weights}$ denotes the true \glspl{modelparam} of the \gls{probdist} 
		of $\datapoint$. The MSEE is 
		defined as the \gls{expectation} $\expect \big\{ \big\| \Delta \weights \big\|^{2} \big\}$ of the 
		squared \gls{euclidnorm} of the \gls{esterr} \cite{LC}, \cite{kay}.
					\\ 
		See also: \gls{rv}, \gls{esterr}, \gls{probmodel}, \gls{sqerrloss}.},
	first={mean squared estimation error (MSEE)},
	text={MSEE} 
}

\newglossaryentry{gtvmin}
{name={generalized total variation minimization (GTVMin)},
	description={GTVMin\index{generalized total variation minimization (GTVMin)} is an instance of \gls{rerm} 
		using the \gls{gtv} of local \glspl{modelparam} as a \gls{regularizer} \cite{ClusteredFLTVMinTSP}.
					\\ 
		See also: \gls{rerm}, \gls{gtv}, \gls{regularizer}.},
	first={generalized total variation minimization (GTVMin)},
	text={GTVMin} 
}

\newglossaryentry{networklasso}
{name={network Lasso},
sort={networkLasso},
	description={The network \gls{lasso}\index{network Lasso} is a special case of 
		\gls{gtvmin}, obtained by using a \gls{norm}-based \gls{discrepancy} 
		\gls{measure} for comparing local \glspl{modelparam} \cite{NetworkLasso}. 
		It can also be viewed as a \gls{generalization} of the \gls{lasso} 
		method to \glspl{dataset} and \glspl{model} with an intrinsic network 
		structure.
			\\
		See also: \gls{lasso}, \gls{gtvmin}, \gls{gtv}, \gls{rerm}, \gls{regularizer}.},
	first={network Lasso},
	text={network Lasso}
}

\newglossaryentry{regression}
{name={regression},
	description={Regression\index{regression} problems revolve around the 
		\gls{prediction} of a numeric \gls{label} solely from the \glspl{feature} of a \gls{datapoint} \cite[Ch. 2]{MLBasics}.
					\\ 
		See also: \gls{prediction}, \gls{label}, \gls{feature}, \gls{datapoint}.},
	first={regression},
	text={regression} 
}

\newglossaryentry{acc}
{name={accuracy},
	description={Consider\index{accuracy} \glspl{datapoint} characterized by \glspl{feature} $\featurevec \in \featurespace$ and 
		a categorical \gls{label} $\truelabel$ that takes on values from a finite \gls{labelspace} $\labelspace$. The 
		accuracy of a \gls{hypothesis} $\hypothesis: \featurespace \rightarrow \labelspace$, when applied to the \glspl{datapoint} in a \gls{dataset} 
		$\dataset = \big\{ \big(\featurevec^{(1)}, \truelabel^{(1)} \big), \,\ldots, \,\big(\featurevec^{(\samplesize)},\truelabel^{(\samplesize)}\big) \big\}$, 
		is then defined as $1 - (1/\samplesize)\sum_{\sampleidx=1}^{\samplesize} \lossfunczo{\big(\featurevec^{(\sampleidx)},\truelabel^{(\sampleidx)}\big)}{\hypothesis}$ 
		using the \gls{zerooneloss} $\lossfunczo{\cdot}{\cdot}$.
					\\ 
		See also: \gls{zerooneloss}, \gls{loss}, \gls{metric}.},
	first={accuracy},
	text={accuracy} 
}

\newglossaryentry{expert}
{name={expert},
	description={\gls{ml}\index{expert} aims to learn a \gls{hypothesis} $\hypothesis$ that accurately predicts the \gls{label} 
		of a \gls{datapoint} based on its \glspl{feature}. We measure the \gls{prediction} error using 
		some \gls{lossfunc}. Ideally, we want to find a \gls{hypothesis} that incurs minimal \gls{loss} 
		on any \gls{datapoint}. We can make this informal goal precise via the \gls{iidasspt} 
		and by using the \gls{bayesrisk} as the \gls{baseline} for the (average) \gls{loss} of a \gls{hypothesis}. 
		An alternative approach to obtaining a \gls{baseline} is to use the \gls{hypothesis} $\hypothesis'$ learned 
		by an existing \gls{ml} method. We refer to this \gls{hypothesis} $\hypothesis'$ as an expert \cite{PredictionLearningGames}. 
		\Gls{regret} minimization methods learn a \gls{hypothesis}
		that incurs a \gls{loss} comparable to that of the best expert \cite{PredictionLearningGames}, \cite{HazanOCO}.
					\\ 
		See also: \gls{lossfunc}, \gls{baseline}, \gls{regret}.},
	first={expert},
	text={expert} 
}

\newglossaryentry{nfl}
{name={networked federated learning (NFL)},
sort={networkedfederated},
	description={NFL\index{networked federated learning (NFL)} refers 
		to methods that learn personalized \glspl{model} in a distributed fashion. These methods learn from \glspl{localdataset} 
		that are related by an intrinsic network structure.
					\\ 
		See also: \gls{model}, \gls{localdataset}, \gls{fl}.},
	first={networked federated learning (NFL)},
	text={NFL} 
}

\newglossaryentry{fedprox}
{name={federated proximal (FedProx)},
	description={FedProx\index{federated proximal (FedProx)} refers to an 
		iterative \gls{fl} \gls{algorithm} that alternates between separately 
		\gls{training} \glspl{localmodel} and combining the updated 
		local \glspl{modelparam}. In contrast to \gls{fedavg}, which uses 
		\gls{stochGD} to train \glspl{localmodel}, FedProx uses a \gls{proxop} 
		for the \gls{training} \cite{FedProx2020}.
					\\ 
		See also: \gls{fl}, \gls{algorithm}, \gls{localmodel}, \gls{modelparam}, \gls{fedavg}, \gls{stochGD}, \gls{proxop}.}, 
	first={FedProx}, 
	text={FedProx} 
}

\newglossaryentry{relu}
{name={rectified linear unit (ReLU)},
	description={The\index{rectified linear unit (ReLU)} ReLU is 
		a popular choice for the \gls{actfun} of a \gls{neuron} within an \gls{ann}. It is defined 
		as $\actfun(z) = \max\{0,z\}$, with $z$ being the weighted input of the \gls{artificialneuron}.
					\\ 
		See also: \gls{actfun}, \gls{ann}.}, 
	first={rectified linear unit (ReLU)}, 
	text={ReLU} 
}

\newglossaryentry{hypothesis}
{name={hypothesis},
	description={A\index{hypothesis} hypothesis refers to a \gls{map} (or \gls{function}) $\hypothesis: \featurespace \rightarrow \labelspace$ 
		from the \gls{featurespace} $\featurespace$ to the \gls{labelspace} $\labelspace$. 
		Given a \gls{datapoint} with \glspl{feature} $\featurevec$, we use a hypothesis 
		\gls{map} $\hypothesis$ to estimate (or approximate) the \gls{label} $\truelabel$ 
		using the \gls{prediction} $\hat{\truelabel} = \hypothesis(\featurevec)$. 
		\begin{figure}[H]
		\centering
			\begin{tikzpicture}[
 		  	>=Latex, node distance=2.0cm,
 		  	box/.style={draw, rounded corners=2pt, inner sep=6pt},
  		 	label/.style={font=\footnotesize},
  		 	thinline/.style={line width=0.6pt}
		 	]
		 	\node[minimum width=3.8cm, minimum height=1.6cm] (audio) {};
		 	\node[label] at (audio.north) [yshift=0mm] {audio \glspl{sample} $\featurevec \in \mathbb{R}^{\nrfeatures}$};
		 	\begin{scope}
		 	\draw[thinline]
		 		($(audio.west)+(0.2,0)$) .. controls +(.3,.35) and +(-.3,.35) .. ++(0.8,0)
		 		.. controls +(.3,-.35) and +(-.3,-.35) .. ++(0.8,0)
		 		.. controls +(.3,.25) and +(-.3,.25) .. ++(0.8,0)
		 		.. controls +(.3,-.25) and +(-.3,-.25) .. ++(0.8,0);
		 	\end{scope}
		 	\node[box,right=1.0cm of audio, minimum width=2.2cm, minimum height=1.6cm] (model)
		 	{$\hypothesis$};
		 	\draw[->,thinline] (audio) -- (model) ; 
		 	\node[right=1.0cm of model, minimum width=3.2cm, minimum height=1.6cm] (rating) {};
		 	\node[label, align=center] at ($(rating.north)+(0,-6mm)$)
		 		{};
		 	\coordinate (barL) at ($(rating.west)+(0.4,0)$);
		 	\coordinate (barR) at ($(rating.east)+(-0.4,0)$);
		 	\def\score{0.82}
		 	\coordinate (ptr) at ($(barL)!{\score}!(barR)$);
		 	\node[label, above=0pt of ptr] {$\hypothesis(\featurevec)=0.82 (\approx \mbox{Freddie level})$};
		 	\draw[->,thinline] (model) -- (rating);
		 	\end{tikzpicture}
		\caption{\label{fig:hypothesis_dict} Hypothesis $\hypothesis: \featurespace \rightarrow \labelspace$ mapping the \glspl{feature} 
			$\featurevec \in \featurespace$ of a \gls{datapoint} to a \gls{prediction} $\hypothesis(\featurevec) \in \labelspace$ of the \gls{label}. 
			For example, the \gls{ml} application \url{https://freddiemeter.withyoutube.com/} uses the \glspl{sample} of an audio 
			recording as \glspl{feature} to predict how closely a person’s singing resembles that of Freddie Mercury. }
		\end{figure}
		\Gls{ml} is all about learning (or finding) a hypothesis \gls{map} $\hypothesis$ 
		such that $\truelabel \approx \hypothesis(\featurevec)$ for any \gls{datapoint} 
		(with \glspl{feature} $\featurevec$ and \gls{label} $\truelabel$). Practical \gls{ml} methods, 
		limited by finite computational resources, must restrict learning to a subset of all possible 
		hypothesis \glspl{map}. This subset is called the \gls{hypospace} or simply the \gls{model} underlying 
		the method.
					\\ 
		See also: \gls{map}, \gls{function}, \gls{prediction}, \gls{model}.},
	first={hypothesis},
	firstplural={hypotheses},
	plural={hypotheses},
	text={hypothesis}  
}

\newglossaryentry{effdim}
{name={effective dimension},
	description={The\index{effective dimension} effective \gls{dimension} $\effdim{\hypospace}$ of 
		an infinite \gls{hypospace} $\hypospace$ is a \gls{measure} of its size. Loosely speaking, the 
		effective \gls{dimension} is equal to the effective number of independent tunable \glspl{modelparam}. 
		These \glspl{parameter} might be the coefficients used in a \gls{linearmap} or the 
		\glspl{weight} and \gls{bias} terms of an \gls{ann}.
					\\ 
		See also: \gls{hypospace}, \gls{modelparam}, \gls{ann}.},
	first={effective dimension},
	text={effective dimension}  
}

\newglossaryentry{labelspace}
{name={label space},
sort={labelspace},
	description={In an \gls{ml} application\index{label space}, each \gls{datapoint} is described by a 
		set of \glspl{feature} together with an associated \gls{label}. 
		The set of all admissible \gls{label} values is called the \gls{label} space,
		denoted by $\labelspace$. Importantly, $\labelspace$ may include values that no 
		observed \gls{datapoint} has as its \gls{label} value. 
		To a large extent, the choice of $\labelspace$ is up to the \gls{ml} engineer 
		and depends on the problem formulation. Fig.\ \ref{fig_label_spaces_dict} shows some examples 
		of \gls{label} spaces that are commonly used in \gls{ml} applications.
		\begin{figure}[H]
		\centering
		\begin{tikzpicture}[>=Stealth, font=\small]
			\begin{scope}[shift={(0,0)}]
				\draw[->] (-2,0) -- (2,0);
				\node[below=6pt] at (0,-0.7) {$\labelspace\!=\!\mathbb{R}$ (\gls{regression})};
				\node at (0,-2) {(a)};
			\end{scope}
			\begin{scope}[shift={(7,0)}]
				\fill[gray!20] (-1,-0.5) rectangle (1,0.5);
				\draw[->] (-2,0) -- (2,0);
				\draw[->] (0,-1) -- (0,1);
				\node[below=6pt] at (0,-0.7) {$\labelspace\!=\!\mathbb{R}^{2}$ (multi-label \gls{regression})};
				\node at (0,-2) {(b)};
			\end{scope}
			\begin{scope}[shift={(0,-3)}]
				\fill (-1,0) circle (1.2pt) node[below=2pt] {\text{``hot''}};
				\fill ( 1,0) circle (1.2pt) node[below=2pt] {\text{``cold''}};
				\node[below=14pt] at (0,-0.7) {$|\labelspace|=2$ (\gls{binclass})};
				\node at (0,-2.3) {(c)};
			\end{scope}
			\begin{scope}[shift={(7,-3)}]
				\node[circle, inner sep=1pt, draw] (n1) at (-1.5,0) {};
				\node[circle, inner sep=1pt, draw] (n2) at (-0.5,0) {};
				\node[circle, inner sep=1pt, draw] (n3) at ( 0.5,0) {};
				\node[circle, inner sep=1pt, draw] (n4) at ( 1.5,0) {};
				\draw[->] (n1) -- (n2);
				\draw[->] (n2) -- (n3);
				\draw[->] (n3) -- (n4);
				\node[below=2pt of n1] {1};
				\node[below=2pt of n2] {2};
				\node[below=2pt of n3] {3};
				\node[below=2pt of n4] {4};
				\node[below=14pt] at (0,-0.7) {$\labelspace\!=\!\{1,\,2,\,3,\,4\}$ (ordinal \gls{regression})};
				\node at (0,-2.3) {(d)};
			\end{scope}
		\end{tikzpicture}
		\caption{\label{fig_label_spaces_dict} Examples of \gls{label} spaces and the corresponding types of \gls{ml}.
			(a) \Gls{regression}. (b) Multi-label \gls{regression}. (c) \Gls{binclass}. (d) Ordinal \gls{regression}.}
		\end{figure}
		The choice of the \gls{label} space $\labelspace$ determines the type of \gls{ml} methods 
		appropriate for the application at hand. \Gls{regression} methods use the $\labelspace = \mathbb{R}$, 
		while \gls{binclass} methods use a \gls{label} space $\labelspace$ that consists of two different 
		elements, i.e., $|\labelspace|=2$. Ordinal \gls{regression} methods use a finite, ordered set of 
		\gls{label} values, e.g., $\labelspace = \{1,\,2,\,3,\,4\}$ with the natural ordering $1 < 2 < 3 < 4$. 
					\\ 
		See also:  \gls{datapoint}, \gls{label}, \gls{regression}, \gls{binclass}.}, 
	first={label space},
	text={label space}  
}

\newglossaryentry{prediction}
{name={prediction},
	description={A\index{prediction} prediction is an estimate or approximation for some 
		quantity of interest. \Gls{ml} revolves around learning or finding a \gls{hypothesis} \gls{map} $\hypothesis$ 
		that reads in the \glspl{feature} $\featurevec$ of a \gls{datapoint} and delivers a prediction 
		$\widehat{\truelabel} \defeq \hypothesis(\featurevec)$ for its \gls{label} $\truelabel$.
					\\ 
		See also: \gls{ml}, \gls{hypothesis}, \gls{map}, \gls{feature}, \gls{datapoint}, \gls{label}.},
	first={prediction},
	plural={predictions},
	text={prediction}  
}

\newglossaryentry{histogram}
{name={histogram},
	description={Consider\index{histogram} a \gls{dataset} $\dataset$ that consists of 
		$\samplesize$ \glspl{datapoint} $\datapoint^{(1)}, \,\ldots, \,\datapoint^{(\samplesize)}$, 
		each of them belonging to some cell $[-U,U] \times \ldots \times [-U,U] \subseteq \mathbb{R}^{\featuredim}$ with side 
		length $U$. We partition this cell evenly into smaller elementary cells with side 
		length $\Delta$. The histogram of $\dataset$ assigns each elementary cell to 
		the corresponding fraction of \glspl{datapoint} in $\dataset$ that fall into this 
		elementary cell. A visual example of such a histogram is provided in Fig.\ \ref{fig:histogram_dict}.
		\begin{figure}[H]
		\centering
		\begin{tikzpicture}
		\pgfplotsset{compat=1.18}
		\begin{axis}[
		    ybar,
		    ymin=0,
		    ymax=6,
		    bar width=22pt,
		    width=10cm,
		    height=6cm,
		    xlabel={Value},
		    ylabel={Frequency},
		    ytick={1,2,3,4,5,6},
		    xtick={1,2,3,4,5},
		    xticklabels={{[0,1)}, {[1,2)}, {[2,3)}, {[3,4)}, {[4,5)}},
		    enlarge x limits=0.15,
		    title={Histogram of \gls{sample} \gls{data}}
			]
		\addplot+[fill=blue!40] coordinates {(1,2) (2,5) (3,4) (4,3) (5,1)};
		\end{axis}
		\end{tikzpicture}
		\caption{Histogram consisting of the fractions of \glspl{datapoint} that 
			fall within different value ranges (i.e., bins). Each bar height shows 
			the count of \glspl{sample} in the corresponding interval.}
			\label{fig:histogram_dict}
		\end{figure}
		See also: \gls{dataset}, \gls{datapoint}, \gls{sample}.},
	first={histogram},
	text={histogram}  
}

\newglossaryentry{bootstrap}
{name={bootstrap},
	description={For\index{bootstrap} the analysis of \gls{ml} methods, it is often 
	    	useful to interpret a given \gls{dataset} $\dataset = \big\{ \datapoint^{(1)}, \,\ldots, \,\datapoint^{(\samplesize)}\big\}$ 
		as a collection of (\glspl{realization} of) \gls{iid} \glspl{rv} with 
		common \gls{probdist} $\probdist$. In practice, the \gls{probdist} $\probdist$ 
		is unknown and must be estimated from $\dataset$. The idea of the bootstrap 
		method is to use the \gls{empiricaldistribution} $\probdist^{(\dataset)}$ 
		of $\dataset$ as an \gls{estimator} for $\probdist$ \cite{hastie01statisticallearning}:
		$$ \frac{1}{\samplesize} \big| \sampleidx: \datapoint^{(\sampleidx)} \in \mathcal{A} \big| \approx \prob{\mathcal{A}}.$$
		Repeatedly sampling from the \gls{empiricaldistribution}, which is equivalent 
		to sampling with replacement from $\dataset$ \cite{BoostrapBook}, 
		results in new \glspl{dataset} $\dataset^{(1)}, \,\ldots, \,\dataset^{(\nrbootstraps)}$, each containing 
		$\samplesize$ \glspl{datapoint}. We then use each of 
		those \glspl{dataset} for \gls{model} \gls{training} (e.g., via \gls{erm}), 
		resulting in the learned \glspl{hypothesis}
		$\widehat{\hypothesis}^{(1)}, \,\ldots, \,\learnthypothesis^{(\nrbootstraps)}.$ 
        		We can use these learned \glspl{hypothesis} to estimate important characteristics 
		of an \gls{ml} method such as \gls{bias}, \gls{variance}, or \gls{gengap} \cite{hastie01statisticallearning}.
		\\
		See also: \gls{iid}, \gls{rv}, \gls{probdist}, \gls{histogram}.},
	first={bootstrap},
	text={bootstrap}  
}

\newglossaryentry{featurespace}
{name={feature space},
	description={The\index{feature space} \gls{feature} space of a given \gls{ml} application 
		or method is constituted by all potential values that the \gls{featurevec} of a \gls{datapoint} can take on. 
		For \glspl{datapoint} described by a fixed number $\nrfeatures$ of numerical \glspl{feature}, 
		a common choice for the \gls{feature} space is the \gls{euclidspace} $\mathbb{R}^{\nrfeatures}$. 
		However, the mere presence of $\nrfeatures$ numeric \glspl{feature} does not imply that $\mathbb{R}^{\nrfeatures}$ 
		is the most appropriate representation of the \gls{feature} space. Indeed, the numerical \glspl{feature}  
		might be assigned to \glspl{datapoint} in a largely arbitrary or random manner, resulting 
		in \glspl{datapoint} that are randomly scattered throughout $\mathbb{R}^{\nrfeatures}$ 
		without any meaningful geometric structure. \Gls{featlearn} methods try to learn a 
		transformation of the original (potentially nonnumeric) \glspl{feature} to ensure a 
		more meaningful arrangement of \glspl{datapoint} in $\mathbb{R}^{\nrfeatures}$. 
		Three examples of \gls{feature} spaces are shown in Fig.\ \ref{fig_featurespace_dict}.
		\begin{figure}[H]
			\centering
			\begin{tikzpicture}[scale=0.6]
			\begin{scope}[xshift=0cm]
 	 			\draw[->] (-0.5, 0) -- (3.5, 0) node[right] {$x_1$};
  				\foreach \x/\lbl in {0.5/$\featurevec^{(1)}$, 1.5/$\featurevec^{(2)}$, 2.8/$\featurevec^{(3)}$}
    				\filldraw[blue] (\x,0) circle (2pt) node[above] {\lbl};
  				\node at (1.5, -4.0)  {$\featurespace^{(1)}$};
				\node at (1.5, -6) {(a)};
			\end{scope}
			\begin{scope}[xshift=8cm]
  				\draw[thick] (0,0) circle (1.8);
  				\fill[gray!5] (0,0) circle (1.8);
  				\filldraw[red] (0.8, 0.9) circle (2pt) node[anchor=south west] {$\featurevec^{(1)}$};
  				\filldraw[blue] (-1.2, 0.5) circle (2pt) node[anchor=south east] {$\featurevec^{(2)}$};
  				\filldraw[green!50!black] (0.3, -1.4) circle (2pt) node[anchor=north west] {$\featurevec^{(3)}$};
  				\node at (0.5, -4) {$\featurespace^{(2)}$};
				\node at (0.5, -6) {(b)};
			\end{scope}
			\begin{scope}[xshift=14cm, yshift=0.3cm]
 	 			\filldraw[blue!60] (0,0) circle (2pt) node[anchor=north east] {$\featurevec^{(1)}$};
 	 			\filldraw[red!70] (2,1.2) circle (2pt) node[anchor=south west] {$\featurevec^{(2)}$};
  				\filldraw[green!50!black] (1,2.5) circle (2pt) node[anchor=south east] {$\featurevec^{(3)}$};
  				\filldraw[orange!80] (3,2.5) circle (2pt) node[anchor=south west] {$\featurevec^{(4)}$};
  				\draw[-] (0,0) -- (2,1.2);
  				\draw[-] (2,1.2) -- (1,2.5);
  				\draw[-] (1,2.5) -- (3,2.5);
  				\node at (1.5, -4.2) {$\featurespace^{(3)}$};
				\node at (1.5, -6.2) {(c)};
			\end{scope}
			\end{tikzpicture}
		\caption{Three different \gls{feature} spaces. (a) Linear space $\featurespace^{(1)} = \mathbb{R}$. (b) Bounded 
			\gls{convex} set $\featurespace^{(2)} \subseteq \mathbb{R}^{2}$. (c) Discrete space 
			$\featurespace^{(3)}$ whose elements are nodes of an \gls{undirectedgraph}. \label{fig_featurespace_dict}}
		\end{figure}
		See also: \gls{featurevec}, \gls{euclidspace}.},
	first={feature space},
	text={feature space}  
}

\newglossaryentry{missingdata}
{name={missing data},
	description={Consider\index{missing data} a \gls{dataset} constituted by \glspl{datapoint} collected via 
		some physical \gls{device}. Due to imperfections and failures, some of the \gls{feature} 
		or \gls{label} values of \glspl{datapoint} might be corrupted or simply missing. 
		\Gls{data} imputation aims to estimate these missing values \cite{Abayomi2008DiagnosticsFM}. 
		We can interpret \gls{data} imputation as an \gls{ml} problem where the \gls{label} of a \gls{datapoint} is 
		the value of the corrupted \gls{feature}.
				\\
		See also: \gls{feature}, \gls{label}. },
	first={missing data},
	text={missing data}  
}

\newglossaryentry{hyperparameter}
{name={hyperparameter},
	description={A hyperparameter\index{hyperparameter} associated with an \gls{ml} method is 
		a quantity that is used to select among a family of \glspl{model}. 
		Typical examples include the \gls{learnrate} used in a \gls{gdmethod}, 
		the number of \glspl{feature} used in a \gls{linmodel}, or the 
		\gls{maximum} depth of a \gls{decisiontree}. The usefulness of a specific 
		hyperparameter choice can be assessed via \gls{validation}. Similar 
		to learning (or tuning) \glspl{modelparam} by \gls{erm} on a \gls{trainset}, 
		we can learn (or tune) hyperparameters via minimizing the \gls{valerr}. Thus, 
		in a sense, hyperparameters are higher-level \glspl{modelparam} that are learned 
		via a higher-level form of \gls{erm}, i.e., minimizing the \gls{valerr} obtained by 
		the trained \gls{model} with a given hyperparameter value. 
			\\ 
		See also: \gls{model}, \gls{validation}, \gls{modelparam}.},
	first={hyperparameter},
	text={hyperparameter},
	plural={hyperparameters},
	firstplural={hyperparameters}
}

\newglossaryentry{dataimputation}
{name={data imputation},
	description={See\index{data imputation} \gls{missingdata}.},
	first={data imputation},
	text={data imputation}  
}

\newglossaryentry{feature}
{name={feature},
	description={A\index{feature} feature of a \gls{datapoint} is one of its properties 
	    	that can be measured or computed easily without the need for human supervision. 
		For example, if a \gls{datapoint} is a digital image (e.g., stored as a \texttt{.jpeg} file), 
		then we could use the red–green–blue (RGB) intensities of its pixels as features. 
		Another example is shown in Fig.\ \ref{fig:audio_features_dict}, where the signal 
		\glspl{sample} of a finite-duration audio signal are used as its features.
		\begin{figure}[H]
		\centering
		\begin{tikzpicture}[scale=1]
		\draw[thick, blue, domain=0:6.28, smooth, variable=\x] 
			plot ({\x}, {sin(\x r)});
		\foreach \x [count=\i] in {0,0.5,...,6.28} {
			\fill[red] (\x, {sin(\x r)}) circle (2pt);
			\ifnum\i=1
				\node[above left] at (\x, {sin(\x r)}) {$x_1$};
			\fi
			\ifnum\i=2
				\node[above left] at (\x, {sin(\x r)}) {$x_2$};
			\fi
		}
		\end{tikzpicture}
		\caption{Audio signal (blue waveform) and its discretized signal \glspl{sample} (red dots), which 
			can be used as features $x_{1},\,\ldots,\,x_{\nrfeatures}$. \label{fig:audio_features_dict}}
		\end{figure}
		Domain-specific synonyms for the term feature are "\gls{covariate}," "explanatory variable," 
		"independent variable," "input (variable)," "\gls{predictor} (variable)," or "regressor" 
		\cite{Everitt2010}, \cite{Gujarati2021}, \cite{Dodge2003}. 
		\\
		See also: \gls{datapoint}.}, 
	first={feature},
	plural={features},
	text={feature}  
}

\newglossaryentry{featurevec}
{name={feature vector},
	description={\Gls{feature} \gls{vector} refers to a\index{feature vector} \gls{vector} $\vx = \big(x_{1}, \,\ldots, \,x_{\nrfeatures}\big)\,^{T}$ 
		whose entries are individual \glspl{feature} $x_{1}, \,\ldots, \,x_{\nrfeatures}$. Many \gls{ml} methods 
		use \gls{feature} \glspl{vector} that belong to some finite-dimensional \gls{euclidspace} $\mathbb{R}^{\nrfeatures}$. 
		For some \gls{ml} methods, however, it can be more convenient to work with \gls{feature} 
		\glspl{vector} that belong to an infinite-dimensional \gls{vectorspace} (e.g., see \gls{kernelmethod}). 
			\\
		See also: \gls{feature}, \gls{vector}, \gls{ml}, \gls{euclidspace}, \gls{vectorspace}.}, 
	first={feature vector},
	plural={feature vectors},
	firstplural={feature vectors},
	text={feature vector}  
}

\newglossaryentry{label}
{name={label},
	description={A label\index{label} is a higher-level fact or quantity of interest associated 
		with a \gls{datapoint}. For example, if the \gls{datapoint} is an image, the label 
		could indicate whether the image contains a cat 
		\cite{Everitt2010}, \cite{Gujarati2021}, \cite{Dodge2003}.
				\\
		See also: \gls{datapoint}, \gls{labelspace}.},
	first={label},
	plural={labels},
	text={label}  
}

\newglossaryentry{data}
{name={data},
	description={In the context of \gls{ml}, the term 
 		data\index{data} is often used as a synonym for \gls{dataset}
  		\cite{Everitt2010}, \cite{OxfordStatisticsDictionary}. 
  		The ISO/IEC 2382:2015 standard \cite{ISO2382} defines data as a \begin{quote} reinterpretable representation of 
  		information in a formalized manner suitable for communication, interpretation, or processing.\end{quote}
  		See also: \gls{dataset}, \gls{datapoint}, \gls{sample}.}, 
  	first={data},
	text={data}
}

\newglossaryentry{relational model}
{name={relational model},
	description={A relational \gls{model}\index{relational model} is a mathematical representation of \gls{data}. 
		The core idea is to ogranize \gls{data} as a collection of 
		tables (or relations) \cite{silberschatz2019database}, \cite{codd1970relational}. 
		A table consists of rows and columns, where each 
		row corresponds to a single \gls{datapoint}, and each 
		column represents a specific attribute of a \gls{datapoint}. 
		\gls{ml} methods use these attributes as the \glspl{feature} 
		and \gls{label} of a \gls{datapoint}. Table~\ref{tab:relmodel_dict} shows 
		a relational representation of a \gls{dataset} that consists of cows. 
		In the relational \gls{model}, the order of rows is immaterial, and each 
		attribute (i.e., column) is associated with a \gls{domain} that specifies the set 
		of admissible values. 
		In \gls{ml} applications, these attribute \glspl{domain} correspond to the 
		\gls{featurespace} and the \gls{labelspace}.
				\begin{table}[H]
					\refstepcounter{table}
					\caption*{
						\centering 
						\scshape TABLE \thetable \\[0.5ex]
						\scshape Relation (or Table) Representing the \gls{dataset} in Fig.\ \ref{fig_cows_dataset_dict} 
					}
					\label{tab:relmodel_dict} 
					\centering
					\begin{tabular}{lcccc}
						\hline
						\textbf{Name} & \textbf{Weight} & \textbf{Age} & \textbf{Height} & \textbf{Stomach temperature} \\
						\hline
						Zenzi & 100 & 4 & 100 & 25 \\
						Berta & 140 & 3 & 130 & 23 \\
						Resi  & 120 & 4 & 120 & 31 \\
						\hline
					\end{tabular}
				\end{table}
		See also: \gls{data}, \gls{datapoint}, \glspl{feature}, \gls{label}, \gls{dataset}.}, 
	first={relational model},
	text={relational model}
}
  		
\newglossaryentry{tabulardata}
{name={tabular data},
	description={Tabular \gls{data}\index{tabular data} consist of \glspl{datapoint} 
		that are characterized by a common set of attributes. These attributes 
		can be used as the \glspl{feature} or \glspl{label} of \glspl{datapoint}. 
		If the attributes are numeric, we can represent a \gls{dataset} by 
		a \gls{datamatrix} $\mathbf{D} \in \mathbb{R}^{\samplesize \times \nrfeatures}$, 
		where each of the $\samplesize$ rows corresponds to a single \gls{datapoint},
		and each of the $\nrfeatures$ columns represents a specific attribute \cite{Everitt2022}.
		\\
		See also: \gls{data}, \gls{datapoint}. }, 
	first={tabular data},
	text={tabular data}
}

\newglossaryentry{datamatrix}
{name={data matrix},
	description={See \gls{tabulardata}\index{data matrix}.}, 
	first={data matrix},
	firstplural={data matrices},
	plural={data matrices},
	text={data matrix}
}
		
\newglossaryentry{dataset}
{name={dataset},
	description={A\index{dataset} dataset is a set of distinct \glspl{datapoint}. 
		Strictly speaking, a dataset is an unordered collection of \glspl{datapoint} 
		that does not contain any repetitions. However, in \gls{ml} literature, the 
		term dataset is often used as a synonym for \gls{sample}, i.e., 
		a \gls{sequence} (or finite list) of \glspl{datapoint} that may contain repetitions. 
		\gls{ml} methods use datasets for \gls{model} \gls{training} and \gls{validation}. 
		The notion of a dataset is broad, i.e., \glspl{datapoint} may represent concrete 
		physical entities (such as humans or animals) or abstract objects (such as numbers). 
		For illustration, Fig.\ \ref{fig_cows_dataset_dict} depicts a dataset whose 
		\glspl{datapoint} are cows.	
		\begin{figure}[H]
			\begin{center}
			\label{fig:cowsintheswissalps_dict}
			\includegraphics[width=0.5\textwidth]{assets/CowsAustria.jpg}
		  	\end{center}
			\caption{\label{fig_cows_dataset_dict}Cow herd somewhere in the Alps.}
	 	\end{figure}
		Quite often, an \gls{ml} engineer does not have direct access to the underlying dataset. 
		For instance, accessing the dataset in Fig.\ \ref{fig_cows_dataset_dict} would require 
		visiting the cow herd. In practice, we work with a more convenient 
		representation (or approximation) of the dataset. Various mathematical \glspl{model} 
		have been developed for this purpose \cite{silberschatz2019database}, \cite{abiteboul1995foundations}, 
		\cite{hoberman2009data}, \cite{ramakrishnan2002database}. One of the most widely used is 
		the \gls{relational model}, which organizes \gls{data} as a table (or relation) 
		\cite{silberschatz2019database}, \cite{codd1970relational}. A table consists of 
		rows and columns, where each row corresponds to a single \gls{datapoint}, and each 
		column represents a specific attribute of a \gls{datapoint}. 
		\gls{ml} methods typically interpret these attributes as \glspl{feature} 
		or as a \gls{label} of a \gls{datapoint}. As an illustration, 
		Table~\ref{tab:cowdata_dict} shows a relational representation of the 
		dataset from Fig.\ \ref{fig_cows_dataset_dict}. In the \gls{relational model}, 
		the order of rows is immaterial, and each attribute (i.e., column) is associated with a 
		\gls{domain} that specifies the set of admissible values. In \gls{ml} applications, 
		these attribute \glspl{domain} correspond to the \gls{featurespace} and the \gls{labelspace}.
		\begin{table}[H]
			\refstepcounter{table}
			\caption*{
				\centering 
				\scshape TABLE \thetable \\[0.5ex]
				\scshape Relation (or Table) Representing the Dataset in Fig.\ \ref{fig_cows_dataset_dict} 
			}
			\label{tab:cowdata_dict} 
			\centering
			\begin{tabular}{lcccc}
				\hline
				\textbf{Name} & \textbf{Weight} & \textbf{Age} & \textbf{Height} & \textbf{Stomach temperature} \\
				\hline
				Zenzi & 100 & 4 & 100 & 25 \\
				Berta & 140 & 3 & 130 & 23 \\
				Resi  & 120 & 4 & 120 & 31 \\
				\hline
			\end{tabular}
		\end{table}
 		While the \gls{relational model} is useful for the study of many \gls{ml} applications, 
		it may be insufficient regarding the requirements for \gls{trustAI}. Modern 
 		approaches like datasheets for datasets provide more comprehensive 
 		documentation, including details about the \gls{data} collection process, intended 
 		use, and other contextual information \cite{DatasheetData2021}.
 		\\
		See also: \gls{datapoint}, \gls{sample}, \gls{data}, \gls{feature}, \gls{featurespace}, \gls{labelspace}.},
	first={dataset},
	plural={datasets},
	text={dataset}  
}

\newglossaryentry{predictor}
{name={predictor},
	description={A\index{predictor} predictor is a real-valued \gls{hypothesis} \gls{map}. 
		Given a \gls{datapoint} with \glspl{feature} $\featurevec$, the value 
		$\hypothesis(\featurevec) \in \mathbb{R}$ is used as a \gls{prediction} for the true 
		numeric \gls{label} $\truelabel \in \mathbb{R}$ of the \gls{datapoint}.
				\\
		See also: \gls{hypothesis}, \gls{map}, \gls{datapoint}, \gls{feature}, \gls{prediction}, \gls{label}. },
	first={predictor},
	text={predictor}  
}

\newglossaryentry{labeled datapoint}
{name={labeled data point},
 	description={A labeled \gls{datapoint}\index{labeled data point} is a \gls{datapoint} whose \gls{label} is known 
		or has been determined by some means that might require human labor.
			\\
		See also: \gls{datapoint}, \gls{label}.},
 	first={labeled data point},
	plural={labeled data points},
	firstplural={labeled data points},
 	text={labeled data point}  
}
 
\newglossaryentry{samplespace}
{name={sample space}, 
  	description={A\index{sample space} \gls{sample} space is the set of all possible 
		\glspl{outcome} of a \gls{randomexperiment} \cite{BillingsleyProbMeasure}, 
		\cite{BertsekasProb}, \cite{AshProbMeasure}, \cite{papoulis}. 
		\\
 		See also: \gls{probspace}.},  
  	first={sample space}, 
 	firstplural={sample spaces},
 	plural={sample spaces},
  	text={sample space}
} 
	
\newglossaryentry{realization}
{name={realization},
	description={Consider\index{realization} an \gls{rv} $\featurevec$ that maps each \gls{outcome} 
		$\outcome \in \samplespace$ of a \gls{probspace} to an element $a$ of a \gls{measurable} space $\mathcal{N}$ 
		\cite{RudinBookPrinciplesMatheAnalysis}, \cite{BillingsleyProbMeasure}, \cite{HalmosMeasure}. 
		A realization of $\featurevec$ is any element $\va \in \mathcal{N}$ such that there exists 
		an element $\outcome' \in \samplespace$ with $\featurevec(\outcome') = \va$.
			\\
		See also: \gls{rv}, \gls{outcome}, \gls{probspace}, \gls{measurable}.}, 
	first={realization},
	plural={realizations},
	text={realization}  
}

\newglossaryentry{trainset}
{name={training set},
	description={A\index{training set} \gls{training} set is a \gls{dataset} $\dataset$ that consists of some \glspl{datapoint} used in \gls{erm} 
		to learn a \gls{hypothesis} $\learnthypothesis$. The average \gls{loss} of $\learnthypothesis$ on the 
		\gls{training} set is referred to as the \gls{trainerr}. The comparison of the \gls{trainerr} with the 
		\gls{valerr} of $\learnthypothesis$ allows us to diagnose the \gls{ml} method and informs how to improve 
		the \gls{valerr} (e.g., using a different \gls{hypospace} or collecting more \glspl{datapoint}) \cite[Sec. 6.6]{MLBasics}.
			\\
		See also: \gls{training}, \gls{dataset}, \gls{datapoint}, \gls{erm}, \gls{hypothesis}, \gls{loss}, \gls{trainerr}, \gls{valerr}, \gls{ml}, \gls{hypospace}.},
	first={training set},
	plural={training sets},
	firstplural={training sets},
	text={training set}  
}

\newglossaryentry{netmodel}
{name={networked model},
sort={networkedmodel},
 	description={A\index{networked model} networked \gls{model} over an \gls{flnetwork} $\graph = \pair{\nodes}{\edges}$ assigns 
   		a \gls{localmodel} (i.e., a \gls{hypospace}) to each node $\nodeidx \in \nodes$ of the \gls{flnetwork} $\graph$.
   		\\
		See also: \gls{model}, \gls{flnetwork}, \gls{localmodel}, \gls{hypospace}.}, 
   first={networked model},
   text={networked model}  
}

\newglossaryentry{batch}
{name={batch},
	description={In\index{batch} the context of \gls{stochGD}, a batch refers to a randomly 
		chosen subset of the overall \gls{trainset}. We use the \glspl{datapoint} in this subset 
		to estimate the \gls{gradient} of \gls{trainerr} and, in turn, to update the \glspl{modelparam}.
		The number of \glspl{datapoint} in a batch is the \emph{batch size} $\batchsize$,
		which is a \gls{hyperparameter} that controls a trade-off in \gls{training}.
		A small batch size yields a noisy \gls{gradient} estimate but allows
		frequent \gls{modelparam} updates and can help escape local \glspl{minimum}.
		A large batch size yields a more accurate \gls{gradient} estimate and
		enables greater \gls{dataparallelism} across \glspl{device}, but requires
		more memory per update and may slow \gls{convergence} per \gls{iteration}.
			\\
		See also: \gls{stochGD}, \gls{trainset}, \gls{datapoint}, \gls{gradient}, 
		\gls{trainerr}, \gls{modelparam}, \gls{hyperparameter}, \gls{dataparallelism}.}, 
 	first={batch},
 	firstplural={batches}, 
 	plural={batches}, 
 	text={batch}  
}

\newglossaryentry{epoch}
{name={epoch},
	description={An epoch\index{epoch} represents one complete pass of the entire \gls{trainset} through some learning 
		\gls{algorithm}. It refers to the point at which a \gls{model} has processed every \gls{datapoint} in the \gls{trainset} once. 
		\Gls{training} a \gls{model} usually requires multiple epochs, since each \gls{iteration} allows the \gls{model} to refine the 
		\glspl{parameter} and improve \glspl{prediction}. The number of epochs is something predefined by the user,  
		and thus a \gls{hyperparameter}, which plays a crucial role in determining how the \gls{model} will generalize to unseen \gls{data}. 
		Too few epochs will result in \gls{underfitting}, while too many epochs can result in \gls{overfitting}.
		\\
		See also: \gls{trainset}, \gls{algorithm}, \gls{model}, \gls{datapoint}, \gls{parameter}, \gls{prediction}, \gls{underfitting}, \gls{overfitting}.},
	first={epoch},
	text={epoch}
} 

\newglossaryentry{netdata}
{name={networked data},
sort={networkeddata},
	description={Networked\index{networked data} \gls{data} consist of \glspl{localdataset} 
		that are related by some notion of pairwise similarity. We can represent networked 
		\gls{data} using a \gls{graph} whose nodes carry \glspl{localdataset} and whose edges encode 
		pairwise similarities. An example of networked \gls{data} can be found in \gls{fl} applications 
		where \glspl{localdataset} are generated by spatially distributed \glspl{device}.
			\\
		See also: \gls{data}, \gls{localdataset}, \gls{graph}, \gls{fl}, \gls{device}.}, 
	first={networked data},
	text={networked data}  
}

\newglossaryentry{trainerr}
{name={training error},
	description={\Gls{training} error is the\index{training error} average \gls{loss} of a \gls{hypothesis} when 
		predicting the \glspl{label} of the \glspl{datapoint} in a \gls{trainset}. 
		We sometimes also refer to \gls{training} error as the minimal average \gls{loss} 
		that is achieved by a solution of \gls{erm}.
				\\
		See also: \gls{loss}, \gls{hypothesis}, \gls{label}, \gls{datapoint}, \gls{trainset}, \gls{erm}.},
	first={training error},
	text={training error}  
}

\newglossaryentry{datapoint}
{name={data point},
	description={A\index{data point} \gls{data} point is any object that conveys information~\cite{coverthomas}. 
		Examples include students, radio signals, trees, images, \glspl{rv}, real numbers, 
 		or proteins. We describe \gls{data} points of the same type by two categories 
		of properties. The first category includes \glspl{feature} that are \gls{measurable} or 
		computable properties of a \gls{data} point. These attributes can be automatically extracted 
		or computed using sensors, computers, or other \gls{data} collection systems. For a \gls{data} 
		point that represents a patient, one \gls{feature} could be the body weight.
		The second category includes \glspl{label} that are higher level facts (or quantities of interest)—that is, 
		facts that typically require human expertise or domain knowledge to determine rather than being directly 
		\gls{measurable}—associated with the \gls{data} point. For a \gls{data} point that represents a patient, 
		a cancer diagnosis provided by a physician would serve as the \gls{label}. 
		Fig.\ \ref{fig:datapoint_cowherd_dict} depicts an image as an example of a \gls{data} 
		point along with its \glspl{feature} and \glspl{label}. Importantly, what constitutes 
		a \gls{feature} or a \gls{label} is not inherent to the \gls{data} point itself—it is a design 
		choice that depends on the specific \gls{ml} application.
		\begin{figure}[H]
    		\centering
    			\begin{minipage}[t]{0.95\textwidth}
        			\centering
        			\includegraphics[width=\textwidth]{assets/CowsAustria.jpg}
        			\caption*{A single \gls{data} point.}
        			\vspace{5mm}
    			\end{minipage}
    			\begin{minipage}[t]{0.95\textwidth}
        			\Glspl{feature}:
        			\begin{itemize}
            			\item $x_{1}, \,\ldots, \,x_{\nrfeatures}$: Color intensities of all image pixels.
            			\item $x_{\nrfeatures+1}$: Time-stamp of the image capture.
            			\item $x_{\nrfeatures+2}$: Spatial location of the image capture.
			\end{itemize}
			\Glspl{label}:
            		\begin{itemize}
               	 		\item $\truelabel_{1}$: Number of cows depicted. 
                			\item $\truelabel_{2}$: Number of wolves depicted. 
                			\item $\truelabel_{3}$: Condition of the pasture (e.g., healthy, overgrazed).
            		\end{itemize}
    			\end{minipage}
    		\caption{Illustration of a \gls{data} point consisting of an image. We can use 
			different properties of the image as \glspl{feature} and higher level facts
			about the image as \glspl{label}.\label{fig:datapoint_cowherd_dict}}
		\end{figure}
 		The distinction between \glspl{feature} and \glspl{label} is not always clear-cut. 
 		A property that is considered a \gls{label} in one setting (e.g., a cancer diagnosis) 
 		may be treated as a \gls{feature} in another setting—particularly if reliable automation (e.g., 
 		via image analysis) allows it to be computed without human intervention.
   		\Gls{ml} broadly aims to predict the \gls{label} of a \gls{data} point based on its \glspl{feature}. 
				\\
		See also: \gls{data}, \gls{feature}, \gls{label}, \gls{dataset}.}, 
	first={data point},
	plural={data points},
	firstplural={data points},
	text={data point}  
}

\newglossaryentry{valerr}
{name={validation error},
 	description={Consider\index{validation error} a \gls{hypothesis} $\learnthypothesis$ that is 
 		obtained by some \gls{ml} method, e.g., using \gls{erm} on a \gls{trainset}. The average \gls{loss} 
 		of $\learnthypothesis$ on a \gls{valset}, which is different from the \gls{trainset}, is referred 
 		to as the \gls{validation} error.
			\\
		See also: \gls{hypothesis}, \gls{ml}, \gls{erm}, \gls{trainset}, \gls{loss}, \gls{valset}, \gls{validation}.},
	first={validation error},
	plural={validation errors},
	firstplural={validation errors},
	text={validation error}  
}

\newglossaryentry{validation} 
{name={validation},
	description={Consider\index{validation} a \gls{hypothesis} $\learnthypothesis$ 
		that has been learned via some \gls{ml} method, e.g., by solving \gls{erm} 
	   	on a \gls{trainset} $\dataset$. 
		Validation refers to the process of evaluating the \gls{loss} incurred by the 
		\gls{hypothesis} $\learnthypothesis$ on a set of 
		\glspl{datapoint} that are not contained in the \gls{trainset} $\dataset$. This 
		set of \glspl{datapoint} is called the \gls{valset}. The average \gls{loss} of 
		$\learnthypothesis$ on the \gls{valset} is referred to as the \gls{valerr}. 
		An example of validation is shown in Fig.\ \ref{fig_validation_dict}.
		\begin{figure}[H] 
			\centering
			\begin{tikzpicture}[scale=1.2,x=1.5cm]
				\draw[thick, domain=-0.5:2.5, samples=10, smooth]
				plot (\x,{0.5*\x*\x})
				node[pos=0, above left]
				{$\learnthypothesis(\feature)$};
				\fill[blue] (0,0) circle (4pt);
				\fill[blue] (2,2) circle (4pt);
				\node[below right] at (0,0) {\gls{trainset}};
				\fill[red] (1,3) circle (4pt);
				\node[above,yshift=2pt] at (1,3) {\gls{valset}};
			\end{tikzpicture}
		\caption{Illustration of validation. The blue points represent the \glspl{datapoint} in the 
			\gls{trainset}, while the red point represents a \gls{datapoint} in the \gls{valset}. The 
			\gls{hypothesis} $\learnthypothesis$ (black curve) fits the \glspl{datapoint} in the \gls{trainset} perfectly, 
			but incurs a large \gls{loss} on the \gls{datapoint} in the \gls{valset}. \label{fig_validation_dict}}
		\end{figure} 
		See also: \gls{trainset}, \gls{valset}, \gls{valerr}, \gls{overfitting}, \gls{generalization}, \gls{kfoldcv}, \gls{loo}.},
 first={validation},
 text={validation}  
}

\newglossaryentry{quadfunc}
{name={quadratic function},
	description={A quadratic \gls{function}\index{quadratic function} is a \gls{function} $f: \mathbb{R}^{\nrfeatures} \rightarrow \mathbb{R}$ 
		of the following form: 
		$$f(\weights) =  \weights\,^{T} \mathbf{Q} \mathbf{w} + \mathbf{q}\,^{T} \weights+a$$ with 
		some \gls{matrix} $\mQ \in \mathbb{R}^{\nrfeatures \times \nrfeatures}$, \gls{vector} $\vq \in \mathbb{R}^{\nrfeatures}$, 
		and scalar $a \in \mathbb{R}$.
		\\
		See also: \gls{function}, \gls{matrix}, \gls{vector}. },
	first={quadratic function},
	text={quadratic function}  
}

\newglossaryentry{valset}
{name={validation set},
 	description={A \gls{validation} set\index{validation set} refers to a set of \glspl{datapoint} used to estimate 
  		the \gls{risk} of a \gls{hypothesis} $\learnthypothesis$ that has been learned by some 
  		\gls{ml} method (e.g., solving \gls{erm}). The average \gls{loss} of $\learnthypothesis$ 
  		on the \gls{validation} set is referred to as the \gls{valerr} and can be used to diagnose an 
  		\gls{ml} method (see \cite[Sec. 6.6]{MLBasics}). The comparison between \gls{trainerr} 
  		and \gls{valerr} can inform directions for the improvement of the \gls{ml} method (such as 
  		using a different \gls{hypospace}).
			\\
		See also: \gls{validation}, \gls{datapoint}, \gls{risk}, \gls{hypothesis}, \gls{loss}, \gls{valerr}, \gls{trainerr}, \gls{kfoldcv}, \gls{loo}.},
	first={validation set},
	text={validation set},
	plural={validation sets}
}

\newglossaryentry{testset}
{name={test set},
	description={A test set\index{test set} refers to a set of \glspl{datapoint} that have  
		been used neither to train a \gls{model} (e.g., via \gls{erm}) nor 
		to choose between different \glspl{model} in a \gls{valset}.
				\\
		See also: \gls{datapoint}, \gls{model}, \gls{erm}, \gls{valset}.},
	first={test set},
	text={test set}  
}

\newglossaryentry{modelsel}
{name={model selection},
sort={modelselection},
	description={In\index{model selection} \gls{ml}, \gls{model} selection refers to the 
		process of choosing between different candidate \glspl{model}. In its most 
		basic form, \gls{model} selection amounts to the following steps \cite[Ch. 6]{MLBasics}: 
		\begin{enumerate}[label=\arabic*)]
			\item \gls{training} each candidate \gls{model}; 
			\item computing the \gls{valerr} for each trained \gls{model}; 
			\item choosing the \gls{model} with the smallest \gls{valerr}. 
		\end{enumerate}
		See also: \gls{ml}, \gls{model}, \gls{training}, \gls{valerr}.},
	first={model selection},
	text={model selection}  
}

\newglossaryentry{linclass}
{name={linear classifier}, 
	description={Consider\index{linear classifier} \glspl{datapoint} characterized by numeric \glspl{feature} $\featurevec \in \mathbb{R}^{\nrfeatures}$ 
	    	and a \gls{label} $\truelabel \in \labelspace$ from some finite \gls{labelspace} $\labelspace$. 
		A linear \gls{classifier} is characterized by having \glspl{decisionregion} that are 
		separated by \glspl{hyperplane} in $\mathbb{R}^{\featuredim}$ \cite[Ch. 2]{MLBasics}.
				\\
		See also: \gls{datapoint}, \gls{feature}, \gls{label}, \gls{labelspace}, \gls{classifier}, \gls{decisionregion}.},
	first={linear classifier},
	text={linear classifier} 
}

\newglossaryentry{erm}
{name={empirical risk minimization (ERM)}, 
	description={ERM\index{empirical risk minimization (ERM)} is the \gls{optproblem} of 
		selecting a \gls{hypothesis} $\learnthypothesis \in \hypospace$ that minimizes the 
		average \gls{loss} (or \gls{emprisk}) on a \gls{trainset} $\dataset$. 
		The \gls{hypothesis} is chosen from a \gls{hypospace} (or \gls{model}) $\hypospace$. 
		The \gls{dataset} $\dataset$ is referred to as \gls{trainset}. A plethora of ERM-based 
		\gls{ml} methods is obtained for different design choices for the \gls{dataset}, \gls{model}, 
		and \gls{loss} \cite[Ch. 3]{MLBasics}. Fig.\ \ref{fig_erm_dict} illustrates ERM for a \gls{linmodel} 
		and \glspl{datapoint} that are characterized by a single \gls{feature} $\feature$ and a \gls{label} $\truelabel$. 
		The \gls{hypothesis} $\hypothesis$ is a \gls{linearmap} that predicts the \gls{label} 
		of a \gls{datapoint} as a linear \gls{function} of its \gls{feature} $\feature$, i.e., 
		$\hypothesis(\feature) = \weight_{1} \feature + \weight_{0}$, where $\weight_{1}$ 
		and $\weight_{0}$ are the \glspl{modelparam} of the \gls{hypothesis} $\hypothesis$. 
		The ERM problem is to find the \glspl{modelparam} $\weight_{1}$ and $\weight_{0}$ that 
		minimize the average \gls{loss} (or \gls{emprisk}) incurred by the \gls{hypothesis} 
		$\hypothesis$ on the \gls{trainset} $\dataset$.
    		\begin{figure}[H]
			\begin{center}
			\begin{tikzpicture}[scale=1]
  			\def\slope{0.4}
 	 		\def\intercept{2.0}
  			\def\xmin{0}
 		 	\def\xmax{6.5}
  			\draw[color=black, thick, dashed, domain=\xmin:\xmax, variable=\x] plot ({\x},{\slope*\x + \intercept}); 
   			\coordinate (hend) at (\xmax,{\slope*\xmax + \intercept});
			\node[above right] at (hend) {$\hypothesis(\feature)$};
   			\foreach \i/\xval/\yval in {1/1.2/1.8, 2/3.0/2.6, 3/5.0/5.7} {
      			\coordinate (l\i) at (\xval, {\slope*\xval + \intercept});
      			\coordinate (n\i) at (\xval, \yval);
      			\node[circle,draw,fill=blue,minimum size=6pt,scale=0.6] (pt\i) at (n\i) {};
      			\draw[<->, color=blue, thick] (l\i) -- (n\i);
  			}
  			\node[anchor=west,below,yshift=-2pt] at (n1) {$\big(\feature^{(1)},\,\truelabel^{(1)}\big)$};
  			\node[anchor=west,below]        at (n2) {$\big(\feature^{(2)},\,\truelabel^{(2)}\big)$};
  			\node[anchor=west,above]   at (n3) {$\big(\feature^{(3)},\,\truelabel^{(3)}\big)$};
			\end{tikzpicture}
		\caption{ERM learns a \gls{hypothesis} $\hypothesis \in \hypospace$ 
			out of a \gls{model} $\hypospace$ by minimizing the average \gls{loss} (or \gls{emprisk}) 
			$1/\samplesize \sum_{\sampleidx=1}^\samplesize \lossfunc{\pair{\featurevec^{(\sampleidx)}}{ \truelabel^{(\sampleidx)}}}				{\hypothesis}$ 
			incurred on a \gls{trainset} $\dataset$.} \label{fig_erm_dict}
			\end{center}
		\end{figure}
		See also: \gls{optproblem}, \gls{loss}, \gls{emprisk}, \gls{trainset}, \gls{optmethod}.},
	first={empirical risk minimization (ERM)},
	text={ERM} 
}

\newglossaryentry{sampleweighting}
{name={sample weighting}, 
	description={Consider\index{sample weighting} an \gls{erm}-based method that 
		learns a \gls{hypothesis} by minimizing the average \gls{loss} on a \gls{trainset}. 
		In its basic form, \gls{erm} treats all \glspl{datapoint} equally important. 
		However, in some applications, it can be useful to put different emphasis on the 
		\gls{prediction} errors obtained for different \glspl{datapoint}. For example, 
		if a \gls{datapoint} is considered an \gls{outlier} we should reduce its influence 
		on the learned \gls{hypothesis}. 
		We can implement this idea by assigning a nonnegative weight 
		$\sampleweight{\sampleidx}$ to each \gls{datapoint} 
		$\pair{\featurevec^{(\sampleidx)}}{\truelabel^{(\sampleidx)}}$ in the \gls{trainset}. 
    		This results in the following weighted \gls{erm} principle:
		\begin{equation}
			\label{equ_weighted_erm_dict}
			\min_{\hypothesis \in \hypospace} \sum_{\sampleidx=1}^{\samplesize} 
			  \sampleweight{\sampleidx} \, \lossfunc{\pair{\featurevec^{(\sampleidx)}}{\truelabel^{(\sampleidx)}}}{\hypothesis}. 
		\end{equation}
		Fig.\ \ref{fig_sampleweighting_dict} illustrates the concept of a 
		\gls{trainset} of three \glspl{datapoint} that contribute unequally to the \gls{emprisk}.
    		\begin{figure}[H]
    			\begin{center}
        			\begin{tikzpicture}[yscale=0.3]
      			\def\slope{1.55}
      			\def\intercept{-2.05}
      			\draw[thick,dashed,domain=0:6.5,variable=\x]
        			plot({\x},{\slope*\x+\intercept});
      			\node[] at (6.5,{\slope*6.5+\intercept}) {$\hypothesis(\feature)$};
      			\foreach \i/\xval/\yval/\scaleval in {1/1.2/1.8/0.6, 2/3.0/2.6/1.0, 3/5.0/5.7/1.5} {
        			\node[circle,draw,fill=blue!60,minimum size=6pt,scale=\scaleval]
          		(pt\i) at (\xval,\yval) {};
        			\draw[<->,color=blue!70] (\xval,{\slope*\xval+\intercept}) -- (pt\i);
      			}
      			\node[below left] at (pt1) {small $\sampleweight{1}$};
      			\node[below, yshift=-5pt]  at (pt2) {medium $\sampleweight{2}$ };
      			\node[below right]at (pt3) {large $\sampleweight{3}$};
    			\end{tikzpicture}
      		\caption{\Gls{sample} weighting assigns each \gls{datapoint} of a \gls{trainset} 
	  		a weight $\sampleweight{\sampleidx}$. Assigning a small weight 
	  		(such as $\sampleweight{1}$ in this example) to a 
	  		\gls{datapoint} decreases its influence on the \gls{hypothesis} 
	 		learned via solving \eqref{equ_weighted_erm_dict}.}
      			\label{fig_sampleweighting_dict}
    			\end{center}
    		\end{figure}
		See also: \gls{erm}, \gls{outlier}, \gls{adaboost}.},
	first={sample weighting}, 
	text={sample weighting}
}

\newglossaryentry{multilabelclass}
{name={multi-label classification}, 
	description={Multi-\gls{label} 
		\gls{classification}\index{multi-label classification} problems and methods use \glspl{datapoint} 
		that are characterized by several \glspl{label}. As an example, consider a \gls{datapoint} 
		representing a picture with two \glspl{label}. One \gls{label} indicates the presence of a human 
		in this picture and another \gls{label} indicates the presence of a car.
				\\
		See also: \gls{label}, \gls{classification}, \gls{datapoint}.},
	    first={multi-label classification},
	    text={multi-label classification} 
}

\newglossaryentry{inference}
{name={inference}, 
	description={In the context of \gls{ml}, inference\index{inference} refers to the process 
		of evaluating a learned \gls{hypothesis} (or trained \gls{model}) $\learnthypothesis(\featurevec)$ 
		based on the \glspl{feature} of a \gls{datapoint} \cite{BishopBook}, \cite{Murphy2012}. 
		A basic \gls{ml} workflow starts with \gls{model} \gls{training} and then 
		uses the trained \gls{model} for inference. 
				\\
		See also: \gls{model}, \gls{loss}, \gls{erm}.},
	first={inference},
	text={inference}
}

\newglossaryentry{training}
{name={training}, 
	description={In the context of \gls{ml}, training\index{training} refers to the process 
		of learning a useful \gls{hypothesis} $\learnthypothesis$ out of a \gls{model} $\hypospace$. 
		The training of a \gls{model} $\hypospace$ is guided by the \gls{loss} incurred on a set 
		of \glspl{datapoint}, which serve as the \gls{trainset}. For \glspl{parammodel}, 
		where each \gls{hypothesis} $\hypothesis^{(\weights)}$ is characterized by a specific 
		choice for the \glspl{modelparam}, training amounts to finding an optimal choice for the 
		\glspl{modelparam} $\weights$. A widely-used approach to training is \gls{erm}, which 
		learns a \gls{hypothesis} by minimizing the average \gls{loss} incurred on a \gls{trainset}. 
		One of the main challenges in \gls{ml} is to control the \gls{discrepancy} between the 
		\gls{loss} incurred on the \gls{trainset} and the \gls{loss} incurred on other (unseen) \glspl{datapoint}.
				\\
		See also: \gls{model}, \gls{loss}, \gls{erm}.},
	first={training},
	text={training}, 
	user1={trained}
}

\newglossaryentry{ssl}
{name={semi-supervised learning (SSL)}, 
	description={SSL\index{semi-supervised learning (SSL)} methods use unlabeled \glspl{datapoint}
		to support the learning of a \gls{hypothesis} from \glspl{labeled datapoint} \cite{SemiSupervisedBook}. 
		This approach is particularly useful for \gls{ml} applications that offer a large number of 
		unlabeled \glspl{datapoint}, but only a limited number of \glspl{labeled datapoint}.
			\\
		See also: \gls{datapoint}, \gls{hypothesis}, \gls{labeled datapoint}, \gls{ml}.}, 
	first={semi-supervised learning (SSL)},
	text={SSL} 
}
	
\newglossaryentry{objfunc}
{name={objective function},
	description={An\index{objective function} objective \gls{function} is a \gls{map} that assigns a numeric 
		objective value $f(\weights)$ to each choice $\weights$ of some variable that we want to 
		optimize (see Fig.\ \ref{fig_obj_func_dict}). In the context of \gls{ml}, the \gls{optimization} variable could 
		be the \glspl{modelparam} of a \gls{hypothesis} $\hypothesis^{(\weights)}$. 
		Common objective \glspl{function} include the \gls{risk} (i.e., expected \gls{loss}) or the \gls{emprisk} 
		(i.e., average \gls{loss} over a \gls{trainset}). \gls{ml} methods apply \gls{optimization} 
		techniques, such as \glspl{gdmethod}, to find the choice $\weights$ with the 
		optimal value (e.g., the \gls{minimum} or the \gls{maximum}) of the objective \gls{function}.
		\begin{figure}[H]
			\begin{center}
			\begin{tikzpicture}[scale=1.0]
				\draw[->] (-0.5,0) -- (4.5,0) node[right] {$\weights$};
				\draw[->] (0,-0.5) -- (0,3.5);
				\draw[thick,domain=0.3:4,smooth,variable=\x] 
				plot ({\x}, {0.5*(\x-2)^2 + 0.5});
				\node at (3.5,2.8) {$f(\weights)$};
			\end{tikzpicture} 
			\end{center}
		\caption{Objective \gls{function} mapping each possible value $\weights$ of an 
			\gls{optimization} variable, such as the \glspl{modelparam} of an \gls{ml} \gls{model}, 
			to a value $f(\weights)$ that measures the usefulness of $\weights$.\label{fig_obj_func_dict}}
		\end{figure} 
		See also: \gls{loss}, \gls{emprisk}, \gls{erm}, \gls{optproblem}.},
	first={objective function},
	plural={objective functions}, 
	firstplural={objective functions}, 
	text={objective function} 
}
	
\newglossaryentry{regularizer}
{name={regularizer}, 
	description={A regularizer\index{regularizer} 
		assigns each \gls{hypothesis} $\hypothesis$ from a \gls{hypospace} $\hypospace$ a quantitative 
		\gls{measure} $\regularizer{\hypothesis}$ conveying to what extent its \gls{prediction} errors might differ 
		on \glspl{datapoint} on and outside a \gls{trainset}. \Gls{ridgeregression} 
		uses the regularizer $\regularizer{\hypothesis} \defeq \normgeneric{\weights}{2}^{2}$ for linear \gls{hypothesis} 
		\glspl{map} $\hypothesis^{(\weights)}(\featurevec) \defeq \weights\,^{T} \featurevec$ \cite[Ch. 3]{MLBasics}. 
		\Gls{lasso} uses the regularizer $\regularizer{\hypothesis} \defeq \normgeneric{\weights}{1}$ 
		for linear \gls{hypothesis} \glspl{map} $\hypothesis^{(\weights)}(\featurevec) \defeq \weights\,^{T} \featurevec$ \cite[Ch. 3]{MLBasics}.
				\\
		See also: \gls{ridgeregression}, \gls{lasso}, \gls{loss}, \gls{objfunc}. },
	first={regularizer},
	text={regularizer} 
}

\newglossaryentry{regularization}
{name={regularization}, 
	description={A\index{regularization} key challenge of modern \gls{ml} applications is that they often 
		use large \glspl{model}, which have an \gls{effdim} in the order of billions. 
		\Gls{training} a high-dimensional \gls{model} using basic \gls{erm}-based methods
		is prone to \gls{overfitting}, i.e., the learned \gls{hypothesis} performs well on the \gls{trainset} 
		but poorly outside the \gls{trainset}. Regularization refers to modifications of a given instance 
		of \gls{erm} in order to avoid \gls{overfitting}, i.e., to ensure that the learned \gls{hypothesis} does 
		not perform much worse outside the \gls{trainset}. There are three routes for implementing 
		regularization: 
		\begin{enumerate}[label=\arabic*)]
			\item {\Gls{model} pruning:} We prune the original \gls{model} $\hypospace$ to obtain a 
			smaller \gls{model} $\hypospace'$. For a \gls{parammodel}, the pruning can be 
			implemented via constraints on the \glspl{modelparam} (such as $w_{1} \in [0.4,0.6]$ for 
			the \gls{weight} of \gls{feature} $x_{1}$ in \gls{linreg}).
			\item {\Gls{loss} penalization:} We modify the \gls{objfunc} of \gls{erm} by adding a 
			\gls{penaltyterm} to the \gls{trainerr}. The \gls{penaltyterm} estimates how much higher the 
			expected \gls{loss} (or \gls{risk}) is compared with the average \gls{loss} on the \gls{trainset}. 
			\item {\Gls{dataaug}:} We can enlarge the \gls{trainset} $\dataset$ by adding 
			perturbed copies of the original \glspl{datapoint} in $\dataset$. One example for such 
			a perturbation is to add the \gls{realization} of an \gls{rv} to the \gls{featurevec} 
			of a \gls{datapoint}. 
		\end{enumerate} 
		Fig.\ \ref{fig_equiv_dataaug_penal_dict} illustrates the above three routes to regularization. 
		These routes are closely related and sometimes fully equivalent. \Gls{dataaug} using \glspl{gaussrv} 
		to perturb the \glspl{featurevec} in the \gls{trainset} of \gls{linreg} 
		has the same effect as adding the penalty 
		$\lambda \normgeneric{\weights}{2}^2$ to the \gls{trainerr} (which is nothing but \gls{ridgeregression}). 
        		The decision on which route to use for regularization can be based on the 
        		available computational infrastructure. For example, it might be much easier to 
        		implement \gls{dataaug} than \gls{model} pruning. 
		\begin{figure}[H]
			\begin{center} 
				\begin{tikzpicture}[scale = 1]
					\draw[->, very thick] (0,0.5) -- (7.7,0.5) node[right] {\gls{feature} $\feature$};       
					\draw[->, very thick] (0.5,0) -- (0.5,4.2) node[above] {\gls{label} $\truelabel$};   
					\draw[color=black, thick, dashed, domain = -1: 6.2, variable = \x]  plot ({\x},{\x*0.4 + 2.0}) ;     
					\draw[color=black, thick, dashed, domain = -1: 6.2, variable = \x]  plot ({\x},{\x*0.6 + 2.0}) ;     
					\draw[blue, thick] (5, 4.5) ellipse [x radius=0.2cm, y radius=1cm];
					\node at (5, 5.8) [text=black, font=\small] {$\{ \hypothesis: \hypothesis(x)\!=\!w_{1}x\!+\!w_{0}; w_{1} \in [0.4,0.6]\}$};
					\node at (6.7,4.5) {$\hypothesis(\feature)$};    
					\coordinate (l1)   at (1.2, 2.48);
					\coordinate (l2) at (1.4, 2.56);
					\coordinate (l3)   at (1.7,  2.68);
					\coordinate (l4)   at (2.2, 2.2*0.4+2.0);
					\coordinate (l5) at (2.4, 2.4*0.4+2.0);
					\coordinate (l6)   at (2.7,  2.7*0.4+2.0);
					\coordinate (l7)   at (3.9,  3.9*0.4+2.0);
					\coordinate (l8) at (4.2, 4.2*0.4+2.0);
					\coordinate (l9)   at (4.5,  4.5*0.4+2.0);
					\coordinate (n1)   at (1.2, 1.8);
					\coordinate (n2) at (1.4, 1.8);
					\coordinate (n3)   at (1.7,  1.8);
					\coordinate (n4)   at (2.2, 3.8);
					\coordinate (n5) at (2.4, 3.8);
					\coordinate (n6)   at (2.7,  3.8);
					\coordinate (n7)   at (3.9, 2.6);
					\coordinate (n8) at (4.2, 2.6);
					\coordinate (n9)   at (4.5,  2.6);
					\node at (n1)  [circle,draw,fill=red,minimum size=6pt,scale=0.6, name=c1] {};
					\node at (n2)  [circle,draw,fill=blue,minimum size=6pt, scale=0.6, name=c2] {};
					\node at (n3)  [circle,draw,fill=red,minimum size=6pt,scale=0.6,  name=c3] {};
					\node at (n4)  [circle,draw,fill=red,minimum size=12pt, scale=0.6, name=c4] {};  
					\node at (n5)  [circle,draw,fill=blue,minimum size=12pt,scale=0.6,  name=c5] {};
					\node at (n6)  [circle,draw,fill=red,minimum size=12pt, scale=0.6, name=c6] {};  
					\node at (n7)  [circle,draw,fill=red,minimum size=12pt,scale=0.6,  name=c7] {};
					\node at (n8)  [circle,draw,fill=blue,minimum size=12pt, scale=0.6, name=c8] {};
					\node at (n9)  [circle,draw,fill=red,minimum size=12pt, scale=0.6, name=c9] {};
					\draw [<->] ($ (n7) + (0,-0.3) $)  --  ($ (n9) + (0,-0.3) $) node [pos=0.4, below] {$\sqrt{\regparam}$}; ; 
					\draw[<->, color=red, thick] (l1) -- (c1);  
					\draw[<->, color=blue, thick] (l2) -- (c2);  
					\draw[<->, color=red, thick] (l3) -- (c3);  
					\draw[<->, color=red, thick] (l4) -- (c4);  
					\draw[<->, color=blue, thick] (l5) -- (c5);  
					\draw[<->, color=red, thick] (l6) -- (c6);  
					\draw[<->, color=red, thick] (l7) -- (c7);  
					\draw[<->, color=blue, thick] (l8) -- (c8);  
					\draw[<->, color=red, thick] (l9) -- (c9);  
					\draw[fill=blue] (6.2, 3.7)  circle (0.1cm) node [black,xshift=2.3cm] {original \gls{trainset} $\dataset$};
					\draw[fill=red] (6.2, 3.2)  circle (0.1cm) node [black,xshift=1.3cm] {augmented};
					\node at (4.6,1.2)  [minimum size=12pt, font=\fontsize{12}{0}\selectfont, text=blue] {$\frac{1}{\samplesize} \sum\limits_{\sampleidx=1}^\samplesize \lossfunc{\pair{\featurevec^{(\sampleidx)}}{ \truelabel^{(\sampleidx)}}}{\hypothesis}$};
					\node at (7.8,1.2)  [minimum size=12pt, font=\fontsize{12}{0}\selectfont, text=red] {$+\regparam \regularizer{\hypothesis}$};
				\end{tikzpicture}
				\caption{Three approaches to regularization: 1) \gls{dataaug}; 2) \gls{loss} penalization; and 3) \gls{model} 
				pruning (via constraints on \glspl{modelparam}). \label{fig_equiv_dataaug_penal_dict} }
			\end{center}
		\end{figure} 
		See also: \gls{overfitting}, \gls{dataaug}, \gls{validation}, \gls{ridgeregression}, \gls{lasso}, \gls{modelsel}.},
	first={regularization},
	text={regularization} 
}

\newglossaryentry{rerm}
{name={regularized empirical risk minimization (RERM)}, 
	description={Basic \gls{erm} learns a \gls{hypothesis} (or trains a \gls{model}) $\hypothesis \in \hypospace$ 
		based solely on the \gls{emprisk} $\emprisk{\hypothesis}{\dataset}$ incurred on a \gls{trainset} $\dataset$. 
		To make \gls{erm} less prone to \gls{overfitting}, we can implement \gls{regularization} by 
		including a (scaled) \gls{regularizer} $\regularizer{\hypothesis}$ in the learning objective. 
		This leads to RERM\index{regularized empirical risk minimization (RERM)} such that
		\begin{equation}
			\label{equ_def_rerm_dict}
			\learnthypothesis \in \argmin_{\hypothesis \in \hypospace} \emprisk{\hypothesis}{\dataset} + \regparam \regularizer{\hypothesis}.
		\end{equation}
		The \gls{parameter} $\regparam \geq 0$ controls the \gls{regularization} strength. 
		For $\regparam = 0$, we recover standard \gls{erm} without \gls{regularization}. As $\regparam$ increases, the 
		learned \gls{hypothesis} is increasingly biased toward small values of $\regularizer{\hypothesis}$. 
		The component $\regparam \regularizer{\hypothesis}$ in the \gls{objfunc} of \eqref{equ_def_rerm_dict} 
		can be intuitively understood as a surrogate for the increased average \gls{loss} that may 
		occur when predicting \glspl{label} for \glspl{datapoint} outside the \gls{trainset}. This intuition  
		can be made precise in various ways. For example, consider a \gls{linmodel} trained using \gls{sqerrloss} 
		and the \gls{regularizer} $\regularizer{\hypothesis} = \normgeneric{\weights}{2}^{2}$. 
		In this setting, $\regparam \regularizer{\hypothesis}$ corresponds to the expected increase in \gls{loss} 
		caused by adding \glspl{gaussrv} to the \glspl{featurevec} in the \gls{trainset} 
		\cite[Ch. 3]{MLBasics}.
		A principled construction for the \gls{regularizer} $\regularizer{\hypothesis}$ 
		arises from approximate upper bounds on the \gls{generalization} error. The resulting 
		RERM instance is known as \gls{srm} \cite[Sec. 7.2]{ShalevShwartz2009}.
				\\
		See also: \gls{erm}, \gls{regularization}, \gls{loss}, \gls{srm}.}, 
	first={regularized empirical risk minimization (RERM)},
	text={RERM} 
}

\newglossaryentry{generalization}
{name={generalization},
	description={Generalization\index{generalization} refers to the ability of a \gls{model} trained on a \gls{trainset} to make accurate 
		\glspl{prediction} on new unseen \glspl{datapoint}. This is a central goal of \gls{ml} and \gls{ai}, i.e., 
		to learn patterns that extend beyond the \gls{trainset}. Most \glspl{mlsystem}  
		use \gls{erm} to learn a \gls{hypothesis} $\learnthypothesis \in \hypospace$ by minimizing 
		the average \gls{loss} over a \gls{trainset} of \glspl{datapoint} $\datapoint^{(1)}, \,\ldots, \,\datapoint^{(\samplesize)}$, 
		which is denoted by $\trainset$. However, success on the \gls{trainset} does not guarantee success on 
		unseen \gls{data}—this discrepancy is the challenge of generalization. \\ To study generalization 
		mathematically, we need to formalize the notion of ``unseen'' \gls{data}. A widely used 
		approach is to assume a \gls{probmodel} for \gls{data} generation, such as the \gls{iidasspt}. 
		Here, we interpret \glspl{datapoint} as independent \glspl{rv} with an identical 
		\gls{probdist} $p(\datapoint)$. This \gls{probdist}, which is assumed fixed but unknown, 
		allows us to define the \gls{risk} of a trained \gls{model} $\learnthypothesis$ as the expected \gls{loss}:
		\[
		\risk{\learnthypothesis}=\expect_{\datapoint \sim p(\datapoint)} \big\{ \loss(\learnthypothesis, \datapoint) \big\}.
		\]
		The difference between \gls{risk} $\risk{\learnthypothesis}$ and \gls{emprisk} $\emprisk{\learnthypothesis}{\trainset}$ 
		is known as the \gls{gengap}. Tools from \gls{probability} theory, such as \glspl{concentrationinequ} 
		and uniform \gls{convergence}, allow us to bound this gap under certain conditions \cite{ShalevMLBook}.\\
		Generalization without \gls{probability}: \Gls{probability} theory is one way to study how well a 
		\gls{model} generalizes beyond the \gls{trainset}, but it is not the only way. Another option is to use 
		simple deterministic changes to the \glspl{datapoint} in the \gls{trainset}. The basic idea is that a 
		good \gls{model} $\learnthypothesis$ should be robust, i.e., its \gls{prediction} $\learnthypothesis(\featurevec)$ 
		should not change much if we slightly change the \glspl{feature} $\featurevec$ of a \gls{datapoint} $\datapoint$. 
		For example, an object detector trained on smartphone photos should still detect the object if a few 
		random pixels are masked \cite{OnePixelAttack}. Similarly, it should deliver the same result if we rotate 
		the object in the image \cite{MallatUnderstandingDeepLearning}. See Fig.\ \ref{fig:polynomial_fit_dict} for a 
		visual illustration.
		\begin{figure}[H]
			\centering
			\begin{tikzpicture}[scale=0.8]
			\draw[lightblue, fill=lightblue, opacity=0.5] (3, 2) ellipse (6cm and 2cm);
			\node[black] at (6, 3) {$p(\datapoint)$};
			\fill[blue] (1, 3) circle (4pt) node[below, xshift=0pt, yshift=0pt] {$\datapoint^{(1)}$};
			\fill[blue] (5, 1) circle (4pt) node[below] {$\datapoint^{(2)}$};
			\fill[blue] (1.6, 3) circle (3pt);
			\fill[blue] (0.4, 3) circle (3pt);
			\draw[<->, thin] (1, 3) -- (1.6, 3);
			\draw[<->, thin] (1, 3) -- (0.4, 3);
			\fill[blue] (5.6, 1) circle (3pt);
			\fill[blue] (4.4, 1) circle (3pt);
			\draw[<->, thin] (5, 1) -- (5.6, 1);
			\draw[<->, thin] (5, 1) -- (4.4, 1);
			\draw[black, thick, domain=0:6, smooth] plot (\x, {- 1*\x + 5});
			\node[black] at (3, 2.5) [right] {$\learnthypothesis$};
			\end{tikzpicture}
		\caption{Two \glspl{datapoint} $\datapoint^{(1)},\datapoint^{(2)}$ used as a \gls{trainset} 
			to learn a \gls{hypothesis} $\learnthypothesis$ via \gls{erm}. We can evaluate $\learnthypothesis$ 
			outside $\trainset$ either by an \gls{iidasspt} with some underlying \gls{probdist} $p(\datapoint)$ 
			or by perturbing the \glspl{datapoint}.}
			\label{fig:polynomial_fit_dict}
		\end{figure}
		See also: \gls{erm}, \gls{iidasspt}, \gls{overfitting}, \gls{validation}.},
	first={generalization},
	plural={generalizations}, 
	text={generalization} 
}

\newglossaryentry{nonparametric}
{name={nonparametric}, 
	description={Nonparametric\index{nonparametric} \gls{ml} methods do not assume a fixed 
		\gls{model} structure with a finite number of \glspl{modelparam}. Instead, 
		the complexity of the learned \gls{hypothesis} can grow with the number of \glspl{datapoint} in the	
		\gls{trainset} \cite{MLBasics}, \cite{BishopBook}. Examples of nonparametric 
		\gls{ml} methods include \gls{locallyweightedlearning} and \glspl{GaussProc}. 
				\\
		See also: \gls{ml}, \gls{locallyweightedlearning}, \gls{GaussProc}.}, 
	first={nonparametric},
	text={nonparametric} 
}

\newglossaryentry{knnregression}
{name={$k$-nearest neighbors regression (KNNR)},
sort={knearest},
	description={KNNR\index{$k$-nearest neighbors regression (KNNR)} is a widely used instance of 
		\gls{locallyweightedlearning} for \gls{regression} \cite{MLBasics}, \cite{BishopBook}. 
		To make a \gls{prediction} for a given \gls{datapoint} $\datapoint$, KNNR identifies the 
		$k$ \glspl{datapoint} in the \gls{trainset} that are closest to $\datapoint$ according 
		to some \gls{metric}. The \gls{prediction} is then obtained as the average of the 
		\glspl{label} of these $k$ \glspl{nn}. 
				\\
		See also: \gls{locallyweightedlearning}, \gls{regression}, \gls{datapoint}, \gls{trainset}, \gls{nn}.}, 
	first={$k$-nearest neighbors regression (KNNR)},
	text={KNNR} 
}

\newglossaryentry{locallyweightedlearning}
{name={locally weighted learning (LWL)},
sort={locallyweighted},
	description={LWL\index{locally weighted learning (LWL)} is a \gls{nonparametric} \gls{ml} method 
		that makes \glspl{prediction} for a given \gls{datapoint} $\datapoint$ based on a 
		weighted combination of the \glspl{label} of the \glspl{datapoint} in the \gls{trainset}. 
		The weights reflect the similarity between the \gls{datapoint} $\datapoint$ and each 
		\gls{datapoint} in the \gls{trainset}. A widely used instance of LWL is the
		\gls{knnregression} \cite{MLBasics}, \cite{BishopBook}, \cite{LocallyWeightedLearning}.
				\\
		See also: \gls{nonparametric}, \gls{datapoint}, \gls{trainset}, \gls{knnregression}.}, 
	first={locally weighted learning (LWL)},
	text={LWL} 
}

\newglossaryentry{multilayerperceptron}
{name={multilayer perceptron (MLP)},
	description={An\index{multilayer perceptron} MLP is a type of 
		\gls{ann} that is composed of multiple \glspl{layer} of affine 
		transformations followed by pointwise nonlinear \glspl{actfun} \cite{Goodfellow-et-al-2016}.
		Mathematically, an MLP implements a composition of nonlinear \glspl{operator} of the form
		$\weights \mapsto \actfun(\mA \weights + \vb)$, where $\mA$ and $\vb$ are 
		learnable \glspl{modelparam} and the \gls{actfun} $\actfun$ is applied elementwise. 
		\\
		See also: \gls{ann}, \gls{actfun}.},
	first={multilayer perceptron (MLP)},
	plural={multilayer perceptrons},
	firstplural={multilayer perceptrons (MLPs)},
	text={multilayer perceptron}
}

\newglossaryentry{embedding}
{name={embedding},
	description={An \index{embedding} embedding is a \gls{map} $\hypothesis$ that represents 
		\glspl{datapoint} using the elements (or \glspl{vector}) of a given \gls{vectorspace}. 
		The \gls{map} is constructed such that similar objects are represented by
		similar \glspl{vector}, according to some \gls{metric} in the \gls{vectorspace}. 
		\Gls{ml} methods learn the \gls{map} $\hypothesis \in \hypospace$, out of a \gls{model} $\hypospace$, 
		by optimizing a quantitative \gls{measure} of how well it captures the intrinsic structure of a given 
		\gls{trainset}. For \glspl{datapoint} belonging to a \gls{metricspace}, a widely used approach is to 
		minimize the reconstruction error incurred by the embedding \cite{Goodfellow-et-al-2016}, \cite[Ch. 13]{BishopDeepLearning}.		
		\\
		See also: \gls{vectorspace}, \gls{featlearn}.},
	first={embedding},
	plural={embeddings},
	firstplural={embeddings},
	text={embedding}
}

\newglossaryentry{gnn}
{name={graph neural network (GNN)},
	description={A\index{graph neural network (GNN)} GNN is a 
 		special type of \gls{ann} that is defined via a given 
 		\gls{graph}. In a GNN, node-associated representations are updated 
		by aggregating and transforming \glspl{embedding} of neighboring 
		nodes \cite{DeepLearningBishop}, \cite{GeometricDeepLearning2017}, \cite{SSLGCN}.
		\\
		See also: \gls{ann}, \gls{graph}.},
	first={graph neural network (GNN)},
	plural={GNNs},
	firstplural={graph neural networks (GNNs)},
	text={GNN}
}

\newglossaryentry{gengap}
{name={generalization gap}, 
	description={\Gls{generalization} gap is the difference\index{generalization gap} 
		between the performance of a \gls{hypothesis} $\hypothesis \in \hypospace$ 
	  	on the \gls{trainset} $\trainset$ and its performance 
	  	on \glspl{datapoint} outside $\trainset$. We can make this notion 
	  	precise by using a \gls{probmodel} that allows us to compute 
	  	the \gls{risk} (or expected \gls{loss}) $\risk{\learnthypothesis}$ 
	  	of a \gls{hypothesis} $\hypothesis$. 
	 	 \begin{figure}[H]
			\begin{center}
			\begin{tikzpicture}[x=3cm, y=1cm]
  			\def\hmin{0.2}   
  			\def\hmax{0.8}   
  			\def\ymin{-2.1}  
  			\def\ymax{4.2}   
  			\fill[gray!40, opacity=0.35] (\hmin,\ymin) rectangle (\hmax,\ymax);
  			\draw[gray!60] (\hmin,\ymin) -- (\hmin,\ymax);
  			\draw[gray!60] (\hmax,\ymin) -- (\hmax,\ymax);
  			\node[rotate=0] at ({(\hmin+\hmax)/2}, {\ymax-0.2}) {$\hypospace$};
			\node[anchor=west] at (2, 4) {$\risk{\hypothesis}$};
			\draw[line width=1, domain=-2:1, samples=100,dashed] plot  ({\x+1}, {\x*\x -0.5}) node[right] {$\emprisk{\dataset}{\hypothesis}$};
			\draw[line width=1, domain=-1:2, samples=100] plot ({\x}, {\x*\x});
			\draw[<->, thick] (1, -0.5) -- (1, 1) node[midway, right] {gap};
            		\draw[->, thick] (-1.2,-1) -- (2.2,-1) node[below right] {$\hypothesis$};
			\end{tikzpicture}
			\end{center}
		\caption{\Gls{generalization} gap defined as the difference between the 
			\gls{risk} $\risk{\hypothesis}$ and the average \gls{loss} (or \gls{emprisk}) 
			$\emprisk{\hypothesis}{\trainset}$ computed on a \gls{trainset}.}
		\end{figure}
	  	In practice, the \gls{probdist} underlying this \gls{expectation} is 
	  	unknown. Thus, we need to estimate the \gls{expectation} based on 
	  	observed \glspl{datapoint}. \Gls{validation} techniques 
	  	use different constructions of a \gls{valset}, which is different from 
	  	the \gls{trainset}, to estimate the \gls{generalization} gap.
		\\
		See also: \gls{generalization}, \gls{validation}, \gls{erm}, \gls{lossfunc}.}, 
	first={generalization gap}, 
	text={generalization gap}
}

\newglossaryentry{lda} 
{name={linear discriminant analysis (LDA)}, 
	description={LDA\index{linear discriminant analysis (LDA)} is a classical 
		\gls{featlearn} method \cite{BishopBook}, \cite{Fisher1936Taxonomic}. 
		In the context of \gls{binclass} problems, 
		LDA seeks a linear \gls{featuremap} $\featuremap^{(\weights)}: \mathbb{R}^{\nrfeatures} 
		\rightarrow \mathbb{R}: \featurevec \mapsto \weights^{T} \featurevec$ 
		such that the new \gls{feature} $\featuremap^{(\weights)}(\featurevec)$ 
		optimally allows us to predict the \gls{label} of a \gls{datapoint}. 
		\\
		See also: \gls{featlearn}, \gls{dimred}, \gls{selfsupervisedlearning}. }, 
 	first={linear discriminant analysis (LDA)},
 	text={LDA}, 
 	firstplural={linear discriminant analyses (LDAs)},
 	plural={LDAs}
}

\newglossaryentry{randomprojection} 
{name={random projection}, 
	description={A random \gls{projection}\index{random projection} uses a random \gls{widematrix} 
		$\mathbf{A}\!\in\!\mathbb{R}^{\nrfeatures' \times \nrfeatures}$, with 
		$\nrfeatures' < \nrfeatures$, to map a \gls{featurevec} 
		$\featurevec \!\in\! \mathbb{R}^{\nrfeatures}$ to a shorter 
		\gls{featurevec} $\mathbf{A}\featurevec \!\in\! \mathbb{R}^{\nrfeatures'}$. 
		It is a basic method for \gls{featlearn} and \gls{dimred}. 
		The \gls{projection} \gls{matrix} $\mathbf{A}$ is typically generated entrywise 
		by \gls{iid} \glspl{rv} with a common \gls{probdist} $\probdist$. 
		For a broad class of such \glspl{probdist}, a random \gls{projection} 
		approximately preserves pairwise \glspl{eucliddist} between \glspl{featurevec} 
		of a given finite \gls{dataset}. The \gls{johnsonlindenstrausslemma}
		guarantees the existence of such a distance-preserving \gls{dimred} \gls{map}
		but does not itself involve randomness. Random \glspl{projection} provide a 
		probabilistic construction that realizes this guarantee with high \gls{probability}. 
		Roughly speaking, for many relevant applications, random \glspl{projection} 
		preserve the most relevant information contained in the original (typically 
		very long) \gls{featurevec}. Fig.\ \ref{fig:randomprojection_dict} illustrates this 
		behavior for an RGB image. The left panel shows the original image. The middle panel 
		shows a masked image where a randomly selected five percent of the original pixels are kept, and 
		the remaining pixels are set to a fixed light-gray color. The right panel shows the result of a 
		simple reconstruction based on repeated averaging of nearby retained pixels. 
		\begin{figure}[H]
			\centering
			\begin{minipage}[t]{0.32\textwidth}
				\includegraphics[width=\linewidth]{assets/pythonsnacks/randomprojection/randomprojection_original.png}
				\caption*{original}
				\vspace{2ex}
				\centering
				{\selectfont (a)}
			\end{minipage}\hfill
			\begin{minipage}[t]{0.32\textwidth}
				\includegraphics[width=\linewidth]{assets/pythonsnacks/randomprojection/randomprojection_masked.png}
				\caption*{95 \% masked}
				\vspace{2ex}
				\centering
				{\selectfont (b)}
			\end{minipage}\hfill
			\begin{minipage}[t]{0.32\textwidth}
				\includegraphics[width=\linewidth]{assets/pythonsnacks/randomprojection/randomprojection_reconstructed.png}
				\caption*{reconstructed}
				\vspace{2ex}
				\centering
				{\selectfont (c)}
			\end{minipage}
		\caption{Illustration of a random \gls{projection} in the form of removing (or masking) 
			all image pixels, except those in a small random subset. 
			(a) Left panel showing the original RGB image. (b) Middle panel showing a version 
			with only a random five percent subset of pixels retained. (c) Right panel showing a 
			simple convolution-based reconstruction that diffuses information from 
			the known pixels into masked regions.}
			\label{fig:randomprojection_dict}
		\end{figure}
		See also: \gls{featlearn}, \gls{dimred}, \gls{johnsonlindenstrausslemma}. 
		\\
		Python demo: \href{https://github.com/AaltoDictionaryofML/AaltoDictionaryofML.github.io/blob/main/assets/pythonsnacks/randomprojection/randomprojection.py}{click me}},
	first={random projection},
	plural={random projections}, 
	firstplural={random projections}, 
	text={random projection}
}

\newglossaryentry{boosting}
{name={boosting}, 
	description={Boosting\index{boosting} is an iterative \gls{optmethod} to learn 
		an accurate \gls{hypothesis} \gls{map} (or strong learner) by sequentially 
		combining less accurate \glspl{baselearner} (referred to as weak learners) 
		\cite[Ch. 10]{hastie01statisticallearning}, \cite{pmlr-vR2-ridgeway99a}, \cite{Schapire1999}, \cite{Drucker1997}.
		Boosting can be understood as a \gls{generalization} of \glspl{gdmethod} 
		for \gls{erm} using \glspl{parammodel} and \gls{smooth} \glspl{lossfunc} 
		\cite{Friedman2001}. In particular, starting from an initialization $\widetilde{\hypothesis}$, 
		boosting methods construct a \gls{sequence} of \glspl{hypothesis} 
		$\widetilde{\hypothesis}^{(\iteridx)}$, $\iteridx=1,\,\ldots$, 
		via a generalized \gls{gradstep} 
		$$ \widetilde{\hypothesis}^{(\iteridx)} = \widetilde{\hypothesis}^{(\iteridx-1)}+
		\lrate^{(\iteridx)} 
		\learnthypothesis^{(\iteridx)}.$$ 
		Here, $\lrate^{(\iteridx)}$ denotes a \gls{learnrate} and 
		$\learnthypothesis^{(\iteridx)}$ is provided 
		by the $\iteridx$th \gls{baselearner}. Comparing the above update with the 
		plain \gls{gradstep} suggests that we view $\learnthypothesis^{(\iteridx)}$ 
		as a (negative) generalized \gls{gradient}. 
		Boosting methods differ in their choice of \glspl{baselearner} for computing 
		the generalized \glspl{gradient} $\learnthypothesis^{(\iteridx)}$. 
		\begin{figure}[H]
			\begin{center}
				\begin{tikzpicture}[scale=1.2]
					\draw[->] (-0.5,0) -- (5.5,0) node[right] {$\hypothesis$};
					\draw[->] (0,-0.5) -- (0,4.5) node[above] {$\emprisk{\hypothesis}{\dataset}$};
					\draw[thick,domain=0.2:5,smooth,variable=\x,blue!60] plot ({\x},{(4 - 1.3*\x + 0.15*\x*\x)});
					\foreach \x/\label in {0.7/$\widetilde{\hypothesis}^{(0)}$, 1.5/$\widetilde{\hypothesis}^{(1)}$, 2.3/$\widetilde{\hypothesis}^{(2)}$, 3.0/$\widetilde{\hypothesis}^{(3)}$} {
						\draw[dashed, gray] (\x, 0) -- (\x, {4 - 1.3*\x + 0.15*\x*\x}); 
						\filldraw[black] (\x, {4 - 1.3*\x + 0.15*\x*\x}) circle (2pt);   
						\node[below] at (\x, -0.1) {\label};                             
					}
				\end{tikzpicture}
			\end{center} 
		\caption{Boosting methods construct a \gls{sequence} of \gls{hypothesis} \glspl{map} 
			via a generalized \gls{gradstep}. This generalized \gls{gradstep} uses the \gls{output} of 
			\glspl{baselearner}.\label{fig_boosting_dict}}
     		\end{figure} 
     		See also: \gls{ensemble}, \gls{adaboost}, \gls{gradientboosting}.}, 
	first={boosting}, 
	text={boosting}
} 

\newglossaryentry{rsquared}
{name={$R^2$ (coefficient of determination)}, 
sort={Rtwo},
	description={$R^2$\index{$R^2$ (coefficient of determination)} is a \gls{metric} 
		for assessing a trained \gls{regression} \gls{model}. 
		Formally, it is defined as 
		$R^2 = 1 - \widehat{\varepsilon}/\widehat{v}$, where 
		$\widehat{\varepsilon}$ denotes the average \gls{sqerrloss} 
		on the \gls{trainset} and $\widehat{v}$ is the \gls{samplevariance} 
		of the \glspl{label} in the \gls{trainset} \cite{COLINCAMERON1997329}. 
				 \\ 
		See also: \gls{model}, \gls{mse}, \gls{risk}.}, 
	first={$R^2$ (coefficient of determination)}, 
	text={$R^2$}
}

\newglossaryentry{mse} 
{name={mean squared error (MSE)}, 
	description={The MSE\index{mean squared error (MSE)} of a \gls{hypothesis} 
         	is the average \gls{sqerrloss} computed over a given \gls{dataset}
		\cite{MLBasics}, \cite{hastie01statisticallearning}. 
		When using a \gls{probmodel} for the \gls{data} generation, the 
		term MSE can also refer to the \gls{risk}, i.e., the \gls{expectation} of the 
		\gls{sqerrloss} \cite{LC}, \cite{kay}.
		\\ 
		See also: \gls{sqerrloss}, \gls{risk}.}, 
 	first={mean squared error (MSE)}, 
 	text={MSE}
}

\newglossaryentry{mae} 
{name={mean absolute error (MAE)}, 
	description={The MAE\index{mean absolute error (MAE)} of a \gls{hypothesis} 
        		is the average \gls{abserr} computed over a given \gls{dataset}. 
               	In theoretical analyses, MAE also denotes the expected \gls{abserr}, 
               	i.e., the corresponding \gls{risk}.
		\\ 
		See also: \gls{abserr}, \gls{risk}.}, 
 	first={mean absolute error (MAE)}, 
 	text={MAE}
}

\newglossaryentry{adaboost}
{name={adaptive boosting (AdaBoost)},
	description={AdaBoost\index{adaptive boosting (AdaBoost)} is a specific 
		\gls{boosting} \gls{algorithm} that combines \glspl{baselearner} sequentially 
		\cite{pmlr-vR2-ridgeway99a}, \cite{Schapire1999}, \cite{Drucker1997}. 
	    	The core idea of AdaBoost is to use the \gls{prediction} errors of the current  
		\gls{baselearner} for \gls{sampleweighting} in the next \gls{baselearner}. 
		In particular, the $\iteridx$th \gls{baselearner} learns a \gls{hypothesis} 
		$\learnthypothesis^{(\iteridx)}$ by weighted \gls{erm} with  
		\gls{sample} \glspl{weight} $\sampleweight{\sampleidx}$. The \gls{prediction} errors of 
		$\learnthypothesis^{(\iteridx)}$ are then used to update the 
		\gls{sample} \glspl{weight} by increasing the \glspl{weight} of \glspl{datapoint} 
		that have been predicted poorly (i.e., with large \gls{loss}) 
		by $\learnthypothesis^{(\iteridx)}$. The updated \gls{sample} \glspl{weight} are 
		then used in the next \gls{baselearner} to learn $\learnthypothesis^{(\iteridx+1)}$. 
		The ultimate \gls{hypothesis} $\widetilde{\hypothesis}^{(\nriter)}$ delivered after 
		$\nriter$ \glspl{iteration} is a linear combination of the \glspl{hypothesis} 
		$\learnthypothesis^{(1)},\,\ldots,\,\learnthypothesis^{(\nriter)}$. 
		AdaBoost can be interpreted as a generalized \gls{gradstep} \cite[Ch. 10.4]{hastie01statisticallearning}
		$$\widetilde{\hypothesis}^{(\iteridx)}=\widetilde{\hypothesis}^{(\iteridx-1)}+\lrate^{(\iteridx)} \learnthypothesis^{(\iteridx)}.$$   
		This generalized \gls{gradstep} involves a \gls{learnrate} $\lrate^{(\iteridx)}$, which 
		controls the amount of modification of the current \gls{hypothesis}. 
		\\
		See also: \gls{boosting}, \gls{sampleweighting}, \gls{erm}.}, 
	first={adaptive boosting (AdaBoost)}, 
	text={AdaBoost}
}

\newglossaryentry{gradientboosting}
{name={gradient boosting},
sort={gradientboosting},
 	description={\Gls{gradient} \gls{boosting}\index{gradient boosting} is a \gls{boosting} 
 		\gls{algorithm} that learns a \gls{hypothesis} $\widetilde{\hypothesis}$ 
 		by sequentially combining the \glspl{hypothesis} $\learnthypothesis^{(\iteridx)}$ 
 		\cite[Algorithm 10.3]{hastie01statisticallearning}, \cite{Friedman2001}. 
 		Similar to \gls{adaboost}, \gls{gradient} \gls{boosting} uses a generalized \gls{gradstep} 
 		to combine the results of the \glspl{baselearner}:
        		$$\widetilde{\hypothesis}^{(\iteridx)} = \widetilde{\hypothesis}^{(\iteridx-1)}-\lrate^{(\iteridx)} 
 		\learnthypothesis^{(\iteridx)},$$
 		where the generalized \gls{gradient} $\learnthypothesis{\iteridx}$ is constructed from 
		the $\iteridx$th \gls{baselearner}.
 		The difference between \gls{adaboost} and \gls{gradient} \gls{boosting} is in the construction 
		of $\learnthypothesis{\iteridx}$. While \gls{adaboost} uses weighted \gls{erm} for this construction, 
		\gls{gradient} \gls{boosting} uses \gls{erm} on a modified \gls{trainset}. This modification is obtained 
		by leaving the \glspl{featurevec} untouched but replacing the \glspl{label} with the 
		\gls{partialderivative} of the \gls{lossfunc} with respect to the \glspl{prediction} 
		of the previous \gls{baselearner}.   
 		 \\
 		See also: \gls{boosting}, \gls{adaboost}, \gls{gd}.},
 	first={gradient boosting}, 
 	text={gradient boosting}
}
	
\newglossaryentry{gtv}
{name={generalized total variation (GTV)}, 
	description={GTV is a\index{generalized total variation (GTV)} 
		\gls{measure} of the variation of trained \glspl{localmodel} $\localhypothesis{\nodeidx}$ 
		(or their \glspl{modelparam} $\localparams{\nodeidx}$) assigned to the nodes $\nodeidx=1, \,\ldots, \,\nrnodes$ 
		of an undirected weighted \gls{graph} $\graph$ with edges $\edges$. Given a \gls{measure} $\discrepancy{\hypothesis}{\hypothesis'}$ 
		of the \gls{discrepancy} between \gls{hypothesis} \glspl{map} $\hypothesis,\hypothesis'$, the GTV is 
		\begin{equation} 
			\nonumber
			\sum_{\edge{\nodeidx}{\nodeidx'}\in \edges} \edgeweight_{\nodeidx,\nodeidx'} 
			\discrepancy{\localhypothesis{\nodeidx}}{\localhypothesis{\nodeidx'}}.
		\end{equation}
		Here, $\edgeweight_{\nodeidx,\nodeidx'}>0$ denotes the weight of the undirected edge $\edge{\nodeidx}{\nodeidx'}\in \edges$.
				\\
		See also: \gls{localmodel}, \gls{modelparam}, \gls{graph}, \gls{discrepancy}, \gls{hypothesis}, \gls{map}.},
	first={generalized total variation (GTV)},
	text={GTV} 
}
	
\newglossaryentry{srm}
{name={structural risk minimization (SRM)}, 
	description={SRM\index{structural risk minimization (SRM)} is an
		instance of \gls{rerm} with which the \gls{model} $\hypospace$ can be expressed 
		as a \gls{countable} union of submodels such that $\hypospace = \bigcup_{n=1}^{\infty} \hypospace^{(n)}$. 
		Each submodel $\hypospace^{(n)}$ permits the derivation of an approximate upper bound 
		on the \gls{generalization} error incurred when applying \gls{erm} to train $\hypospace^{(n)}$. 
		These individual bounds—one for each submodel—are then combined to form a \gls{regularizer} 
		used in the \gls{rerm} objective. 
        		These approximate upper bounds (one for each $\hypospace^{(n)}$) are then combined 
		to construct a \gls{regularizer} for \gls{rerm} \cite[Sec.\ 7.2]{ShalevMLBook}.
				\\
		See also: \gls{rerm}, \gls{model}, \gls{generalization}, \gls{erm}, \gls{regularizer}, \gls{risk}.},
	first={structural risk minimization (SRM)},
	text={SRM}
}

\newglossaryentry{rlm}
{name={regularized loss minimization (RLM)},
 	description={See\index{regularized loss minimization (RLM)} \gls{rerm}.},
	first={regularized loss minimization (RLM)}, 
 	text={RLM}
}

\newglossaryentry{clustasspt}
{name={clustering assumption}, 
	description={The\index{clustering assumption} 
		\gls{clustering} assumption postulates that \glspl{datapoint} in a \gls{dataset} form a (small) number of 
		groups or \glspl{cluster}. \Glspl{datapoint} in the same \gls{cluster} are more similar to each 
		other than those outside the \gls{cluster} \cite{SemiSupervisedBook}. We obtain different 
		\gls{clustering} methods by using different notions of similarity between \glspl{datapoint}.
				\\
		See also: \gls{clustering}, \gls{datapoint}, \gls{dataset}, \gls{cluster}.},
	first={clustering assumption},
	text={clustering assumption} 
}

\newglossaryentry{netexpfam}
{name={networked exponential families (nExpFam)}, 
sort={networkedexponential}, 
	description={nExpFam is a\index{networked exponential families (nExpFam)} collection of exponential 
		families, each of them assigned to a node of an \gls{flnetwork}. The \glspl{modelparam} are coupled 
	   	via the network structure by requiring them to have a small \gls{gtv} \cite{JungNetExp2020}.
	   		\\
		See also: \gls{flnetwork}, \gls{modelparam}, \gls{gtv}.},
	first={networked exponential family (nExpFam)},
	text={nExpFam} 
}

\newglossaryentry{scatterplot}
{name={scatterplot}, 
	description={A\index{scatterplot} 
		visualization technique that depicts \glspl{datapoint} using markers in a 2-D plane. 
		Fig.\ \ref{fig_scatterplot_temp_FMI_dict} depicts an example of a scatterplot.  
		\begin{figure}[H]
			\begin{center}
				\begin{tikzpicture}[scale=1]
					\tikzset{x=2cm,y=2cm,every path/.style={>=latex},node style/.style={circle,draw}}
					\begin{axis}[axis x line=none,
						axis y line=none,
						ylabel near ticks,
						xlabel near ticks,
						enlarge y limits=true,
						xmin=-5, xmax=30,
						ymin=-5, ymax=30,
						width=6cm, height=6cm ]
						\addplot[only marks] table [x=mintmp, y=maxtmp, col sep = semicolon] {assets/FMIData1.csv};
						\node at (axis cs:26,2) [anchor=west] {$\feature$};
						\node at (axis cs:0,30) [anchor=west] {$\truelabel$};
						\draw[->] (axis cs:-5,0) -- (axis cs:30,0);
						\draw[->] (axis cs:0,-5) -- (axis cs:0,30);
					\end{axis}
				\end{tikzpicture}
				\vspace*{-10mm}
			\end{center}
		\caption{Scatterplot with circle markers, where the \glspl{datapoint} represent daily weather conditions in Finland. 
			Each \gls{datapoint} is characterized by its \gls{minimum} daytime temperature $\feature$ 
			as the \gls{feature} and its \gls{maximum} daytime temperature $\truelabel$ as the \gls{label}. 
			The temperatures have been measured at the \gls{fmi} weather station Helsinki Kaisaniemi 
			during 1 September 2024—28 October 2024.}
			\label{fig_scatterplot_temp_FMI_dict}
			\vspace*{-3mm}
		\end{figure}
		A scatterplot can enable the visual inspection of \glspl{datapoint} that are naturally 
		represented by \glspl{featurevec} in high-dimensional spaces.
		\\
		See also: \gls{datapoint}, \gls{minimum}, \gls{feature}, \gls{maximum}, \gls{label}, \gls{fmi}, \gls{featurevec}, \gls{dimred}.},
	first={scatterplot},
	text={scatterplot} 
}

\newglossaryentry{stepsize}
{name={step size}, 
	description={See\index{step size} \gls{learnrate}.}, 
	first={step size},
	text={step size} 
}

\newglossaryentry{learnrate}
{name={learning rate}, 
	description={Consider\index{learning rate} an iterative \gls{ml} method for finding or 
	    	learning a useful \gls{hypothesis} $\hypothesis \in \hypospace$. 
		Such an iterative method repeats similar computational (update) steps that adjust or 
		modify the current \gls{hypothesis} to obtain an improved \gls{hypothesis}. 
		A key \gls{parameter} of an iterative method is the learning rate. The learning rate 
		controls the extent to which the current \gls{hypothesis} can be modified during a 
		single \gls{iteration}. Consider, for example, the \gls{gradstep} \cite[Ch. 5]{MLBasics} 
		\begin{equation}
			\label{equ_def_basic_gradstep_lrate_dict}
			\weights^{(\iteridx\!+\!1)} = \weights^{(\iteridx)} - \lrate \nabla f(\weights^{(\iteridx)})
       		\end{equation} 
		of a \gls{gdmethod} for \gls{erm}, where the \gls{objfunc} $f(\weights)$ 
		is the \gls{emprisk} incurred by $\hypothesis^{(\weights)}$ on a \gls{trainset}. 
		Given the current \glspl{modelparam} $\weights^{(\iteridx)}$ at \gls{iteration} $\iteridx$, 
		the \gls{gradstep} produces updated \glspl{modelparam} $\weights^{(\iteridx\!+\!1)}$ 
		by moving in the opposite direction of the \gls{gradient} $\nabla f(\weights^{(\iteridx)})$. 
		\begin{figure}[H]
			\begin{center}
			\begin{minipage}{0.45\columnwidth}
			\begin{tikzpicture}[xscale=0.4,yscale=0.6]
				\draw[blue, ultra thick, domain=-4.1:4.1] plot (\x,  {(1/4)*\x*\x});
				\draw[] (1,0.25) circle [radius=0.1] node [right] (A) {$f(\weights^{(\iteridx)})$} ;
				\draw[] (-2,1) circle [radius=0.1] node [left] (B) {$f(\weights^{(\iteridx\!+\!1)})$} ;
				\draw[] (3,2.25) circle [radius=0.1]  node  [right] (C) {$f(\weights^{(\iteridx\!+\!2)})$} ;
				\draw[->,dashed] (-2,1) -- (3,2.25) node [midway,above] {\eqref{equ_def_basic_gradstep_lrate_dict}};
				\draw[->,dashed]  (1,0.25) -- (-2,1) node [midway,above] {\eqref{equ_def_basic_gradstep_lrate_dict}};
				\node [below] at (0,-0.2) {(a)};
			\end{tikzpicture}
			\end{minipage}
			\begin{minipage}{0.45\columnwidth}
			\begin{tikzpicture}[xscale=0.4,yscale=0.6]
				\draw[blue, ultra thick, domain=-4.1:4.1] plot (\x,  {(1/4)*\x*\x});
				\draw[] (4,4) circle [radius=0.1];
				\node [right] at (4,4) {$f(\weights^{(\iteridx)})$};
				\draw[] (3.8,3.61) circle [radius=0.1];
				\node [left] at (3.8,3.61) {$f(\weights^{(\iteridx\!+\!1)})$};
				\draw[] (3.65,3.33) circle [radius=0.1];
				\node [right] at (3.65,3.33) {$f(\weights^{(\iteridx\!+\!2)})$};
				\node [below] at (0,-0.2) {(b)};
			\end{tikzpicture}
			\end{minipage}
			\end{center}
		\caption{Effect of using an inadequate learning rate $\lrate$ in the \gls{gradstep} 
			\eqref{equ_def_basic_gradstep_lrate_dict}. (a) If $\lrate$ is too large, 
			the \glspl{gradstep} can ``overshoot'' such that the iterates $\weights^{(\iteridx)}$ 
			diverge away from the optimum, i.e., $f(\weights^{(\iteridx\!+\!1)}) > f(\weights^{(\iteridx)})$. 
			(b) If $\lrate$ is too small, the \glspl{gradstep} make too little progress towards 
			the optimum within the available number of \glspl{iteration} (due to limited computational budget). 
			\label{fig_small_large_lrate_dict}}
		\end{figure}
		See also: \gls{ml}, \gls{hypothesis}, \gls{parameter}, \gls{gd}, \gls{stochGD}, \gls{projgd}, \gls{stepsize}.},
	first={learning rate},
	text={learning rate} 
}

\newglossaryentry{featuremap}
{name={feature map}, 
	description={A \gls{feature} \gls{map}\index{feature map} refers to a \gls{function} 
		$$
		\featuremapvec: \featurespace \rightarrow \featurespace', \qquad \featurevec \mapsto \featurevec'
		$$
		that transforms a \gls{featurevec} $\featurevec \in \featurespace$ of 
 		a \gls{datapoint} into a new \gls{featurevec} $\featurevec' \in \featurespace'$, 
 		where $\featurespace'$ is typically different from $\featurespace$.
 		The transformed representation $\featurevec'$ is often more useful than the original 
 		$\featurevec$. For instance, the geometry of \glspl{datapoint} may become more linear 
 		in $\featurespace'$, allowing the application of a \gls{linmodel} to $\featurevec'$. 
 		This idea is central to the design of \glspl{kernelmethod}~\cite{LearningKernelsBook}.
 		Other benefits of using a \gls{feature} \gls{map} include reducing \gls{overfitting} and 
 		improving \gls{interpretability}~\cite{Ribeiro2016}. A common use case is \gls{data} 
 		visualization, where a \gls{feature} \gls{map} with two output dimensions allows the representation 
 		of \glspl{datapoint} in a 2-D \gls{scatterplot}. Some \gls{ml} methods employ trainable 
 		\gls{feature} \glspl{map}, whose \glspl{parameter} are learned from \gls{data}. An example is 
 		the use of hidden \glspl{layer} in a \gls{deepnet}, which act as successive \gls{feature} \glspl{map} 
 		\cite{MallatUnderstandingDeepLearning}. A principled way to train a \gls{feature} \gls{map} 
 		is through \gls{erm}, using a \gls{lossfunc} that measures reconstruction quality, 
 		e.g., $\lossfun = \|\featurevec - r(\featurevec')\|^2$, where $r(\cdot)$ is a trainable
 		\gls{map} that attempts to reconstruct $\featurevec$ from the transformed \gls{featurevec} $\featurevec'$.
				\\
		See also: \gls{feature}, \gls{map}, \gls{kernelmethod}, \gls{featlearn}, \gls{pca}.},
	first={feature map},
	text={feature map} 
}
 
\newglossaryentry{lasso}
{name={least absolute shrinkage and selection operator (Lasso)}, 
	description={The Lasso \index{least absolute shrinkage and selection operator (Lasso)} is an 
		instance of \gls{eerm} \cite{hastie01statisticallearning}. It learns the \glspl{weight} 
		$\weights$ of a \gls{linearmap} $\hypothesis(\featurevec) = \weights\,^{T} \featurevec$ from a \gls{trainset}. 
		Lasso is obtained from \gls{linreg} by adding the scaled $\ell_{1}$-\gls{norm} 
		$\regparam \normgeneric{\weights}{1}$ to the average \gls{sqerrloss} incurred on 
		the \gls{trainset} \cite{Tibshirani1996}. Using the $\ell_{1}$-\gls{norm} as a \gls{regularizer}, 
		instead of the squared $\ell_{2}$-\gls{norm} used in \gls{ridgeregression}, encourages the 
		learned \glspl{weight} to have many entries set to zero \cite{Wain2019}, \cite{BuhlGeerBook}.
				\\
		See also: \gls{eerm}, \gls{weight}, \gls{linearmap}, \gls{trainset}, \gls{linreg}, \gls{norm}, \gls{sqerrloss}.},
	first={Lasso},
	text={Lasso} 
}
 
\newglossaryentry{simgraph}
{name={similarity graph}, 
 	description={Some\index{similarity graph} \gls{ml} applications generate \glspl{datapoint} that 
 		are related by a domain-specific notion of similarity. These similarities can be 
 		represented conveniently using a similarity \gls{graph} $\graph = \big(\nodes \defeq \{1, \,\ldots, \,\samplesize\}, \edges\big)$. 
 		The node $\sampleidx \in \nodes$ represents the $\sampleidx$th \gls{datapoint}. Two 
 		nodes are \gls{connected} by an undirected edge if the corresponding \glspl{datapoint} are similar. 
				\\
		See also: \gls{ml}, \gls{datapoint}, \gls{graph}, \gls{connected}.},
 	first={similarity graph},
	text={similarity graph} 
}

\newglossaryentry{kernel}
{name={kernel (kernel method)}, 
	description={Consider\index{kernel (kernel method)} a set of \glspl{datapoint}, each represented by a \gls{featurevec} 
	 	$\featurevec \in \featurespace$, where $\featurespace$ denotes the \gls{featurespace}. 
	 	A (real-valued) kernel is a \gls{function} 
	 	$\kernel: \featurespace \times \featurespace \rightarrow \mathbb{R}$ that assigns to every pair of 
     		\glspl{featurevec} $\featurevec, \featurevec' \in \featurespace$ a real number $\kernelmap{\featurevec}{\featurevec'}$. 
     		This value is typically interpreted as a similarity \gls{measure} between $\featurevec$ and $\featurevec'$. 
	 	The defining property of a kernel is that it is symmetric, i.e.,
	 	$\kernelmap{\featurevec}{\featurevec'} = \kernelmap{\featurevec'}{\featurevec}$, and that 
	 	for any finite set of \glspl{featurevec} $\featurevec_1, \,\ldots, \,\featurevec_n \in \featurespace$, the \gls{matrix} 
	  	\begin{equation}
	 		\nonumber
	 		\mathbf{K} = \begin{pmatrix}
	 			\kernelmap{\featurevec_1}{\featurevec_1} & \kernelmap{\featurevec_1}{\featurevec_2} & \ldots & \kernelmap{\featurevec_1}{\featurevec_n} \\
	 			\kernelmap{\featurevec_2}{\featurevec_1} & \kernelmap{\featurevec_2}{\featurevec_2} & \ldots & \kernelmap{\featurevec_2}{\featurevec_n} \\
	 			\vdots											
	 			& \vdots & \ddots & \vdots \\
	 			\kernelmap{\featurevec_n}{\featurevec_1} & \kernelmap{\featurevec_n}{\featurevec_2} & \ldots & \kernelmap{\featurevec_n}{\featurevec_n} 
	 		\end{pmatrix} \in \mathbb{R}^{n \times n}
	 	\end{equation}
	 	is \gls{psd}. 
     		A kernel naturally defines a transformation of a \gls{featurevec} $\featurevec$ into a 
	 	\gls{function} $\vz = \kernelmap{\featurevec}{\cdot}$. The \gls{function} $\vz$ maps an  
	 	input $\featurevec' \in \featurespace$ to the value $\kernelmap{\featurevec}{\featurevec'}$. 
	 	We can view the \gls{function} $\vz$ as a new \gls{featurevec} that belongs to a 
	 	\gls{featurespace} $\featurespace'$ that is typically different from $\featurespace$. 
	 	This new \gls{featurespace} $\featurespace'$ has a particular mathematical structure, i.e., it is a 
	 	reproducing kernel \gls{hilbertspace} (RKHS)~\cite{LearningKernelsBook}, \cite{LampertNowKernel}.
     		Since $\vz$ belongs to a RKHS, which is a \gls{vectorspace}, we can interpret it as a generalized 
	 	\gls{featurevec}. Note that a finite-length \gls{featurevec} $\featurevec=\big(\feature_{1}, \,\ldots, \,\feature_{\nrfeatures} \big)\,^{T} \in \mathbb{R}^{\nrfeatures}$ 
	 	can be viewed as a \gls{function} $\featurevec: \{1, \,\ldots, \,\nrfeatures\} \rightarrow \mathbb{R}$ 
	 	that assigns a real value to each index $\featureidx \in \{1, \,\ldots, \,\nrfeatures\}$.
          		\\
		See also: \gls{featurevec}, \gls{featurespace}, \gls{hilbertspace}, \gls{kernelmethod}.},
	first={kernel},
	text={kernel} 
}
	
\newglossaryentry{kernelmethod}
{name={kernel method},
	description={A\index{kernel method} \gls{kernel} method is an \gls{ml} method that uses a 
		\gls{kernel} $\kernel$ to map the original (i.e., raw) \gls{featurevec} $\featurevec$ of a 
		\gls{datapoint} to a new (transformed) \gls{featurevec} $\vz = \kernelmap{\featurevec}{\cdot}$ \cite{LearningKernelsBook}, \cite{LampertNowKernel}.
		The motivation for transforming the \glspl{featurevec} is that, by using a suitable \gls{kernel}, 
		the \glspl{datapoint} have a more "pleasant" geometry in the transformed \gls{featurespace}. 
		For example, in a \gls{binclass} problem, using transformed \glspl{featurevec} $\vz$ might 
		allow us to use \glspl{linmodel}, even if the \glspl{datapoint} are not linearly 
		separable in the original \gls{featurespace} (see Fig.\ \ref{fig_linsep_kernel_dict}). 
		\begin{figure}[H]
			\begin{center}
 			\begin{tikzpicture}[auto,scale=0.6]
        			\draw [thick] (-6,2) circle (0.1cm) node[anchor=west] {\hspace*{0mm}$\featurevec^{(5)}$};
       			\draw [thick] (-8,1.6) circle (0.1cm) node[anchor=west] {\hspace*{0mm}$\featurevec^{(4)}$};
        			\draw [thick] (-7.4,-1.7) circle (0.1cm) node[anchor=west] {\hspace*{0mm}$\featurevec^{(3)}$};
        			\draw [thick] (-6,-1.9) circle (0.1cm) node[anchor=west] {\hspace*{0mm}$\featurevec^{(2)}$};
        			\draw [thick] (-6.5,0.0) rectangle ++(0.1cm,0.1cm) node[anchor=west,above] {\hspace*{0mm}$\featurevec^{(1)}$};
        			\draw [thick] (4,0) circle (0.1cm) node[anchor=north] {\hspace*{0mm}$\vz^{(5)}$};
        			\draw [thick] (5,0) circle (0.1cm) node[anchor=north] {\hspace*{0mm}$\vz^{(4)}$};
        			\draw [thick] (6,0) circle (0.1cm) node[anchor=north] {\hspace*{0mm}$\vz^{(3)}$};
        			\draw [thick] (7,0) circle (0.1cm) node[anchor=north] {\hspace*{0mm}$\vz^{(2)}$};
        			\draw [thick] (2,0) rectangle ++(0.1cm,0.1cm) node[anchor=west,above] {\hspace*{0mm}$\vz^{(1)}$};
       			\draw[->,bend left=30] (-3,0) to node[midway,above] {$\vz = \kernelmap{\featurevec}{\cdot}$} (1,0);
    			\end{tikzpicture}
			\end{center}
		\caption{Five \glspl{datapoint} characterized by \glspl{featurevec} $\featurevec^{(\sampleidx)}$ 
			and \glspl{label} $\truelabel^{(\sampleidx)} \in \{ \circ, \square \}$ for $\sampleidx=1, \,\ldots, \,5$. 
			With these \glspl{featurevec}, there is no way to separate the two classes 
			by a straight line (representing the \gls{decisionboundary} of a \gls{linclass}). 
			In contrast, the transformed \glspl{featurevec} $\vz^{(\sampleidx)} = \kernelmap{\featurevec^{(\sampleidx)}}{\cdot}$ 
			allow us to separate the \glspl{datapoint} using a \gls{linclass}.  \label{fig_linsep_kernel_dict}}
		\end{figure}
		See also: \gls{kernel}, \gls{featurevec}, \gls{featurespace}, \gls{linclass}.},
	first={kernel method},
	plural={kernel methods},
	text={kernel method} 
}
	
\newglossaryentry{cm}
{name={confusion matrix}, 
	description={Consider\index{confusion matrix} a finite \gls{dataset} 
		with $\samplesize$ \glspl{datapoint}, each characterized 
		by a \gls{featurevec} $\featurevec$ and a \gls{label} 
		$\truelabel \in \labelspace$ with a finite \gls{labelspace} $\labelspace = \{1, \,\ldots, \,\nrcluster\}$. 
		For a given \gls{hypothesis} $\hypothesis$, the confusion \gls{matrix} is a 
		$\nrcluster \times \nrcluster$ \gls{matrix} where each row corresponds 
		to a specific value of the true \gls{label} $\truelabel \in \labelspace$ 
		and each column to a specific value of the \gls{prediction} $\hypothesis(\featurevec) \in \labelspace$. 
		The \gls{matrix} entry in the $\clusteridx$th row and $\clusteridx'$th 
		column is the number of \glspl{datapoint} with the true \gls{label} 
		$\truelabel = \clusteridx$ that are predicted as 
		$\hypothesis(\featurevec) = \clusteridx'$. The sum of the main 
		diagonal entries is the number of correctly classified 
		\glspl{datapoint}, i.e., those for which $\truelabel = \hypothesis(\featurevec)$. 
		Summing the off-diagonal entries results in the total number of \glspl{datapoint} 
		that are misclassified by $\hypothesis$.
				\\
		See also: \gls{label}, \gls{labelspace}, \gls{hypothesis}, \gls{matrix}, \gls{classification}.},
	first={confusion matrix},
	text={confusion matrix} 
}

\newglossaryentry{precision}
{name={precision}, 
	description={Precision\index{precision} is a \gls{metric} commonly used in \gls{binclass} for 
		the assessment of trained \glspl{model}. It measures the proportion of correctly 
		predicted \glspl{datapoint} among those with a positive \gls{label} 
		\cite{Juba_Le_2019}, \cite{Rijsbergen1979}.
		\\
		See also: \gls{metric}, \gls{classification}, \gls{cm}, \gls{recall}.},
	first={precision}, 
	text={precision}
}

\newglossaryentry{recall}
{name={recall}, 
	description={Recall\index{recall} is a \gls{metric} commonly used in \gls{binclass} for 
		the assessment of trained \glspl{model}. It is the ratio between the number of 
		true positives (i.e., correctly predicted positive \glspl{datapoint}) and the number 
		of \glspl{datapoint} with a positive \gls{label} \cite{Juba_Le_2019}, \cite{Rijsbergen1979}.
		\\ 
		See also: \gls{metric}, \gls{classification}, \gls{cm}, \gls{precision}.},
	first={recall}, 
	text={recall}
}

\newglossaryentry{sensitivity}
{name={sensitivity}, 
	description={See\index{sensitivity} \gls{recall}.},
	first={sensitivity}, 
	text={sensitivity}
}

\newglossaryentry{transferlearning}
{name={transfer learning},
	description={Transfer learning\index{transfer learning} aims at leveraging information 
		obtained while solving an existing \gls{learningtask} to 
   		solve another \gls{learningtask}.
		\\ 
   		See also: \gls{learningtask}, \gls{multitask learning}},
  	first={transfer learning},
  	text={transfer learning}
}

\newglossaryentry{featuremtx}
{name={feature matrix}, 
	description={Consider\index{feature matrix} a \gls{dataset} $\dataset$ 
		with $\samplesize$ \glspl{datapoint} with \glspl{featurevec} 
		$\featurevec^{(1)}, \,\ldots, \,\featurevec^{(\samplesize)} \in \mathbb{R}^{\nrfeatures}$. 
		It is convenient to collect the individual \glspl{featurevec} into 
		the \gls{feature} \gls{matrix}:
		$$\mX \defeq \big(\featurevec^{(1)}, \,\ldots, \,\featurevec^{(\samplesize)}\big)\,^{T} =
		\begin{pmatrix}
			x^{(1)}_{1} & x^{(1)}_{2} & \cdots & x^{(1)}_{\nrfeatures} \\
			x^{(2)}_{1} & x^{(2)}_{2} & \cdots & x^{(2)}_{\nrfeatures} \\
			\vdots      & \vdots      & \ddots & \vdots \\
			x^{(\samplesize)}_{1} & x^{(\samplesize)}_{2} & \cdots & x^{(\samplesize)}_{\nrfeatures}
		\end{pmatrix}
		\in \mathbb{R}^{\samplesize \times \nrfeatures}.$$
		Note that the \gls{feature} \gls{matrix} is of size $\samplesize \times \nrfeatures$, i.e., 
		it has $\samplesize$ rows and $\nrfeatures$ columns.
				\\
		See also: \gls{dataset}, \gls{datapoint}, \gls{featurevec}, \gls{feature}, \gls{matrix}.},
	first={feature matrix},
	text={feature matrix} 
}

\newglossaryentry{dbscan}
{name={density-based spatial clustering of applications with noise (DBSCAN)}, 
	description={DBSCAN\index{density-based spatial clustering of applications with noise (DBSCAN)} 
		refers to a \gls{clustering} \gls{algorithm} for \glspl{datapoint} that are characterized by numeric \glspl{featurevec}. 
		Like \gls{kmeans} and \gls{softclustering} via \gls{gmm}, DBSCAN also uses the \glspl{eucliddist} 
		between \glspl{featurevec} to determine the \glspl{cluster}. However, in contrast to \gls{kmeans} 
		and \gls{gmm}, DBSCAN uses a different notion of similarity between \glspl{datapoint}. 
		DBSCAN considers two \glspl{datapoint} as similar if they are connected 
		via a \gls{sequence} (i.e., path) of nearby intermediate \glspl{datapoint}. Thus, DBSCAN might consider 
		two \glspl{datapoint} as similar (and therefore belonging to the same \gls{cluster}) even if 
		their \glspl{featurevec} have a large \gls{eucliddist}.
				\\
		See also: \gls{clustering}, \gls{kmeans}, \gls{gmm}, \gls{cluster}, \gls{graph}.},
	first={density-based spatial clustering of applications with noise (DBSCAN)},
	text={DBSCAN} 
}

\newglossaryentry{fl}
{name={federated learning (FL)}, 
	description={FL\index{federated learning (FL)} 
		is an umbrella term for \gls{ml} methods that train \glspl{model} in a collaborative 
		fashion using decentralized \gls{data} and computation.
				\\
		See also: \gls{ml}, \gls{model}, \gls{data}.},
	first={federated learning (FL)},
	text={FL} 
}
	
\newglossaryentry{cfl}
{name={clustered federated learning (CFL)}, 
	description={CFL\index{clustered federated learning (CFL)} trains \glspl{localmodel} for the 
 		\glspl{device} in an \gls{fl} application by using a \gls{clustasspt}, i.e., the \glspl{device} 
 		of an \gls{flnetwork} form \glspl{cluster}. Two \glspl{device} in the same \gls{cluster} generate 
 		\glspl{localdataset} with similar statistical properties. CFL pools the \glspl{localdataset} of \glspl{device} 
 		in the same \gls{cluster} to obtain a \gls{trainset} for a \gls{cluster}-specific \gls{model}. 
 		\Gls{gtvmin} clusters \glspl{device} implicitly by enforcing approximate similarity of \glspl{modelparam} 
 		across well-connected nodes of the \gls{flnetwork}.\\ 
 		See also: \gls{fl}, \gls{clustasspt}, \gls{flnetwork}, \gls{cluster}, \gls{graphclustering}.},
	first={clustered federated learning (CFL)},
	text={CFL} 
}

\newglossaryentry{coreset}
{name={coreset}, 
	description={A coreset\index{coreset} is a small subset of a larger \gls{dataset} 
		that approximates certain properties of the original \gls{dataset} \cite{Chai2023}. 
		The construction of a coreset typically involves selecting representative \glspl{datapoint}
		and assigning them \glspl{weight} to reflect their importance in the original \gls{dataset} (Fig.\ \ref{fig_coreset_dict}).
        		\begin{figure}[H]
			\centering
			\begin{tikzpicture}
  			\foreach \x/\y in {0.5/0.7, 1.2/1.4, 1.8/0.9, 2.2/1.8, 2.6/1.2, 3.1/1.6}
    			{\fill (\x,\y) circle (1.5pt);}
  			\foreach \x/\y in {1.2/1.4, 2.6/1.2}{
    			\fill[blue] (\x,\y) circle (2pt);
    			\draw[blue, thick] (\x,\y) circle (6pt);
  			}
  			\node[blue] (label) at (0.6,2.2) {\small coreset};
  			\draw[->,blue] (label) -- (1.2,1.5);
  			\draw[->,blue] (label) -- (2.6,1.3);
			\end{tikzpicture}
		\caption{Coreset (highlighted in blue) representing a small subset of a larger \gls{dataset}. \label{fig_coreset_dict}}
		\end{figure}
		Coresets are particularly useful for \gls{ml} applications (such as \gls{clustering}) 
		involving large \glspl{dataset}, as they allow for efficient computation while 
		preserving the essential characteristics of the \gls{data} \cite{DistkmeansBalcan2013}.
				\\
		See also: \gls{dataset}, \gls{datapoint}, \gls{clustering}.},
	first={coreset},
	firstplural={coresets},
	plural={coresets},
	text={coreset} 
}

\newglossaryentry{outlier}
{name={outlier}, 
	description={Many\index{outlier} \gls{ml} methods 
		are motivated by the \gls{iidasspt}, which interprets \glspl{datapoint} as \glspl{realization} of 
		\gls{iid} \glspl{rv} with a common \gls{probdist}. The \gls{iidasspt} is useful for applications  
		where the statistical properties of the \gls{data} generation process are stationary (or time-invariant) \cite{Brockwell91}. 
		However, in some applications, the \gls{data} consist of a majority of regular \glspl{datapoint} 
		that conform with the \gls{iidasspt} as well as a small number of \glspl{datapoint} that have fundamentally different 
       		statistical properties compared with the regular \glspl{datapoint}. We refer to 
		a \gls{datapoint} that substantially deviates from the statistical properties of 
		most \glspl{datapoint} as an outlier. Different methods for outlier detection 
		use different \glspl{measure} of this deviation. 
        		Statistical learning theory studies fundamental limits on the ability to 
		mitigate outliers reliably \cite{doi:10.1137/0222052}, \cite{10.1214/20-AOS1961}.
        		\\
		See also: \gls{robustness}, \gls{stability}, \gls{huberreg}, \gls{probmodel}.},
	 first={outlier},
	 text={outlier} 
}

\newglossaryentry{ensemble}
{name={ensemble}, 
	description={An ensemble method\index{ensemble} combines multiple 
		\gls{ml} methods, each of those referred to as a \gls{baselearner}, 
		to improve overall performance. The \glspl{baselearner} can be \gls{erm}-based 
        		using different choices for the \gls{loss}, \gls{model}, and \gls{trainset}. 
		By aggregating the \glspl{prediction} of \glspl{baselearner}, ensemble methods can 
		often achieve better performance than any single \gls{baselearner}. The 
		aggregation can amount to averaging the \glspl{prediction} of \glspl{baselearner} 
		(in \gls{regression}) or using a majority vote (for \gls{classification} methods). 
		\begin{figure}[H]
		\begin{center}
			\begin{tikzpicture}[
				scale=1.1, transform shape,
				node distance=11mm and 10mm,
				dataset/.style={draw, rounded corners, inner sep=2pt},
				learner/.style={draw, rounded corners,inner sep=2pt},
				op/.style={draw, circle, inner sep=1pt},
				flow/.style={->, >=latex},
				feedback/.style={->, >=latex, dashed, very thin},
				lab/.style={font=\scriptsize}
				]
				\node[dataset] (D) {$\dataset$};
				\node[dataset, below left=of D]  (D1) {\small $\widetilde{\dataset}^{(1)}$};
				\node[dataset, below=of D]       (D2) {\small $\widetilde{\dataset}^{(2)}$};
				\node[dataset, below right=of D] (D3) {\small $\widetilde{\dataset}^{(3)}$};
				\draw[flow] (D) -- (D1) node[midway, lab, above left=-1pt] {resample};
				\draw[flow] (D) -- (D2) node[midway, lab, right] {};
				\draw[flow] (D) -- (D3) node[midway, lab, above right=-1pt] {};
				\node[learner, below=5mm of D1] (L1) {\small $\hypospace^{(1)}$};
				\node[learner, below=5mm of D2] (L2) {\small $\hypospace^{(2)}$};
				\node[learner, below=5mm of D3] (L3) {\small $\hypospace^{(3)}$};
				\draw[flow] (D1) -- (L1);
				\draw[flow] (D2) -- (L2);
				\draw[flow] (D3) -- (L3);
				\draw[feedback] (L1.east) .. controls +(+7mm,0mm) and +(-7mm,0mm) .. (L2.west)
				node[midway, lab, above] {};
				\draw[feedback] (L2.east) .. controls +(+7mm,0mm) and +(-7mm,0mm) .. (L3.west);
				\draw[feedback] (L1.east) to[out=60, in=120] (L3.west);
				\node[op, below=8mm of L2] (agg) {\small $\phi_{\text{agg}}$};
				\draw[flow] (L1) -- (agg);
				\draw[flow] (L2) -- (agg);
				\draw[flow] (L3) -- (agg);
				\node[below=5mm of agg, align=right, inner sep=2pt] (yhat)
				{\small $\learnthypothesis= \phi_{\text{agg}}\!\big(\learntlocalhypothesis{1},\,\learntlocalhypothesis{2},\,\learntlocalhypothesis{3}\big)$};
				\draw[flow] (agg) -- (yhat);
				\node[lab, below left=-1pt and -1pt of L1.south] {$\learntlocalhypothesis{1}$};
				\node[lab, below           =-1pt of L2.south]    {$\learntlocalhypothesis{2}$};
				\node[lab, below right=-1pt and -1pt of L3.south]{$\learntlocalhypothesis{3}$};
			\end{tikzpicture}
		\caption{Generic ensemble with three \glspl{baselearner}, each  
			using \gls{erm} to learn $\hypospace^{(\featureidx)} \in \hypospace^{(\featureidx)}$ 
			based on the \gls{trainset} $\widetilde{\dataset}^{(\featureidx)}$. A \gls{baselearner} might also 
			use the \gls{output} of other \glspl{baselearner}. The final \gls{hypothesis} $\learnthypothesis$ 
			is obtained by aggregating the \glspl{hypothesis} generated by the \glspl{baselearner}.}
		\end{center}
		\end{figure}
		Different ensemble methods use different constructions for the \glspl{baselearner}. 
		For example, \gls{bagging} methods (such as a \gls{randomforest}) use random sampling to 
		construct slightly different \glspl{trainset} for each \gls{baselearner}. 
	    	On the other hand, \gls{boosting} methods run the \glspl{baselearner} 
		sequentially, i.e., each \gls{baselearner} tries to correct the \gls{prediction} 
		errors of the previous ones. A third family of ensemble methods is \gls{stacking}, 
		where \glspl{baselearner} are trained on the same \gls{trainset} but with 
		different \glspl{model}.
		\\
	 	See also: \gls{bagging}, \gls{boosting}, \gls{stacking}.},
	 first={ensemble},
	 text={ensemble} 
}

\newglossaryentry{stacking}
{name={stacking}, 
	description={Stacking\index{stacking} is one of the main types of \gls{ensemble} methods.  
		In stacking, a finite number $\numlearners$ of \glspl{baselearner} are trained on the 
		same \gls{dataset} but with different \glspl{model} $\hypospace^{(\featureidx)}$ 
		or \glspl{lossfunc} $\loss^{(\featureidx)}$ for $\featureidx=1,\,\ldots,\,\numlearners$ \cite[Ch. 8.8]{hastie01statisticallearning},
		\cite{WOLPERT1992241}, \cite{ZhouEnsemble2012}. 
		The $\featureidx$th \gls{baselearner} delivers a learned \gls{hypothesis} 
		$\learntlocalhypothesis{\featureidx} \in \hypospace^{(\featureidx)}$. 
		The final \gls{prediction} for a \gls{datapoint} is obtained by aggregating the 
		\glspl{prediction} of the \glspl{baselearner} via an aggregation rule $\phi_{\rm agg}$, 
		such as majority voting for \gls{classification} or averaging for \gls{regression}. 
		We can interpret stacking as a form of \gls{featlearn}, where each \gls{baselearner} 
		extracts a new \gls{feature}. The aggregation rule can be obtained by another instance 
		of \gls{erm} that learns a \gls{hypothesis} \gls{map} $\phi_{\rm agg} \in \hypospace$ from 
		a meta-\gls{model} $\hypospace$. The \gls{hypothesis} $\phi_{\rm agg}$ is applied to the 
		transformed \gls{featurevec} $$\featuremapvec(\featurevec)= \big( \learnthypothesis^{(1)}(\featurevec), 
       		\,\ldots, \,\learnthypothesis^{(\numlearners)}(\featurevec) \big)^{T}.$$
       		\begin{figure}[H]
			\begin{center}
			\begin{tikzpicture}[
			font=\small,
			scale=1.0, transform shape,
			node distance=7mm and 10mm,
			dataset/.style={draw, rounded corners, inner sep=2pt},
			learner/.style={draw, rounded corners, minimum width=14mm, minimum height=7mm, inner sep=6pt,align=center},
			op/.style={draw, circle, inner sep=1pt},
			>=latex
			]
			\node[dataset] (D) {$\dataset$};
			\node[learner, below left=of D]  (L1) {\gls{erm} \\ $\hypospace^{(1)},\loss^{(1)}$};
			\node[learner, below=of D]       (L2) {\gls{erm} \\ $\hypospace^{(2)},\loss^{(2)}$};
			\node[learner, below right=of D] (L3) {\gls{erm} \\ $\hypospace^{(3)},\loss^{(3)}$};
			\draw[->] (D) -- (L1) node[midway, above left=-1pt] {};
			\draw[->] (D) -- (L2) node[midway, right]           {};
			\draw[->] (D) -- (L3) node[midway, above right=-1pt]{};
			\node[op, below=8mm of L2] (agg) {$\phi_{\text{agg}}$};
			\node[below=5mm of agg, align=right, inner sep=2pt] (yhat)
			{$\predictedlabel = \phi_{\text{agg}}\!\big(\learntlocalhypothesis{1}(\featurevec),\learntlocalhypothesis{2}(\featurevec),\learntlocalhypothesis{3}(\featurevec)\big)$};
			\draw[->] (L1) -- (agg);
			\draw[->] (L2) -- (agg);
			\draw[->] (L3) -- (agg);
			\draw[->] (agg) -- (yhat);
			\node[below left=-1pt and -1pt of L1.south] {$\learntlocalhypothesis{1}$};
			\node[below           =-1pt of L2.south]    {$\learntlocalhypothesis{2}$};
			\node[below right=-1pt and -1pt of L3.south]{$\learntlocalhypothesis{3}$};
			\end{tikzpicture}
		\caption{Three \glspl{baselearner} using \gls{erm} with 
			different \glspl{model} and \glspl{lossfunc} to obtain learned \glspl{hypothesis} 
			$\learntlocalhypothesis{1},\,\learntlocalhypothesis{2},\,\learntlocalhypothesis{3}$. 
			For a \gls{datapoint} with \gls{featurevec} $\featurevec$, each \gls{baselearner} delivers a \gls{prediction}  
			$\learntlocalhypothesis{\featureidx}(\featurevec)$ for $\featureidx=1,\,2,\,3$. These 
			\glspl{prediction} are then used as new \glspl{feature} for an aggregation 
			rule $\phi_{\rm agg}$ that delivers the overall \gls{prediction} $\predictedlabel$. 
			The aggregation rule can be obtained by \gls{training} a meta-\gls{model} $\hypospace$.}
			\end{center}
		\end{figure}
	 	See also: \gls{ensemble}, \gls{bagging}.},
	 first={stacking},
	 text={stacking} 
}

\newglossaryentry{auc}
{name={area under the curve (AUC)}, 
	description={The AUC\index{area under the curve (AUC)} is a quantitative \gls{measure} of the 
 		usefulness of a binary \gls{classifier} \cite{Hanley:1982aa}. It 
		is defined (using the natural \gls{measure} of the \gls{euclidspace} $\mathbb{R}^{2}$) as the area  
		under the \gls{roc} curve.
		\\
		See also: \gls{classifier}, \gls{euclidspace}, \gls{roc}. } , 
 	first={area under the curve (AUC)}, 
 	text={AUC} 
}

\newglossaryentry{roc} 
{name={receiver operating characteristic (ROC)}, 
	description={Consider a \gls{labelspace} $\labelspace=\{-1,1\}$ and a \gls{classifier} 
 		that uses a real-valued \gls{hypothesis} $\hypothesis(\featurevec)$. For a given 
 		threshold $\eta \in \mathbb{R}$, the ultimate \gls{prediction} is $\predictedlabel=1$ 
 		if $\hypothesis(\featurevec)\geq \eta$ and $\predictedlabel=-1$ otherwise. 
 		On a \gls{testset} $\testset$, we compute, for each value of $\eta$, 
		the following two quantities: 1) true positive rate ${\rm TPR}^{(\eta)}$; 
		and 2) false positive rate ${\rm FPR}^{(\eta)}$. 
 		The ROC curve\index{receiver operating characteristic (ROC)} is  
 		the following \gls{function} \cite{IntroROCFawcett}:
 		$${\rm ROC}: \mathbb{R} \rightarrow \mathbb{R}^{2}: \eta \mapsto \big({\rm TPR}^{(\eta)}, {\rm FPR}^{(\eta)} \big).$$ 
 	        One important characteristic of the ROC curve is the \gls{auc}.
	        \\ 
 		See also: \gls{classifier}, \gls{auc}.}, 
  	first={receiver operating characteristic (ROC)}, 
  	text={ROC}
}

\newglossaryentry{bagging}
{name={bagging},
	description={Bagging\index{bagging} is an \gls{ensemble} technique where \glspl{baselearner} 
		use perturbed copies $$\widetilde{\dataset}^{(1)},\,\ldots,\,\widetilde{\dataset}^{(\numlearners)}$$ 
		of the original \gls{trainset} $\dataset$ \cite{Breiman:1996aa}. Each 
		\gls{baselearner} delivers a potentially different \gls{hypothesis} $$\learnthypothesis^{(1)},\,\ldots,\,\learnthypothesis^{(\numlearners)}.$$
		The \gls{hypothesis} delivered by the overall method is obtained by 
		aggregating the \glspl{hypothesis} $\learnthypothesis^{(1)},\,\ldots,\,\learnthypothesis^{(\numlearners)}$ 
		using some aggregation rule. For \gls{classification} methods, the rule is typically 
		a majority vote, while for \gls{regression} methods, it amounts to averaging. 
		\begin{figure}[H]
			\begin{center}
			\begin{tikzpicture}[
			scale=1.0, transform shape,
			node distance=10mm and 10mm,
			dataset/.style={draw, rounded corners, inner sep=2pt},
			learner/.style={draw, rounded corners, minimum width=14mm, minimum height=7mm, inner sep=2pt},
			op/.style={draw, circle, inner sep=1pt},
			>=latex
			]
			\node[dataset] (D) {$\dataset$};
			\node[dataset, below left=of D]  (D1) {$\widetilde{\dataset}^{(1)}$};
			\node[dataset, below=of D]       (D2) {$\widetilde{\dataset}^{(2)}$};
			\node[dataset, below right=of D] (D3) {$\widetilde{\dataset}^{(3)}$};
			\draw[->] (D) -- (D1) node[midway, above left=-1pt] {resample};
			\draw[->] (D) -- (D2) node[midway, right]           {resample};
			\draw[->] (D) -- (D3) node[midway, above right=-1pt]{resample};
			\node[learner, below=5mm of D1] (L1) {$\hypospace^{(1)}$};
			\node[learner, below=5mm of D2] (L2) {$\hypospace^{(2)}$};
			\node[learner, below=5mm of D3] (L3) {$\hypospace^{(3)}$};
			\draw[->] (D1) -- (L1);
			\draw[->] (D2) -- (L2);
			\draw[->] (D3) -- (L3);
			\node[op, below=8mm of L2] (agg) {$\phi_{\text{agg}}$};
			\node[below=5mm of agg, align=right, inner sep=2pt] (yhat)
			{$\learnthypothesis = \phi_{\text{agg}}\!\big(\learntlocalhypothesis{1},\learntlocalhypothesis{2},\learntlocalhypothesis{3}\big)$};
			\draw[->] (L1) -- (agg);
			\draw[->] (L2) -- (agg);
			\draw[->] (L3) -- (agg);
			\draw[->] (agg) -- (yhat);
			\node[below left=-1pt and -1pt of L1.south] {$\learntlocalhypothesis{1}$};
			\node[below           =-1pt of L2.south]    {$\learntlocalhypothesis{2}$};
			\node[below right=-1pt and -1pt of L3.south]{$\learntlocalhypothesis{3}$};
			\end{tikzpicture}
		\caption{Example of bagging in which three \glspl{baselearner} use perturbations  
			$\widetilde{\dataset}^{(1)},\,\ldots,\,\widetilde{\dataset}^{(3)}$ of 
			the original \gls{trainset} $\dataset$ to learn the \glspl{hypothesis} 
			$\learntlocalhypothesis{1},\,\ldots,\,\learntlocalhypothesis{3}$. The final 
			\gls{hypothesis} is obtained by aggregating these individual \glspl{hypothesis} 
			via some aggregation rule $\phi_{\rm agg}$.}
			\end{center}
		\end{figure}
		See also: \gls{ensemble}, \gls{robustness}, \gls{bootstrap}.},
	first={bagging},
	text={bagging}
}

\newglossaryentry{bootstrap aggregation}
{name={bootstrap aggregation}, 
	description={See \gls{bagging}\index{bootstrap aggregation}.},
 	first={bootstrap aggregation},
 	text={bootstrap aggregation} 
}	

\newglossaryentry{decisionregion}
{name={decision region},
	description={Consider\index{decision region} 
		a \gls{hypothesis} \gls{map} $\hypothesis$ that delivers values from a finite set $\labelspace$. 
		For each \gls{label} value (i.e., category) $a \in \labelspace$, the \gls{hypothesis} $\hypothesis$ 
		determines a subset of \gls{feature} values $\featurevec \in \featurespace$ that result 
		in the same \gls{output} $\hypothesis(\featurevec)=a$. We refer to this subset as a decision 
		region of the \gls{hypothesis} $\hypothesis$. 
		\\
		See also: \gls{hypothesis}, \gls{map}, \gls{label}, \gls{feature}.},
	first={decision region},
	plural={decision regions}, 
	firstplural={decision regions}, 
	text={decision region} 
}

\newglossaryentry{baselearner}
{name={base learner}, 
	description={A base learner\index{base learner} is an \gls{ml} method 
         	that is part of an \gls{ensemble} method. 
		\\
		See also: \gls{ensemble}, \gls{bagging}, \gls{stacking}, \gls{boosting}.},
	first={base learner},
	firstplural={base learners},
	plural={base learners},
	text={base learner} 
}	

\newglossaryentry{decisionboundary}
{name={decision boundary}, 
	description={Consider\index{decision boundary} a 
		\gls{hypothesis} \gls{map} $\hypothesis$ that reads in a \gls{featurevec}  
		$\featurevec \in \mathbb{R}^{\featuredim}$ and delivers a value from a finite set $\labelspace$. 
		The decision \gls{boundary} of $\hypothesis$ is the set of \glspl{vector} $\featurevec \in \mathbb{R}^{\featuredim}$ 
		that lie between different \glspl{decisionregion}. More precisely, a 
		\gls{vector} $\featurevec$ belongs to the decision \gls{boundary} if and only 
		if each \gls{neighborhood} $\{ \featurevec': \| \featurevec - \featurevec' \| \leq \varepsilon \}$, 
		for any $\varepsilon >0$, contains at least two \glspl{vector} with different \gls{function} values.
				\\
		See also: \gls{hypothesis}, \gls{map}, \gls{featurevec}, \gls{boundary}, \gls{vector}, \gls{decisionregion}, 
		\gls{neighborhood}, \gls{function}.},
	first={decision boundary},
	firstplural={decision boundaries},
	plural={decision boundaries},
	text={decision boundary} 
}

\newglossaryentry{normalequations}
{name={normal equations}, 
	description={Consider \gls{linleastsquares}\index{normal equations}, which learns the \glspl{parameter} of a 
		\gls{linmodel} by minimizing the average \gls{sqerrloss} on a 
		\gls{trainset} $\dataset$. 
		The learned \glspl{parameter} $\widehat{\weights}$ are characterized by the 
		linear system of equations, i.e.,
		\begin{equation} 
			\label{equ_normal_equations_dict}
			\featuremtx^{T} \featuremtx \widehat{\weights} = \featuremtx^{T} \labelvec.
		\end{equation} 
		Here, $\featuremtx=\big(\featurevec^{(1)},\,\ldots,\,\featurevec^{(\samplesize)}\big)^{T} \in \mathbb{R}^{\samplesize \times \nrfeatures}$ 
		is the \gls{featuremtx} of $\dataset$ and $\labelvec=\big(\truelabel^{(1)},\,\ldots,\,\truelabel^{(\samplesize)} \big)^{T} \in \mathbb{R}^{\samplesize}$ 
		is the \gls{labelvec} of $\dataset$. This linear system of equations is referred 
		to as normal equations, as it amounts to an orthogonality condition. Indeed, \eqref{equ_normal_equations_dict} 
		can be rewritten as $$ \featuremtx^{T} \big( \featuremtx \widehat{\weights} - \labelvec \big) = \mathbf{0}.$$
		This means that the \gls{prediction} error \gls{vector} $\featuremtx \widehat{\weights} - \labelvec$ 
		is orthogonal to the columns of $\featuremtx$ and, in turn, to the \gls{subspace} $\linspan{\featuremtx}$
		spanned by them.
		\\ 
		See also: \gls{linleastsquares}, \gls{modelparam}. },
	first={normal equations},
	text={normal equations} 
}

\newglossaryentry{eerm}
{name={explainable empirical risk minimization (EERM)}, 
	description={EERM is an\index{explainable empirical risk minimization (EERM)} 
		in-\linebreak stance of \gls{srm} that adds a \gls{regularization} term to the 
		average \gls{loss} in the \gls{objfunc} of \gls{erm}. 
		The \gls{regularization} term is chosen to favor \gls{hypothesis} \glspl{map} that are intrinsically 
		explainable for a specific user. This user is characterized by their \glspl{prediction} provided 
		for the \glspl{datapoint} in a \gls{trainset} \cite{Zhang:2024aa}.
				\\
		See also: \gls{srm}, \gls{regularization}, \gls{erm}, \gls{trainset}.},
	first={explainable empirical risk minimization (EERM)},
	text={EERM} 
}
	
\newglossaryentry{kmeans}
{name={$k$-means}, 
sort={kmeans},
	description={The\index{$k$-means} $k$-\glspl{mean} principle is an \gls{optimization}-based 
	    	approach to the \gls{clustering} of \glspl{datapoint} with a numeric \gls{featurevec} 
		\cite[Ch. 8]{MLBasics}. As a \gls{hardclustering} approach, $k$-\glspl{mean} partitions 
		a \gls{dataset} into $k$ disjoint subsets (or \glspl{cluster}), which are indexed 
		by $\clusteridx=1,\,\ldots,\,\nrcluster$. Each \gls{cluster} $\cluster$ is represented 
		by the average \gls{featurevec} of \glspl{datapoint} that belong to it. 
		This average (or \gls{mean}) \gls{featurevec} is referred to as the \gls{clustercentroid} 
		$\clustercentroid{\clusteridx}$. A visual illustration is provided in Fig.\ \ref{fig_kmeans_dict}.
		\begin{figure}[H]
		\begin{center}
		\begin{tikzpicture}[scale=1]
			\tikzset{
			data/.style={circle, fill=black, inner sep=1.2pt},
			centroid/.style={thick, cross out, draw, minimum size=6pt, inner sep=0pt}
			}
			\node[data] (xi) at (1.0,0.2) {};
			\node[right=0pt of xi,yshift=2pt] {$\featurevec^{(\sampleidx)}$};
			\foreach \p in {(-0.3,0.0),(0.2,-0.4),(0.6,0.8),(0.0,0.9),(1.1,-0.2)}
			\node[data] at \p {};
			\foreach \p in {(2.5,1.0),(3.7,1.2),(2.6,2.3),(3.8,2.5),(3.0,2.9),(3.6,1.6)}
			\node[data] at \p {};
			\node[centroid] (mu1) at (0.55,0.4) {};
			\node[centroid] (mu2) at (3.1,1.85) {};
			\node[above of=mu1,yshift=-25pt,xshift=-10pt] {$\clustercentroid{1}$};
			\node[above of=mu2,yshift=-25pt,xshift=-10pt] {$\clustercentroid{2}$};
		\end{tikzpicture}
		\end{center}
		\caption{\Gls{scatterplot} of \glspl{datapoint}, indexed by $\sampleidx=1,\,\ldots,\,\samplesize$ and 
			characterized by \glspl{featurevec} $\featurevec^{(\sampleidx)} \in \mathbb{R}^{2}$. 
			The \gls{scatterplot} also includes two \glspl{clustercentroid} $\clustercentroid{1}, 
			\clustercentroid{2} \in \mathbb{R}^{2}$. \label{fig_kmeans_dict}}
		\end{figure} 
		The quality (or usefulness) of a specific \gls{clustering} is measured by some form of 
		\gls{clusteringerror} such as the average squared \gls{eucliddist} between the \gls{featurevec} 
		of a \gls{datapoint} and the \gls{clustercentroid} of the \gls{cluster} to which the \gls{datapoint} belongs. 
		In general, solving the $k$-\glspl{mean} \gls{optproblem} exactly is challenging (or NP-hard) 
		\cite{Mahajan2009Springer}. However, there are simple iterative methods for finding 
		approximately optimal \glspl{clustercentroid}. One such method is referred to as 
		\gls{lloydalgorithm}.
		\\
		See also: \gls{hardclustering}, \gls{cluster}, \gls{lloydalgorithm}.},
	first={$k$-means},
	text={$k$-means} 
}

\newglossaryentry{lloydalgorithm} 
{name={Lloyd's algorithm}, 
	description={Lloyd's\index{Lloyd's algorithm} \gls{algorithm} \cite{Lloyd1982} is an iterative 
		\gls{optmethod} for finding \glspl{clustercentroid} that are approximately 
	   	optimal for the \gls{kmeans} \gls{objfunc}. Lloyd's \gls{algorithm}  
	   	alternates between: 1) updating the \gls{cluster} assignment of each \gls{datapoint} 
		based on the nearest current \gls{clustercentroid}; and 
		2) recalculating the \glspl{clustercentroid} given the updated 
		\gls{cluster} assignments \cite{Lloyd1982}.
		\begin{figure}[H]
			\begin{center}
			\begin{tikzpicture}[
			assignment/.style={-Latex, very thin},
			move/.style={-Latex, thick},
			center/.style={draw, circle, inner sep=1.2pt, fill=white},
			pointA/.style={circle, inner sep=1.1pt, fill=black},
			pointB/.style={rectangle, inner sep=1.1pt, fill=black},
			cross/.style={line width=0.4pt},
			lab/.style={font=\scriptsize}
			]
			\begin{scope}[shift={(0,0)}]
				\coordinate (Ba1) at (0.2, 1.1);
				\coordinate (Ba2) at (0.4, 0.6);
				\coordinate (Ba3) at (0.8, 1.0);
				\coordinate (Ba4) at (0.6, 0.2);
				\coordinate (Bb1) at (2.7, 1.4);
				\coordinate (Bb2) at (3.2, 0.9);
				\coordinate (Bb3) at (2.6, 0.3);
				\coordinate (Bb4) at (3.3, 0.2);
				\node[pointA] at (Ba1) {}; \node[pointA] at (Ba2) {}; \node[pointA] at (Ba3) {}; \node[pointA] at (Ba4) {};
				\node[pointB] at (Bb1) {}; \node[pointB] at (Bb2) {}; \node[pointB] at (Bb3) {}; \node[pointB] at (Bb4) {};
				\coordinate (Bc1old) at (0.9,0.65);
				\coordinate (Bc2old) at (2.8,1.8);
				\node[center,label={[lab, xshift=-2pt]right:$\clustercentroiditer{1}{\iteridx}$}] (B-C1) at (Bc1old) {};
				\node[center,label={[lab, xshift=-2pt]right:$\clustercentroiditer{2}{\iteridx}$}] (B-C2) at (Bc2old) {};
				\path let 
				\p1=(B-C1), \p2=(B-C2),
				\n1={(\x2-\x1)}, \n2={(\y2-\y1)},
				\n3={veclen(\n1,\n2)}
				in coordinate (B-M)    at ($ (B-C1)!0.5!(B-C2) $)
				coordinate (B-Nhat) at ($ (0,0) + ({-\n2/\n3},{\n1/\n3}) $);
				\draw[dashed, thick] ($ (B-M) + 10.4*(B-Nhat) $) -- ($ (B-M) - 10.4*(B-Nhat) $);
				\draw[assignment] (Ba2) -- (B-C1);
				\draw[assignment] (Ba3) -- (B-C1);
				\draw[assignment] (Bb2) -- (B-C2);
				\draw[assignment] (Bb3) -- (B-C2);
				\node[lab] at (1.6,-0.35) {Assign to the nearest \gls{clustercentroid}.};
				\node at (1.6,-1.3) {(a)};
			\end{scope}
			\begin{scope}[shift={(5.5,0)}]
				\coordinate (Ca1) at (0.2, 1.1);
				\coordinate (Ca2) at (0.4, 0.6);
				\coordinate (Ca3) at (0.8, 1.0);
				\coordinate (Ca4) at (0.6, 0.2);
				\coordinate (Cb1) at (2.7, 1.4);
				\coordinate (Cb2) at (3.2, 0.9);
				\coordinate (Cb3) at (2.6, 0.3);
				\coordinate (Cb4) at (3.3, 0.2);
				\node[pointA] at (Ca1) {}; \node[pointA] at (Ca2) {}; \node[pointA] at (Ca3) {}; \node[pointA] at (Ca4) {};
				\node[pointB] at (Cb1) {}; \node[pointB] at (Cb2) {}; \node[pointB] at (Cb3) {}; \node[pointB] at (Cb4) {};
				\coordinate (Cc1old) at (0.9,0.65);
				\coordinate (Cc2old) at (2.8,1.8);
				\coordinate (Cc1new) at (0.5,0.725);
				\coordinate (Cc2new) at (2.95,0.70);
				\node[center, opacity=0.5] at (Cc1old) {};
				\node[center, opacity=0.5] at (Cc2old) {};
				\draw[cross] ($(Cc1new)+(-0.07,0)$) -- ($(Cc1new)+(0.07,0)$);
				\draw[cross] ($(Cc1new)+(0,-0.07)$) -- ($(Cc1new)+(0,0.07)$);
				\draw[cross] ($(Cc2new)+(-0.07,0)$) -- ($(Cc2new)+(0.07,0)$);
				\draw[cross] ($(Cc2new)+(0,-0.07)$) -- ($(Cc2new)+(0,0.07)$);
				\draw[move] (Cc1old) -- (Cc1new);
				\draw[move] (Cc2old) -- (Cc2new);
				\node[lab] at (1.6,-0.35) {Recompute \glspl{clustercentroid}.};
				\node at (1.6,-1.3) {(b)};
			\end{scope}
			\begin{scope}[shift={(3,-4)}]
				\coordinate (Da1) at (0.2, 1.1);
				\coordinate (Da2) at (0.4, 0.6);
				\coordinate (Da3) at (0.8, 1.0);
				\coordinate (Da4) at (0.6, 0.2);
				\coordinate (Db1) at (2.7, 1.4);
				\coordinate (Db2) at (3.2, 0.9);
				\coordinate (Db3) at (2.6, 0.3);
				\coordinate (Db4) at (3.3, 0.2);
				\node[pointA] at (Da1) {}; \node[pointA] at (Da2) {}; \node[pointA] at (Da3) {}; \node[pointA] at (Da4) {};
				\node[pointB] at (Db1) {}; \node[pointB] at (Db2) {}; \node[pointB] at (Db3) {}; \node[pointB] at (Db4) {};
				\coordinate (Dc1new) at (0.5,0.725);
				\coordinate (Dc2new) at (2.95,0.70);
				\node[center, label={[lab, xshift=-2pt]right:$\clustercentroiditer{1}{\iteridx+1}$}] (D-C1n) at (Dc1new) {};
				\node[center, label={[lab]right:$\clustercentroiditer{2}{\iteridx+1}$}] (D-C2n) at (Dc2new) {};
				\path let 
				\p1=(D-C1n), \p2=(D-C2n),
				\n1={(\x2-\x1)}, \n2={(\y2-\y1)},
				\n3={veclen(\n1,\n2)}
				in coordinate (D-M)    at ($ (D-C1n)!0.5!(D-C2n) $)
				coordinate (D-Nhat) at ($ (0,0) + ({-\n2/\n3},{\n1/\n3}) $);
				\draw[dashed, thick] ($ (D-M) + 10.4*(D-Nhat) $) -- ($ (D-M) - 10.4*(D-Nhat) $);
				\node[lab] at (1.6,-0.35) {Assign to the nearest \gls{clustercentroid}.};
				\node at (1.6,-1.3) {(c)};
			\end{scope}
			\end{tikzpicture}
			\end{center}
		\caption{Lloyd's \gls{algorithm} alternates between (a) and (c) assigning \glspl{datapoint} to 
			the nearest \gls{clustercentroid} and, in turn, (b) recomputing the \glspl{clustercentroid} 
			based on the new \gls{cluster} assignments. 
			\label{fig_lloyd_algorithm_dict}}
		\end{figure}
		See also: \gls{clustercentroid}, \gls{kmeans}, \gls{clustering}.},
	first={Lloyd's algorithm},
	text={Lloyd's algorithm} 
}

\newglossaryentry{qlearning} 
{name={Q-learning}, 
	description={Q-learning\index{Q-learning} is a popular \gls{reinforcementlearning} 
	     	\gls{algorithm} that learns an optimal \gls{policy} by estimating the optimal \gls{action}-value 
		\gls{function} (or Q-\gls{function}) \cite{SuttonEd2}.  
		\\
		See also: \gls{reinforcementlearning}, \gls{fixedpointiter}.},
	first={Q-learning},
	text={Q-learning} 
}

\newglossaryentry{iteration}
{name={iteration},
	description={The elementary computational step during the execution of an 
		\gls{algorithm} is referred to as iteration\index{iteration} 
		\cite{Cormen:2022aa}, \cite{KleinbergTardos2006}. 
		For example, the elementary computational step of \glspl{gdmethod} 
		is a \gls{gradstep}. 
		More generally, the elementary computational step of a \gls{fixedpointiter} 
		is the evaluation of an underlying \gls{operator} $\fixedpointop$, which might 
		vary across iterations (see Fig.\ \ref{fig_iteration_dict}). 
		Many important \gls{ml} \glspl{algorithm}, including 
		\gls{lloydalgorithm} and \gls{gd}, are \glspl{fixedpointiter}.  
		\begin{figure}[H]
			\centering
			\begin{tikzpicture}[>=Latex, font=\small,scale=1]
			\node[circle,draw,inner sep=1.2pt,fill=white,label={above:$\weights^\ast$}] (xstar) at (8,0) {};
			\node[circle,fill,inner sep=1pt,label={below:$\weights^{(0)}$}] (x0) at (0.3,0) {};
			\node[circle,fill,inner sep=1pt,label={below:$\weights^{(1)}$}] (x1) at (4.3,0) {};
			\node[circle,fill,inner sep=1pt,label={below:$\weights^{(2)}$}] (x2) at (6.5,0) {};
			\draw[->] (x0) to[bend left=12] node[above,sloped] {$\fixedpointop$} (x1);
			\draw[->] (x1) to[bend left=12] node[above,sloped] {$\fixedpointop$} (x2);
			\end{tikzpicture}
		\caption{\Gls{fixedpointiter} consisting of the repeated application of an 
			\gls{operator} $\fixedpointop$ with some \gls{fixedpoint} $\weights^\ast$, i.e., 
			$\fixedpointop\weights^\ast=\weights^\ast$. \label{fig_iteration_dict}}
		\end{figure} 
		See also: \gls{algorithm}, \gls{gradstep}, \gls{gd}.  },
	first={iteration},
	text={iteration},
	firstplural={iterations},
	plural={iterations}
}

\newglossaryentry{clustercentroid}
{name={cluster centroid}, 
	description={\Gls{clustering} methods\index{cluster centroid} decompose a given 
		\gls{dataset} into a few \glspl{cluster}. Different \gls{clustering} methods use 
		different representations for these \glspl{cluster}. If \glspl{datapoint} are 
		characterized by numerical \glspl{featurevec} $\featurevec \in \mathbb{R}^{\nrfeatures}$, 
		we can use some \gls{vector} ${\bm \mu} \in \mathbb{R}^{\nrfeatures}$, referred to 
		as \gls{cluster} centroid, to represent a \gls{cluster}. For example, if a \gls{cluster} 
		consists of a set of \glspl{datapoint}, we use the average of their \glspl{featurevec} 
		as a \gls{cluster} centroid. However, there are also other choices for how to construct 
		a \gls{cluster} centroid, e.g., using the \gls{geometricmedian} \cite{PARK20093336} 
		or via non-Euclidean geometry \cite{JMLR:v6:banerjee05b}.
		\\
		See also: \gls{clustering}, \gls{featurevec}, \gls{kmeans}.},
	first={cluster centroid},
	firstplural={cluster centroids}, 
	text={cluster centroid}, 
	plural={cluster centroids}
}

\newglossaryentry{xml}
{name={explainable machine learning (XML)}, 
	description={XML\index{explainable machine learning (XML)} 
		methods aim to complement each \gls{prediction} with an \gls{explanation} of 
		how the \gls{prediction} has been obtained. The construction of an explicit \gls{explanation} 
		might not be necessary if the \gls{ml} method uses a sufficiently simple (or interpretable) \gls{model} \cite{rudin2019stop}.
				\\
		See also: \gls{prediction}, \gls{explanation}, \gls{ml}, \gls{model}.},
	first={explainable machine learning (XML)},
	text={XML} 
}
	
\newglossaryentry{samplemean}
{name={sample mean}, 
	description={The\index{sample mean} \gls{sample} \gls{mean} 
		$\vm \in \mathbb{R}^{\nrfeatures}$ for a given \gls{dataset}, with \glspl{featurevec} 
		$\featurevec^{(1)}, \,\ldots, \,\featurevec^{(\samplesize)} \in \mathbb{R}^{\nrfeatures}$, is defined as 
		$$\vm = \frac{1}{\samplesize} \sum_{\sampleidx=1}^{\samplesize} \featurevec^{(\sampleidx)}.$$ 
					\\
		See also: \gls{sample}, \gls{mean}, \gls{dataset}, \gls{featurevec}.},
	first={sample mean},
	text={sample mean} 
}
	
\newglossaryentry{highdimregime}
{name={high-dimensional regime}, 
	description={The\index{high-dimensional regime} 
		high-dimensional regime of \gls{erm} is characterized by the \gls{effdim} of the \gls{model} 
		being larger than the \gls{samplesize}, i.e., the number of (labeled) \glspl{datapoint} in the \gls{trainset}. 
		For example, \gls{linreg} methods operate in the high-dimensional regime whenever the number $\featuredim$ of \glspl{feature} 
		used to characterize \glspl{datapoint} exceeds the number of \glspl{datapoint} in the \gls{trainset}. 
		Another example of \gls{ml} methods that operate in the high-dimensional regime is large \glspl{ann}, which have 
		far more tunable \glspl{weight} (and bias terms) than the total number of \glspl{datapoint} in the \gls{trainset}. 
		High-dimensional statistics is a recent main thread of \gls{probability} theory that studies the 
		behavior of \gls{ml} methods in the high-dimensional regime \cite{Wain2019}, \cite{BuhlGeerBook}.
				\\
		See also: \gls{erm}, \gls{effdim}, \gls{overfitting}, \gls{regularization}.},
   	first={high-dimensional regime},
	text={high-dimensional regime} 
}

\newglossaryentry{gmm}
{name={Gaussian mixture model (GMM)}, 
	description={A GMM\index{Gaussian mixture model (GMM)} is a \gls{probmodel} 
		for (the generation of) \glspl{datapoint} with numeric \glspl{featurevec} 
		$\featurevec \in \mathbb{R}^\nrfeatures$ \cite{BishopBook}, \cite{Murphy2012}. 
		It assumes that each \gls{datapoint} is generated by first drawing 
		a latent \gls{cluster} index $I \in \{1,\,\ldots,\,\nrcluster\}$
		according to \gls{cluster} \glspl{probability} 
		$$\prob{I=\clusteridx} = p_{\clusteridx}, 
		\qquad 
		\sum_{\clusteridx=1}^{\nrcluster} p_{\clusteridx}=1.$$
		Conditioned on $I=\clusteridx$, the \gls{featurevec} $\featurevec$ 
		is drawn from a \gls{mvndist}
		$\probdist^{(\clusteridx)}= \mvnormal{\meanvec{\clusteridx}}{\covmtx{\clusteridx}}$.
		The resulting \gls{marginaldist} of $\featurevec$ is therefore 
		a weighted sum of \glspl{mvndist}, i.e.,
		$$\probdist
		=\sum_{\clusteridx=1}^{\nrcluster}
		p_{\clusteridx} \mvnormal{\meanvec{\clusteridx}}{\covmtx{\clusteridx}}.$$
		\begin{figure}[H]
		\begin{center}
		\begin{tikzpicture}[scale=0.4]
			\draw (0,0) ellipse (3 and 1.5);
			\fill (0,0) circle (2pt);
			\node[below left] at (0,0) {$\meanvec{1}$};
			\node[above] at (0,1.8) {$\covmtx{1}$};
			\draw (4,0.5) ellipse (2 and 1);
			\fill (4,0.5) circle (2pt);
			\node[below right] at (4,0.5) {$\meanvec{2}$};
			\node[above] at (4,1.7) {$\covmtx{2}$};
			\begin{scope}[rotate around={30:(-3,1)}]
			\draw (-3,1) ellipse (2.5 and 1);
			\end{scope}
			\fill (-3,1) circle (2pt);
			\node[below] at (-3,1) {$\meanvec{3}$};
			\node[above left] at (-3,2.5) {$\covmtx{3}$};
		\end{tikzpicture}
		\caption{Illustration of a GMM with three components. \label{fig_gmm_dict}}
		\end{center}
		\end{figure}
		A GMM is parameterized by the \gls{cluster}-specific \gls{probability}  
		$p_{\clusteridx}$, \gls{mean} $\meanvec{\clusteridx}$, and \gls{covmtx}  
		$\covmtx{\clusteridx}$ for $\clusteridx=1,\,\ldots,\,\nrcluster$.
		\\		
		See also: \gls{probmodel}, \gls{mvndist}, \gls{clustering}.},
	first={Gaussian mixture model (GMM)},
	plural={GMMs}, 
	firstplural={Gaussian mixture models (GMMs)}, 
	text={GMM} 
}
	
\newglossaryentry{polyreg}
{name={polynomial regression}, 
	description={Polynomial\index{polynomial regression} 
		\gls{regression} is an instance of \gls{erm} that learns a polynomial \gls{hypothesis} 
		\gls{map} to predict a numeric \gls{label} based on the numeric \glspl{feature} of a \gls{datapoint}. 
		 For \glspl{datapoint} characterized by a single numeric \gls{feature}, polynomial \gls{regression} uses the \gls{hypospace} 
		$\hypospace^{(\rm poly)}_{\nrfeatures} \defeq \{ \hypothesis(x) = \sum_{\featureidx=0}^{\nrfeatures-1} x^{\featureidx} \weight_{\featureidx} \}.$
		The quality of a polynomial \gls{hypothesis} \gls{map} is measured using the average \gls{sqerrloss} 
		incurred on a set of \glspl{labeled datapoint} (which we refer to as the \gls{trainset}).
					\\
		See also: \gls{regression}, \gls{erm}, \gls{sqerrloss}.},
	first={polynomial regression},
	text={polynomial regression}
}

\newglossaryentry{leastsquares}
{name={least squares},
	description={Least squares\index{least squares} refers to \gls{erm}-based 
        		methods that use the average \gls{sqerrloss} 
		$$ \frac{1}{\samplesize} \sum_{\sampleidx=1}^{\samplesize} \big( \truelabel^{(\sampleidx)} 
		- \hypothesis(\featurevec^{(\sampleidx)}) \big)^{2} $$   
		on a \gls{trainset} $\dataset = \big\{ \pair{\featurevec^{(1)}}{\truelabel^{(1)}}, \,\ldots, \,
		\pair{\featurevec^{(\samplesize)}}{\truelabel^{(\samplesize)}} \big\}$	
		to measure the quality of a \gls{hypothesis} \gls{map} $\hypothesis \in \hypospace$. 
		We obtain different least squares methods by using different \glspl{model} 
		in \gls{erm}. For example, the least squares variant of a \gls{linmodel} 
		is a least squares method that uses a \gls{linmodel}.
			   \\
		See also: \gls{erm}, \gls{sqerrloss}, \gls{linmodel}, \gls{linreg}.},
	first={least squares},
	text={least squares}
}

\newglossaryentry{designmatrix}
{name={design matrix},
	description={The\index{design matrix} term design \gls{matrix} is a synonym 
 		for the \gls{featuremtx}, particularly used in statistics \cite{OxfordStatisticsDictionary}, \cite{Everitt2022}. 
 		It collects the \glspl{featurevec} of the \glspl{datapoint} in a \gls{dataset} 
 		that is used for \gls{model} \gls{training} or \gls{validation}.
		\\
		See also: \gls{featuremtx}, \glspl{featurevec}, \gls{datapoint}, \gls{dataset}. }, 
 	first={design matrix},
 	text={design matrix} 
}

\newglossaryentry{labelvec}
{name={label vector},
sort={labelvector},
	description={Given\index{label vector} a \gls{dataset} of $\samplesize$ \glspl{labeled datapoint} 
 		$$\pair{\featurevec^{(1)}}{\truelabel^{(1)}}, \,\ldots, \,\pair{\featurevec^{(\samplesize)}}{\truelabel^{(\samplesize)}},$$ it 
 		is convenient to collect the corresponding \glspl{label} into a single 
 		\gls{label} \gls{vector} $\labelvec \defeq \big( \truelabel_{1},\,\ldots,\,\truelabel_{\samplesize}\big)^{T}$ 
 		\cite{BishopBook}, \cite{hastie01statisticallearning}. 
 		\\
 		See also: \gls{dataset}, \gls{labeled datapoint}, \gls{label}, \gls{datapoint}.},
 	first={label vector},
 	text={label vector}, 
 	firstplural={label vectors},
 	plural={label vectors}
}

\newglossaryentry{inputvec}
{name={input vector},
	description={The term\index{input vector} input \gls{vector} is often used 
		as a synonym for the \gls{featurevec} of a \gls{datapoint}. In settings where \glspl{datapoint} 
 		arise from a \gls{dynamicalsystem} observed over time, \glspl{feature} are obtained 
 		from measuring input variables. These input variables are then used 
 		by \gls{ml} methods to predict the system’s \gls{output} (which is a \gls{label} in \gls{ml} terminology).
		\\
		See also: \gls{vector}, \gls{featurevec}, \gls{datapoint}, \gls{output}. }, 
 	first={input vector}, 
 	text={input vector}, 
 	firstplural={input vectors}, 
 	plural={input vectors}
}

\newglossaryentry{outputvec}
{name={output vector},
	description={The term\index{output vector} \gls{output} \gls{vector} is used 
 		as a synonym for the \gls{labelvec} of a \gls{dataset} \cite{Murphy2012}.
		\\
		See also: \gls{output}, \gls{labelvec}, \gls{dataset}. },  
 	first={output vector}, 
 	text={output vector}, 
 	firstplural={output vectors}, 
 	plural={output vectors}
}

\newglossaryentry{output}
{name={output},
	description={The term\index{output} output is sometimes used 
 		as a synonym for the \gls{label} of a \gls{datapoint} \cite{Murphy2012}.
		\\
		See also: \gls{label}, \gls{datapoint}. },  
 	first={output}, 
 	text={output}, 
 	firstplural={outputs}, 
 	plural={outputs}
}

\newglossaryentry{targetvec}
{name={target vector},
	description={The term\index{target vector} \gls{target} \gls{vector} is used 
 		as a synonym for the \gls{labelvec} of a \gls{dataset} \cite{Goodfellow-et-al-2016}, \cite{BishopBook}.
		\\
		See also: \gls{target}, \gls{labelvec}, \gls{dataset}. },  
 	first={target vector}, 
 	text={target vector}, 
 	firstplural={target vectors}, 
 	plural={target vectors}
}

\newglossaryentry{target}
{name={target},
	description={The term\index{target} target is sometimes used 
 		as a synonym for the \gls{label} of a \gls{datapoint} \cite{Goodfellow-et-al-2016}, \cite{BishopBook}.
		\\
		See also: \gls{label}, \gls{datapoint}. },  
 	first={target}, 
 	text={target}, 
 	firstplural={targets}, 
 	plural={targets}
}

\newglossaryentry{responsevec}
{name={response vector},
	description={The term\index{response vector} \gls{response} \gls{vector} is used 
 		as a synonym for the \gls{labelvec} of a \gls{dataset} \cite{hastie01statisticallearning}.
		\\
		See also: \gls{response}, \gls{labelvec}, \gls{dataset}, \gls{targetvec}.},  
 	first={response vector}, 
 	text={response vector}, 
 	firstplural={response vectors}, 
 	plural={response vectors}
}

\newglossaryentry{response}
{name={response},
	description={The term\index{response} response is sometimes used 
 		as a synonym for the \gls{label} of a \gls{datapoint} \cite{hastie01statisticallearning}.
		\\
		See also: \gls{label}, \gls{datapoint}, \gls{target}. },  
 	first={response}, 
 	text={response}, 
 	firstplural={responses}, 
 	plural={responses}
}

\newglossaryentry{linleastsquares}
{name={linear least squares},
	description={Linear \gls{leastsquares}\index{linear least squares} refers to the 
 		variant of \gls{linreg} that uses the \gls{sqerrloss} to measure the quality of 
 		a linear \gls{hypothesis} \gls{map}. Conversely, it can also be viewed as the 
 		variant of \gls{leastsquares} that restricts the \gls{hypothesis} to a \gls{linmodel}. 
 		In particular, linear \gls{leastsquares} learns the \glspl{parameter} $\weights$ of a linear \gls{hypothesis} 
  		$\hypothesis^{(\weights)}(\featurevec) = \weights^{T} \featurevec$ by solving the following \gls{optproblem}:
  		\begin{equation} 
 			\label{eq_linleastsquares_dic}
 			\min_{\weights \in \mathbb{R}^{\nrfeatures}} \normgeneric{\labelvec - \featuremtx \weights}{2}^2.
  		\end{equation} 
  		Here, the \gls{featuremtx} is 
  		$\featuremtx = \big(\featurevec^{(1)},\,\ldots,\,\featurevec^{(\samplesize)}\big)$ 
  		and the \gls{labelvec} is 
  		$\labelvec= \big(\truelabel^{(1)},\ldots,\truelabel^{(\samplesize)}\big)^{T}$. 
  		Both are constructed from the \gls{trainset}
  		$$\dataset = \left\{ \pair{\featurevec^{(1)}}{\truelabel^{(1)}}, \,\ldots, \,
 			   \pair{\featurevec^{(\samplesize)}}{\truelabel^{(\samplesize)}} \right\}.$$ 
  		The \gls{optproblem} in \eqref{eq_linleastsquares_dic} admits a clear 
  		geometric interpretation, i.e., we seek the \gls{vector} $\featuremtx \widehat{\weights}$ in the 
 	 	\gls{columnspace} of $\featuremtx$ that is closest to the \gls{labelvec} $\vy$ 
  		(see Fig.\ \ref{fig_linleastsquares_dict}) \cite[Ch. 8]{BoydConvexBook}. 
  		A necessary and sufficient condition for $\widehat{\weights}$ to minimize 
  		\eqref{eq_linleastsquares_dic} is the \gls{normalequations}:
  		$$ \featuremtx^{T} \featuremtx \widehat{\weights} = \featuremtx^{T} \labelvec. $$
		\begin{figure}[H]
 			\centering
  			\begin{tikzpicture}[scale=1]
   			\draw[thick] (-1,-0.333) -- (3,1) node[pos=0.0,below right] {$\linspan{\featuremtx}$};
   			\coordinate (y) at (1,2);
   			\fill (y) circle (1.6pt) node[above] {$\labelvec$};
   			\coordinate (xw) at (1.5,0.5);
   			\fill (xw) circle (1.6pt) node[below right] {$\featuremtx\widehat{\weights}$};
   			\draw[dashed] (y) -- (xw);
    			\begin{scope}[xshift=6.2cm]
       				\node[anchor=west] at (0,1.2) {\gls{normalequations}};
       				\node[anchor=west] at (0,0.4) {$\featuremtx^{T}\featuremtx\,\widehat{\weights} \;=\; \featuremtx^{T}\labelvec$};
				\node at (-5, -2) {(a)};
				\node at (1.5, -2) {(b)};
     			\end{scope}
 			\end{tikzpicture}
 		\caption{Linear \gls{leastsquares} has both geometric and algebraic interpretations. 
 			(a) Geometrically, it finds the orthogonal \gls{projection} of the 
 			\gls{labelvec} $\labelvec$ onto the \gls{columnspace} of the 
 			\gls{featuremtx} $\featuremtx$ \cite[Ch. 8]{BoydConvexBook}. (b) Algebraically, it solves a linear system 
 			known as \gls{normalequations}. 
 			\label{fig_linleastsquares_dict}}
 		\end{figure}
		See also: \gls{leastsquares}, \gls{linreg}, \gls{sqerrloss}, \gls{linmodel}, \gls{erm}.},
	first={linear least squares},
	text={linear least squares}
}

\newglossaryentry{weightedleastsquares}
{name={weighted least squares},
	description={Weighted \gls{leastsquares}\index{weighted least squares} 
 		refers to \gls{erm}-based methods that use the weighted average \gls{sqerrloss} 
		$$ \frac{1}{\samplesize} \sum_{\sampleidx=1}^{\samplesize} \sampleweight{\sampleidx} \big( \truelabel^{(\sampleidx)} 
		- \hypothesis(\featurevec^{(\sampleidx)}) \big)^{2} $$   
		on a \gls{trainset} $\dataset = \big\{ \pair{\featurevec^{(1)}}{\truelabel^{(1)}}, \,\ldots, \,
		\pair{\featurevec^{(\samplesize)}}{\truelabel^{(\samplesize)}} \big\}$	
		to measure the quality of a \gls{hypothesis} \gls{map} $\hypothesis \in \hypospace$. 
		The \glspl{weight} $\sampleweight{1}, \,\ldots, \,\sampleweight{\samplesize} \in \mathbb{R}_{+}$
		allow us to emphasize or de-emphasize the contribution of individual \glspl{datapoint} 
		in the \gls{trainset}. Ideally, we assign a small weight $\sampleweight{\sampleidx}$ 
		to the $\sampleidx$th \gls{datapoint} if it is an \gls{outlier} 
		(see Fig.\ \ref{fig_weighted_least_squares_dict}). 
		We obtain different weighted \gls{leastsquares} methods by 
		using different \glspl{model} in \gls{erm}.  
				\begin{figure}[H]
			\centering
		  	\begin{tikzpicture}[scale=0.7, y=0.5cm, x=0.5cm]
		  		\foreach \x/\y in {
		  		1/2, 4/3, 7/10
		  		} {
		  		\draw[dashed, gray] (\x, 0) -- (\x, \y);
		  		\filldraw[blue] (\x, \y) circle (2pt);
		  		\node[circle, inner sep=0pt] (ptB\x) at (\x, \y) {};
		  		}
		  		\draw[dashed, thick] (0.5, 3.5) -- (10.5, 3.5) node[right] 
		  		{$\hypothesis$};
		  		\node[above right=2pt and 2pt, red] at (ptB7) {\gls{outlier}};
				\coordinate (lineStart) at (0.5, 3.5);
				\coordinate (lineEnd)   at (10.5, 3.5);
				\draw[dashed, thick] (lineStart) -- (lineEnd) node[right] {$\hypothesis$};
				\path let \p1 = (ptB7) in coordinate (proj7) at (\x1, 3.5);
				\draw[<->, very thick, red]
					(proj7) -- node[right, xshift=2pt, fill=white, inner sep=1pt]
					{$\sampleweight{\sampleidx} \big( \truelabel^{(\sampleidx)}\!-\!\hypothesis\big(\featurevec^{(\sampleidx)}\big) \big)^{2}$}
					(ptB7);
				\fill[red] (proj7) circle (1.2pt);
				\fill[red] (ptB7)  circle (1.2pt);
		  	\end{tikzpicture}
		\caption{Weighted \gls{leastsquares} can be used to mitigate the effect of \gls{outlier} 
			\glspl{datapoint} in a \gls{trainset}. 
			\label{fig_weighted_least_squares_dict}}
		\end{figure}
		See also: \gls{erm}, \gls{sqerrloss}, \gls{linreg}, \gls{linmodel}.},
	first={weighted least squares},
	text={weighted least squares}
}

\newglossaryentry{linreg}
{name={linear regression}, 
	description={Linear\index{linear regression} 
		\gls{regression} methods learn a linear \gls{hypothesis} \gls{map} 
		$\hypothesis^{(\weights)}(\featurevec) =\weights^{T}\featurevec$ that is used 
		to predict the numeric \gls{label} $\truelabel \in \mathbb{R}$ of a \gls{datapoint} 
		based on its numeric \glspl{featurevec} $\featurevec=\big(\feature_{1},\,\ldots,
		\,\feature_{\nrfeatures}\big)^{T} \in \mathbb{R}^{\nrfeatures}$. 
		The least-squares variant of linear \gls{regression} measures the quality 
		of a linear \gls{hypothesis} \gls{map} via the average 
		\gls{sqerrloss} incurred on a \gls{trainset} 
		$$\pair{\featurevec^{(1)}}{\truelabel^{(1)}},\,\ldots,
		\,\pair{\featurevec^{(\samplesize)}}{\truelabel^{(\samplesize)}}.$$ 
		As an instance of \gls{erm}, linear (least-squares) \gls{regression} learns 
		the \glspl{modelparam} $\weights$ by solving the following \gls{optproblem}:
		$$\min_{\weights \in \mathbb{R}^{\nrfeatures}} \frac{1}{\samplesize} \sum_{\sampleidx=1}^{\samplesize} \big( \truelabel^{(\sampleidx)} 
		- \weights^{T} \featurevec^{(\sampleidx)} \big)^{2}.$$
		\begin{figure}[H]
		 \centering
		 \begin{tikzpicture}[scale=0.7, y=0.5cm, x=0.5cm]
		 	\begin{scope}
		 		\foreach \x/\y in {
		 			1/2, 4/3, 7/4
		 		} {
		 			\draw[dashed, gray] (\x, 0) -- (\x, \y);
		 			\filldraw[blue] (\x, \y) circle (2pt);
		 			\node[circle, inner sep=0pt] (ptA\x) at (\x, \y) {};
		 		}
		 		\draw[dashed, thick] (0.5, 3) -- (10.5, 3) node[right] {$\widehat{w}$};
		 		\node at (7.5, -4) {(a)};
		 	\end{scope}
		 	\begin{scope}[xshift=10cm]
		 		\foreach \x/\y in {
		 			1/2, 4/3, 7/10
		 		} {
		 			\draw[dashed, gray] (\x, 0) -- (\x, \y);
		 			\filldraw[blue] (\x, \y) circle (2pt);
		 			\node[circle, inner sep=0pt] (ptB\x) at (\x, \y) {};
		 		}
		 		\draw[dashed, thick] (0.5, 7.5) -- (10.5, 7.5) node[right] 
		 		{$\widehat{w}$};
		 		\node[above right=2pt and 2pt, red] at (ptB7) {\gls{outlier}};
		 		\node at (7.5, -4) {(b)};
		 	\end{scope}
		 \end{tikzpicture}
		 \caption{For a \gls{linmodel} with $\nrfeatures=1$ and using 
			the trivial \gls{feature} $\feature=1$ for any \gls{datapoint}, 
		 	linear \gls{regression} reduces to computing the average 
		 	$\widehat{w} = (1/\samplesize) \sum_{\sampleidx=1}^{\samplesize} \truelabel^{(\sampleidx)}$. 
			(a) Clean \gls{trainset} and the resulting \gls{parameter} (given by the average). 
		 	(b) Perturbed \gls{dataset} (including an \gls{outlier}) 
		 	and the resulting \gls{parameter}. \label{fig_linreg_dict}}
		\end{figure}
		We can rewrite the above \gls{optproblem} more compactly using the 
		\gls{featuremtx} $\featuremtx = \big(\featurevec^{(1)},\,\ldots,\,\featurevec^{(\samplesize)}\big)^{T} \in \mathbb{R}^{\samplesize \times \nrfeatures}$ 
		and the \gls{label} \gls{vector} $\labelvec = \big( \truelabel^{(1)},\,\ldots, \,\truelabel^{(\samplesize)} \big)^{T} \in \mathbb{R}^{\samplesize}$. 
		This allows us to rewrite the above \gls{optproblem} as
		$$\min_{\weights \in \mathbb{R}^{\nrfeatures}} \frac{1}{\samplesize} \normgeneric{\labelvec - \featuremtx \weights}{2}^{2}.$$
		By the \gls{zerogradientcondition}, a necessary and sufficient condition for a 
		\gls{vector} $\widehat{\weights}$ to be a solution to the above \gls{optproblem} is the 
		linear system of equations \cite{StrangLinAlg2016}:
		\begin{equation} 
			\label{eq_linreg_normal_eq_dict}
			\featuremtx^{T}\featuremtx \widehat{\weights} = \featuremtx^{T} \labelvec.
		\end{equation}
		Instead of solving \eqref{eq_linreg_normal_eq_dict} directly 
		(via computing the \gls{inverse} or \gls{pseudoinverse}), 
		many \gls{ml} methods use variants of \gls{gd} to construct a \gls{sequence} 
		$\weights^{(0)},\,\weights^{(1)},\,\ldots$ of increasingly accurate approximations 
		of a solution $\widehat{\weights}$ to \eqref{eq_linreg_normal_eq_dict}. These 
		\glspl{gdmethod} can be interpreted as a \gls{fixedpointiter} for the following 
		reformulation of \eqref{eq_linreg_normal_eq_dict}:
		$$ \big(\mathbf{I} - \mA \featuremtx^{T}\featuremtx\big) \widehat{\weights} + \featuremtx^{T} 
		\labelvec = \widehat{\weights} \quad \mbox{with some \gls{invertible} \gls{matrix} } \mA.$$
		This equation is solved by a \gls{vector} $\widehat{\weights}$ if and only if 
		this \gls{vector} also solves \eqref{eq_linreg_normal_eq_dict}. The optimality 
		condition \eqref{eq_linreg_normal_eq_dict} is also useful for the study of 
		the \gls{stability} of linear \gls{regression}. Ideally, we would like the solutions of 
		\eqref{eq_linreg_normal_eq_dict} to be insensitive to small perturbations 
		of the \gls{trainset}. We can capture these perturbations via a 
		perturbed \gls{featuremtx} $\widetilde{\featuremtx} = \featuremtx + \Delta \featuremtx$
		and perturbed \gls{label} 
		\gls{vector} $\widetilde{\labelvec} = \labelvec + \Delta \labelvec$. 
		Here, $\Delta \featuremtx$ and $\Delta \labelvec$ represent small perturbations
		to the \glspl{featurevec} and \glspl{label} of the \glspl{datapoint} in the 
		original \gls{trainset}. \Gls{matrix} perturbation theory allows us to evaluate 
		how much the solutions of the perturbed linear \gls{regression} problem \cite[Sec.\ 2.6]{GolubVanLoanBook}
		$$\widetilde{\featuremtx}^{T} \widetilde{\featuremtx} \widetilde{\weights} = 
		\widetilde{\featuremtx}^{T} \widetilde{\labelvec}$$
		deviate from the solutions $\widetilde{\weights}$ 
		of the original linear \gls{regression} problem.
		\\
		See also: \gls{regression}, \gls{erm}, \gls{linmodel}.},
	first={linear regression},
	text={linear regression}
}
        
\newglossaryentry{ridgeregression}
{name={ridge regression}, 
	description={Consider a \gls{regression} problem where the goal is to learn a \gls{hypothesis} $\hypothesis^{(\weights)}$ 
		for predicting the numeric \gls{label} of a \gls{datapoint} based on its \gls{featurevec}. 
	  	Ridge\index{ridge regression} \gls{regression} learns the \glspl{parameter} $\weights$ 
	  	by minimizing the penalized average \gls{sqerrloss}. The average \gls{sqerrloss} is measured 
		on a set of \glspl{labeled datapoint} (i.e., the \gls{trainset}) 
		$$\pair{\featurevec^{(1)}}{\truelabel^{(1)}}, \,\ldots, \,\pair{\featurevec^{(\samplesize)}}{\truelabel^{(\samplesize)}}.$$ 
		The \gls{penaltyterm} is the scaled squared \gls{euclidnorm} $\regparam \normgeneric{\weights}{2}^{2}$ with a 
		\gls{regularization} \gls{parameter} $\regparam > 0$. The purpose of the \gls{penaltyterm} is \gls{regularization}, 
		i.e., to prevent \gls{overfitting} in the \gls{highdimregime}, where the number of \glspl{feature} 
		$\featuredim$ exceeds the number of \glspl{datapoint} $\samplesize$ in the \gls{trainset}. 
		For the \gls{training} of a \gls{linmodel}, adding $\regparam \normgeneric{\weights}{2}^{2}$ to the average 
		\gls{sqerrloss} is equivalent to computing the average \gls{sqerrloss} on an augmented \gls{trainset}. 
		\begin{figure}[H]
			\begin{center} 
				\begin{tikzpicture}[scale = 1]
					\draw[->, very thick] (0,0.5) -- (7.7,0.5) node[right] {\gls{feature} $\featurevec$};       
					\draw[->, very thick] (0.5,0) -- (0.5,4.2) node[above] {\gls{label} $\truelabel$};   
					\draw[color=black, thick, dashed, domain = -1: 6.2, variable = \x]  plot ({\x},{\x*0.4 + 2.0}) ;     
					\node at (6.7,4.5) {$\hypothesis(\featurevec)$};    
					\coordinate (l1)   at (1.2, 2.48);
					\coordinate (l2) at (1.4, 2.56);
					\coordinate (l3)   at (1.7,  2.68);
					\coordinate (l4)   at (2.2, 2.2*0.4+2.0);
					\coordinate (l5) at (2.4, 2.4*0.4+2.0);
					\coordinate (l6)   at (2.7,  2.7*0.4+2.0);
					\coordinate (l7)   at (3.9,  3.9*0.4+2.0);
					\coordinate (l8) at (4.2, 4.2*0.4+2.0);
					\coordinate (l9)   at (4.5,  4.5*0.4+2.0);
					\coordinate (n1)   at (1.2, 1.8);
					\coordinate (n2) at (1.4, 1.8);
					\coordinate (n3)   at (1.7,  1.8);
					\coordinate (n4)   at (2.2, 3.8);
					\coordinate (n5) at (2.4, 3.8);
					\coordinate (n6)   at (2.7,  3.8);
					\coordinate (n7)   at (3.9, 2.6);
					\coordinate (n8) at (4.2, 2.6);
					\coordinate (n9)   at (4.5,  2.6);
					\node at (n1)  [circle,draw,fill=red,minimum size=6pt,scale=0.6, name=c1] {};
					\node at (n2)  [circle,draw,fill=blue,minimum size=6pt, scale=0.6, name=c2] {};
					\node at (n3)  [circle,draw,fill=red,minimum size=6pt,scale=0.6,  name=c3] {};
					\node at (n4)  [circle,draw,fill=red,minimum size=12pt, scale=0.6, name=c4] {};  
					\node at (n5)  [circle,draw,fill=blue,minimum size=12pt,scale=0.6,  name=c5] {};
					\node at (n6)  [circle,draw,fill=red,minimum size=12pt, scale=0.6, name=c6] {};  
					\node at (n7)  [circle,draw,fill=red,minimum size=12pt,scale=0.6,  name=c7] {};
					\node at (n8)  [circle,draw,fill=blue,minimum size=12pt, scale=0.6, name=c8] {};
					\node at (n9)  [circle,draw,fill=red,minimum size=12pt, scale=0.6, name=c9] {};
					\draw [<->] ($ (n7) + (0,-0.3) $)  --  ($ (n9) + (0,-0.3) $) node [pos=0.4, below] {$\sqrt{\regparam}$}; ; 
					\draw[<->, color=red, thick] (l1) -- (c1);  
					\draw[<->, color=blue, thick] (l2) -- (c2);  
					\draw[<->, color=red, thick] (l3) -- (c3);  
					\draw[<->, color=red, thick] (l4) -- (c4);  
					\draw[<->, color=blue, thick] (l5) -- (c5);  
					\draw[<->, color=red, thick] (l6) -- (c6);  
					\draw[<->, color=red, thick] (l7) -- (c7);  
					\draw[<->, color=blue, thick] (l8) -- (c8);  
					\draw[<->, color=red, thick] (l9) -- (c9);  
					\draw[fill=blue] (6.2, 3.7)  circle (0.1cm) node [anchor=west,black,xshift=0.1cm] {original \gls{datapoint} $\pair{\featurevec}{\truelabel}$};
					\draw[fill=red] (6.2, 3.2)  circle (0.1cm) node [anchor=west,black,xshift=0.1cm] {augmented $\pair{\mvnormal{\featurevec}{\regparam\mathbf{I}}}{\truelabel}$};
				\end{tikzpicture}
			\caption{For a \gls{linmodel}, adding the \gls{penaltyterm} $\regparam \normgeneric{\weights}{2}^2$ to 
				the \gls{objfunc} in \gls{erm} is equivalent to \gls{erm} on an augmented \gls{dataset}.  \label{fig_ridge_regression_dict} }
			\end{center}
		\end{figure} 
		This augmented \gls{trainset} is obtained by replacing each \gls{datapoint} 
		$\pair{\featurevec^{(\sampleidx)}}{\truelabel^{(\sampleidx)}}$ 
		in the original \gls{trainset} by the \gls{realization} of infinitely 
		many \gls{iid} \glspl{rv} whose \gls{probdist} is centered around 
		$\pair{\featurevec^{(\sampleidx)}}{\truelabel^{(\sampleidx)}}$.
	    		\\
		See also: \gls{regression}, \gls{regularization}, \gls{map}, \gls{dataaug}.},
	first={ridge regression},
	text={ridge regression}
}

\newglossaryentry{expectation}
{name={expectation},
	description={Consider\index{expectation} a numeric \gls{featurevec} $\featurevec \in \mathbb{R}^{\featuredim}$ 
		that we interpret as the \gls{realization} of an \gls{rv} with \gls{probdist} $p(\featurevec)$. 
		The expectation of $\featurevec$ is defined as the integral $\expect \{ \featurevec \} \defeq \int \featurevec p(\featurevec)$. 
		Note that the expectation is only defined if this integral exists, i.e., if the \gls{rv} is \gls{integrable} 
		\cite{RudinBookPrinciplesMatheAnalysis}, \cite{BillingsleyProbMeasure}, \cite{HalmosMeasure}. 
		Fig.\ \ref{fig_expect_discrete_dict} illustrates the expectation of a scalar \gls{discreteRV} $x$ that takes on values 
		from a finite set only. 
   		\begin{figure}[H]
   			\begin{center}
   			\begin{tikzpicture}
			\begin{axis}[
			ybar,
			y=5cm,
			x=2cm,                          
			bar width=0.6cm,                   
			xlabel={$x_i$},
			clip=false,
			ylabel={$p(x_i)$},
			y label style={rotate=-90, anchor=west, xshift=-1cm},
			xtick={1,2,3,4,5},
			ymin=0, ymax=0.6,
			grid=both,
			major grid style={gray!20},
			tick align=outside,
			axis line style={black!70},
			]
			\addplot+[ybar, fill=blue!50] coordinates {
			(1,0.1) 
			(2,0.2) 
			(3,0.4) 
			(4,0.2)
			(5,0.1)
			};
			\node[font=\footnotesize,xshift=7pt] at (axis cs:1,0.13) {$p(x_i)\!\,\cdot\!\,x_i\!=\!0.1$};
			\node[font=\footnotesize]at (axis cs:2,0.23) {$0.4$};
			\node[font=\footnotesize]at (axis cs:3,0.43) {$1.2$};
			\node[font=\footnotesize] at (axis cs:4,0.23) {$0.8$};
			\node[font=\footnotesize]at (axis cs:5,0.13) {$0.5$};
			\node[font=\footnotesize]at (axis cs:3.8,0.53) {$\expect\{x\}\!=\!0.1\!+\!0.4\!+\!1.2\!+\!0.8\!+\!0.5\!=\!3$};
			\end{axis}
			\end{tikzpicture}
			\end{center}
			\vspace*{-5mm}
		\caption{Expectation of a \gls{discreteRV} $x$ obtained by summing its possible values $x_{i}$, weighted 
			by the corresponding \gls{probability} $p(x_i) = \prob{x= x_i}$. \label{fig_expect_discrete_dict}}
 		\end{figure}
		See also: \gls{featurevec}, \gls{realization}, \gls{rv}, \gls{probdist}, \gls{probability}.},
	first={expectation},
	plural={expectations},
	text={expectation}
}

\newglossaryentry{logreg}
{name={logistic regression}, 
	description={Logistic\index{logistic regression} \gls{regression} learns a 
		linear \gls{hypothesis} \gls{map} (or \gls{classifier}) $\hypothesis(\featurevec) = \weights\,^{T} \featurevec$ 
		to predict a binary \gls{label} $\truelabel$ based on the numeric 
		\gls{featurevec} $\featurevec$ of a \gls{datapoint} \cite{BishopBook}, \cite{hastie01statisticallearning}. 
		The quality of a linear \gls{hypothesis} \gls{map} is measured by the average \gls{logloss} 
		on some \glspl{labeled datapoint} (i.e., the \gls{trainset}).
				\\
		See also: \gls{regression}, \gls{hypothesis}, \gls{map}, \gls{classifier}, \gls{label}, \gls{featurevec}, \gls{datapoint}, \gls{logloss}, \gls{labeled datapoint}, \gls{trainset}.},
	first={logistic regression},
	text={logistic regression}
}
	
\newglossaryentry{logloss}
{name={logistic loss}, 
	description={Consider\index{logistic loss} 
		a \gls{datapoint} characterized by \glspl{feature} $\featurevec$ and a 
		binary \gls{label} $\truelabel \in \{-1,1\}$. We use a real-valued \gls{hypothesis} 
		$\hypothesis$ to predict the \gls{label} $\truelabel$ from the \glspl{feature} 
		$\featurevec$. The logistic \gls{loss} incurred by this \gls{prediction} is 
		defined as \cite{BishopBook}
		\begin{equation} 
			\label{equ_log_loss_gls_dict}
			\lossfunc{(\featurevec,\truelabel)}{\hypothesis} \defeq  \log\, ( 1 + \exp\,(- \truelabel \hypothesis(\featurevec))).
		\end{equation}
		\begin{figure}[H]
		\begin{center}
			\begin{tikzpicture}
			\begin{axis}[
				axis lines=middle,
				xlabel={$\truelabel\hypothesis(\featurevec)$},
				ylabel={$\lossfunc{(\featurevec,\truelabel)}{\hypothesis}$},
				xlabel style={at={(axis description cs:1.,0.3)}, anchor=north},  
				ylabel style={at={(axis description cs:0.5,1.1)}, anchor=center}, 
				xmin=-3.5, xmax=3.5,
				ymin=-0.5, ymax=2.5,
				xtick={-3, -2, -1, 0, 1, 2, 3},
				ytick={0, 1, 2},
				domain=-3:3,
				samples=100,
				width=10cm, height=6cm,
				grid=both,
				major grid style={line width=.2pt, draw=gray!50},
				minor grid style={line width=.1pt, draw=gray!20},
				legend pos=south west 
				]
					\addplot [red, thick] {ln(1 + exp(-x))};    
			\end{axis}
			\end{tikzpicture}
		\caption{Logistic \gls{loss} incurred by the \gls{prediction} $\hypothesis(\featurevec) \in \mathbb{R}$ 
			for a \gls{datapoint} with \gls{label} $\truelabel \in \{-1,1\}$.}
			\label{fig_logloss_dict}
		\end{center}
		\end{figure}
		Note that the expression \eqref{equ_log_loss_gls_dict} 
		for the logistic \gls{loss} applies only for the \gls{labelspace} $\labelspace = \{ -1,1\}$ 
		and when using the thresholding rule \eqref{equ_def_threshold_bin_classifier_dict}. 
		\\
		See also: \gls{loss}, \gls{classification}, \gls{classifier}, \gls{linmodel}.},
	first={logistic loss},
	text={logistic loss}
}
	
\newglossaryentry{hingeloss}
{name={hinge loss}, 
	description={Consider\index{hinge loss} a \gls{datapoint} 
		characterized by a \gls{featurevec} $\featurevec \in \mathbb{R}^{\featuredim}$ and a 
		binary \gls{label} $\truelabel \in \{-1,1\}$. The hinge \gls{loss} incurred by a real-valued 
		\gls{hypothesis} \gls{map} $\hypothesis(\featurevec)$ is defined as 
		\begin{equation} 
			\label{equ_hinge_loss_gls_dict}
			\lossfunc{(\featurevec,\truelabel)}{\hypothesis} \defeq \max \{ 0 , 1 - \truelabel \hypothesis(\featurevec) \}. 
		\end{equation}
		\begin{figure}[H]
		\begin{center}
		\begin{tikzpicture}
    			\begin{axis}[
        			axis lines=middle,
        			xlabel={$\truelabel\hypothesis(\featurevec)$},
        			ylabel={$\lossfunc{(\featurevec,\truelabel)}{\hypothesis}$},
 			xlabel style={at={(axis description cs:1.,0.3)}, anchor=north},  
        			ylabel style={at={(axis description cs:0.5,1.1)}, anchor=center}, 
        			xmin=-3.5, xmax=3.5,
        			ymin=-0.5, ymax=2.5,
        			xtick={-3, -2, -1, 0, 1, 2, 3},
        			ytick={0, 1, 2},
        			domain=-3:3,
        			samples=100,
        			width=10cm, height=6cm,
        			grid=both,
        			major grid style={line width=.2pt, draw=gray!50},
        			minor grid style={line width=.1pt, draw=gray!20},
        			legend pos=south west 
    			]
        			\addplot[blue, thick] {max(0, 1-x)};
    			\end{axis}
		\end{tikzpicture}
		\caption{Hinge \gls{loss} incurred by the \gls{prediction} $\hypothesis(\featurevec) \in \mathbb{R}$ 
			for a \gls{datapoint} with \gls{label} $\truelabel \in \{-1,1\}$. A regularized variant of the hinge 
			\gls{loss} is used by the \gls{svm} \cite{LampertNowKernel}.}
			\label{fig_hingeloss_dict}
		\end{center}
		\end{figure} 	    
		See also: \gls{svm}, \gls{classification}, \gls{classifier}.},
	first={hinge loss},
	text={hinge loss}
}

\newglossaryentry{iidasspt}
{name={independent and identically distributed assumption (i.i.d.\ assumption)}, 
	description={The \gls{iid} assumption\index{independent and identically distributed assumption (i.i.d.\ assumption)} 
		is a widely used \gls{probmodel} for the generation of \glspl{datapoint}. 
		In particular, \glspl{datapoint} are represented as \gls{iid} \glspl{rv}.
				\\
		See also: \gls{iid}, \gls{probmodel}, \gls{datapoint}, \gls{rv}.},
	first={independent and identically distributed assumption (i.i.d.\ assumption)},
	text={i.i.d.\ assumption} 
}

\newglossaryentry{hypospace}
{name={hypothesis space},
	description={A \gls{hypothesis} space\index{hypothesis space} is a mathematical \gls{model} 
		that characterizes the learning capacity of an \gls{ml} method. The goal of such a method is 
		to learn a \gls{hypothesis} \gls{map} that maps \glspl{feature} of a \gls{datapoint} to a 
		\gls{prediction} of its \gls{label}. Given a finite amount of computational resources, a 
		practical \gls{ml} method typically explores only a restricted set of all possible \glspl{map} 
		from the \gls{featurespace} to the \gls{labelspace}. Such a restricted set is referred to as 
		a \gls{hypothesis} space $\hypospace$ underlying the \gls{ml} method (see Fig.\ \ref{fig_hypospace_dict}).
    		For the analysis of a given \gls{ml} method, the choice of a \gls{hypothesis} space $\hypospace$ is not 
		unique, i.e., any superset containing all \glspl{map} the method can learn is also a valid \gls{hypothesis} space. 
		\begin{figure}[H]
		\begin{center}
			\begin{tikzpicture}[allow upside down, scale=0.4]
			\node [below] at (5,-3) {$\labelspace^{\featurespace}$};
			\draw [ultra thick] (5,0) circle (5cm);
			\draw [ultra thick,fill=black!20] (5,0) circle (1cm);
			\node [] at (5,0) {$\hypospace$};
			\end{tikzpicture}
		\end{center}
		\caption{\Gls{hypothesis} space $\hypospace$ of an \gls{ml} method as a (typically very small) 
			subset of the (typically very large) set $\labelspace^{\featurespace}$ of all possible \glspl{map} 
			from the \gls{featurespace} $\featurespace$ into the \gls{labelspace} $\labelspace$. \label{fig_hypospace_dict}}
		\end{figure}
		On the other hand, from an \gls{ml} engineering perspective, the \gls{hypothesis} space $\hypospace$ is a design 
		choice for \gls{erm}-based methods. This design choice can be guided by the available computational 
		resources and \glspl{statasp}. For instance, if efficient \gls{matrix} operations are feasible and a 
		roughly linear relation exists between \glspl{feature} and \glspl{label}, a \gls{linmodel} can be a 
		useful choice for $\hypospace$.
				\\
		See also: \gls{hypothesis}, \gls{model}, \gls{map}, \gls{linmodel}.},
	first={hypothesis space},
	plural={hypothesis spaces}, 
	firstplural={hypothesis spaces}, 
	text={hypothesis space} 
}
	
\newglossaryentry{model}
{name={model (machine learning)},
sort={modelmachine},
	description={The study and design of \gls{ml} methods is often based on a mathematical model\index{model} 
		\cite{bender1978mathematical}. One of the most widely used examples of a mathematical model for \gls{ml} is a \gls{hypospace}. 
		A \gls{hypospace} consists of \gls{hypothesis} \glspl{map} that are used by an \gls{ml} method to 
		predict \glspl{label} from the \glspl{feature} of \glspl{datapoint}. Another important type of 
		mathematical model is a \gls{probmodel}, which consists of \glspl{probdist} that describe 
		how \glspl{datapoint} are generated. Unless stated otherwise, we use the term model to 
		refer specifically to the \gls{hypospace} underlying an \gls{ml} method. We illustrate one example 
		of a \gls{hypospace} and a \gls{probmodel} in Fig.\ \ref{fig_model_dict}.
		\begin{figure}[H]
			\centering
			\begin{tikzpicture}[scale=1]
			\draw[->] (-1,0) -- (3,0) node[right] {$\feature$};
			\draw[->] (0,-1) -- (0,3) node[above] {$\truelabel$};
			\draw[thick, red] (-0.5,0) -- (2.5,2) node[right] {$\hypothesis^{(1)}$};
			\draw[thick, blue] (-0.5,1) -- (2.5,1) node[right] {$\hypothesis^{(2)}$};
			\draw[thick, green!60!black] (-0.5,2) -- (2.5,0.5) node[right] {$\hypothesis^{(3)}$};
			\node at (1.5,-1.2) {(a)};
			\end{tikzpicture}
			\hspace{2cm}
			\begin{tikzpicture}[scale=1]
			\draw[->] (-1,0) -- (3,0) node[right] {$\feature$};
			\draw[->] (0,-1) -- (0,3) node[above] {$\truelabel$};
			\draw[thick, red] (1,1) ellipse [x radius=1, y radius=0.5];
			\draw[thick, blue] (2,2) ellipse [x radius=0.7, y radius=0.3];
			\node[red] at (1,0.3) {$p^{(1)}(\feature,\truelabel)$};
			\node[blue] at (2,2.7) {$p^{(2)}(\feature,\truelabel)$};
			\node at (1.5,-1.2) {(b)};
			\end{tikzpicture}
		\caption{Two types of mathematical models used in \gls{ml}. (a) \Gls{hypospace} consisting of 
			three \glspl{linearmap}. (b) \Gls{probmodel} consisting of a \gls{probdist} 
			over the plane spanned by the possible values of the \gls{feature} and \gls{label} 
			of a \gls{datapoint}. \label{fig_model_dict}}
		\end{figure}
		See also: \gls{hypospace}, \gls{probmodel}, \gls{probdist}.},
	first={model},
	plural={models},
	text={model} 
}

\newglossaryentry{modelparam}
{name={model parameter}, 
sort={modelparameter}, 
	description={The elements of a \gls{parammodel} are specified by 
		quantities that are referred to as \gls{model} \glspl{parameter}\index{model parameter}. 
    		In the context of \gls{ml}, a \gls{parammodel} consists of \gls{hypothesis} 
		\glspl{map} that are specified by a list of \gls{model} \glspl{parameter} 
		$\weight_{1}, \,\weight_{2}, \,\ldots, \,\weight_{\nrfeatures}$. It is often 
		convenient to stack these \gls{model} \glspl{parameter} 
		into a \gls{vector} 
		$\weights=\big(\weight_{1},\,\ldots,\,\weight_{\nrfeatures}\big)^{T} \in \mathbb{R}^{\nrfeatures}$.
		\begin{figure}[H]
			\centering
			\begin{tikzpicture}[scale=1]
  				\draw (0,0) ellipse (1.6 and 1.1);
  				\node at (0,1.5) {\gls{paramspace} $\subseteq \mathbb{R}^{\nrfeatures}$};
  				\fill (-0.6,0.2) circle (1.5pt) node[above left] {$\weights'$};
  				\fill (0.7,-0.3) circle (1.5pt) node[below right] {$\weights''$};
  				\draw (6,0) ellipse (1.8 and 1.2);
  				\node at (6,1.6) {\gls{model} $\hypospace$};
  				\fill (5.4,0.1) circle (1.5pt) node[above left] {$\hypothesis^{(\weights')}$};
  				\fill (6.7,-0.2) circle (1.5pt) node[below right] {$\hypothesis^{(\weights'')}$};
  				\draw[->] (-0.6,0.2) .. controls (2,0.9) .. (5.4,0.1);
  				\draw[->] (0.7,-0.3) .. controls (2,-0.9) .. (6.7,-0.2);
  				\node at (3,1.2) {$\weights \mapsto \hypothesis^{(\weights)}$};
			\end{tikzpicture}
		\caption{\Gls{model} \glspl{parameter} $\weights$ selecting a well-defined 
			\gls{hypothesis} $\hypothesis^{(\weights)}$ out of the \gls{model} $\hypospace$.}
		\end{figure}
    		We can think of \gls{model} \glspl{parameter} as an identifier for 
		a \gls{hypothesis} \gls{map}, similar to how a social security number 
		identifies a person.
		\\
		See also: \gls{model}, \gls{parameter}, \gls{hypothesis}, \gls{map}.},
	first={model parameter},
	plural={model parameters},
	firstplural={model parameters},
	text={model parameter} 
}

\newglossaryentry{generalizedadditivemodel}
{name={generalized additive model (GAM)},
	description={A GAM\index{generalized additive model (GAM)} 
		is obtained from a \gls{linmodel} by replacing the original \glspl{feature} 
		$\feature_\featureidx$, for $\featureidx=1,\,\ldots,\,\nrfeatures$, 
		of a \gls{datapoint} with nonlinear \glspl{function} 
		$\featuremap_{\featureidx}\big(\feature_\featureidx\big)$ \cite{Hastie1986}.
		More formally, a GAM consists of \gls{hypothesis} \glspl{map} of the form
		$$\hypothesis(\featurevec) = \weight_0 + \sum_{\featureidx=1}^{\nrfeatures} \weight_{\featureidx} 
		\featuremap_{\featureidx}\big(\feature_\featureidx\big).$$
		See also: \gls{linmodel}, \gls{feature}, \gls{function}. },
 	first={generalized additive model (GAM)},
 	plural={GAMs},
 	firstplural={generalized additive models (GAMs)},
 	text={GAM}
}

\newglossaryentry{brierscore}
{name={Brier score},
	description={The Brier score\index{Brier score} evaluates probabilistic \glspl{prediction} for binary \glspl{outcome}. 
		Consider $\samplesize$ \glspl{datapoint}, indexed by $\sampleidx=1,\ldots,\samplesize$, 
		each representing a binary \gls{outcome} $\truelabel^{(\sampleidx)} \in \{0,1\}$. 
		Let $\predictedlabel^{(\sampleidx)} \in [0,1]$ denote the predicted \gls{probability} 
		of success for the $\sampleidx$th \gls{datapoint}. The Brier score is defined as the 
		average squared deviation between the predicted \gls{probability} $\predictedlabel^{(\sampleidx)}$ 
		and the observed \gls{outcome} $\truelabel^{(\sampleidx)}$ \cite{Brier1950}, i.e., 
		\[
			\frac{1}{\samplesize}\sum_{\sampleidx=1}^{\samplesize} (\predictedlabel^{(\sampleidx)} - \truelabel^{(\sampleidx)})^2.
		\]
		See also: \gls{prediction}, \glspl{outcome}, \gls{probability}. },
	first={Brier score},
	plural={Brier scores},
	firstplural={Brier scores},
	text={Brier score}
}

\newglossaryentry{ai}
{name={artificial intelligence (AI)}, 
	description={AI\index{artificial intelligence (AI)} refers to systems that behave rationally, in the sense of 
		maximizing a long-term \gls{reward}. The \gls{ml}-based approach to AI is to train a \gls{model} to  
		predict optimal actions. These \glspl{prediction} are computed from observations about the \gls{state} of the 
		\gls{environment}. The choice of \gls{lossfunc} sets AI applications apart from more basic \gls{ml} applications. 
		\Glspl{aisystem} rarely have access to a labeled \gls{trainset} that allows the average \gls{loss} to be 
		measured for any possible choice of \glspl{modelparam}. Instead, \glspl{aisystem} use observed \gls{reward} 
		signals to estimate the \gls{loss} incurred by the current choice of \glspl{modelparam}.
				\\
		See also: \gls{ml}, \gls{aisystem}, \gls{reinforcementlearning}.},
	first={AI},
	text={AI} 
}
 
\newglossaryentry{reward}
{name={reward}, 
	description={A reward refers to some\index{reward} observed 
		(or measured) quantity that allows us to estimate the \gls{loss} incurred by the \gls{prediction} 
		(or decision) of a \gls{hypothesis} $\hypothesis(\featurevec)$. For example, in an 
		\gls{ml} application to self-driving vehicles, $\hypothesis(\featurevec)$ could represent 
		the current steering direction of a vehicle. We could construct a reward from the 
		measurements of a collision sensor that indicate if the vehicle is moving toward 
		an obstacle. We define a low reward for the steering direction 
		$\hypothesis(\featurevec)$ if the vehicle moves dangerously toward an obstacle.
			\\
		See also: \gls{loss}, \gls{mab}, \gls{reinforcementlearning}.},
	first={reward}, 
	text={reward}
} 

\newglossaryentry{clusteringerror}
{name={clustering error}, 
	description={Consider a \gls{clustering} method\index{clustering error} 
		that decomposes a given \gls{dataset} into \glspl{cluster}. 
		The \gls{clustering} error is a quantitative \gls{measure} of the 
		usefulness of the \glspl{cluster}. Different \gls{clustering} 
		methods use different choices for the \gls{clustering} error. 
	 	For example, the \gls{hardclustering} method \gls{kmeans} measures 
		the \gls{clustering} error via the average squared \gls{eucliddist} 
	 	between the \gls{featurevec} of a \gls{datapoint} 
	 	and the nearest \gls{clustercentroid} (see Fig.\ \ref{fig_clustering_error_dict}). 
	 	Another construction for the \gls{clustering} error can be based on a \gls{probmodel} 
	 	such as the \gls{gmm} where the \glspl{clustercentroid} are \glspl{parameter} 
	 	of the underlying \gls{probdist}.
		\begin{figure}[H]
			\centering
			\begin{tikzpicture}[scale=1]
			\coordinate (c1) at (0.8,0.7);
			\fill[black] ($(c1)+(-0.05,-0.05)$) rectangle ($(c1)+(0.05,0.05)$);
			\node[above left] at (c1) {$\clustercentroid{1}$};
			\coordinate (c2) at (6.6,1.6);
			\fill[black] ($(c2)+(-0.05,-0.05)$) rectangle ($(c2)+(0.05,0.05)$);
			\node[above left] at (c2) {$\clustercentroid{2}$};
			\begin{scope}[shift={(c1)}, xscale=1.3, yscale=1.3, rotate=0]
				\foreach \dx/\dy in {-0.6/-0.4, 0.1/0.9, 0.7/-0.6} {
				\fill (\dx,\dy) circle (1.5pt);
				\draw[dashed, gray] (\dx,\dy) -- (0,0);
				}
			\end{scope}
			\begin{scope}[shift={(c2)}, xscale=1.5, yscale=1.5, rotate=0]
				\foreach \dx/\dy in {-1.1/-0.5, -0.2/0.6, 0.6/-0.2} {
				\fill (\dx,\dy) circle (1.5pt);
				\draw[dashed, gray] (\dx,\dy) -- (0,0);
				}
			\end{scope}
			\end{tikzpicture}
		\caption{For \glspl{datapoint} with numeric \glspl{featurevec}, we can 
	     		use the average squared \gls{eucliddist} to the nearest 
			\glspl{clustercentroid} as a \gls{measure} of the \gls{clustering} error.\label{fig_clustering_error_dict}}
		\end{figure}
	 	See also: \gls{kmeans}, \gls{probmodel}, \gls{gmm}, \gls{maxlikelihood}. }, 
	first={clustering error}, 
	firstplural={clustering errors}, 
	plural={clustering errors}, 
	text={clustering error}
}

\newglossaryentry{hardclustering}
{name={hard clustering}, 
	description={Hard \gls{clustering}\index{hard clustering} 
		refers to the task of partitioning a given set of \glspl{datapoint} 
		into (a few) nonoverlapping \glspl{cluster}. This requirement 
		allows us to represent a \gls{cluster} by a subset of \glspl{datapoint}, 
		i.e., precisely those belonging to the \gls{cluster}. In contrast to 
		hard \gls{clustering}, \gls{softclustering} methods allow for overlapping 
		\glspl{cluster} and specify, for each \gls{datapoint}, a numeric \gls{dob} 
		to each \gls{cluster}. Hard \gls{clustering} is an extreme case of \gls{softclustering} 
		where the \glspl{dob} take only two values, indicating either no belonging or 
		full belonging. 
		For \glspl{datapoint} characterized by numeric \glspl{featurevec} $\featurevec \in \mathbb{R}^{\nrfeatures}$, 
		a widely used hard \gls{clustering} method is \gls{kmeans}. Any hard
		\gls{clustering} method for numeric \glspl{featurevec} $\featurevec \in \mathbb{R}^{\nrfeatures}$ can 
		be adapted to nonnumerical \gls{data} using \gls{featlearn} methods. 
		One important example of this approach is \gls{specclustering}, where 
		\glspl{datapoint} have a similarity structure in the form of an  
		\gls{undirectedgraph}. The nodes of this \gls{graph} represent \glspl{datapoint} 
		while undirected (possibly weighted) edges represent similarities (and their 
		extend) between \glspl{datapoint}. We can then use the entries of the 
		\glspl{eigenvector} of the \gls{LapMat} as numeric \glspl{feature} for 
		each \gls{datapoint}. 
		\\
		See also: \gls{clustering}, \gls{datapoint}, \gls{cluster}, \gls{kmeans}.},
	first={hard clustering},
	text={hard clustering} 
}
	
\newglossaryentry{softclustering}
{name={soft clustering}, 
	description={Soft \gls{clustering}\index{soft clustering} 
		refers to the task of partitioning a given set of \glspl{datapoint} into (a few) 
		overlapping \glspl{cluster}. Each \gls{datapoint} is assigned to several different 
		\glspl{cluster} with varying \glspl{dob}. Soft \gls{clustering} methods determine the 
		\gls{dob} (or soft \gls{cluster} assignment) for each \gls{datapoint} and each \gls{cluster}.
		A principled approach to soft \gls{clustering} for \glspl{datapoint} characterized by 
		numerical \glspl{featurevec} is via a \gls{probmodel} such as the \gls{gmm}. 
		The conditional \gls{probability} of a \gls{datapoint} belonging to a specific mixture 
		component is then a natural choice for the \gls{dob}. \Gls{gmm} soft \gls{clustering} 
		methods can be applied to nonnumeric \gls{data} by using \gls{featlearn} methods 
		to provide numerical \glspl{feature} (such as in \gls{specclustering}). 
				\\
		See also: \gls{clustering}, \gls{cluster}, \gls{dob}, \gls{gmm}, \gls{specclustering}.},
	first={soft clustering},
	text={soft clustering} 
}

\newglossaryentry{kroneckerproduct}
{name={Kronecker product}, 
	description={The Kronecker product\index{Kronecker product} of two \glspl{matrix} $\mA \in \mathbb{R}^{m \times n}$ 
		and $\mB \in \mathbb{R}^{p \times q}$ is a block \gls{matrix} denoted by $\mA \otimes \mB$ 
	 	and defined as \cite{GolubVanLoanBook}, \cite{HornMatAnalysis}
    		\[
      		\mA \otimes \mB =
      		\begin{pmatrix}
        		a_{11}\mB & \cdots & a_{1n}\mB \\
        		\vdots & \ddots & \vdots \\
        		a_{m1}\mB & \cdots & a_{mn}\mB
      		\end{pmatrix}
      		\in \mathbb{R}^{mp \times nq}.
    		\]
    		The Kronecker product is a special case of the tensor product for \glspl{matrix} and is 
		widely used in multivariate statistics, linear algebra, and structured \gls{ml} \glspl{model}. 
		It satisfies the identity $(\mA \otimes \mB)(\vx \otimes \vy) = (\mA\vx) \otimes (\mB\vy)$ 
		for \glspl{vector} $\vx$ and $\vy$ of compatible dimensions.
		\\
		See also: \gls{matrix}, \gls{ml}, \gls{model}, \gls{vector}. },
	first={Kronecker product},
	text={Kronecker product} 
}
	
\newglossaryentry{clustering}
{name={clustering}, 
	description={Clustering\index{clustering} methods decompose a given 
		set of \glspl{datapoint} into a few subsets, which are referred to as \glspl{cluster}. 
		Each \gls{cluster} consists of \glspl{datapoint} that are more similar to each 
		other than to \glspl{datapoint} outside the \gls{cluster}. Different clustering methods 
		use different \glspl{measure} for the similarity between \glspl{datapoint} and different 
		forms of \gls{cluster} representations. The clustering method \gls{kmeans} uses the 
		average \gls{featurevec} of a \gls{cluster} (i.e., the \gls{cluster} \gls{mean}) as its representative. 
		A popular \gls{softclustering} method based on \gls{gmm} represents 
		a \gls{cluster} by a \gls{mvndist}.
				\\
		See also: \gls{cluster}, \gls{kmeans}, \gls{softclustering}, \gls{gmm}.},
	first={clustering},
	text={clustering} 
}
	
\newglossaryentry{cluster}
{name={cluster},
	description={A\index{cluster} cluster is a subset of 
		\glspl{datapoint} that are more similar to each other than to the \glspl{datapoint} outside the cluster. 
		The quantitative \gls{measure} of similarity between \glspl{datapoint} is a design choice. If \glspl{datapoint} 
		are characterized by Euclidean \glspl{featurevec} $\featurevec \in \mathbb{R}^{\nrfeatures}$, 
		we can define the similarity between two \glspl{datapoint} via the \gls{eucliddist} between 
		their \glspl{featurevec}. An example of such clusters is shown in Fig.\ \ref{fig:clusters_dict}.
		\begin{figure}[H]
		\centering
		\begin{tikzpicture}
		\pgfplotsset{compat=1.18}
		\begin{axis}[
		    width=10cm,
		    height=8cm,
		    xlabel={$x_1$},
		    ylabel={$x_2$},
		    title={Clusters of \glspl{datapoint}},
		    xmin=0, xmax=10,
		    ymin=0, ymax=10,
		    axis lines=left,
		    legend style={at={(0.5,-0.25)}, anchor=north, legend columns=3}
		]
		\addplot[only marks, color=blue, mark=*, mark size=3pt] coordinates {
		    (1,1) (2,1.2) (1.8,2) (2.2,1.5) (1.5,2.5)
		};
		\addplot[only marks, color=red, mark=square*, mark size=3pt] coordinates {
		    (7,8) (8,7.5) (7.5,8.5) (8.2,7.8) (7.7,7)
		};
		\addplot[only marks, color=green!60!black, mark=triangle*, mark size=3pt] coordinates {
		    (5,3) (5.5,3.2) (5.2,2.8) (4.8,3.5) (5.1,3.1)
		};
		\legend{Cluster 1, Cluster 2, Cluster 3}
		\end{axis}
		\end{tikzpicture}
		\caption{Illustration of three clusters in a 2-D \gls{featurespace}. Each cluster groups \glspl{datapoint} 
			that are more similar to each other than to those in other clusters based on the \gls{eucliddist}.}
			\label{fig:clusters_dict}
		\end{figure}
		See also: \gls{datapoint}, \gls{featurevec}, \gls{eucliddist}, \gls{featurespace}.},
	first={cluster},
	plural={clusters}, 
	text={cluster} 
}

\newglossaryentry{huberloss}
{name={Huber loss}, 
	description={The\index{Huber loss} 
		Huber \gls{loss} unifies the \gls{sqerrloss} and the \gls{abserr}.
				\\
		See also: \gls{loss}, \gls{sqerrloss}, \gls{abserr}.},
	first={Huber loss},
	text={Huber loss} 
}

\newglossaryentry{svm}
{name={support vector machine (SVM)}, 
	description={The\index{support vector machine (SVM)} 
		SVM is a \gls{binclass} meth\-od that 
		learns a linear \gls{hypothesis} \gls{map}. Thus, like \gls{linreg} and \gls{logreg}, 
		it is also an instance of \gls{erm} for the \gls{linmodel}. However, the 
		SVM uses a different \gls{lossfunc} from the one used in those methods. As illustrated in 
		Fig.\ \ref{fig_svm_gls_dict}, it aims to maximally separate \glspl{datapoint} from 
		the two different classes in the \gls{featurespace} (i.e., \gls{maximum} margin principle). 
		Maximizing this separation is equivalent to minimizing a regularized 
		variant of the \gls{hingeloss} \eqref{equ_hinge_loss_gls_dict} \cite{BishopBook}, \cite{LampertNowKernel}, \cite{Cristianini_Shawe-Taylor_2000}.
		\begin{figure}[H]
			\begin{center}
				\begin{tikzpicture}[auto,scale=0.8]
					\draw [thick] (1,2) circle (0.1cm)node[anchor=west] {\hspace*{0mm}$\featurevec^{(5)}$};
					\draw [thick] (0,1.6) circle (0.1cm)node[anchor=west] {\hspace*{0mm}$\featurevec^{(4)}$};
					\draw [thick] (0,3) circle (0.1cm)node[anchor=west] {\hspace*{0mm}$\featurevec^{(3)}$};
					\draw [thick] (2,1) circle (0.1cm)node[anchor=east,above] {\hspace*{0mm}$\featurevec^{(6)}$};
					\node[] (B) at (-2,0) {support \gls{vector}};
					\draw[->,dashed] (B) to (1.9,1) ; 
					\draw [|<->|,thick] (2.05,0.95)  -- (2.75,0.25)node[pos=0.5] {$\xi$} ; 
					\draw [thick] (1,-1.5) -- (4,1.5) node [right] {$\hypothesis^{(\weights)}$} ; 
					\draw [thick] (3,-1.9) rectangle ++(0.1cm,0.1cm) node[anchor=west,above]  {\hspace*{0mm}$\featurevec^{(2)}$};
					\draw [thick] (4,.-1) rectangle ++(0.1cm,0.1cm) node[anchor=west,above] {\hspace*{0mm}$\featurevec^{(1)}$};
				\end{tikzpicture}
		\caption{SVM learning a \gls{hypothesis} (or \gls{classifier}) $\hypothesis^{(\weights)}$ with 
			minimal average soft-margin \gls{hingeloss}. Minimizing this \gls{loss} is equivalent 
			to maximizing the margin $\xi$ between the \gls{decisionboundary} of $\hypothesis^{(\weights)}$ 
			and each class of the \gls{trainset}.}
			\label{fig_svm_gls_dict}
			\end{center}
		\end{figure}
		The above basic variant of SVM is only useful if the \glspl{datapoint} from different categories can be  
		(approximately) linearly separated. 
				\\
		See also: \gls{classification}, \gls{linmodel}, \gls{classifier}, \gls{hingeloss}.},
	first={support vector machine (SVM)},
	text={SVM} 
}

\newglossaryentry{cvxclustering}
{name={convex clustering}, 
 	description={Consider\index{convex clustering} a \gls{dataset} 
 		$\featurevec^{(1)}, \,\ldots, \,\featurevec^{(\samplesize)} \in \mathbb{R}^{\nrfeatures}$. 
 		\Gls{convex} \gls{clustering} learns \glspl{vector} $\weights^{(1)}, \,\ldots, \,\weights^{(\samplesize)}$ by minimizing:
 		$$\sum_{\sampleidx=1}^{\samplesize} \normgeneric{\featurevec^{(\sampleidx)} - \weights^{(\sampleidx)}}{2}^2 + 
 		\regparam \sum_{\nodeidx,\nodeidx' \in \nodes} \normgeneric{\weights^{(\nodeidx)} - \weights^{(\nodeidx')}}{p}.$$ 
		Here, $\normgeneric{\vu}{p} \defeq \big( \sum_{\featureidx=1}^{\dimlocalmodel} |u_{\featureidx}|^{p} \big)^{1/p}$ 
		denotes the $p$-\gls{norm} (for $p\geq1$).  
		It turns out that many of the optimal \glspl{vector} $\widehat{\weights}^{(1)}, \,\ldots, \,\widehat{\weights}^{(\samplesize)}$ 
		coincide. A \gls{cluster} then consists of those \glspl{datapoint} $\sampleidx \in \{1, \,\ldots, \,\samplesize\}$ 
		with identical $\widehat{\weights}^{(\sampleidx)}$ \cite{JMLR:v22:18-694}, \cite{Pelckmans2005}. 
			\\
		See also: \gls{dataset}, \gls{convex}, \gls{clustering}, \gls{vector}, \gls{norm}, \gls{cluster}, \gls{datapoint}. },
 	first={convex clustering},
	text={convex clustering} 
}

\newglossaryentry{gdmethod}
{name={gradient-based method}, 
sort={gradientbased method}, 
	description={A \gls{gradient}-based\index{gradient-based method} 
		method is an iterative technique for finding the \gls{minimum} (or \gls{maximum}) 
		of a \gls{differentiable} \gls{objfunc} $f(\weights)$ of the \glspl{modelparam} $\weights$. 
		Such a method constructs a \gls{sequence} of approximations to an optimal choice for $\weights$. 
		As the name indicates, a \gls{gradient}-based method uses the \glspl{gradient} 
		of the \gls{objfunc} evaluated during previous \glspl{iteration} to construct new, 
		(hopefully) improved \glspl{modelparam}. One important example of a \gls{gradient}-based 
		method is \gls{gd}.
				\\
		See also: \gls{gradient}, \gls{differentiable}, \gls{objfunc}, \gls{gd}, \gls{optmethod}.},
	first={gradient-based method},
	firstplural={gradient-based methods},
	plural={gradient-based methods},
	text={gradient-based method} 
}

\newglossaryentry{sgd}
{name={subgradient descent}, 
	description={\Gls{subgradient}\index{subgradient descent} 
		descent is a \gls{generalization} of \gls{gd} that does not require differentiability of the 
		\gls{function} to be minimized. This \gls{generalization} is obtained by replacing the concept 
		of a \gls{gradient} with that of a \gls{subgradient}. Similar to \glspl{gradient}, \glspl{subgradient} 
		allow us to construct local approximations of an \gls{objfunc}. The \gls{objfunc} 
		might be the \gls{emprisk} $\emperror\big( \hypothesis^{(\weights)} \big| \dataset \big)$ viewed 
		as a \gls{function} of the \glspl{modelparam} $\weights$ that select a \gls{hypothesis} $\hypothesis^{(\weights)} \in \hypospace$.
				\\
		See also: \gls{subgradient}, \gls{generalization}, \gls{gd}, \gls{function}, \gls{gradient}, \gls{objfunc}, \gls{emprisk}, \gls{modelparam}, \gls{hypothesis}.},
	first={subgradient descent},
	text={subgradient descent} 
}
	
\newglossaryentry{stochGD}
{name={stochastic gradient descent (SGD)}, 
	description={SGD\index{stochastic gradient descent (SGD)} 
		is obtained from \gls{gd} by replacing the \gls{gradient} of the \gls{objfunc} 
		with a \gls{stochastic} approximation. A main application of SGD
		is to train a parameterized \gls{model} via \gls{erm} on a \gls{trainset} $\dataset$ that 
		is either very large or not readily available (e.g., when \glspl{datapoint} are stored 
		in a database distributed globally). To evaluate the \gls{gradient} of the 
		\gls{emprisk} (as a \gls{function} of the \glspl{modelparam} $\weights$), 
		we need to compute a sum $\sum_{\sampleidx=1}^{\samplesize} \nabla_{\weights} \lossfunc{\datapoint^{(\sampleidx)}}{\weights}$  
		over all \glspl{datapoint} in the \gls{trainset}. We obtain a \gls{stochastic} 
		approximation to the \gls{gradient} by replacing the sum $\sum_{\sampleidx=1}^{\samplesize} \nabla_{\weights} \lossfunc{\datapoint^{(\sampleidx)}}{\weights}$ 
		with a sum $\sum_{\sampleidx \in \batch} \nabla_{\weights} \lossfunc{\datapoint^{(\sampleidx)}}{\weights}$ 
		over a randomly chosen subset $\batch \subseteq \{1, \,\ldots, \,\samplesize\}$ (see Fig.\ \ref{fig_sgd_approx_dict}). 
		We often refer to these randomly chosen \glspl{datapoint} as a \gls{batch}. 
		The \gls{batch} size $|\batch|$ is an important \gls{parameter} of SGD. 
		SGD with $|\batch|> 1$ is referred to as mini-\gls{batch} SGD \cite{Bottou99}. 		
		\begin{figure}[H]
			\centering
			\begin{tikzpicture}[scale=1.5, >=stealth]
				\draw[thick, blue, domain=0.5:2.5, samples=100] plot (\x, {(\x-1.5)^2 + 1});
				\node[blue,above] at (0.5, 2) {$\sum\limits_{\sampleidx=1}^{\samplesize}$};
				\draw[thick, red, domain=1:3, samples=100] plot (\x, {(\x-2)^2 + 0.5});
				\node[red] at (3.3, 1.5) {$\sum\limits_{\sampleidx \in \batch}$};
			\end{tikzpicture}
		\caption{SGD for \gls{erm} approximates the \gls{gradient} by replacing the sum 
		$\sum_{\sampleidx=1}^{\samplesize} \nabla_{\weights} \lossfunc{\datapoint^{(\sampleidx)}}{\weights}$
		over all \glspl{datapoint} in the \gls{trainset} (indexed by $\sampleidx=1, \,\ldots, \,\samplesize$) 
		with a sum over a randomly chosen subset $\batch \subseteq \{1, \,\ldots, \,\samplesize\}$.\label{fig_sgd_approx_dict}}
		\end{figure}
		See also: \gls{gd}, \gls{gradient}, \gls{objfunc}, \gls{stochastic}, \gls{model}, \gls{erm}, \gls{trainset}, \gls{datapoint}, \gls{emprisk}, \gls{function}, \gls{modelparam}, \gls{batch}, \gls{parameter}.},
	first={stochastic gradient descent (SGD)},
	text={SGD} 
}

\newglossaryentry{onlineGD}
{name={online gradient descent (online GD)}, 
	description={Consider \index{online gradient descent (online GD)} an \gls{ml} 
		method that learns \glspl{modelparam} $\weights$ from some 
		\gls{paramspace} $\paramspace \subseteq \mathbb{R}^{\dimlocalmodel}$. 
		The learning process uses \glspl{datapoint} $\datapoint^{(\timeidx)}$ 
		that arrive at consecutive time instants $\timeidx=1, \,2, \,\dots$. 
		Let us interpret the (generation of) \glspl{datapoint} $\datapoint^{(\timeidx)}$ 
		as \gls{iid} \glspl{rv} with a common \gls{probdist} $\probdist^{(\datapoint)}$. 
		The \gls{risk} $\expect\{ \lossfunc{\datapoint}{\weights} \}$ of a 
		\gls{hypothesis} $\hypothesis^{(\weights)}$ can then (under mild conditions) 
		be obtained as the limit:
		$$\lim_{T\rightarrow \infty} \frac{1}{T}\sum_{\timeidx=1}^{T} \lossfunc{\datapoint^{(\timeidx)}}{\weights}.$$ 
		We might use this limit as the \gls{objfunc} for learning the 
		\glspl{modelparam} $\weights$. Unfortunately, the above limit can 
		only be evaluated if we wait infinitely long in order to collect all \glspl{datapoint}. 
		However, many \gls{ml} applications require methods that learn online, i.e.,
		as soon as a new \gls{datapoint} $\datapoint^{(\timeidx)}$ arrives at 
		time $\timeidx$, we update the current \glspl{modelparam} $\weights^{(\timeidx)}$. 
		Note that the new \gls{datapoint} $\datapoint^{(\timeidx)}$ contributes 
		the component $\lossfunc{\datapoint^{(\timeidx)}}{\weights}$ to the \gls{risk}. 
		As its name suggests, online \gls{gd} updates $\weights^{(\timeidx)}$ via 
		a (projected) \gls{gradstep} such that
		\begin{equation} 
			\label{equ_def_ogd_dict}
 			\weights^{(\timeidx+1)} \defeq \projection{\paramspace}{\weights^{(\timeidx)} - \lrate_{\timeidx} \nabla_{\weights} \lossfunc{\datapoint^{(\timeidx)}}{\weights}}. 
		\end{equation} 
		Note that \eqref{equ_def_ogd_dict} is a \gls{gradstep} for the 
		current component $\lossfunc{\datapoint^{(\timeidx)}}{\cdot}$ 
		of the \gls{risk}. The update \eqref{equ_def_ogd_dict} ignores all 
		previous components $\lossfunc{\datapoint^{(\timeidx')}}{\cdot}$
		for $\timeidx' < \timeidx$. It might therefore happen that, compared 
		with $\weights^{(\timeidx)}$, the updated \glspl{modelparam} 
		$\weights^{(\timeidx+1)}$ increase the retrospective average 
		\gls{loss} $\sum_{\timeidx'=1}^{\timeidx-1} \lossfunc{\datapoint^{(\timeidx')}}{\cdot}$. 
		However, for a suitably chosen \gls{learnrate} $\lrate_{\timeidx}$, 
		online \gls{gd} can be shown to be optimal in practically relevant 
		settings. By optimal, we mean that the \glspl{modelparam} 
		$\weights^{(T+1)}$ delivered by online \gls{gd} after observing 
		$T$ \glspl{datapoint} $\datapoint^{(1)}, \,\ldots, \,\datapoint^{(T)}$ 
		are at least as good as those delivered by any other learning method 
		\cite{HazanOCO}, \cite{GDOptimalRakhlin2012}. 
		\begin{figure}[H]
		\begin{center}
		\begin{tikzpicture}[x=1.5cm,scale=1.5, every node/.style={font=\footnotesize}]
			\draw[->] (0.5, 0) -- (5.5, 0) node[below] {};
			\foreach \x in {1, 2, 3, 4, 5} {
				\draw (\x, 0.1) -- (\x, -0.1) node[below] {$t=\x$};
			}
			\foreach \x/\y in {1/2.5, 2/1.8, 3/2.3, 4/1.5, 5/2.0} {
				\fill[black] (\x, \y) circle (2pt) node[above right] {$\datapoint^{(\x)}$};
			}
			\foreach \x/\y in {1/1.0, 2/1.6, 3/1.8, 4/2.2, 5/1.9} {
				\fill[blue] (\x, \y) circle (2pt) node[below left] {$\weights^{(\x)}$};
			}
			\foreach \x/\y/\z in {1/2.5/1.0, 2/1.8/1.6, 3/2.3/2.0, 4/1.5/1.8, 5/2.0/1.9} {
				\draw[dashed, gray] (\x, \y) -- (\x, \z);
			}
		\end{tikzpicture}
		\end{center} 
		\caption{Instance of online \gls{gd} updating the \glspl{modelparam} $\weights^{(\timeidx)}$ 
			using the \gls{datapoint} $\datapoint^{(\timeidx)} = \feature^{(\timeidx)}$ arriving at time $\timeidx$. 
			This instance uses the \gls{sqerrloss} $\lossfunc{\datapoint^{(\timeidx)}}{\weight} = (\feature^{(\timeidx)} - \weight)^{2}$.}
		\end{figure}
		See also: \gls{objfunc}, \gls{gd}, \gls{gradstep}, \gls{onlinelearning}.},
	first={online gradient descent (online GD)},
	text={online GD}
}

\newglossaryentry{pca}
{name={principal component analysis (PCA)}, 
	description={Consider a \gls{dataset} 
		$$\dataset = \big\{ \featurevec^{(1)}, \,\ldots, \,\featurevec^{(\samplesize)} \big\}$$
		consisting of \glspl{datapoint} characterized by \glspl{featurevec} 
		$\featurevec^{(\sampleidx)} \in \mathbb{R}^{\nrfeatures}$ for $\sampleidx=1,\,\ldots,\,\samplesize$.
		PCA\index{principal component analysis (PCA)} determines, for a given number 
		$\nrfeatures' < \nrfeatures$, a linear \gls{featuremap} 
		$$\featuremapvec^{(\mW)}: \mathbb{R}^{\nrfeatures} \rightarrow 
		\mathbb{R}^{\nrfeatures'}: \featurevec \mapsto \mW \featurevec$$ 
		such that the new \glspl{featurevec} 
		$\mathbf{z}^{(\sampleidx)} = \mW \featurevec^{(\sampleidx)}$ 
		allow us to reconstruct the original \glspl{feature} with  
		\gls{minimum} linear reconstruction error \cite{MLBasics}, \cite{BishopBook}, \cite{hastie01statisticallearning}:
		$$\min_{\mR \in \mathbb{R}^{\nrfeatures \times \nrfeatures'}} 
		\sum_{\sampleidx=1}^{\samplesize} 
		\normgeneric{\featurevec^{(\sampleidx)}- \mR \mW \featurevec^{(\sampleidx)}}{2}^{2}.$$ 
		We can view PCA as a form of \gls{erm} using the \gls{lossfunc} 
		$\lossfunc{\featurevec}{\mW} = \normgeneric{\featurevec- \widehat{\mR} \mW \featurevec}{2}^{2}$ with a 
		reconstruction \gls{matrix} $\widehat{\mR}$ 
		that achieves the above \gls{minimum} reconstruction error. It turns 
		out that this \gls{erm} problem can be solved by a \gls{matrix}  
		$\mW = \big(\eigvecCov^{(1)},\,\ldots,\,\eigvecCov^{(\nrfeatures')}\big)^{T}$ 
		whose rows are given by $\nrfeatures'$ 
		\glspl{eigenvector} corresponding to the $\nrfeatures'$ largest 
		\glspl{eigenvalue} of the \gls{matrix}:
		$$\widehat{\mQ} = \frac{1}{\samplesize} \sum_{\sampleidx=1}^{\samplesize} 
		\featurevec^{(\sampleidx)} \big( \featurevec^{(\sampleidx)}\big)^{T}
		=\featuremtx^{T} \featuremtx.$$ 
		Note that $\widehat{\mQ}$ coincides with the \gls{samplecovmtx} of 
		$\dataset$ if its \gls{samplemean} vanishes. The \gls{psd} \gls{matrix} 
		$\widehat{\mQ}$ allows for an \gls{evd} of the following form
		\cite{HornMatAnalysis}, \cite{MeyerMatrixAnalysis}:
		$$\widehat{\mQ}\!=\!\sum_{\featureidx=1}^{\nrfeatures} \eigval{\featureidx} 
			\eigvecCov^{(\nrfeatures)} \big( \eigvecCov^{(\nrfeatures)} \big)^{T}=
			\begin{pmatrix}
			\eigvecCov^{(1)} & \cdots & \eigvecCov^{(\nrfeatures)}
			\end{pmatrix}
			\begin{pmatrix}
			\eigval{1}        &        & 0 \\
			& \ddots &   \\
			0                 &        & \eigval{\nrfeatures}
			\end{pmatrix}
			\begin{pmatrix}
			\big(\eigvecCov^{(1)}\big)^{T} \\
			\vdots                         \\
			\big(\eigvecCov^{(\nrfeatures)}\big)^{T}
		\end{pmatrix}.$$
		This decomposition consists of decreasing nonnegative \glspl{eigenvalue} 
		$\eigval{1} \geq \eigval{2} \geq \,\ldots \,\geq \eigval{\nrfeatures} 
		\geq 0$ and corresponding \glspl{eigenvector} $\eigvecCov^{(1)},\,\ldots,\,\eigvecCov^{(\nrfeatures)}$ 
		that form an orthonormal \gls{basis} of $\mathbb{R}^{\nrfeatures}$.
				 \\
		See also: \gls{featuremap}, \gls{featlearn}, \gls{dimred}, \gls{evd}.},
	first={principal component analysis (PCA)},
	text={PCA} 
}
	
\newglossaryentry{loss}
{name={loss}, 
	description={\gls{ml}\index{loss} methods use a 
		\gls{lossfunc} $\lossfunc{\datapoint}{\hypothesis}$ to measure the error incurred 
		by applying a specific \gls{hypothesis} to a specific \gls{datapoint}. With a
		slight abuse of notation, we use the term loss for both the \gls{lossfunc} $\loss$ 
		itself and the specific value $\lossfunc{\datapoint}{\hypothesis}$ for a \gls{datapoint} $\datapoint$ 
		and \gls{hypothesis} $\hypothesis$.
				\\
		See also: \gls{lossfunc}, \gls{emprisk}.},
	first={loss},
	plural={losses},
	firstplural={losses},
	text={loss} 
}

\newglossaryentry{lossfunc}
{name={loss function}, 
	description={A\index{loss function} \gls{loss} \gls{function} is a \gls{map}:
		$$\lossfun: \featurespace \times \labelspace \times \hypospace \rightarrow \mathbb{R}_{+}: \big( \big(\featurevec,\truelabel\big),
		 \hypothesis\big) \mapsto  \lossfunc{(\featurevec,\truelabel)}{\hypothesis}.$$
		It assigns a nonnegative real number (i.e., the \gls{loss}) $\lossfunc{(\featurevec,\truelabel)}{\hypothesis}$
		to a pair that consists of a \gls{datapoint}, with \glspl{feature} $\featurevec$ 
		and \gls{label} $\truelabel$, and a \gls{hypothesis} $\hypothesis \in \hypospace$. The 
		value $\lossfunc{(\featurevec,\truelabel)}{\hypothesis}$ quantifies the discrepancy 
		between the true \gls{label} $\truelabel$ and the \gls{prediction} $\hypothesis(\featurevec)$. 
		Lower (closer to zero) values $\lossfunc{(\featurevec,\truelabel)}{\hypothesis}$ indicate a smaller 
		discrepancy between \gls{prediction} $\hypothesis(\featurevec)$ and \gls{label} $\truelabel$. 
		Fig.\ \ref{fig_loss_function_gls_dict} depicts a \gls{loss} \gls{function} for a given \gls{datapoint}, 
		with \glspl{feature} $\featurevec$ and \gls{label} $\truelabel$, as a \gls{function} of the \gls{hypothesis} $\hypothesis \in \hypospace$. 
		\begin{figure}[H]
			\begin{center}
				\begin{tikzpicture}[scale = 0.7,
        				every axis/.append style={
          			axis line style={-Latex, thick},
          			tick style={thick}
        				}]
					\begin{axis}
						[axis x line=center,
						axis y line=center,
						xlabel={},
						xlabel style={below right},
						ylabel style={above right},
						xtick=\empty,
						ytick=\empty,
						xmin=-5,
						xscale = 1.4, 
						xmax=5,
						ymin=-0.5,
						ymax=2.5
						]
  						\addplot[red, line width=0.5mm,thick] {ln(1 + exp(-x))};
  						\addplot[blue, line width=0.5mm, dashed, domain=-4:4] {0.3*abs(x)};
  						\addplot[green!50!black, line width=0.5mm, dotted, domain=-4:4] {0.3*x^2};
					\end{axis}
					\node [below] at (10,1) {$\hypothesis(\featurevec)$};
					\node [right] at (4,6) {$\lossfunc{(\featurevec,\truelabel)}{\hypothesis}$};
				\end{tikzpicture}
			\end{center}
			\vspace*{-7mm}
		\caption{Some \gls{loss} \gls{function} $\lossfunc{(\featurevec,\truelabel)}{\hypothesis}$ for a fixed \gls{datapoint}, with 
			\gls{featurevec} $\featurevec$ and \gls{label} $\truelabel$, and a varying \gls{hypothesis} $\hypothesis$. 
			\gls{ml} methods try to find (or learn) a \gls{hypothesis} that incurs minimal \gls{loss}.}
			\label{fig_loss_function_gls_dict}
		\end{figure}
		See also: \gls{loss}, \gls{label}, \gls{featurevec}, \gls{erm}.},
 	first={loss function},
 	text={loss function} 
}

\newglossaryentry{covariate}
{name={covariate},
	description={See\index{covariate} \gls{feature}.}, 
	first={covariate},
	firstplural={covariates},
	plural={covariates}, 
	text={covariate}
}

\newglossaryentry{deeplearning}
{name={deep learning},
	description={See\index{deep learning} \gls{deepnet}.},
	first={deep learning},
	text={deep learning}
}

\newglossaryentry{decisiontree}
{name={decision tree},
	description={A\index{decision tree} decision tree is a flowchart-like representation 
		of a \gls{hypothesis} \gls{map} $\hypothesis$. More formally, a decision tree 
		is a \gls{dag} containing a root node that reads in the \gls{featurevec} 
		$\featurevec$ of a \gls{datapoint}. The root node then forwards 
		the \gls{datapoint} to one of its child nodes based on some elementary test on the \glspl{feature} $\featurevec$. 
		If the receiving child node is not a leaf node, i.e., it has child nodes itself, 
	  	it represents another test. Based on the test result, the \gls{datapoint} is forwarded 
	   	to one of its descendants. This testing and forwarding of the \gls{datapoint} is continued 
	  	until the \gls{datapoint} ends up in a leaf node without any children. See Fig.\ \ref{fig_decision_tree_dict} for visual illustrations.
		\begin{figure}[H]
		\begin{minipage}{.45\textwidth}
		\scalebox{1}{
		\begin{tikzpicture}
			\node[fill=black, circle, inner sep=2pt, label=above:{$\| \featurevec-\mathbf{u} \| \leq \varepsilon$?}] (A) {};	
			\node[fill=black, circle, inner sep=2pt, below left=1.5cm and 1cm of A, label=left:{$\hypothesis(\featurevec) = \predictedlabel_1$}] (B) {};
			\node[fill=black, circle, inner sep=2pt, below right=1.5cm and 1cm of A, label=right:{$\| \featurevec - \mathbf{v} \| \leq \varepsilon$?}] (C) {};
			\node[fill=black, circle, inner sep=2pt, below left=1.5cm and 1cm of C, label=left:{$\hypothesis(\featurevec) = \predictedlabel_2$}] (D) {};
			\node[fill=black, circle, inner sep=2pt, below right=1.5cm and 1cm of C, label=right:{$\hypothesis(\featurevec) =\predictedlabel_3$}] (E) {};
			\draw[line width=1.5pt, ->] (A) -- (B) node[midway, left] {no};
			\draw[line width=1.5pt, ->] (A) -- (C) node[midway, right] {yes};
			\draw[line width=1.5pt, ->] (C) -- (D) node[midway, left] {no};
			\draw[line width=1.5pt, ->] (C) -- (E) node[midway, right] {yes};
			\node at (0.7,-4.5) {\selectfont (a)};
		\end{tikzpicture}
		}
		\end{minipage}	
		\hspace*{15mm}
		\begin{minipage}{.45\textwidth}
		\hspace*{15mm}
		\begin{tikzpicture}
		\draw (-2,2) rectangle (2,-2);
		\begin{scope}
			\clip (-0.5,0) circle (1cm);
			\clip (0.5,0) circle (1cm);
			\fill[color=gray] (-2,1.5) rectangle (2,-1.5);
		\end{scope}
		\draw (-0.5,0) circle (1cm);
		\draw (0.5,0) circle (1cm);
		\draw[fill] (-0.5,0) circle [radius=0.025];
		\node [below right, red] at (-0.5,0) {$\predictedlabel_{3}$};
		\node [below left, blue] at (-0.7,0) {$\predictedlabel_{2}$};
		\node [above left] at (-0.7,1) {$\predictedlabel_{1}$};
		\node [left] at (-0.4,0) {$\mathbf{u}$};
		\draw[fill] (0.5,0) circle [radius=0.025];
		\node [right] at (0.6,0) {$\mathbf{v}$};
		\node at (0,-3.5) {\selectfont (b)};
		\end{tikzpicture}
		\end{minipage}
		\caption{(a) Decision tree as a flowchart-like representation of a piecewise constant \gls{hypothesis} 
			$\hypothesis: \featurespace \rightarrow \mathbb{R}$. Each piece is a \gls{decisionregion} 
			$\decreg{\predictedlabel} \defeq \big\{ \featurevec \in  \featurespace: \hypothesis(\featurevec) = \predictedlabel \big\}$. 
			The depicted decision tree can be applied to numeric \glspl{featurevec}, i.e., $\featurespace \subseteq \mathbb{R}^{\dimlocalmodel}$. 
			It is parameterized by the threshold $\varepsilon>0$ and the \glspl{vector} $\vu, \vv \in \mathbb{R}^{\dimlocalmodel}$. 
			(b) Decision tree partitioning the \gls{featurespace} $\featurespace$ into \glspl{decisionregion}. Each \gls{decisionregion}  
			$\decreg{\hat{\truelabel}} \!\subseteq\!\featurespace$ corresponds to a specific leaf node in the decision tree.}
			\label{fig_decision_tree_dict}
		\end{figure} 
		See also: \gls{decisionregion}.},
	  first={decision tree},
	  plural={decision trees},
	  text={decision tree} 
}

\newglossaryentry{samplesize}
{name={sample size},
	description={The \gls{sample} size\index{sample size} is the number of individual \glspl{datapoint} contained in a 
	    	\gls{sample} or \gls{dataset}. Consider an \gls{erm}-based method that uses a 
		\gls{trainset} $\dataset$ with \gls{sample} size $\samplesize$ 
		and a \gls{model} $\hypospace$ with \gls{effdim} $\effdim{\hypospace}$. If the \gls{trainset} can 
		be well approximated by the \gls{iidasspt}, then the ratio between $\samplesize$ 
		and $\effdim{\hypospace}$ can be a useful indicator of the occurrence of \gls{overfitting} \cite[Ch. 6]{MLBasics}.
				\\
		See also: \gls{datapoint}, \gls{dataset}.},
	first={sample size},
	text={sample size}
}

\newglossaryentry{skipconnection}
{name={skip connection},
	description={Consider a \gls{deepnet}\index{skip connection} with \glspl{neuron} that are organized 
		in consecutive \glspl{layer}. A skip connection links the \gls{output} of a \gls{neuron} in 
		some \gls{layer} to the input of a \gls{neuron} in a nonconsecutive \gls{layer} \cite{HeResidual2016}.
		\\
		See also: \gls{deepnet}, \gls{dag}.},
 	first={skip connection},
 	firstplural={skip connections}, 
 	plural={skip connections}, 
 	text={skip connection}
}

\newglossaryentry{ann}
{name={artificial neural network (ANN)}, 
	description={An\index{artificial neural network (ANN)} ANN is a graphical (signal-flow) 
		representation of a \gls{hypothesis} that maps \glspl{feature} of a 
		\gls{datapoint} at its input to a \gls{prediction} for the corresponding 
		\gls{label} at its \gls{output}. The fundamental computational unit of an 
		ANN is the \gls{artificialneuron}, which applies an \gls{actfun} $\actfun(\cdot)$ 
		to the weighted sum of its inputs. The \gls{output} of a \gls{neuron} can be used 
		either as the final \gls{output} of the ANN or as an input to other \glspl{neuron}. 
		A key design aspect of an ANN is its connectivity structure (or architecture), 
		i.e., which \gls{neuron} \glspl{output} are connected to which other \gls{neuron}'s inputs. 
		As illustrated in Fig.\ \ref{fig_ANN_DAG_dict}, we can represent an 
		ANN as a \gls{dag}.
    		\begin{figure}[H]
 		\centering
 		\begin{tikzpicture}[>=stealth, node distance=2.3cm and 2.4cm]
  			\node[circle, fill=black, inner sep=1.5pt, label=left:{$\feature_1$}] (x1) {};
  			\node[circle, fill=black, inner sep=1.5pt, label=left:{$\feature_2$}] (x2) [below=of x1] {};
  			\node[circle, fill=black, inner sep=1.5pt, label=above:{$a_{1}$}] (h1) [right=of x1, yshift=5mm] {};
  			\node[circle, fill=black, inner sep=1.5pt, label=below:{$a_{2}$}] (h2) [right=of x2, yshift=-5mm] {};
  			\node[circle, fill=black, inner sep=1.5pt, label=right:{$\hypothesis^{(\weights)}(\featurevec)$}] (y) [right=of h1, yshift=-2cm] {};
  			\draw[->] (x1) -- (h1) node [midway, above] {$w_{1}$};
  			\draw[->] (x2) -- (h1)node [pos=0.1, above] {$w_{2}$};
  			\draw[->] (x1) -- (h2) node [pos=0.8, above] {$w_{3}$};
  			\draw[->] (x2) -- (h2) node [midway, above] {$w_{4}$};
  			\draw[->] (h1) -- (y) node [midway, above] {$w_{5}$};
  			\draw[->] (h2) -- (y) node [midway, above] {$w_{6}$};
			\draw[->] (x1) -- (y) node [midway, above] {$w_{7}$};
 		\end{tikzpicture}
 		\caption{ANN represented as a weighted \gls{dag} with 
			nodes corresponding to \glspl{neuron} or \glspl{feature} of a \gls{datapoint}. \Glspl{feature} 
			can be viewed as trivial \glspl{neuron} without input and with a fixed \gls{output} given by the 
			\gls{feature} value. The weighted directed edges indicate how \gls{neuron} \glspl{output} are used 
			as inputs to other \glspl{neuron}. The \glspl{edgeweight} are tunable \glspl{modelparam} and are 
			used to scale the inputs to the \glspl{neuron}. The \gls{output} of some \glspl{neuron} is used as the 
			\gls{prediction} $\hypothesis^{(\weights)}(\featurevec)$. \label{fig_ANN_DAG_dict}}
 		\end{figure}
		One widely used type of ANN is \glspl{deepnet} 
		where \glspl{neuron} form consecutive \glspl{layer}. In a \gls{deepnet}, the 
		\glspl{output} of \glspl{neuron} in a given \gls{layer} are typically only connected 
		to the inputs of the \glspl{neuron} in a consecutive \gls{layer}. Sometimes it 
		is useful to add shortcut or \glspl{skipconnection} that directly connect the \glspl{output} of 
		\glspl{neuron} in one \gls{layer} to the inputs of \glspl{neuron} in a nonconsecutive \gls{layer} 
		\cite{Goodfellow-et-al-2016}, \cite{HeResidual2016}.	
		\\
		See also: \gls{label}, \gls{artificialneuron}, \gls{actfun}, \gls{deepnet}, \gls{layer}.},
	first={artificial neural network (ANN)},
	plural={ANNs},
	text={ANN}
}

\newglossaryentry{neuron}
{name={neuron (machine learning)},
	description={See\index{neuron} \gls{artificialneuron}.},
	first={neuron},
	firstplural={neurons},
	plural={neurons},
	text={neuron}
}

\newglossaryentry{artificialneuron}
{name={artificial neuron},
	description={An\index{artificial neuron} artificial neuron is the basic building block 
 		of an \gls{ann} \cite{Goodfellow-et-al-2016}. In the context of \gls{ml}, an artificial 
		neuron is often also referred to as neuron. A neuron applies a (typically nonlinear) 
		\gls{actfun} to a weighted sum of numeric input values.
		In particular, a neuron with $\nrfeatures$ inputs $z_{1},\,\ldots,\,z_{\nrfeatures}$ 
		computes 
		\[
		a = \actfun\!\Big( \sum_{\featureidx=1}^{\nrfeatures} \weight_\featureidx z_{\featureidx} + b \Big).
		\]
		Here, $\weight_\featureidx$ and $b$ are tunable \glspl{modelparam} that are 
		learned during the \gls{training} of the \gls{ann}. 
		The inputs of a neuron typically correspond to \glspl{feature} of a 
		\gls{datapoint} or to \glspl{output} of other neurons within an \gls{ann}. 
		The resulting \gls{output} can serve as input to further neurons 
		or contribute directly to the \gls{prediction} 
		$\hypothesis^{(\weights)}(\featurevec)$ delivered by an \gls{ann}.
		\begin{figure}[H]
		\centering
			\begin{tikzpicture}[>=stealth, node distance=1.8cm]
			\node[circle, fill=black, inner sep=1.5pt, 
			label=left:{$z_{1}$}] (z1) {};
			\node[circle, fill=black, inner sep=1.5pt, 
			label=left:{$z_{2}$}] (z2) [below=of z1] {};
			\node[circle, fill=black, inner sep=1.5pt, 
			label=left:{$z_{\nrfeatures}$}] (zm) [below=of z2, yshift=-2mm] {};
			\node[circle, draw, minimum size=12mm, 
			right=2.5cm of z2] (neuron) {$g(\cdot)$};
			\node[circle, fill=black, inner sep=1.5pt, 
			label=right:{$a$}] (out) 
			[right=2.5cm of neuron] {};
			\draw[->] (z1) -- (neuron) node[midway, above] {$w_{1}$};
			\draw[->] (z2) -- (neuron) node[midway, above] {$w_{2}$};
			\draw[->] (zm) -- (neuron) node[midway, above] {$w_{\nrfeatures}$};
			\draw[->] (neuron) -- (out);
			\node at ($(neuron)+(0,-1.4)$) {$+\,b$};
			\end{tikzpicture}
              \caption{Artificial neuron as a weighted nonlinear mapping.}
              \end{figure}
              See also: \gls{ann}, \gls{actfun}, \gls{layer}.},
	first={artificial neuron},
	firstplural={artificial neurons},
	plural={artificial neurons},
	text={artificial neuron}
}

\newglossaryentry{randomforest}
{name={random forest},
	description={A\index{random forest} random forest is a set of different \glspl{decisiontree}. 
		Each of these \glspl{decisiontree} is obtained by fitting a perturbed copy of 
		the original \gls{dataset}.
				\\
		See also: \gls{decisiontree}, \gls{dataset}.},
	first={random forest}, 
	text={random forest}
}

\newglossaryentry{gd}
{name={gradient descent (GD)},
sort={gradientdescent},
	description={GD\index{gradient descent (GD)} is an iterative method for 
	 	finding the \gls{minimum} of a \gls{differentiable} \gls{function} $f: \mathbb{R}^{\nrfeatures} \rightarrow \mathbb{R}$. 
	 	GD generates a \gls{sequence} of estimates $\weights^{(0)}, \,\weights^{(1)}, \,\weights^{(2)}, \,\ldots$ 
	 	that (ideally) converge to a \gls{minimum} of $f$. At each \gls{iteration} $\itercntr$, GD refines 
	 	the current estimate $\weights^{(\itercntr)}$ by taking a step in the direction of the
	 	steepest descent of a local linear approximation. This direction is given by the negative 
	 	\gls{gradient} $\nabla f(\weights^{(\itercntr)})$ of the \gls{function} $f$ at the current estimate $\weights^{(\itercntr)}$. 
	 	The resulting update rule is given by
		\begin{equation} 
			\label{equ_def_GD_step_dict}
			\weights^{(\itercntr\!+\!1)} = \weights^{(\itercntr)} - \lrate \nabla f(\weights^{(\itercntr)}),
		\end{equation} 
		where $\lrate > 0$ is a suitably small \gls{stepsize}. For a suitably choosen \gls{stepsize} $\lrate$, 
		the update typically reduces the \gls{function} value, i.e., $f(\weights^{(\itercntr+1)}) < f(\weights^{(\itercntr)})$. 
		Fig.\ \ref{fig_basic_GD_step_dict} illustrates a single GD step.
		\begin{figure}[H]
			\begin{center}
				\begin{tikzpicture}[scale=0.9]
					\draw[loosely dotted] (-4,0) grid (4,4);
					\draw[blue, ultra thick, domain=-4.1:4.1] plot (\x,  {(1/4)*\x*\x});
					\draw[red, thick, domain=2:4.7] plot (\x,  {2*\x - 4});
					\draw[<-] (4,4) -- node[right] {$\big( \Delta \weights\big)\,^{T} \nabla f(\weights^{(\itercntr)})$} (4,2);
					\draw[->] (4,4) -- node[above] {$-\lrate \nabla f(\weights^{(\itercntr)})$} (1,4);
					\draw[<-] (4,2) -- node[below] {$\Delta \weights$} (3,2) ;
					\draw[->] (-4.25,0) -- (4.25,0) node[right] {$\weights$};
					\draw[->] (0,-2pt) -- (0,4.25) node[above] {$f(\weights)$};
					\draw[shift={(0,0)}] (0pt,2pt) -- (0pt,-2pt) node[below] {$\overline{\weights}$};
					\draw[shift={(4,0)}] (0pt,2pt) -- (0pt,-2pt) node[below] {$\weights^{(\itercntr)}$};
					\draw[shift={(1,0)}] (0pt,2pt) -- (0pt,-2pt) node[below] {$\weights^{(\itercntr\!+\!1)}$};
					\foreach \y/\ytext in {1/1, 2/2, 3/3, 4/4}
					\draw[shift={(0,\y)}] (2pt,0pt) -- (-2pt,0pt) node[left] {$\ytext$};  
				\end{tikzpicture}
			\end{center}
		\caption{Single \gls{gradstep} \eqref{equ_def_GD_step_dict} toward the minimizer $\overline{\weights}$ of $f(\weights)$.}
			\label{fig_basic_GD_step_dict}
		\end{figure}
		See also: \gls{minimum}, \gls{differentiable}, \gls{gradient}, \gls{stepsize}, \gls{gradstep}. },
	first={gradient descent (GD)},
	text={GD}
}

\newglossaryentry{abserr}
{name={absolute error loss},
	description={Consider a \gls{datapoint} with \glspl{feature} $\featurevec \in \featurespace$ and numeric 
		\gls{label} $\truelabel \in \mathbb{R}$. As its name suggests, the absolute error \gls{loss}\index{absolute error loss} 
		incurred by a \gls{hypothesis} $\hypothesis: \featurespace \rightarrow \mathbb{R}$ 
		is defined as $$\lossfunc{\pair{\featurevec}{\truelabel}}{\hypothesis} = |\truelabel - \hypothesis(\featurevec)|.$$
		Fig.\ \ref{fig_abs_err_dict} depicts the absolute error \gls{loss} for a fixed \gls{datapoint} 
		with \gls{featurevec} $\featurevec$ and \gls{label} $\truelabel$. It also indicates the \gls{loss} values 
		incurred by two different \glspl{hypothesis} $\hypothesis'$ and $\hypothesis''$. Similar to the \gls{sqerrloss}, 
		the absolute error \gls{loss} is also a \gls{convex} \gls{function} of the \gls{prediction} $\predictedlabel=\hypothesis(\featurevec)$. 
		However, in contrast to the \gls{sqerrloss}, the absolute error \gls{loss} is \gls{nonsmooth}, as it is not 
		\gls{differentiable} at the optimal \gls{prediction} $\predictedlabel=\truelabel$. This property 
		makes \gls{erm}-based methods using the absolute error \gls{loss} computationally more demanding \cite{nesterov04}, \cite{OptMLBook}. 
	    	To build intuition, it is useful to consider the two \glspl{hypothesis} depicted in Fig.\ \ref{fig_abs_err_dict}. 
	    	Just by inspecting the slope of $\loss$ around $\hypothesis'(\featurevec)$ and $\hypothesis''(\featurevec)$,
	     	it is impossible to determine whether we are very close to the optimum (at $\hypothesis'$) or still far away (at $\hypothesis''$). 
	     	As a result, any \gls{optmethod} that is based on local approximations of the \gls{lossfunc} (such as \gls{sgd}) 
	     	must use a decreasing \gls{learnrate} to avoid overshooting when approaching the optimum. This required 
	     	decrease in \gls{learnrate} tends to slow down the \gls{convergence} of the 
	     	\gls{optmethod}. Besides the increased computational complexity, using absolute error \gls{loss} in \gls{erm} can 
	     	be beneficial in the presence of \glspl{outlier} in the \gls{trainset}. In contrast to the \gls{sqerrloss}, 
	     	the slope of the absolute error \gls{loss} does not increase with increasing \gls{prediction} error 
	     	$\truelabel - \hypothesis(\featurevec)$. As a result, the effect of introducing an \gls{outlier} with 
	     	large \gls{prediction} error on the solution $\learnthypothesis$ of \gls{erm} with absolute error \gls{loss} 
	     	is much smaller compared with the effect on the solution of \gls{erm} with \gls{sqerrloss}. 
		\begin{figure}[H]
		\begin{center}
		\begin{tikzpicture}[x=3cm,y=1.6cm]
			\pgfmathsetmacro{\ytrue}{0.0}
			\pgfmathsetmacro{\hp}{0.6}
			\pgfmathsetmacro{\hpp}{3.7}
			\begin{axis}[axis lines=middle,xtick=\empty,ytick=\empty,width=15cm,height=5cm,xmin=-4,xmax=4,ymin=-0.2,ymax=3,domain=-4:4,samples=100,clip=false,enlarge x limits=0.15,enlarge y limits=0.15]			
				\addplot[very thick, red] {abs(x - \ytrue)};
				\addplot[only marks, mark=*, mark options={fill=white}] coordinates {(\ytrue, 0)};
				\node[anchor=west,below,xshift=-2mm] at (axis cs:\ytrue, 0) {$\truelabel$};
				\pgfmathsetmacro{\Lhp}{abs(\hp - \ytrue)}
				\addplot[only marks, mark=*, mark options={scale=1.1}] coordinates {(\hp, \Lhp)};
				\node[anchor=west] at (axis cs:\hp, \Lhp) {$\lossfunc{\pair{\featurevec}{\truelabel}}{\hypothesis'}$};
				\node[yshift=-4mm] at (axis cs:\hp, 0) {$\hypothesis'(\featurevec)$};
				\addplot[densely dashed] coordinates {(\hp, 0) (\hp, \Lhp)};
				\pgfmathsetmacro{\Lhpp}{abs(\hpp - \ytrue)}
				\addplot[only marks, mark=*, mark options={scale=1.1}] coordinates {(\hpp, \Lhpp)};
				\node[yshift=-4mm] at (axis cs:\hpp, 0) {$\hypothesis''(\featurevec)$};
				\addplot[densely dashed] coordinates {(\hpp, 0) (\hpp, \Lhpp)};
				\node[anchor=west] at (axis cs:\hpp, \Lhpp) {$\lossfunc{\pair{\featurevec}{\truelabel}}{\hypothesis''}$};
				\node[xshift=18mm] at (axis cs:4,0) {$\hat{\truelabel}$};
			\end{axis}
		\end{tikzpicture}
		\end{center}
		\caption{For a \gls{datapoint} with numeric \gls{label} $\truelabel\!\in\!\mathbb{R}$, the absolute 
			error $|\truelabel\!-\!\hypothesis(\featurevec)|$ can be used as a \gls{lossfunc} to guide the 
			learning of a \gls{hypothesis} $\hypothesis$. \label{fig_abs_err_dict}}
		\end{figure} 
		See also: \gls{datapoint}, \gls{feature}, \gls{label}, \gls{loss}, \gls{erm}, \gls{sgd}, \gls{ladregression}.},
	first={absolute error loss},
	text={absolute error loss}
}

\newglossaryentry{device}
{name={device},
	description={The term device\index{device} refers to a physical system that can store 
		and process \gls{data}. In the context of \gls{ml}, 
		the term typically refers to a computer capable of 
		reading \glspl{datapoint} from different sources and 
		using them to train an \gls{ml} \gls{model} \cite{Patterson2013}.
						\\
		See also: \gls{data}, \gls{ml}, \gls{datapoint}, \gls{model}.},
	first={device},
	plural={devices}, 
	text={device}
}

\newglossaryentry{huberreg}
{name={Huber regression},
	description={Huber \gls{regression}\index{Huber regression} \cite{huber1992robust} refers to 
		\gls{erm}-based methods that use the \gls{huberloss} as a \gls{measure} of the \gls{prediction} error. 
		Two important special cases of Huber \gls{regression} are \gls{ladregression} and 
		\gls{linreg}. Tuning the threshold \gls{parameter} of the \gls{huberloss} allows the user
		to trade the \gls{robustness} of the \gls{abserr} 
		against the computational benefits of the \gls{smooth} \gls{sqerrloss}.
					\\
		See also: \gls{ladregression}, \gls{linreg}, \gls{abserr}, \gls{sqerrloss}.},
	first={Huber regression},
	text={Huber regression}
}

\newglossaryentry{ladregression}
{name={least absolute deviation regression},
	description={Least\index{least absolute deviation regression} absolute 
		deviation \gls{regression} is an instance of \gls{erm} using the \gls{abserr}. 
		It is a special case of \gls{huberreg}. 
		For the \gls{parammodel} $\hypothesis^{(\weight)}(\featurevec) = \weight$, 
		\gls{erm} with \gls{abserr} is solved by the \gls{median} (see Fig.\ \ref{fig_lad_dict}). For the same 
		\gls{parammodel}, using the \gls{sqerrloss} instead makes \gls{erm} compute the \gls{mean}. 
		\begin{figure}[H]
		\centering
		\begin{tikzpicture}[scale=0.7, y=0.5cm, x=0.5cm]
			\begin{scope}
				\foreach \x/\y in {
					1/2, 4/3, 7/4
				} {
					\draw[dashed, gray] (\x, 0) -- (\x, \y);
					\filldraw[blue] (\x, \y) circle (2pt);
					\node[circle, inner sep=0pt] (ptA\x) at (\x, \y) {};
				}
				\draw[dashed, thick, darkgreen] (0.5, 3) -- (10.5, 3) node[right] {$\widehat{\weight} = \med\,(\dataset)$};
				\node at (7.5, -4) {(a)};
			\end{scope}
			\begin{scope}[xshift=10cm]
				\foreach \x/\y in {
					1/2, 4/3, 7/10
				} {
					\draw[dashed, gray] (\x, 0) -- (\x, \y);
					\filldraw[blue] (\x, \y) circle (2pt);
					\node[circle, inner sep=0pt] (ptB\x) at (\x, \y) {};
				}
				\draw[dashed, thick, darkgreen] (0.5, 3) -- (10.5, 3) node[right] {$\widehat{\weight} = \med\,\big(\widetilde{\dataset}\big)$};
				\node[above right=2pt and 2pt, red] at (ptB7) {\gls{outlier}};
				\node at (7.5, -4) {(b)};
			\end{scope}
		\end{tikzpicture}
		\caption{For the simple \gls{parammodel} $\hypothesis^{(\weight)}(\featurevec) = \weight$, 
			\gls{erm} with \gls{abserr} amounts to computing the \gls{median}. 
			(a) Original \gls{dataset} $\dataset$. (b) Noisy 
			\gls{dataset} $\widetilde{\dataset}$ including an \gls{outlier}. \label{fig_lad_dict}}
		\end{figure}
		See also: \gls{erm}, \gls{abserr}, \gls{huberreg}.},
	first={least absolute deviation regression},
	text={least absolute deviation regression}
}

\newglossaryentry{bayesrisk}
{name={Bayes risk},
	description={Consider a \gls{probmodel} for an \gls{ml} application where each 
		\gls{datapoint} is interpreted as an \gls{rv}. The \gls{probmodel} includes 
		a \gls{probdist} for the \glspl{feature} $\featurevec$ and \gls{label} $\truelabel$ 
		of a \gls{datapoint}. The\index{Bayes risk} Bayes \gls{risk} is the \gls{minimum} 
		possible \gls{risk} that can be achieved by any \gls{hypothesis} $\hypothesis: \featurespace \rightarrow \labelspace$. 
		Any \gls{hypothesis} that achieves the Bayes \gls{risk} is referred to as a 
		\gls{bayesestimator} \cite{LC}.
		\\
		See also: \gls{probmodel}, \gls{risk}, \gls{bayesestimator}.},
	first={Bayes risk},
	text={Bayes risk}
}
	
\newglossaryentry{bayesestimator}
{name={Bayes estimator},
	description={Consider\index{Bayes estimator} a \gls{probmodel} with a joint \gls{probdist} 
		$p(\featurevec,\truelabel)$ over the \glspl{feature} $\featurevec$ and the \gls{label} $\truelabel$ 
		of a \gls{datapoint}. For a given \gls{lossfunc} $\lossfunc{\cdot}{\cdot}$, we refer to a \gls{hypothesis} 
		$\hypothesis$ as a Bayes \gls{estimator} if its \gls{risk} 
		$\expect\left\{\lossfunc{\pair{\featurevec}{\truelabel}}{\hypothesis}\right\}$
		is the \gls{minimum} achievable \gls{risk} \cite{LC}. 
		Note that whether a \gls{hypothesis} qualifies as a Bayes \gls{estimator} depends on 
		the underlying \gls{probdist} and the choice for the \gls{lossfunc} $\lossfunc{\cdot}{\cdot}$.
		\\
		See also: \gls{probmodel}, \gls{hypothesis}, \gls{estimator}, \gls{risk}.},
	first={Bayes estimator},
	text={Bayes estimator}
}

\newglossaryentry{weight}
{name={weight},
	description={Consider\index{weight} a parameterized \gls{hypospace} $\hypospace$. We use the term 
		weights for numeric \glspl{modelparam} that are used to scale \glspl{feature} or their transformations 
		in order to compute $\hypothesis^{(\weights)} \in \hypospace$. A \gls{linmodel} uses weights 
		$\weights=\big(\weight_{1}, \,\ldots, \,\weight_{\nrfeatures}\big)\,^{T}$ to compute the linear combination 
		$\hypothesis^{(\weights)}(\featurevec)= \weights\,^{T} \featurevec$. Weights are also used in \glspl{ann} 
		to form linear combinations of \glspl{feature} or the \glspl{output} of \glspl{neuron} in hidden \glspl{layer} 
		(see Fig.\ \ref{fig_weights_dict}).
		\begin{figure}[H]
			\begin{center}
			\begin{tikzpicture}[neuron/.style={circle, draw, minimum size=1cm}, 
                    	thick, >=stealth]
			\node[neuron] (h1) at (0, 2) {$h_1$};
			\node[neuron] (h2) at (0, 0) {$h_2$};
			\node[neuron] (h3) at (0, -2) {$h_3$};
			\coordinate[right=3cm of h2] (outpoint);
			\node[anchor=west] at ([xshift=0.2cm]outpoint) 
    			{$z = w_1 h_1 + w_2 h_2 + w_3 h_3$};
			\draw[->] (h1) -- node[above] {$w_1$} (outpoint);
			\draw[->] (h2) -- node[above] {$w_2$} (outpoint);
			\draw[->] (h3) -- node[below] {$w_3$} (outpoint);
			\end{tikzpicture}
			\end{center}
		\caption{Section of an \gls{ann} containing a hidden \gls{layer} with \glspl{output} (or \glspl{activation}) 
			$h_{1}$, $h_{2}$, and $h_{3}$. These \glspl{output} are combined linearly to compute $z$, which can 
			be used either as \gls{output} of the \gls{ann} or as input to another \gls{layer}.\label{fig_weights_dict}}
		\end{figure}
		See also: \gls{hypospace}, \gls{modelparam}, \gls{feature}, \gls{linmodel}, \gls{ann}, \gls{layer}, \gls{activation}.},
	firstplural={weights},
	plural={weights},
	first={weight},
	text={weight}
}

\newglossaryentry{parameter}
{name={parameter},
	description={The\index{parameter} parameter of an \gls{ml} \gls{model} is a tunable (i.e., learnable or adjustable) quantity that 
		allows us to choose between different \gls{hypothesis} \glspl{map}. For example, the \gls{linmodel} 
		$\hypospace \defeq \{\hypothesis^{(\weights)}: \hypothesis^{(\weights)}(\feature)= \weight_{1} \feature + \weight_{2}\}$ 
		consists of all \gls{hypothesis} \glspl{map} $\hypothesis^{(\weights)}(\feature)= \weight_{1} \feature + \weight_{2}$ 
		with a particular choice for the parameters $\weights = \big(\weight_{1},\weight_{2}\big)\,^{T} \in \mathbb{R}^{2}$. 
		Another example of a \gls{model} parameter is the \glspl{weight} assigned to a connection between two \glspl{neuron} of an \gls{ann}.
				\\
		See also: \gls{ml}, \gls{model}, \gls{hypothesis}, \gls{map}, \gls{linmodel}, \gls{weight}, \gls{ann}.},
	first={parameter},
	text={parameter},
	firstplural={parameters}, 
 	plural={parameters}
}
    
\newglossaryentry{stopcrit}
{name={stopping criterion},
	description={Many\index{stopping criterion} \gls{ml} methods use iterative \glspl{algorithm} 
		that construct a sequence of \glspl{modelparam} in order to minimize the \gls{trainerr}. 
		For example, \glspl{gdmethod} iteratively update the \glspl{parameter} of a \gls{parammodel}, 
		such as a \gls{linmodel} or a \gls{deepnet}. Given a finite amount of computational 
		resources, we need to stop updating the \glspl{parameter} after a finite number of \glspl{iteration}. 
		A stopping criterion is any well-defined condition for deciding when to stop updating.
				\\
		See also: \gls{algorithm}, \gls{gdmethod}.},
	first={stopping criterion},
	firstplural={stopping criteria},
	plural={stopping criteria}, 
	text={stopping criterion}
}

\newglossaryentry{jacobimethod}
{name={Jacobi method},
	description={The Jacobi method\index{Jacobi method} is an \gls{algorithm}  
		for solving systems of linear equations (i.e., a linear system) of the form $\mA\vx= \mathbf{b}$.  
		Here, $\mA \in \mathbb{R}^{\nrfeatures \times \nrfeatures}$ is a square \gls{matrix} with 
		nonzero main diagonal entries. The method constructs a \gls{sequence} $\vx^{(0)}, \,\vx^{(1)}, \,\ldots$ 
		by updating each entry of $\mathbf{x}^{(\iteridx)}$ as follows:
		\[
		x_i^{(\iteridx+1)} = \frac{1}{a_{ii}} \left( b_i - \sum_{j \neq i} a_{ij} x_j^{(\iteridx)} \right).
		\]
		Note that all entries $x^{(k)}_{1}, \,\ldots, \,x^{(k)}_{\nrfeatures}$ are updated simultaneously.
		The above \gls{iteration} converges to a solution, i.e., $\lim_{\iteridx \rightarrow \infty} \vx^{(\iteridx)}=\vx$, 
		under certain conditions on the \gls{matrix} $\mA$, e.g., being strictly 
		diagonally dominant or symmetric positive  definite \cite{GolubVanLoanBook}, \cite{StrangLinAlg2016}, \cite{Horn91}. 
		Jacobi-type methods are appealing for large linear systems due to their parallelizable structure \cite{ParallelDistrBook}.
		We can interpret the Jacobi method as a \gls{fixedpointiter}. Indeed, using the decomposition $\mA = \mD + \mR$, with $\mD$ being the 
		diagonal of $\mA$, allows us to rewrite the linear equation $\mA \vx = \vb$ as a \gls{fixedpointeq}: 
		\[
		\mathbf{x} = \underbrace{\mD^{-1}(\mathbf{b} - \mR \mathbf{x})}_{\fixedpointop \vx},
		\]
		which leads to the \gls{iteration} $\vx^{(\iteridx+1)} = \mD^{-1}(\mathbf{b} - \mR \vx^{(\iteridx)})$.
		\\
		As an example, for the linear equation $\mA \mathbf{x} = \mathbf{b}, \text{where}$
		 \[
		 \mA = \begin{pmatrix}
		 	a_{11} & a_{12} & a_{13} \\
		 	a_{21} & a_{22} & a_{23} \\
		 	a_{31} & a_{32} & a_{33}
		 \end{pmatrix}, \qquad
		 \mathbf{b} = \begin{pmatrix}
		 	b_1 \\
		 	b_2 \\
		 	b_3
		 \end{pmatrix},
		 \]
		 the Jacobi method updates each component of \( \mathbf{x} \) as follows:
		 \[
		 \begin{aligned}
		 	x_1^{(k+1)} &= \frac{1}{a_{11}} \left( b_1 - a_{12} x_2^{(k)} - a_{13} x_3^{(k)} \right), \\
		 	x_2^{(k+1)} &= \frac{1}{a_{22}} \left( b_2 - a_{21} x_1^{(k)} - a_{23} x_3^{(k)} \right), \\
		 	x_3^{(k+1)} &= \frac{1}{a_{33}} \left( b_3 - a_{31} x_1^{(k)} - a_{32} x_2^{(k)} \right).
		 \end{aligned}
		 \]
		See also: \gls{algorithm}, \gls{matrix}, \gls{fixedpointiter}, \gls{optmethod}.},
	text={Jacobi method}, 
	first={Jacobi method}
}

\newglossaryentry{parammodel}
{name={parametric model},
	description={A\index{parametric model} parametric \gls{model} is a mathematical \gls{model} characterized 
		by a finite set of variable quantities called \glspl{parameter}.
		An important example is the \gls{probmodel} consisting, for a given 
		\gls{dimension} $\nrfeatures$, of all \glspl{mvndist} (on \gls{samplespace} 
		$\mathbb{R}^{\nrfeatures}$) with some \gls{mean} \gls{vector} 
		$\meanvecgeneric \in \mathbb{R}^{\nrfeatures}$ and \gls{covmtx} $\covmtxgeneric \in \mathbb{R}^{\nrfeatures \times \nrfeatures}$.
        		In the context of \gls{ml}, a parametric \gls{model} defines a 
		\gls{hypospace} $\hypospace$ parameterized by a finite number of \glspl{modelparam}.
        		Each \gls{hypothesis} $\hypothesis \in \hypospace$ is uniquely identified by a 
		list of \glspl{modelparam} $\weight_{1}, \,\weight_{2}, \,\ldots, \,\weight_{\dimlocalmodel}$ 
		(see Fig.\ \ref{fig_param_space_dict}). We can stack these \glspl{parameter} into a 
		\gls{vector} $\weights \in \mathbb{R}^{\dimlocalmodel}$. Two widely used examples 
		of parametric \glspl{model} are the \gls{linmodel} and the \gls{ann}. The 
		corresponding \gls{paramspace} is typically a subset of $\mathbb{R}^{\dimlocalmodel}$.
		\begin{figure}[H]
			\begin{center}
			\begin{tikzpicture}
				\node[ellipse, minimum width=3cm, minimum height=2cm, draw, thick] (paramspace) {};
				\node[below=0.1cm of paramspace] {\gls{paramspace} $\paramspace$};
				\node[black, circle, inner sep=2pt, fill] (theta1) at ($(paramspace.north west) + (1, -1)$) {};
				\node[left=0.01cm of theta1] {$\weights$};
				\node[black, circle, inner sep=2pt, fill] (theta2) at ($(paramspace.south east) + (-1.5, 1)$) {};
				\node[left=0.01cm of theta2] {$\weights'$};
				\node[ellipse, minimum width=7cm, minimum height=3cm, draw, thick, right=4cm of paramspace] (plotcloud) {};
				\node[above=0.2cm of plotcloud] {\gls{model} $\hypospace$};
				\node (plot1start) at ($(plotcloud.south west) + (0.2, 0.2)$) {};
				\draw[thick, red] (plot1start) .. controls ++(0.8, 1) and ++(-0.8, -0.8) .. ($(plotcloud.south west) + (2.8, 0.8)$) node[anchor=west] {$\hypothesis^{(\weights)}$};
				\node (plot2start) at ($(plotcloud.south west) + (1.0, 1.2)$) {};
				\draw[thick, blue] (plot2start) .. controls ++(0.8, 0.5) and ++(-0.8, -0.8) .. ($(plotcloud.south west) + (2.8, 2.1)$) node[anchor=west] {$\hypothesis^{(\weights')}$};
				\draw[thick, ->, bend right=20] (theta1) to ($(plot1start) + (0,0)$);
				\draw[thick, ->, bend left=20] (theta2) to (plot2start);
			\end{tikzpicture}
			\end{center} 
		\caption{\Gls{paramspace} $\paramspace$ of an \gls{ml} \gls{model} $\hypospace$ consisting of all 
			feasible choices for the \glspl{modelparam}. Each choice $\weights$ for the \glspl{modelparam} 
			selects a \gls{hypothesis} \gls{map} $\hypothesis^{(\weights)} \in \hypospace$.
			\label{fig_param_model_dict}} 
		\end{figure}
		See also: \gls{model}, \gls{modelparam}, \gls{paramspace}.},
	first={parametric model},
	firstplural={parametric models},
	plural={parametric models},
	text={parametric model}
}

\newglossaryentry{paramspace}
{name={parameter space},
	description={The\index{parameter space} \gls{parameter} space $\paramspace$ of 
		an \gls{ml} \gls{model} $\hypospace$ is the set of all feasible choices for the 
		\glspl{modelparam} (see Fig.\ \ref{fig_param_space_dict}). Many important \gls{ml} methods 
		use a \gls{model} that is parameterized by \glspl{vector} of the \gls{euclidspace} $\mathbb{R}^{\dimlocalmodel}$. 
		Two widely used examples of parameterized \glspl{model} are \glspl{linmodel} 
		and \glspl{deepnet}. The \gls{parameter} space is then often a subset $\paramspace \subseteq \mathbb{R}^{\dimlocalmodel}$, 
		e.g., all \glspl{vector} $\weights \in \mathbb{R}^{\dimlocalmodel}$ with a \gls{norm} smaller than one.
		\begin{figure}[H]
			\begin{center}
			\begin{tikzpicture}
				\node[ellipse, minimum width=3cm, minimum height=2cm, draw, thick] (paramspace) {};
				\node[below=0.1cm of paramspace] {\gls{parameter} space $\paramspace$};
				\node[black, circle, inner sep=2pt, fill] (theta1) at ($(paramspace.north west) + (1, -1)$) {};
				\node[left=0.01cm of theta1] {$\weights$};
				\node[black, circle, inner sep=2pt, fill] (theta2) at ($(paramspace.south east) + (-1.5, 1)$) {};
				\node[left=0.01cm of theta2] {$\weights'$};
				\node[ellipse, minimum width=7cm, minimum height=3cm, draw, thick, right=4cm of paramspace] (plotcloud) {};
				\node[above=0.2cm of plotcloud] {\gls{model} $\hypospace$};
				\node (plot1start) at ($(plotcloud.south west) + (0.2, 0.2)$) {};
				\draw[thick, red] (plot1start) .. controls ++(0.8, 1) and ++(-0.8, -0.8) .. ($(plotcloud.south west) + (2.8, 0.8)$) node[anchor=west] {$\hypothesis^{(\weights)}$};
				\node (plot2start) at ($(plotcloud.south west) + (1.0, 1.2)$) {};
				\draw[thick, blue] (plot2start) .. controls ++(0.8, 0.5) and ++(-0.8, -0.8) .. ($(plotcloud.south west) + (2.8, 2.1)$) node[anchor=west] {$\hypothesis^{(\weights')}$};
				\draw[thick, ->, bend right=20] (theta1) to ($(plot1start) + (0,0)$);
				\draw[thick, ->, bend left=20] (theta2) to (plot2start);
			\end{tikzpicture}
			\end{center} 
		\caption{\Gls{parameter} space $\paramspace$ of an \gls{ml} \gls{model} $\hypospace$ consisting of all 
			feasible choices for the \glspl{modelparam}. Each choice $\weights$ for the \glspl{modelparam} 
			selects a \gls{hypothesis} \gls{map} $\hypothesis^{(\weights)} \in \hypospace$.
			\label{fig_param_space_dict}} 
		\end{figure}
		See also: \gls{parameter}, \gls{model}, \gls{modelparam}.},
	first={parameter space},
	text={parameter space}
}

\newglossaryentry{datanorm}
{name={data normalization},
	description={\Gls{data} normalization\index{data normalization} refers to transformations 
		applied to the \glspl{featurevec} of \glspl{datapoint} to improve the \gls{ml} method's 
		\glspl{statasp} or \glspl{compasp}. For example, in \gls{linreg} with \glspl{gdmethod} using 
		a fixed \gls{learnrate}, \gls{convergence} depends on controlling the \gls{norm} of \glspl{featurevec} 
		in the \gls{trainset}. A common approach is to normalize \glspl{featurevec} such that their 
		\gls{norm} does not exceed one \cite[Ch.\ 5]{MLBasics}.
				\\
		See also:  \gls{gdmethod}, \gls{learnrate}, \gls{featuremap}.},
	first={data normalization},
	text={data normalization}
}

\newglossaryentry{dataaug}
{name={data augmentation},
	description={\Gls{data} augmentation\index{data augmentation} methods add synthetic \glspl{datapoint} 
		to an existing set of \glspl{datapoint}. These synthetic \glspl{datapoint} are obtained by 
		perturbations (e.g., adding noise to physical measurements) or transformations 
		(e.g., rotations of images) of the original \glspl{datapoint}. These perturbations and 
		transformations are such that the resulting synthetic \glspl{datapoint} should 
		still have the same \gls{label}. As a case in point, a rotated cat image is still 
		a cat image even if their \glspl{featurevec} (obtained by \gls{stacking} pixel color intensities) 
		are very different (see Fig.\ \ref{fig_symmetry_dataaug_dict}). \Gls{data} augmentation can be an 
		efficient form of \gls{regularization}.
		\begin{figure}[H]
		\begin{center}
			\begin{tikzpicture}
				\newcommand{\xshift}{0.5}
				\newcommand{\yshift}{2}
  				\draw[very thick, blue] plot[smooth, tension=1] coordinates {(0,0) (2,1) (4,0) (6,-1) (8,0)};
  				\node[blue, right] at (0,0) {\textbf{cat}};
  				\draw[very thick, red, dashed] plot[smooth, tension=1] coordinates {(0 + \xshift,0 + \yshift) (2 + \xshift,1 + \yshift) (4 + \xshift,0 + \yshift) (6 + \xshift,-1 + \yshift) (8 + \xshift,0 + \yshift)};
  				\node[red, right] at (8 + \xshift,0 + \yshift) {\textbf{no cat}};
				\fill[blue] (2,1) circle (2pt) node[above] {$\featurevec^{(1)}$};
				\fill[blue] (6,-1) circle (2pt) node[above] {$\featurevec^{(2)}$};
				  \draw[->, thin, >=latex, line width=0.5pt] (2,1) to[out=240, in=240] node[midway, below] {$\mathcal{T}^{(\eta)}$} (6,-1);
			  \end{tikzpicture}
			  \vspace*{-11mm}
		\end{center}
		\caption{\Gls{data} augmentation exploits intrinsic symmetries of \glspl{datapoint} in 
			some \gls{featurespace} $\featurespace$. We can represent a symmetry by 
		    	an \gls{operator} $\mathcal{T}^{(\eta)}: \featurespace \rightarrow \featurespace$,
		    	parameterized by some number $\eta \in \mathbb{R}$. For example, $\mathcal{T}^{(\eta)}$ 
		    	might represent the effect of rotating a cat image by $\eta$ degrees. A \gls{datapoint} 
		    	with \gls{featurevec} $\featurevec^{(2)} = \mathcal{T}^{(\eta)} \big(\featurevec^{(1)} \big)$ must 
		    	have the same \gls{label} $\truelabel^{(2)}=\truelabel^{(1)}$ as a \gls{datapoint} 
		   	 with \gls{featurevec} $\featurevec^{(1)}$.\label{fig_symmetry_dataaug_dict}}
		 \end{figure}
		See also: \gls{data}, \gls{datapoint}, \gls{label}, \gls{featurevec}, \gls{stacking}, \gls{regularization}, \gls{featurespace}. },
	first={data augmentation},
	text={data augmentation}
}
	
\newglossaryentry{localdataset}
{name={local dataset},
	description={The\index{local dataset} concept of a local \gls{dataset} is 
		in between the concept of a \gls{datapoint} and a \gls{dataset}. A local \gls{dataset} consists of several 
		individual \glspl{datapoint} characterized by \glspl{feature} and \glspl{label}. 
		In contrast to a single \gls{dataset} used in basic \gls{ml} methods, a local \gls{dataset} is also 
		related to other local \glspl{dataset} via different notions of similarity. These similarities 
		might arise from \glspl{probmodel} or communication infrastructure and 
		are encoded in the edges of an \gls{flnetwork}.
				\\
		See also: \gls{dataset}, \gls{datapoint}, \gls{feature}, \gls{label}, \gls{ml}, \gls{probmodel}, \gls{flnetwork}.},
	first={local dataset},
	plural={local datasets}, 
	firstplural={local datasets}, 
	text={local dataset}
}
	
\newglossaryentry{localmodel}
{name={local model},
sort={localmodel},
	description={Consider\index{local model} a collection of \glspl{device} that are represented 
		as nodes $\nodes$ of an \gls{flnetwork}. A local \gls{model} $\localmodel{\nodeidx}$ 
		is a \gls{hypospace} assigned to a node $\nodeidx \in \nodes$. Different nodes can have 
		different \glspl{hypospace}, i.e., in general, $\localmodel{\nodeidx} \neq \localmodel{\nodeidx'}$ 
		for different nodes $\nodeidx, \nodeidx' \in \nodes$. 
				\\
		See also: \gls{device}, \gls{flnetwork}, \gls{model}, \gls{hypospace}. },
	first={local model},
	plural={local models}, 
	firstplural={local models}, 
	text={local model}
}
	
\newglossaryentry{mutualinformation}
{name={mutual information (MI)},
 	description={The\index{mutual information (MI)} MI $\mutualinformation{\featurevec}{\truelabel}$ 
 		between two \glspl{rv} $\featurevec$, $\truelabel$ defined on the same \gls{probspace} 
 		is given by \cite{coverthomas} $$\mutualinformation{\featurevec}{\truelabel} \defeq 
		\expect \left\{ \log \frac{p (\featurevec,\truelabel)}{p(\featurevec)p(\truelabel)} \right\}.$$ 
		It is a \gls{measure} of how well we can estimate $\truelabel$ based 
		solely on $\featurevec$. A large value of $\mutualinformation{\featurevec}{\truelabel}$ indicates that 
		$\truelabel$ can be well predicted solely from $\featurevec$. This \gls{prediction} could be obtained by a 
		\gls{hypothesis} learned by an \gls{erm}-based \gls{ml} method. 
				\\
		See also: \gls{rv}, \gls{probspace}, \gls{prediction}, \gls{hypothesis}, \gls{erm}, \gls{ml}. }, 
	first={MI}, 
	text={MI} 
}

\newglossaryentry{zerogradientcondition}
{name={zero-gradient condition},
	description={Consider\index{zero-gradient condition} the unconstrained 
		\gls{optproblem} $\min_{\weights \in \mathbb{R}^{\dimlocalmodel}} f(\weights)$ with 
		a \gls{smooth} and \gls{convex} \gls{objfunc} $f(\weights)$. 
		A necessary and sufficient condition for a \gls{vector} $\widehat{\weights} \in \mathbb{R}^{\dimlocalmodel}$ 
		to solve this problem is that the \gls{gradient} $\nabla f \big( \widehat{\weights} \big)$ 
		is the zero \gls{vector} $\nabla f \big( \widehat{\weights} \big) = \mathbf{0}$ (see Fig.\ \ref{fig_zerogradientcondition_dict}). 
		\begin{figure}[H] 
			\centering
			\begin{tikzpicture}[scale=1.1,x=2cm]
				\pgfmathsetmacro{\xmin}{1}
				\pgfmathsetmacro{\ymin}{0.5}
				\pgfmathsetmacro{\xA}{-0.2}
				\pgfmathsetmacro{\xB}{2.4}
				\pgfmathsetmacro{\yB}{(\xB-\xmin)^2 + \ymin}
				\draw[thick, domain=\xA:\xB, samples=40]
				plot (\x,{(\x-\xmin)^2 + \ymin});
				\node[anchor=west] at (\xB,\yB) {$f(\weights)$};
				\coordinate (xstar) at (\xmin,\ymin);
				\fill (xstar) circle (1.6pt);
				\draw[dashed] (\xmin,0) -- (xstar) node[below,yshift=-18pt] {$\widehat{\weights}$};
				\draw[dashed] (0,\ymin) -- (xstar);
				\draw[thick] (xstar) ++(-0.7,0) -- ++(1.4,0);
				\node [xshift=20pt,yshift=-5pt] at ($(xstar)+(1.0,0.18)$) {$\nabla f\big(\widehat{\weights}\big)\!=\!\mathbf{0}$};
			\end{tikzpicture}
		\caption{\Gls{vector} $\widehat{\weights}$ solves the \gls{optproblem} if the \gls{gradient} 
			satisfies $\nabla f \big( \widehat{\weights} \big) = \mathbf{0}$. \label{fig_zerogradientcondition_dict}}
		\end{figure}
		In other words \cite[p. 140]{BoydConvexBook}, 
		$$\nabla f \big( \widehat{\weights} \big) = \mathbf{0} \Leftrightarrow  f \big( \widehat{\weights} \big) = \min_{\weights \in \mathbb{R}^{\dimlocalmodel}} f(\weights).$$ 
		By defining the \gls{gradient} \gls{operator} $\nabla f: \mathbb{R}^{\nrfeatures} \rightarrow \mathbb{R}^{\nrfeatures}$, we 
		can rewrite the zero-gradient condition as a \gls{fixedpointeq}:
		$$ \big( \mathcal{I} - \regparam \nabla f\big) \widehat{\weights} = \widehat{\weights}.$$ 
		Here, $\mathcal{I}$ denotes the identity \gls{operator} (i.e., $\mathcal{I}(\weights) = \weights$) 
		and $\regparam$ is an arbitrary positive number. 
					\\
		See also: \gls{optproblem}, \gls{smooth}, \gls{convex}, \gls{objfunc}, \gls{vector}, \gls{gradient}.}, 
	first={zero-gradient condition},
	text={zero-gradient condition}
}

\newglossaryentry{edgeweight}
{name={edge weight},
	description={Each\index{edge weight} edge $\edge{\nodeidx}{\nodeidx'}$ of an \gls{flnetwork} is 
		assigned a nonnegative edge weight $\edgeweight_{\nodeidx,\nodeidx'}\geq0$. 
		A zero edge weight $\edgeweight_{\nodeidx,\nodeidx'}=0$ indicates the absence 
		of an edge between nodes $\nodeidx, \nodeidx' \in \nodes$.
				\\
		See also: \gls{flnetwork}.}, 
	first={edge weight},
	firstplural={edge weights},
	plural={edge weights},
	text={edge weight}
}

\newglossaryentry{dataminprinc}
{name={data minimization principle},
	description={European\index{data minimization principle} \gls{data} protection regulation 
		includes a \gls{data} minimization principle. This principle requires a \gls{data} controller to 
		limit the collection of personal information to what is directly relevant and necessary 
		to accomplish a specified purpose. The \gls{data} should be retained only for as long as 
		necessary to fulfill that purpose \cite[Article 5(1)(c)]{GDPR2016}, \cite{EURegulation2018}.
				\\
		See also: \gls{data}.}, 
	first={data minimization principle},
	text={data minimization principle}
}

\newglossaryentry{layer}
{name={layer},
	description={A \gls{deepnet}\index{layer} is an \gls{ann} that consists of consecutive 
	    	layers, indexed by $\ell=1,\,2,\,\ldots,\,L$. The $\ell$th layer consists of 
		\glspl{artificialneuron} $a^{(\ell)}_{1},\,\ldots,\,a^{(\ell)}_{\featuredim^{(\ell)}}$ with 
		the layer width $\featuredim^{(\ell)}$. Each of these \glspl{artificialneuron} evaluates 
		an \gls{actfun} for a weighted sum of the \glspl{output} (or \glspl{activation}) of the 
		previous layer $\ell-1$. The input to layer $\ell=1$ is formed from weighted sums 
		of the \glspl{feature} of the \gls{datapoint} for which the \gls{deepnet} computes 
		a \gls{prediction}. The \glspl{output} of the \glspl{neuron} in layer $\ell$ are then, in turn, 
		used to form the inputs for the \glspl{neuron} in the next layer. The final (\gls{output}) layer 
		consists of a single \gls{neuron} whose 
		\gls{output} is used as the \gls{prediction} delivered by the \gls{deepnet}.  
		\\
		See also: \gls{deepnet}, \gls{ann}.}, 
	first={layer},
	firstplural={layers},
	plural={layers}, 
	text={layer}
}

\newglossaryentry{activation}
{name={activation},
	description={The \gls{output}\index{activation} of an \gls{artificialneuron} within an \gls{ann} is referred 
		to as its activation. In particular, the activation is obtained by applying a (typically nonlinear) \gls{actfun} 
		to a weighted sum of its inputs.
		\\
		See also: \gls{ann}, \gls{deepnet}.}, 
	first={activation},
	firstplural={activations},
	plural={activations}, 
	text={activation}, 
}

\newglossaryentry{cav}
{name={concept activation vector (CAV)},
	description={Consider a \gls{deepnet}, consisting\index{concept activation vector (CAV)} of 
		several hidden \glspl{layer}, trained to predict the \gls{label} of a 
		\gls{datapoint} from its \gls{featurevec}. One way to explain the behavior 
		of the trained \gls{deepnet} is by using the \glspl{activation} of a 
		hidden \gls{layer} as a new \gls{featurevec} $\vz$. We then probe 
		the geometry of the resulting new \gls{featurespace} by applying the 
		\gls{deepnet} to \glspl{datapoint} that represent a specific concept 
		$\mathcal{C}$. By applying the \gls{deepnet} also to \glspl{datapoint} 
		that do not belong to this concept, we can train a binary \gls{linclass} 
		$g(\vz)$ that distinguishes between concept and non-concept \glspl{datapoint} 
		based on the \glspl{activation} of the hidden \gls{layer}. The resulting 
		\gls{decisionboundary} is a \gls{hyperplane} whose normal \gls{vector} is 
		the CAV \cite{kim2018interpretability} for the concept $\mathcal{C}$. 
				\\
		 See also: \gls{deepnet}, \gls{linmodel}, \gls{trustAI}, \gls{interpretability}, \gls{transparency}.}, 
	first={concept activation vector (CAV)},
	text={CAV}
}

\newglossaryentry{backpropagation}
{name={backpropagation},
 	description={Backpropagation\index{backpropagation} is an \gls{algorithm} for computing the \gls{gradient} 
	 	$\nabla_{\weights} f(\weights)$ of an \gls{objfunc} $f(\weights)$ that depends on the 
	  	\glspl{modelparam} $\weights$ of an \gls{ann}. One example of such an \gls{objfunc} 
	  	is the average \gls{loss} incurred by the \gls{ann} on a \gls{batch} of \glspl{datapoint}. 
	  	This \gls{algorithm} is a direct application of the chain rule from calculus to 
	  	efficiently compute \glspl{partialderivative} of the \gls{lossfunc} with respect to the \glspl{modelparam}.
		Backpropagation consists of two consecutive phases, also illustrated in Fig.\ \ref{fig:backprop_schematic_dict}. 
		The first phase includes the forward pass, where a \gls{batch} of 
		\glspl{datapoint} is fed into the \gls{ann}. The \gls{ann} processes the input through its \glspl{layer} using its current 
		\glspl{weight}, ultimately producing a \gls{prediction} at its \gls{output}. The \gls{prediction} of the \gls{batch} is compared to 
	   	the true \gls{label} using a \gls{lossfunc}, which quantifies the \gls{prediction} error. The second phase includes the 
		backward pass (i.e., backpropagation), where the error is backpropagated through the \gls{ann} \glspl{layer}. 
	   	The obtained \glspl{partialderivative} with respect to the \gls{ann} \glspl{parameter} $\weight_{1},\,\ldots,\,\weight_{\nrfeatures}$ 
	   	constitute the \gls{gradient} $\nabla f(\weights)$, which can be used, in turn, to implement a \gls{gradstep}.
	   	\begin{figure}[H]
		\centering
			\begin{tikzpicture}[
			>=Stealth,
			neuron/.style={circle,draw,minimum size=9mm,inner sep=0pt},
			conn/.style={-Stealth, line width=0.7pt},
			grad/.style={Stealth-, dashed, line width=0.7pt},
			lbl/.style={font=\small},
			box/.style={draw, rounded corners, inner sep=4pt}
			]
			\node[neuron] (x1) at (0,1.8) {$x_1$};
			\node[neuron] (x2) at (0,0)   {$x_2$};
			\node[neuron] (x3) at (0,-1.8){$x_3$};
			\node[neuron, right=2.8cm of x1] (h1) {$h_1$};
			\node[neuron, right=2.8cm of x2] (h2) {$h_2$};
			\node[neuron, right=2.8cm of x3] (h3) {$h_3$};
			\node[neuron, right=2.8cm of h2] (yhat) {$\predictedlabel=\hypothesis(\featurevec)$};
			\node[neuron, right=2.6cm of yhat, minimum size=10mm] (loss) {$\loss$};
			\foreach \i in {1,2,3}{
		  	\foreach \j in {1,2,3}{
			\draw[conn] (x\i) -- (h\j);
		  	}
			}
			\foreach \j in {1,2,3}{
		  	\draw[conn] (h\j) -- (yhat);
			}
			\draw[conn] (yhat) -- (loss);
			\draw[grad] (yhat) to[bend left=10] node[above, lbl] {$\nabla_{\hat{y}}\loss$} (loss);
			\foreach \j in {1,2,3}{
		  	\draw[grad] (h\j) to[bend left=10] (yhat);
			}
			\foreach \i in {1,2,3}{
		  	\foreach \j in {1,2,3}{
			\draw[grad] (x\i) to[bend left=10] (h\j);
		  	}
			}
			\node[lbl] at ($(x1)!0.5!(h1)+(0,1.0)$) {$W^{(1)},\,b^{(1)}$};
			\node[lbl] at ($(h1)!0.5!(yhat)+(0,1.0)$) {$W^{(2)},\,b^{(2)}$};
			\node[draw, rounded corners, inner sep=3pt, below=1.4cm of yhat, align=left] (legend) {%
			\begin{tikzpicture}[baseline={(0,0)}]
		  	\draw[conn] (0,0) -- (1.0,0); \node[right=2mm] at (1.5,0) {\small forward pass (compute $\hypothesis(\featurevec)$)};
		  	\draw[grad] (0,-0.5) -- (1.0,-0.5); \node[right=2mm] at (1.5,-0.5) {\small backward pass (compute $\nabla_{\weights}\loss$)};
			\end{tikzpicture}
			};
			\node[lbl] at ($(x2)+(0,-2.8)$) {$a^{(0)}{=}\mathbf{x}$,\; $a^{(1)}{=}\sigma(z^{(1)})$,\; $a^{(2)}{=}\hypothesis(\featurevec)$};
			\end{tikzpicture}
			\caption{Solid arrows show the forward pass (i.e., \gls{data} flow and \gls{loss} calculation), while dashed arrows show the 
			\gls{gradient} correction flow during the backward pass for updating the \glspl{parameter} {$W^{(x)},\,b^{(x)}$}.}
			\label{fig:backprop_schematic_dict}
		\end{figure}
 	  	See also: \gls{ann}, \gls{lossfunc}, \gls{gd}, \gls{optmethod}. },
 	first={backpropagation},
 	text={backpropagation}
}

\newglossaryentry{vcdim}
{name={Vapnik–Chervonenkis dimension (VC dimension)},
	description={The statistical properties of an \gls{erm}-based method depend critically on the expressive capacity 
		of its \gls{hypospace} (or \gls{model}) $\hypospace$. A standard \gls{measure} of this capacity is the 
		\index{Vapnik–Chervonenkis dimension (VC dimension)} VC \gls{dimension} $\vcdim{\hypospace}$ 
		\cite{vapnik2015uniform}. Formally, it is the largest integer $\samplesize$ such that there exists a \gls{dataset} 
		$\dataset = \{\featurevec^{(1)},\,\dots,\,\featurevec^{(\samplesize)}\} \subseteq \featurespace$ 
		that can be perfectly classified (or shattered) by some $\hypothesis \in \hypospace$. 
		Formally, this means that for every one of the $2^\samplesize$ possible assignments of binary \glspl{label} to  
		each \gls{featurevec} in $\dataset$, there exists some \gls{hypothesis} $\hypothesis \in \hypospace$ that realizes 
		this labeling. Intuitively, the VC \gls{dimension} quantifies how well $\hypospace$ can fit arbitrary \gls{label} assignments, 
		and thus captures its approximate power. It plays a central role in deriving bounds on the \gls{gengap}. 
		Fig.\ \ref{fig:vcdim_dict} illustrates the definition of the VC \gls{dimension} for a \gls{linmodel} $\linmodel{2}$ with 
		$\featurelen=2$ \glspl{feature}. Fig.\ \ref{fig:vcdim_dict}(a) and ~\ref{fig:vcdim_dict}(b) show the same set of three 
		noncollinear \glspl{featurevec} under two different binary labelings. In both cases, a separating \gls{hyperplane} 
		exists that realizes the labeling. Since this holds for all $2^3 = 8$ possible binary labelings 
		of the three \glspl{featurevec}, the set is shattered.
		Fig.\ \ref{fig:vcdim_dict}(c) depicts four \glspl{featurevec} with a specific labeling. 
		No linear separator can correctly classify all \glspl{datapoint} in this case. Thus, $\vcdim{\linmodel{2}}=3$. \\
		\begin{figure}[H]
		\centering
		\vspace*{3mm}
			\begin{subfigure}{.3\textwidth}
			\centering
			\begin{tikzpicture}[scale =.4]
				\filldraw [red]  (-1,1) circle (3pt);
				\filldraw [blue] (1,-1) circle (3pt);
				\filldraw [blue] (-1,-1) circle (3pt);
				\draw [black] (-1.5,0) -- (1.5,0);
				\node at (0,-3.4) {(a)};
			\end{tikzpicture}  
			\label{fig:VC_a}
			\end{subfigure}%
			\begin{subfigure}{.3\textwidth}
			\centering
			\begin{tikzpicture}[scale =.4]
				\filldraw [red]  (-1,1) circle (3pt);
				\filldraw [red]  (1,-1) circle (3pt);
				\filldraw [blue] (-1,-1) circle (3pt);
				\draw [black] (-1.2,0.3) -- (0.3,-1.2);
				\node at (0,-3.4) {(b)};
			\end{tikzpicture}  
			\label{fig:VC_b} 
			\end{subfigure}
			\begin{subfigure}{.3\textwidth}
			\centering
			\begin{tikzpicture}[ scale =.4]
				\filldraw [red]  (-1,1) circle (3pt);
				\filldraw [blue] (1,1) circle (3pt);
				\filldraw [red]  (1,-1) circle (3pt);
				\filldraw [blue] (-1,-1) circle (3pt);
				\node at (0,-3.4) {(c)};
			\end{tikzpicture}  
			\label{fig:VC_c}
			\end{subfigure}
			\caption{Illustration of the VC \gls{dimension} for a \gls{linmodel} $\linmodel{2}$ that is used 
			to learn a \gls{linclass} in the \gls{featurespace} $\mathbb{R}^2$. \label{fig:vcdim_dict} }
		\end{figure}
		More  generally, for a \gls{linmodel} $\linmodel{\nrfeatures}$, the VC \gls{dimension} equals $\nrfeatures+1$. 
		In other words, for \glspl{linmodel}, the VC \gls{dimension} essentially matches the \gls{dimension} of the 
		underlying \gls{paramspace} $\mathbb{R}^{\nrfeatures}$. 
		For more complex \glspl{hypospace}, such as \glspl{decisiontree} or \glspl{ann}, the relation 
		between VC \gls{dimension} and the \gls{dimension} of the \gls{featurespace} is far less direct. In 
		these cases, alternative complexity \glspl{measure}, such as the \gls{rademachercomplexity}, 
		can be more useful for analyzing \gls{erm}-based methods.
		\\
		See also: \gls{hypospace}, \gls{rademachercomplexity}, \gls{generalization}, \gls{ml}, \gls{effdim}. },
	first={Vapnik–Chervonenkis dimension (VC dimension)},
	text={VC dimension}
}

\newglossaryentry{rademachercomplexity}
{name={Rademacher complexity},
 	description={Similar to the \gls{vcdim}, the\index{Rademacher complexity} Rademacher 
		complexity \cite{bartlett2002rademacher} is a quantitative \gls{measure} 
		of the size of a \gls{hypospace} $\hypospace$. It is based on the 
		empirical Rademacher complexity, which is defined for a given \gls{dataset} 
		$\dataset$ as
		\begin{equation}
			\nonumber
    			\mathcal{R}_\dataset(\hypospace) 
    			= 
    			\expect_{\varepsilon_{1},\,\ldots,\,\varepsilon_{\samplesize} } 
			\sup_{\hypothesis \in \hypospace} \frac{1}{\samplesize} \sum_{\sampleidx=1}^\samplesize \varepsilon_\sampleidx 
			\hypothesis\big(\featurevec^{(\sampleidx)} \big).
    		\end{equation}
		Here, the \gls{expectation} is taken with respect to 
		the \glspl{rv} $\varepsilon_{1},\,\ldots,\,\varepsilon_{\samplesize}$,
		which are \gls{iid} and take values in $\{ -1, +1 \}$ with equal 
		\gls{probability} $1/2$. The Rademacher complexity of $\hypospace$ is 
		then defined as the \gls{expectation} of the empirical Rademacher complexity 
		of a random \gls{dataset} $\dataset = \{\featurevec^{(1)}, \,\hdots, \,\featurevec^{(\samplesize)}\}$
		that consists of $\samplesize$ \gls{iid} \glspl{rv} $\featurevec^{(\sampleidx)} \in \featurespace$ 
		for $\sampleidx=1,\,\ldots,\,\samplesize$.  
		\\
        		See also: \gls{vcdim}, \gls{hypospace}, \gls{generalization}, \gls{ml}, \gls{effdim}. },
	first={Rademacher complexity},
	text={Rademacher complexity}  
}

\newglossaryentry{penaltyterm}
{name={penalty term},
	description={Consider\index{penalty term} an \gls{erm}-based \gls{ml} method that learns 
  		\glspl{modelparam} $\weights$ by minimizing the average \gls{loss} (or \gls{emprisk}) 
   		$\emprisk{\weights}{\dataset}$ on a \gls{trainset} $\dataset$. 
   		To avoid \gls{overfitting} and control the \gls{gengap}, 
 	 	it is common to augment the \gls{objfunc} $\emprisk{\weights}{\dataset}$ 
     		with a penalty term $\regparam \regularizer{\weights}$. We refer to the 
	 	resulting modified \gls{erm} as \gls{rerm}. 
	 	\begin{figure}[H]
			\begin{center} 
				\begin{tikzpicture}[scale = 1]
					\draw[->, very thick] (0,0.5) -- (7.7,0.5) node[right] {\gls{feature} $\feature$};       
					\draw[->, very thick] (0.5,0) -- (0.5,4.2) node[above] {\gls{label} $\truelabel$};   
					\draw[color=black, thick, dashed, domain = -1: 6.2, variable = \x] 
					 plot ({\x},{\x*0.4 + 2.0}) ;     
					\node at (6.7,4.5) {$\hypothesis^{(\weights)}(\feature)$};    
					\coordinate (l1)   at (1.2, 2.48);
					\coordinate (l2) at (1.4, 2.56);
					\coordinate (l3)   at (1.7,  2.68);
					\coordinate (l4)   at (2.2, 2.2*0.4+2.0);
					\coordinate (l5) at (2.4, 2.4*0.4+2.0);
					\coordinate (l6)   at (2.7,  2.7*0.4+2.0);
					\coordinate (l7)   at (3.9,  3.9*0.4+2.0);
					\coordinate (l8) at (4.2, 4.2*0.4+2.0);
					\coordinate (l9)   at (4.5,  4.5*0.4+2.0);
					\coordinate (n1)   at (1.2, 1.8);
					\coordinate (n2) at (1.4, 1.8);
					\coordinate (n3)   at (1.7,  1.8);
					\coordinate (n4)   at (2.2, 3.8);
					\coordinate (n5) at (2.4, 3.8);
					\coordinate (n6)   at (2.7,  3.8);
					\coordinate (n7)   at (3.9, 2.6);
					\coordinate (n8) at (4.2, 2.6);
					\coordinate (n9)   at (4.5,  2.6);
					\node at (n1)  [circle,draw,fill=red,minimum size=6pt,scale=0.6, name=c1] {};
					\node at (n2)  [circle,draw,fill=blue,minimum size=6pt, scale=0.6, name=c2] {};
					\node at (n3)  [circle,draw,fill=red,minimum size=6pt,scale=0.6,  name=c3] {};
					\node at (n4)  [circle,draw,fill=red,minimum size=12pt, scale=0.6, name=c4] {};  
					\node at (n5)  [circle,draw,fill=blue,minimum size=12pt,scale=0.6,  name=c5] {};
					\node at (n6)  [circle,draw,fill=red,minimum size=12pt, scale=0.6, name=c6] {};  
					\node at (n7)  [circle,draw,fill=red,minimum size=12pt,scale=0.6,  name=c7] {};
					\node at (n8)  [circle,draw,fill=blue,minimum size=12pt, scale=0.6, name=c8] {};
					\node at (n9)  [circle,draw,fill=red,minimum size=12pt, scale=0.6, name=c9] {};
					\draw [<->] ($ (n7) + (0,-0.3) $)  --  ($ (n9) + (0,-0.3) $) node [pos=0.4, below] {$\sqrt{\regparam}$}; ; 
					\draw[<->, color=red, thick] (l1) -- (c1);  
					\draw[<->, color=blue, thick] (l2) -- (c2);  
					\draw[<->, color=red, thick] (l3) -- (c3);  
					\draw[<->, color=red, thick] (l4) -- (c4);  
					\draw[<->, color=blue, thick] (l5) -- (c5);  
					\draw[<->, color=red, thick] (l6) -- (c6);  
					\draw[<->, color=red, thick] (l7) -- (c7);  
					\draw[<->, color=blue, thick] (l8) -- (c8);  
					\draw[<->, color=red, thick] (l9) -- (c9);  
					\draw[fill=blue] (6.2, 3.7)  circle (0.1cm) node [black,xshift=2.3cm] {original \gls{trainset} $\dataset$};
					\draw[fill=red] (6.2, 3.2)  circle (0.1cm) node [black,xshift=1.3cm] {augmented};
					\node at (4.6,1.2)  [minimum size=12pt, font=\fontsize{12}{0}\selectfont, text=blue] {$\frac{1}{\samplesize} \sum\limits_{\sampleidx=1}^\samplesize \lossfunc{\pair{\featurevec^{(\sampleidx)}}{ \truelabel^{(\sampleidx)}}}{\hypothesis^{(\weights)}}$};
					\node at (7.8,1.2)  [minimum size=12pt, font=\fontsize{12}{0}\selectfont, text=red] {$+\regparam \regularizer{\weights}$};
				\end{tikzpicture}
		\caption{Adding a penalty term $\regparam \regularizer{\weights}$ 
			to the \gls{objfunc} in \gls{erm} is equivalent to including 
			perturbations of the \gls{trainset} $\dataset$ during \gls{training}. 
			The \gls{regularization} \gls{parameter} $\regparam$ controls 
			the extent of the perturbations. \label{fig_penal_term_dict} }
			\end{center}
		\end{figure} 
  		The penalty term depends only on the \glspl{modelparam} but not on the \glspl{datapoint} in 
  		the \gls{trainset}. For some combinations of \gls{model} and \gls{lossfunc}, the penalty term 
  		can be obtained as the average \gls{loss} incurred on (an infinite number of) perturbed copies 
  		of the \gls{trainset}. In other words, adding a penalty term in \gls{erm} can 
  		be viewed as a form of \gls{dataaug}. 
		\\ 
  		See also: \gls{overfitting}, \gls{rerm}, \gls{regularization}, \gls{dataaug}. },
	first={penalty term},
  	text={penalty term},
  	firstplural={penalty terms},
  	plural={penalty terms}
}

\newglossaryentry{perceptron}
{name={perceptron}, 
	description={The perceptron\index{perceptron} is one of the oldest \gls{ml} \glspl{algorithm}; 
		it was developed by Frank Rosenblatt in 1957 \cite{rosenblatt1958perceptron}. 
		The perceptron \gls{algorithm} works on \gls{data} with numeric \glspl{feature} 
		$\featurevec_\sampleidx \in \mathbb{R}^{\nrfeatures}$ 
		and binary \glspl{label} $\truelabel_\sampleidx \in \{-1, 1\}$ and returns a \gls{linclass}. 
		It is guaranteed to find such a \gls{linclass} if the classes are linearly separable. 
		As the \gls{algorithm} tries to find a \gls{hyperplane} $g(\featurevec) \defeq \weights^\mathsf{T} \featurevec$ 
		that separates the classes, i.e., 
		$\text{sgn}\,(\weights^\mathsf{T} \featurevec_\sampleidx) = \truelabel_\sampleidx \; \forall \sampleidx \in \{1, \,\ldots,\, \samplesize \}$, 
		at each \gls{iteration} it selects a \gls{sample} $\sampleidx'$ that is misclassified, i.e., 
		$\text{sgn}\,(\weights^\mathsf{T} \featurevec_{\sampleidx}') \neq \truelabel_{\sampleidx'}$. 
		Then, the \glspl{weight} $\weights$ are updated by 
		$\weights \leftarrow \weights + \truelabel_{\sampleidx'} \featurevec_{\sampleidx'}$. 
		While this does not guarantee that the \gls{sample} is now correctly classified, the error margin has decreased. 
		\\
		See also: \gls{probmodel}, \gls{maxlikelihood}, \gls{optproblem}. },
	first={perceptron},
	text={perceptron}
}

\newglossaryentry{binclass}
{name={binary classification}, 
	description={Binary \gls{classification}\index{binary classification} refers to \gls{classification} with two \glspl{label}. 
		The \glspl{label} are usually defined as $\{-1, 1\}$ or $\{0, 1\}$. 
		\\ 
		See also: \gls{classification}, \gls{label}, \gls{datapoint}, \gls{feature}. },
	first={binary classification},
	text={binary classification}
}

\newglossaryentry{softmax}
{name={softmax function},
	description={The softmax\index{softmax function} \gls{function} \cite{bridle1989training}, \cite{bridle1990probabilistic}
		$S: \mathbb{R}^\nrcluster \rightarrow \simplex{\nrcluster-1}$ maps an 
		arbitrary \gls{inputvec} $\weights \in \mathbb{R}^\nrcluster$ (often referred to as logits) 
		onto the \gls{probabilitysimplex}. The $i$th component of the output is defined using 
		the exponential \gls{function} as
		\begin{equation}\label{eq:softmax_function_dict}
			\nonumber
			S(\weights)_i 
            =   \frac{\exp\,(\weight_i)}{\sum\limits_{j=1}^{\nrcluster} \exp\,(\weight_j)} \quad \text{for } i=1,\,\ldots,\,\nrcluster.
		\end{equation}
		Since the resulting components are nonnegative and sum to one, the output of the 
		softmax \gls{function} can be interpreted as a \gls{pmf} over $\nrcluster$ distinct \glspl{outcome}.
		The softmax \gls{function} is a standard tool in \gls{ml} for modeling discrete \glspl{probdist}. 
		For instance, in an \gls{ann}-based \gls{classifier}, it is typically applied 
		to the \glspl{output} of the final \gls{layer} to obtain the predicted \gls{probdist} over the possible class \gls{label}. 
		Similarly, in \gls{reinforcementlearning} with a discrete 
		\gls{actionspace}, a \gls{policy} can be defined using a softmax \gls{function} to 
		describe the \gls{probdist} for the next \gls{action} to take.
		\\
		See also: \gls{probabilitysimplex}, \gls{pmf}, \gls{ann}, \gls{classifier}, \gls{layer}, \gls{policy}.},
	first={softmax function},
	text={softmax function}
}

\newglossaryentry{cossim}
{name={cosine similarity},
	description={The cosine\index{cosine similarity} similarity quantifies the similarity of any two \glspl{featurevec} in the 
		\gls{euclidspace} as the cosine of the angle $\theta$ between them, i.e., irrespectively of their magnitude measured by their \gls{euclidnorm}. 
		Equivalently, for any two \glspl{featurevec} $\featurevec^{(1)}, \featurevec^{(2)} \in \mathbb{R}^\featuredim$, 
		the cosine similarity can also be defined as the \gls{innerproduct} of the \glspl{vector}, normalized by the product of their \glspl{norm}, i.e.,
		$$
		       \cos(\theta)  
		=     \frac{\innerprod{\featurevec^{(1)}}{\featurevec^{(2)}}}{\normgeneric{\featurevec^{(1)}}{2}\normgeneric{\featurevec^{(1)}}{2} } 
                  \; \in [-1,1].
		$$
		A value of $1$ indicates that the \glspl{featurevec} point in the exact same direction, $0$ indicates orthogonality (i.e., no similarity), and 
		$-1$ indicates they point in opposite directions, as illustrated in Fig.\ \ref{fig_cossim_dict}.
		\begin{figure}[H]
			\centering
			\begin{tikzpicture}[scale=0.8]
			\begin{scope}[xshift=0cm]
				\draw[->, thick, blue] (0,0) -- (2,1) node[above left] {$\featurevec^{(1)}$};
				\draw[->, thick, red] (0,0) -- (1.5,0.75) node[below right] {$\featurevec^{(2)}$};
				\node at (1, -1.5) {$\cos(\theta) = 1$};
				\node at (1, -2.5) {(a)};
			\end{scope}
			\begin{scope}[xshift=5.5cm]
				\draw[->, thick, blue] (0,0) -- (2,0) node[right] {$\featurevec^{(1)}$};
				\draw[->, thick, red] (0,0) -- (0,1.5) node[above] {$\featurevec^{(2)}$};
				\draw (0.3,0) -- (0.3,0.3) -- (0,0.3);
				\node at (1, -1.5) {$\cos(\theta) = 0$};
				\node at (1, -2.5) {(b)};
			\end{scope}
			\begin{scope}[xshift=11.5cm]
				\draw[->, thick, blue] (0,0) -- (1.5,0) node[right] {$\featurevec^{(1)}$};
				\draw[->, thick, red] (0,0) -- (-1.5,0) node[left] {$\featurevec^{(2)}$};
				\node at (0, -1.5) {$\cos(\theta) = -1$};
				\node at (0, -2.5) {(c)};
			\end{scope}
			\end{tikzpicture}
		\caption{Three cases of cosine similarity between two \glspl{featurevec}
                         $\featurevec^{(1)}, \featurevec^{(2)} \in \mathbb{R}^2$. \label{fig_cossim_dict}}
		\end{figure}
		This similarity measure is particularly useful in \gls{ml} applications where the directional orientation of the \glspl{featurevec} 
		is more informative than their \gls{norm} or their \gls{eucliddist}. 
		\\
		See also: \gls{datapoint}, \gls{featurespace}, \gls{feature}, \gls{metric}, \gls{nn}.},
	first={cosine similarity},
	text={cosine similarity}
}   
\newglossaryentry{psd}
{name={positive semi-definite (psd)},
	description={A\index{positive semi-definite (psd)} (real-valued) 
		\gls{symmetricmatrix} $\mQ = \mQ\,^{T} \in \mathbb{R}^{\featuredim \times \featuredim}$ 
	 	is referred to as psd if $\featurevec\,^{T} \mQ \featurevec \geq 0$ 
		for every \gls{vector} $\featurevec \in \mathbb{R}^{\featuredim}$. 
		A psd \gls{matrix} $\mQ$ admits a \gls{spectraldecomp} with nonnegative 
		\glspl{eigenvalue} $\eigval{1}, \,\ldots, \,\eigval{\featuredim} \geq 0$
		\cite{HornMatAnalysis}, \cite{Axler2025}. 
	 	The notion of being psd can be extended from \glspl{matrix} to (real-valued) 
	 	symmetric \gls{kernel} \glspl{map} $\kernel: \featurespace \times \featurespace \rightarrow \mathbb{R}$ 
	 	(with $\kernel(\featurevec,\featurevec') = \kernel(\featurevec',\featurevec)$)
	 	as follows: For any finite set of \glspl{featurevec} $\featurevec^{(1)}, \,\ldots, \,\featurevec^{(\samplesize)}$, 
	 	the resulting \gls{matrix} $\mQ \in \mathbb{R}^{\samplesize \times \samplesize}$ with 
		entries $Q_{\sampleidx,\sampleidx'} = \kernelmap{\featurevec^{(\sampleidx)}}{\featurevec^{(\sampleidx')}}$ 
		is psd \cite{LearningKernelsBook}.
		\\
		The psd property induces a partial order on \glspl{symmetricmatrix}:
		We write $\mA \succeq \mB$ (for symmetric $\mA, \mB \in \mathbb{R}^{\featuredim \times \featuredim}$)
		if the difference $\mA - \mB$ is psd \cite[Sec. 2.4]{BoydConvexBook}. In particular, $\mA \succeq \mathbf{0}$ means
		that $\mA$ is psd.
			\\
		See also: \gls{symmetricmatrix}, \gls{kernel}, \gls{pd}.},
	first={positive semi-definite (psd)},
	type=math, 
	text={psd}  
}

\newglossaryentry{pd}
{name={positive definite (pd)},
	description={A\index{positive definite (pd)} (real-valued)
		\gls{symmetricmatrix} $\mQ = \mQ\,^{T} \in \mathbb{R}^{\featuredim \times \featuredim}$
		is referred to as pd if $\featurevec\,^{T} \mQ \featurevec > 0$
		for every nonzero \gls{vector} $\featurevec \in \mathbb{R}^{\featuredim}$.
		Every pd \gls{matrix} is \gls{psd}, but not vice versa.
		A pd \gls{matrix} $\mQ$ admits a \gls{spectraldecomp} with strictly positive
		\glspl{eigenvalue} $\eigval{1}, \,\ldots, \,\eigval{\featuredim} > 0$
		\cite{HornMatAnalysis}. In particular, a pd \gls{matrix} is \gls{invertible}.
		\\
		Analogously to the \gls{psd} ordering, we write $\mA \succ \mB$
		if $\mA - \mB$ is pd \cite[Sec. 2.4]{BoydConvexBook}.
		\\
		See also: \gls{symmetricmatrix}, \gls{psd}, \gls{invertible}.},
	first={positive definite (pd)},
	type=math,
	text={pd}
}

\newglossaryentry{normalmatrix}
{name={normal matrix},
	description={A normal \gls{matrix} is a square \gls{matrix} $\mA \in \mathbb{C}^{\nrfeatures \times \nrfeatures}$ 
		that commutes with its \gls{conjugatetranspose}, i.e., $\mA\mA^{H}=\mA^{H}\mA$. 
		Normal \glspl{matrix}\index{normal matrix} admit an orthonormal \gls{basis} of \glspl{eigenvector} and are 
		unitarily \gls{diagonalizable}.
                 \\
                 See also: \gls{matrix}, \gls{diagonalizable}. },
	first={normal matrix},
    	type=math, 
	plural={normal matrices},
	firstplural={normal matrices},
	text={normal matrix}
}

\newglossaryentry{spectraldecomp}
{name={spectral decomposition},
	description={An \gls{evd} of a \gls{normalmatrix} is called a spectral decomposition\index{spectral decomposition}. 
		It has the special property that the \glspl{eigenvector} of the \gls{normalmatrix} are orthonormal. 
		This also works the other way, i.e., any \gls{matrix} that has orthonormal \glspl{eigenvector} is a \gls{normalmatrix}.
		\\
		See also: \gls{evd}, \gls{normalmatrix}, \gls{eigenvector}. },
	first={spectral decomposition},
    	type=math,
	text={spectral decomposition}
}

\newglossaryentry{symmetricmatrix}
{name={symmetric matrix},
	description={A symmetric \gls{matrix} is a square\index{symmetric matrix} \gls{matrix} $\mA$ 
		with real-valued entries that is equal to its \gls{transpose}, i.e., $\mA=\mA^{T}$. Every 
		symmetric \gls{matrix} is a \gls{normalmatrix}.
                 \\
                 See also: \gls{transpose}, \gls{normalmatrix}. },
	first={symmetric matrix},
	plural={symmetric matrices},
    	type=math,
	firstplural={symmetric matrices},
	text={symmetric matrix}
}

\newglossaryentry{transpose}
{name={transpose},
	description={The transpose\index{transpose} of a real-valued \gls{matrix} is obtained by exchanging 
		rows and columns. For a \gls{matrix} $\mA \in \mathbb{R}^{\samplesize \times \nrfeatures}$, 
		its transpose is denoted by $\mA^{T}$ and satisfies 
		$\big(\mA^{T}\big)_{\featureidx,\featureidx'}=\mA_{\featureidx',\featureidx}$.
		\\
		See also: \gls{matrix}, \gls{symmetricmatrix}. },
 	first={transpose},
    	type=math, 
 	text={transpose}
 }

\newglossaryentry{conjugatetranspose}
{name={conjugate transpose},
	description={The conjugate \gls{transpose}\index{conjugate transpose} of a 
		\gls{matrix} is obtained by transposing the \gls{matrix} 
		and taking the complex conjugate of each entry.
		For a \gls{matrix} $\mA \in \mathbb{C}^{\samplesize \times \nrfeatures}$, its
		conjugate \gls{transpose} is denoted by $\mA^{H} \in 
		\mathbb{C}^{\nrfeatures \times \samplesize}$ and is defined entrywise by
		\[
                 (\mA^{H})_{\featureidx,\sampleidx}
                 = \complexconjugate{\big(\mA\big)_{\sampleidx,\featureidx}},
		\]
		where $\complexconjugate{(\cdot)}$ denotes complex conjugation.
		\\
		See also: \gls{transpose}, \gls{matrix}, \gls{hermitian}. },
 	first={conjugate transpose},
    	type=math, 
 	text={conjugate transpose}
 }

\newglossaryentry{hermitian}
 {name={Hermitian},
 	description={A square\index{Hermitian} \gls{matrix} $\mA \in \mathbb{C}^{\nrfeatures \times \nrfeatures}$ is 
		Hermitian if it coincides with its \gls{conjugatetranspose}, i.e., $\mA=\mA^{H}$. 
                 Trivially, a Hermitian \gls{matrix} is also a \gls{normalmatrix}.
                 \\
                 See also: \gls{normalmatrix}. },
 	first={Hermitian},
     	type=math,
 	text={Hermitian}
}

\newglossaryentry{dimension}
{name={dimension},
	description={The\index{dimension} dimension $\dim \mathcal{A}$ of a 
		\gls{vectorspace} $\mathcal{A}$ is the cardinality of any 
		\gls{basis} of $\mathcal{A}$ \cite{StrangLinAlg2016}. 
		Strictly speaking, this definition applies only to finite-dimensional \glspl{vectorspace}, 
		i.e., those that possess a finite \gls{basis}. 
		\begin{figure}[H]
			\begin{tikzpicture}[scale=1]
  			\coordinate (O) at (0,0);
  			\draw[->, thick] (O) -- (1.8,0) node[below right] {$\ve^{(1)}$};
  			\draw[->, thick] (O) -- (0,1.6) node[above left] {$\ve^{(2)}$};
			\draw[->, thick, dashed, shift={(3.5,0.5)}] (0,0) -- (1.2,1.2) node[above right] {$\vu^{(1)}$};
			\draw[->, thick, dashed, shift={(3.5,0.5)}] (0,0) -- (-1.2,1.2) node[above left] {$\vu^{(2)}$};
  			\draw[->, thick, dotted, shift={(-2.5,-2.5)}] (O) -- (2.0,0.6) node[above right] {$\vw^{(1)}$};
  			\draw[->, thick, dotted, shift={(-2.5,-2.5)}] (O) -- (0.4,1.8) node[left] {$\vw^{(2)}$};
			\end{tikzpicture}
		\caption{Three \glspl{basis}, $\big\{\ve^{(1)},\ve^{(2)} \big\}, \big\{\vu^{(1)},\vu^{(2)} \big\},
			\big\{\vw^{(1)},\vw^{(2)} \big\}$, for the \gls{vectorspace} $\mathbb{R}^{2}$.} 
		\end{figure}
		For such spaces, all \glspl{basis} have the same cardinality, which is the dimension of the space 
		\cite[Ch.~2]{Axler2025}.	
   			 \\
		See also: \gls{vectorspace}, \gls{basis}. }, 
	text={dimension}, 
	type=math,
	first={dimension}  
}

\newglossaryentry{linearlyindep}
{name={linearly independent},
	description={A subset $\{\va^{(1)}, \,\ldots, \,\va^{(\nrfeatures)}\} \in \mathcal{V}$ 
		of a \gls{vectorspace} is linearly independent\index{linearly independent} 
		if there is no nontrivial linear combination of these \glspl{vector} that 
		equals the zero \gls{vector} \cite{StrangLinAlg2016}. 
		In other words, $$\sum_{\featureidx=1}^{\nrfeatures} \alpha_{\featureidx} \va^{(\featureidx)} = \mathbf{0}	
		\quad \text{implies} \quad \alpha_{1} = \alpha_{2} = \ldots = \alpha_{\nrfeatures} = 0.$$ 
			\\ 
		See also: \gls{vectorspace}, \gls{vector}, \gls{dimension}, \gls{basis}, \gls{linearlydep}.}, 
	text={linearly independent}, 
	type=math,
	first={linearly independent}  
}

\newglossaryentry{basis}
{name={basis},
	description={A basis\index{basis} of a finite-dimensional \gls{vectorspace} $\mathcal{V}$ 
		is a set of \gls{linearlyindep} \glspl{vector} $\mathcal{B} = \{\vb^{(1)}, \,\ldots, \,\vb^{(\nrfeatures)}\} 
                 \subseteq \mathcal{V}$ such that any \gls{vector} $\vv \in \mathcal{V}$ 
		can be expressed as a linear combination of the basis \glspl{vector} $\vb^{(1)}, \,\ldots, \,\vb^{(\nrfeatures)}$, i.e.,	
		$$ \vv = \sum_{\featureidx=1}^{\nrfeatures} \alpha_{\featureidx} \vb^{(\featureidx)}
			\quad \text{for some } \alpha_{1}, \,\ldots, \,\alpha_{\nrfeatures}. $$
                 The  scalars $\alpha_{1}, \,\ldots, \,\alpha_{\nrfeatures} $ can be regarded 
		as the coordinates of $\vv$ with respect to the basis $\mathcal{B}$. 
		Any basis of $\mathcal{V}$ has the same number $\nrfeatures$ of elements, 
		which is the \gls{dimension} of $\mathcal{V}$.
			\\
		See also: \gls{vectorspace}, \gls{linearlyindep}, \gls{vector}. },
	text={basis}, 
	firstplural={bases}, 
	plural={bases}, 
	type=math,
	first={basis} 
}

\newglossaryentry{widematrix}
{name={wide matrix},
	description={A\index{wide matrix} \gls{matrix} 
   		$\featuremtx \in \mathbb{R}^{\samplesize \times \nrfeatures}$ 
		is referred to as wide if it has more columns than rows, 
		i.e., when $\nrfeatures > \samplesize$ (see Fig.\ \ref{fig_widematrix_dict}).
		\begin{figure}[H]
			\centering
			\begin{tikzpicture}
				\def\matHeight{2}
				\def\matWidth{5} 
				\draw[thick, fill=gray!20] (0,0) rectangle (\matWidth, \matHeight);
				\node at (0.5*\matWidth, 0.5*\matHeight) {\Large $\featuremtx \in \mathbb{R}^{\samplesize \times \nrfeatures}$};
				\draw[decorate, decoration={brace, amplitude=5pt}, thick] 
  				(-0.2, 0) -- (-0.2, \matHeight)
  				node[midway, left=8pt] {$\samplesize$};
				\draw[decorate, decoration={brace, amplitude=5pt, mirror}, thick] 
        				(0, -0.2) -- (\matWidth, -0.2) 
        				node[midway, below=8pt] {$\nrfeatures \quad (\nrfeatures > \samplesize)$};
			\end{tikzpicture}
		\caption{Illustration of a wide \gls{matrix}. \label{fig_widematrix_dict}}
		\end{figure}
		See also: \gls{matrix}, \gls{tallmatrix}, \gls{johnsonlindenstrausslemma}, \gls{randomprojection}. },
	text={wide matrix}, 
	firstplural={wide matrices}, 
	plural={wide matrices}, 
	type=math,
   	first={wide matrix} 
}

\newglossaryentry{tallmatrix}
{name={tall matrix},
	description={A\index{tall matrix} \gls{matrix} $\featuremtx \in \mathbb{R}^{\samplesize \times \nrfeatures}$ 
		is referred to as tall if it has more rows than columns, i.e., 
		when $\samplesize > \nrfeatures$ (see Fig.\ \ref{fig_tallmatrix_dict}).
		\begin{figure}[H]
			\centering
			\begin{tikzpicture}
				\def\matHeight{5}
				\def\matWidth{3} 
				\draw[thick, fill=gray!20] (0,0) rectangle (\matWidth, \matHeight);
				\node at (0.5*\matWidth, 0.5*\matHeight) {\Large $\featuremtx \in \mathbb{R}^{\samplesize \times \nrfeatures}$};
				\draw[decorate, decoration={brace, amplitude=5pt}, thick] 
  				(-0.2, 0) -- (-0.2, \matHeight)
  				node[midway, left=8pt] {$\samplesize$};
				\draw[decorate, decoration={brace, amplitude=5pt, mirror}, thick] 
        				(0, -0.2) -- (\matWidth, -0.2) 
        				node[midway, below=8pt] {$\nrfeatures$};
			\end{tikzpicture}
		\caption{Illustration of a tall \gls{matrix}. \label{fig_tallmatrix_dict}}
		\end{figure}
		See also: \gls{matrix}, \gls{widematrix}. },
	text={tall matrix}, 
	firstplural={tall matrices}, 
	plural={tall matrices}, 
	type=math,
   	first={tall matrix} 
}

\newglossaryentry{randomexperiment}
{name={random experiment},
	description={A random experiment\index{random experiment} is a physical (or abstract) process 
    	 	that produces an \gls{outcome} $\outcome$ from a set $\samplespace$ of possibilities. 
	 	This set of all possible \glspl{outcome} is referred to as the \gls{samplespace} of 
	 	the experiment. The key characteristic of a random experiment is that its 
	 	\gls{outcome} is unpredictable (or uncertain). Any measurement or observation 
	 	of the \gls{outcome} is an \gls{rv}, i.e., a \gls{function} of the \gls{outcome} $\outcome \in \samplespace$. 
	 	\Gls{probability} theory uses a \gls{probspace} as a mathematical structure for the study of 
	 	random experiments. A key conceptual property of a random experiment is that it can 
	 	be repeated under identical conditions. Strictly speaking, repeating a random experiment 
	 	a given number of $\samplesize$ times defines a new random experiment. The \glspl{outcome} 
	 	of this new experiment are length-$\samplesize$ \glspl{sequence} of \glspl{outcome} 
	 	from the original experiment (see Fig.\ \ref{fig_randomexperiment_dict}). While the \gls{outcome} of a single experiment is 
	 	uncertain, the long-run behavior of the \glspl{outcome} of repeated experiments 
	 	tends to become increasingly predictable. This informal claim can be made 
	 	precise via fundamental results of \gls{probability} theory, such as the \gls{lln} 
	 	and the \gls{clt}.
	 	\begin{figure}[H]
		\begin{center}
	 		\begin{tikzpicture}[>=Stealth, node distance=1.5cm and 2cm, every node/.style={font=\small}]
			\node (experiment) [draw, rectangle, rounded corners, minimum width=2.6cm, align=center] {random\\experiment};
			\node (omega) [right=of experiment] {$\outcome \in \samplespace$};
			\coordinate (rightpad) at ($(omega.east) + (0.2,0)$);
			\draw[->] (experiment) -- (omega);
			\node (sequence) [below=of experiment, yshift=-0.5cm] {$(\outcome^{(1)}, \,\outcome^{(2)}, \,\dots, \,\outcome^{(\samplesize)})$};
			\node (sequence1) [below=of sequence, yshift=-0.5cm] {$(\datapoint^{(1)}, \,\datapoint^{(2)}, \,\dots, \,\datapoint^{(\samplesize)})$};
			\draw[->, thick] (experiment.south) -- node[midway, right, xshift=3pt] {repeat $\samplesize$ times} (sequence.north);
			\draw[->, thick] (sequence.south) -- node[midway, right, xshift=3pt] {\glspl{rv}} (sequence1.north);
			\path (experiment.south) -- (sequence.north) coordinate[pos=0.6] (repeatpoint);
        			\node[draw=black, rounded corners, dotted, fit={(experiment) (repeatpoint) (rightpad)}, inner sep=8pt, label=above:{new random experiment with $\samplespace' = \samplespace \times \ldots \times \samplespace$}] {};
	 		\end{tikzpicture}
	     	\end{center}
		\caption{Random experiment producing an \gls{outcome} $\outcome \in \samplespace$ from a set 
			of possibilities (i.e., a \gls{samplespace}) 
			$\samplespace$. Repeating the experiment $\samplesize$ times yields another random 
			experiment whose \glspl{outcome} are \glspl{sequence} 
			$(\outcome^{(1)}, \,\outcome^{(2)}, \,\dots, \,\outcome^{(\samplesize)}) \in \samplespace\times\ldots\times \samplespace$. 
			One example of a random experiment arising in many \gls{ml} applications is the gathering 
			of a \gls{trainset} $\datapoint^{(1)},\,\ldots,\,\datapoint^{(\samplesize)}$. \label{fig_randomexperiment_dict}}
	 	\end{figure} 
	 	Examples for random experiments arising in \gls{ml} applications include the following: 
	 	\begin{itemize} 
			\item \Gls{data} collection: The \glspl{datapoint} collected in \gls{erm}-based methods 
			can be interpreted as \glspl{rv}, i.e., as \glspl{function} of the \gls{outcome} $\outcome \in \samplespace$ 
			of a random experiment. 
			\item \Gls{stochGD} uses a random experiment at each \gls{iteration} to select a subset of 
			the \gls{trainset}. 
			\item \Gls{privprot} methods use random experiments to perturb  
			the \glspl{output} of an \gls{ml} method to ensure \gls{diffpriv}. 
	 	\end{itemize} 
		See also: \gls{outcome}, \gls{samplespace}, \gls{rv}, \gls{probability}, \gls{probspace}.},
 	firstplural={random experiments},
 	plural={random experiments},
	type=math,
 	first={random experiment},
 	text={random experiment}
}

\newglossaryentry{pseudoinverse}
{name={pseudoinverse},
	description={The \index{pseudoinverse}Moore–Penrose pseudoinverse $\mA^{+}$ 
 		of a \gls{matrix} $\featuremtx \in \mathbb{R}^{\samplesize \times \nrfeatures}$ 
		generalizes the notion of an \gls{inverse} \cite{GolubVanLoanBook}. 
		The pseudoinverse arises naturally in \gls{ridgeregression} for a 
		\gls{dataset} with \gls{featuremtx} $\featuremtx$ and \gls{labelvec} 
		$\labelvec$ \cite[Ch.\ 3]{hastie01statisticallearning}. 
		The \glspl{modelparam} learned by \gls{ridgeregression} 
  		are given by
  		\[
  		\widehat{\weights}^{(\regparam)}  = \big(\featuremtx^{T} \featuremtx + \regparam \mI \big)^{-1} \featuremtx^{T} \vy, \qquad \regparam > 0.
  		\]
  		We can then define the pseudoinverse $\featuremtx^{+} \in \mathbb{R}^{\nrfeatures \times \samplesize}$ via 
  		the limit \cite[Ch. 3]{benisrael2003generalized}:
  		\[
  		\lim_{\regparam \to 0^+} \widehat{\weights}^{(\regparam)} = \featuremtx^+ \vy.
  		\]
		See also: \gls{matrix}, \gls{inverse}, \gls{ridgeregression}. },
 	first={pseudoinverse},
 	type=math, 
 	text={pseudoinverse}
}

\newglossaryentry{mgf}
{name={moment generating function (MGF)}, 
	description={Consider the\index{moment generating function (MGF)} MGF $\mgf{x}{t}$ 
	 	of a real-valued \gls{rv} $x$, which is defined as
	 	$\mgf{x}{t} = \expect \{ \exp(t \cdot x) \}$ for any $t \in \mathbb{R}$ 
	 	for which this \gls{expectation} exists \cite[Sec. 21]{BillingsleyProbMeasure}. 
		As its name indicates, the MGF allows us to compute the moments 
	 	$\expect\{ x^{k} \}$ for $k \in \mathbb{N}$. 
	 	In particular, the $k$th moment is obtained by evaluating the $k$th 
	 	\gls{derivative} of $\mgf{x}{t}$ for $t=0$, i.e., $\expect\{ x^{k} \} = \mgfder{x}{k}{0}$. 
	 	This fact can be verified by the following identities: 
	 	\begin{align}
			\mgf{x}{t} & =\expect\{ \exp(t \cdot x)  \} \nonumber \\ 
			& \stackrel{(a)}{=} \expect\!\bigg\{\sum_{k=0}^{\infty} \frac{t^{k}}{k!} x^{k}\bigg\}  \nonumber \\ 
			& \stackrel{(b)}{=}  \sum_{k=0}^{\infty} \frac{t^{k}}{k!}\, \expect\!\big\{ x^{k} \big\}. \nonumber
	 	\end{align}
	 	Here, step $(a)$ is due to the Taylor series expansion of 
	 	$\exp\,(t \cdot x)$ and step $(b)$ is valid when the MGF exists 
	 	for all $t$ in some interval $(-t_{0},t_{0})$ \cite[p. 278]{BillingsleyProbMeasure}.
	 	\begin{figure}[H]
			\centering
			\begin{tikzpicture}
			\begin{axis}[
				width=9cm, height=4.2cm,
				domain=-1:1,
				samples=200,
				xlabel={$x$},
				ylabel={},
				ytick=\empty,
				ytick={-1,-0.5,0,0.5,1}, 
            			yticklabels={$-1$,$-0.5$,$0$,$0.5$,$1$},
				xtick={-1,-0.5,0,0.5,1}, 
            			xticklabels={$-1$,$-0.5$,$0$,$0.5$,$1$},
				xmin=-1, xmax=1,
				legend style={at={(1.5,0.02)},anchor=south east}
				]
				\addplot[thick, dashed] {x};
				\addlegendentry{$x$}
				\addplot[thick, dotted] {x^2};
				\addlegendentry{$x^{2}$}
				\addplot[thick, dashdotdotted] {x^3};
				\addlegendentry{$x^{3}$}
			\end{axis}
			\end{tikzpicture}
		\caption{First few powers of an \gls{rv} $x$. The MGF 
			encodes the moments of $x$, which are the \glspl{expectation} 
			of the powers $x^{k}$ for $k=1,\,2,\,\ldots$.}
		\end{figure}
		The MGF is a useful tool for the study of sums of independent 
		\glspl{rv}. As a case in point, if $x$ and $y$ are independent 
		\glspl{rv}, then the MGF of their sum $z = x + y$ typically 
		satisfies $\mgf{z}{t} = \mgf{x}{t}\,\mgf{y}{t}$,
        		i.e., the MGF of the sum is typically the pointwise product of the 
		individual MGFs \cite[p.~280]{BillingsleyProbMeasure}.
		\\
 		See also: \gls{rv}, \gls{expectation}. }, 
 	first={moment generating function (MGF)}, 
 	firstplural={moment generating functions (MGFs)}, 
 	plural={MGFs}, 
 	type=math, 
 	text={MGF}
} 

\newglossaryentry{chernoffbound}
{name={Chernoff bound}, 
	description={The Chernoff bound\index{Chernoff bound} is a \gls{concentrationinequ} 
		derived as a direct application of \gls{markovsinequality} \cite[Ch.\ 2]{vershynin2018high}. 
		Let $\feature$ be a real-valued \gls{rv} such that its \gls{mgf} 
		$\mgf{\feature}{t}=\expect\{\exp\,(t x)\}$ exists for some $t>0$. 
		Applying \gls{markovsinequality} to the nonnegative 
		\gls{rv} $\exp\,(t \feature)$ yields, for any $\eta\in\mathbb{R}$, 
		\begin{equation}
			\prob{ \feature \ge \eta }
			= \prob{ \exp(t \feature) \ge \exp(t\eta)}
			\le \exp(-t\eta)\, \expect\{\exp\,(t \feature)\}\nonumber.
		\end{equation}
		Note that this is actually an entire family of upper bounds, parameterized 
		by all valid choices for $t>0$ (i.e., $\mgf{\feature}{t}$ must exist). 
		\\
		See also: \gls{concentrationinequ}, \gls{markovsinequality}, \gls{chebyshevsinequality}, 
		\gls{hoeffdingsinequality}.}, 
 	first={Chernoff bound}, 
 	type=math, 
 	text={Chernoff bound}
}

\newglossaryentry{rankdeficient}
{name={rank-deficient},
	description={A \gls{matrix} $\mA \in \mathbb{R}^{\samplesize \times \nrfeatures}$ 
         	is \gls{rank}-deficient\index{rank-deficient} if it is not 
		\gls{fullrank}, i.e., when $\rank{\mA} < \min\{\samplesize,\nrfeatures\}$.
 		\begin{figure}[H]
			\begin{center}
			\begin{tikzpicture}[x=2cm]
				\begin{scope}
					\draw[->, thick] (0,0) -- (1,0) node[below] {$\vu^{(1)}$};
					\draw[->, thick] (0,0) -- (0,1) node[above] {$\vu^{(2)}$};
				\end{scope}
				\begin{scope}[shift={(3.2,0)}]
					\coordinate (A) at (0.2,0.0);
					\coordinate (B) at (2.0,0.0);
					\draw[->, very thick, red] (0,0) -- (A) node[below,yshift=-2pt] {$\mA \vu^{(1)}$};
					\draw[->, very thick, red] (0,0) -- (B) node[above,yshift=2pt] {$\mA \vu^{(2)}$};
				\end{scope}
				\draw[->, thick] (1.6,0.5) to[bend left] node[midway, above] {$\mA$} (2.7,0.5);
			\end{tikzpicture}
			\end{center}
		\caption{Example of a \gls{rank}-deficient \gls{matrix} 
			$\mA \in \mathbb{R}^{2 \times 2}$.	\label{fig_matrix_rank_defdict}} 
		\end{figure} 
		A \gls{rank}-deficient \gls{matrix} has \gls{linearlydep} columns,
		meaning that some columns can be expressed as linear combinations of others.
		As a consequence, the linear system $\mA \vw = \vy$ has either no solution
		or infinitely many solutions, since different choices of $\vw$ can yield
		the same product $\mA \vw$.
		In \gls{linreg}, the \glspl{modelparam} are found by solving the
		\gls{normalequations} $\featuremtx^{\top}\featuremtx \weights = \featuremtx^{\top} \vy$.
		If the \gls{featuremtx} $\featuremtx$ is \gls{rank}-deficient
		(e.g., because two \glspl{feature} are perfectly correlated),
		then $\featuremtx^{\top}\featuremtx$ is also \gls{rank}-deficient
		and the \gls{normalequations} have infinitely many solutions, i.e.,
		multiple \glspl{modelparam} $\weights$ fit the \gls{trainset} equally well.
		\\
   		See also: \gls{fullrank}, \gls{dimension}, \gls{vectorspace}. }, 
	first={rank-deficient}, 
   	type=math,
   	text={rank-deficient}
}

\newglossaryentry{fullrank}
{name={full-rank},
 	description={A \gls{matrix} $\mA \in \mathbb{R}^{\samplesize \times \nrfeatures}$ 
  		is\index{full-rank} full-\gls{rank} if it has \gls{maximum} \gls{rank} \cite{StrangLinAlg2016}. 
  		For a \gls{tallmatrix}, i.e., when $\nrfeatures < \samplesize$, being 
  		full-\gls{rank} means that its \gls{rank} is equal to $\nrfeatures$. 
 		\begin{figure}[H]
			\centering
			\begin{tikzpicture}[every node/.style={font=\small}]
			\node at (0,2) {$
			\mA =
			\begin{pmatrix}
			1 & 2\\
			3 & 4
			\end{pmatrix}
			$};
			\node[below=0.8cm of {$(0,2)$}] {\small full-\gls{rank} square};
			\node at (4.5,2) {$
			\mB =
			\begin{pmatrix}
			1 & 2\\
			2 & 4
			\end{pmatrix}
			$};
			\node[below=0.8cm of {$(4.5,2)$}] {\small \gls{rankdeficient} square};
			\node at (0,-1.0) {$
			\mC =
			\begin{pmatrix}
			1 & 0\\
			0 & 1\\
			1 & 1
			\end{pmatrix}
			$};
			\node[below=1.2cm of {$(0,-1.0)$}] {\small full-\gls{rank} \gls{tallmatrix}};
			\node at (4.5,-1.0) {$
			\mD =
			\begin{pmatrix}
			1 & 2 & 3\\
			2 & 4 & 6
			\end{pmatrix}
			$};
			\node[below=1.2cm of {$(4.5,-1.0)$}] {\small \gls{rankdeficient} \gls{widematrix}};
			\end{tikzpicture}
		\caption{Examples of full-\gls{rank} and \gls{rankdeficient} \glspl{matrix}.}
		\end{figure} 
  		A square \gls{matrix} is full-\gls{rank} if and only if it is \gls{invertible}. 
		\\ 
  		See also: \gls{matrix}, \gls{rank}, \gls{dimension}, \gls{linearmap}, \gls{columnspace}.}, 
	text={full-rank}, 
  	type=math,
  	first={full-rank} 
}

\newglossaryentry{rank}
{name={rank},
	description={The rank\index{rank} of a \gls{matrix} $\mA \in \mathbb{R}^{\samplesize \times \nrfeatures}$, 
 		denoted by $\rank{\mA}$, is the \gls{maximum} number of \gls{linearlyindep} columns 
 		of $\mA$ \cite{StrangLinAlg2016}. Equivalently, the rank can be defined as the 
 		\gls{dimension} of the \gls{columnspace} $\linspan{\mA} = \big\{ \mA \weights \mbox{ for some } 
 		\weights \in \mathbb{R}^{\nrfeatures} \big\}$. The rank of a \gls{matrix} 
 		$\mA \in \mathbb{R}^{\samplesize \times \nrfeatures}$ can neither exceed the 
 		number of rows nor the number of columns of $\mA$ \cite{Horn91}, \cite{MeyerMatrixAnalysis}, 
		i.e., $\rank{\mA} \leq \min \{ \samplesize, \nrfeatures \}$.  
		\\ 
 		See also: \gls{matrix}, \gls{dimension}, \gls{columnspace}, \gls{linearmap}.}, 
	text={rank}, 
	type=math,
	first={rank} 
}

\newglossaryentry{invertible}
{name={invertible (matrix)}, 
	description={A square \gls{matrix} $\mA \in \mathbb{R}^{n \times n}$ is 
		called invertible\index{invertible} if there exists a 
		\gls{matrix} $\mB \in \mathbb{R}^{n \times n}$ such that 
		$\mA \mB = \mB \mA = \mI$, where $\mI$ is the $n \times n$ 
		identity \gls{matrix} \cite{Strang2007}, \cite{Axler2025}, \cite{StrangLinAlg2016}. 
		In this case, $\mB$ is called the \gls{inverse} of $\mA$ and is denoted by $\mA^{-1}$.
		\\
		See also: \gls{matrix}, \gls{inverse}. },
	first={invertible},
	type=math,
	text={invertible}
}

\newglossaryentry{inverse}
{name={inverse matrix},
	description={An inverse \gls{matrix}\index{inverse matrix} $\mA^{-1}$ is defined for a 
 		square \gls{matrix} $\mA \in \mathbb{R}^{n \times n}$ that is of \gls{fullrank}, meaning its 
 		columns are \gls{linearlyindep}. In this case, $\mA$ is said to be \gls{invertible}, 
 		and its inverse satisfies
 		\[
 		\mA \mA^{-1} = \mA^{-1} \mA = \mI.
 		\]  	
     		A square \gls{matrix} is \gls{invertible} if and only if its \gls{det} is nonzero. Inverse \glspl{matrix} are 
     		fundamental in solving systems of linear equations and in the closed-form solution of 
     		\gls{linreg} \cite{Strang2007}, \cite{Horn91}.  The concept of an inverse \gls{matrix} can be extended 
     		to \glspl{matrix} that are not square or do not have \gls{fullrank}. One may define a ``left inverse'' $\mB$ 
     		satisfying $\mB \mA = \mI$ or a ``right inverse'' $\mC$ satisfying $\mA \mC = \mI$. 
     		For general rectangular or singular \glspl{matrix}, the Moore–Penrose \gls{pseudoinverse}
     		$\mA^{+}$ provides a unified concept of a generalized inverse \gls{matrix} \cite{GolubVanLoanBook}.
 		 \begin{figure}[H]
 			\centering
 			\begin{tikzpicture}[x=2cm,y=2cm]
 			\begin{scope}
 				\draw[->, thick] (0,0) -- (1,0) node[below right] {$\vx$};
 				\draw[->, thick] (0,0) -- (0,1) node[above left] {$\vy$};
 			\end{scope}
 			\begin{scope}[shift={(2.0,0)}]
 				\coordinate (A) at (1.5,0.5);
 				\coordinate (B) at (-0.2,1.2);
				\draw[->, very thick, red] (0,0) -- (A) node[pos=0.5, below right] {$\mA \vx$};
 				\draw[->, very thick, red] (0,0) -- (B) node[above right] {$\mA \vy$};
 			\end{scope}
 			\begin{scope}[shift={(4.9,0)}]
 				\draw[->, very thick, blue] (0,0) -- (1,0) node[pos=0.5, below] {$\mA^{-1} (\mA \vx) = \vx$};
 				\draw[->, very thick, blue] (0,0) -- (0,1) node[above] {$\mA^{-1} (\mA \vy) = \vy$};
 			\end{scope}
 			\draw[->, thick, bend left=20] (1.2,0.4) to node[above] {$\mA$} (1.8,0.4);
 			\draw[->, thick, bend left=20] (3.8,0.4) to node[below] {$\mA^{-1}$} (4.4,0.4);
 			\end{tikzpicture}
 		\caption{\Gls{matrix} $\mathbf{A}$ representing a linear transformation of $\mathbb{R}^{2}$; inverse \gls{matrix} $\mathbf{A}^{-1}$ 
 			representing the inverse transformation. \label{fig_matrix_inverse_dict}} 
 		\end{figure}
		See also: \gls{matrix}, \gls{det}, \gls{linreg}, \gls{pseudoinverse}.},
	first={inverse matrix},
	type=math,
	text={inverse matrix}
}

\newglossaryentry{matrix}
{name={matrix},
	description={A matrix\index{matrix} of size $\samplesize \times \nrfeatures$ is a 2-D array of numbers, 
 		which is denoted by 
		$$
  		\mA = \begin{pmatrix}
   		A_{1,1} & A_{1,2} & \dots  & A_{1,\nrfeatures} \\
		A_{2,1} & A_{2,2} & \dots  & A_{2,\nrfeatures} \\
		\vdots  & \vdots  & \ddots & \vdots \\
		A_{\samplesize,1} & A_{\samplesize,2} & \dots  & A_{\samplesize,\nrfeatures}
		\end{pmatrix} \in \mathbb{R}^{\samplesize \times \nrfeatures}.
		$$
		Here, $A_{\sampleidx,\featureidx}$ denotes the matrix entry in the $\sampleidx$th row and the 
		$\featureidx$th column. Matrices are useful representations of various mathematical objects \cite{StrangLinAlg2016},
		including the following:
		\begin{itemize}
			\item Systems of linear equations: We can use a matrix to represent a system of linear equations 
			$$ \begin{pmatrix}
			A_{1,1} & A_{1,2} \\
			A_{2,1} & A_{2,2}
			\end{pmatrix}
			\begin{pmatrix}
				w_1 \\
				w_2
			\end{pmatrix}
			=\begin{pmatrix}
				y_1 \\
				y_2
			\end{pmatrix}
			\text{ compactly as } \mA \vw = \vy.
			$$
    			One important example of systems of linear equations is the optimality condition for the 
    			\glspl{modelparam} within \gls{linreg}. 
			\item \Glspl{linearmap}:
			Consider a $\nrfeatures$-dimensional \gls{vectorspace} $\mathcal{U}$ and a $\samplesize$-dimensional \gls{vectorspace} $\mathcal{V}$. 
			If we fix a \gls{basis} $\mathbf{u}^{(1)},\,\ldots,\,\mathbf{u}^{(\nrfeatures)}$ for $\mathcal{U}$ and a \gls{basis} $\mathbf{v}^{(1)},\,\ldots,\,\mathbf{v}^{(\samplesize)}$ 
			for $\mathcal{V}$, each matrix $\mA \in \mathbb{R}^{\samplesize \times \nrfeatures}$ naturally defines a 
			\gls{linearmap} $\alpha: \mathcal{U} \rightarrow \mathcal{V}$ (see Fig.\ \ref{fig_matrix_dict}) such that
   			$$\vu^{(\featureidx)} \mapsto \sum_{\sampleidx=1}^{\samplesize} A_{\sampleidx,\featureidx} \vv^{(\sampleidx)}.$$
			\item \Glspl{dataset}: We can use a matrix to represent a \gls{dataset}. Each row 
			corresponds to a single \gls{datapoint}, and each column corresponds to a specific 
			\gls{feature} or \gls{label} of a \gls{datapoint}. 
		\end{itemize}
		\begin{figure}[H]
		\begin{center}
		\begin{tikzpicture}[x=2cm]
			\begin{scope}
				\draw[->, thick] (0,0) -- (1,0) node[below] {$\vu^{(1)}$};
				\draw[->, thick] (0,0) -- (0,1) node[above] {$\vu^{(2)}$};
			\end{scope}
			\begin{scope}[shift={(3.2,0)}]
				\draw[->, thick] (0,0) -- (1,0) node[below] {$\vv^{(1)}$};
				\draw[->, thick] (0,0) -- (0,1) node[above] {$\vv^{(2)}$};
				\coordinate (A) at (0.2,-1.0);
				\coordinate (B) at (0.4,1.2);
				\draw[->, very thick, red] (0,0) -- (A) node[below,right] {$\mA \vu^{(1)}\!=\!A_{1,1}\vv^{(1)}\!+\!A_{2,1}\vv^{(2)}$};
				\draw[->, very thick, red] (0,0) -- (B) node[right,xshift=1pt] {$\mA \vu^{(2)}$};
			\end{scope}
			\draw[->, thick] (1.6,0.5) to[bend left] node[midway, above] {$\mA$} (2.7,0.5);
		\end{tikzpicture}
		\end{center}
		\caption{Matrix $\mA$ defining a \gls{linearmap} between two \glspl{vectorspace}. \label{fig_matrix_dict}} 
		\end{figure}
		See also: \gls{linearmap}, \gls{dataset}, \gls{linmodel}. },
	first={matrix},
	firstplural={matrices},
	type=math,
	plural={matrices},
	text={matrix}
}

\newglossaryentry{hyperplane}
{name={hyperplane},
	description={A hyperplane\index{hyperplane} is a $(\nrfeatures-1)$-dimensional affine 
		\gls{subspace} of a $\nrfeatures$-dimensional \gls{vectorspace}. In the 
		context of a \gls{euclidspace} $\mathbb{R}^{\nrfeatures}$, a 
		hyperplane is a set of the following form:
 		\[ 
		\{ \vx \in \mathbb{R}^\nrfeatures : \vw^\top \vx = b \},
  		\]
		where $\vw \in \mathbb{R}^d \setminus \{0\}$ is a normal \gls{vector} 
		and $b \in \mathbb{R}$ is an offset. Such a hyperplane divides
		$\mathbb{R}^\nrfeatures$ into two \glspl{halfspace}:
		\[\{ \vx \in \mathbb{R}^\nrfeatures : \vw^\top \vx > b \}  \quad
		\text{and}  \quad  \{ \vx \in \mathbb{R}^\nrfeatures : \vw^\top \vx > b \}.\]
 		Hyperplanes arise as the \glspl{decisionboundary} of \glspl{linclass}.
		\\
		See also: \gls{subspace}, \gls{vectorspace}, \gls{euclidspace}, \gls{decisionboundary}, \gls{innerproduct}. },
	first={hyperplane},
	type=math,
	plural={hyperplanes}, 
	firstplural={hyperplanes},
	text={hyperplane}
}

\newglossaryentry{normalvector}
{name={normal vector},
	description={See\index{normal vector} \gls{hyperplane}.},
	first={normal vector},
	type=math,
	plural={normal vectors}, 
	firstplural={normal vectors},
	text={normal vector}
}

\newglossaryentry{halfspace}
{name={halfspace},
	description={See\index{halfspace} \gls{hyperplane}.},
	first={halfspace},
	type=math,
	plural={halfspaces}, 
	firstplural={halfspaces},
	text={halfspace}
}

\newglossaryentry{subspace}
{name={subspace},
	description={A subset of a \gls{vectorspace} $\mathcal{V}$ is a subspace\index{subspace} of $\mathcal{V}$ if it is also a 
		\gls{vectorspace} with respect to the same operations as $\mathcal{V}$.
		Fig.\ \ref{fig:subspace} illustrates this for the 
		\gls{vectorspace} $\mathbb{R}^2$ (over the \gls{field} $\mathbb{R}$).
		\begin{figure}[H] 
			\centering
			\begin{tikzpicture}[>=stealth,scale=0.9]
			\draw[->,thick](-2.5,0)--(2.5,0)node[right]{$x$};
			\draw[->,thick](0,-2.5)--(0,2.5)node[above]{$y$};
			\draw[ultra thick,blue!70!black](-2.2,-1.1)--(2.2,1.1)node[right]{$\mathcal{U}_1$};
			\draw[ultra thick,red!70!black](-1,2)--(1,-2)node[below]{$\mathcal{U}_2$};
			\filldraw[black](0,0)circle(1.5pt)node[anchor=north east]{$\mathbf{0}$};
			\end{tikzpicture}
		\caption{Any $\featurevec \in \mathbb{R}^2$ defines a subspace of $\mathcal{U}_\featurevec \subset \mathbb{R}^2$ 
			by $\mathcal{U}_\featurevec = \{ \lambda \cdot \featurevec \; : \; \lambda \in \mathbb{R}\} \subset \mathbb{R}^2$. 
			For $\featurevec \neq \mathbf{0}$, this corresponds to straight lines through the origin (illustrated by $\mathcal{U}_1$ 
			and $\mathcal{U}_2$). For the zero \gls{vector} $\featurevec =\mathbf{0}$, this leads to the trivial subspace $\{\mathbf{0}\}$ 
			just containing the zero \gls{vector}. \label{fig:subspace}}
		\end{figure}
		See also: \gls{vectorspace}, \gls{dimension}, \gls{basis}.},
	type=math, 
	plural={subspaces},
	first={subspace},
	text={subspace}
}

\newglossaryentry{columnspace}
{name={column space},
	description={The column space\index{column space} of a \gls{matrix} 
		$\mA \in \mathbb{R}^{\samplesize \times \nrfeatures}$,
		denoted by $\linspan{\mA}$, is the set of all linear combinations of the 
		columns of $\mA$. In other words, 
		$$ \linspan{\mA} = \{ \mA \weights : \weights \in \mathbb{R}^{\nrfeatures} \}. $$
		The column space $\linspan{\mA}$ of the \gls{matrix} $\mA$ 
		is a \gls{subspace} of the \gls{euclidspace} $\mathbb{R}^{\samplesize}$.
			   \\
		See also: \gls{matrix}, \gls{vectorspace}.},
	type=math,
	first={column space},
	text={column space}
}

\newglossaryentry{mvndist}
{name={multivariate normal distribution}, 
	description={The\index{multivariate normal distribution} multivariate \gls{normaldist}, 
		which is denoted by $\mvnormal{\meanvecgeneric}{\covmtxgeneric}$, is a fundamental 
		\gls{probmodel} for numerical \glspl{featurevec} of fixed dimension $\nrfeatures$. 
		It defines a family of \glspl{probdist} over \gls{vector}-valued \glspl{rv} 
		$\featurevec \in \mathbb{R}^{\nrfeatures}$~\cite{BertsekasProb}, \cite{GrayProbBook}, \cite{Lapidoth09}. 
		Each distribution in this family is fully specified by its \gls{mean} \gls{vector} 
		$\meanvecgeneric \in \mathbb{R}^{\nrfeatures}$ and \gls{covmtx} 
		$\covmtxgeneric \in \mathbb{R}^{\nrfeatures \times \nrfeatures}$. When the 
		\gls{covmtx} $\covmtxgeneric$ is \gls{invertible}, the corresponding \gls{probdist} is 
		characterized by the following \gls{pdf}:
		\[p(\featurevec) = 
 		\frac{1}{\sqrt{(2\pi)^{\nrfeatures} \det\,(\covmtxgeneric)}} 
 		\exp\left[ -\frac{1}{2} 
 		(\featurevec - \meanvecgeneric)\,^{T}\, \covmtxgeneric^{-1} 
 		(\featurevec - \meanvecgeneric) \right].
 		\]
		Note that this \gls{pdf} is only defined when $\covmtxgeneric$ is \gls{invertible}.
   		More generally, any \gls{rv} $\featurevec \sim \mvnormal{\meanvecgeneric}{\covmtxgeneric}$ 
   		admits the following representation:
  		\[
    		\featurevec = \mA \vz + \meanvecgeneric,
   		\]
   		where $\vz \sim \mvnormal{\mathbf{0}}{\mathbf{I}}$ is a \gls{stdnormvec} 
   		and $\mA \in \mathbb{R}^{\nrfeatures \times \nrfeatures}$ satisfies $\mA \mA^\top = \covmtxgeneric$. 
   		This representation remains valid even when $\covmtxgeneric$ is singular, in which case $\mA$ 
   		is not \gls{fullrank}~\cite[Ch. 23]{Lapidoth2017}.
   		The family of multivariate \glspl{normaldist} is exceptional among \glspl{probmodel} for numerical 
   		quantities, at least for the following reasons. First, the family is closed under affine 
   		transformations, i.e.,
		\[ 
		\featurevec \sim \mathcal{N}(\meanvecgeneric,\covmtxgeneric) \mbox{ implies } 
		\mB\featurevec\!+\!\vc \sim \mathcal{N}\big( \mB\meanvecgeneric+\vc,\mB \covmtxgeneric \mB\,^{T} \big). 
		\]
		Second, the \gls{probdist} $\mathcal{N}(\mathbf{0},\covmtxgeneric)$ maximizes the 
		\gls{diffentropy} among all distributions with the same \gls{covmtx} $\covmtxgeneric$~\cite{coverthomas}. 
		\\ 
		See also: \gls{probmodel}, \gls{probdist}, \gls{stdnormvec}, \gls{diffentropy}, \gls{gaussrv}.}, 
	first={multivariate normal distribution},
	type=math, 
	text={multivariate normal distribution}
}

\newglossaryentry{stdnormvec}
{name={standard normal random vector}, 
	description={A\index{standard normal random vector} standard normal random \gls{vector} 
		is an \gls{rv} $\featurevec=\big(\feature_{1}, \,\ldots, \,\feature_{\nrfeatures}\big)\,^{T}$ 
		whose entries are \gls{iid} \glspl{gaussrv} $x_{\featureidx} \sim \mathcal{N}(0,1)$. 
		The \gls{probdist} of a standard normal random \gls{vector} is a special case 
		of a \gls{mvndist} $\featurevec \sim \mathcal{N}(\mathbf{0},\mathbf{I})$.
		\\ 
		See also: \gls{vector}, \gls{iid}, \gls{gaussrv}, \gls{mvndist}, \gls{rv}.}, 
	first={standard normal random vector},
	type=math, 
	text={standard normal random vector}
}

\newglossaryentry{continuous}
{name={continuous}, 
	description={A \gls{function}\index{continuous} $f: \mathbb{R}^{\nrfeatures} \to \mathbb{R}$ is 
		continuous at a point $\featurevec' \in \mathbb{R}^{\nrfeatures}$ if, for 
	 	every $\epsilon > 0$, there is a $\delta > 0$ such that, for all 
	 	$\featurevec \in \mathbb{R}^{\nrfeatures}$ with $\normgeneric{\featurevec - \featurevec'}{2} < \delta$, 
	 	it holds that $|f(\featurevec) - f(\featurevec')| < \epsilon$ \cite{RudinBookPrinciplesMatheAnalysis}. 
	 	In other words, we can make $f(\featurevec)$ arbitrarily close to $f(\featurevec')$ 
	 	by choosing $\featurevec$ sufficiently close to $\featurevec'$. 
		\begin{figure}[H]
			\centering
	 		\begin{tikzpicture}[
				>=stealth, 
    				thick,
    				declare function={f(\x) = 0.3*(\x-2)^3 + 2.5;}
				]
				\def\xprime{2}      
    				\def\epsilonval{1.2}   
				\def\deltaval{1.3}     
				\def\xmax{5.5}
				\def\ymax{5}
				\fill[blue!5] (0, {f(\xprime)-\epsilonval}) rectangle (\xmax, {f(\xprime)+\epsilonval});
				\draw[blue!30, dashed] (0, {f(\xprime)-\epsilonval}) -- (\xmax, {f(\xprime)-\epsilonval});
				\draw[blue!30, dashed] (0, {f(\xprime)+\epsilonval}) -- (\xmax, {f(\xprime)+\epsilonval});
				\fill[red!10] ({\xprime-\deltaval}, 0) rectangle ({\xprime+\deltaval}, \ymax);
				\draw[red!40, dashed] ({\xprime-\deltaval}, 0) -- ({\xprime-\deltaval}, \ymax);
				\draw[red!40, dashed] ({\xprime+\deltaval}, 0) -- ({\xprime+\deltaval}, \ymax);
				\fill[purple!20] ({\xprime-\deltaval}, {f(\xprime)-\epsilonval}) rectangle ({\xprime+\deltaval}, {f(\xprime)+\epsilonval});
				\draw[purple, thick] ({\xprime-\deltaval}, {f(\xprime)-\epsilonval}) rectangle ({\xprime+\deltaval}, {f(\xprime)+\epsilonval});
				\draw[->] (-0.5,0) -- (\xmax,0) node[right] {$\feature$};
				\draw[->] (0,-0.5) -- (0,\ymax) node[above] {};
				\draw[line width=1.5pt, black!80] plot[domain=0.2:4.5, samples=100] (\x, {f(\x)}) 
				node[right] {$f(\feature)$};
				\draw[dashed] (\xprime, 0) -- (\xprime, {f(\xprime)}) -- (0, {f(\xprime)});
				\fill[black] (\xprime, {f(\xprime)}) circle (2pt);
				\node[below,yshift=-5pt] at (\xprime, 0) {$\feature'$};
				\node[left] at (0, {f(\xprime)}) {$f(\feature')$};
				\draw[decorate, decoration={brace, amplitude=4pt}, blue!80] 
				(-0.2, {f(\xprime)}) -- (-0.2, {f(\xprime)+\epsilonval}) 
				node[midway, left=4pt] {$\varepsilon$};
				\draw[decorate, decoration={brace, amplitude=4pt}, blue!80] 
				(-0.2, {f(\xprime)-\epsilonval}) -- (-0.2, {f(\xprime)}) 
				node[midway, left=4pt] {$\varepsilon$};
				\draw[decorate, decoration={brace, amplitude=4pt, mirror}, red!80] 
				(\xprime, -0.2) -- ({\xprime+\deltaval}, -0.2) 
				node[midway, below=4pt] {$\delta$};
				\draw[decorate, decoration={brace, amplitude=4pt, mirror}, red!80] 
				({\xprime-\deltaval}, -0.2) -- (\xprime, -0.2) 
				node[midway, below=4pt] {$\delta$};
				\node[align=left, purple!40!black, font=\small] at (6.2, 1.5) 
				{If $| \feature - \feature' | < \delta$,\\
				then $| f(\feature) - f(\feature') | < \varepsilon$.};
			\end{tikzpicture}
		\caption{\Gls{function} $f(\feature) = 0.3(\feature-2)^3 + 2.5$, continuous 
		         at every $\feature'$.}
		\end{figure}
		If $f$ is continuous at every point $\featurevec' \in \mathbb{R}^{\nrfeatures}$, then $f$ is said to be 
	 	continuous on $\mathbb{R}^{\nrfeatures}$. The notion of a continuous 
	 	\gls{function} can be naturally extended to \glspl{function} between general \glspl{metricspace} 
		\cite{RudinBookPrinciplesMatheAnalysis}.
		\\
		See also: \gls{euclidspace}, \gls{metric}.},
	first={continuous},
	type=math,
	text={continuous}
}

\newglossaryentry{minimum}
{name={minimum},
	description={Given a set of real numbers, the minimum\index{minimum} is the smallest of those numbers.
		Note that for some sets, such as the set of negative real numbers, the minimum does not exist.
		\\
		See also: \gls{maximum}. },
	firstplural={minima}, 
 	plural={minima},
	type=math, 
	first={minimum},
	text={minimum}
}

\newglossaryentry{co-domain}
{name={co-domain}, 
	description={The co-\gls{domain}\index{co-domain} of a \gls{function} 
		$f: \mathcal{U} \rightarrow \mathcal{V}$ is the set $\mathcal{V}$ 
		into which $f$ maps elements of its \gls{domain} $\mathcal{U}$.  
		\begin{figure}[H]
			\centering
			\begin{tikzpicture}[
			>=stealth, 
			node distance=2cm,
			scale=1.0, every node/.style={transform shape} 
			]
			\draw[thick, fill=blue!5] (0,0) ellipse (0.8cm and 1.4cm);
			\node[] at (0, 1.6) {\gls{domain}};
			\draw[thick, fill=red!5] (3.5,0) ellipse (1cm and 1.6cm);
			\node[] at (3.5, 1.8) {co-\gls{domain}};
			\draw[dashed, fill=green!10, thick, green!60!black] (3.5, -0.3) circle (0.6cm);
			\node[] at (3.5, -1.1) {range};
			\fill (0, 0.5) circle (1.5pt) coordinate (a1);
			\fill (0, -0.5) circle (1.5pt) coordinate (a2);
			\fill (3.5, 0) circle (1.5pt) coordinate (b1);
			\fill (3.5, -0.5) circle (1.5pt) coordinate (b2);
			\fill (3.5, 1.0) circle (1.5pt) coordinate (b_miss) node[right, font=\tiny, gray] {unused};
			\draw[->, semithick] (a1) -- (b1);
			\draw[->, semithick] (a2) -- (b2);
			\node[font=\footnotesize] at (1.75, 0.5) {$f$};
			\end{tikzpicture}
		\caption{\Gls{domain} and co-\gls{domain} of a \gls{function} $f: \mathcal{U} \rightarrow \mathcal{V}$. }
		\end{figure}
		See also: \gls{domain}, \gls{function}, \gls{map}.},
	first={co-domain},
	firstplural={co-domains}, 
	type=math, 
	plural={co-domains},
	text={co-domain}
}

\newglossaryentry{cdf}
{name={cumulative distribution function (cdf)},
	description={The\index{cumulative distribution function (cdf)} cdf 
		$\cdf{\feature}{\eta}$ of a real-valued \gls{rv} $\feature$ is \cite{AshProbMeasure}, \cite{papoulis}
		$$\cdf{\feature}{\eta} \defeq \prob{\feature \leq \eta}.$$
					\\ 
		See also: \gls{rv}, \gls{pdf}, \gls{probdist}.},
	first={cumulative distribution function (cdf)},
	firstplural={cumulative distribution functions (cdfs)}, 
	plural={cdfs}, 
	type=math,
	text={cdf} 
}

\newglossaryentry{weightedgraph}
{name={weighted graph},
	description={A weighted \gls{graph} is a \gls{graph}\index{weighted graph} whose edges 
		are assigned numeric weights. Typically, these \glspl{edgeweight} 
		are nonnegative real numbers. For example, if a \gls{graph} represents 
		a road network with nodes corresponding to intersections and edges representing 
		road segments, the \gls{edgeweight} could represent the capacity (measured 
		in \gls{maximum} vehicles per hour) of the road segment \cite{NewmannBook}.  
		\begin{figure}[H] 
			\centering
			\begin{tikzpicture}[scale=1.5,
				node/.style={circle, fill=black, inner sep=1.9pt},
				lab/.style={anchor=west, xshift=3pt}
				]
				\node[node] (v1) at (0,0) {};
				\node[node] (v2) at (2,0) {};
				\node[node] (v3) at (1,1.5) {};
				\node[node] (v4) at (3,1.5) {};
				\node[anchor=north] at (v1) {$\nodeidx_1$};
				\node[anchor=north] at (v2) {$\nodeidx_2$};
				\node[anchor=south] at (v3) {$\nodeidx_3$};
				\node[anchor=south] at (v4) {$\nodeidx_4$};
				\draw [line width=1pt] (v1) -- node[midway, above] {$\edgeweight_{\nodeidx_1,\nodeidx_2}$} (v2);
				\draw (v2) -- node[midway, right] {$\edgeweight_{\nodeidx_2,\nodeidx_3}$} (v3);
				\draw (v3) -- node[midway, left] {$\edgeweight_{\nodeidx_3,\nodeidx_1}$} (v1);
				\draw (v2) -- node[midway, right] {$\edgeweight_{\nodeidx_2,\nodeidx_4}$} (v4);
			\end{tikzpicture}
		\caption{Weighted \gls{graph} with four nodes 
			$\nodes = \{\nodeidx_1, \nodeidx_2, \nodeidx_3, \nodeidx_4\}$ 
			and four edges $\edges = \{\{\nodeidx_1,\nodeidx_2\},
			\{\nodeidx_2,\nodeidx_3\}, \{\nodeidx_3,\nodeidx_1\}, 
			\{\nodeidx_2,\nodeidx_4\}\}$. Each edge is assigned a weight.}
		\end{figure}		
		See also: \gls{graph}.},
	first={weighted graph},
	type=math,
	firstplural={weighted graphs}, 
	plural={weighted graphs}, 
	text={weighted graph} 
}

\newglossaryentry{graph}
{name={graph},
	description={A graph\index{graph} $\graph = \pair{\nodes}{\edges}$ 
		consists of a node set $\nodes$ and an edge set $\edges$.
		Each edge $\edgeidx \in \edges$ is characterized by the nodes to which 
		it is \gls{connected} and in what precise sense. For example, 
		an edge of a \gls{directedgraph} is leaving one node 
		and pointing to another node. An edge of an \gls{undirectedgraph} 
		connects two nodes without any sense of 
		direction \cite{NewmannBook}, \cite{RockNetworks}. 
		In principle, there can also be several (parallel) edges that are 
		\gls{connected} to the same nodes in the same way \cite{RockNetworks}. 
		Moreover, edges may connect a node to itself, resulting in 
		so-called self-loops \cite{NewmannBook}. 
		A simple \gls{undirectedgraph} contains no 
		parallel edges and no self-loops \cite{WilsonGraph2010}. 
		Each edge $\edgeidx \in \edges$ of a simple \gls{undirectedgraph}
		can be identified with a set of two nodes ${\nodeidx,\nodeidx'}$. 
		\begin{figure}[H] 
			\centering
			\begin{tikzpicture}[scale=1,
				node/.style={circle, fill=black, inner sep=1.9pt},
				lab/.style={anchor=west, xshift=3pt}
				]
				\node[node] (v1) at (0,0) {};
				\node[node] (v2) at (2,0) {};
				\node[node] (v3) at (1,1.5) {};
				\node[node] (v4) at (3,1.5) {};
				\node[anchor=north] at (v1) {$\nodeidx_1$};
				\node[anchor=north] at (v2) {$\nodeidx_2$};
				\node[anchor=south] at (v3) {$\nodeidx_3$};
				\node[anchor=south] at (v4) {$\nodeidx_4$};
				\draw [line width=1pt] (v1) -- (v2);
				\draw (v2) -- (v3);
				\draw (v3) -- (v1);
				\draw (v2) -- (v4);
			\end{tikzpicture}
		\caption{Simple \gls{undirectedgraph} with four nodes 
			$\nodes = \{\nodeidx_1, \nodeidx_2, \nodeidx_3, \nodeidx_4\}$ 
			and four edges $\edges = \{\{\nodeidx_1,\nodeidx_2\},
			\{\nodeidx_2,\nodeidx_3\}, \{\nodeidx_3,\nodeidx_1\}, 
			\{\nodeidx_2,\nodeidx_4\}\}$.}
		\end{figure}
		\Glspl{weightedgraph} assign a numerical value $\edgeweight_{\edgeidx}$, 
		referred to as \gls{edgeweight}, to each edge $\edgeidx \in \edges$.
					\\ 
		See also: \gls{directedgraph}, \gls{undirectedgraph}, \gls{weightedgraph}, \gls{edgeweight}.},
 	first={graph},
 	text={graph}, 
 	firstplural={graphs}, 
	plural={graphs}, 
 	type=math
}

\newglossaryentry{markovchain}
{name={Markov chain},
	description={A Markov chain\index{Markov chain} is a \gls{stochproc} $\{X_\timeidx \}_{\timeidx \in \mathbb{N}}$ 
		defined on a common \gls{probspace} and using the index set $\mathbb{N}$. The 
		\gls{rv} $X_\timeidx$ might represent (the generation of) a \gls{state} of a physical system 
		at the time instant $\timeidx$. The defining property of a Markov chain 
		is the \gls{markovprop} \cite{durrett2010probability}, \cite{papoulis}, \cite{NorrisMarkovChains1997}.
		For all $\timeidx \in \mathbb{N}$,
 		\begin{equation}
 			\nonumber \probdist^{(X_{\timeidx+1} \mid X_\timeidx,\,\ldots,\,X_1)} = \probdist^{(X_{\timeidx+1} \mid X_\timeidx)}. 
 		\end{equation}
 		In other words, the \gls{condprobdist} of the next \gls{state} $X_{\timeidx+1}$ 
		depends on the past $X_{\timeidx},\,X_{\timeidx-1},\,\dots,\,X_{1}$ 
		only through the current \gls{state} $X_{\timeidx}$. The concept of a 
		Markov chain can be generalized from discrete time (with index set $\mathbb{N}$) 
		to continuous time (with index set $\mathbb{R}$) \cite{NorrisMarkovChains1997}. 
		\\ 
		See also: \gls{stochproc}, \gls{condprobdist}, \gls{mdp}.},
	first={Markov chain},
	type=math,
    	text={Markov chain}, 
	plural={Markov chains}, 
	firstplural={Markov chains}
}

\newglossaryentry{markovprop}
{name={Markov property},
	description={See \gls{markovchain}\index{Markov property}.},
	first={Markov property},
	type=math,
	text={Markov property}
}

\newglossaryentry{em}
{name={expectation–maximization (EM)}, 
	description={\index{expectation–maximization (EM)}
	    	The EM \gls{algorithm} \cite{dempster1977maximum} is an iterative \gls{optmethod} for 
		approximately solving certain \gls{maxlikelihood} \glspl{optproblem} 
		that are difficult to solve directly \cite[Sec. 9.4]{BishopBook}, \cite[Sec. 11.4.7]{Murphy2012}.
	    	To motivate the EM \gls{algorithm} and explain its construction, 
		consider an \gls{ml} application involving a single observed \gls{datapoint} 
		with \gls{feature} $\feature \in \featurespace$, where $\featurespace$ 
		is a finite \gls{featurespace}. The \gls{data} generation 
		is modeled via a \gls{probmodel} that consists of an \gls{rv} $\feature'$ 
		with a \gls{pmf} $\pmf{\feature'}{\cdot;\weights}$. Here, 
		the actual \glspl{modelparam} $\weights \in \paramspace$—used for the \gls{data} 
		generation via sampling from $\pmf{\feature'}{\cdot;\weights}$—are unknown. A widely 
		used approach for estimating these \glspl{modelparam} is via the solutions of the following 
		\gls{maxlikelihood} problem:
		\begin{equation}
			\label{equ_def_ML_EM_dict}
			\min_{\weights \in \paramspace} - \log \pmf{\feature'}{\feature;\weights}.
		\end{equation}
		For some \glspl{probmodel}, such as a \gls{gmm}, this \gls{optproblem} can be 
		difficult to solve directly. As a work-around, one can often introduce an auxiliary 
		attribute $\truelabel \in \labelspace$, generated via some \gls{rv} $\truelabel'$, 
		such that the corresponding \gls{probmodel} $\pmf{\feature',\truelabel'}{\cdot,\cdot;\weights}$ 
		yields the following, much easier \gls{maxlikelihood} problem:
		\begin{equation}
			\label{equ_def_complete_EM_dict}
			\min_{\weights \in \paramspace} 
			- \log \pmf{\feature',\truelabel'}{\feature,\truelabel;\weights}.
		\end{equation}
		The attribute $\truelabel$ is introduced solely to simplify 
		\eqref{equ_def_complete_EM_dict}, but it is not observed in practice—only the 
		\gls{feature} $\feature$ is available. Thus, we cannot solve \eqref{equ_def_complete_EM_dict} 
		directly, as we do not know which value $\truelabel$ to plug into the \gls{pmf} 
		$ \pmf{\feature',\truelabel'}{\feature,\truelabel;\weights}$. 
		The EM method resolves this dilemma by alternating between the following two steps:
		1) an E-step in which a “soft’’ estimate of the auxiliary attribute $\truelabel$ is computed 
		in the form of the \gls{posteriordist} $\pmf{\truelabel'|\feature'}{\cdot;\widehat{\weights}}$
		using the current choice $\widehat{\weights}$ for the \glspl{modelparam}; and 
		2) an M-step in which a surrogate \gls{objfunc} derived from this \gls{posteriordist} is minimized.
		The completion of these two steps constitutes one full \gls{iteration} of 
		the EM method. 
		In more detail, the E-step produces the following \gls{function}:
		\[
		Q(\weights \mid \widehat{\weights})
		\defeq 
		- \sum_{\truelabel \in \labelspace} 
		\pmf{\truelabel'|\feature'}{\truelabel;\widehat{\weights}}
		\log\!\Big(
		\pmf{\feature',\truelabel'}{\feature,\truelabel;\weights}
		/\pmf{\truelabel'|\feature'}{\truelabel;\widehat{\weights}}
		\Big),
		\]
		and the M-step minimizes $Q(\weights \mid \widehat{\weights})$ over 
		$\weights \in \paramspace$.
		This \gls{function} satisfies the following two key properties 
		\cite[Sec. 9.4]{BishopBook},\cite[Sec. 11.4.7]{Murphy2012}:
		1) upper bound
		$Q(\weights\mid \widehat{\weights})
		\geq
		- \log \pmf{\feature'}{\feature;\weights}$
		for all $\weights \in \paramspace$;
		and 2) tightness
		$Q(\widehat{\weights}\mid \widehat{\weights})
		=- \log \pmf{\feature'}{\feature;\widehat{\weights}}.$
		To summarize, during each \gls{iteration}, EM minimizes an upper-bounding 
		surrogate \gls{objfunc} that is tight at the current iterate $\widehat{\weights}$. 
		Thus, EM is a \gls{majmin} method for approximately solving \eqref{equ_def_ML_EM_dict}. 
		The above construction and analysis of EM can be extended to more 
		general settings involving multiple \glspl{datapoint} 
		and infinite \glspl{featurespace} such as $\mathbb{R}^{\featuredim}$ 
		(see \cite[Sec. 11.4.7]{Murphy2012} for further details). 
			\\
		See also: \gls{maxlikelihood}, \gls{optproblem}, \gls{probmodel}. },
	first={expectation–maximization (EM)},
	type=math, 
	text={EM}
}

\newglossaryentry{ppca}
{name={probabilistic principal component analysis (PPCA)}, 
	description={PPCA\index{probabilistic principal component analysis (PPCA)} 
		extends\linebreak basic \gls{pca} by using a \gls{probmodel} for \glspl{datapoint}. 
		Using a \gls{probmodel} allows us to cast \gls{dimred} as an estimation problem 
		that can be solved using \gls{em} \cite{TippingProbPCA}.
				\\
		See also: \gls{pca}, \gls{probmodel}, \gls{dimred}, \gls{em}.},
	first={probabilistic principal component analysis (PPCA)},
	type=math, 
	text={PPCA}
}

\newglossaryentry{contractop}
{name={contractive operator},
	description={An\index{contractive operator} \gls{operator} 
		$\fixedpointop: \mathcal{V} \rightarrow  \mathcal{V}$ on a \gls{normedspace} 
		$(\mathcal{V}, \normgeneric{\,\cdot \,}{})$
		is a contraction (or contractive) if, for some $\contractfac \in [0,1)$, 
 	   	\cite{BausckeCombette}, \cite{ProximalMethods}
		\begin{equation} 
			\nonumber
			\normgeneric{ \fixedpointop \weights\!-\!\fixedpointop \weights'}{}  \leq 
			 \contractfac	\normgeneric{\weights\!-\!\weights'}{} \quad \mbox{holds for any } \weights,\weights' \in \mathcal{V}.
		\end{equation} 
		The notion of a contractive \gls{operator} generalizes naturally from \glspl{normedspace} 
		to arbitrary \glspl{metricspace} \cite{BausckeCombette}.
		\begin{figure}[H]
    			\centering
    			\begin{tikzpicture}[>=Latex, font=\small]
        			\tikzset{
            		space/.style={draw, thick, circle, minimum size=4.0cm},
            		pt/.style={circle, inner sep=1.5pt, draw, fill=black},
            		maparrow/.style={->, very thick},
            		distline/.style={dashed, thick}
        			}
        			\node[space,label=above:{$\featurespace$}] (X) at (0,0) {};
        			\coordinate (x1)  at (-1.0,0.8);
        			\coordinate (x2)  at ( 1.0,0.8);
        			\coordinate (Tx1) at (-0.5,0.1);
        			\coordinate (Tx2) at ( 0.5,0.1);
        			\node[pt,label=above left:{$\featurevec$}] at (x1) {};
        			\node[pt,label=above right:{$\featurevec'$}] at (x2) {};
        			\node[pt] (TX1) at (Tx1) {};
        			\node[pt] (TX2) at (Tx2) {};
        			\node[anchor=east] at ($(TX1.west)+(-4pt,-2pt)$) {$\fixedpointop \featurevec$};
        			\node[anchor=west] at ($(TX2.east)+(4pt,-2pt)$) {$\fixedpointop \featurevec'$};
        			\draw[distline] (x1) -- (x2)
            		node[midway, above=4pt] {$\metric{\featurevec}{\featurevec'}$};
        			\draw[distline] (Tx1) -- (Tx2)
            		node[midway, below=4pt] {$\metric{\fixedpointop \featurevec}{\fixedpointop \featurevec'}$};
        			\coordinate (xs) at (0,-0.9);
        			\node[pt,label=below:{$\featurevec^\star$}] (XS) at (xs) {};
        			\node at (0,-1.6) {$\fixedpointop \featurevec^\star = \featurevec^\star$};
    			\end{tikzpicture}
    		\caption{Contractive \gls{operator} $\fixedpointop:\featurespace\to\featurespace$ 
      			with a unique \gls{fixedpoint} $\featurevec^\star$ satisfying $\fixedpointop \featurevec^\star=\featurevec^\star$.
      			For any two points $\featurevec,\featurevec'$ in the same space, the \gls{distance} between their images
      			$\fixedpointop\featurevec$ and $\fixedpointop\featurevec'$ is strictly smaller.}
    			\label{fig:contract_dict}
		\end{figure}
		Intuitively, a contractive \gls{operator} brings any two points 
		from its \gls{domain} closer together by at least a factor of $\contractfac$.
		\\ 
		See also: \gls{operator}, \gls{metricspace}, \gls{banachfixedpoint}, \gls{bellmanoperator}. },
	first={contractive operator},
	text={contractive operator}, 
	type=math,
	firstplural={contractive operators}, 
	plural={contractive operators}
}

\newglossaryentry{pseudocontractop}
{name={pseudocontractive operator},
	description={An\index{pseudocontractive operator} \gls{operator}
		$\fixedpointop:\hilbertspace\rightarrow\hilbertspace$ on a \gls{hilbertspace}
		$\hilbertspace$ is called pseudocontractive if \cite{BausckeCombette}
		\begin{equation}
			\nonumber
			\normgeneric{\fixedpointop\weights-\fixedpointop\weights'}{}^{2}
			\le
			\normgeneric{\weights-\weights'}{}^{2}
			+
			\normgeneric{(\identityop-\fixedpointop)\weights
			-(\identityop-\fixedpointop)\weights'}{}^{2}
		\end{equation}
		holds for all $\weights,\weights'\in\hilbertspace$, where
		$\identityop$ denotes the identity \gls{operator}.
		This condition is equivalent to the monotonicity of the \gls{operator}
		$\identityop-\fixedpointop$ \cite[Example 20.8]{BausckeCombette}.
		Pseudocontractive \glspl{operator} include \glspl{nonexpansiveop} and \glspl{contractop} as 
		special cases. 
		\medskip
		\noindent \\
		Example: Let $\hilbertspace=\mathbb{R}$ and define
		\[
		\fixedpointop(\weight)=\weight-\phi(\weight),
		\]
		where $\phi:\mathbb{R}\to\mathbb{R}$ is any nondecreasing \gls{function}.
		Then, $\identityop-\fixedpointop=\phi$ is monotone, and hence
		$\fixedpointop$ is pseudocontractive \cite{BausckeCombette}.
		For instance, choosing $\phi(\weight)=\max\{0,\weight\}$ yields a pseudocontractive
		\gls{operator} that is neither contractive nor non-expansive.
		\medskip
		\noindent \\
		Alternative meaning \cite{Feyzmahdavian2023AsynchJMLR}, \cite{ParallelDistrBook}:
		In the literature on asynchronous and parallel \gls{optimization},
		an \gls{operator} $\fixedpointop$ with a \gls{fixedpoint} $\weights^\star$ is called
		pseudocontractive if there exists $\contractfac\in(0,1)$ such that \cite{Feyzmahdavian2023AsynchJMLR}, \cite{ParallelDistrBook}
		\begin{equation}
			\label{equ_pseudocontractive_Bertsekas_dict}
			\normgeneric{\fixedpointop\weights-\weights^\star}{}
			\le
			\contractfac\,\normgeneric{\weights-\weights^\star}{}
			\quad\text{for all }\weights\in\hilbertspace.
		\end{equation}
		This notion expresses contraction with respect to the \gls{fixedpoint} and
		implies linear \gls{convergence} of the corresponding \gls{fixedpointiter}
		$\weights^{(\iteridx+1)}=\fixedpointop\weights^{(\iteridx)}$ \cite[Proposition 1.2]{ParallelDistrBook}.
		\medskip
		\noindent \\
		Example: Consider the \gls{operator} $\fixedpointop:\mathbb{R}\to\mathbb{R}$ defined by
		\[
			\fixedpointop(\weight)=
			\begin{cases}
			\frac12 \weight, & \weight\neq 1,\\
			0.9, & \weight=1.
			\end{cases}
		\]
		The unique \gls{fixedpoint} is $\weight^\star=0$, and
		$\lvert\fixedpointop(\weight)-\weight^\star\rvert\le
		0.9\,\lvert \weight-\weight^\star\rvert$ holds for all $\weight \in\mathbb{R}$,
		so $\fixedpointop$ is pseudocontractive in the sense of \eqref{equ_pseudocontractive_Bertsekas_dict}.
		However, $\fixedpointop$ fails to be a \gls{nonexpansiveop}, since arbitrarily small
		perturbations around $\weight=1$ can produce large changes in $\fixedpointop(\weight)$.
		\\
		See also: \gls{operator}, \gls{hilbertspace}, \gls{fixedpoint}.},
	first={pseudocontractive operator},
	text={pseudocontractive operator},
	type=math,
	firstplural={pseudocontractive operators},
	plural={pseudocontractive operators}
}

\newglossaryentry{proxalg}
{name={proximal algorithm},
	description={A\index{proximal algorithm} proximal \gls{algorithm} refers to an \gls{optmethod} 
		that uses \glspl{proxop} as the main computational primitive \cite{ProximalMethods}. 
		Proximal \glspl{algorithm} generalize \glspl{gdmethod} by replacing 
		\glspl{gradstep} with the application of \glspl{proxop} induced 
		by the \gls{objfunc} or some approximation. 
		It is interesting to note that the \gls{gradstep} for a \gls{smooth} 
		and \gls{convex} \gls{function} coincides with the \gls{proxop} 
		of a local linear approximation. Proximal \glspl{algorithm} can 
		be applied to a wide range of \glspl{optproblem}, including those 
		with a \gls{nonsmooth} \gls{objfunc} \cite{ProximalMethods}.
		\\
		See also: \gls{operator}, \gls{hilbertspace}, \gls{fixedpoint}.},
	first={proximal algorithm},
	text={proximal algorithm},
	type=math,
	firstplural={proximal algorithms},
	plural={proximal algorithms}
}

\newglossaryentry{proxop}
{name={proximal operator},
	description={Given\index{proximal operator} a \gls{convex} and \gls{continuous}
		\gls{function} $f(\weights')$, we define its proximal \gls{operator} as \cite{ProximalMethods}, \cite{Bauschke:2017}  
		$$\proximityop{f}{\weights}\defeq \argmin_{\weights' \in \mathbb{R}^{\dimlocalmodel}} \bigg[ f(\weights')\!+\!(1/2) \normgeneric{\weights- \weights'}{2}^{2}\bigg].$$ 
		It is often necessary to evaluate the proximal \gls{operator} for a scaled version $\lrate f$, 
		with some factor $\lrate > 0$, 
		$$\proximityop{\lrate f}{\weights} \defeq \argmin_{\weights' \in \mathbb{R}^{\dimlocalmodel}} 
		\bigg[ f(\weights')\!+\!\frac{1}{2\lrate} \normgeneric{\weights- \weights'}{2}^{2}\bigg].$$ 
		The factor $\lrate$ controls the relative importance of the \gls{function} $f$ 
		and the \gls{penaltyterm} in the \gls{optproblem} defining the proximal 
		\gls{operator}. From an \gls{algorithm} design perspective, the role of the 
		scaling factor $\lrate$ is similar to the \gls{learnrate} in \glspl{gdmethod}. 
		As illustrated in Fig.\ \ref{fig_proxoperator_opt_dict}, evaluating the proximal \gls{operator} 
		amounts to minimizing a penalized variant of $f$. The \gls{penaltyterm} is the 
		scaled squared \gls{eucliddist} to a given \gls{vector} $\weights$ (which is the input to the proximal \gls{operator}). 
		The proximal \gls{operator} can be interpreted as a \gls{generalization} of the \gls{gradstep}, which is defined 
		for a \gls{smooth} \gls{convex} \gls{function} $f(\weights')$. Indeed, taking a 
		\gls{gradstep} with \gls{stepsize} $\lrate$ at the current \gls{vector} $\weights$ 
		is the same as applying the (scaled, by $\lrate$) proximal \gls{operator} of the \gls{function} 
		$\tilde{f}(\weights')= \big( \nabla f(\weights)\big)\,^{T} (\weights'-\weights)$. 
			\begin{figure}[H]
			\begin{center}
				\begin{tikzpicture}[scale=0.8]
					\draw[blue, ultra thick, domain=-4.1:4.1] plot (\x, {(1/4)*\x*\x}) node[above right] {$f(\weights')$};		
					\draw[red, thick, domain=1:3] plot (\x, {2*(\x - 2)*(\x - 2)}) node[below right] {$\frac{1}{2\lrate}\normgeneric{\weights-\weights'}{2}^{2}$};
					\draw[shift={(2,0)}] (0pt,2pt) -- (0pt,-2pt) node[below] {$\weights$};
				\end{tikzpicture}
			\end{center}
			\caption{Proximal \gls{operator} updating a \gls{vector} $\weights$ by minimizing a penalized version 
				of the \gls{function} $f$. The \gls{penaltyterm} is the scaled squared \gls{eucliddist} between the \gls{optimization} 
				variable $\weights'$ and the given \gls{vector} $\weights$.	\label{fig_proxoperator_opt_dict}}
		\end{figure}
		See also: \gls{convex}, \gls{function}, \gls{gradstep}.},
	first={proximal operator},
	type=math,
	firstplural={proximal operators},
	plural={proximal operators},
	text={proximal operator}
}

\newglossaryentry{connected}
{name={connected}, 
	description={An\index{connected} 
		\gls{undirectedgraph} $\graph=\pair{\nodes}{\edges}$ is connected if, for every 
		non-empty subset $\nodes' \subset \nodes$, we can find at least one edge 
		connecting a node in $\nodes'$ with some node in $\nodes \setminus \nodes'$.
		We illustrate two examples of \glspl{undirectedgraph} in Fig.\ \ref{fig_connected_dict}.
		\begin{figure}[H]
			\centering
			\begin{tikzpicture}
			\node[circle, fill=black, inner sep=1.5pt, label=above:{1}] (A1) at (0, 1.5) {};
			\node[above=0.5cm of A1, align=center] {disconnected};
			\node[circle, fill=black, inner sep=1.5pt, label=below right:{2}] (B1) [below right=0.8cm and 0.5cm of A1] {};
			\node[circle, fill=black, inner sep=1.5pt, label=below left:{3}] (C1) [below left=0.8cm and 0.5cm of A1] {};
			\draw [line width=1 pt]  (A1) -- (B1);
			\node at (0, -1) {(a)};
			\begin{scope}[xshift=3.5cm]
				\node[circle, fill=black, inner sep=1.5pt, label=above:{1}] (A2) at (0, 1.5) {};
				\node[above=0.5cm of A2, align=center] {connected};
				\node[circle, fill=black, inner sep=1.5pt, label=below right:{2}] (B2) [below right=0.8cm and 0.5cm of A2] {};
				\node[circle, fill=black, inner sep=1.5pt, label=below left:{3}] (C2) [below left=0.8cm and 0.5cm of A2] {};
				\draw [line width=1 pt]  (A2) -- (B2);
				\draw [line width=1 pt]  (B2) -- (C2);
				\node at (0, -1) {(b)};
			\end{scope}
			\end{tikzpicture}
		\caption{(a) Disconnected \gls{graph}. (b) Connected \gls{graph}. \label{fig_connected_dict}}
		\end{figure} 
		See also: \gls{undirectedgraph}, \gls{algconn}.}, 
	type=math, 
	first={connected},
	text={connected}
}

\newglossaryentry{GaussProc}
{name={Gaussian process (GP)},
	description={A\index{Gaussian process (GP)} GP is a collection of \glspl{rv} 
  		$\{f(\featurevec)\}_{\featurevec \in \featurespace}$ indexed by input values $\featurevec$ 
  		from some input space $\featurespace$ such that, for any finite subset 
  		$\featurevec^{(1)}, \,\ldots, \,\featurevec^{(\samplesize)} \in \featurespace$, 
  		the corresponding \glspl{rv} $f(\featurevec^{(1)}, \,\ldots, \,\featurevec^{(\samplesize)})$ 
		have a joint \gls{mvndist} 
  		\[
  		f \left( \featurevec^{(1)}, \,\ldots, \,\featurevec^{(\samplesize)} \right) \sim \mathcal{N}(\boldsymbol{\mu}, \mathbf{K}).
  		\]
  		For a fixed input space $\featurespace$, a GP is 
		fully specified (or parameterized) by: 1) a \gls{mean} \gls{function} 
		$\mu(\featurevec) = \expect\{ f(\featurevec)\}$; and 2) a \gls{covariancefunction}
		$\kernelmap{\featurevec}{\featurevec'}= \expect\{ \big(f(\featurevec)-\mu(\featurevec)\big) \big(f(\featurevec')-\mu(\featurevec')\big) \big\}$.\\
  		Example: We can interpret the temperature distribution across Finland (at a specific 
  		point in time) as the \gls{realization} of a GP $f(\featurevec)$, where each input 
		$\featurevec = (\text{lat}, \text{lon})$ denotes a geographic location. Temperature 
		observations from \gls{fmi} weather stations provide values $f(\featurevec)$ at specific 
		locations (see Fig.\ \ref{fig_gp_FMI_dict}). A GP allows us to predict the temperature 
		nearby \gls{fmi} weather stations and to quantify the \gls{uncertainty} 
  		of these \glspl{prediction}. 
  		\begin{figure}[H]
  		\begin{center}
  			\begin{tikzpicture}
			\begin{axis}[
			axis equal,
			hide axis,
			scale=1.2,
			xmin=17, xmax=32,
			ymin=55, ymax=71,
			clip=true
			]
			\addplot[
			color=black,
			thick
			] table [x=lon, y=lat, col sep=comma] {assets/finland_border.csv};
			\addplot[
			only marks,
			mark=*,
			mark options={fill=blue},
			color=black
			] table [x=lon, y=lat, col sep=comma] {assets/fmi_stations_subset.csv};
			\draw[->, thick] (axis cs:19,59) -- (axis cs:25.5,59) node[anchor=west] {longitude (lon)};
			\draw[->, thick] (axis cs:19,59) -- (axis cs:19,65.5) node[anchor=south] {latitude (lat)};
			\end{axis}
			\end{tikzpicture}
			\vspace*{-5mm}
		\end{center}
		\caption{For a given point in time, we can interpret the current temperature distribution 
		over Finland as a \gls{realization} of a GP indexed by geographic coordinates and 
		sampled at \gls{fmi} weather stations. The weather stations are indicated by blue dots. \label{fig_gp_FMI_dict}}
		\end{figure}
		See also: \gls{mvndist}, \gls{uncertainty}, \gls{gaussrv}.}, 
  	first={Gaussian process (GP)},
	plural={GPs},
	firstplural={Gaussian processes (GPs)},
  	type=math, 
  	text={GP}
}

\newglossaryentry{normaldist} 
{name={normal distribution}, 
	description={See\index{normal distribution} \gls{gaussrv}, \gls{mvndist}, \gls{clt}.},
 	first={normal distribution}, 
 	firstplural={normal distributions},
 	plural={normal distributions},
	type=math, 
 	text={normal distribution}
}

\newglossaryentry{gaussrv}
{name={Gaussian random variable (Gaussian RV)}, 
	description={A\index{Gaussian random variable (Gaussian RV)} standard \gls{gaussian} \gls{rv} is a 
		real-valued \gls{rv} $\feature$ with a \gls{pdf} \cite{BertsekasProb}, \cite{GrayProbBook}, \cite{papoulis}
		\begin{equation}
			\nonumber
			\pdf{\feature}{\eta} = \frac{1}{\sqrt{2\pi}} \exp\,(-\eta^2/2). 
		\end{equation}
		Given a standard \gls{gaussian} \gls{rv} $\feature$, we can construct a general \gls{gaussian} 
		\gls{rv} $\feature'$ with \gls{mean} $\mu$ and \gls{variance} $\sigma^2$ via 
		$\feature' \defeq \sigma \feature + \mu$. 
		The \gls{probdist} of a \gls{gaussian} \gls{rv} is referred to as \gls{normaldist}, 
		denoted by $\mvnormal{\mu}{\sigma^{2}}$. 
		\\ 
		A \gls{gaussian} random \gls{vector} $\featurevec \in \mathbb{R}^{\featuredim}$ with 
		\gls{covmtx} $\mathbf{C}$ and \gls{mean} ${\bm \mu}$ can be constructed as \cite{GrayProbBook}, \cite{papoulis}, \cite{Lapidoth09}
		\[
		\featurevec \defeq \mathbf{A} \vz + {\bm \mu},
		\]
		where $\vz \defeq \big( z_{1}, \,\ldots, \,z_{\featuredim} \big)\,^{T}$ is a \gls{vector} 
		of \gls{iid} standard \gls{gaussian} \glspl{rv}, and $\mA \in \mathbb{R}^{\featuredim \times \featuredim}$ is any \gls{matrix} satisfying $\mA \mA\,^{T} = \mC$. 
		The \gls{probdist} of a \gls{gaussian} random \gls{vector} is referred to as the \gls{mvndist}, 
		denoted by $\mvnormal{\bm \mu}{\mathbf{C}}$.
		\\
		We can interpret a \gls{gaussian} random \gls{vector} $\featurevec=\big(\feature_{1},\,\ldots,\,\feature_{\nrfeatures}\big)^{T}$ as a \gls{stochproc} 
		indexed by the set $\mathcal{I}=\{1,\,\ldots,\,\nrfeatures\}$. A \gls{GaussProc} is a 
		\gls{stochproc} over an arbitrary index set $\mathcal{I}$ such that any restriction 
		to a finite subset $\mathcal{I}' \subseteq \mathcal{I}$ yields a \gls{gaussian} 
		random \gls{vector} \cite{Rasmussen2006Gaussian}.
  		\\
        		\Gls{gaussian} \glspl{rv} are widely used \glspl{probmodel} in the 
		statistical analysis of \gls{ml} methods. Their significance arises 
		partly from the \gls{clt}, which provides conditions under which 
		the average of many independent \glspl{rv} (not necessarily \gls{gaussian} themselves) 
		tends toward a \gls{gaussian} \gls{rv} \cite{ross2013first}.
		\\ 
		The \gls{mvndist} is also distinct in that it represents \gls{maximum} \gls{uncertainty}. 
		Among all \gls{vector}-valued \glspl{rv} with a given \gls{covmtx} $\mC$, the 
		\gls{rv} $\vx \sim \mvnormal{\bm \mu}{\mathbf{C}}$ maximizes \gls{diffentropy} 
		\cite[Th. 8.6.5]{coverthomas}. This makes \glspl{GaussProc} a natural choice for 
		capturing \gls{uncertainty} or the lack of (domain) knowledge.
		\\ 
		See also: \gls{mvndist}, \gls{GaussProc}, \gls{probmodel}, \gls{clt}, \gls{diffentropy}.},
	first={Gaussian random variable (Gaussian RV)},
	type=math, 
	plural={Gaussian RVs}, 
	firstplural={Gaussian random variables (Gaussian RVs)}, 
	text={Gaussian RV}
}

\newglossaryentry{gaussian} 
{name={Gaussian}, 
	description={See\index{Gaussian} \gls{gaussrv}.},
 	first={Gaussian},
 	type=math,
 	text={Gaussian},
	firstplural={Gaussians},  
 	plural={Gaussians}
}

\newglossaryentry{clt}
{name={central limit theorem (CLT)},
	description={Consider a \gls{sequence} of \gls{iid} \glspl{rv} \( \feature^{(\sampleidx)} \), for \( \sampleidx = 1, \,2, \,\ldots \), 
		each with \gls{mean} zero and finite \gls{variance} \( \sigma^2 > 0 \). 
		The\index{central limit theorem (CLT)} CLT states that the normalized sum 
		\[
		s^{(\samplesize)} \defeq \frac{1}{\sqrt{\samplesize}} \sum_{\sampleidx = 1}^{\samplesize} \feature^{(\sampleidx)} 
		\]
		converges in distribution to a \gls{gaussrv} with \gls{mean} zero and \gls{variance} \( \sigma^2 \) as \( \samplesize \to \infty \) \cite[Proposition~2.17]{AsympVanderVaartBook}.
		One elegant way to derive the CLT is via the \gls{characteristicfunc} of the normalized sum \( s^{(\samplesize)} \). 
		Let $ \phi(t) = \expect \big\{ \exp \big( j t \feature \big) \big\}$ (with the imaginary unit $j = \sqrt{-1}$) 
		be the common \gls{characteristicfunc} of each sum and $\feature^{(\sampleidx)}$, and let \( \phi^{(\samplesize)}(t) \) 
		denote the \gls{characteristicfunc} of \( s^{(\samplesize)} \). Define an \gls{operator} \( \mathcal{T} \) acting on \glspl{characteristicfunc} 
		such that
		\[
		\phi^{(\samplesize)}(t) = \mathcal{T}(\phi^{(\samplesize-1)})(t) \defeq \phi\left( \frac{t}{\sqrt{\samplesize}} \right) \cdot \phi^{(\samplesize-1)}\left( \frac{\sqrt{\samplesize-1}}{\sqrt{\samplesize}} t \right).
		\]
		This \gls{fixedpointiter} captures the effect of recursively adding an \gls{iid} \gls{rv} $\featurevec^{(\samplesize)}$ 
		and rescaling. Iteratively applying \( \mathcal{T} \) leads to \gls{convergence} of \( \phi^{(\samplesize)}(t) \) toward the \gls{fixedpoint}
		\[
		\phi^*(t) = \exp\,(-t^2 \sigma^2 / 2),
		\]
		which is the \gls{characteristicfunc} of a \gls{gaussrv} with \gls{mean} zero and \gls{variance} 
		\( \sigma^2 \). \Glspl{generalization} of the CLT allow for dependent or nonidentically distributed \glspl{rv} \cite[Sec.~2.8]{AsympVanderVaartBook}.
		\begin{figure}[H]
			\centering
			\begin{tikzpicture}
			\begin{axis}[
			width=10cm,
			height=6cm,
			xlabel={},
			ylabel={},
			legend style={at={(0.97,0.97)}, anchor=north west},
			domain=-3:3,
			ylabel style={
			yshift=10pt   
			},
			samples=400,
			ymin=-0.2, ymax=1.1,
			axis lines=middle,
			clip=false,
			grid=both,
			]
			\addplot[thick, blue,domain=-2:2] {cos(x/sqrt(1) r)^1};
			\addlegendentry{$m=1$}
			\addplot[thick, red] {cos(x/sqrt(2) r)^2};
			\addlegendentry{$m=2$}
			\addplot[thick, green!60!black] {cos(x/sqrt(3) r)^3};
			\addlegendentry{$m=3$}
			\addplot[thick, dashed, black] {exp(-x^2/2)};
			\addlegendentry{$\exp\,(-t^2/2)$}
			\node[anchor=south, rotate=0] at (axis cs:-0.08,1.05) {$\phi^{(m)}(t)$};
			\node[anchor=north, rotate=0] at (axis cs: 3.2,0.1) {$t$};
			\end{axis}
			\end{tikzpicture}
			\caption{\Glspl{characteristicfunc} of normalized sums of \gls{iid} \glspl{rv} $x^{(\sampleidx)} \in \{-1,1\}$ 
			for $\sampleidx=1,\,\ldots,\,\samplesize$ compared to the \gls{gaussian} limit.}
		\end{figure}
		See also: \gls{rv}, \gls{gaussrv}.},
	first={central limit theorem (CLT)},
	type=math, 
	text={CLT}
}

\newglossaryentry{indicatorfunc}
{name={indicator function}, 
	description={Consider some process\index{indicator function} that can result in different possible 
		\glspl{outcome} (e.g., survival or death of a patient). Such 
		a process can be modeled as a \gls{randomexperiment} with 
		\gls{samplespace} $\samplespace$ containing all possible \glspl{outcome}. 
		The indicator \gls{function} $\indicatorfunc{\mathcal{A}}$ of an \gls{event} 
		$\mathcal{A} \subseteq \samplespace$ is a \gls{function} defined as \cite{BertsekasProb} 
		\[
			\indicatorfunc{\mathcal{A}}(\outcome) = 
			\begin{cases}
			1 \quad \text{if } \outcome \in \mathcal{A}, \\
			0 \quad \text{if } \outcome \notin \mathcal{A}.
			\end{cases}
		\]
		The notion of an indicator \gls{function} is not limited to \glspl{outcome} of 
		a \gls{randomexperiment}. Ultimately, it is just a principled way to 
		represent a set $\mathcal{A}$ by a \gls{function} $\indicatorfunc{\mathcal{A}}$ \cite{BoydConvexBook}.
		For a \gls{probspace} $(\samplespace, \sigmaalgebra, \prob{\cdot})$ and $\mathcal{A} \in \sigmaalgebra$, 
		the indicator \gls{function} $\indicatorfunc{\mathcal{A}}$ is an \gls{rv} and it holds that 
		$\expect\{\indicatorfunc{\mathcal{A}}\} = \prob{\mathcal{A}}$.
		\\
		See also: \gls{outcome}, \gls{randomexperiment}, \gls{function}. }, 
	first={indicator function},
	plural={indicator functions}, 
	type=math, 
	text={indicator function}
}

\newglossaryentry{probmodel}
{name={probabilistic model}, 
	description={A probabilistic \gls{model}\index{probabilistic model} for the 
		generation of \glspl{datapoint} consists of \glspl{rv} 
		with a joint \gls{probdist} \cite{BertsekasProb}. 
		This joint \gls{probdist} typically involves \glspl{parameter} (or \glspl{modelparam}) 
		that are either chosen manually or learned via statistical \gls{inference} 
		methods such as \gls{maxlikelihood} estimation \cite{LC}.
					\\ 
		See also: \gls{model}, \gls{datapoint}, \gls{rv}, \gls{probdist}, \gls{parameter}, \gls{maxlikelihood}, \gls{realization}. }, 
	first={probabilistic model}, 
	plural={probabilistic models},
	firstplural={probabilistic models},
	type=math,
	text={probabilistic model} 
}

\newglossaryentry{LapMat}
{name={Laplacian matrix},
	description={The\index{Laplacian matrix} structure of a \gls{graph} $\graph$, with 
		nodes $\nodeidx=1, \,\ldots, \,\nrnodes$, can be analyzed using the properties of 
		special \glspl{matrix} that are associated with $\graph$. One such \gls{matrix} is the 
		\gls{graph} Laplacian \gls{matrix} $\mL^{(\graph)} \in \mathbb{R}^{\nrnodes \times \nrnodes}$, 
		which is defined for an undirected and weighted \gls{graph} \cite{Luxburg2007}, \cite{Ng2001}. 
		It is defined elementwise as (see Fig.\ \ref{fig_lap_mtx_dict})
		\begin{equation}
			\nonumber
			\LapMatEntry{\graph}{\nodeidx}{\nodeidx'} \defeq \begin{cases} - \edgeweight_{\nodeidx,\nodeidx'}, & \mbox{ for } \nodeidx\neq \nodeidx', \edge{\nodeidx}{\nodeidx'}\!\in\!\edges, \\ 
			\sum\limits_{\nodeidx'' \neq \nodeidx} \edgeweight_{\nodeidx,\nodeidx''}, & \mbox{ for } \nodeidx = \nodeidx', \\ 
							0, & \mbox{ else.} \end{cases}
	 	\end{equation}
  		Here, $\edgeweight_{\nodeidx,\nodeidx'}$ denotes the \gls{edgeweight} of an edge $\edge{\nodeidx}{\nodeidx'} \in \edges$. 
  		\begin{figure}[H]
  		\begin{center}
   		\begin{minipage}{0.45\textwidth}
		\begin{tikzpicture}
	 	 		\begin{scope}[every node/.style={circle, draw, minimum size=1cm}]
	 				\node (1) at (0,0) {1};
	 				\node (2) [below left=of 1] {2};
	 				\node (3) [below right=of 1] {3};
	 				\draw (1) -- (2);
	 				\draw (1) -- (3);
	 			\end{scope}
				\node at (0,-3) {(a)};
	 	\end{tikzpicture}
	 	\end{minipage} 
	 	\hspace*{-15mm}
 		\begin{minipage}{0.45\textwidth}
	 			 \begin{equation} 
	 				 \LapMat{\graph} = \begin{pmatrix} 2 & -1& -1 \\ -1& 1 & 0 \\  -1 & 0 & 1 \end{pmatrix}  
	 				 \nonumber
	 			\end{equation} 
				\begin{minipage}{\textwidth}
				\vspace{3.5ex}
				\centering
				{\selectfont (b)}
				\end{minipage}
	 	\end{minipage}
	 	\caption{\label{fig_lap_mtx_dict} (a) \Gls{undirectedgraph} $\graph$ with three nodes $\nodeidx=1,\,2,\,3$. 
	 		 (b) Laplacian \gls{matrix} $\LapMat{\graph}  \in \mathbb{R}^{3 \times 3}$ of $\graph$.} 
	 	\end{center}
	 	\end{figure}
		See also: \gls{graph}, \gls{matrix}, \gls{edgeweight}.},
	first={Laplacian matrix},
	type=math,
	plural={Laplacian matrices},
	firstplural={Laplacian matrices},
	text={Laplacian matrix}
}

\newglossaryentry{spectrum}
{name={spectrum},
	description={The spectrum\index{spectrum} of a \gls{linearoperator} $\fixedpointop$ 
  		is the set of values $\lambda$ for which the \gls{operator}
  		$\fixedpointop - \lambda \identityop$ fails to be invertible \cite{Muller2003}. 
  		The notion of a spectrum can then be naturally extended to any kind of 
		object for which an associated \gls{linearoperator} can be defined. 
		For example, the spectrum of a square \gls{matrix} $\mA \in \mathbb{R}^{\dimlocalmodel \times \dimlocalmodel}$ 
  		is obtained via the spectrum of the associated \gls{linearoperator} \cite{Horn91}.
  		As another example, we can define the spectrum of a \gls{stochproc} via a  
		\gls{linearoperator} constructed from its \gls{covariancefunction} \cite{Rasmussen2006Gaussian}.
		\\
		See also: \gls{linearoperator}, \gls{matrix}, \gls{stochproc}. },
	first={spectrum},
	type=math,
	plural={spectra},
	firstplural={spectra},
	text={spectrum}
}

\newglossaryentry{kernelmap}
{name={kernel (linear map)}, 
 	description={See\index{kernel (linear map)} \gls{nullspace}.},
 	first={kernel},
	type=math, 
	plural={kernels},
	firstplural={kernels},
	text={kernel} 
}

\newglossaryentry{kld}
{name={Kullback–Leibler divergence (KL divergence)}, 
 	description={The\index{Kullback–Leibler divergence (KL divergence)} KL divergence 
		is a quantitative \gls{measure} of how different a \gls{probdist} $\probdist^{(1)}$ is from another 
		\gls{probdist} $\probdist^{(2)}$, which is defined on the same \gls{probspace} \cite{coverthomas}. 
		The KL divergence $\kld{\probdist^{(1)}}{\probdist^{(2)}}$ 
		is well defined only if $\probdist^{(1)}$ is absolutely \gls{continuous} with respect to $\probdist^{(2)}$.\\
		Special case: If $\probdist^{(1)}$ 
		and $\probdist^{(2)}$ are described by the \glspl{pmf} $\pmf{x}{\cdot}$ and $\pmf{y}{\cdot}$ 
		on a \gls{samplespace} $\samplespace$, then 
		\[
			\kld{\pmf{x}{\cdot}}{\pmf{y}{\cdot}} \defeq \sum_{\omega \in \samplespace} 
			\pmf{x}{\omega} \log \frac{\pmf{x}{\omega}}{\pmf{y}{\omega}}.
		\]
		In this discrete setting, the KL divergence coincides with the 
		\gls{bregmandivergence} generated by the negative \gls{entropy} \gls{function} \cite[Sec. 4.3]{CvxBubeck2015}
		\[
			\phi(p) \defeq \sum_{\omega \in \samplespace} 
			\pmf{x}{\omega} \log \pmf{x}{\omega}.
		\]
		See also: \gls{probdist}, \gls{bregmandivergence}.},
 	first={Kullback–Leibler divergence (KL divergence)},
	type=math, 
	plural={KL divergences},
	firstplural={Kullback–Leibler divergences (KL divergences)},
	text={KL divergence} 
}

\newglossaryentry{mean}
{name={mean},
	description={The\index{mean} mean of an \gls{rv} $\featurevec$, which takes 
 		on values in a \gls{euclidspace} $\mathbb{R}^{\dimlocalmodel}$, is its 
 		\gls{expectation} $\expect\{\featurevec\}$. It is defined as the \gls{LebesgueIntegral}
 		of $\featurevec$ with respect to the underlying \gls{probdist} $\prob{\cdot}$ (e.g., 
		see \cite{RudinBookPrinciplesMatheAnalysis} or \cite{BillingsleyProbMeasure}), i.e.,
		\[
			\expect\{\featurevec\} = \int_{\mathbb{R}^{\dimlocalmodel}} \vx \, \mathrm{d}P(\vx).
		\]  
		We also use the term to refer to the \gls{samplemean} of a finite \gls{dataset}
		$\dataset = \left\{ \vx^{(1)}, \,\ldots, \,\vx^{(\samplesize)} \in \mathbb{R}^{\dimlocalmodel}\right\}$. 
		However, these two definitions are essentially the same. Indeed, we can use 
		a \gls{dataset} to construct a \gls{discreteRV} $\widetilde{\vx}^{(\dataset)}=\vx^{(I)}$ on 
		the \gls{samplespace} $\{1, \,\ldots, \,\samplesize\}$. Here, the index $I$ is 
		chosen uniformly at random, i.e., $\prob{I=\sampleidx}=1/\samplesize$ for all 
		$\sampleidx=1,\ldots,\samplesize$. The mean of $\widetilde{\vx}^{(\dataset)}$ is 
		precisely the average $({1}/{\samplesize}) \sum_{\sampleidx=1}^{\samplesize} \vx^{(\sampleidx)}$.
		For an \gls{rv} with a finite second-order moment, i.e., 
		$\expect\{ \normgeneric{\featurevec}{2}^{2} \}$ is well defined and finite, 
		the mean is characterized as the solution of the 
		following \gls{risk} minimization problem \cite{BertsekasProb}:
		\[
			\expect\{\featurevec\} = \argmin_{\vc \in \mathbb{R}^{\nrfeatures}} 
			\expect \big\{\normgeneric{\featurevec - \vc}{2}^{2}\big \}.
		\]
		For the \gls{rv} $\widetilde{\vx}^{(\dataset)}$, associated with a \gls{dataset} $\dataset$, 
		this \gls{optproblem} reduces to \gls{erm} with \gls{sqerrloss} on $\dataset$. 
		\\ 
		See also: \gls{rv}, \gls{expectation}, \gls{probdist}, \gls{erm}.}, 
	first={mean}, 
	type=math,
	plural={means},
	text={mean} 
}

\newglossaryentry{median}
{name={median},
	description={A\index{median} median $\med\,(x)$ of a real-valued \gls{rv} $x$ 
 		is any number $M \in \mathbb{R}$ such that $\prob{ x \leq M} \geq 1/2$ and 
		$\prob{ x \geq M} \geq 1/2$ 
		(see Fig.\ \ref{fig_median1_dict}) \cite{LC}. 
 		\begin{figure}[H]
			\begin{center}
			\begin{tikzpicture}
 			\begin{axis}[
    			axis lines=middle,
    			xlabel={},
    			ylabel={},
    			ymin=0, ymax=1.1,
    			xmin=-2, xmax=6,
    			xtick=\empty,
    			ytick={0,1/2,1},
    			domain=-2:6,
    			samples=200,
    			width=10cm,
    			height=6cm,
    			smooth,
    			enlargelimits=true,
    			clip=false
  			]
			\addplot[thick, blue, name path=cdf] {1/(1 + exp(-(x - 1)))} node[pos=0.5, above, yshift=15pt] {$\prob{x \leq \eta}$};    
    			\draw[dashed, gray] (axis cs:1,0) -- (axis cs:1,0.5); 
    			\draw[dashed, gray] (axis cs:-2,0.5) -- (axis cs:1,0.5); 
   			\filldraw[red] (axis cs:1,0.5) circle (2pt);
  		    	\node[below] at (axis cs:1,0) {$M$};
			\node[above right] at (axis cs:6.3,0) {$\eta$};
  			\end{axis}
			\end{tikzpicture} 
			\end{center}
		\caption{Median of a real-valued \gls{rv} defined as any number $M$ 
			that partitions $\mathbb{R}$ into two rays with equal \gls{probability}. \label{fig_median1_dict}}
 		\end{figure}  
 		We can define the median $\med\,(\dataset)$ 
 		of a \gls{dataset} $\dataset = \{ x^{(1)}, \,\ldots, \,x^{(\samplesize)} \in \mathbb{R} \}$ 
 		via a specific \gls{rv} $\tilde{x}^{(\dataset)}$ that is naturally associated with $\dataset$. 
 		In particular, this \gls{rv} is defined on the \gls{samplespace} $\{1, \,\ldots, \,\samplesize\}$ 
		via $\tilde{x}^{(\dataset)} \defeq x^{(I)}$. Here, the index $I$ is chosen uniformly 
		at random, i.e., $\prob{I = \sampleidx}=1/\samplesize$ for 
 		all $\sampleidx=1, \,\ldots, \,\samplesize$. If the \gls{rv} $x$ is \gls{integrable}, any 
		median of $x$ solves the following \gls{optproblem}:
 		$$\min_{x' \in \mathbb{R}} \expect{|x - x'|}.$$ 
		For the above \gls{rv} $\tilde{x}$ (constructed from a \gls{dataset} $\dataset$), 
		this \gls{optproblem} is \gls{erm} on $\dataset$ using \gls{abserr}. 
 		Like the \gls{mean}, the median of a \gls{dataset} $\dataset$ can also be used 
 		to estimate \glspl{parameter} of an underlying \gls{probmodel}. Compared 
 		with the \gls{mean}, the median is more robust to \glspl{outlier}. For example, 
 		the median of a \gls{dataset} $\dataset$ with more than one \gls{datapoint} does not 
 		change even if we arbitrarily increase the largest element of $\dataset$ (see Fig.\ \ref{fig_median2_dict}). 
		In contrast, the \gls{mean} will increase arbitrarily.
		\begin{figure}[H]
		\centering
		\begin{tikzpicture}[scale=0.7, y=0.5cm, x=0.5cm]
			\begin{scope}
				\foreach \x/\y in {
					1/2, 4/3, 7/4
				} {
					\draw[dashed, gray] (\x, 0) -- (\x, \y);
					\filldraw[blue] (\x, \y) circle (2pt);
					\node[circle, inner sep=0pt] (ptA\x) at (\x, \y) {};
				}
				\draw[dashed, thick] (0.5, 3) -- (10.5, 3) node[right] {$\med\,(\dataset)\!=\!{\rm mean}(\dataset)$};
				\node at (7.5, -4) {(a)};
			\end{scope}
			\begin{scope}[xshift=12cm]
				\foreach \x/\y in {
					1/2, 4/3, 7/10
				} {
					\draw[dashed, gray] (\x, 0) -- (\x, \y);
					\filldraw[blue] (\x, \y) circle (2pt);
					\node[circle, inner sep=0pt] (ptB\x) at (\x, \y) {};
				}
				\draw[dashed, thick] (0.5, 7.5) -- (10.5, 7.5) node[right] 
				{${\rm mean}\,\big(\widetilde{\dataset}\big)$};
				\draw[dashed, thick] (0.5, 3) -- (10.5, 3) node[right] 
				{$\med\,\big(\widetilde{\dataset}\big)$};
				\node[above right=2pt and 2pt, red] at (ptB7) {\gls{outlier}};
				\node at (7.5, -4) {(b)};
			\end{scope}
		\end{tikzpicture}
		\caption{Median \gls{robustness} to \gls{outlier} contamination. (a) Original \gls{dataset} $\dataset$. (b) Noisy 
			\gls{dataset} $\widetilde{\dataset}$ including an \gls{outlier}. \label{fig_median2_dict}}
		\end{figure}
		See also: \gls{mean}, \gls{outlier}, \gls{robustness}, \gls{ladregression}.}, 
	first={median}, 
	type=math,
	plural={medians},
	text={median} 
}

\newglossaryentry{variance}
{name={variance},
	description={The\index{variance} variance of a real-valued \gls{rv} $\feature$ is defined 
	   	as the \gls{expectation} $\expect\big\{ \big( x - \expect\{x \} \big)^{2} \big\}$ of 
	   	the squared difference between $\feature$ and its \gls{expectation} $\expect\{x \}$. 
	   	We extend this definition to \gls{vector}-valued \glspl{rv} $\featurevec$ 
	   	as $\expect\big\{ \big\| \featurevec - \expect\{\featurevec \} \big\|_{2}^{2} \big\} = \tr{\covmtx{\featurevec}}$,
          	i.e., the sum of the variances of each entry of $\featurevec$, which can be written compactly 
		as the \gls{trace} of the \gls{covmtx} $\covmtx{\featurevec}$ of $\featurevec$. 
          	\\ 	 
		See also: \gls{rv}, \gls{expectation}, \gls{vector}.},
	first={variance},
	type=math,
	text={variance} 
}

\newglossaryentry{probabilitysimplex}
{name={probability simplex},
	description={The\index{probability simplex} \gls{probability} simplex
   		$\simplex{\nrcluster-1}$ is the set of all \glspl{vector} 
   		in $\mathbb{R}^\nrcluster$ with nonnegative entries 
 		that sum to one \cite{BoydConvexBook}. Each element of $\simplex{\nrcluster-1}$ 
		represents a \gls{pmf} of an \gls{rv} $y \in \{1,\,\ldots,\,\nrcluster\}$.
		\\
		See also: \gls{probability}, \gls{vector}, \gls{pmf}, \gls{rv}. },
  	first={probability simplex},
  	plural={probability simplexes},
  	firstplural={probability simplexes},
  	type=math,
  	text={probability simplex}
}

\newglossaryentry{orthogonalitycondition}
{name={orthogonality condition},
	description={Consider\index{orthogonality condition} a linear \gls{subspace} 
		$\paramspace \subseteq \mathbb{R}^{\dimlocalmodel}$ and a \gls{vector} 
		$\weights \in \mathbb{R}^{\dimlocalmodel}$. A \gls{vector} $\weights^\star \in \paramspace$ 
		is the \gls{projection} $\projection{\paramspace}{\weights}$ if and only if
		\begin{equation}
			\nonumber
			\label{eq:orthogonality_condition_dict}
			\vv^{T} \big( \weights - \weights^\star\big) = 0 
			\quad \text{for all } \vv \in \paramspace .
		\end{equation}
		In other words, the \gls{projection} error $\weights-\weights^\star$ is orthogonal to the
		\gls{subspace} $\paramspace$. The notion of orthogonality can be extended to general 
            	\glspl{hilbertspace}, i.e., \glspl{vectorspace} with \gls{innerproduct}.
		\\
		See also: \gls{vector}, \gls{projection}, \gls{euclidspace}.},
	first={orthogonality condition},
	plural={orthogonality conditions},
	firstplural={orthogonality conditions},
	type=math,
	text={orthogonality condition}
}

\newglossaryentry{projection}
{name={projection}, 
       	description={Consider\index{projection} a bounded subset 
		$\paramspace \subseteq \mathbb{R}^{\dimlocalmodel}$ of 
		the $\dimlocalmodel$-dimensional \gls{euclidspace}. We define the projection 
		$\projection{\paramspace}{\weights}$ of a \gls{vector} 
		$\weights \in \mathbb{R}^{\dimlocalmodel}$ onto $\paramspace$ as
	   	\begin{equation} 
   	   		\nonumber
			\label{equ_def_proj_generic_dict}
  	    		\projection{\paramspace}{\weights} = \argmin_{\weights' \in \paramspace} \normgeneric{\weights - \weights'}{2}. 
        	    	\end{equation}
	    	In other words, $\projection{\paramspace}{\weights}$ is the \gls{vector} in $\paramspace$ 
	    	that is closest to $\weights$. The projection is only well defined for subsets $\paramspace$ 
	    	for which the above \gls{minimum} exists \cite{BoydConvexBook}.
		 			\\ 
	    	See also: \gls{euclidspace}, \gls{vector}, \gls{minimum}.},
	first={projection},
	plural={projections}, 
	firstplural={projections},
	type=math,
	text={projection}
}

\newglossaryentry{constrainedoptprob} 
{name={constrained optimization problem}, 
	description={A\index{constrained optimization problem} constrained \gls{optproblem} is an 
		\gls{optproblem} that involves constraints on the \gls{optimization} variable. 
		For example, consider the \gls{optproblem}
		\[
			\min_{\weights \in \paramspace} f(\weights),
		\]
		where the \gls{objfunc} $f: \mathbb{R}^{\dimlocalmodel} \to \mathbb{R}$ is to be minimized 
		over a constraint set $\paramspace \subseteq \mathbb{R}^{\dimlocalmodel}$. Here, the constraint 
		set $\paramspace$ restricts the feasible values of the \gls{optimization} variable $\weights$. 
		If $\paramspace = \mathbb{R}^{\dimlocalmodel}$, the above \gls{optproblem} reduces to an 
		unconstrained \gls{optproblem}. 
					\\ 
		See also: \gls{optproblem}.}, 
	first={constrained optimization problem}, 
	plural={constrained optimization problems},
	firstplural={constrained optimization problems},
	type=math,
	text={constrained optimization problem} 
}

\newglossaryentry{projgd}
{name={projected gradient descent (projected GD)},
	description={Consider an \gls{erm}-based method that uses a parameterized \gls{model} with  
		\gls{paramspace} $\paramspace \subseteq \mathbb{R}^{\dimlocalmodel}$. Even if 
		the \gls{objfunc} of \gls{erm} is \gls{smooth}, we cannot use basic \gls{gd}, as 
		it does not take into account constraints on the \gls{optimization} variable (i.e., the \glspl{modelparam}). 
		Projected\index{projected gradient descent (projected GD)} \gls{gd} 
		extends basic \gls{gd} to address this issue. 
		A single iteration of projected \gls{gd} consists of first taking a \gls{gradstep} 
		and then projecting the result back onto the \gls{paramspace}. 
		See Fig.\ \ref{fig_projected_GD_dict} for a visual illustration.
		\begin{figure}[H]
		\begin{center}
			\begin{tikzpicture}[scale=0.9]
			\node [right] at (-5.1,1.7) {$f(\weights)$} ;
			\draw[ultra thick, domain=-4.1:4.1] plot (\x,  {(1/8)*\x*\x});
			\draw [fill] (2.83,1) circle [radius=0.1] node[right] {$\weights$};
			\draw[line width =0.5mm,dashed,->] (2.83,1) -- node[midway,above] {\gls{gradstep}} (-1.5,1);
			\draw[line width =0.2mm,dashed] (-1.5,1) --(-1.5,-1.5)  node [below, left]{$\widehat{\weights}=\weights\!-\!\lrate \nabla f\big(\weights\big)$} ;
			\draw[line width =0.5mm,dashed,->] (-1.5,-1.5)  -- node[midway,above] {} (1,-1.5) ; 
			\draw [fill] (1,-1.5) circle [radius=0.1] node[below] {$\projection{\paramspace}{\widehat{\weights}}$};
			\draw[line width=1mm] (1,-1.5) -- (3,-1.5) node[midway, above] {$\paramspace$};
			\end{tikzpicture}
		\vspace*{-5mm}
		\end{center}
		\caption{Projected \gls{gd} augmenting a basic \gls{gradstep} with a \gls{projection} back 
			onto the constraint set $\paramspace$.}
			\label{fig_projected_GD_dict}
		\end{figure}
		See also: \gls{erm}, \gls{model}, \gls{paramspace}, \gls{objfunc}, \gls{smooth}, \gls{gd}, \gls{modelparam}, \gls{gradstep}, \gls{projection}.},
	first={projected gradient descent (projected GD)},
	type=math, 
	text={projected GD}
}

\newglossaryentry{proximable}
{name={proximable},
	description={A\index{proximable} 
		\gls{convex} \gls{function} for which the \gls{proxop} can be computed efficiently is 
		sometimes referred to as proximable or simple \cite{Condat2013}.
					\\ 
		See also: \gls{convex}, \gls{function}, \gls{proxop}.},
	first={proximable},
	type=math,
	text={proximable}
}

\newglossaryentry{jacobianmatrix}
{name={Jacobian matrix},
	description={The Jacobian \gls{matrix}\index{Jacobian matrix} of a \gls{vector}-valued \gls{function} 
		$$\vf:\mathbb{R}^{\nrfeatures}\rightarrow\mathbb{R}^{\nrfeatures'}: 
		\featurevec \mapsto \vf(\featurevec) = \big(f_{1}(\featurevec),\,\ldots,\,f_{\nrfeatures'}(\featurevec) \big)^{T}$$
		is the \gls{matrix} constituted by all first-order \glspl{partialderivative} $\vf$
		\cite{RudinBookPrinciplesMatheAnalysis}, \cite{HolderEichholz}. More explicitly, 
		\[
			D_{f}(\featurevec)
			\defeq
			\begin{bmatrix}
				\frac{\partial f_{1}}{\partial \feature_{1}}(\featurevec) & \cdots &
				\frac{\partial f_{1}}{\partial \feature_{\nrfeatures}}(\featurevec) \\
				\vdots & \ddots & \vdots \\
				\frac{\partial f_{\nrfeatures'}}{\partial \feature_{1}}(\featurevec) & \cdots &
				\frac{\partial f_{\nrfeatures'}}{\partial \feature_{\nrfeatures}}(\featurevec)
			\end{bmatrix}.
		\]
		The Jacobian \gls{matrix} represents the best linear approximation of $\vf$
		around a given argument \gls{vector} $\featurevec$. Note that the $i$th row of the Jacobian \gls{matrix} 
		coincides with the \gls{transpose} of the \gls{gradient} of one component \gls{function} 
		$f_i$ evaluated at $\featurevec$.
		\\
		See also: \gls{matrix}, \gls{function}, \gls{partialderivative}, \gls{gradient}.},
	first={Jacobian matrix},
	type=math,
	plural={Jacobian matrices},
	firstplural={Jacobian matrices},
	text={Jacobian matrix}
}

\newglossaryentry{linearoperator}
{name={linear operator},
sort={linearoperator},
	description={A \index{linear operator} linear \gls{operator} is an 
		\gls{operator} $\fixedpointop$ whose \gls{domain} and \gls{co-domain} 
		are \glspl{vectorspace} and which satisfies \cite{DunfordSchwartz1988}
		\begin{equation}
			\nonumber
			\fixedpointop(\alpha \weights + \beta \weights')
			= \alpha \fixedpointop(\weights) + \beta \fixedpointop(\weights')
		\end{equation}
		for all $\weights,\weights'$ in the \gls{domain} and all scalars $\alpha,\beta$.
							\\
		See also: \gls{operator}, \gls{vectorspace}, \gls{matrix}.},
	first={linear operator},
	type=math,
	plural={linear operators},
	firstplural={linear operators},
	text={linear operator}
}

\newglossaryentry{operator} 
{name={operator}, 
	description={An\index{operator} operator is a \gls{function} 
		whose \gls{domain} and \gls{co-domain} have a specific 
		mathematical structure such as a \gls{vectorspace}, a \gls{hilbertspace},
		or a \gls{metricspace} \cite{Bauschke:2017}, \cite{DunfordSchwartz1988}. 
		Many \gls{ml} methods involve operators whose \gls{domain} and \gls{co-domain} 
		are \glspl{euclidspace}.
							\\ 
		See also: \gls{function}, \gls{vectorspace}, \gls{hilbertspace}.},
	first={operator},
	type=math, 
	plural={operators},
	firstplural={operators},
	text={operator}
}

\newglossaryentry{ergraph}
{name={Erd\H{o}s–R\'enyi graph (ER graph)},
	description={An ER \gls{graph}\index{Erd\H{o}s–R\'enyi graph (ER graph)} \cite{erdds1959random}, \cite{gilbert1959random} 
            	is a \gls{probmodel} for \glspl{graph} defined over a given node set $\nodeidx=1, \,\ldots, \,\nrnodes$. 
            	One way to define the ER \gls{graph} is via the collection of \gls{iid} binary 
            	\glspl{rv} $b^{(\edge{\nodeidx}{\nodeidx'})} \in \{0,1\}$
		for each pair of different nodes $\nodeidx, \nodeidx'$. A specific \gls{realization}  
		of an ER \gls{graph} contains an edge $\edge{\nodeidx}{\nodeidx'}$ if and only if 
		$b^{(\edge{\nodeidx}{\nodeidx'})}=1$. The ER \gls{graph} is parameterized by the 
		number $\nrnodes$ of nodes and the \gls{probability} $\prob{b^{(\edge{\nodeidx}{\nodeidx'})}=1}$.	
		\\
		See also: \gls{graph}, \gls{probmodel}, \gls{iid}, \gls{rv}, \gls{realization}, \gls{probability}.},
	first={Erd\H{o}s–R\'enyi graph (ER graph)},
	type=math,
	text={ER graph}
}

\newglossaryentry{condprobdist}
{name={conditional probability distribution}, 
	description={Consider\index{conditional probability distribution} 
     		a \gls{stochproc} consisting of two \glspl{rv} $\featurevec$ and $\truelabel$ 
    		with \gls{probdist} $\probdist^{(\featurevec,\truelabel)}$. The conditional 
    		\gls{probdist} of $\truelabel$ given (or conditioned on) $\featurevec$ is 
    		denoted by $\probdist^{(\truelabel\mid\featurevec)}$. It is defined via the 
    		\glspl{conditionalexpect} of the \glspl{indicatorfunc} of  
    		\gls{measurable} sets in the \gls{sigmaalgebra} generated by 
    		the \gls{rv} $\truelabel$ \cite[Sec. 34]{BillingsleyProbMeasure}, \cite{KallenbergBook}.
		\\
		See also: \gls{probdist}, \gls{conditionalexpect}. },
  	first={conditional probability distribution}, 
  	plural={conditional probability distributions},
  	type=math, 
  	text={conditional probability distribution}
}

\newglossaryentry{linearmap}
{name={linear map},
sort={linearmap},
	description={A\index{linear map} linear \gls{map} 
		$f: \mathbb{R}^\nrfeatures \rightarrow \mathbb{R}^\samplesize$ 
	    	is a \gls{function} that satisfies additivity, i.e.,
		$f(\vx + \vy) = f(\vx) + f(\vy)$, and homogeneity, i.e.,
		$f(c\vx) = c f(\vx)$, for all \glspl{vector} $\vx, \vy \in \mathbb{R}^\nrfeatures$ 
		and scalars $c \in \mathbb{R}$. In particular, $f(\mathbf{0}) = \mathbf{0}$. 
		Any linear \gls{map} can be represented as a \gls{matrix} multiplication 
		$f(\vx) = \mA \vx$ for some \gls{matrix} $\mA \in \mathbb{R}^{m \times n}$. 
		The collection of real-valued linear \glspl{map} (where $\samplesize=1$), 
		for a given dimension $\nrfeatures$, constitutes a \gls{linmodel}. The notion 
		of a linear \gls{map} can be generalized from the \gls{domain} $\mathbb{R}^{\nrfeatures}$ 
		and \gls{co-domain} $\mathbb{R}^{\samplesize}$ to arbitrary \glspl{vectorspace}.
		\\
		See also: \gls{map}, \gls{function}, \gls{vector}, \gls{matrix}, \gls{linmodel}.},
	first={linear map},
	type=math, 
	plural={linear maps}, 
	firstplural={linear maps}, 
	text={linear map}
}

\newglossaryentry{vector}
{name={vector},
	description={A\index{vector} vector is an element of a \gls{vectorspace}. 
		In the context of \gls{ml}, a particularly important example of a \gls{vectorspace} 
		is the \gls{euclidspace} $\mathbb{R}^{\nrfeatures}$, where $\nrfeatures \in \mathbb{N}$ 
		is the (finite) \gls{dimension} of the space. A vector $\vx \in \mathbb{R}^{\nrfeatures}$ 
		can be represented as a list or 1-D array of real numbers, i.e., 
		$x_1, \,\ldots, \,x_{\nrfeatures}$ with $x_\featureidx \in \mathbb{R}$ for 
		$\featureidx = 1, \,\ldots, \,\nrfeatures$. The value $x_\featureidx$ is the $\featureidx$th 
		entry of the vector $\vx$. It can also be useful to view a vector $\vx \in \mathbb{R}^{\nrfeatures}$ 
		as a \gls{function} that maps each index $\featureidx \in \{1, \,\ldots, \,\nrfeatures\}$ 
		to a value $x_\featureidx \in \mathbb{R}$, i.e., $\vx: \featureidx \mapsto x_\featureidx$. 
		This perspective is particularly useful for the study of \glspl{kernelmethod}. See Fig.\ 
		\ref{fig:vector-function-dual_dict} for the two views of a vector.
		\begin{figure}[H]
			\begin{minipage}[c]{0.48\textwidth}
				\centering 
				2, --1, 3, 0, --2, 1
				\begin{minipage}{\textwidth}
				\vspace{5ex}
				\centering
				{\selectfont (a)}
				\end{minipage}
			\end{minipage}
			\hfill
			\begin{minipage}{0.48\textwidth}
			\centering
			\begin{tikzpicture}
			\begin{axis}[
    				width=6.5cm,
    				height=5cm,
    				title={},
    				xlabel={index $\featureidx$},
    				ylabel={$x_\featureidx$},
   		 		ymin=-3.5, ymax=3.5,
    				xmin=0.5, xmax=6.5,
   	 			xtick={1,2,3,4,5,6},
    				ytick={-3,-2,-1,0,1,2,3},
    				axis x line=bottom,        
    				axis y line=left,          
    				grid=both,
    				major grid style={dotted, gray!60},
    				enlargelimits=0.1
			]
			\addplot+[ycomb, thick, mark=*]
    			coordinates {
        				(1,2)
        				(2,-1)
       	 			(3,3)
        				(4,0)
        				(5,-2)
        				(6,1)
    			};
			\end{axis}
			\node at (2,-2.5) {(b)};
			\end{tikzpicture}
			\end{minipage}
		\caption{Two equivalent views of a vector $\vx= \big( 2, -1, 3, 0, -2, 1 \big)^{T} \in \mathbb{R}^{6}$.
			(a) As a numeric array. (b) As a \gls{map} $\featureidx \mapsto x_\featureidx$, where
			$\featureidx \in \{1, \,\dots, \,6\}$.}
			\label{fig:vector-function-dual_dict}
		\end{figure}
		See also: \gls{vectorspace}, \gls{euclidspace}, \gls{basis}, \gls{matrix}, \gls{linearmap}.},
	first={vector},
	firstplural={vectors},
	type=math,
	plural={vectors},
	text={vector}
}

\newglossaryentry{vectorspace}
{name={vector space},
	description={A\index{vector space} \gls{vector} space $\mathcal{V}$ (also called linear space) 
		is a collection of elements, called \glspl{vector}, along with the following two operations 
		(see also Fig.\ \ref{fig:vector-ops_dict}): 
    		1) addition (denoted by $\vv+\vw$) of two \glspl{vector} $\vv,\vw$; and 2) multiplication 
		(denoted by $c \,\cdot \,\vv$) of a \gls{vector} $\vv$ with a scalar $c$ that belongs to some 
		number \gls{field} (such as the real numbers $\mathbb{R}$ or the complex numbers $\mathbb{C}$). The defining 
		property of a \gls{vector} space is that it is closed under two specific operations. First, 
		if $\vv, \vw \in \mathcal{V}$, then $\vv + \vw \in \mathcal{V}$. Second, if $\vv \in \mathcal{V}$ 
		and $c \in \mathbb{R}$, then $c \vv \in \mathcal{V}$.
		\begin{figure}[H]
		\centering
			\begin{tikzpicture}[>=Stealth, scale=1.2]
  			\coordinate (O) at (0,0);            
  			\coordinate (V) at (2,1.5);          
  			\coordinate (W) at (1,3);            
  			\coordinate (VplusW) at (3,4.5);     
  			\coordinate (HalfV) at (1,0.75);     
  			\draw[->, thick, blue] (O) -- (V) node[pos=1, right] {$\vv$};
  			\draw[->, thick, red] (O) -- (W) node[pos=1, left] {$\vw$};
  			\draw[->, thick, purple] (O) -- (VplusW) node[pos=0.99, above right] {$\vv+\vw$};
  			\draw[dashed, red] (V) -- (VplusW);
  			\draw[dashed, blue] (W) -- (VplusW);
  			\draw[->, thick, orange] (O) -- (HalfV) node[midway, right] {$ \alpha \vv$};
  			\filldraw[black] (O) circle (2pt) node[below left] {$\mathbf{0}$};  
  			\filldraw[blue] (V) circle (2pt);         
  			\filldraw[red] (W) circle (2pt);          
  			\filldraw[purple] (VplusW) circle (2pt);  
  			\filldraw[orange] (HalfV) circle (2pt);   
			\end{tikzpicture}
		\caption{\Gls{vector} space $\mathcal{V}$ is a collection of \glspl{vector} such that 
			scaling and adding them always yields another \gls{vector} in $\mathcal{V}$.}
			\label{fig:vector-ops_dict}
		\end{figure}
		A common example of a \gls{vector} space is the \gls{euclidspace} $\mathbb{R}^n$, which is 
		widely used in \gls{ml} to represent \glspl{dataset}. We can also use $\mathbb{R}^n$ 
		to represent, either exactly or approximately, the \gls{hypospace} used by an \gls{ml} method.  
		Another example of a \gls{vector} space, which is naturally associated with every \gls{probspace} 
		$\big(\samplespace,\eventspace,\prob{\cdot} \big)$, is the collection of all 
		real-valued \glspl{rv} $x: \samplespace \rightarrow \mathbb{R}$ \cite{RudinBook}, \cite{folland1999real}.  
		\\
		See also: \gls{vector}, \gls{euclidspace}, \gls{basis}, \gls{hilbertspace}, \gls{linmodel}, \gls{linearmap}.},
	first={vector space},
	plural={vector spaces}, 
	firstplural={vector spaces}, 
	type=math,
	text={vector space}
}

\newglossaryentry{stochastic}
{name={stochastic},
	description={We refer to a\index{stochastic} method as stochastic if it involves a 
		random component or is governed by probabilistic laws. \Gls{ml} methods use randomness 
		to reduce computational complexity (e.g., see \gls{stochGD}) or 
		to capture \gls{uncertainty} in \glspl{probmodel}.
		\\
		See also: \gls{stochGD}, \gls{uncertainty}, \gls{probmodel}.},
	first={stochastic},
	type=math, 
	text={stochastic}
}

\newglossaryentry{stochproc}
{name={stochastic process},
	description={A \gls{stochastic} process\index{stochastic process} is a collection of 
		\glspl{rv} defined on a common \gls{probspace} and indexed by some set 
		$\indexset$ \cite{GrayProbBook}, \cite{papoulis}, \cite{Brockwell91}. The index set 
		$\indexset$ typically represents time or space, allowing us to represent 
		random phenomena that evolve across time or spatial dimensions—for example, 
		sensor noise or financial time series. \Gls{stochastic} processes are not limited 
		to temporal or spatial settings. For instance, random \glspl{graph} such as 
		the \gls{ergraph} or the \gls{sbm} can also be viewed as \gls{stochastic} processes. 
		Here, the index set $\indexset$ consists of node pairs that index \glspl{rv} whose values 
		encode the presence or weight of an edge between two nodes. Moreover, \gls{stochastic} 
		processes naturally arise in the analysis of \glspl{stochalgorithm}, 
		such as \gls{stochGD}, which construct a \gls{sequence} of \glspl{rv}. 
		\\
		See also:  \gls{rv}, \gls{sbm}, \gls{stochGD}, \gls{uncertainty}, \gls{probmodel}.},
	first={stochastic process},
	firstplural={stochastic processes},
	type=math, 
	plural={stochastic processes},
	text={stochastic process}
}

\newglossaryentry{characteristicfunc}
{name={characteristic function},
	description={The characteristic \gls{function}\index{characteristic function} 
		of a real-valued \gls{rv} $x$ is the following \gls{function} \cite[Sec. 26]{BillingsleyProbMeasure}:
		$$ \phi_{x}(t) \defeq \expect { \exp\,(j t x) } \quad \mbox{with } j = \sqrt{-1}.$$
	 	The characteristic \gls{function} uniquely determines the \gls{probdist} of $x$. 
		\\
		See also: \gls{rv}, \gls{probdist}.},
	first={characteristic function},
	firstplural={characteristic functions}, 
	type=math, 
	plural={characteristic functions},
	text={characteristic function}
}

\newglossaryentry{entropy}
{name={entropy},
	description={Entropy\index{entropy} quantifies the \gls{uncertainty} or 
		unpredictability associated with an \gls{rv} \cite{coverthomas}. 
		For a \gls{discreteRV} $x$ taking on values in a finite set 
		$\mathcal{S} = \{x_1, \,\ldots, \,x_\nrcluster\}$ with 
		a \gls{pmf} $\pmf{x}{x_{\clusteridx}} (=\prob{x = x_{\clusteridx}})$, 
		the entropy is defined as \cite{coverthomas}
		\[
		   \entropy{x} \defeq -\sum_{\clusteridx=1}^{\nrcluster} \pmf{x}{x_{\clusteridx}}  \log \pmf{x}{x_{\clusteridx}}.
		\]
		For a given set of values $\mathcal{S}$, the entropy is maximized for a 
		uniformly distributed \gls{rv}, where $\pmf{x}{x_{\clusteridx}}=1/\nrcluster$. 
		The minimal entropy, which is zero, is obtained when $\pmf{x}{x_{\clusteridx}}=1$ 
		for some $x_{\clusteridx} \in \mathcal{S}$.
		\Gls{diffentropy} generalizes the concept of entropy from \glspl{discreteRV} to 
		\gls{continuous} \glspl{rv}. 
		\\
		See also: \gls{uncertainty}, \gls{probmodel}.},
	first={entropy},
	type=math, 
	text={entropy}
}

\newglossaryentry{diffentropy}
{name={differential entropy},
	description={For\index{differential entropy} an 
		\gls{rv} $\featurevec \in \mathbb{R}^{\nrfeatures}$ 
		with a \gls{pdf} $\pdf{x}{\cdot}$, the differential \gls{entropy} 
		is defined as \cite{coverthomas}
		\[
			h(\featurevec) \defeq - \int_{\featurevec' \in \mathbb{R}^{\nrfeatures}} 
			\log p(\featurevec') \pdf{\featurevec}{\featurevec'} \, d \featurevec'.
		\]
		Differential \gls{entropy} can be negative and lacks some properties of 
		\gls{entropy} for discrete-valued \glspl{rv}, such as invariance under 
		a change of variables \cite{coverthomas}. Among all \glspl{rv} with a 
		given \gls{mean} $\meanvecgeneric$ and \gls{covmtx} $\covmtxgeneric$, 
		$h(\featurevec)$ is maximized by $\featurevec \sim \mvnormal{\meanvecgeneric}{\covmtxgeneric}$. 
		\\
		See also: \gls{uncertainty}, \gls{probmodel}.},
	first={differential entropy},
	type=math,
	text={differential entropy}
}

\newglossaryentry{domain}
{name={domain}, 
	description={The domain\index{domain} of a \gls{function} 
		$f: \mathcal{U} \rightarrow \mathcal{V}$ is the set $\mathcal{U}$ 
		from which $f$ takes its inputs.  
		\\
		See also: \gls{function}, \gls{co-domain}, \gls{map}.},
	first={domain},
	firstplural={domains}, 
	type=math, 
	plural={domains},
	text={domain}
}

\newglossaryentry{function}
{name={function}, 
	description={A function\index{function} between two sets $\mathcal{U}$ and $\mathcal{V}$ assigns  
		each element $u \in \mathcal{U}$ exactly one element $f(u) \in \mathcal{V}$ \cite{RudinBookPrinciplesMatheAnalysis}.
		We write this as $$f: \mathcal{U} \rightarrow \mathcal{V}: u \mapsto f(u),$$ 
		where $\mathcal{U}$ is the \gls{domain} and $\mathcal{V}$ the \gls{co-domain} of $f$. 
		That is, a function $f$ defines a unique \gls{output} $f(u) \in \mathcal{V}$ for every 
		input $u \in \mathcal{U}$ (see Fig.\ \ref{fig_function_dict}).
		\begin{figure}[H]
			\centering
			\begin{tikzpicture}[>=stealth, node distance=1.2cm and 2.5cm]
				\tikzset{dot/.style={circle, fill=black, inner sep=1.2pt}}
				\node (A) [dot, label=left:$a$] {};
				\node (B) [dot, below=of A, label=left:$b$] {};
				\node (C) [dot, below=of B, label=left:$c$] {};
				\node (1) [dot, right=4cm of A, label=right:$\star$] {};
				\node (2) [dot, below=of 1, label=right:$\circ$] {};
				\node (3) [dot, below=of 2, label=right:$\otimes$] {};
				\node[draw=blue!70, thick, ellipse, inner sep=0.5cm, fit=(A)(B)(C), label=above:$\mathcal{U}$] {};
				\node[draw=green!70!black, thick, ellipse, inner sep=0.5cm, fit=(1)(2)(3), label=above:$\mathcal{V}$] {};
				\draw[->] (A) -- (2);
				\draw[->] (B) -- (1);
				\draw[->] (C) -- (2);
			\end{tikzpicture}
		\caption{Function \( f \colon \mathcal{U} \to \mathcal{V} \) mapping each element 
			of the \gls{domain} $\mathcal{U} =  \{a,b,c\}$ to exactly one element of 
			the \gls{co-domain} $\mathcal{V} = \{\star,\circ,\otimes\}$. \label{fig_function_dict}}
		\end{figure} 
		See also: \gls{domain}, \gls{co-domain}, \gls{output}. },
	first={function},
	firstplural={functions}, 
	type=math, 
	plural={functions},
	text={function}
}

\newglossaryentry{map}
{name={map}, 
	description={We\index{map} use the term map as a synonym for \gls{function}.
		\\
		See also: \gls{function}.},
	first={map},
	firstplural={maps},	
	type=math, 
	plural={maps},
	text={map}
}

\newglossaryentry{event}
{name={event}, 
	description={Consider\index{event} an \gls{rv} $\featurevec$, defined on some \gls{probspace}, 
		which takes values in a \gls{measurable} space $\featurespace$. An 
		event $\mathcal{A} \subseteq \featurespace$ is a subset of $\featurespace$ 
		such that the \gls{probability} $\prob{\featurevec \in \mathcal{A}}$ is well 
		defined. In other words, the \gls{preimage} $\featurevec^{-1}(\mathcal{A})$ 
		of an event belongs to the underlying \gls{sigmaalgebra}, i.e., the \gls{preimage} 
		is a \gls{measurable} subset of the \gls{samplespace} 
		\cite{RudinBook}, \cite{BillingsleyProbMeasure}, \cite{durrett2010probability}.	
		Roughly speaking, an event represents a set of possible \glspl{outcome} of some 
		process. One example of such a process could also be the treatment of a 
		health-care patient.
				\\
		See also: \gls{rv}, \gls{datapoint}, \gls{iidasspt}, \gls{probmodel}.},
	first={event},
	firstplural={events},
	plural={events},
	type=math,
	text={event} 
}

\newglossaryentry{countable}
{name={countable},
	description={A set is called countable\index{countable} if its 
		elements can be put into a one-to-one correspondence with the natural numbers 
		$\mathbb{N}=\{1,\,2,\,3,\,\ldots\}$ or with a finite subset of $\mathbb{N}$ \cite{HalmosSet}. 
		Equivalently, a set $\mathcal{A}$ is countable if there exists an \gls{injective} 
		\gls{function} $f:\mathcal{A}\rightarrow\mathbb{N}$. 
		\begin{figure}[H]
			\centering
			\begin{tikzpicture}[>=stealth, node distance=1.0cm, thick]
  			\node (a1) {$a_1$};
  			\node[below=of a1] (a2) {$a_2$};
  			\node[below=of a2] (a3) {$a_3$};
  			\node[left=0.4cm of a2, align=center] {$\mathcal{A}$};
  			\begin{scope}[on background layer]
    			\draw[rounded corners, dashed, gray] ($(a1)+(-0.5,0.4)$) rectangle ($(a3)+(0.5,-0.4)$);
  			\end{scope}
  			\node[right=3.0cm of a1] (n1) {$1$};
  			\node[below=of n1] (n2) {$2$};
  			\node[below=of n2] (n3) {$3$};
  			\node[below=of n3] (n4) {$4$};
  			\node[below=of n4] (ndots) {$\vdots$};
  			\node[right=0.4cm of n2, align=center] {$\mathbb{N}$};
  			\draw[->] (a1) -- (n3);
  			\draw[->] (a2) -- (n1);
  			\draw[->] (a3) -- (n4);
			\end{tikzpicture}
		\caption{\Gls{injective} \gls{function} mapping the elements of a finite set 
			$\mathcal{A}$ to the natural numbers $\mathbb{N}$, which implies that $\mathcal{A}$ is countable.}
		\end{figure}
		Typical examples include the set of integers $\mathbb{Z}$ and rational 
		numbers $\mathbb{Q}$. In contrast, the set of real numbers $\mathbb{R}$ 
		is not countable, meaning no such one-to-one correspondence with $\mathbb{N}$ exists.
		\\ 
		See also: \gls{injective}, \gls{function}.}, 
	first={countable}, 
	type=math, 
	text={countable}
}

\newglossaryentry{pmf}
{name={probability mass function (pmf)}, 
	description={The pmf\index{probability mass function (pmf)} 
		of a \gls{discreteRV} $\feature$ is a \gls{function} 
		$\pmf{\feature}{\cdot}: \featurespace \rightarrow [0,1]$ that assigns to each 
		possible value $\feature' \in \featurespace$ of the \gls{rv} $\feature$ 
		the \gls{probability} $\pmf{\feature}{\feature'} = \prob{\feature' = \feature}$ \cite{papoulis}. 
		Fig.\ \ref{fig_pmf_dict} illustrates the pmf of a \gls{discreteRV} $\feature$. 
		\begin{figure}[H]
			\centering
			\begin{tikzpicture}[>=stealth, thick,y=2cm]
  			\foreach \x/\p in {1/0.3, 4/0.7}{
			\draw[gray] (\x,0) -- (\x,\p);
			\fill[blue] (\x,\p) circle (2pt);
			}
  			\node[anchor=south,align=center] at (1,0.3) {\small $\pmf{\feature}{\star}=\frac{3}{10}$};
  			\node[anchor=north] at (1,0) {\small $\star$};
  			\node[anchor=north] at (4,0) {\small $\otimes$};
  			\node[anchor=west,text width=11cm] at (-1.2,-0.80) {
    			$\dataset = (\star,\,\star,\,\star,\,\otimes,\,\otimes,\,\otimes,\,\otimes,\,\otimes,\,\otimes,\,\otimes)$
  			};
  			\node[anchor=west,text width=11cm] at (-5.2,1.18) {\small
    			$\dataset' = (\otimes,\,\star,\,\otimes,\,\star,\,\otimes,\,\otimes,\,\star,\,\otimes,\,\otimes,\,\otimes)$
  			};
  			\node[anchor=west,text width=11cm] at (3.2,1.56) {
    			$\dataset'' = (\otimes,\,\otimes,\,\otimes,\,\star,\,\otimes,\,\star,\,\otimes,\,\otimes,\,\star,\,\otimes)$
  			};
			\end{tikzpicture}
		\caption{\Gls{discreteRV} $\feature$ taking values in the set $\featurespace = \{\star,\otimes\}$
			with a pmf $\pmf{\feature}{\cdot}$. Three \glspl{dataset} 
			are also shown whose relative frequencies of \glspl{datapoint} match 
			this pmf exactly. Such \glspl{dataset} could arise as \glspl{realization} of 
			\gls{iid} \glspl{rv} sharing a common pmf $\pmf{\feature}{\cdot}$. 
			\label{fig_pmf_dict}}
		\end{figure}
		A pmf always satisfies $\sum_{\feature' \in \featurespace} \pmf{\feature}{\feature'} = 1$. 
		We can view a pmf as representing a collection of (sufficiently large)  
		\glspl{dataset}. This collection contains any 
		$\dataset = \{\feature^{(1)}, \,\ldots, \,\feature^{(\samplesize)}\}$, 
		with the relative frequencies of every value $\feature' \in \featurespace$ 
		being close to the corresponding pmf value $\pmf{\feature}{\feature'}$:
		$$ \frac{\big|\sampleidx \in \{1,\,\ldots,\,\samplesize\}: \feature^{(\sampleidx)}= \feature' \big|}
		{\samplesize} \approx \pmf{\feature}{\feature'}.$$ 
		Note that requiring relative frequencies to be close to the pmf values 
		implies that the empirical \gls{entropy} of such a \gls{dataset} is close to the 
		\gls{entropy} of the pmf $\pmf{\feature}{\cdot}$. Information theory refers 
		to the collection of such \glspl{dataset} as the typical set corresponding to the pmf 
		$\pmf{\feature}{\cdot}$ \cite{coverthomas}. A main result of information theory states 
		that a \gls{dataset} generated by \gls{iid} sampling from $\pmf{\feature}{\cdot}$ 
		belongs, with high \gls{probability}, to the typical set with respect to $\pmf{\feature}{\cdot}$ \cite[Th. 3.1.2]{coverthomas}.
				\\
		See also: \gls{discreteRV}, \gls{probability}, \gls{probdist}, \gls{probmodel}.},
	first={probability mass function (pmf)},
	firstplural={probability mass functions (pmfs)},
	plural={pmfs},
	type=math,
	text={pmf} 
}

\newglossaryentry{discreteRV}
{name={discrete random variable (discrete RV)}, 
 	description={An\index{discrete random variable (discrete RV)} \gls{rv}, i.e., 
		a \gls{function} that maps the \glspl{outcome} of a \gls{randomexperiment} 
		to elements of a \gls{measurable} space $\featurespace$, 
		is referred to as discrete if its value space 
		$\featurespace$ is \gls{countable} \cite{BillingsleyProbMeasure}. 
			\\
		See also: \gls{rv}, \gls{probability}, \gls{probdist}.},
 	first={discrete random variable (discrete RV)},
	firstplural={discrete random variables (discrete RVs)}, 
	plural={discrete RVs},
	type=math, 
 	text={discrete RV}  
}

\newglossaryentry{rv}
{name={random variable (RV)},
 	description={An RV\index{random variable (RV)} is a \gls{function} that maps the 
		\glspl{outcome} of a \gls{randomexperiment} to elements of a \gls{measurable} space 
		\cite{BillingsleyProbMeasure}, \cite{GrayProbBook}. 
 		Mathematically, an RV is a \gls{function} $x: \samplespace \rightarrow \featurespace$ 
		whose \gls{domain} is the \gls{samplespace} $\samplespace$ of a \gls{probspace} and 
		whose \gls{co-domain} is a \gls{measurable} space $\featurespace$. 
 		Different types of RVs include  
 		\begin{itemize} 
 			\item {binary RVs}, which map each \gls{outcome} to an element of a binary 
			set (e.g., $\{-1,1\}$ or $\{\text{cat}, \text{no cat}\}$); 
			\item {\glspl{discreteRV}}, which take on values in a \gls{countable} set (which can 
			be finite or countably infinite); 
 			\item {real-valued RVs}, which take on values in the real numbers $\mathbb{R}$;  
 			\item {\gls{vector}-valued RVs}, which map \glspl{outcome} to the \gls{euclidspace} $\mathbb{R}^{\featuredim}$.  
 		\end{itemize} 
 		\Gls{probability} theory uses the concept of \gls{measurable} spaces to rigorously define 
 		and study the properties of collections of RVs \cite{BillingsleyProbMeasure}.
			\\
		See also: \gls{function}, \gls{randomexperiment}, \gls{samplespace}, \gls{probspace}, \gls{vector}, \gls{euclidspace}, \gls{probability}, \gls{measurable}.}, 
	first={random variable (RV)},
	firstplural={random variables (RVs)},
	plural={RVs},
	type=math,
	text={RV}  
}

\newglossaryentry{outcome}
{name={outcome}, 
  	description={Outcome is one possible\index{outcome} result of a physical process.  
		Such a process could be the observation of a physical phenomenon,  
		a computation performed by an \gls{algorithm}, or a \gls{randomexperiment}  
		\cite{BillingsleyProbMeasure}.   
		\\
 		See also: \gls{samplespace}.},  
	type=math, 
  	first={outcome}, 
 	firstplural={outcomes},
 	plural={outcomes},
  	text={outcome}
}

\newglossaryentry{probspace}
{name={probability space}, 
 	description={A\index{probability space} \gls{probability} space is a mathematical 
 		structure that allows us to reason about a \gls{randomexperiment}, e.g., 
		the observation of a physical phenomenon. 
 	   	Formally, a \gls{probability} space $\mathcal{P}$ is a triplet 
		$(\samplespace, \eventspace, \prob{\cdot})$ where
 		\begin{itemize} 
 			\item  $\samplespace$ is a \gls{samplespace} containing all possible \glspl{outcome} 
			of a \gls{randomexperiment};
 			\item  $\eventspace$ is a \gls{sigmaalgebra}, i.e., a collection of subsets of 
			$\samplespace$ (called \glspl{event}) that satisfies certain closure properties 
			under set operations;
 			\item $\prob{\cdot}$ is a \gls{probdist}, i.e., a \gls{function} that assigns 
			a \gls{probability} $\prob{\mathcal{A}} \in [0,1]$ to each \gls{event} $\mathcal{A} 
			\in \eventspace$. This \gls{function} must satisfy $\prob{\samplespace} = 1$ and 
			$\prob{\bigcup_{i=1}^{\infty} \mathcal{A}_i} = \sum_{i=1}^{\infty} \prob{\mathcal{A}_i}$ 
			for any \gls{countable} \gls{sequence} of pairwise disjoint \glspl{event} $\mathcal{A}_1, \,\mathcal{A}_2, \,\ldots$ in $\mathcal{F}$.
 		\end{itemize}
 		\Gls{probability} spaces provide the foundation of \glspl{probmodel} 
		that can be used to study the behavior of \gls{ml} methods \cite{BillingsleyProbMeasure}, \cite{GrayProbBook}, \cite{ross2013first}.
				\\
		See also: \gls{probability}, \gls{randomexperiment}, \gls{samplespace}, \gls{event}, \gls{probdist}, \gls{function}, \gls{probmodel}, \gls{ml}.},  
 	first={probability space}, 
	plural={probability spaces},
	firstplural={probability spaces},
	type=math, 
 	text={probability space}
}

\newglossaryentry{integrable}
{name={integrable},
	description={A \gls{measurable} \gls{function} $f\!:\!\samplespace \!\to\! \mathbb{R}$ 
		defined on a \gls{measurespace} $(\samplespace, \,\sigmaalgebra, \,\mu)$ 
		is called integrable\index{integrable} if the \gls{LebesgueIntegral} 
		of its absolute value is finite, i.e.,
		\[
		\int_{\samplespace} |f(x)|\,\mathrm{d}\mu < \infty.
		\]
		In this case, the \gls{LebesgueIntegral} $\int_{\samplespace} f(x)\,\mathrm{d}\mu$ 
		is well defined and finite. An \gls{rv} $x$ defined on the \gls{samplespace} 
		of a \gls{probspace} $(\samplespace, \,\sigmaalgebra, \,\probdist)$ 
		is integrable if
		\[
		\expect\{|x|\}
		= \int_{\samplespace} |x(\omega)|\,\mathrm{d}\probdist 
		< \infty,
		\]
		which is equivalent to the existence of the \gls{expectation} $\expect\{x\}$ (i.e., it is finite). 
		\\ 
		See also: \gls{measurespace}, \gls{measure}.},
	first={integrable},
	type=math, 
	text={integrable}
}

\newglossaryentry{measurespace}
{name={measure space},
	description={A \gls{measure} space\index{measure space} is a triple $(\samplespace, \,\sigmaalgebra, \,\mu)$ consisting of 
        		a set $\samplespace$, a \gls{sigmaalgebra} $\sigmaalgebra$ of subsets 
		of $\samplespace$, and a \gls{measure} $\mu\!:\sigmaalgebra\!\to\![0,\infty)$. 
        		The \gls{measure} $\mu$ assigns a nonnegative number to each \gls{measurable} 
		set $\mathcal{A} \in \sigmaalgebra$, generalizing the 
		notions of length, area, or volume in \glspl{euclidspace} \cite{RudinBookPrinciplesMatheAnalysis}, \cite{HalmosMeasure}. 
		\Gls{measure} spaces provide the mathematical foundation for the 
		\gls{LebesgueIntegral} or the definition of \glspl{rv} as \gls{measurable} mappings 
        		between \gls{measure} spaces.  
        		A \gls{probspace} is a special case of a \gls{measure} space 
		where the total \gls{measure} of the \gls{samplespace} is normalized to one, 
        		i.e., $\mu(\samplespace) = 1$. In this case, $\mu$ is called a \gls{probdist}.
		\\ 
		See also: \gls{measurable}, \gls{probspace}, \gls{probdist}. },
    	first={measure space},
	plural={measure spaces}, 
	firstplural={measure spaces}, 
	type=math, 
    	text={measure space}
}

\newglossaryentry{measure}
{name={measure},
	description={A measure\index{measure} $\mu$ on a set $\samplespace$ equipped with a 
		\gls{sigmaalgebra} $\sigmaalgebra$ is a \gls{function} $\mu: \sigmaalgebra \to [0, \infty)$ 
		that assigns a nonnegative value to each \gls{measurable} set 
		$\mathcal{A} \in \sigmaalgebra$ such that 
		\cite{RudinBookPrinciplesMatheAnalysis}, \cite{BillingsleyProbMeasure}, \cite{HalmosMeasure}: 
		1) $\mu(\emptyset) = 0$; and 
		2) for any \gls{countable} collection $\{\mathcal{A}_i\}_{i=1}^{\infty}$ of 
		pairwise disjoint sets in $\sigmaalgebra$,
		\[
		\mu\!\left(\bigcup_{i=1}^{\infty} A_i\right) = \sum_{i=1}^{\infty} \mu(A_i),
		\]
		which is referred to as ``\gls{countable} additivity''.
		\\
		See also: \gls{measurable}, \gls{countable}. },
	first={measure},
	type=math, 
	plural={measures}, 
	firstplural={measures}, 
	text={measure}
}

\newglossaryentry{LebesgueIntegral}
{name={Lebesgue integral},
	description={The Lebesgue integral\index{Lebesgue integral} 
		assigns each \gls{integrable} \gls{function} 
		$f: \mathbb{R}^{\nrfeatures} \rightarrow \mathbb{R}$ a number 
		$\int_{\mathbb{R}^{\nrfeatures}} f(\featurevec) d\featurevec$ 
		that is referred to as the integral of $f$. 
		The integral of $f$ can be interpreted as the volume that 
		is enclosed by the \gls{function} $f$ in the space 
		$\mathbb{R}^{\nrfeatures+1}$. We can compute it by increasingly 
		accurate approximations using \glspl{simplefunction} 
		\cite[Ch. 1]{RudinBook}. 
		\begin{figure}[H]
			\centering
			\begin{tikzpicture}[scale=1.5]
  			\draw[->] (-0.2,0) -- (3.4,0) node[right] {$x$};
  			\draw[thick,domain=0:3,smooth] plot(\x,{0.5+0.3*\x+0.1*\x*\x}) node[right] {$f(x)$};
  			\fill[gray!20] (0,0) -- plot[domain=0:3] (\x,{0.5+0.3*\x+0.1*\x*\x}) -- (3,0) -- cycle;
  			\draw[fill=gray!50,opacity=0.35] (0,0) rectangle (1,0.5);
  			\draw[fill=gray!50,opacity=0.35] (1,0) rectangle (2,0.9);
  			\draw[fill=gray!50,opacity=0.35] (2,0) rectangle (3,1.5);
  			\draw[thick] (0,0.5)--(1,0.5);
  			\draw[thick] (1,0.9)--(2,0.9);
  			\draw[thick] (2,1.5)--(3,1.5);
 	 		\node[above left] at (0.8,0.5) {};
  			\foreach \x in {0,1,2,3} \draw[dashed,gray] (\x,0) -- (\x,2.5);
  			\node[anchor=north] at (0,-0.05) {$a$};
  			\node[anchor=north] at (1,-0.05) {$b$};
  			\node[anchor=north] at (2,-0.05) {$c$};
  			\node[anchor=north] at (3,-0.05) {$d$};
  			\node at (1.6,3.0) {$\displaystyle \int_a^d f(x)\,dx \approx \underbrace{f(a)(b-a) + f(b)(c-b)+f(c)(d-c)}_{=:\int_a^d s(x)dx}$};
  			\node [anchor=west] at (3.2,1.5) {\gls{simplefunction} $s(x)$};
			\end{tikzpicture}
		\end{figure}
 		It is useful to think of the Lebesgue integral as a \gls{function} that maps 
 		an \gls{integrable} \gls{function} $f$ to the value of its integral:
		$$ f \mapsto \int_{\mathbb{R}^{\nrfeatures}} f(\featurevec) d\featurevec.$$ 
		The precise definition of this \gls{function}, whose \gls{domain} 
		consists of the \gls{integrable} \glspl{function}, is a cornerstone of 
		\gls{measure} theory \cite[Ch. 1]{RudinBook}.
					\\ 
		See also: \gls{function}.},
	first={Lebesgue integral},
	text={Lebesgue integral},
	type=math, 
	plural={Lebesgue integrals},
	firstplural={Lebesgue integrals}
}

\newglossaryentry{conditionalexpect}
{name={conditional expectation}, 
	description={Consider a numeric \gls{rv} $\featurevec \in \mathbb{R}^{\nrfeatures}$ defined on a 
                 \gls{probspace} $(\samplespace,\,\sigmaalgebra,\,\probmeasure)$. 
                 Let $\sigmaalgebra' \subseteq \sigmaalgebra$ be a (sub-)\gls{sigmaalgebra} that 
                 represents partial information about the \gls{outcome} of a \gls{randomexperiment}. 
                 The conditional \gls{expectation}\index{conditional expectation} of 
                 $\featurevec$ given (or conditioned on) $\sigmaalgebra'$, denoted by
                 $\expect\{ \featurevec \mid \sigmaalgebra'\}$, is a numeric \gls{rv} 
		that \cite[Sec. 34]{BillingsleyProbMeasure}, \cite{durrett2010probability}:
                	1) is \gls{measurable} with respect to $\sigmaalgebra'$; and
                 2) satisfies
		$$
    		\int_{\mathcal{A}} \expect\{\featurevec \mid \sigmaalgebra'\} {\rm d} \probmeasure =
    		\int_{\mathcal{A}} \featurevec {\rm d} \probmeasure \quad \text{for any } \mathcal{A} \in \sigmaalgebra'.
		$$
               	Intuitively, $\expect\{\featurevec \mid \sigmaalgebra'\}$ summarizes 
               	the average value of $\featurevec$ using only information contained 
               	in the (typically smaller) \gls{sigmaalgebra} $\sigmaalgebra'$ 
		\cite{BillingsleyProbMeasure}, \cite{GrayProbBook}, \cite{ross2013first}. 
		\\
              	See also: \gls{probspace}, \gls{sigmaalgebra}, \gls{expectation}.}, 
 	first={conditional expectation},
 	plural={conditional expectations}, 
 	type=math, 
 	firstplural={conditional expectations},  
 	text={conditional expectation}
}

\newglossaryentry{conditionalpmf}
{name={conditional probability mass function (conditional pmf)}, 
	description={Consider two \glspl{discreteRV} $\truelabel$ and $\feature$ defined on 
                 the same \gls{probspace} $(\samplespace,\,\eventspace,\,\prob{\cdot})$. 
                 The conditional \gls{pmf}\index{conditional probability mass function (conditional pmf)} 
                 of $\truelabel$ given (or conditioned on) $\feature$ is denoted by
		$\pmf{\truelabel \mid \feature}{\cdot \mid \cdot}$ and is defined by
                \[
                  \pmf{\truelabel \mid \feature}{\truelabel' \mid \feature'}
                  :=
                  \prob{\truelabel=\truelabel' \mid \feature=\feature'}
               \]
                for all \glspl{realization} $\truelabel',\feature'$ with 
                $\prob{\feature=\feature'}>0$. 
                Equivalently, the conditional \gls{pmf} can be expressed using 
                \gls{conditionalexpect} as
                \[
                  \pmf{\truelabel \mid \feature}{\truelabel' \mid \feature'}
                  =
                  \expect\{\indicatorfunc{\truelabel=\truelabel'}\mid \sigmaalgebra(\feature)\}(\feature'),
                \]
                where $\sigmaalgebra(\feature)$ denotes the \gls{sigmaalgebra} generated by 
                the \gls{rv} $\feature$. 
                \\
              See also: \gls{probspace}, \gls{pmf}, \gls{conditionalexpect}.}, 
 	first={conditional probability mass function (conditional pmf)},
 	firstplural={conditional probability mass functions (conditional pmfs)}, 
 	type=math, 
 	plural={conditional pmfs},  
 	text={conditional pmf}
}

\newglossaryentry{iid}
{name={independent and identically distributed (i.i.d.)}, 
	description={A collection of \glspl{rv}\linebreak $\datapoint^{(1)}, \,\ldots, \,\datapoint^{(\samplesize)}$ 
		defined on the same \gls{probspace} $(\samplespace, \sigmaalgebra, \prob{\cdot})$ is 
		referred to as i.i.d.\index{independent and identically distributed (i.i.d.)} 
		if each $\datapoint^{(\sampleidx)}$ follows the same \gls{probdist}, and 
		the \glspl{rv} are mutually independent. That is, for any collection of 
		\glspl{event} $\mathcal{A}_1, \,\ldots, \,\mathcal{A}_\samplesize \in \sigmaalgebra$, we have
       		\[
          		\prob{ \datapoint^{(1)} \in \mathcal{A}_1, \,\ldots, \,\datapoint^{(\samplesize)} \in \mathcal{A}_{\samplesize}} 
         		= \prod_{\sampleidx=1}^{\samplesize} \prob{ \datapoint^{(\sampleidx)} \in \mathcal{A}_\sampleidx}.
         	\]
		Probability theory also guarantees the existence and construction of independent copies of \glspl{rv} 
		\cite[Th. 2.19]{klenke2020probability}.
				\\
		See also: \gls{rv}, \gls{probdist}, \gls{event}, \gls{datapoint}, \gls{iidasspt}.},
	first={independent and identically distributed (i.i.d.)},
	type=math, 
	text={{i.i.d.}} 
}

\newglossaryentry{likelihood}
{name={likelihood function},
	description={The\index{likelihood function} likelihood \gls{function} of a \gls{probmodel} 
		with \glspl{parameter} $\weights$ is obtained by evaluating the 
		\gls{pdf} (or \gls{pmf}) for the observed \gls{dataset} \cite{LC}, \cite{CasellaBergerStatInf}. 
		More precisely, consider a \gls{dataset} 
		$\dataset' = \{\datapoint^{(1)}, \,\ldots, \,\datapoint^{(\samplesize)}\}$ 
		that is assumed to be generated via \gls{iid} 
		\glspl{rv} with a common \gls{pdf} $\pdf{\datapoint;\weights}{\cdot}$. 
		The common \gls{pdf} is parameterized by the \glspl{modelparam} $\weights$. 
		The likelihood \gls{function} is then defined by 
		$L^{(\dataset')}(\weights) = \prod_{\sampleidx=1}^{\samplesize} \pdf{\datapoint;\weights}{\datapoint^{(\sampleidx)}}$.
		The likelihood \gls{function} can be used to construct a \gls{lossfunc} 
		for \gls{erm}, which results in \gls{maxlikelihood}. 
		\\
		See also: \gls{probmodel}, \gls{lossfunc}, \gls{maxlikelihood}.},
	first={likelihood function},
	plural={likelihood functions},
	type=math,
	text={likelihood function},
	firstplural={likelihood functions}
}

\newglossaryentry{maxlikelihood}
{name={maximum likelihood}, 
	description={Consider\index{maximum likelihood} \glspl{datapoint} 
		$\dataset=\big\{ \datapoint^{(1)}, \,\ldots, \,\datapoint^{(\samplesize)} \}$ 
		that are interpreted as the \glspl{realization} of \gls{iid} \glspl{rv} 
		with a common \gls{probdist} $\probdist^{(\dataset;\weights)}$, which depends 
		on the \glspl{modelparam} $\weights \in \mathcal{W} \subseteq \mathbb{R}^{n}$. 
		\Gls{maximum} likelihood methods learn \glspl{modelparam} $\weights$ by maximizing 
		the \gls{pdf} (or \gls{pmf}) 
		$\pdf{\dataset;\weights}{\dataset'} = \prod_{\sampleidx=1}^{\samplesize} 
		\pdf{\datapoint;\weights}{\datapoint^{(\sampleidx)}}$ 
		of the observed \gls{dataset} $\dataset'$. Thus, the \gls{maximum} 
		likelihood \gls{estimator} is a solution to the \gls{optproblem} 
		$\max_{\weights \in \mathcal{W}} \pdf{\dataset;\weights}{\dataset'}$.
				\\
		See also: \gls{probdist}, \gls{optproblem}, \gls{probmodel}.},
	first={maximum likelihood},
	type=math, 
	text={maximum likelihood}
}

\newglossaryentry{preimage}
{name={preimage}, 
	description={Consider a \gls{function}\index{preimage} $f\colon \mathcal{U} \rightarrow \mathcal{V}$ 
		between two sets. The preimage $f^{-1}(\mathcal{B})$ of a subset $\mathcal{B} \subseteq \mathcal{V}$ is the set 
		of all inputs $u \in \mathcal{U}$ that are mapped into $\mathcal{B}$ by $f$, i.e.,
		\[
		f^{-1}(\mathcal{B}) \defeq \{ u \in \mathcal{U} \mid f(u) \in \mathcal{B} \}.
		\]
		The preimage is well defined even if the \gls{function} $f$ is non-invertible \cite{RudinBookPrinciplesMatheAnalysis}.
		\\
		See also: \gls{function}. },
	first={preimage},
	type=math, 
	text={preimage}
}

\newglossaryentry{measurable}
{name={measurable}, 
	description={Consider\index{measurable} a \gls{randomexperiment}, such as recording 
		the air temperature at an \gls{fmi} weather station. The corresponding \gls{samplespace} 
		$\samplespace$ consists of all possible \glspl{outcome} $\outcome$ (e.g., 
		all possible temperature values in degree Celsius). In many \gls{ml} 
		applications, we are not interested in the exact \gls{outcome} $\outcome$, but only 
		whether it belongs to a subset $\mathcal{A} \subseteq \samplespace$ 
		(e.g., determining whether the temperature is below zero degrees). 
		We call such a subset $\mathcal{A}$ measurable if it is possible to 
		decide, for any \gls{outcome} $\outcome$, whether $\outcome \in \mathcal{A}$ 
		(see Fig.\ \ref{fig_measurable_dict}).
		\begin{figure}[H]
		\begin{center}
		\begin{tikzpicture}
			\draw[->] (0,0) -- (8.5,0) node[right] {temperature ($^\circ$C)};
			\foreach \x/\label in {0/--20, 1/--10, 2/0, 3/10, 4/20, 5/30, 6/40, 7/50, 8/60} {
			\draw (\x,0.1) -- (\x,-0.1);
			\node[below] at (\x,-0.1) {\label};
			}
			\fill[blue!20] (0,0.3) rectangle (2,0.6);
			\node[above] at (1,0.6) {$\outcome < 0\;^\circ$C};
			\fill[red!20] (5.5,0.3) rectangle (7.5,0.6);
			\node[above] at (6,0.6) {$35\;^\circ$C $< \outcome < 55\;^\circ$C};
			\vspace*{10mm}
			\end{tikzpicture}
			\vspace*{10mm}
			\end{center}
		\caption{\Gls{samplespace} constituted by all possible temperature values $\outcome$ 
			that can occur at an \gls{fmi} station. Two measurable subsets of temperature 
			values, denoted by $\mathcal{A}^{(1)}$ and $\mathcal{A}^{(2)}$, are highlighted. For any 
			actual temperature value $\outcome$, it is possible to determine (via some equipment) 
			whether $\outcome \in \mathcal{A}^{(1)}$ and whether $\outcome \in \mathcal{A}^{(2)}$. 
			\label{fig_measurable_dict}} 
		\end{figure}
		In principle, measurable sets could be chosen freely (e.g., depending on the resolution of the 
		measuring equipment). However, it is often useful to impose certain completeness requirements 
		on the collection of measurable sets. For example, the \gls{samplespace} itself should be 
		measurable, and the union of two measurable sets should also be measurable. These completeness 
		requirements can be formalized via the concept of a \gls{sigmaalgebra} (or \gls{sigmafield}) 
		\cite{RudinBook}, \cite{BillingsleyProbMeasure}, \cite{durrett2010probability}. 
		A measurable space is a pair $\big(\featurespace,\eventspace\big)$ that consists of an arbitrary 
		set $\featurespace$ and a collection $\eventspace$ of measurable subsets of $\featurespace$ 
		that form a \gls{sigmaalgebra}. 
		\\
		See also: \gls{samplespace}, \gls{outcome}, \gls{sigmaalgebra}, \gls{probability}.},
	first={measurable},
	type=math, 
	text={measurable} 
}

\newglossaryentry{sigmaalgebra}
{name={$\sigma$-algebra}, 
sort={sigma-algebra},
	description={Consider a \gls{randomexperiment} with a \gls{samplespace} $\samplespace$. 
		A $\sigma$-algebra\index{$\sigma$-algebra} (or \gls{sigmafield}) $\sigmaalgebra$ 
		is a collection of subsets of $\samplespace$ with the following properties 
		\cite{RudinBook}, \cite{BillingsleyProbMeasure}, \cite{durrett2010probability}:
		\begin{itemize}
		 	\item The empty set $\emptyset$ and the entire \gls{samplespace} 
		 	$\samplespace$ belong to $\sigmaalgebra$, i.e., $\emptyset \in \sigmaalgebra$ and $\samplespace \in \sigmaalgebra$.
		 	\item If a set $\mathcal{A}$ belongs to $\sigmaalgebra$, then its complement 
		 	$\samplespace \setminus \mathcal{A}$ also belongs to $\sigmaalgebra$, i.e., 
		 	$\mathcal{A} \in \sigmaalgebra$ implies $\samplespace \setminus \mathcal{A} \in \sigmaalgebra$.
		 	\item If a \gls{countable} collection of sets $\mathcal{A}_1, \,\mathcal{A}_2, \,\ldots$ belongs 
			to $\sigmaalgebra$, 
		 	then their union also belongs to $\sigmaalgebra$, i.e.,
		 	$\mathcal{A}_1, \,\mathcal{A}_2, \,\ldots \in \sigmaalgebra$ implies 
		 	$\bigcup_{i=1}^{\infty} \mathcal{A}_i \in \sigmaalgebra$.	
		 \end{itemize}			 
		See also: \gls{samplespace}, \gls{rv}, \gls{probspace}.},
	first={$\sigma$-algebra},
	type=math, 
	text={$\sigma$-algebra} 
}

\newglossaryentry{sigmafield}
{name={$\sigma$-field}, 
sort={sigma-field},
	description={See \gls{sigmaalgebra}\index{$\sigma$-field}.}, 
	first={$\sigma$-field},
	type=math,
	text={$\sigma$-field} 
}

\newglossaryentry{injective}
{name={injective}, 
	description={A \gls{function} $f: \mathcal{U} \rightarrow \mathcal{V}$ is injective\index{injective} 
		if it maps distinct elements of its \gls{domain} to distinct elements 
		of its \gls{co-domain}, 
    		i.e., if $f(u_1) = f(u_2)$ implies $u_1 = u_2$ for all $u_1, u_2 \in \mathcal{U}$ 
        		\cite{HalmosSet}. 
    		Equivalently, no two different \gls{function} inputs are mapped to the same \gls{function} \gls{output}.
				\\
		See also: \gls{function}.},
	first={injective},
	type=math,
	text={injective} 
}

\newglossaryentry{typicalset}
{name={typical set}, 
	description={See \gls{pmf}\index{typical set}.}, 
 	first={typical set},
 	firstplural={typical sets},
 	type=math,
 	plural={typical sets},
 	text={typical set} 
}

\newglossaryentry{majmin}
{name={majorize-minimize (MM)}, 
	description={Consider an \gls{optproblem} $\min_{\weights \in \paramspace} f(\weights)$ with 
		some complicated (potentially non-\gls{convex} and \gls{nonsmooth}) \gls{objfunc}. 
		One important example of such an \gls{optproblem} is \gls{erm}, which is used to learn the 
		\glspl{modelparam} of a nonlinear \gls{model}. 
		An MM method is\index{majorize-minimize (MM)} an iterative \gls{optmethod} 
		that constructs a \gls{sequence} $\weights^{(1)},\,\weights^{(2)},\,\ldots \in \paramspace$ 
		of \glspl{modelparam} as follows \cite{BishopBook}, \cite{Lange2016MM}, \cite{Hunter01022004}
		(see also Fig.\ \ref{fig:majmin_dict}): 
		During the $\iteridx$th \gls{iteration}, the \gls{objfunc} $f(\cdot)$ 
		is approximated by another \gls{function} $g\big(\cdot;\weights^{(\iteridx)}\big)$.
		This approximation must be an upper bound for (i.e., must majorize) the original
		\gls{objfunc}, i.e., $g\big(\weights;\weights^{(\iteridx)}\big) \geq f(\weights)$ 
		for all $\weights \in \paramspace$, and it must be tight for $\weights^{(\iteridx)}$, i.e., 
		$g\big(\weights^{(\iteridx)};\weights^{(\iteridx)}\big) = f\big(\weights^{(\iteridx)}\big)$.
		The new \glspl{modelparam}  $\weights^{(\iteridx+1)}$ are then obtained by 
		minimizing the approximation, i.e., 
		$\weights^{(\iteridx+1)} \in \argmin_{\weights \in \paramspace}g\big(\weights;\weights^{(\iteridx)}\big)$. 
		\begin{figure}[H]
			 \centering
			 \begin{tikzpicture}[x=1.2cm,y=1cm]
				\def\xa{0}
				\def\xb{2*pi}
				\def\xo{3*pi/4}     
				\def\mL{0.4}        
				\def\mR{0.7}        
				\draw[->] (\xa-0.2,-2) -- (\xb+0.3,-2) node[right] {$\weights$};
				\draw[thick,samples=100,domain=\xa:\xb]
				plot (\x,{sin(\x r)}) node[pos=0.1,above left,black] {\small $f(\weights)$};
				\pgfmathsetmacro\yxo{sin(\xo r)}
				\def\xL{2.70}
				\def\xR{3.80}
				\def\yO{0.70}
				\draw[dashed,samples=2,domain=0:\xL]
				plot (\x,{\yO + (-0.7)*(\x - \xo)});
				\draw[dashed]
				(\xL,{\yO+(-0.7)*(\xL-\xo)}) -- (\xR,{\yO+(-0.7)*(\xL-\xo)});
				\draw[dashed,samples=2,domain=\xR:2*pi]
				plot (\x,{\yO+(-0.7)*(\xL-\xo) + (0.7)*(\x - \xR)});
				\pgfmathparse{\xR + 0.65*(2*pi - \xR)} \let\xmid\pgfmathresult
				\pgfmathparse{\yO + 0.7*(\xmid - \xR)} \let\ymid\pgfmathresult
				\node[right,black] at (\xmid,\ymid) {$g\big( \weights; \weights^{(\iteridx)} \big)$};
				\fill[red] (2.35619449,0.70710678) circle (1.2pt);
				\draw[densely dotted,gray] (\xo,-2) -- (\xo,\yO);
				\node[below] at (\xo,-2) {$\weights^{(\iteridx)}$};
			\end{tikzpicture}
		\caption{Construction of \glspl{modelparam} based on the iterative MM method.}
			\label{fig:majmin_dict}
		\end{figure}
		Similar to \glspl{gdmethod}, the MM principle is also based on approximating an 
		\gls{objfunc} locally, around the current \glspl{modelparam}, and then optimizing 
		this approximation to obtain new \glspl{modelparam}. However, the construction 
		of local approximations is very different. While \glspl{gdmethod} use linear \glspl{function} 
		for these approximations, MM methods can use nonlinear \glspl{function} as long as 
		they are upper bounds for the original \gls{objfunc}. 
			 \\ 
		See also: \gls{gdmethod}, \gls{em}.},
	first={majorize–minimize (MM)},
	type=math, 
	text={MM}
}

\newglossaryentry{markovsinequality}
{name={Markov's inequality},
	description={Consider a real-valued nonnegative \gls{rv} $x$ for which 
		the \gls{expectation} $\expect\{ x\}$ exists. \index{Markov's inequality} 
	 	Markov's inequality provides an upper bound on the \gls{probability} 
	 	$\prob{x\geq a}$ that $x$ exceeds a given positive threshold $a>0$.  
	 	In particular,          
	 	\begin{equation}
            		\nonumber
			\prob{x \geq a} \leq \frac{\expect \{ x\}}{a} \quad \mbox{holds for any } a > 0. 
    		\end{equation}
    		This inequality can be verified by noting that $\prob{x \geq a}$ is the 
		\gls{expectation} $\expect\{g(x)\}$ with the following \gls{indicatorfunc}:
	 	$$g: \mathbb{R} \rightarrow \mathbb{R}: x' \mapsto \indicatorfunc{\{x \geq a\}}(x').$$ 
	 	As illustrated in Fig.\ \ref{fig:markovsinequality_dict}, for any positive $a>0$, 
	 	$$ g(x') \leq x'/a \quad \mbox{for all } x' \in \mathbb{R}.$$ 
	 	This implies Markov's inequality via the monotonicity property 
	 	of the \gls{LebesgueIntegral} \cite[p. 50]{folland1999real}. 
    		\begin{figure}[H]
			\centering
			\begin{tikzpicture}[scale=1, x=0.8cm, y=0.8cm]
			\def\a{3.24}      
			\def\xmax{10}     
			\draw[-{Latex}] (0,0) -- (\xmax+1,0) node[below right] {$x'$};
                		\draw[-{Latex}] (0,0) -- (0,3.1)  node[left, text=blue!70!black] {$\pdf{x}{x'}$};
			\fill[blue!15, opacity=0.3]
			plot[samples=400, domain=0:\xmax, smooth]
			(\x,{ (6/sqrt(2*pi)) * (\x)^(1.5) * exp(-\x/2) }) -- (\xmax,0) -- (0,0) -- cycle;
			\draw[blue!70!black, very thick, samples=400, domain=0:\xmax, smooth]
			plot (\x,{ (6/(sqrt(2*pi))) * (\x)^(1.5) * exp(-\x/2) })
			node[pos=0.9, above right, xshift=2pt] {};
			\draw[dashed, gray] (\a,0) -- (\a,1.05);
			\node[below] at (\a,0) {$a$};
			\node[above] at (1*\a,3) {$\prob{x \geq a} \leq \frac{\expect \{ x\}}{a}$}; 
			\node[below] at (0,0) {$0$};
			\draw[green!60!black, ultra thick] (0,0) -- (\a,0);
			\filldraw[white, draw=green!60!black, line width=0.8pt] (\a,0) circle (2pt);
			\draw[green!60!black, ultra thick] (\a,1) -- (\xmax,1)
			node[pos=0.9, above, yshift=2pt] {$\indicatorfunc{\{x \ge a\}}(x')$};
			\filldraw[green!60!black] (\a,1) circle (2pt);
			\draw[red!70, very thick, samples=2, domain=0:\xmax]
			plot (\x,{ \x*(1/\a) }); 
			\node[align=right,red!70,yshift=20pt] at ({2.5*\a+0.2},{2.5}) {$f(x') = x'/a$};  
			\draw (0,1) -- ++(-0.12,0) node[left] {$1$};
			\end{tikzpicture}
            	\caption{\Gls{expectation} $\expect\{x\}$ and \gls{probability} $\prob{x \geq a}$ 
			of a nonnegative \gls{rv} $x$ with a \gls{pdf} $\pdf{x}{\cdot}$,           
			obtained via \glspl{LebesgueIntegral} of 
			$f(x') = x'/a$ and $g(x') = \indicatorfunc{\{x \geq a\}}(x')$, respectively.}
            		\label{fig:markovsinequality_dict}
        		\end{figure} 
		See also: \gls{expectation}, \gls{probability}, \gls{concentrationinequ}.},
	first={Markov's inequality},
	type=math, 
    	text={Markov's inequality}  
}

\newglossaryentry{chebyshevsinequality}
{name={Chebyshev's inequality},
	description={Consider a real-valued \gls{rv} $x$ for which 
		the second moment $\expect\{ x^{2} \}$ exists (and is finite). 
 	 	The existence of the second moment implies the existence 
 	 	of a finite \gls{expectation} $\mu \defeq \expect\{ x\}$ and a finite 
 	 	\gls{variance} $\sigma^{2}\!\defeq\!\expect \big\{ \big( x\!-\!\mu \big)^{2} \big\}$ \cite[Proposition 6.12]{folland1999real}. 
 	 	Chebyshev's inequality\index{Chebyshev's inequality} refers 
 	 	to the following upper bound on the \gls{probability} that $x$ 
 	 	deviates from $\mu$ by more than a given threshold $\eta$ 
 	 	\cite[Ch. 4]{FellerBook}. 
 	 	In particular,   
 		\begin{equation}
           		\nonumber
			\prob{\big|x - \mu \big| \geq \eta } \leq \frac{\sigma^2}{\eta^2} \quad \mbox{for any } \eta > 0.
     		\end{equation}
 		This upper bound can be obtained by applying \gls{markovsinequality} to 
 		the new \gls{rv} $\tilde{x}\!\defeq\! \big(x\!-\!\mu \big)^2$. 
  		\begin{figure}[H]
   			\centering
			\begin{tikzpicture}
     			\begin{axis}[
      			width=9cm, height=4.2cm,
      			samples=300,
      			axis lines=left,
	  		ylabel={$\pdf{x}{x'}$},
      			xlabel={$x'$},
      			x label style={at={(axis description cs:1,0)}, anchor=west},
	  		ylabel style={rotate=270,anchor=south,at={(axis description cs:0,1.02)}},
      			xtick={-1.5,0,1.5},
      			xticklabels={$-\eta$,$\mu$,$\eta$},
      			ytick=\empty,
      			ymin=0, ymax=0.45,
      			domain=-4:4,
      			clip=false
    			]
      			\addplot[name path=pdf, black, very thick] {exp(-0.5*x^2)/sqrt(2*pi)};
      			\addplot[name path=axis, draw=none] {0};
      			\addplot[red!70, opacity=0.25] fill between[of=pdf and axis, soft clip={domain=1.5:4}];
      			\addplot[red!70, opacity=0.25] fill between[of=pdf and axis, soft clip={domain=-4:-1.5}];
      			\draw[densely dashed] (axis cs:0,0) -- (axis cs:0,0.42);
      			\draw[dashed] (axis cs:1.5,0) -- (axis cs:1.5,{exp(-0.5*1.5^2)/sqrt(2*pi)});
      			\draw[dashed] (axis cs:-1.5,0) -- (axis cs:-1.5,{exp(-0.5*1.5^2)/sqrt(2*pi)});
    			\end{axis}
  			\end{tikzpicture}
  		\caption{Chebyshev's inequality provides an upper bound on 
              		the tail \gls{probability} $\prob{|x-\mu|\ge \eta}$ (i.e., the shaded area) of
			a real-valued \gls{rv} $x$ with a finite second moment.\label{fig:chebyshev_minimal_dict}} 
		\end{figure}
     		See also: \gls{expectation}, \gls{markovsinequality}, \gls{concentrationinequ}.},
    	first={Chebyshev's inequality},
	type=math, 
     	text={Chebyshev's inequality}  
}

\newglossaryentry{hoeffdingsinequality}
{name={Hoeffding's inequality},
	description={\index{Hoeffding's inequality} Hoeffding's inequality \cite{Hoeffding1963} is a fundamental \gls{concentrationinequ} 
		that provides an upper bound on the \gls{probability} that a sum (or average) of independent, bounded \glspl{rv} 
		deviates from its \gls{mean} by more than some threshold. 
		Let $x_1,\,\dots,\,x_n$ be independent real-valued \glspl{rv} $x_i$ taking values in $[a_i,b_i] \subset \mathbb{R}$, 
		and $S_n \defeq \sum_{i=1}^n x_i$ and $\expect\{S_n\} = \sum_{i=1}^n \expect\{x_i\}$. 
		Then, Hoeffding's inequality \cite[Th. 2.2.6]{vershynin2018high} states that
        		\begin{equation}
                		\nonumber
			\prob{ |S_n - \expect\{S_n\}| \geq t } 
            		\leq 2 \exp \left( -\frac{2 t^2}{\sum\limits_{i=1}^n (b_i - a_i)^2} \right) \qquad \forall t > 0.
        		\end{equation}
    		Hoeffding's inequality typically provides sharper bounds than \gls{chebyshevsinequality}, 
		but it is more restrictive by assuming bounded \glspl{rv}.
    		In the context of \gls{ml}, Hoeffding's inequality can be used for the analysis of \gls{erm} or \gls{mab} methods.
		 \\
    		See also: \gls{concentrationinequ}, \gls{probability}, \gls{rv}, \gls{chebyshevsinequality}, \gls{expectation}, \gls{probspace}, 
		\gls{markovsinequality}.},
    	first={Hoeffding's inequality},
	type=math, 
    	text={Hoeffding's inequality}  
}

\newglossaryentry{rgg}
{name={random geometric graph (RGG)},
	description={An RGG\index{random geometric graph (RGG)} is a \gls{probmodel} for \glspl{graph} 
		built from nodes randomly placed in a \gls{metricspace}. Given a \gls{metricspace}, the RGG is characterized by 
		the number of nodes, the connection radius, and the \gls{probdist} describing the node placement. 
    		More precisely, for a node set $\nodeidx=1, \,\ldots, \,\nrnodes$, each node $\nodeidx$ is assigned to a random position 
		$\featurevec^{(\nodeidx)} \in \featurespace$, typically as \glspl{realization} of \gls{iid} \glspl{rv} taking values in a set $\featurespace$. 
		Together with a \gls{metric}, this forms a \gls{metricspace}, often with the \gls{metric} induced by a \gls{norm}. 
		A specific \gls{realization} of an RGG contains an edge $\edge{\nodeidx}{\nodeidx'}$ if and only if the \gls{distance} between the nodes 
		with respect to the \gls{metric} is smaller than some threshold, i.e., when $\metric{\featurevec^{(\nodeidx)}}{\featurevec^{(\nodeidx')}}\leq r$ 
		for some threshold $r>0$, as illustrated in Fig.\ \ref{fig:rgg}.
         	\begin{figure}[H]
    			\centering
    			\begin{tikzpicture}[x=0.6cm,y=0.6cm]
        				\draw (-5,-5) rectangle (5,5);
        				\node[below left] at (-5,-5) {\small $-5$};
        				\node[above left]  at (-5, 5) {\small $5$};
        				\node[below right] at ( 5,-5) {\small $5$};
        				\fill (-2.0,-3.5) circle (2pt);
        				\draw (-2.0,-3.5) circle (2.5);
        				\node[anchor=east] at (-2.0,-3.5) {\small 1};
        				\fill (-0.5,-3.3) circle (2pt);
        				\draw (-0.5,-3.3) circle (2.5);
        				\node[anchor=west] at (-0.5,-3.3) {\small 2};
        				\draw (-2.0,-3.5) -- (-0.5,-3.3);
        				\fill (3.0,3.5) circle (2pt);
        				\draw (3.0,3.5) circle (2.5);
        				\node[anchor=west] at (3.0,3.5) {\small 3};
    			\end{tikzpicture}
   		\caption{Illustration of an RGG with $ \featurespace = [-5,5]\times[-5,5] \subset \mathbb{R}^2$ and radius $r=2.5$ 
			(with respect to the \gls{eucliddist}), where nodes 1 and 2 (corresponding to $(-2.0, -3.5)^T$ and $(-0.5, -3.3)^T$, 
			within radius $r=2.5$) are connected, while node 3 (corresponding to $(3.0, 3.5)^T$) has no other node within that \gls{distance}.}
    			\label{fig:rgg}
    		\end{figure}
        		See also: \gls{graph}, \gls{sbm}, \gls{ergraph}.},
    	first={random geometric graph (RGG)},
	type=math, 
    	text={RGG}
}

\newglossaryentry{banachfixedpoint}
{name={Banach's fixed-point theorem},
	description={Banach's fixed-point theorem\index{Banach's fixed-point 
		theorem} (also referred to as the contraction principle \cite[Th. 9.23]{RudinBookPrinciplesMatheAnalysis},\cite{Kelley1995}) 
		states that every \gls{contractop} $\fixedpointop$ on a complete 
		\gls{metricspace} $\paramspace$ has a unique \gls{fixedpoint}. 
		Formally, let $(\paramspace,\metric{\cdot}{\cdot})$ be a non-empty 
		complete \gls{metricspace} and let $\fixedpointop:\paramspace\to\paramspace$ satisfy
    		\[
    		\metric{\fixedpointop \weights}{\fixedpointop \weights'} \leq \contractfac \cdot \metric{\weights}{\weights'} \qquad \forall \weights,\weights'\in\paramspace
    		\]
    		for some constant $\contractfac \in [0,1)$. Then, $\fixedpointop$ has a unique \gls{fixedpoint}, 
		i.e., there exists a unique $\weights^\star\in\paramspace$ with 
		$\fixedpointop \weights^\star =\weights^\star$. Moreover, for any initial 
		$\weights^{(0)}\in\paramspace$, the \gls{fixedpointiter} 
		$\weights^{(\iteridx)}\!=\!\fixedpointop \weights^{(\iteridx-1)}$, for $\iteridx=1,\,2,\,\dots$, 
		converges to $\weights^\star$ at a rate governed by $\contractfac$. 
		In particular, 
		\[ \metric{\weights^{(\iteridx)}}{\weights^\star} \leq \contractfac^{\iteridx}
		 \cdot \metric{\weights^{(0)}}{\weights^\star} \mbox{ \quad for }\iteridx=1,\,2,\,\dots. \]   		
		\begin{figure}[H]
    			\centering
    			\begin{tikzpicture}[>=Latex, font=\small]
    			\tikzset{space/.style={draw, thick, circle, minimum size=4.0cm},pt/.style={circle, inner sep=1.5pt, draw, fill=black},maparrow/.style={->, very thick},link/.style={->, thick},distline/.style={dashed, thick}}
    			\def\dx{6.5}
    			\node[space,label=above:{$\paramspace$}] (XL) at (0,0) {};
    			\node[space,label=above:{$\paramspace$}] (XR) at (\dx,0) {};
    			\draw[maparrow] (XL.east) -- node[above=2pt] {$\fixedpointop$} (XR.west);
    			\coordinate (x1) at (-0.8,0.9);
    			\coordinate (x2) at (0.9,0.5);
    			\node[pt,label=above left:{$\weights$}] at (x1) {};
    			\node[pt,label=right:{$\weights'$}] at (x2) {};
    			\coordinate (fx1) at (\dx-0.3,0.5);
    			\coordinate (fx2) at (\dx+0.6,0.5);
    			\node[pt] (FX1) at (fx1) {};
    			\node[pt] (FX2) at (fx2) {};
    			\node[anchor=east] at ($(FX1.west)+(-6pt,-2pt)$) {$\fixedpointop \weights$};
    			\node[anchor=west] at ($(FX2.east)+(6pt,-2pt)$) {$\fixedpointop \weights'$};
    			\coordinate (xsL) at (0,-0.9);
    			\coordinate (xsR) at (\dx,-0.9);
    			\node[pt,label=below:{$\weights^\star$}] at (xsL) {};
    			\node[pt,label=below:{$\fixedpointop \weights^\star\!=\!\weights^\star$}] at (xsR) {};
    			\end{tikzpicture}
   	 	\caption{\Gls{contractop} $\fixedpointop:\paramspace\to\paramspace$  
	  		with a unique \gls{fixedpoint} $\weights^\star$ satisfying 
			$\fixedpointop \weights^\star\!=\!\weights^\star$.}
    			\label{fig:banach_dict}
    		\end{figure}
    		See also: \gls{contractop}, \gls{fixedpointiter}.},
    	first={Banach's fixed-point theorem},
	text={Banach's fixed-point theorem},
    	type=math
}

\newglossaryentry{cvxoptproblem}
{name={convex optimization problem}, 
	description={See\index{convex optimization problem} \gls{convexopt}.},
	first={convex optimization problem},
	firstplural={convex optimization problems}, 
	type=math,
	plural={convex optimization problems}, 
	text={convex optimization problem}
}

\newglossaryentry{optimization}
{name={optimization}, 
	description={See\index{optimization} \gls{convexopt}, \gls{optmethod}, \gls{optproblem}.},
	first={optimization},
	firstplural={optimizations}, 
	type=math,
	plural={optimizations}, 
	text={optimization}
}

\newglossaryentry{optproblem}
{name={optimization problem}, 
	description={An\index{optimization problem} \gls{optimization} problem is a mathematical 
		   structure consisting of an \gls{objfunc} $f: \mathcal{U} \rightarrow \mathcal{V}$ 
		   defined over an \gls{optimization} variable $\weights \in \mathcal{U}$, together with a 
		   feasible set $\paramspace \subseteq \mathcal{U}$. The \gls{co-domain} $\mathcal{V}$ is 
		   assumed to be ordered, meaning that for any two elements $\mathbf{a}, \mathbf{b} \in \mathcal{V}$, 
		   we can determine whether $\mathbf{a} \leq \mathbf{b}$.
		   Here, ``$\leq$'' denotes a general partial order relation, which may differ 
		   from the intrinsic ordering of real numbers \cite[Sec. 2.4]{BoydConvexBook}. 
		   The goal in \gls{optimization} is to find those values $\widehat{\weights} \in \paramspace$ 
		   for which the objective $f(\weights)$ is extremal, i.e., either minimal or maximal 
		   \cite{BoydConvexBook}, \cite{nesterov04}, \cite{BertsekasNonLinProgr}.
		   \\
		   See also: \gls{objfunc}.},
	first={optimization problem},
	firstplural={optimization problems}, 
	type=math,
	plural={optimization problems}, 
	text={optimization problem}
}

\newglossaryentry{optmethod}
{name={optimization method},
	description={An\index{optimization method} \gls{optimization} method is 
		an \gls{algorithm} that takes a representation of an 
		\gls{optproblem} as input and computes an (approximate) 
		solution as its \gls{output}. A central example of an \gls{optproblem} 
		in \gls{ml} is \gls{erm}. By applying an appropriate 
		\gls{optimization} method to \gls{erm}, we obtain a concrete learning 
		\gls{algorithm} \cite{BoydConvexBook}, \cite{nesterov04}, 
		\cite{BertsekasNonLinProgr}.
		 \\
		 See also: \gls{algorithm}, \gls{optproblem}.},
	type=math, 
	first={optimization method},
	firstplural={optimization methods}, 
	plural={optimization methods}, 
	text={optimization method}
}

\newglossaryentry{convexopt}
{name={convex optimization},
 	description={\Gls{convex} \gls{optimization}\index{convex optimization} studies the 
		formulation, properties, and efficient solution methods for \gls{convex} \glspl{optproblem} \cite{BoydConvexBook}.  
		A \gls{convex} \gls{optproblem} (defined on the \gls{euclidspace} $\mathbb{R}^{\nrfeatures}$) 
		consists of a \gls{convex} \gls{objfunc} $f: \mathbb{R}^{\nrfeatures} \rightarrow \mathbb{R}$  
    		and a \gls{convex} constraint set $\mathcal{C}$ for the \gls{optimization} variable $\weights$.  
    		It can be written compactly as \cite{BoydConvexBook}
		$$ \min_{\weights \in \mathcal{C}}  f(\weights).$$ 
		Alternatively, a \gls{convex} \gls{optproblem} can be expressed in 
		terms of \gls{convex} constraint \glspl{function} $g_1,\,\ldots,\,g_k$ as
		\begin{align} 
			\min_{\weights \in \mathbb{R}^{\nrfeatures}} & f(\weights)   \nonumber \\ 
			\mbox{s.t.} \quad & g_{\sampleidx}(\weights) \leq 0, \qquad \sampleidx=1,\,\ldots,\,k. \label{equ_def_convx_opt_constr_dict}
		\end{align} 
		\begin{figure}[H]
			\centering
			\begin{tikzpicture}[>=stealth, scale=1.0]
  			\draw[->] (-3,0) -- (5.2,0) node[below] {$\vc$};
  			\draw[->] (0,-0.2) -- (0,4.2) node[left] {$t$};
  			\draw[thick, domain=-1:3, smooth, variable=\x, name path=boundary]
    			plot ({\x},{exp(-\x)});
  			\path[fill=gray!40, opacity=0.4]
    			(-1,3) -- (-1,{exp(1)}) -- plot[domain=-1:3] ({\x},{exp(-\x)})
   			-- (3,3) -- cycle;
  			\fill (0,1) circle (1.6pt) node[left] {$f^{\ast}$};
  			\draw[dashed, domain=-1:3, smooth, variable=\x]
    			plot ({\x},{(2/exp(1)) - (1/exp(1))*\x}); 
  			\fill (1,{1/exp(1)}) circle (1.2pt);
  			\node at (2.6,2.5) {$\mathcal{A}$};
  			\node [below,yshift=-3pt] at (0,-0.2) {$\mathbf{0}$};
  			\end{tikzpicture}
  		\caption{\Gls{convex} \gls{optproblem} \eqref{equ_def_convx_opt_constr_dict} represented 
  			by a set $\mathcal{A}$ that consists of objective values $t$ and constraint values 
  			$\vc=\big(c_{1},\,\ldots,\,c_{\nrfeatures}\big)^{T}$ that are achievable, i.e., 
  			$f(\weights) \leq t, g_{1}(\weights)\leq c_{1},\,\ldots, \,g_{k}(\weights) \leq c_{k}$ 
  			for some $\weights \in \mathbb{R}^{\nrfeatures}$. The optimal value $f^{\ast}$ of the 
  			\gls{optproblem} is the smallest $t$ for which $(\mathbf{0},t) \in \mathcal{A}$. }
		\end{figure}
		The formulation \eqref{equ_def_convx_opt_constr_dict} lends, in turn, to the 
		following \gls{epigraph} form \cite[Sec. 5.3]{BoydConvexBook}:
		$$\inf \big\{ t \in \mathbb{R} : (\mathbf{0}, t) \in \mathcal{A} \big\}$$ 
		with the set
		\begin{align} 
			\mathcal{A} \defeq 
    			\big\{ (\vc,t) & \in \mathbb{R}^{\nrfeatures} \times \mathbb{R} : 
    			f(\weights) \leq t, \, \nonumber \\ 
   			&  g_{\sampleidx}(\weights) \leq c_{\sampleidx}, \,
    			\sampleidx = 1,\,\ldots,\,k 
    			\text{ for some } \weights \in \mathbb{R}^{\nrfeatures} \big\}. \nonumber
		\end{align}
		It can be shown that, since $f,g_{1},\,\ldots,\,g_{k}$ are \gls{convex} \glspl{function}, 
    		$\mathcal{A}$ is a \gls{convex} set \cite[Ch. 2]{BoydConvexBook}.
		The set $\mathcal{A}$ fully characterizes the \gls{optproblem}~\eqref{equ_def_convx_opt_constr_dict} 
		and can be interpreted as the \gls{epigraph} of the \gls{objfunc}~$f$ over the 
		feasible region defined by the constraint \glspl{function} $g_1,\,\ldots,\,g_k$.
		\\ 
    		See also: \gls{convex}, \gls{optproblem}, \gls{optmethod}. },
	first={convex optimization},
	type=math,
  	text={convex optimization}
}

\newglossaryentry{newtonmethod}
{name={Newton's method},
	description={Newton's method\index{Newton's method} is an iterative \gls{optmethod} for finding 
		local \glspl{minimum} or \glspl{maximum} of a \gls{differentiable} \gls{objfunc} $f(\weights)$. 
		Like \glspl{gdmethod}, Newton's method also 
		computes a new estimate $\estparams{\iteridx+1}$ by optimizing a local approximation of $f(\weights)$ around 
		the current estimate $\estparams{\iteridx}$. In contrast to \glspl{gdmethod}, which use the \gls{gradient} to build 
		a local linear approximation, Newton's method uses the \gls{hessian} \gls{matrix} to build a local quadratic approximation. 
		In particular, starting from an initial estimate $\estparams{0}$, Newton's method iteratively updates the estimate as follows:
		\[
		\estparams{\iteridx+1}= \estparams{\iteridx}- \big( \nabla^2 f\big(\estparams{\iteridx}\big) \big)^{-1} \nabla f\big( \estparams{\iteridx} \big) \quad \mbox{for } \iteridx=0,\,1,\,\ldots.
		\]
		Here, $\nabla f\big(\estparams{\iteridx} \big)$ is the \gls{gradient}, and $\nabla^2 f(\weights^{(\iteridx)})$ is 
		the \gls{hessian} of the \gls{objfunc} $f$. Since using a \gls{quadfunc} as a local approximation is more accurate 
		than using a linear \gls{function} (which is a special case of a \gls{quadfunc}), Newton's method tends to 
		converge faster than \glspl{gdmethod} (see Fig.\ \ref{fig_newtonmethod_dict}). However, this faster \gls{convergence} comes at the increased computational 
		complexity of the \glspl{iteration}. Indeed, each \gls{iteration} of Newton's method requires the inversion of the \gls{hessian}. 
		\begin{figure}[H]
		\centering
		\begin{tikzpicture}[samples=200,smooth]
			\begin{scope}
				\clip(-5,-2) rectangle (5,5);
				\draw[thin] plot[domain=0:360] ({1.5*cos(\x)*sqrt(20/(sin(2*\x)+2))},{1.5*sin(\x)*sqrt(20/(sin(2*\x)+2))});
				\draw[thin] plot[domain=0:360] ({1.5*cos(\x)*sqrt(16/(sin(2*\x)+2))},{1.5*sin(\x)*sqrt(16/(sin(2*\x)+2))});
				\draw[thin] plot[domain=0:360] ({1.5*cos(\x)*sqrt(12/(sin(2*\x)+2))},{1.5*sin(\x)*sqrt(12/(sin(2*\x)+2))});
				\draw[thin] plot[domain=0:360] ({1.5*cos(\x)*sqrt(8/(sin(2*\x)+2))},{1.5*sin(\x)*sqrt(8/(sin(2*\x)+2))});
				\draw[thin] plot[domain=0:360] ({1.5*cos(\x)*sqrt(4/(sin(2*\x)+2))},{1.5*sin(\x)*sqrt(4/(sin(2*\x)+2))});
				\draw[thin] plot[domain=0:360] ({1.5*cos(\x)*sqrt(1/(sin(2*\x)+2))},{1.5*sin(\x)*sqrt(1/(sin(2*\x)+2))});
				\draw[thin] plot[domain=0:360] ({1.5*cos(\x)*sqrt(0.0625/(sin(2*\x)+2))},{1.5*sin(\x)*sqrt(0.0625/(sin(2*\x)+2))});
				\draw[->,blue,ultra thick] (-3.5,4.5) -- (-3.3,3.8);
				\draw[->,blue,ultra thick] (-3.3,3.8) -- (-3,3.1);
				\draw[->,blue,ultra thick] (-3,3.1) -- (-2.6,2.4);
				\draw[->,blue,ultra thick] (-2.6,2.4) -- (-2.1,1.8);
				\draw[->,blue,ultra thick] (-2.1,1.8) -- (-1.5,1.3);
				\draw[->,blue,ultra thick] (-1.5,1.3) -- (-0.9,0.9);
				\draw[->,blue,ultra thick] (-0.9,0.9) -- (-0.4,0.5);
				\draw[->,blue,ultra thick] (-0.4,0.5) -- (-0.1,0.2);
				\node[blue,above left] at (-3.5,4.5) {start};
				\node[blue,above] at (0.5,3) {\gls{gd}};
				\draw[->,red,ultra thick] (-3.5,4.5) -- (-2,2.5);
				\draw[->,red,ultra thick] (-2,2.5) -- (-0.8,0.8);
				\draw[->,red,ultra thick] (-0.8,0.8) -- (0,0.1);
				\node[red,above] at (1.5,4) {Newton's method};
				\node at (0,0) [circle,fill,inner sep=1pt,label=below:\gls{minimum}] {};
			\end{scope}
		\end{tikzpicture}
		\caption{Comparison of \gls{gd} (blue) and Newton's method (red) paths toward the \gls{minimum} of a 
			\gls{lossfunc}. \label{fig_newtonmethod_dict}}
		\end{figure}
		See also: \gls{optmethod}, \gls{gradient}, \gls{hessian}, \gls{gd}. },
  	first={Newton's method},
 	type=math,
  	text={Newton's method} 
}

\newglossaryentry{hilbertspace}
{name={Hilbert space},
	description={A\index{Hilbert space} Hilbert space $\hilbertspace$ is a complete 
    		\gls{innerproduct} space. Thus, $\hilbertspace$ is a \gls{vectorspace} 
		equipped with an \gls{innerproduct} $\innerprod{\cdot}{\cdot}$. The \gls{innerproduct} 
		induces a \gls{norm} $\normgeneric{\cdot}{\hilbertspace}$ via 
		$$\normgeneric{\weights}{\hilbertspace} = \sqrt{\innerprod{\weights}{\weights}}.$$ 
  		Furthermore, $\hilbertspace$ is complete, in the sense that every 
 		\gls{cauchysequence} $\big( \weights^{(\sampleidx)} \big)_{\sampleidx \in \mathbb{N}}$ 
		in $\hilbertspace$ converges to a limit $\lim_{\sampleidx\rightarrow \infty} \weights^{(\sampleidx)}$ 
		that is also contained in $\hilbertspace$. 
		\begin{figure}[H]
			\centering
			\begin{tikzpicture}[scale=3]
			\draw[gray!40] (0,0) circle (1);
			\def\ang{35} 
			\draw[->,thick] (0,0) -- (1,0) node[below right] {$\vu$};
			\draw[->,thick] (0,0) -- ({cos(\ang)},{sin(\ang)}) node[above] {$\vv$};
			\coordinate (P) at ({cos(\ang)},0);
			\draw[dashed] ({cos(\ang)},{sin(\ang)}) -- (P);
			\draw[->,thick] (0,0) -- (P) node[pos=0.5,below] {$\innerprod{\vv}{\vu} \vu$};
			\draw ($(P)+(0,0.06)$) -- (P) -- ($(P)+(-0.06,0)$);
			\node[right] at ({cos(-\ang)},{sin(-\ang)}) {$\sphere{\nrfeatures} = \{ \weights \in \mathbb{R}^{\nrfeatures}: \innerprod{\weights}{\weights}=1\}$};
			\end{tikzpicture}
		\caption{For two unit-\gls{norm} \glspl{vector} $\vu, \vv \in \sphere{\nrfeatures} \subseteq \mathbb{R}^{\nrfeatures}$, 
			the \gls{innerproduct} $\innerprod{\vu}{\vv}$ is the expansion coefficient for the \gls{projection} 
			of $\vv$ onto the \gls{subspace} $\{ c \vu: c \in \reals \}$ spanned by $\vu$. 
			The absolute value $|\innerprod{\vu}{\vv}|$ measures the \gls{norm} of this \gls{projection}.\label{fig_hilbertspace_dict}}
		\end{figure}
		One important example of a Hilbert space 
 		is the \gls{euclidspace} $\mathbb{R}^{\nrfeatures}$ with the 
 		\gls{innerproduct} $\innerprod{\weights}{\weights'} = \weights^{\top} \weights'$. 
		\\
   		See also: \gls{innerproduct}, \gls{vectorspace}.},
	type=math, 
	first={Hilbert space}, 
	text={Hilbert space},
	plural={Hilbert spaces},
	firstplural={Hilbert spaces}
}

\newglossaryentry{estimator} 
{name={estimator},
	description={\Glspl{mlsystem} are\index{estimator} often analyzed using a 
		\gls{probmodel} $\probdist^{(\featurevec;\weights)}$ for the 
		generation of \gls{data} $\featurevec$ \cite{MLBasics}. The true 
		value of the \glspl{modelparam} $\weights$ is typically unknown and needs 
		to be estimated. An estimator is a \gls{measurable} \gls{function} 
		$\hypothesis(\featurevec)$ that reads in the \gls{data} $\featurevec$ 
		and delivers an estimate (or approximation) $\widehat{\weights} = \hypothesis(\featurevec)$ 
		of the true value of the \glspl{modelparam} $\weights$ \cite{LC},  \cite{kay}.
		\\
		See also: \gls{modelparam}, \gls{measurable}, \gls{function}. },
	first={estimator},
	type=math,
	text={estimator},
	plural={estimators},
	firstplural={estimators}
}

\newglossaryentry{unbiasedest} 
{name={unbiased estimator},
	description={An \gls{estimator}\index{unbiased estimator} for the \glspl{parameter} of a \gls{probmodel} 
	    	is unbiased if its \gls{expectation} equals the true value of the \glspl{parameter} \cite{LC}, \cite{kay}.
		\\
		See also: \gls{estimator}, \gls{parameter}, \gls{expectation}. },
	first={unbiased estimator},
	type=math,
	text={unbiased estimator},
	plural={unbiased estimators},
	firstplural={unbiased estimators}
}

\newglossaryentry{rkhs}
{name={reproducing kernel Hilbert space (RKHS)},
	description={An\index{reproducing kernel Hilbert space (RKHS)} RKHS $\hilbertspace$ is a  
		special type of \gls{hilbertspace} that consists of 
 		\glspl{function} defined on a set $\featurespace$ \cite{aronszajn1950}. 
 		The defining property of an RKHS is that point evaluations are continuous 
 		linear functionals. Each RKHS is associated with a \gls{kernel} 
  		$$\kernel :\featurespace \times\featurespace\to\mathbb{R},$$ 
 		satisfying the reproducing property
 		\[
 			f(\featurevec)=\langle f, \kernel(\featurevec,\cdot)\rangle_{\hilbertspace}
 			\quad \text{for all } f\in\hilbertspace,\; \featurevec\in\featurespace .
 		\]
 		RKHSs arise naturally in the analysis of \glspl{kernelmethod}, 
 		including \gls{kernel}-\gls{ridgeregression}, \gls{svm}, and \gls{GaussProc} 
 		\gls{regression} \cite{Murphy2012}, \cite{LearningKernelsBook}. Another application of 
		RKHSs is for the analysis of \gls{minimum} \gls{variance} estimation \cite{RKHSExpFamIT2012}.
		\\
		See also: \gls{hilbertspace}, \gls{kernel}. }, 
 	first={reproducing kernel Hilbert space (RKHS)},
 	type=math,
 	text={RKHS}, 
 	plural={RKHSs},
 	firstplural={reproducing kernel Hilbert spaces (RKHSs)}
}

\newglossaryentry{cauchysequence}
{name={Cauchy sequence},
sort={Cauchysequence},
	description={A\index{Cauchy sequence} Cauchy \gls{sequence} is a \gls{sequence}
		$\big( \featurevec^{(\sampleidx)}\big)_{\sampleidx \in \mathbb{N}}$ 
		in a \gls{metricspace} $\pair{\featurespace}{\metric{\cdot}{\cdot}}$ such 
		that the elements $\featurevec^{(\sampleidx)}\in \featurespace$ 
		become arbitrarily close to each other eventually. 
		In other words \cite[Definition 3.8]{RudinBookPrinciplesMatheAnalysis}, 
		\[
		\forall \epsilon > 0, \exists N \in \mathbb{N} \text{ such that } 
		\forall \sampleidx, \sampleidx' \geq N, \ \metric{\featurevec^{(\sampleidx)}}{\featurevec^{(\sampleidx')}} < \epsilon.
		\] 
		Fig.\ \ref{fig:fpit_cauchy_sqrt2_dict} shows a Cauchy \gls{sequence} in the \gls{metricspace} $\pair{\mathbb{Q}}{|\cdot|}$ of 
		rational numbers.  
		\begin{figure}[H]
			\centering
		  	\begin{tikzpicture}[x=1cm,y=4cm]
			\def\srtwo{1.4142}
			\def\eps{0.15}
			\def\nmax{7}
			\fill[gray!30, opacity=0.35]
			(-0.3,{\srtwo-\eps}) rectangle (\nmax+0.6,{\srtwo+\eps});
			\draw[<->, thick] (4.5,{\srtwo-\eps}) -- (4.5,{\srtwo+\eps})
			node[above, right] {\footnotesize{$\varepsilon$}};
			\draw[dashed] (-0.3,\srtwo) -- (\nmax+0.6,\srtwo)
			node[right] {\footnotesize{$\sqrt{2}$}};
			\fill (0,1.0000) circle (1.2pt) node [right] {$\feature^{(1)}$};
			\fill (1,1.5000) circle (1.2pt);
			\fill (2,1.45) circle (1.2pt);
			\fill (3,1.39) circle (1.2pt);
			\fill (4,1.41) circle (1.2pt);
			\fill (5,1.4142136) circle (1.2pt);
			\fill (6,1.41421356) circle (1.2pt);
			\def\N{1}
			\draw[densely dashed, thick, red] (\N,0.88) -- (\N,1.655);
			\node[right, red] at (\N,0.8) {$N$};
			\node[align=right, anchor=north, font=\small]
			at (5,0.9)
			{\footnotesize{$\displaystyle \feature^{(\sampleidx+1)}=\frac{1}{2}\!\left(\feature^{(\sampleidx)}+\frac{2}{\feature^{(\sampleidx)}}\right),\, \feature^{(1)}=1.$}};
			\end{tikzpicture}
		\caption{Cauchy \gls{sequence} $\big(\feature^{(\sampleidx)}\big)_{\sampleidx\in\mathbb{N}}$ 
			in the \gls{metricspace} $\pair{\mathbb{Q}}{|\cdot|}$. 
			This \gls{sequence} is generated by a \gls{fixedpointiter} used to approximate 
			$\sqrt{2}$. For all $\sampleidx \ge N$, the \gls{sequence} elements lie within a band of width $\varepsilon$. 
			Note that the \gls{sequence} does not converge in $\mathbb{Q}$, 
			since $\sqrt{2} \notin \mathbb{Q}$ \cite[Example 1.1]{RudinBookPrinciplesMatheAnalysis}.\label{fig:fpit_cauchy_sqrt2_dict}}
		\end{figure}
		See also: \gls{sequence}, \gls{metricspace}, \gls{convergence}. }, 
	type=math,
	first={Cauchy sequence},
	text={Cauchy sequence},
	plural={Cauchy sequences},
	firstplural={Cauchy sequences}
}

\newglossaryentry{nonexpansiveop}
{name={non-expansive operator},
 	description={An\index{non-expansive operator} \gls{operator} 
		$\fixedpointop: \mathcal{V} \rightarrow \mathcal{V}$ defined on a 
		\gls{normedspace} $(\mathcal{V}, \normgeneric{\,\cdot \,}{})$ is called non-expansive if it does 
		not increase \glspl{distance}. In other words,
 		\begin{equation} 
 			\nonumber
 			\normgeneric{ \fixedpointop \weights - \fixedpointop \weights'}{} 
			\leq 	\normgeneric{\weights - \weights'}{} \quad \mbox{for any } \weights, \weights' \in \mathcal{V}. 
 		\end{equation} 
		Non-expansiveness is typically not sufficient to guarantee \gls{convergence} 
		of a \gls{fixedpointiter} that uses $\fixedpointop$ (see Fig.\ \ref{fig_nonexpansiveop_dict}).
		\begin{figure}[H]
			\centering
			\begin{tikzpicture}[thick,font=\small]
			\node[circle,fill=black,inner sep=1.6pt,label=below:{$\weights^\star$}] at (0,0) {};
			\draw[dashed] (0,0) circle[radius=2.7];
			\node[circle,fill=black,inner sep=1.6pt,label=right:{$\weights$}] at (2.3,1.4) {};
			\node[circle,fill=black,inner sep=1.6pt,label=below:{$\fixedpointop \weights$}] at (1.27,0.77) {};
			\node[circle,fill=black,inner sep=1.6pt,label=above:{$\fixedpointop' \weights$}] at (-0.23,2.68) {};
			\node[circle,fill=black,inner sep=1.6pt,label=below:{$\fixedpointop''\weights$}] at (1.03,2.04) {};
			\end{tikzpicture}
		\caption{Result of applying a \gls{contractop} $\fixedpointop$, a \gls{nonexpansiveop} 
		         $\fixedpointop'$, and a \gls{firmlynonexpansiveop} $\fixedpointop''$, all 
			sharing the \gls{fixedpoint} $\weights^\star$. \label{fig_nonexpansiveop_dict}}
       		\end{figure}
 		See also: \gls{fixedpointiter}, \gls{contractop}.},
 	type=math, 
	first={non-expansive operator},
 	plural={non-expansive operators},
	firstplural={non-expansive operators},
 	text={non-expansive operator}
}

\newglossaryentry{tv}
{name={total variation}, 
	description={See \gls{gtv}\index{total variation}.},
	first={total variation},
	type=math,
	text={total variation} 
}

\newglossaryentry{firmlynonexpansiveop}
{name={firmly non-expansive operator},
 	description={An\index{firmly non-expansive operator} \gls{operator} 
		$\fixedpointop: \hilbertspace \rightarrow \hilbertspace$ defined on a 
		\gls{hilbertspace} $\hilbertspace$ is called firmly non-expansive if it satisfies
 		\begin{equation} 
 			\nonumber
 			\normgeneric{ \fixedpointop \weights - \fixedpointop \weights'}{\hilbertspace}^2  
			\leq 	\innerprod{ \weights - \weights'}{ \fixedpointop \weights - \fixedpointop \weights'} 
			\quad \mbox{for any } \weights, \weights' \in \hilbertspace. 
 		\end{equation}
		By the \gls{cauchyschwarzinequ}, any firmly \gls{nonexpansiveop} is necessarily 
		also a \gls{nonexpansiveop}. 
		Every \gls{fixedpointiter} that uses a firmly \gls{nonexpansiveop} converges 
		to a \gls{fixedpoint} of the \gls{operator} \cite{BausckeCombette}.
		\\ 
 		See also: \gls{fixedpointiter}, \gls{contractop}.},
 	type=math, 
	first={firmly non-expansive operator},
 	plural={firmly non-expansive operators},
	firstplural={firmly non-expansive operators},
 	text={firmly non-expansive operator}
}

\newglossaryentry{fixedpointiter}
{name={fixed-point iteration},
	description={A\index{fixed-point iteration} fixed-point \gls{iteration} is an iterative method 
	     	for solving a \gls{fixedpointeq}, which, in an \gls{ml} context, often arises from an \gls{optproblem}. 
		In case of \glspl{operator} $\fixedpointop: \mathbb{R}^\featuredim \to \mathbb{R}^\featuredim$,
		it constructs a \gls{sequence} $\weights^{(0)}, \,\weights^{(1)}, \,\ldots$ in $\mathbb{R}^\featuredim$ by 
		repeatedly applying an \gls{operator} $\fixedpointop$, i.e.,
		\begin{equation} 
		 	\label{equ_def_fixed_point_dict} 
		 	\weights^{(\iteridx+1)} = \fixedpointop \weights^{(\iteridx)} \quad \mbox{for } \iteridx=0, \,1, \,\ldots.
		\end{equation} 
		The \gls{operator} $\fixedpointop$ is chosen such that any of its \glspl{fixedpoint} is a solution 
		$\widehat{\weights}$ to the given \gls{optproblem}. For example, given a \gls{differentiable} and 
		\gls{convex} \gls{function} $f(\weights)$, the \glspl{fixedpoint} of the \gls{operator} $\fixedpointop: \weights \mapsto \weights - \nabla f(\weights)$ 
		coincide with the minimizers of $f(\weights)$. In general, for a given \gls{optproblem} 
		with solution $\widehat{\weights}$, there are many different \glspl{operator} 
		$\fixedpointop$ whose \glspl{fixedpoint} are $\widehat{\weights}$. Clearly, we should use an 
		\gls{operator} $\fixedpointop$ in \eqref{equ_def_fixed_point_dict} that reduces the 
		distance (with respect to the \gls{euclidnorm} or another \gls{norm}) to a solution such that
		\begin{equation} 
			\nonumber
			\underbrace{\normgeneric{ \weights^{(\iteridx+1)} - \widehat{\weights}}{2}}_{\stackrel{\eqref{equ_def_fixed_point_dict}}{=} \normgeneric{ \fixedpointop \weights^{(\iteridx)} - \fixedpointop\widehat{\weights}}{2}}  \leq 	\normgeneric{ \weights^{(\iteridx)} - \widehat{\weights}}{2}. 
		\end{equation}
		Thus, we require $\fixedpointop$ to be at least a \gls{nonexpansiveop}, i.e., the \gls{iteration} \eqref{equ_def_fixed_point_dict} 
		should not result in worse \glspl{modelparam} that have a larger distance to a solution $\widehat{\weights}$. 
		Furthermore, each \gls{iteration} \eqref{equ_def_fixed_point_dict} should also make some progress, i.e., 
		reduce the distance to a solution $\widehat{\weights}$. This requirement can be made precise using 
		the notion of a \gls{contractop} \cite{Bauschke:2017}, \cite{fixedpoinIsta}. 
		The \gls{operator} $\fixedpointop$ is a \gls{contractop} (with respect to the \gls{euclidnorm}) if, 
		for some $\contractfac \in [0,1)$,
		\begin{equation} 
			\nonumber
			\normgeneric{ \fixedpointop \weights\!-\!\fixedpointop \weights'}{2}  \leq  \contractfac	\normgeneric{\weights\!-\!\weights'}{2} \quad \mbox{holds for any } \weights,\weights'.
		\end{equation}
		For a \gls{contractop} $\fixedpointop$, the fixed-point \gls{iteration} \eqref{equ_def_fixed_point_dict} generates 
		a \gls{sequence} $\weights^{(\iteridx)}$ that converges quite rapidly. In particular \cite[Th. 9.23]{RudinBookPrinciplesMatheAnalysis}, 
		\begin{equation} 
			\nonumber
			\normgeneric{ \weights^{(\iteridx)} - \widehat{\weights}}{2} \leq \contractfac^{\iteridx} 	\normgeneric{ \weights^{(0)} - \widehat{\weights}}{2}. 
		\end{equation} 
		Here, $\normgeneric{ \weights^{(0)} - \widehat{\weights}}{2}$ is the distance between 
		the initialization $\weights^{(0)}$ and the solution $\widehat{\weights}$. 
		It turns out that a fixed-point \gls{iteration} \eqref{equ_def_fixed_point_dict} with a 
		\gls{firmlynonexpansiveop} $\fixedpointop$ is guaranteed to converge to a 
		\gls{fixedpoint} of $\fixedpointop$ \cite[Corollary 5.16]{Bauschke:2017}. 
		Fig.\ \ref{fig_examples_nonexp_dict} depicts examples of a \gls{nonexpansiveop}, 
		a \gls{firmlynonexpansiveop}, and a \gls{contractop}. All of these \glspl{operator} are 
		defined on the one-dimensional (1-D) space $\mathbb{R}$. Another example of a \gls{firmlynonexpansiveop} 
		is the \gls{proxop} of a \gls{convex} \gls{function} \cite{ProximalMethods}, \cite{Bauschke:2017}. 
		\definecolor{darkgreen}{rgb}{0.0, 0.5, 0.0}
		\begin{figure}[H]
			\begin{center} 
				\begin{tikzpicture}[scale=1.5]
					\draw[line width=1pt, ->] (-2,0) -- (2,0) node[right] {$\weight^{(\iteridx)}$};
					\draw[line width=1pt, ->] (0,-2) -- (0,2) node[above] {$\weight^{(\iteridx+1)}$};
					\node at (2.1,2.2) {$\fixedpointop^{(3)}$};
					\node at (1.9,-1.5) {$\fixedpointop^{(1)}$};
					\node at (1.5,1.2) {$\fixedpointop^{(2)}$};
					\draw[dashed] (1,-2) -- (1,2); 
					\draw[dashed] (-2,1) -- (2,1); 
					\draw[dashed] (-2,-1) -- (2,-1); 
					\draw[dashed] (-1,-2) -- (-1,2); 
					\node[above,xshift=4pt,yshift=-1pt] at (1,0) {$1$};
					\node[above,xshift=8pt,yshift=-1pt] at (0,-1) {$-1$};
					\draw[line width=2,domain=-2:2,smooth,blue] plot(\x,{0.5*\x + 1});
					\draw[line width=2,domain=-2:2,smooth,red] plot(\x,{-\x});
					\draw[line width=2, domain=-2:-1,smooth,darkgreen] plot(\x,{-1});
					\draw[line width=2,domain=-1:1,smooth,darkgreen] plot(\x,{\x});
					\draw[line width=2,domain=1:2,smooth,darkgreen] plot(\x,{1});
				\end{tikzpicture}
			\end{center} 
		\caption{Examples of a \gls{nonexpansiveop} $\fixedpointop^{(1)}$, a \gls{firmlynonexpansiveop} 
			$\fixedpointop^{(2)}$, and a \gls{contractop} $\fixedpointop^{(3)}$. \label{fig_examples_nonexp_dict}}
		\end{figure} 
		See also: \gls{contractop}, \gls{proxop}, \gls{bellmanoperator}, \gls{banachfixedpoint}, \gls{policyevaluation}, 
		\gls{valueiteration}.},
	first={fixed-point iteration},
	text={fixed-point iteration},
	type=math, 
	firstplural={fixed-point iterations}, 
	plural={fixed-point iterations}
}

\newglossaryentry{graphoffunction}
{name={graph of a function},
	description={Given\index{graph of a function} a \gls{function} $f\!:\!\featurespace \to \labelspace$
		with \gls{domain} $\featurespace$ and \gls{co-domain} $\labelspace$,
		the graph of $f$ is the subset of $\featurespace \times \labelspace$ 
		defined as
		\[
		\{ (\featurevec,f(\featurevec)) : \featurevec \in \featurespace \,\}.
		\]
		The graph of a \gls{function} provides a geometric representation 
		that is widely used in analysis, topology, and \gls{optimization}
		\cite{MunkresTopology2000}, \cite{RockafellarBook}.
		\\
		See also: \gls{function}, \gls{epigraph}. }, 
	first={graph},
	text={graph},
	type=math, 
	firstplural={graphs},
	plural={graphs}
}

\newglossaryentry{epigraph}
{name={epigraph},
	description={The epigraph\index{epigraph} of a real-valued \gls{function} 
		$f : \mathbb{R}^n \to \mathbb{R} \cup \{+\infty\}$ 
  		is the set of points lying on or above its \gls{graphoffunction} 
		(see Fig.\ \ref{fig_epigraph_dict}), i.e., 
		\[
		\operatorname{epi}(f) = \left\{ (\mathbf{x}, t) \in \mathbb{R}^n \times \mathbb{R} \,\middle|\, f(\mathbf{x}) \leq t \right\}.
		\]
		A \gls{function} is \gls{convex} if and only if its epigraph is a 
		\gls{convex} set \cite{BoydConvexBook}, \cite{BertCvxAnalOpt}.
		\begin{figure}[H]
			\centering
			\begin{tikzpicture}[scale=1.0]
				\begin{axis}[
					axis lines = middle,
					xlabel = $x$,
					ylabel = {},
					xmin=-2, xmax=2,
					ymin=0, ymax=4.5,
					samples=100,
					domain=-1.5:1.5,
					thick,
					width=8cm,
					height=6cm,
					grid=none,
					axis on top,
					]
					\addplot [blue, thick, domain=-1.5:1.5] {x^2} node [pos=0.85, anchor=south west, xshift=5pt] {$f(\weights)$};
					\addplot [
					name path=f,
					draw=none,
					ytick=\empty,
					domain=-1.5:1.5,
					] {x^2};
					\path[name path=top] (axis cs:-1.5,4) -- (axis cs:1.5,4);
					\addplot [
					blue!20,
					opacity=0.6,
					draw=none,
					] fill between [
					of=f and top,
					soft clip={domain=-1.5:1.5},
					];
					    \node[font=\small] at (axis cs:-1.0,2.3) {$\operatorname{epi}(f)$};
				\end{axis}
			\end{tikzpicture}
		\caption{Epigraph of the \gls{function} $f(x) = x^2$ (i.e., the shaded area). \label{fig_epigraph_dict}}
		\end{figure}
		See also: \gls{function}, \gls{convex}.},
	first={epigraph},
	text={epigraph},
	type=math,
	firstplural={epigraphs},
	plural={epigraphs}
}

\newglossaryentry{nullspace}
{name={nullspace},
	description={The nullspace\index{nullspace} of a \gls{matrix} $\mA \in \mathbb{R}^{\nrfeatures' \times \nrfeatures}$, 
		denoted by $\nullspace{\mA}$, is the set of all \glspl{vector} $\mathbf{n} \in \mathbb{R}^\nrfeatures$ 
    		such that $$\mA \mathbf{n} = \mathbf{0}.$$ 
		Consider a \gls{featlearn} method that uses the \gls{matrix} $\mA$ to transform 
		a \gls{featurevec} $\mathbf{x} \in \mathbb{R}^{\nrfeatures}$ of a \gls{datapoint} 
		into a new \gls{featurevec} $\vz = \mA \mathbf{x} \in \mathbb{R}^{\nrfeatures'}$. 
		The nullspace $\nullspace{\mA}$ characterizes all directions in the original 
    		\gls{featurespace} $\mathbb{R}^{\nrfeatures}$ along which the transformation 
		$\mA \mathbf{x}$ remains unchanged. In other words, adding any \gls{vector} from 
		the nullspace to a \gls{featurevec} $\featurevec$ does not affect the transformed 
		representation $\vz$. This property can be exploited to enforce invariances in the 
		\glspl{prediction} (computed from $\mA \mathbf{x}$). Fig.\ \ref{fig:nullspace-rotation-dict} 
		illustrates one such invariance. It shows rotated versions of two handwritten digits, 
		which approximately lie along 1-D curves in the original \gls{featurespace}. 
		These curves are aligned with a direction \gls{vector} $\mathbf{n} \in \mathbb{R}^{\nrfeatures}$. 
    		To ensure that the trained \gls{model} is invariant to such rotations, we can 
		choose the transformation \gls{matrix} $\mA$ such that $\mathbf{n} \in \nullspace{\mA}$. 
		This ensures that $\mA \mathbf{x}$, and hence the resulting \gls{prediction}, 
		is approximately insensitive to rotations of the input image.
		\begin{figure}[H]
      			\centering
      			\includegraphics[width=0.6\textwidth]{assets/pythonsnacks/nullspace/nullspace_0_1.png}
	  		\caption{
			Rotated handwritings of two different digits. The rotations are approximately 
			aligned along straight lines parallel to the \gls{vector} $\mathbf{n}$. For a 
			binary \gls{classifier} distinguishing between these digits, a natural choice is 
			a linear \gls{featuremap} $\featurevec \mapsto \mathbf{A}\featurevec$ with a 
			\gls{matrix} $\mA$ whose nullspace contains $\mathbf{n}$, i.e., $\mathbf{n} \in \mathrm{null}(\mA)$.
                \label{fig:nullspace-rotation-dict}}	
	       	\end{figure}
		See also: \gls{matrix}, \gls{featuremap}, \gls{featlearn}. \\ 
		Python demo: \href{https://github.com/AaltoDictionaryofML/AaltoDictionaryofML.github.io/blob/main/assets/pythonsnacks/nullspace/nullspace.py}{click me}},
 	first={nullspace},
 	firstplural={nullspaces},
	type=math,
 	plural={nullspaces},
 	text={nullspace}
}

\newglossaryentry{fixedpoint}
{name={fixed point},
sort={fixedpoint},
	description={Consider some \gls{operator} $\fixedpointop: \hilbertspace \rightarrow \hilbertspace$ 
		defined on a \gls{hilbertspace} $\hilbertspace$. A 
		\gls{vector} $\widehat{\weights} \in \hilbertspace$ 
		is called a fixed point\index{fixed point} of the \gls{operator} $\fixedpointop$ 
		if it satisfies
		\[
			\fixedpointop \widehat{\weights} = \widehat{\weights}.
		\]
		In other words, applying the \gls{operator} $\fixedpointop$ to its fixed 
		point $\widehat{\weights}$ returns the same \gls{vector} $\widehat{\weights}$. 
		Finding a fixed point of a suitable \gls{operator} $\fixedpointop$ is a common 
		approach to solving various \glspl{optproblem} (e.g., an instance of \gls{erm}). 
		A popular method for computing approximations of a fixed point is the \gls{fixedpointiter}.
		\\
		See also: \gls{fixedpointiter}.},
	first={fixed point},
	text={fixed point},
	type=math, 
	firstplural={fixed points}, 
	plural={fixed points}
}

\newglossaryentry{fixedpointeq}
{name={fixed-point equation},
	description={A fixed-point equation\index{fixed-point equation} is an equation of the following form:
		\[
		\weights = \fixedpointop(\weights),
		\]
		where $\fixedpointop:\hilbertspace \to \hilbertspace$ is an \gls{operator} defined 
		on a \gls{hilbertspace} $\hilbertspace$.  
		Solving a fixed-point equation amounts to finding the \glspl{fixedpoint} of 
		$\fixedpointop$. Many \glspl{optproblem}, including instances of \gls{erm}, 
		can be cast in this form. For example, minimizing a \gls{smooth} \gls{convex} 
		\gls{function} $f$ is equivalent to solving the following fixed-point equation:
		$$\weights = \fixedpointop \weights \quad \mbox{with }  
		\fixedpointop: \weights \mapsto \weights - \eta \nabla f(\weights).$$
		Here, $\lrate>0$ can be choosen freely. The above fixed-point 
		equation is nothing but the \gls{zerogradientcondition} for the 
		minimizer of $f$ \cite{BoydConvexBook}. Similarly, one can reformulate the
		optimality conditions (i.e., the \gls{kktconditions}) of \glspl{optproblem} with constraints as a 
		fixed-point equation \cite{BoydConvexBook}, \cite{pock_chambolle_2016}.
		\\
		See also: \gls{fixedpoint}, \gls{optproblem}. }, 
	type=math, 
	first={fixed-point equation},
 	text={fixed-point equation}, 
	plural={fixed-point equations}, 
	firstplural={fixed-point equations}
}

\newglossaryentry{diagonalizable}
{name={diagonalizable},
	description={A square \gls{matrix} $\mA \in \mathbb{C}^{\nrfeatures \times \nrfeatures}$ 
		is called diagonalizable\index{diagonalizable} if it is similar  
 		to a diagonal \gls{matrix} \cite{HornMatAnalysis}, \cite{Axler2025}. 
 		Formally, $\mA$ is diagonalizable if there exists an \gls{invertible} \gls{matrix} 
		$\mP \in \mathbb{C}^{\nrfeatures \times \nrfeatures}$ such that 
  		\[
  			\mA = \mP \mD \mP^{-1},
  		\]
		where $\mD \in \mathbb{C}^{\nrfeatures \times \nrfeatures}$ is a 
 		diagonal \gls{matrix} whose main diagonal entries are the \glspl{eigenvalue} 
 		of $\mA$. A \gls{matrix} $\mA \in \mathbb{R}^{\nrfeatures \times \nrfeatures}$ 
 		is diagonalizable if and only if it has $\nrfeatures$ \gls{linearlyindep}
 		\glspl{eigenvector} \cite{HornMatAnalysis}. 
 		\\	
 		See also: \gls{matrix}, \gls{eigenvalue}, \gls{evd}.},
 	first={diagonalizable},
 	type=math,
 	text={diagonalizable}
}

\newglossaryentry{schurdecomp}
{name={Schur decomposition},
	description={Every square \gls{matrix} $\mA \in \mathbb{C}^{\nrfeatures \times \nrfeatures}$ 
		admits a Schur decomposition\index{Schur decomposition} as follows \cite[Th. 7.1.3]{GolubVanLoanBook}:
		\[
		\mA = \mU \mT \mU^{H}. 
		\]
		Here, $\mU \in \mathbb{C}^{\nrfeatures \times \nrfeatures}$ is a \gls{unitary} 
		(i.e., $\mU^{H}\mU = \mI$) and $\mT$ is upper triangular 
		with the \glspl{eigenvalue} of $\mA$ on its diagonal.
		Carefully note that the Schur decomposition exists also 
		for a \gls{matrix} $\mA$ that is not \gls{diagonalizable}.
		The identity
		\[
			\big(\mU^{(1)}\big)^{H} \mA \mU^{(1)}
			=
			\begin{pmatrix}
			\big(\vx^{(1)}\big)^{H} \\
			\big(\mX^{(2)}\big)^{H}
			\end{pmatrix}
			\mA
			\begin{pmatrix}
			\vx^{(1)} & \mX^{(2)}
			\end{pmatrix}
			=
			\begin{pmatrix}
			\eigval{1} & \vb^{H} \\
			0 & \mA^{(1)}
			\end{pmatrix}
		\]
		represents the first step in the construction of the Schur decomposition.
		Every \gls{matrix} $\mA \in \mathbb{C}^{n\times n}$ has at least one 
		\gls{eigenvalue} $\eigval{1}$ with a unit-\gls{norm} \gls{eigenvector} 
		$\vx^{(1)}$, $\mA \vx^{(1)} = \eigval{1} \vx^{(1)}$. This 
		\gls{eigenvector} allows us to decompose $\mA$ as shown above. Here, 
		we extend $\vx^{(1)}$ to an orthonormal \gls{basis}
		$\mQ^{(1)} = \big( \vx^{(1)} \,\big|\, \mX^{(2)} \big)$
		and use $\vb^{H} := \big(\vx^{(1)}\big)^{H} \mA \mX^{(2)}$ and 
		$\mA^{(1)} := \big(\mX^{(2)}\big)^{H} \mA \mX^{(2)}$.
		Applying the same construction recursively to $\mA^{(1)}$ yields the 
		Schur decomposition.
		\\
 		See also: \gls{matrix}, \gls{eigenvalue}, \gls{evd}.},
	first={Schur decomposition},
	type=math,
	text={Schur decomposition}
}

\newglossaryentry{unitary}
{name={unitary matrix},
	description={A square \gls{matrix} $\mU \in \mathbb{C}^{\nrfeatures \times \nrfeatures}$ 
		is called unitary\index{unitary matrix} if its \gls{conjugatetranspose} 
		(or \gls{hermitian} \gls{transpose}) $\mU^{H}$ is also its \gls{inverse}, i.e., if
 		\[
 			\mU^{H} \mU = \mU \mU^{H} = \mI.
 		\]
 		Equivalently, the columns (and rows) of a unitary \gls{matrix} form an 
		orthonormal \gls{basis} of $\mathbb{C}^{\nrfeatures}$ with respect to the standard 
		\gls{innerproduct} \cite{HornMatAnalysis}, \cite{Axler2025}. 
 		\\
 		See also: \gls{matrix}.},
 	first={unitary matrix},
 	type=math,
 	text={unitary matrix}
}

\newglossaryentry{innerproduct}
{name={inner product},
	description={Consider a \gls{vectorspace} $\featurespace$ over the 
		\gls{field} $\mathbb{F}$, where $\mathbb{F}$ is either the \gls{field} of 
		real numbers $\mathbb{R}$ or the \gls{field} of complex numbers $\mathbb{C}$. 
		An inner product\index{inner product} in $\featurespace$ is 
		a \gls{function} 
 		\[
 			\innerprod{\cdot}{\cdot}: \featurespace \times \featurespace \to \mathbb{F}
 		\]
 		that satisfies the following properties for all 
		$\featurevec, \featurevec', \featurevec'' \in \featurespace$ 
		and all scalars $\alpha \in \mathbb{F}$ \cite{Axler2025}:
 		\begin{itemize}
 			\item Conjugate symmetry: $\innerprod{\featurevec}{\featurevec'} = \overline{\innerprod{\featurevec'}{\featurevec}}$;
 			\item Linearity in the first argument: $\innerprod{\alpha \featurevec + \featurevec''}{\featurevec'} = \alpha \innerprod{\featurevec}{\featurevec'} + \innerprod{\featurevec''}{\featurevec'}$;
 			\item Positive-definiteness: $\innerprod{\featurevec}{\featurevec} \geq 0$, with equality if and only if $\featurevec = \mathbf{0}$.
 		\end{itemize}
 		The pair $(\featurespace, \innerprod{\cdot}{\cdot})$ is called an inner product space. 
		Each inner product induces a \gls{norm} via 
		$\norm{\featurevec} \defeq \sqrt{\innerprod{\featurevec}{\featurevec}}$ for all 
		$\featurevec \in \featurespace$, which in turn induces a \gls{metric} via 
		$\metric{\featurevec}{\featurevec'} \defeq \norm{\featurevec - \featurevec'}$
		for all $\featurevec, \featurevec' \in \featurespace$.
 		\\
 		See also: \gls{field}, \gls{norm}, \gls{vector}, \gls{hilbertspace}, \gls{orthogonalitycondition}, \gls{metricspace}.},
 	first={inner product},
 	type=math,
 	text={inner product}
}

\newglossaryentry{trace}
{name={trace},
 	description={The\index{trace} trace $\tr{\mA}$ of a square \gls{matrix} 
		$\mA \in \mathbb{R}^{\nrfeatures \times \nrfeatures}$ is the 
		sum of its diagonal entries \cite{StrangLinAlg2016}. 
		Formally, it is the following \gls{linearmap}:
		$$\tr{\cdot}: \mathbb{R}^{\nrfeatures \times \nrfeatures} 
		 \rightarrow \mathbb{R}: \mA \mapsto \sum_{\featureidx=1}^{\nrfeatures} A_{\featureidx,\featureidx}.$$     
    	 	It satisfies the cyclic property $\tr{\mA\mB}=\tr{\mB\mA}$ for any \glspl{matrix} 
    	 	$\mA,\mB \in \mathbb{R}^{\nrfeatures \times \nrfeatures}$ \cite[p. 301]{Axler2025}.
		\begin{figure}[H]
			\begin{center}
			\begin{tikzpicture}[font=\small, every node/.style={inner sep=1pt}]
				\node (a11) at (0,0)   {$A_{1,1}$};
				\node (a12) at (1,0)   {$A_{1,2}$};
				\node (a13) at (2,0)   {$A_{1,3}$};
				\node (a21) at (0,-1)  {$A_{2,1}$};
				\node (a22) at (1,-1)  {$A_{2,2}$};
				\node (a23) at (2,-1)  {$A_{2,3}$};
				\node (a31) at (0,-2)  {$A_{3,1}$};
				\node (a32) at (1,-2)  {$A_{3,2}$};
				\node (a33) at (2,-2)  {$A_{3,3}$};
				\draw[black] (-0.4,0.4) rectangle (2.4,-2.4);
				\coordinate (C) at ($(a11)!0.5!(a33)$);
				\draw[thick] (C) ellipse [x radius=2.1cm, y radius=0.35cm,rotate=-45];
			\end{tikzpicture}
			\end{center}
		\caption{Trace $\tr{\mA}$ of a $3 \times 3$ \gls{matrix} $\mA \in \mathbb{R}^{3\times 3}$ as 
			the sum of three main diagonal entries $A_{1,1}, \,A_{2,2}, \,A_{3,3}$.}
		\end{figure}
		Furthermore, if $\mA$ has \glspl{eigenvalue} $\eigval{1},\,\ldots,\,\eigval{\nrfeatures}$ 
		(each repeated according to its algebraic multiplicity), then
		\[
		\tr{\mA} = \sum_{\featureidx=1}^{\nrfeatures} \eigval{\featureidx}.
		\]
		This identity follows from the invariance of the trace under similarity 
		transformations \cite[Ch.~10]{Axler2025}.
		\\
    		See also: \gls{matrix}, \gls{eigenvalue}.},
    first={trace},
    type=math,
    text={trace}
}
 
\newglossaryentry{stddev}
{name={standard deviation},
	description={The\index{standard deviation} standard deviation of a 
               	real-valued \gls{rv} $\feature$ is defined as the square 
		root of its \gls{variance}, i.e., 
		$\sqrt{\expect\big\{ \big( \feature - \expect\{\feature \} \big)^{2} \big\}}$. 
		\\
    		See also: \gls{rv}, \gls{variance}, \gls{expectation}.},
    first={standard deviation},
    type=math,
    text={standard deviation}
}

\newglossaryentry{sequence}
{name={sequence},
	description={A sequence\index{sequence} is an ordered collection 
		of values from a set $\mathcal{A}$. For example, a sequence of 
	   	values from the set $\mathcal{A} = \{ \star, \otimes \}$ could be 
	   	$$ a = \big( \star, \,\otimes, \,\star, \,\star, \,\otimes, \,\ldots \big). $$
	   	Formally, a sequence $a$ is a \gls{function} \cite{RudinBookPrinciplesMatheAnalysis} 
        		\[
            	a: \mathbb{N} \rightarrow \mathcal{A}: \sampleidx \mapsto a_\sampleidx.
        		\]        
		We denote a sequence by $\big( a_{\sampleidx} \big)_{\sampleidx \in \mathbb{N}}$ 
		or $\big( a^{(\sampleidx)} \big)_{\sampleidx \in \mathbb{N}}$. Sometimes we also 
		use the notation $\big\{ a^{(\sampleidx)} \big\}_{\sampleidx \in \mathbb{N}}$. 
        		Note that the same value $a \in \mathcal{A}$ can appear multiple times 
		in the sequence at different positions $\sampleidx$. 		
        		Sequences are fundamental for the study of \gls{ml} methods, 
        		for instance when describing successive iterates  
        		$\{\weights^{(\iteridx)}\}_{\iteridx \in \mathbb{N}}$ of an iterative 
		\gls{algorithm} updating a \gls{parameter} \gls{vector}. 
		We can also use a sequence to represent an infinite \gls{dataset} 
		$$\dataset = \big\{ \pair{\featurevec^{(1)}}{\truelabel^{(1)}},\,\pair{\featurevec^{(2)}}{\truelabel^{(2)}},\,\ldots \big\}.$$ 
		See also: \gls{function}, \gls{dataset}, \gls{convergence}, \gls{cauchysequence}. }, 
	first={sequence},
	text={sequence},
	type=math, 
	firstplural={sequences},
	plural={sequences}
}

\newglossaryentry{convergence}
{name={convergence},
	description={Consider a \gls{sequence}\index{convergence} $\big( a_{\sampleidx} \big)_{\sampleidx \in \mathbb{N}}$ 
		with numeric values $a_{\sampleidx} \in \mathbb{R}$. This \gls{sequence} 
       	 	is said to converge to a value $a^\star$ if the values $a_{\sampleidx}$ become 
		arbitrarily close to $a^\star$ for sufficiently large indices $\sampleidx$. 
        		Mathematically speaking, the \gls{sequence} converges to $a^\star$ if 
		\cite{RudinBook}, \cite{RudinBookPrinciplesMatheAnalysis}
        		\[
           	\forall \epsilon > 0, \; \exists N \in \mathbb{N} : \sampleidx > N \Rightarrow |a_{\sampleidx} - a^\star| < \epsilon.
        		\]
		We denote the convergence of a \gls{sequence} to $a^\star$ 
		by 
		\[
			\lim_{\sampleidx \to \infty} a_{\sampleidx} = a^\star.
		\]
		\begin{figure}[H] 
			\centering
			\begin{tikzpicture}[x=1.2cm, y=2cm, >=stealth]
			\draw[->] (0.5,0) -- (6.5,0) node[right] {$\sampleidx$};
			\draw[->] (0.5,0) -- (0.5,1.6) node[above] {$a_\sampleidx$};
  			\def\eps{0.3}
  			\fill[gray!30, opacity=0.4] (0, {1-\eps}) rectangle (6.3, {1+\eps});
  			\draw[dotted] (0,{1+\eps}) -- (6.3,{1+\eps}) node[right] {$1+\varepsilon$};
  			\draw[dotted] (0,{1-\eps}) -- (6.3,{1-\eps}) node[right] {$1-\varepsilon$};
			\draw[dashed] (0,1) -- (6.3,1) node[right] {$\lim_{\sampleidx \to \infty} a_\sampleidx = 1$};
			\foreach \t in {1,...,6} {
			\pgfmathsetmacro{\at}{1 - 0.6^(\t)}
			\fill (\t,\at) circle (2pt);
			}
 		     	\foreach \t in {1,2,3} {
    		    	\draw (\t,0.02) -- (\t,-0.02) node[below] {$\t$};
  		    	}
  			\def\N{3}
  			\draw[densely dashed, thick, red] (\N,0) -- (\N,1.7);
  			\node[above] at (\N,1.7) {$N$};
			\end{tikzpicture}
		\caption{Real-valued \gls{sequence} $\big( a_{\sampleidx} \big)_{\sampleidx \in \mathbb{N}}$ 
			converging to the limit $a^\star = 1$.\label{fig:convergence_dict}}
		\end{figure} 
		The concept of convergence of a real-valued \gls{sequence} 
		(where $\mathcal{A}=\mathbb{R}$) extends naturally to a \gls{sequence} 
		in an arbitrary \gls{metricspace} $\mathcal{A}$. Indeed, we just need to 
		replace the absolute difference $|a_{\sampleidx} - a^\star|$ 
		with the \gls{metric} $\metric{a_{\sampleidx}}{a^\star}$. 
		Note that a \gls{sequence} can only converge if it is a 
		\gls{cauchysequence} \cite{RudinBookPrinciplesMatheAnalysis}. 
		However, not every \gls{cauchysequence} is converging unless 
		the underlying \gls{metricspace} is complete.   
		\\ 
		See also: \gls{sequence}, \gls{metricspace}, \gls{cauchysequence}.},
	first={convergence},
	type=math,
	text={convergence}, 
}

\newglossaryentry{johnsonlindenstrausslemma}
{name={Johnson--Lindenstrauss lemma (JL lemma)},
	description={The\index{Johnson--Lindenstrauss lemma (JL lemma)} JL lemma describes conditions for 
  		the existence of a \gls{featuremap} $\featuremapvec: \mathbb{R}^{\nrfeatures} \to \mathbb{R}^{\nrfeatures'}$ 
  		with $\nrfeatures' \ll \nrfeatures$ such that the pairwise \gls{eucliddist} 
  		between \glspl{featurevec} of a finite \gls{dataset} is approximately preserved 
  		\cite{vershynin2018high}, \cite{JMLR:v19:18-264}, \cite{johnson1984extensions}. 
  		Consider a \gls{dataset} $\dataset = \big\{\featurevec^{(1)}, \,\dots, \,\featurevec^{(\samplesize)} \big\}$ 
  		with \glspl{datapoint} characterized by \glspl{featurevec} in $\mathbb{R}^{\nrfeatures}$. 
  		Then, for any $\nrfeatures'$ that satisfies
		\[
	  	\nrfeatures' \ge \frac{4 \ln(\samplesize)}{\varepsilon^2/2 - \varepsilon^3/3} \quad \mbox{for some } 0 < \varepsilon < 1,
		\]  
   		there is a \gls{featuremap} $\featuremapvec$ such that \cite{ProofJLlemma}
  		\begin{equation} 
 			\label{equ_def_approx_norm_JL_lemma_dict}
 			(1\!-\!\varepsilon)\normgeneric{\featurevec^{(\sampleidx)}\!-\!\featurevec^{(\sampleidx')}}{2}  
 			\!\le\!\normgeneric{\featuremapvec\big(\featurevec^{(\sampleidx)}\big)\!-\!\featuremapvec\big(\featurevec^{(\sampleidx')}\big)}{2} 
			\!\le\!(1\!+\!\varepsilon)\normgeneric{\featurevec^{(\sampleidx)}\!-\!\featurevec^{(\sampleidx')}}{2}
 		\end{equation}
		for all $\featurevec^{(\sampleidx)}, \featurevec^{(\sampleidx')} \in \dataset$. 
		\begin{figure}[H]
			\centering
			\begin{tikzpicture}
			\begin{scope}
				\coordinate (x1) at (0.5,-0.6);   
				\coordinate (x2) at (2.0,0.9);    
				\coordinate (x3) at (1.1,0.3);    
				\foreach \p in {x1,x2,x3} \fill (\p) circle (1.7pt);
				\node[below left]  at (x1) {\small $\featurevec^{(1)}$};
				\node[above right] at (x2) {\small $\featurevec^{(2)}$};
				\node[above left]  at (x3) {\small $\featurevec^{(3)}$};
				\node [anchor=east] at (1.2,2.2) {$\mathbb{R}^{\nrfeatures}$};
			\end{scope}
			\begin{scope}[xshift=1cm]
				\draw[->, thick] (2.9,2.2) -- (4.1,2.2)
				node[midway, above] {$\featurevec \mapsto \vz \!\defeq\! \featuremapvec(\featurevec)$};
			\end{scope}
			\begin{scope}[xshift=2cm]
				\coordinate (y1) at (4.7,-0.7);   
				\coordinate (y2) at (6.1,0.5);    
				\coordinate (y3) at (5.3,-0.1);   
				\foreach \p in {y1,y2,y3} \fill (\p) circle (1.7pt);
				\node[below left]  at (y1) {\small $\vz^{(1)}$};
				\node[above right] at (y2) {\small $\vz^{(2)}$};
				\node[above left]  at (y3) {\small $\vz^{(3)}$};
				\node [anchor=west] at (6.0,2.2) {$\mathbb{R}^{\nrfeatures'}$};
			\end{scope}
			\end{tikzpicture}
		\caption{JL lemma offering precise conditions that guarantee the existence of 
		         a \gls{featuremap} $\featuremapvec: \mathbb{R}^{\nrfeatures} \rightarrow \mathbb{R}^{\nrfeatures'}$ 
		         such that pairwise \glspl{eucliddist} between (the \glspl{featurevec} of) \glspl{datapoint} are 
			approximately preserved. Roughly speaking, 
			$\featuremapvec$ maps neighboring points in the original \gls{featurespace} 
			to neighboring points in the new \gls{featurespace}.}
		\end{figure}
		The \gls{featuremap} $\featuremapvec$ can be obtained from a random \gls{matrix} 
		$\mA \in \mathbb{R}^{\nrfeatures' \times \nrfeatures}$ whose entries are \gls{iid} 
		\glspl{gaussrv} $A_{i,j} \sim \mvnormal{0}{1/\nrfeatures'}$. 
		It can be shown that the \gls{featuremap} $\featurevec \mapsto \underbrace{\mA \featurevec}_{\featuremapvec(\featurevec)}$ 
		satisfies \eqref{equ_def_approx_norm_JL_lemma_dict} with \gls{probability} at least $1\!-\!1/\samplesize$ \cite{ProofJLlemma}. 
		\\
  		See also: \gls{matrix}, \gls{norm}, \gls{vectorspace}, \gls{euclidspace}, \gls{dimred}, \gls{pca}.},	
	first={Johnson--Lindenstrauss lemma (JL lemma)},
	type=math, 
  	text={JL lemma}
}

\newglossaryentry{admm}
{name={alternating direction method of multipliers (ADMM)},
	description={The\index{alternating direction method of multipliers (ADMM)} 
		ADMM is an iterative \gls{optmethod} for solving a structured \gls{optproblem}. 
		In particular, the ADMM can be used to solve an \gls{optproblem} of the following form:
		\begin{align}
			\min_{\featurevec \in \mathbb{R}^{\nrfeatures},\featurevec' \in \mathbb{R}^{\nrfeatures'}} \; f(\featurevec) 
			+ g(\featurevec') \nonumber \\ 
			\mbox{ s.t. } \mA \featurevec - \mB \featurevec' = \vc \nonumber
		\end{align}
		for given \glspl{matrix} $\mA \in \mathbb{R}^{p \times \nrfeatures}$ and $\mB \in \mathbb{R}^{p \times \nrfeatures'}$
		and a given \gls{vector} $\vc \in \mathbb{R}^p$.
		\\
		See also: \gls{optmethod}, \gls{methodofmultipliers}. }, 
	first={alternating direction method of multipliers (ADMM)},
	type=math,
	text={ADMM}
}	

\newglossaryentry{methodofmultipliers}
{name={method of multipliers (MoM)},
	description={The\index{method of multipliers (MoM)} MoM is an 
		iterative \gls{optmethod} for solving a \gls{constrainedoptprob} 
		of the following form \cite{DistrOptStatistLearningADMM}:
		\begin{align}
			\min_{\vx \in \mathbb{R}^{\nrfeatures}} \; f(\vx) \nonumber \\ 
			\mbox{ s.t. } \mathbf{A}\vx = \vb. \nonumber
		\end{align}
		Here, $f: \mathbb{R}^{\nrfeatures} \rightarrow \mathbb{R}$ denotes 
		the \gls{objfunc}, $\mathbf{A} \in \mathbb{R}^{m \times \nrfeatures}$ 
		is a given \gls{matrix}, and $\vb \in \mathbb{R}^{m}$ is a given 
		\gls{vector}. The MoM is based on the \gls{augmentedlagrangian}
		\[L_\rho(\vx,\vy) = f(\vx) + \vy^\top \mathbf{A}\vx
		+ \frac{\rho}{2}\normgeneric{\mathbf{A}\vx - \vb}{2}^{2},
		\]
		where $\vy$ denotes the \gls{vector} of Lagrange multipliers and $\rho>0$
		is a penalty \gls{parameter}. The MoM constructs a \gls{sequence} of estimates 
		$\pair{\vx^{(1)}}{\vy^{(1)}},\,\ldots$ that converge to a solution of 
		the \gls{optproblem}. In particular, during each \gls{iteration} $\iteridx$, 
		the current estimates $\vx^{(\itercntr)},\vy^{(\itercntr)}$
		are updated as follows:
		\begin{align*} 
			\vx^{(\itercntr+1)} &= \argmin_{\vx \in \mathbb{R}^{\nrfeatures}} L_\rho(\vx,\vy^{(\itercntr)}), \\
			\vy^{(\itercntr+1)} &= \vy^{(\itercntr)} + \rho\, \big(\mathbf{A}\vx^{(\itercntr+1)} - \vb\big).
		\end{align*}
		The MoM can be written as a \gls{fixedpointiter} of the following form:
		\[
		\big( \vx^{(\itercntr+1)},\vy^{(\itercntr+1)} \big) = \fixedpointop \big(\vx^{(\itercntr)},\vy^{(\itercntr)} \big)
		\]
		with 
		\begin{align} 
			\fixedpointop:\mathbb{R}^{\nrfeatures}\!\times\!\mathbb{R}^{m}\!\to\!
			\mathbb{R}^{\nrfeatures}\!\times\!\mathbb{R}^{m}\!:\!\big(\vx,\vy\big)\!&\mapsto\!\big(\vx',\vy\!+\!\rho \big(\mathbf{A}\vx'\!-\!\vb\big) \big) \nonumber \\ 
			& \mbox{ with } \vx' = \argmin_{\vx \in \mathbb{R}^{\nrfeatures}} L_\rho(\vx,\vy). \nonumber
		\end{align}
		See also: \gls{optproblem}, \gls{augmentedlagrangian}, \gls{fixedpointiter}. }, 
 	first={method of multipliers (MoM)},
 	type=math,
 	text={MoM}
}

\newglossaryentry{augmentedlagrangian}
{name={augmented Lagrangian},
	description={The augmented \gls{lagrangian}\index{augmented Lagrangian} is a modification 
	 	of the standard \gls{lagrangian} that includes an additional quadratic 
		\gls{penaltyterm} for constraint violations. It is used in the \gls{methodofmultipliers} 
		for iteratively solving \glspl{constrainedoptprob}.
		\\
		See also: \gls{methodofmultipliers}.},
	first={augmented Lagrangian},
	text={augmented Lagrangian},
	firstplural={augmented Lagrangians},
	plural={augmented Lagrangians},
	type=math
}

\newglossaryentry{lagrangian}
{name={Lagrangian},
	description={Consider a \gls{constrainedoptprob} of the following form:
		\begin{align} 
			\min_{\weights \in \mathbb{R}^{\nrfeatures}} & f(\weights)   \nonumber \\ 
			\mbox{s.t.} \quad & g_{i}(\weights) \leq 0, \qquad i=1,\,\ldots,\,k \nonumber \\
			& h_j(\weights) = 0, \qquad j=1, \,\ldots,\,l. \nonumber
		\end{align}
		Here, $f: \mathbb{R}^{\nrfeatures} \rightarrow \mathbb{R}$ is the 
		\gls{objfunc}, $g_i: \mathbb{R}^{\nrfeatures} \rightarrow \mathbb{R}$, 
		$i=1,\,\ldots,\,k$, are the inequality constraint \glspl{function}, 
		and $h_j: \mathbb{R}^{\nrfeatures} \rightarrow \mathbb{R}$, 
		$j=1,\,\ldots,\,l$, are the equality constraint \glspl{function}.	
		The Lagrangian\index{Lagrangian} of the above \gls{optproblem} is defined as \cite{BoydConvexBook}
		\begin{equation}
			\nonumber
			L(\weights, {\bm \lambda}, {\bm \nu}) \defeq f(\weights) + \sum_{i=1}^{k} \lambda_i g_i(\weights) + \sum_{j=1}^{l} \nu_j h_j(\weights).
		\end{equation}
		Here, ${\bm \lambda} \in \mathbb{R}_{+}^{k}$ (i.e., $\lambda_i \geq 0$ for all $i=1,\ldots,k$) 
		and ${\bm \nu} \in \mathbb{R}^{l}$ are the multipliers associated with the inequality and equality 
		constraints, respectively. The Lagrangian is a useful tool for the analysis of  
		\glspl{constrainedoptprob} and the design of \glspl{optmethod}. For example, the Lagrangian allows us
		to define a \gls{dualproblem}, which provides a lower bound for the optimal value of a  
		\gls{constrainedoptprob}. This lower bound, in turn, can be used to construct a \gls{stopcrit} for 
		iterative \glspl{optmethod} \cite{BoydConvexBook}. Fig.\ \ref{fig_Langrian_dict} provides a 
		geometric interpretation of the Lagrangian for a \gls{constrainedoptprob} with a single 
		inequality constraint ($k=1$) and no equality constraints ($l=0$) (see \cite[Sec. 5.3]{BoydConvexBook}). 
		Here, the Lagrangian $L(\weights, \lambda)$ is the vertical intercept of a straight line with slope $-\lambda$ 
		that passes through a point $\pair{g_{1}(\weights)}{f(\weights)}$. 
		\begin{figure}[H]
			\centering
			\begin{tikzpicture}[>=stealth, scale=1.0]
			\path[fill=gray!40,opacity=0.4,draw=none,rotate around={-40:(1.6,2.1)}] (1.6,2.1) ellipse [x radius=2.7,y radius=0.65];
			\node at (1.6,2.1) {$\mathcal{G}$};
			\draw [->] (-3,0) -- (5,0) node[right] {};
			\draw [->] (0,-0.2) -- (0,4.2) node[above] {};
			\path[name path=line] (-2.6,2.55) -- (4.8,0.85);
			\path[name path=yaxis] (0,-0.2) -- (0,4.2);
			\draw[thick] (-2.6,2.55) -- (4.8,0.85);
			\draw[] (-2.6,2.55) -- (0,2.55); 
			\node[above] at ($(-2.6,2.55)!0.5!(0,2.55)$) {\small $1$};
			\node[below right,yshift=-2pt] at (0,2.55) {\small $\lambda$};
			\path[name intersections={of=line and yaxis, by=I}];
			\fill (I) circle (1.6pt) node[below left,xshift=-2pt] {$L(\weights,\lambda)$};
			\fill ($( -2.6,2.55)!0.8!(4.8,0.85)$) circle (1.6pt) node[above right] {$\pair{g_{1}(\weights)}{f(\weights)}$};
			\end{tikzpicture}
		\caption{\Gls{optproblem} with an \gls{objfunc} $f(\weights)$ and a single inequality constraint $g_{1}(\weights) \leq 0$, represented 
			by a set $\mathcal{G} \defeq \big\{ \big(g_{1}(\weights),\,f(\weights)\big) : \weights \in \mathbb{R}^{\nrfeatures} \big\}$.
			The value of the Lagrangian $L(\weights, \lambda)$ is the vertical intercept of a line with slope $-\lambda$ 
			that passes through the point $\pair{g_{1}(\weights)}{f(\weights)} \in \mathcal{G}$ \cite{BoydConvexBook}.
			\label{fig_Langrian_dict}}
		\end{figure}	
		See also: \gls{optproblem}, \gls{convexopt}, \gls{dualproblem}.},
	first={Lagrangian},
	text={Lagrangian},
	plural={Lagrangians},
	type=math
}

\newglossaryentry{kktconditions}
{name={Karush--Kuhn--Tucker conditions (KKT conditions)},
	description={Consider a constrained \gls{optproblem} with \gls{lagrangian}
		\[
		L(\vx, {\bm \lambda}, {\bm \nu}) 
		= f(\vx) 
		+ \sum_{i=1}^{k} \lambda_i g_i(\vx) 
		+ \sum_{j=1}^{l} \nu_j h_j(\vx)
		\]
		with \gls{differentiable} \glspl{function} $f, \,g_1, \,\ldots, \,g_k, \,h_1, \,\ldots, \,h_l$.	
		The \index{Karush--Kuhn--Tucker conditions (KKT conditions)} KKT conditions are  
		a system of equations and inequalities for the primal and 
		dual variables $\big(\vx,{\bm \lambda},{\bm \nu}\big)$ \cite[Sec. 5.5.3]{BoydConvexBook}:
		\begin{align}
			 \nabla_{\vx} L(\vx, {\bm \lambda}, {\bm \nu}) &= \mathbf{0}
			  && \text{(stationarity)} \nonumber\\
			  g_i(\vx) &\le 0, \qquad i=1,\,\ldots,\,k
			  && \text{(primal feasibility)} \nonumber\\
			  h_j(\vx) &= 0, \qquad j=1,\,\ldots,\,l
			  && \text{(primal feasibility)} \nonumber\\
			  \lambda_i &\ge 0, \qquad i=1,\,\ldots,\,k
			  && \text{(dual feasibility)} \nonumber\\
			  \lambda_i g_i(\vx) &= 0, \qquad i=1,\,\ldots,\,k
			  && \text{(complementary slackness).} \nonumber
		\end{align}
		If the primal problem results in a zero \gls{dualitygap}, 
		any optimal choice $(\vx^\star,{\bm \lambda}^\star,{\bm \nu}^\star)$ 
		for the primal and dual variables must satisfy the KKT conditions. 
		Conversely, if the primal problem is \gls{convex} and some regularity 
		conditions hold, any choice $(\vx^\star,{\bm \lambda}^\star,{\bm \nu}^\star)$ 
		that satisfies the KKT conditions is also optimal for the primal and \glspl{dualproblem} \cite{BoydConvexBook}.
		The KKT conditions can be written as a \gls{fixedpointeq} 
		and can, in turn, be used to construct \glspl{fixedpointiter} for solving 
		the primal and \glspl{dualproblem} \cite{BoydConvexBook}.
		\\
		See also: \gls{optproblem}, \gls{dualitygap}, \gls{convex}, \gls{dualproblem}, \gls{fixedpointiter}. },
	first={Karush--Kuhn--Tucker conditions (KKT conditions)},
	text={KKT conditions},
	type=math
}

\newglossaryentry{dualitygap}
{name={duality gap},
	description={Consider a \gls{constrainedoptprob} with \gls{lagrangian}
		\[
			L(\vx, {\bm \lambda}, {\bm \nu})
			= f_0(\vx)
			+ \sum_{i=1}^{k} \lambda_i f_i(\vx)
			+ \sum_{j=1}^{l} \nu_j g_j(\vx).
		\]
		The \gls{dualproblem} is defined as
		\[
			\max_{{\bm \lambda}\in \mathbb{R}_+^k,{\bm \nu} \in \mathbb{R}^l} 
			\underbrace{q({\bm \lambda},{\bm \nu})}_{\defeq \inf_{\vx} L(\vx,{\bm \lambda},{\bm \nu})}.
		\]
		Let $p^\star$ denote the optimal value of the primal problem and
		$d^\star$ the optimal value of the associated \gls{dualproblem}. The\index{duality gap} 
		duality gap is defined as
		\[
			p^\star - d^\star \;\ge\; 0.
		\]
		The duality gap is always nonnegative, even for non-\gls{convex} 
		and \gls{nonsmooth} \glspl{optproblem}. When the duality gap is
		zero, i.e., $p^\star = d^\star$, strong duality is said to hold \cite[Ch. 5]{BoydConvexBook}.
		\\
		See also: \gls{dualproblem}, \gls{optproblem}. },
	first={duality gap},
	text={duality gap},
	type=math
}

\newglossaryentry{dualproblem}
{name={dual problem},
sort={dualproblem},
	description={Consider a \gls{constrainedoptprob}, which we 
 		refer to as primal problem in what follows, of the following form:
    		\begin{align} 
			\min_{\vx \in \mathbb{R}^{\nrfeatures}} & f(\vx) \nonumber \\ 
     			\mbox{s.t.} \quad & g_i(\vx)\le 0, \qquad i=1,\,\ldots,\,k \nonumber \\ 
			& h_j(\vx)=0, \qquad j=1,\,\ldots,\,l, \nonumber
        		\end{align}
		and its associated \gls{lagrangedualfunc} $q({\bm \lambda},{\bm \nu})$. 
		For any ${\bm \lambda} \ge \mathbf{0}$, ${\bm \nu} \in \mathbb{R}^{l}$, 
		and any $\vx$ that satisfies the constraints of the primal problem \cite[Ch. 5]{BoydConvexBook},
		$$q({\bm \lambda},{\bm \nu}) \leq f(\vx).$$
		Making this lower bound maximally tight amounts to the following \gls{optproblem}:
              	\[
              	\max_{{\bm \lambda}\geq\mathbf{0},{\bm \nu}}\; q({\bm \lambda},{\bm \nu}). 
              	\]
              	This \gls{optproblem} is referred to as the dual problem\index{dual problem} associated with 
		the original primal problem.
		Fig.\ \ref{fig_dual_problem_dict} illustrates the dual problem for an \gls{optproblem} 
		with a single inequality constraint. This \gls{optproblem} can be characterized by 
		the set $\mathcal{A}=\{(u,t): f(\vx) \leq t, g_1(\vx) \leq u \mbox{ for some } \vx \in \mathbb{R}^{\nrfeatures}\}$ 
		\cite [Sec. 5.3]{BoydConvexBook}. The dual problem amounts to finding a \gls{supportinghyperplane} 
		for the set $\mathcal{A}$ with the largest vertical intercept.
		\begin{figure}[H]
     		  	\centering
     		  	\begin{tikzpicture}[>=stealth, scale=1.0]
     		  		\coordinate (G1) at (-0.8,3.55);
     		  		\coordinate (G3) at ( 1.6,1.25);
     		  		\coordinate (G5) at ( 4.0,1.1);
     		  		\path[fill=gray!35, opacity=0.5, draw=none]
     		  		(G1) --
     		  		(G3) --
     		  		(G5) --
     		  		(4.0,3.95) --
     		  		(-0.8,3.95) --
     		  		cycle;
     		  		\draw[thick] (G1) -- (G3)  -- (G5);
     		  		\draw[thin] (G1) -- (-0.8,3.95) -- (4.0,3.95) -- (G5);
     		  		\node at (1.6,2.1) {$\mathcal{A}$};
     		  		\draw [->] (-3,0) -- (7,0) node [below] {$u$};
     		  		\draw [->] (0,-0.2) -- (0,5.2) node [left] {$t$};
     		  		\path[name path=line1] (-2.6,4.25) -- (G3);
     		  		\path[name path=line] (-2.6,1.65) -- (G3);
     		  		\path[name path=yaxis] (0,-0.2) -- (0,4.2);
     		  		\draw[thick] (-2.6,4.25) -- (G3) -- ++($(G3)-(-2.6,4.25)$);
     		  		\draw[thick] (-2.6,1.65) -- (G3) -- ++($(G3)-(-2.6,1.65)$);
     		  		\path[name intersections={of=line and yaxis, by=I}];
     		  		\fill (I) circle (1.6pt) node[below left,xshift=-2pt,yshift=2pt] {$q(\lambda)$};
     		  		\path[name intersections={of=line1 and yaxis, by=I1}];
     		  		\fill (I1) circle (1.6pt) node[below left,xshift=-2pt,yshift=2pt] {$q(\lambda')$};
     		  	\end{tikzpicture}
     		\caption{Dual problem of a \gls{constrainedoptprob}: finding a \gls{supportinghyperplane} with 
     		  	the largest vertical intercept \cite [Sec. 5.3]{BoydConvexBook}.	\label{fig_dual_problem_dict}}
		\end{figure}
		See also: \gls{optproblem}, \gls{lagrangian}. },
	first={dual problem},
	type=math,
	text={dual problem}
}

\newglossaryentry{euclidspace}
{name={Euclidean space}, 
	description={The\index{Euclidean space} 
		Euclidean space $\mathbb{R}^{\featuredim}$ of dimension $\featuredim \in \mathbb{N}$ consists 
		of \glspl{vector} $\featurevec= \big(\feature_{1}, \,\ldots, \,\feature_{\featurelen}\big)$ with $\featuredim$ 
		real-valued entries $\feature_{1}, \,\ldots, \,\feature_{\featuredim} \in \mathbb{R}$. Such a Euclidean 
		space is equipped with a geometric structure defined by the \gls{innerproduct} 
		$\featurevec\,^{T} \featurevec' = \sum_{\featureidx=1}^{\featuredim} \feature_{\featureidx} \feature'_{\featureidx}$ 
		between any two \glspl{vector} $\featurevec,\featurevec' \in \mathbb{R}^{\featuredim}$ \cite{RudinBookPrinciplesMatheAnalysis}.
		\\
		See also: \gls{vector}. },
	first={Euclidean space},
	type=math, 
	plural={Euclidean spaces}, 
	firstplural={Euclidean spaces},
	text={Euclidean space} 
}

\newglossaryentry{euclidnorm}
{name={Euclidean norm}, 
	description={The\index{Euclidean norm} 
		Euclidean \gls{norm} (or $2$-\gls{norm}, or $\ell_2$-\gls{norm}) $\normgeneric{\weights}{2}$ of a \gls{vector} 
		$\weights = \big(\weight_{1},\,\ldots,\,\weight_{\nrfeatures} \big) \in \mathbb{R}^{\nrfeatures}$
		is defined as 
		$$\normgeneric{\weights}{2} \defeq \sqrt{\sum_{\featureidx=1}^{\nrfeatures}\weight^2_{\featureidx}}.$$ 
		The Euclidean \gls{norm} is distinct among all \glspl{norm} on $\mathbb{R}^{\nrfeatures}$, in the sense 
		that it is induced by the \gls{innerproduct} $\vw^{T} \vv$ \cite{RudinBook}, \cite{BoydConvexBook}, \cite{HalmosFiniteDimVecSpace}. 
		In other words, $\normgeneric{\weights}{2}= \sqrt{\weights^{T} \weights}$.	
		\\
		See also: \gls{norm}, \gls{vector}, \gls{metric}, \gls{euclidspace}, \gls{hilbertspace}. },
	first={Euclidean norm},
	type=math,
	plural={Euclidean norms}, 
	firstplural={Euclidean norms},
	text={Euclidean norm} 
}

\newglossaryentry{eucliddist}
{name={Euclidean distance}, 
	description={The\index{Euclidean distance} Euclidean \gls{distance} between two 
		\glspl{vector} $\vw,\vw' \in \mathbb{R}^{\nrfeatures}$ is the \gls{euclidnorm} 
		$\normgeneric{\vw-\vw'}{2}$ of the difference $\vw-\vw'$.	
		\\
		See also: \gls{distance}, \gls{vector}, \gls{euclidspace}. },
	first={Euclidean distance},
	type=math,
	plural={Euclidean distances},
	firstplural={Euclidean distances},
	text={Euclidean distance} 
}

\newglossaryentry{boundary}
{name={boundary},
	description={Consider\index{boundary} a subset $\mathcal{C} \subseteq \mathbb{R}^n$. 
                 The boundary of $\mathcal{C}$, denoted by $\bd{\mathcal{C}}$, is the 
		set of all points $\vx \in \mathbb{R}^n$ such that every closed ball 
                 $\closedball{\vx}{\varepsilon} = \{\vy \in \mathbb{R}^n: \norm{\vy - \vx} \leq \varepsilon\}$ 
                 of radius $\varepsilon > 0$ centered at $\vx$ intersects both $\mathcal{C}$ 
		and its complement $\mathbb{R}^n \setminus \mathcal{C}$.
		\begin{figure}[H]
			\centering
			\begin{tikzpicture}[scale=1, >=stealth]
			\def\setpath{plot [smooth cycle, tension=0.8] coordinates {(-2, 0.5) (-0.5, 2.2) (1.8, 1.2) (1.5, -1.5) (-1.2, -1.8)}}
			\fill[gray!30] \setpath;
			\draw[thick, dashed] \setpath;
			\coordinate (xPoint) at (1.8, 1.2);
			\draw[thin] (xPoint) circle (0.6cm);
			\fill (xPoint) circle (2pt);
			\node at (-0.2, 0) [font=\large] {$\mathcal{C}$};
			\node[right] at ($(xPoint)+(0.1,0)$) {$\vx$};
			\node (ballLabel) at (2.8, 2.2) {$\closedball{\vx}{\varepsilon}$};
			\draw[->, thin] (ballLabel) -- ($(xPoint)+(45:0.6cm)$);
			\node[anchor=east] (legendText) at (7.2, 0.0) {boundary $\bd{\mathcal{C}}$};
			\draw[thick, dashed] ($(legendText.west)+(-0.6,0)$) -- ($(legendText.west)+(-0.1,0)$);
			\end{tikzpicture}
		\caption{Boundary of the set $\mathcal{C}$ and a closed ball $\closedball{\vx}{\varepsilon}$ around 
			$\vx\!\in\!\bd{\mathcal{C}}$.}
		\end{figure}
		Note that the boundary $\bd{\mathcal{C}}$ 
		might contain elements that do not belong to $\mathcal{C}$ itself.
		The notion of a boundary can be generalized to arbitrary \glspl{metricspace}, including 
		an \gls{undirectedgraph} with a \gls{metric} given by the length of the shortest path.
		\\
		See also: \gls{metricspace}, \gls{neighborhood}. },
	first={boundary}, 
	type=math,
	text={boundary}
}

\newglossaryentry{neighborhood}
{name={neighborhood},
 	description={Consider some \gls{metricspace} $\featurespace$ with a \gls{metric} 
		$d: \featurespace \times \featurespace \rightarrow \mathbb{R}_{+}$.	
 		The\index{neighborhood} neighborhood of a point $\featurevec \in \featurespace$ 
 		is the set of other points having a sufficiently small \gls{distance} to $\featurevec$. 
 		For example, the $\epsilon$-neighborhood of $\featurevec$ is defined as
 		$$ \{ \featurevec' \in \featurespace : d(\featurevec, \featurevec') \leq \epsilon \}.$$
		If $\featurespace$ is an \gls{undirectedgraph}, which is a special case of 
 		a \gls{metricspace}, the neighborhood of a node $\nodeidx \in \nodes$ 
 		is the set of its \glspl{neighbor}.
 				\\ 
 		See also: \gls{metric}, \gls{distance}, \gls{neighbor}.},
 	first={neighborhood},
	type=math,
	firstplural={neighborhoods},
 	plural={neighborhoods},
 	text={neighborhood} 
}

\newglossaryentry{metric}
{name={metric},
	description={A\index{metric} metric is a quantitative \gls{measure} used to compare objects. 
	     	In mathematics, a metric measures the \gls{distance} between two points in a space and must 
		follow specific rules, i.e., the \gls{distance} is always nonnegative, zero only if the 
		points are the same, symmetric, and it satisfies the triangle inequality \cite{RudinBookPrinciplesMatheAnalysis}. In the context of 
		\gls{ml}, the term metric refers to a quantitative \gls{measure} of how well a \gls{model} performs (somewhat 
		similar to a \gls{lossfunc}). Examples include \gls{acc}, \gls{precision}, and the average 
		\gls{zerooneloss} on a \gls{testset} \cite{Goodfellow-et-al-2016}, \cite{BishopBook}. 
		The term \gls{lossfunc} is typically used in the context of \gls{model} \gls{training}, 
		while the term metric is used in the context of \gls{model} \gls{validation}.
		\\ 
		See also: \gls{distance}, \gls{acc}, \gls{precision}, \gls{validation}, \gls{loss}.},
	first={metric}, 
	type=math,
	plural={metrics},
	text={metric}
}

\newglossaryentry{lagrangedualfunc}
{name={Lagrange dual function},
	description={Consider\index{Lagrange dual function} a \gls{constrainedoptprob}
		of the following form:
		\begin{align}
			\min_{\vx \in \mathbb{R}^{\nrfeatures}} & f(\vx) \nonumber \\
			\mbox{s.t.} \quad & g_i(\vx)\le 0, \qquad i=1,\,\ldots,\,k \nonumber \\ 
			& h_j(\vx)=0, \qquad j=1,\,\ldots,\,l. \nonumber
		\end{align}
		A useful tool for the analysis of such an \gls{optproblem} is the 
		Lagrange dual \gls{function}:
		\begin{equation}
			\nonumber
			q: \mathbb{R}^{k}_{+} \times \mathbb{R}^{l}:	q({\bm \lambda},{\bm \nu})=\inf_{\vx \in \mathbb{R}^{\nrfeatures}} L(\vx,{\bm \lambda},{\bm \nu}),
		\end{equation}
		where $L(\vx,{\bm \lambda},{\bm \nu}) \defeq f(\vx) + \sum_{i=1}^{k} \lambda_i g_i(\vx) + \sum_{j=1}^{l} \nu_{j} h_{j}(\vx)$ 
		denotes the \gls{lagrangian} of the original \gls{optproblem}. 
		Fig.\ \ref{fig_dual_function_dict} illustrates the \gls{function} $q(\lambda)$ obtained 
		for an \gls{optproblem} with a single inequality constraint. For a given value $\lambda \geq 0$, 
		$q(\lambda)$ is the vertical intercept of a \gls{supportinghyperplane} with \gls{normalvector} 
		$(\lambda,1)$ to the set $\mathcal{A}=\{(u,t): f(\vx) \leq t, g_{1}(\vx) \leq u \mbox{ for some } \vx \in \mathbb{R}^{\nrfeatures}\}$ 
		\cite [Sec. 5.3]{BoydConvexBook}. 
		\begin{figure}[H]
			\centering
			\begin{tikzpicture}[>=stealth, scale=1.0]
			\coordinate (G1) at (-0.8,3.55);
			\coordinate (G3) at ( 1.6,1.25);
			\coordinate (G5) at ( 4.0,1.1);
			\path[fill=gray!35, opacity=0.5, draw=none]
			(G1) --
			(G3) --
			(G5) --
			(4.0,3.95) --
			(-0.8,3.95) --
			cycle;
			\draw[thick] (G1) -- (G3)  -- (G5);
			\draw[thin] (G1) -- (-0.8,3.95) -- (4.0,3.95) -- (G5);
			\node at (1.6,2.1) {$\mathcal{A}$};
			\draw [->] (-3,0) -- (7,0) node [below] {$u$};
			\draw [->] (0,-0.2) -- (0,5.2) node [left] {$t$};
			\path[name path=line] (-2.6,2.55) -- (G3);
			\path[name path=yaxis] (0,-0.2) -- (0,4.2);
			\draw[thick] (-2.6,2.55) -- (G3) -- ++($(G3)-(-2.6,2.55)$);
			\draw[dashed] (-2.6,2.55) -- (0,2.55);
			\node[above] at ($(-2.6,2.55)!0.5! (0,2.55)$) {\small $1$};
			\node[below right,yshift=-2pt] at (0,2.55) {\small $\lambda$};
			\path[name intersections={of=line and yaxis, by=I}];
			\fill (I) circle (1.6pt) node[below left,xshift=-2pt] {$q(\lambda)$};
			\end{tikzpicture}
		\caption{Lagrange dual \gls{function} $q(\lambda)$, for a given $\lambda\geq0$, represented as the vertical intercept 
			of a \gls{supportinghyperplane} with \gls{normalvector} $(\lambda,1)$ 
			to the set $\mathcal{A}=\{(u,t): f(\vx) \leq t, g_{1}(\vx) \leq u \mbox{ for some } \vx \in \mathbb{R}^{\nrfeatures}\}$ 
			\cite [Sec. 5.3]{BoydConvexBook}.	\label{fig_dual_function_dict}}
		\end{figure}
		See also: \gls{optproblem}, \gls{function}, \gls{lagrangian}, \gls{supportinghyperplane}. },
	first={Lagrange dual function},
	type=math,
	text={Lagrange dual function}
}

\newglossaryentry{supportinghyperplane}
{name={supporting hyperplane},
	description={Let\index{supporting hyperplane} $\mathcal{C} \subseteq \mathbb{R}^d$ be a non-empty set.
		A \gls{hyperplane}
		\[  
			\{ \vx \in \mathbb{R}^d : \langle \va, \vx \rangle = b \},
			\qquad \va \in \mathbb{R}^d \setminus \{ \mathbf{0} \}, \qquad \ b \in \mathbb{R}
		\]
		is called a supporting \gls{hyperplane} of $\mathcal{C}$ if
		\begin{align*}
			\va^{T} \vx \le b 
			\quad \text{for all } \vx \in \mathcal{C}
		\end{align*}
		and there exists at least one point $\vx^\star$ in the \gls{boundary} of $\mathcal{C}$ such that
		$\va^{T} \vx = b$ \cite{BoydConvexBook}. The \gls{vector} $\va$ is then a \gls{normalvector} 
		to the supporting \gls{hyperplane}, pointing toward the \gls{halfspace} that does not 
		contain the set $\mathcal{C}$ (see \cite[Sec. 2.5.2]{BoydConvexBook}). 
		\\
		See also: \gls{hyperplane}, \gls{boundary}. }, 
	first={supporting hyperplane}, 
	text={supporting hyperplane}, 
	type=math
}

\newglossaryentry{statespace}
{name={state space},
	description={The \gls{state} space\index{state space} of a system is constituted  
		by all possible \glspl{state} of the system at any point in time. 
		\\
		See also: \gls{featurevec}, \gls{hypothesis}, \gls{action}.},
	first={state space},
	type=math,
	text={state space}
}

\newglossaryentry{state}
{name={state},
	description={A state\index{state} is a mathematical representation of the minimal 
		information needed to characterize a system at a given time 
		such that, together with the system dynamics, it suffices
		to predict the system’s future behavior \cite{SuttonEd2}, \cite{BertsekasDynOptII}. 	
		\\
		See also: \gls{featurevec}, \gls{hypothesis}, \gls{action}.},
	first={state},
	text={state}, 
	plural={states},
	firstplural={states},
	type=math
}

\newglossaryentry{differentiable}
{name={differentiable},
	description={A\index{differentiable} real-valued \gls{function} $f: \mathbb{R}^{\featuredim} \rightarrow \mathbb{R}$ 
		is differentiable if it can be approximated locally at any point by a linear \gls{function}. 
		The local linear approximation at the point $\mathbf{x}$ is determined 
		by the \gls{gradient} $\nabla f ( \mathbf{x})$ \cite{RudinBookPrinciplesMatheAnalysis}.
					\\ 
		See also: \gls{function}, \gls{gradient}.},
	first={differentiable},
	type=math,
	text={differentiable} 
}

\newglossaryentry{gradient}
{name={gradient},
	description={For\index{gradient} a real-valued \gls{function} 
		$f: \mathbb{R}^{\featuredim} \rightarrow \mathbb{R}: \weights \mapsto f(\weights)$, 
		if a \gls{vector} $\vg$ exists such that 
		$\lim_{\weights \rightarrow \weights'} {f(\weights) - \big(f(\weights') + \vg\,^{T} (\weights - \weights') \big) }/{\| \weights - \weights'\|}=0$, 
		it is referred to as the gradient of $f$ at $\weights'$. If it exists, the gradient is unique and 
		denoted by $\nabla f(\weights')$ or $\nabla f(\weights)\big|_{\weights'}$ \cite{RudinBookPrinciplesMatheAnalysis}.
		For a \gls{differentiable} \gls{function}, the gradient is the \gls{vector} whose entries are the 
		\glspl{partialderivative} of $f$ with respect to each input component, such that 
		$\nabla f(\weights) = [{\partial f}/{\partial \feature_1}, \,\dots, \,{\partial f}/{\partial \feature_{\featuredim}}]^T$. 
		Geometrically, the gradient points in the direction of steepest ascent.
		\\
		See also: \gls{function}, \gls{vector}, \gls{differentiable}, \gls{partialderivative}, \gls{gdmethod}, \gls{gd}, 
		\gls{zerogradientcondition},  \gls{hessian}. },
	first={gradient},
	plural={gradients},
	type=math, 
	text={gradient}
}

\newglossaryentry{subgradient}
{name={subgradient},
	description={For\index{subgradient} a real-valued \gls{function} $f: \mathbb{R}^{\featuredim} \rightarrow \mathbb{R}: \weights \mapsto f(\weights)$, 
		a \gls{vector} $\va$ such that $f(\weights) \geq  f(\weights') +\big(\weights-\weights' \big)\,^{T} \va$ is 
		referred to as a subgradient of $f$ at $\weights'$ \cite{BertCvxAnalOpt}, \cite{BertsekasNonLinProgr}.
		\\
		See also: \gls{function}, \gls{vector}.},
	first={subgradient},
	type=math,
	plural={subgradients},
	text={subgradient} 
}

\newglossaryentry{distance}
{name={distance},
	description={The\index{distance} distance between two points in a \gls{metricspace} 
		is the value of the \gls{metric} evaluated at those points \cite{BasicRealAnalysis}.
		\\
		See also: \gls{metricspace}, \gls{eucliddist}.},
	first={distance},
	firstplural={distances},
	type=math,
	plural={distances},
	text={distance} 
}

\newglossaryentry{Lipschitz}
{name={Lipschitz continuity},
	description={A\index{Lipschitz continuity} \gls{function} between two \glspl{metricspace} is 
		Lipschitz \gls{continuous} if there exists a constant $L > 0$ such that 
		the \gls{distance} between the \gls{function} values at any two points is at 
		most $L$ times the \gls{distance} between those points \cite{BasicRealAnalysis}. 
		\\
		See also: \gls{function}, \gls{vector}.},
	first={Lipschitz continuity},
	type=math,
	plural={Lipschitz continuities},
	firstplural={Lipschitz continuities},
	text={Lipschitz continuity} 
}

\newglossaryentry{strcvx}
{name={strongly convex},
	description={A \gls{function}\index{strongly convex} $f:\mathbb{R}^d \rightarrow \mathbb{R}\cup\{+\infty\}$ is 
		$\mu$-strongly \gls{convex} (with \gls{parameter} $\mu>0$) if \cite{Bauschke:2017}
		\[
			f(\lambda \vx+(1-\lambda)\vy)
			\le
			\lambda f(\vx)+(1-\lambda)f(\vy)
			-
			\frac{\mu}{2}\lambda(1-\lambda)\normgeneric{\vx-\vy}{2}^{2}
		\]
		for all $\vx,\vy\in\mathbb{R}^d$ and $\lambda\in[0,1]$.
		Equivalently, $f$ is $\mu$-strongly \gls{convex} if, for every 
		$\vg\in\partial f(\vx)$, \cite[Sec. 3.4]{CvxBubeck2015}
		\[
			f(\vy) \ge f(\vx) + \vg^\top (\vy-\vx)
			+ \frac{\mu}{2}\normgeneric{\vy-\vx}{2}^{2}
			\quad \mbox{for all } \vx,\vy \in \mathbb{R}^{d}.
		\]
		If $f$ is \gls{differentiable}, this reduces to \cite[Sec. B.1.1]{CvxAlgBertsekas}
		\[
			f(\vy) \ge f(\vx) + \nabla f(\vx)^\top (\vy-\vx)
			+ \frac{\mu}{2}\normgeneric{\vy-\vx}{2}^{2},
		\]
		and if $f$ is twice \gls{differentiable},
		\[
			\nabla^2 f(\vx) \succeq \mu \mathbf{I} 
			\quad \mbox{for all } \vx \in \mathbb{R}^{d}.
		\]
		Strong convexity implies uniqueness of the minimizer of $f(\vx)$ and enables 
		linear \gls{convergence} rates for \glspl{gdmethod}.
					\\ 
		See also: \gls{function}, \gls{convex}, \gls{differentiable}, \gls{strictlyconvex}.},
	first={strongly convex},
	type=math,
	text={strongly convex} 
}

\newglossaryentry{strictlyconvex}
{name={strictly convex},
	description={A\index{strictly convex} real-valued \gls{function} $f: \mathcal{U} \to \mathbb{R}$ 
		with \gls{convex} \gls{domain} $\mathcal{U} \subset  \mathbb{R}^d$ is strictly \gls{convex} if, 
		for any two distinct $\vx,\vy \in \mathcal{U}$ and any $\lambda\in(0,1)$, 
		it satisfies \cite[Definition 3.1.1]{BoydConvexBook}
		\[
			f(\lambda\vx+(1-\lambda)\vy)
			< \lambda f(\vx)+(1-\lambda)f(\vy).
		\]
		Furthermore, if $f$ is \gls{differentiable}, this condition is equivalent to
		\[
			f(\vy) > f(\vx) + \nabla f(\vx)^{T}(\vy-\vx)
		\]
		for any $\vx\neq\vy$, which implies that $f(\cdot)$ admits a unique minimizer on 
		any \gls{convex} subset of its \gls{domain}. Unlike \gls{strcvx} \glspl{function}, strictly \gls{convex} 
		\glspl{function} do not require a uniform quadratic lower bound.
					\\ 
		See also: \gls{convex}, \gls{strcvx}.},
	first={strictly convex},
	type=math, 
	text={strictly convex} 
}

\newglossaryentry{directedcycle}
{name={directed cycle},
	description={A directed cycle\index{directed cycle} in a \gls{directedgraph} $\graph=\pair{\nodes}{\edges}$ 
		is a \gls{sequence} of distinct nodes $(\nodeidx_1, \,\nodeidx_2, \,\ldots, \,\nodeidx_k)$ 
		such that $(\nodeidx_1, \,\nodeidx_2), (\nodeidx_2, \,\nodeidx_3), \,\ldots, (\nodeidx_{k-1}, \,\nodeidx_k), 
     		(\nodeidx_k, \nodeidx_1) \in \edges$. In a directed cycle, following the direction 
	 	of each edge eventually leads back to the starting node, creating a closed loop. 
	 	\begin{figure}[H]
			\centering
			\begin{tikzpicture}[>=Latex, node distance=1.4cm, thick]
	 	 	\coordinate (a1) at (90:1.5);
  	  		\coordinate (a2) at (210:1.5);
      			\coordinate (a3) at (330:1.5);
  	 		\fill (a1) circle (2pt) node[above=3pt] {$\nodeidx_1$};
  	 		\fill (a2) circle (2pt) node[below left=3pt] {$\nodeidx_2$};
  	 		\fill (a3) circle (2pt) node[below right=3pt] {$\nodeidx_3$};
			\draw[->] (a1) -- (a2);
			\draw[->] (a2) -- (a3);
			\draw[->] (a3) -- (a1);
			\node[above=3pt of a1] {$\nodeidx_1$};
			\node[below left=3pt of a2] {$\nodeidx_2$};
			\node[below right=3pt of a3] {$\nodeidx_3$};
			\end{tikzpicture}
		\caption{Directed cycle consisting of three nodes \gls{connected} in a closed loop.}
		\end{figure}
     		The presence of a directed cycle prevents a \gls{directedgraph} from being a \gls{dag}. 
					\\ 
		See also: \gls{directedgraph}, \gls{dag}.},
	first={directed cycle},
	firstplural={directed cycles}, 
	plural={directed cycles}, 
	type=math, 
	text={directed cycle} 
}

\newglossaryentry{dag}
{name={directed acyclic graph (DAG)},
	description={A\index{directed acyclic graph (DAG)} DAG is a \gls{directedgraph} 
    		that contains no \glspl{directedcycle}. Formally, a DAG $\graph = (\nodes, \edges)$ satisfies 
		that, for any \gls{sequence} of distinct nodes $(\nodeidx_1, \,\ldots, \,\nodeidx_k)$, the presence 
		of directed edges $(\nodeidx_1,\,\nodeidx_2), (\nodeidx_2,\,\nodeidx_3), \,\ldots, 
		(\nodeidx_{k-1},\,\nodeidx_k)$ implies that $(\nodeidx_k, \,\nodeidx_1) \notin \edges$. 
		\begin{figure}[H]
			\centering
			\begin{tikzpicture}[>=Latex, node distance=1.4cm, thick, every node/.style={circle, fill=black, inner sep=1.5pt}]
			\node (a1) {};
			\node[right=of a1] (a2) {};
			\node[right=of a2] (a3) {};
			\draw[->] (a1) -- (a2);
			\draw[->] (a2) -- (a3);
			\node[shape=rectangle, draw=none, fill=none] at (1.5,-1.5) {(a)};
			\node[right=3.5cm of a3] (b1) {};
			\node[right=of b1] (b2) {};
			\node[right=of b2] (b3) {};
			\draw[->] (b1) -- (b2);
			\draw[->] (b2) -- (b3);
			\draw[->, bend left=40] (b3) to (b1); 
			\node[shape=rectangle, draw=none, fill=none] at (8.3,-1.5) {(b)};
			\end{tikzpicture}
		\caption{(a) DAG defined on three nodes $\nodes=\{1,\,2,\,3\}$. (b) \Gls{directedgraph} 
			on the same nodes that is not a DAG, since it contains a \gls{directedcycle}.}
		\end{figure}
    		The absence of \glspl{directedcycle} allows for a topological ordering of nodes such that 
    		all edges point from earlier to later nodes in this order. 
		Several \gls{ml} \glspl{model}, such as \glspl{ann} or \glspl{decisiontree}, 
    		are naturally represented as DAGs. 
		\\
		See also: \gls{directedgraph}, \gls{ann}, \gls{decisiontree}. }, 
 	first={directed acyclic graph (DAG)}, 
 	firstplural={directed acyclic graphs (DAGs)}, 
 	plural={DAGs}, 
 	type=math, 
 	text={DAG} 
}

\newglossaryentry{directedgraph}
{name={directed graph},
	description={A directed \gls{graph}\index{directed graph} contains edges that 
		have an orientation (or direction). Mathematically, 
		a directed \gls{graph} $\graph=\pair{\nodes}{\edges}$ consists of 
		nodes $\nodes$ and a set $\edges \in \nodes \times \nodes$ of directed edges. 
		\begin{figure}[H]
			\centering
			\begin{tikzpicture}[>=stealth, node distance=1.8cm]
  				\node[circle, draw, fill=black, inner sep=1.5pt, label=below:{$\nodeidx$}] (i) {};
  				\node[circle, draw, fill=black, inner sep=1.5pt, label=below:{$\nodeidx'$}] (ip) [right=of i] {};
  				\draw[->, >=Latex, line width=1pt, scale=1.3] (i) -- (ip);
			\end{tikzpicture}
		\caption{Edges of a directed \gls{graph} with orientation (or direction), 
			indicated by an arrowhead.}
		\end{figure}
   		We can represent a directed edge from node $\nodeidx \in \nodes$ to node 
   		$\nodeidx' \in \nodes$ by an ordered pair $\pair{\nodeidx}{\nodeidx'}$. 
   		Directed \glspl{graph} are widely used to model interconnected systems or 
   		networks, such as transportation systems, electronic circuits, and biological 
    		processes \cite{NewmannBook}. 
					\\ 
		See also: \gls{graph}.},
	first={directed graph},
	firstplural={directed graphs}, 
	plural={directed graphs}, 
	type=math,
	text={directed graph} 
}

\newglossaryentry{undirectedgraph}
{name={undirected graph},
	description={See \gls{graph}\index{undirected graph}. },
	first={undirected graph},
	firstplural={undirected graphs}, 
	plural={undirected graphs}, 
	type=math,
	text={undirected graph} 
}

\newglossaryentry{simplefunction}
{name={simple function},
	description={A\index{simple function} simple \gls{function} 
 		$f: \mathbb{R}^{\nrfeatures} \rightarrow \mathbb{R}$ is a \gls{measurable} 
 		\gls{function} that takes on only finitely many values. In other words, 
 		$$f(\featurevec) = \sum_{\clusteridx=1}^{\nrcluster} \alpha_{\clusteridx} \indicatorfunc{\cluster^{(\clusteridx)}}(\featurevec),$$ 
 		where $\indicatorfunc{\cluster}$ denotes the \gls{indicatorfunc} of a 
 		subset $\cluster \subset \mathbb{R}^{\nrfeatures}$ and 
 		$\alpha_{1},\,\ldots,\,\alpha_{\nrcluster} \in \mathbb{R}$ are arbitrary coefficients. 
 		The subsets $\cluster^{(1)},\,\ldots,\,\cluster^{(\nrcluster)}$ in the above 
 		decomposition must be \gls{measurable} and must form a partition of $\mathbb{R}^{\nrfeatures}$. 
		\\ 
		See also: \gls{measurable}, \gls{LebesgueIntegral}.}, 
 	first={simple function}, 
	firstplural={simple functions}, 
 	plural={simple functions}, 
 	text={simple function}, 
 	type=math
}

\newglossaryentry{concentrationinequ}
{name={concentration inequality}, 
	description={Concentration inequality refers to an upper bound on the \gls{probability}\index{concentration inequality} that 
		an \gls{rv} deviates more than a prescribed amount from its \gls{expectation} \cite{Wain2019}. 
		\\
		See also: \gls{probability}, \gls{rv}, \gls{expectation}, \gls{markovsinequality}, \gls{chebyshevsinequality}, 
		\gls{hoeffdingsinequality}, \gls{chernoffbound}.}, 
	type=math, 
	first={concentration inequality},
	firstplural={concentration inequalities},
	plural={concentration inequalities},  
	text={concentration inequality}
}

\newglossaryentry{covariancefunction}
{name={covariance function},
	description={A \gls{covariance} \gls{function}\index{covariance function} characterizes the
		second-order dependence structure of a \gls{stochproc}.
		For a real-valued \gls{stochproc} $\{\feature(\timeidx)\}_{\timeidx\in\indexset}$ with
		finite second-order moments, the \gls{covariance} \gls{function} is 
		defined as \cite{Rasmussen2006Gaussian}
		\begin{align}
				\nonumber
				\covfunc{\feature}{\timeidx}{\timeidx'} 
		& \defeq	\cov{\feature(\timeidx)}{\feature(\timeidx')} \nonumber \\ 
		& = 
		\expect\!\big[(\feature(\timeidx)-\expect[\feature(\timeidx)])
		                  (\feature(\timeidx')-\expect[\feature(\timeidx')])\big],
		\qquad \timeidx,\timeidx'\in\indexset. \nonumber
		\end{align}
		See also: \gls{covariance}, \gls{function}, \gls{stochproc}. },
	first={covariance function},
	type=math,
	plural={covariance functions},
	firstplural={covariance functions},
	text={covariance function}
}

\newglossaryentry{covariance}
{name={covariance}, 
	description={The\index{covariance} covariance between two real-valued 
 		\glspl{rv} $x$ and $y$, defined on a common \gls{probspace}, measures their linear 
 		dependence. It is defined as 
			$$
			\cov{x}{y} = \expect\big\{ \big(x - \expect\{ x\} \big)\big(y - \expect\{y\} \big)\big\}.
			$$
		A positive covariance indicates that $x$ and $y$ tend to increase together, 
		while a negative covariance suggests that one tends to increase as the other decreases. 
		If $\cov{x}{y} = 0$, the \glspl{rv} are said to be uncorrelated, though not necessarily statistically 
		independent. See Fig.\ \ref{fig:covariance-examples_dict} for visual illustrations.
		\begin{figure}[H]
		\begin{tikzpicture}
		\begin{scope}[shift={(0,0)}]
			\begin{axis}[
				width=4.5cm, height=4.5cm,
				title={$\cov{x}{y} <0$},
				xlabel={$x$}, ylabel={$y$},
				xmin=-3, xmax=3, ymin=-3, ymax=3,
				xtick=\empty, ytick=\empty,
				axis lines=middle, enlargelimits
				]
				\addplot+[only marks, mark=*, samples=50, domain=-2:2] 
				({x}, {-x + rand});
			\end{axis}
			\node at (1.5,-1) {(a)};
		\end{scope}
		\begin{scope}[shift={(5.2cm,0)}]
			\begin{axis}[
				width=4.5cm, height=4.5cm,
				title={$\cov{x}{y} =0$}, 
				xlabel={$x$}, ylabel={$y$},
				xmin=-3, xmax=3, ymin=-3, ymax=3,
				xtick=\empty, ytick=\empty,
				axis lines=middle, enlargelimits
				]
				\addplot+[only marks, mark=*, samples=50, domain=-2:2] 
				({x}, {rand});
			\end{axis}
			\node at (1.5,-1) {(b)};
		\end{scope}
		\begin{scope}[shift={(10.4cm,0)}]
			\begin{axis}[
				width=4.5cm, height=4.5cm,
				title={$\cov{x}{y} > 0$},
				xlabel={$x$}, ylabel={$y$},
				xmin=-3, xmax=3, ymin=-3, ymax=3,
				xtick=\empty, ytick=\empty,
				axis lines=middle, enlargelimits
				]
				\addplot+[only marks, mark=*, samples=50, domain=-2:2] 
				({x}, {x + rand});
			\end{axis}
			\node at (1.5,-1) {(c)};
		\end{scope}
		\end{tikzpicture}
			\caption{\Glspl{scatterplot} illustrating \glspl{realization} from three different \glspl{probmodel} for two 
				\glspl{rv} with different covariance values. (a) Negative. (b) Zero. (c) Positive.}
			\label{fig:covariance-examples_dict}
		\end{figure}
		See also: \gls{probmodel}, \gls{expectation}. },
	first={covariance},
	text={covariance},
	type=math
}

\newglossaryentry{covmtx}
{name={covariance matrix}, 
	description={The\index{covariance matrix} \gls{covariance} \gls{matrix} of 
		an \gls{rv} $\featurevec \in \mathbb{R}^{\featuredim}$ is defined as 
		the \gls{expectation} (if it exists):
		$$\covmtx{\featurevec} \defeq \expect \bigg \{ \big( \featurevec - \expect \big\{ \featurevec \big\} \big)  
		\big(\featurevec - \expect \big\{ \featurevec \big\} \big)\,^{T} \bigg\}.$$
				\\
		See also: \gls{covariance}, \gls{matrix}, \gls{rv}.},
	first={covariance matrix},
	plural={covariance matrices},
	firstplural={covariance matrices},
	type=math, 
	text={covariance matrix} 
}

\newglossaryentry{samplevariance}
{name={sample variance}, 
	description={Consider a \gls{dataset} $\dataset$ consisting of \glspl{datapoint} 
		characterized by scalar values $\truelabel^{(1)}, \,\ldots, \,\truelabel^{(\samplesize)} 
		\in \mathbb{R}$. The\index{sample variance} \gls{sample} \gls{variance} of $\dataset$ 
		is defined as the \gls{variance} with respect to the \gls{empiricaldistribution} 
		$\probdist^{(\dataset)}$ induced by $\dataset$. It is given explicitly by 
		$$\frac{1}{\samplesize} \sum_{\sampleidx=1}^{\samplesize} (\truelabel^{(\sampleidx)}\!-\!\widehat{\truelabel})^2,$$ 
		where $\widehat{\truelabel}$ is the \gls{samplemean} of $\dataset$. The 
		\gls{sample} \gls{variance} is a special case of the \gls{samplecovmtx} 
		when each \gls{datapoint} is characterized by a single numerical value.
		\\
		See also: \gls{samplecovmtx}.},
	first={sample variance},
	plural={sample variances},
	firstplural={sample variances},
	type=math, 
	text={sample variance}
}

\newglossaryentry{samplecovmtx}
{name={sample covariance matrix}, 
	description={Consider a \gls{dataset} $\dataset$ consisting of \glspl{datapoint} 
		characterized by \glspl{featurevec} $\featurevec^{(1)}, \,\ldots, \,\featurevec^{(\samplesize)} \in \mathbb{R}^{\nrfeatures}$.	
		The\index{sample covariance matrix} \gls{sample} \gls{covmtx} 
		of $\dataset$ is defined as the \gls{covmtx} with 
		respect to the \gls{empiricaldistribution} $\probdist^{(\dataset)}$ 
		induced by $\dataset$. It is given explicitly by 
		$$\samplecovmtxgeneric = \frac{1}{\samplesize} \sum_{\sampleidx=1}^{\samplesize} (\featurevec^{(\sampleidx)}\!-\!\widehat{\vm}) (\featurevec^{(\sampleidx)}\!-\!\widehat{\vm})\,^{T}.$$ 
		Here, we use the \gls{samplemean} $\samplemeanvecgeneric$. 
		\\
		See also: \gls{covariance}, \gls{matrix}, \gls{rv}.},
	first={sample covariance matrix},
	plural={sample covariance matrices},
	firstplural={sample covariance matrices},
	type=math, 
	text={sample covariance matrix} 
}

\newglossaryentry{eigenvalue}
{name={eigenvalue}, 
	description={We\index{eigenvalue} refer to a number $\eigvalgen \in \mathbb{R}$ 
		as an eigenvalue of a square \gls{matrix} $\mathbf{A} \in \mathbb{R}^{\featuredim \times \featuredim}$ 
		if there exists a nonzero \gls{vector} $\vx \in \mathbb{R}^{\featuredim} \setminus \{ \mathbf{0} \}$ 
		such that $\mathbf{A} \vx = \eigvalgen \vx$ (see Fig.\ \ref{fig_eigenvalue_dict}). 
		\begin{figure}[H]
			\centering
			\begin{tikzpicture}[>=stealth, line width=0.8pt]
				\draw[->] (0,0) -- (2,1)
				node[midway, above left] {$\vx$};
				\node[
				minimum width=1.6cm,
				minimum height=1.2cm
				] (A) at (3.0,0.5) {$\mA$};					
				\draw[->] (2,0.5) -- (2.5,0.5);
				\draw[->] (3.5,0.5) -- (4.0,0.5);
				\draw[->] (4.6,0) -- (5.6,0.5)
				node[midway, above] {$\lambda \vx$};
			\end{tikzpicture}
		\caption{\Gls{eigenvector} corresponding to the eigenvalue $\eigvalgen$.
			\label{fig_eigenvalue_dict} }
		\end{figure}
		See also: \gls{matrix}, \gls{eigenvector}. },
	first={eigenvalue},
	plural={eigenvalues},
	firstplural={eigenvalues},
	type=math, 
	text={eigenvalue} 
}

\newglossaryentry{eigenvector}
{name={eigenvector},
	description={An\index{eigenvector} 
		eigenvector of a \gls{matrix} $\mA \in \mathbb{R}^{\featuredim \times \featuredim}$ 
		is a nonzero \gls{vector} $\vx \in \mathbb{R}^{\featuredim} \setminus \{ \mathbf{0} \}$ 
		such that $\mA  \vx = \lambda \vx$ with some \gls{eigenvalue} $\lambda$.
		Eigenvectors to the \gls{eigenvalue} $\lambda$ span a \gls{subspace} of $\mathbb{R}^{\featuredim}$,
		namely the \gls{nullspace} of $\mA  - \lambda \mI$.
				\\
		See also: \gls{matrix}, \gls{vector}, \gls{eigenvalue}.},
	first={eigenvector},
	type=math,
	plural={eigenvectors},
	text={eigenvector} 
}

\newglossaryentry{evd}
{name={eigenvalue decomposition (EVD)}, 
	description={An\index{eigenvalue decomposition (EVD)} EVD
		for a square \gls{matrix} $\mA \in \mathbb{R}^{\dimlocalmodel \times \dimlocalmodel}$ 
		is a factorization of the form
		$$\mA = \mathbf{V} {\bm \Lambda} \mathbf{V}^{-1}.$$ 
		The columns of the \gls{matrix} $\mV = \big( \vv^{(1)}, \,\ldots, \,\vv^{(\dimlocalmodel)} \big)$ are the 
		\glspl{eigenvector} of the \gls{matrix} $\mA$. The diagonal \gls{matrix} 
		${\bm \Lambda} = {\rm diag} \big\{ \eigval{1}, \,\ldots, \,\eigval{\dimlocalmodel} \big\}$ 
		contains the \glspl{eigenvalue} $\eigval{\featureidx}$ corresponding to the \glspl{eigenvector} $\vv^{(\featureidx)}$. 
		\Glspl{matrix} that allow for an EVD are referred to as \gls{diagonalizable}.
				\\
		See also: \gls{matrix}, \gls{eigenvector}, \gls{eigenvalue}, \gls{diagonalizable}.},
	first={eigenvalue decomposition (EVD)},
	type=math, 
	plural={EVDs},
	firstplural={eigenvalue decompositions (EVDs)},
	text={EVD} 
}

\newglossaryentry{singularvalue}
{name={singular value}, 
 	description={See\index{singular value} \gls{svd}.},
 	first={singular value},
	type=math,
	plural={singular values},
	firstplural={singular values},
	text={singular value} 
}

\newglossaryentry{svd}
{name={singular value decomposition (SVD)}, 
  	description={The\index{singular value decomposition (SVD)} SVD  
  		for a \gls{matrix} $\mA \in \mathbb{R}^{\samplesize \times \dimlocalmodel}$ 
		is a factorization of the following form:
		$$\mA = \mathbf{V} {\bm \Lambda} \mathbf{U}\,^{T}$$ 
		with orthonormal \glspl{matrix} $\mV = \big(\vv^{(1)},\,\ldots,\,\vv^{(\samplesize)}\big) \in \mathbb{R}^{\samplesize \times \samplesize}$ 
		and $\mU = \big( \vu^{(1)},\,\ldots,\,\vu^{(\dimlocalmodel)} \big) \in \mathbb{R}^{\dimlocalmodel \times \dimlocalmodel}$ 
		\cite{GolubVanLoanBook} (see Fig.\ \ref{fig_svd_dict}). 
		\begin{figure}[H]
		 	\centering	
		 	\begin{tikzpicture}[>=latex, line cap=round, line join=round, scale=1]
		 	\def\xL{0}
		 	\def\xR{6}
		 	\def\len{2.0}
		 	\def\sone{1.6}
		 	\def\stwo{0.7}
		 	\draw[thick,->] (\xL,0) -- ++(\len,0) node[midway,below] {$\vu^{(1)}$};
			\draw[thick,->] (\xL,0) -- ++(0,\len) node[midway,left] {$\vu^{(2)}$};
			\draw[dashed,->] (\xL+2.9,0) -- (\xR-0.9,0);
			\begin{scope}[shift={(\xR+0.5,0)},rotate=30]
			\draw[thick,->] (0,0) -- (\sone*\len,0) node[pos=1,right] {$\lambda_1 \vv^{(1)}$};
			\draw[thick,->] (0,0) -- (0,\stwo*\len) node[pos=1,above] {$\lambda_2 \vv^{(2)}$};
			\end{scope}
		 	\end{tikzpicture}
		\caption{Orthonormal \glspl{matrix} $\mV$ and $\mU$. \label{fig_svd_dict} }
		\end{figure}
		The \gls{matrix} ${\bm \Lambda} \in \mathbb{R}^{\samplesize \times \dimlocalmodel}$ is 
		only nonzero along the main diagonal, whose entries $\Lambda_{\featureidx,\featureidx}$ 
		are nonnegative and referred to as \glspl{singularvalue}.
		\\
		See also: \gls{matrix}. },
	first={singular value decomposition (SVD)},
	type=math,
	plural={SVDs},
	firstplural={singular value decompositions (SVDs)},
	text={SVD} 
}

\newglossaryentry{marginaldist}
{name={marginal distribution}, 
	description={A\index{marginal distribution} marginal distribution 
		refers to the \gls{probdist} of a subset of the 
		\glspl{rv} belonging to a \gls{stochproc} \cite[p. 266]{BillingsleyProbMeasure}, \cite{papoulis}. 
		For example, consider two \glspl{rv} $(\feature,\truelabel)$ with a joint 
		\gls{probdist} $\probdist^{(\feature,\truelabel)}$. This 
		joint \gls{probdist} fully determines the marginal distributions 
		of $\feature$ and $\truelabel$. The process of computing the marginal 
		distribution of a subset of \glspl{rv} is referred to as \gls{marginalization} \cite{GraphModExpFamVarInfWainJor}.
		\\
		See also: \gls{probdist}, \gls{rv}, \gls{stochproc}, \gls{marginalization}.},
	first={marginal distribution},
	plural={marginal distributions},
	firstplural={marginal distributions},
	type=math,
	text={marginal distribution}
}

\newglossaryentry{marginalization}
{name={marginalization}, 
	description={The term marginalization refers to the\index{marginalization} process of computing the \gls{probdist} 
		of a subset of \glspl{rv} from the joint \gls{probdist} of a \gls{stochproc} 
		\cite{GraphModExpFamVarInfWainJor}.
		\\
		See also: \gls{probdist}, \gls{rv}, \gls{marginaldist}.},
	first={marginalization},
	plural={marginalizations},
	firstplural={marginalizations},
	type=math,
	text={marginalization}
}

\newglossaryentry{probdist}
{name={probability distribution}, 
	description={To\index{probability distribution} analyze \gls{ml} methods, it can be useful 
		to interpret \glspl{datapoint} as \gls{iid} \glspl{realization} of an \gls{rv}. The typical 
		properties of such \glspl{datapoint} are then governed by the \gls{probability} distribution 
		of this \gls{rv}. The \gls{probability} distribution of a binary \gls{rv} $\truelabel \in \{0,1\}$ 
		is fully specified by the \glspl{probability} $\prob{\truelabel = 0}$ and 
		$\prob{\truelabel=1}\!=\!1\!-\!\prob{\truelabel=0}$. The \gls{probability} 
		distribution of a real-valued \gls{rv} $\feature \in \mathbb{R}$ might be specified 
		by a \gls{pdf} $p(\feature)$ such that $\prob{ \feature \in [a,b] } \approx  p(a) |b-a|$. 
	    	In the most general case, a \gls{probability} distribution is defined by a \gls{probability} \gls{measure} 
		\cite{BillingsleyProbMeasure}, \cite{GrayProbBook}.
	    		\\
		See also: \gls{iid}, \gls{realization}, \gls{rv}, \gls{probability}, \gls{pdf}.},
	first={probability distribution},
	plural={probability distributions},
	firstplural={probability distributions},
	type=math,
	text={probability distribution}
}

\newglossaryentry{pdf}
{name={probability density function (pdf)},
	description={The\index{probability density function (pdf)} pdf 
		$\pdf{\feature}{\cdot}$ of a \gls{continuous} real-valued \gls{rv} $\feature \in \mathbb{R}$ 
		allows us to compute the \gls{probability} $\prob{\feature \in \mathcal{B}}$ (of 
		the \gls{event} $\mathcal{B} \subseteq \mathbb{R}$)
		via a \gls{LebesgueIntegral} as follows \cite[Ch. 3]{BertsekasProb}:
	 	$$\prob{\feature \in \mathcal{B}} = \int_{\mathcal{B}} \pdf{\feature}{\eta} d 
	 	\eta.$$ 
	 	This definition extends naturally to a (\gls{continuous}) \gls{vector}-valued \gls{rv} 
		$\featurevec \in \mathbb{R}^{\nrfeatures}$, as the \gls{LebesgueIntegral} 
	 	is defined for $\mathbb{R}^{\nrfeatures}$ with any \gls{dimension} $\nrfeatures$. 
        		\\
		See also: \gls{rv}, \gls{probability}, \gls{vector}, \gls{probdist}, \gls{measurable}.},
	first={probability density function (pdf)},
	plural={pdfs},
	type=math,
	text={pdf},
	firstplural={probability density functions (pdfs)}
}

\newglossaryentry{convex}
{name={convex},
	description={A\index{convex} subset $\mathcal{C} \subseteq \mathbb{R}^{\featuredim}$ of the 
		\gls{euclidspace} $\mathbb{R}^{\featuredim}$ is referred to 
		as convex if it contains the line segment between any two points 
		$\vx, \vy\!\in\!\cluster$ in that set, i.e., 
		$$ \alpha \vx + (1-\alpha) \vy \in \mathcal{C} \quad \mbox{for any } \alpha \in [0,1]. $$
		Similarly, a \gls{function}  
		$f\!:\!\mathbb{R}^{\dimlocalmodel}\!\rightarrow\!\mathbb{R}$ 
		is convex if its \gls{epigraph} $\big\{ \big( \weights\,^{T},t \big)\,^{T}\!\in\!\mathbb{R}^{\dimlocalmodel\!+\!1}\!:\!t\!\geq\!f(\weights) \}$ 
		is a convex set \cite{BoydConvexBook}. We illustrate one example of a convex set 
		and a convex \gls{function} in Fig.\ \ref{fig_convex_set_function_dict}. 
		\begin{figure}[H]
		\begin{center}
			\begin{tikzpicture}
				\fill[blue!20, opacity=0.5] (-3,0) ellipse (2 and 1.2); 
				\draw[thick] (-3,0) ellipse (2 and 1.2);
				\filldraw[black] (-3.7,0.2) circle (2pt) node[left] {$\vw$};
				\filldraw[black] (-2.3,-0.5) circle (2pt) node[right] {$\vw'$};
				\draw[thick] (-3.7,0.2) -- (-2.3,-0.5);
				\node at (-1.2,-1.0) {$\mathcal{C}$};
				\node at (-3,-2.4) {(a)};
				\begin{scope}[shift={(5,-1)}]
					\draw[thick, domain=-2:2, smooth, variable=\x] 
					plot ({\x}, {0.5*\x*\x});
					\fill[blue!30, opacity=0.5] 
					plot[domain=-1.5:1.5, smooth] ({\x}, {0.5*\x*\x}) -- 
					(2, {0.5*2*2}) -- 
					(-2, {0.5*2*2}) -- 
					cycle;
					\node at (0,-0.4) {$f(\weights)$};
					\node at (0,-1.4) {(b)};
				\end{scope}
			\end{tikzpicture}
			\vspace*{-8mm}
		\end{center}
		\caption{(a) Convex set $\cluster \subseteq \mathbb{R}^{\dimlocalmodel}$. 
			(b) Convex \gls{function} $f: \mathbb{R}^{\dimlocalmodel} \rightarrow \mathbb{R}$.\label{fig_convex_set_function_dict}}
		\end{figure}
		See also: \gls{euclidspace}, \gls{function}, \gls{epigraph}.},
	first={convex},
	type=math, 
	text={convex}
}

\newglossaryentry{lln}
{name={law of large numbers},
	description={The\index{law of large numbers} law of large numbers refers to the 
		\gls{convergence} of the average of an increasing (large) number of \gls{iid} \glspl{rv} 
		to the \gls{mean} of their common \gls{probdist}. Different instances of the 
		law of large numbers are obtained by using different notions of \gls{convergence} \cite{papoulis}.
				\\
		See also: \gls{convergence}, \gls{iid}, \gls{rv}, \gls{mean}, \gls{probdist}.},
	first={law of large numbers},
	type=math,
	text={law of large numbers}
}

\newglossaryentry{renyidiv}
{name={R\'enyi divergence}, 
sort={Renyi},
	description={The R\'enyi divergence\index{R\'enyi divergence} measures the (dis)similarity 
		between two \glspl{probdist} \cite{RenyiInfo95}.
				\\
		See also: \gls{probdist}.}, 
	first={R\'enyi divergence}, 
	type=math,
	text={R\'enyi divergence}
} 

\newglossaryentry{nonsmooth}
{name={non-smooth},
	description={We\index{non-smooth} refer to a \gls{function} as non-smooth if it is not 
		\gls{smooth} \cite{nesterov04}.
				\\
		See also: \gls{function}, \gls{smooth}.},
	first={non-smooth},
	type=math,
	text={non-smooth}
}

\newglossaryentry{smooth}
{name={smooth},
	description={A\index{smooth} real-valued \gls{function} $f: \mathbb{R}^{\dimlocalmodel} \rightarrow \mathbb{R}$ 
		is smooth if it is \gls{differentiable} and its \gls{gradient} $\nabla f(\weights)$ is continuous at all $\weights \in \mathbb{R}^{\dimlocalmodel}$  \cite{nesterov04}, \cite{CvxBubeck2015}. 
		A smooth \gls{function} $f$ is referred to as $\beta$-smooth if the \gls{gradient} 
		$\nabla f(\weights)$ is Lipschitz continuous with Lipschitz constant $\beta$, i.e., 
		$$\normgeneric{\nabla f(\weights) - \nabla f(\weights')}{} \leq \beta \normgeneric{\weights - \weights'}{} \quad \mbox{for any } \weights,\weights' \in \mathbb{R}^{\dimlocalmodel}.$$ 
		The constant $\beta$ quantifies the smoothness of the \gls{function} $f$, i.e., the smaller the $\beta$, 
		the smoother $f$ is. \Glspl{optproblem} with a smooth \gls{objfunc} can be solved effectively by 
		\glspl{gdmethod}. Indeed, \glspl{gdmethod} approximate the \gls{objfunc} locally around a current 
		choice $\weights$ using its \gls{gradient}. This approximation works well if the \gls{gradient} does 
	    	not change too rapidly. We can make this informal claim precise by studying the effect of a single 
	    	\gls{gradstep} with \gls{stepsize} $\lrate=1/\beta$ (see Fig.\ \ref{fig_gd_smooth_dict}). 
	    	\begin{figure}[H] 
	    	\begin{center} 
	    	\begin{tikzpicture}[scale=0.8, x=0.6cm,y=0.05cm]
	    		\def\hshift{0.5} 
	    		\draw[thick, dashed, domain=\hshift:8+\hshift, smooth, variable=\x] plot ({\x}, {\x^2}); 
	    		\coordinate (w) at (\hshift,{\hshift*\hshift}); 
	    		\coordinate (wkplus1) at (4+\hshift,{(4+\hshift)^2}); 
	    		\coordinate (wk) at (8+\hshift,{(8+\hshift)^2}); 
  				\draw[line width=1pt, transform canvas={yshift=-1pt}] (wk) -- +(-2, -{4*(8 + \hshift)} ) -- +(1, {2*(8 + \hshift)}); 
 				\draw[line width=1pt, transform canvas={yshift=-1pt}] (w) -- +(-1, {-{2*\hshift}} ) -- +(1, {+{2*\hshift}})  node[below] {$\nabla f(\weights)$};
	    		\filldraw (wk) circle (2pt) node[above left] {$\weights^{(\iteridx)}$} node[below right, xshift=-15pt,yshift=-15pt] {$\nabla f(\weights^{(\iteridx)})$} ;
	    		\filldraw (w) circle (2pt) node[above right] {$\weights$} ;
	    		\filldraw (wkplus1) circle (2pt) node[below right] {$\weights^{(\iteridx+1)}\!=\!\weights^{(\iteridx)}\!-\!(1/\beta)\nabla f(\weights^{(\iteridx)})$};
	    		\draw[dashed] (wk) -- ($(8,0) + (wk)$) ; 
	    		\draw[dashed] (wkplus1) -- ($(12,0) + (wkplus1)$) ; 
	    		 \draw[<->, thick] ($(4,0) + (wk)$) -- ($(8,0) + (wkplus1)$) 
	    		node[midway, right] {$ f\big(\weights^{(\iteridx)}\big)\!-\!f\big(\weights^{(\iteridx+1)}\big)\!\geq\!\frac{1}{2\beta}\normgeneric{\nabla f(\weights^{(\iteridx)})}{2}^{2}$};
	    	\end{tikzpicture}
	    	\end{center}
	    	\caption{Consider an \gls{objfunc} $f(\weights)$ that is $\beta$-smooth. 
	    		Taking a \gls{gradstep}, with \gls{stepsize} $\lrate = 1/\beta$, decreases the 
	    		objective by at least ${1}/{2\beta}\normgeneric{\nabla f(\weights^{(\iteridx)})}{2}^{2}$ \cite{nesterov04}, \cite{CvxBubeck2015}, \cite{CvxAlgBertsekas}. 
	    		Note that the \gls{stepsize} $\lrate = 1/\beta$ becomes larger for smaller $\beta$. Thus, 
	    		for smoother \glspl{objfunc} (i.e., those with smaller $\beta$), 
				we can take larger steps. \label{fig_gd_smooth_dict}}
	    	\end{figure}
		See also: \gls{function}, \gls{differentiable}, \gls{gradient},  \gls{gdmethod}.},
	first={smooth},
	type=math,
	text={smooth}
}

\newglossaryentry{metricspace}
{name={metric space},
	description={A \gls{metric} space\index{metric space} is a set  
        		$\featurespace$ equipped with a \gls{function} (referred to as a \gls{metric})   
        		$\metric{\cdot}{\cdot}: \featurespace \times \featurespace \rightarrow \mathbb{R}_{+}$ 
        		that satisfies the following requirements for all 
		$\featurevec, \featurevec', \featurevec'' \in \featurespace$:
        		\begin{enumerate}
            		\item Nonnegativity: $\metric{\featurevec}{\featurevec'} \ge 0$;
           		\item Identity: $\metric{\featurevec}{\featurevec'} = 0$ 
                  		if and only if $\featurevec = \featurevec'$;
            		\item Symmetry: $\metric{\featurevec}{\featurevec'} 
                  		= \metric{\featurevec'}{\featurevec}$;
            		\item Triangle inequality: 
                  		$\metric{\featurevec}{\featurevec''} \le 
                  		\metric{\featurevec}{\featurevec'} + \metric{\featurevec'}{\featurevec''}$.
        		\end{enumerate}
        		Formally, a \gls{metric} space is a pair $(\featurespace, \metric{\cdot}{\cdot})$ that 
		satisfies the above requirements \cite{BasicRealAnalysis}. 
		\begin{figure}[H]
		\centering
		\begin{tikzpicture}[>=stealth, x=0.5cm, y=1cm]
		\node[font=\footnotesize] at (2.5,3.6) {\gls{euclidspace} $\mathbb{R}^2$};
		\node[font=\footnotesize] at (12.5,3.6) {\gls{undirectedgraph} $\pair{\nodes}{\edges}$};
		\begin{scope}[shift={(0,0)}]
			\draw[->] (0,0) -- (5.2,0) node[below right] {$\feature_1$};
			\draw[->] (0,0) -- (0,3.2) node[left] {$\feature_2$};
			\coordinate (X) at (1.1,0.9);
			\coordinate (Y) at (3.8,2.1);
			\fill (X) circle (1.2pt) node[below left] {$\featurevec$};
			\fill (Y) circle (1.2pt) node[above right] {$\featurevec'$};
			\draw[dashed] (X) -- (Y)
			node[midway, below right, xshift=1pt] {$\normgeneric{\featurevec-\featurevec'}{2}$};
			\node at (3, -1) {(a)};
		\end{scope}
		\begin{scope}[shift={(9.0,0)}]
			\coordinate (A) at (1.0,0.6);
			\coordinate (B) at (3.1,0.9);
			\coordinate (C) at (2.2,2.6);
			\coordinate (D) at (4.8,2.2);
			\coordinate (E) at (0.4,2.1);
			\draw (A)--(B)--(C)--(E)--(A);
			\draw (C)--(D)--(B);
			\fill (A) circle (1.2pt) node[below left] {$\nodeidx$};
			\fill (D) circle (1.2pt) node[above right] {$\nodeidx'$};
			\draw[very thick] (A)--(B)--(D);
			\node[font=\footnotesize, above] at (5.0,0.7) {$\metric{\nodeidx}{\nodeidx'}\!=\!2$};
			\fill (B) circle (1.2pt);
			\fill (C) circle (1.2pt);
			\fill (E) circle (1.2pt);
			\node at (3, -1) {(b)};
		\end{scope}
		\end{tikzpicture}
		\caption{Examples of \gls{metric} spaces. (a) \gls{euclidspace} $\mathbb{R}^2$ with 
			the \gls{eucliddist} as a \gls{metric}. (b) \Gls{undirectedgraph} $\pair{\nodes}{\edges}$ with 
			the shortest-path \gls{distance} as a \gls{metric}.\label{fig:metric_space_examples_dict}}
	    	\end{figure}
		A prominent example of a \gls{metric} space 
		is the \gls{euclidspace} equipped with a \gls{metric} given by the \gls{eucliddist} 
		$\metric{\featurevec}{\featurevec'} = \normgeneric{\featurevec - \featurevec'}{2}$. 
		Another well-known example of a \gls{metric} space is an \gls{undirectedgraph}  
        		$\graph=\pair{\nodes}{\edges}$, with the \gls{metric} $\metric{\nodeidx}{\nodeidx'}$ 
		defined by the length of the shortest path connecting nodes $\nodeidx$ and $\nodeidx'$. 
			\\
		See also: \gls{euclidspace}, \gls{undirectedgraph}, \gls{featurespace}. },
    	text={metric space}, 
	first={metric space}, 
	type=math, 
	firstplural={metric spaces}, 
	plural={metric spaces}
}

\newglossaryentry{gradstep}
{name={gradient step}, 
	description={Given a \gls{differentiable} 
		real-valued \gls{function} $f(\cdot): \mathbb{R}^{\nrfeatures} \rightarrow \mathbb{R}$ 
		and a \gls{vector} $\weights \in \mathbb{R}^{\nrfeatures}$, the \gls{gradient} step\index{gradient step} 
		updates $\weights$ by adding the scaled negative \gls{gradient} 
		$\nabla f(\weights)$ to obtain the new \gls{vector} (see Fig.\ \ref{fig_basic_GD_step_single_dict}) as follows:
		\begin{equation}
			\label{equ_def_gd_basic_dict} 
			\widehat{\weights}  \defeq \weights - \lrate \nabla f(\weights).
		\end{equation}
		\begin{figure}[H]
			\begin{center}
				\begin{tikzpicture}[scale=0.8]
					\draw[loosely dotted] (-4,0) grid (4,4);
					\draw[blue, ultra thick, domain=-4.1:4.1] plot (\x,  {(1/4)*\x*\x});
					\draw[red, thick, domain=2:4.7] plot (\x,  {2*\x - 4});
					\draw[<-] (4,4) -- node[right] {$\normgeneric{\nabla f(\weights^{(\itercntr)})}{}$} (4,2);
					\draw[->] (4,4) -- node[above] {$-\lrate \nabla f(\weights^{(\itercntr)})$} (2,4);
					\draw[<-] (4,2) -- node[below] {$1$} (3,2) ;
					\node[left] at (-4.1, 4.1) {$f(\cdot)$}; 
					\draw[shift={(0,0)}] (0pt,2pt) -- (0pt,-2pt) node[below] {$\overline{\weights}$};
					\draw[shift={(4,0)}] (0pt,2pt) -- (0pt,-2pt) node[below] {$\weights$};
					\draw[shift={(2,0)}] (0pt,2pt) -- (0pt,-2pt) node[below] {$\mathcal{T}^{(f,\lrate)}(\weights)$};
				\end{tikzpicture}
			\end{center}
		\caption{Basic \gls{gradient} step \eqref{equ_def_gd_basic_dict} mapping a given \gls{vector} $\weights$ 
			to the updated \gls{vector} $\weights'$.}
			\label{fig_basic_GD_step_single_dict}
		\end{figure}
		More formally, the \gls{gradient} step amounts to applying the \gls{operator} 
		$$\mathcal{T}^{(f,\lrate)}(\cdot): \mathbb{R}^{\nrfeatures} \rightarrow \mathbb{R}^{\nrfeatures}:
		 	 \weights \mapsto \weights - \lrate \nabla f(\weights).$$ 
		For a \gls{smooth} and \gls{convex} \gls{objfunc} $f$, any of 
		its minimzers $\widehat{\weights}$ is a \gls{fixedpoint} of 
		$\mathcal{T}^{(f,\lrate)}$, i.e., 
		$$ \mathcal{T}^{(f,\lrate)} \widehat{\weights} = \widehat{\weights} \quad \mbox{for any } 
		\lrate >0.$$ What is more, 
		for a sufficiently small \gls{learnrate} $\lrate > 0$, one can verify 
		that in this case $\mathcal{T}^{(f,\lrate)}$ becomes a \gls{firmlynonexpansiveop} 
		\cite[Corollary 18.16]{BausckeCombette}. Moreover, for 
		a \gls{strcvx} and \gls{smooth} \gls{function} $f(\cdot)$ and an 
		appropriate choice of \gls{learnrate}, one can verify that 
		$\mathcal{T}^{(f,\lrate)}$ is a \gls{contractop} \cite[Th. 2.1.14]{nesterov04}.
		Note that the \gls{gradient} step \eqref{equ_def_gd_basic_dict} optimizes 
		locally—in a \gls{neighborhood} whose size is determined by the \gls{stepsize} 
		$\lrate$—a linear approximation to the \gls{function} $f(\cdot)$. 
		A natural \gls{generalization} of \eqref{equ_def_gd_basic_dict} is to locally 
		optimize the \gls{function} itself—instead of its linear approximation—such that
		\begin{align} 
			\label{equ_approx_gd_step_dict}
			\widehat{\weights} = \argmin_{\weights' \in \mathbb{R}^{\dimlocalmodel}} f(\weights')\!+\!\frac{1}{\lrate}\normgeneric{\weights-\weights'}{2}^2. 
		\end{align}
		We intentionally use the same symbol $\lrate$ for the \gls{parameter} in \eqref{equ_approx_gd_step_dict} 
		as we used for the \gls{stepsize} in \eqref{equ_def_gd_basic_dict}. The larger the $\lrate$ we choose in 
		\eqref{equ_approx_gd_step_dict}, the more progress the update will make toward reducing the 
		\gls{function} value $f(\widehat{\weights})$. Note that, much like the \gls{gradient} 
		step \eqref{equ_def_gd_basic_dict}, the update \eqref{equ_approx_gd_step_dict} also 
		defines an \gls{operator} that depends on the \gls{function} $f(\cdot)$ and on
		the \gls{learnrate} $\lrate$. For a \gls{convex} \gls{function} $f(\cdot)$, 
		this \gls{operator} is known as the \gls{proxop} of $f(\cdot)$ \cite{ProximalMethods}. 
					\\ 
		See also: \gls{differentiable}, \gls{gradient}, \gls{stepsize}, \gls{proxop}, \gls{sgd}, \gls{fixedpointeq}.},
	first={gradient step},
	type=math,
	firstplural={gradient steps},
	plural={gradient steps},
	text={gradient step}
}

\newglossaryentry{mirrordescent}
{name={mirror descent},
 	description={Mirror descent\index{mirror descent} is an iterative \gls{optmethod}  
		obtained by generalizing the \gls{gradstep}. The 
 		\gls{gradstep} for minimizing a \gls{differentiable} \gls{objfunc} $f:\mathbb{R}^\nrfeatures \to \mathbb{R}$ 
 		can be written as
		\[
		\weights^{(\itercntr+1)} = \argmin_{\weights \in \mathbb{R}^\nrfeatures}
 		\left \langle \nabla f(\weights^{(\itercntr)}),\weights \right \rangle
 		+ \frac{1}{2\lrate}\normgeneric{\weights-\weights^{(\itercntr)}}{2}^{2}.
 		\]
 		Thus, a \gls{gradstep} minimizes a linearization of $f(\cdot)$ penalized 
 		by a scaled squared \gls{euclidnorm} $\frac{1}{\lrate}\normgeneric{\weights-\weights^{(\itercntr)}}{2}^{2}$. 
 	 	The scaling factor is the inverse of the \gls{learnrate} $\lrate$ used in the \gls{gradstep}. 
		Mirror descent replaces the squared \gls{euclidnorm} with a 
 		\gls{bregmandivergence} $\bregmandiv{\cdot}{\cdot}{\phi}$ 
 		induced by a \gls{strictlyconvex} \gls{function} $\phi(\cdot)$. This 
 		mirror \gls{map} is typically defined on a \gls{convex} set 
 		$\paramspace \subseteq \mathbb{R}^{\nrfeatures}$ and is \gls{differentiable} 
 		and \gls{strictlyconvex} on the interior of $\paramspace$ \cite{CvxBubeck2015}. 
 		The resulting update becomes 
 		\[
 		\weights^{(\itercntr+1)} 
 		= \argmin_{\weights \in \paramspace}
 		\left \langle \nabla f(\weights^{(\itercntr)}),\weights \right \rangle
 		+ \frac{1}{\lrate} \bregmandiv{\weights}{\weights^{(\itercntr)}}{\phi}.
 		\]
 		See also: \gls{gradstep}, \gls{bregmandivergence}, \gls{innerproduct}, \gls{zerogradientcondition}, \gls{proxop}.},
 	first={mirror descent},
 	type=math,
 	text={mirror descent}
 }

\newglossaryentry{bregmandivergence}
{name={Bregman divergence},
	description={The Bregman divergence\index{Bregman divergence} induced by a 
		\gls{convex}, \gls{differentiable} \gls{function} $\phi(\cdot)$ 
		is defined as
		\[
		\bregmandiv{\weights}{\weights'}{\phi}
		\defeq \phi(\weights)-\phi(\weights')-\langle\nabla\phi(\weights'),\weights-\weights'\rangle
		\]
		for $\weights,\weights'$ in the \gls{domain} $\paramspace$ of $\phi$. 
		It measures the deviation of $\phi(\weights)$ from its first-order Taylor 
		approximation around $\weights'$ and is in general neither symmetric nor a \gls{metric}. 
		\begin{figure}[H]
			\centering
			\begin{tikzpicture}[scale=3.4]
			\draw[thick] (0,0)--(1,0)--(0.5,1)--cycle;
			\node at (0.5,0.4) {$\paramspace$};
			\draw[rounded corners=0.12cm] (0.06,0.04)--(0.94,0.04)--(0.50,0.92)--cycle;
			\draw[rounded corners=0.10cm] (0.14,0.09)--(0.86,0.09)--(0.50,0.82)--cycle;
			\draw[rounded corners=0.08cm] (0.24,0.16)--(0.76,0.16)--(0.50,0.68)--cycle;
			\draw[rounded corners=0.06cm] (0.36,0.25)--(0.64,0.25)--(0.50,0.55)--cycle;
			\draw[rounded corners=0.04cm] (0.50,0.36)--(0.50,0.36)--(0.50,0.36)--cycle;
			\end{tikzpicture}
		\caption{Contour lines $\{\weights \in \paramspace \, : \, \phi(\weights) = c\}$ for different values of 
			$c \in \mathbb{R}$ of a \gls{convex} \gls{function} $\phi(\cdot)$ defined on a
			\gls{convex} set $\paramspace$. The density of the contour lines can be used 
			to steer the \gls{learnrate} used in a \gls{gradstep}, i.e., if the 
			current \gls{modelparam} lies in a region with dense contour lines, 
			a smaller \gls{learnrate} is preferable.}
		\end{figure}
		For a twice \gls{differentiable} $\phi$, the divergence $\bregmandiv{\cdot}{\cdot}{\phi}$ 
		behaves locally like a quadratic form:
		\[
		\bregmandiv{\weights}{\weights'}{\phi}\approx\tfrac12 \big(\weights-\weights'\big)^{T} \nabla^{2}\phi(\weights')
		\big(\weights-\weights'\big),
		\]
		which can be interpreted as a squared \gls{norm} of the displacement 
		$\weights-\weights'$ induced by the \gls{hessian} $\nabla^{2}\phi(\weights')$.
		\\
		See also: \gls{gradstep}, \gls{gradient}, \gls{innerproduct}, \gls{proxop}.},
	first={Bregman divergence},
	type=math,
	text={Bregman divergence}
}

\newglossaryentry{derivative} 
{name={derivative}, 
	description={See\index{derivative} \gls{partialderivative}.}, 
 	first={derivative}, 
 	text={derivative}, 
 	type=math, 
 	plural={derivatives}, 
 	firstplural={derivatives}
}

\newglossaryentry{partialderivative}
{name={partial derivative},
	description={Consider a real-valued \gls{function} 
		$f: \mathbb{R}^{\nrfeatures} \rightarrow \mathbb{R}: \featurevec \mapsto f(\featurevec)$. 
	     	The partial \gls{derivative}\index{partial derivative} of $f(\featurevec)$ 
		with respect to the entry $\feature_{\featureidx}$ measures how  
		$f$ changes when $\feature_\featureidx$ varies while all 
        		other entries $x_{\featureidx'}$, for $\featureidx' \in \firstnatural{\nrfeatures} \setminus \{\featureidx \}$, 
		are held fixed. It is defined as \cite{RudinBookPrinciplesMatheAnalysis}
        		$$
            	\frac{\partial f(\featurevec)}{\partial \feature_\featureidx}
            	= \lim_{\varepsilon \rightarrow 0} 
              	\frac{f(\feature_1,\,\ldots,\,\feature_{\featureidx}+\varepsilon,\,\ldots,\,\feature_\nrfeatures) 
			  - f(\feature_1,\,\ldots,\,\feature_\featureidx,\,\ldots,\,\feature_\nrfeatures)}{\varepsilon}.$$
		Note that the partial \gls{derivative} is only defined if this limit exists. 
		For a \gls{differentiable} \gls{function}, the partial \glspl{derivative} of $f$ are the entries 
		of the \gls{gradient} $\nabla f$. 
		\\
        		See also: \gls{function}, \gls{derivative}, \gls{gradient}.},
    	first={partial derivative},
	firstplural={partial derivatives},
	type=math, 
	plural={partial derivatives},
    	text={partial derivative}
}

\newglossaryentry{algconn}
{name={algebraic connectivity},
	description={The\index{algebraic connectivity} algebraic connectivity of 
		an \gls{undirectedgraph} is the second-smallest \gls{eigenvalue} $\eigval{2}$ 
		of its \gls{LapMat}. 
		An \gls{undirectedgraph} is \gls{connected} if and only if $\eigval{2} >0$ 
		(see Fig.\ \ref{fig_algconn_dict}) \cite{ChungSpecGraphTheory}, \cite{Spielman2019}. 
		\begin{figure}[H]
			\centering
			\begin{tikzpicture}
			\draw[->,line width=1pt] (-1, -1) -- (9, -1) node[below] {$\eigval{2}$};
			\draw[thick] (0, -0.8) -- (0, -1.2) node[below] {{$\eigval{2}\!=\!0$}};
			\draw[thick] (7, -0.8) -- (7, -1.2) node[below] {$\eigval{2}\!=\! 3$};
			\draw[thick] (3.5, -0.8) -- (3.5, -1.2) node[below] {$\eigval{2}\!=\! 1$};
			\node[circle, fill=black, inner sep=1.5pt, label=above:{1}] (A1) at (0, 1.5) {};
			\node[above=0.5cm of A1, align=center] {\mbox{disconnected}};
			\node[circle, fill=black, inner sep=1.5pt, label=below right:{2}] (B1) [below right=0.8cm and 0.5cm of A1] {};
			\node[circle, fill=black, inner sep=1.5pt, label=below left:{3}] (C1) [below left=0.8cm and 0.5cm of A1] {};
			\draw [line width=1 pt]  (A1) -- (B1);
			\begin{scope}[xshift=3.5cm]
				\node[circle, fill=black, inner sep=1.5pt, label=above:{1}] (A2) at (0, 1.5) {};
				\node[above=0.5cm of A2, align=center] {\mbox{\gls{connected}}};
				\node[circle, fill=black, inner sep=1.5pt, label=below right:{2}] (B2) [below right=0.8cm and 0.5cm of A2] {};
				\node[circle, fill=black, inner sep=1.5pt, label=below left:{3}] (C2) [below left=0.8cm and 0.5cm of A2] {};
				\draw [line width=1 pt]  (A2) -- (B2);
				\draw [line width=1 pt]  (B2) -- (C2);
			\end{scope}
			\begin{scope}[xshift=7cm]
				\node[circle, fill=black, inner sep=1.5pt, label=above:{1}] (A3) at (0, 1.5) {};
				\node[above=0.5cm of A3, align=center] {\mbox{complete \gls{graph}}};
				\node[circle, fill=black, inner sep=1.5pt, label=below right:{2}] (B3) [below right=0.8cm and 0.5cm of A3] {};
				\node[circle, fill=black, inner sep=1.5pt, label=below left:{3}] (C3) [below left=0.8cm and 0.5cm of A3] {};
				\draw [line width=1 pt] (A3) -- (B3);
				\draw [line width=1 pt]  (B3) -- (C3);
				\draw [line width=1 pt]  (A3) -- (C3);
			\end{scope}
			\end{tikzpicture}
		\caption{Three examples of \glspl{undirectedgraph}. \label{fig_algconn_dict}}
		\end{figure}
		See also: \gls{undirectedgraph}, \gls{eigenvalue}, \gls{LapMat}, \gls{connected}.},
	first={algebraic connectivity},
	type=math,
	text={algebraic connectivity}
}

\newglossaryentry{cfwmaxmin}
{name={Courant–Fischer–Weyl min–max characterization (CFW min–max characterization)}, 
	description={Consider\index{Courant–Fischer–Weyl min–max characterization (CFW min–max characterization)} a \gls{psd} 
		\gls{matrix} $\mQ \in \mathbb{R}^{\nrfeatures \times \nrfeatures}$ with 
		\gls{evd} (or \gls{spectraldecomp}), i.e.,
		$$\mQ = \sum_{\featureidx=1}^{\nrfeatures} \eigval{\featureidx} \vu^{(\featureidx)} \big(  \vu^{(\featureidx)}  \big)\,^{T}.$$ 
		Here, we use the ordered (in ascending order) \glspl{eigenvalue} 
		\begin{equation}
			\nonumber
		 	\eigval{1}  \leq \ldots \leq \eigval{\nrnodes}. 
		\end{equation}
		The CFW min–max characterization \cite[Th. 8.1.2]{GolubVanLoanBook} 
		represents the \glspl{eigenvalue} of $\mQ$ as the solutions to certain \glspl{optproblem}.
			\\
		See also: \gls{psd}, \gls{matrix}, \gls{evd}, \gls{eigenvalue}, \gls{optproblem}.}, 
	first={Courant–Fischer–Weyl min–max characterization (CFW min–max characterization)}, 
	type=math,
	text={CFW min–max characterization}
}

\newglossaryentry{sample}
{name={sample},
	description={In the context of \gls{ml}, a\index{sample} sample is a finite \gls{sequence} 
 		(of length $\samplesize$) of \glspl{datapoint} $\datapoint^{(1)}, \,\ldots, \,\datapoint^{(\samplesize)}$. 
 		The number $\samplesize$ is called the \gls{samplesize}. 
 		\Gls{erm}-based methods use a sample to train a \gls{model} (or learn a 
 		\gls{hypothesis}) by minimizing the average \gls{loss} (i.e., the \gls{emprisk}) over that sample. 
 		Since a sample is defined as a \gls{sequence}, the same \gls{datapoint} may 
		appear more than once. By contrast, some authors in statistics define a sample 
		as a set of \glspl{datapoint}, in which case duplicates are not allowed \cite{Everitt2010}, \cite{OxfordStatisticsDictionary}.
 		These two views (i.e., \gls{sequence} versus set) can be reconciled by regarding a sample as a \gls{sequence} of 
 		\gls{feature}–\gls{label} pairs $\pair{\featurevec^{(1)}}{\truelabel^{(1)}}, \,\ldots, 
 		\,\pair{\featurevec^{(\samplesize)}}{\truelabel^{(\samplesize)}}$. The $\sampleidx$th 
 		pair consists of the \glspl{feature} $\featurevec^{(\sampleidx)}$ and the \gls{label} $\truelabel^{(\sampleidx)}$ 
 		of an unique underlying \gls{datapoint} $\widetilde{\datapoint}^{(\sampleidx)}$. While the 
 		underlying \glspl{datapoint} $\widetilde{\datapoint}^{(1)},\,\ldots,\,\widetilde{\datapoint}^{(\samplesize)}$ 
 		are unique, some of them can have identical \glspl{feature} and \glspl{label}.  
 		\begin{figure}[H]
 		\begin{center}
 			\begin{tikzpicture}[>=Latex, font=\small]
 				\node[draw, rounded corners, inner sep=6pt,
 				minimum width=5.2cm, minimum height=3.4cm] (pop) {};
 				\node[above=0pt of pop.north] {population};
 				\foreach \x/\y [count=\i] in {-2.0/0.3, -1.6/0.9, -1.2/-0.2, -0.8/0.5,
 				-0.3/-0.6, 0.2/0.1, 0.6/0.8, 1.0/-0.4,
 				1.4/0.4, 1.8/-0.1} {
 				\coordinate (p\i) at ($(pop.center)+(\x,\y)$);
 				\fill (p\i) circle (1.6pt);
 				}
 				\coordinate (sampleanchor) at ([xshift=1.8cm,yshift=0.5cm]pop.east);
 				\def\rowgap{0.55cm}
 				\def\ybase{0.95cm}
 				\node[anchor=west] (s1) at ($(sampleanchor)+(0, \ybase - 0*\rowgap)$)
 				{$1:\;(\featurevec^{(1)},\truelabel^{(1)})$};
 				\node[anchor=west] (s2) at ($(sampleanchor)+(0, \ybase - 1*\rowgap)$)
 				{$2:\;(\featurevec^{(2)},\truelabel^{(2)})$};
 				\node[anchor=west] (s3) at ($(sampleanchor)+(0, \ybase - 2*\rowgap)$)
 				{$3:\;(\featurevec^{(3)},\truelabel^{(3)})$};
 				\node[anchor=west] (s4) at ($(sampleanchor)+(0, \ybase - 3*\rowgap)$)
 				{$4:\;(\featurevec^{(4)},\truelabel^{(4)})$};
 				\node[anchor=west] (s5) at ($(sampleanchor)+(0, \ybase - 4*\rowgap)$)
 				{$5:\;(\featurevec^{(5)},\truelabel^{(5)})$};
 				\node[anchor=west] (s6) at ($(sampleanchor)+(0, \ybase - 5*\rowgap)$)
 				{$6:\;(\featurevec^{(6)},\truelabel^{(6)})$};
 				\node[draw, rounded corners, inner sep=6pt, fit=(s1)(s6)] (seqbox) {};
 				\node[above=0pt of seqbox.north]
 				{a sample};
 				\node[font=\scriptsize] at ($(p2)+(1.5mm,3.5mm)$) {$\widetilde{\datapoint}^{(1)}$};
 				\draw[->, thick] (p2) to[out=0,   in=180] ($(s1.west)+(1mm,0mm)$);
 				\draw[->, thick] (p7) to[out=10,  in=180] ($(s3.west)+(1mm,0mm)$);
				\draw[->, thick] (p4) to[out=10,  in=180] ($(s4.west)+(1mm,0mm)$);
 				\draw[->, thick] (p5) to[out=-10, in=180] ($(s5.west)+(1mm,0mm)$);
 				\draw[->, thick, densely dashed] (p3) to[out=0,  in=180] ($(s2.west)+(1mm,0mm)$);
 				\draw[->, thick, densely dashed] (p3) to[out=-5, in=180] ($(s6.west)+(1mm,0mm)$);
 			\end{tikzpicture}
 		\end{center}
 		\caption{Sample viewed as a finite \gls{sequence}. 
		        Each element of this sample consists of the \gls{featurevec} 
			and the \gls{label} of a \gls{datapoint} from an underlying population.
			The same \gls{datapoint} may occur more than once in the sample. 
 			\label{fig:sample-sequence_dict}}
 		\end{figure} 	
 		For the analysis of \gls{ml} methods, it is common to interpret (the generation of) a 
		sample as the \gls{realization} of a \gls{stochproc} indexed by $\{1,\,\ldots,\,\samplesize\}$. 
		A widely used assumption is the \gls{iidasspt}, where sample elements 
		$\pair{\featurevec^{(\sampleidx)}}{\truelabel^{(\sampleidx)}}$, 
		for $\sampleidx=1,\,\ldots,\,\samplesize$, are \gls{iid} \glspl{rv} with a common \gls{probdist}.  
		\\
 		See also: \gls{sequence}, \gls{iidasspt}, \gls{dataset}.},
 	first={sample},
	type=math,
	plural={samples},
 	text={sample}
}

\newglossaryentry{posterior}
{name={posterior (prediction)},
sort={posteriorprediction},
	description={The\index{posterior (prediction)} study and design of \gls{ml} methods is often based on a 
         	\gls{probmodel} for the \gls{data} generation process. Within a 
		\gls{probmodel}, we view (the generation of) a \gls{datapoint} with 
		\glspl{feature} $\featurevec$ and \gls{label} $\truelabel$ as an 
		\gls{rv} with \gls{probdist} $\probdist^{(\featurevec,\truelabel)}$. 
		It turns out that the optimal \gls{prediction} for the \gls{label} 
		$\truelabel$, given the \gls{featurevec} $\featurevec$, is fully 
		determined by the \gls{condprobdist} $\probdist^{(\truelabel \mid \featurevec)}$
		of $\truelabel$ given (or conditioned on) $\featurevec$ \cite{BishopBook}, \cite{papoulis}, \cite{LC}.
		\\
		See also: \gls{probmodel}, \gls{condprobdist}, \gls{posteriordist}. }, 
 	first={posterior}, 
 	type=math, 
 	text={posterior} 
}

\newglossaryentry{det}
{name={determinant},
	description={The\index{determinant} determinant $\det\,(\mA)$ of a square \gls{matrix} 
		$\mA =\linebreak \big( \va^{(1)},\,\ldots,\,\va^{(\nrfeatures)} \big) \in \mathbb{R}^{\nrfeatures \times \nrfeatures}$ is a 
		\gls{function} of its columns $\va^{(1)},\,\ldots,\,\va^{(\nrfeatures)} \in \mathbb{R}^{\nrfeatures}$, i.e., it satisfies 
		the following properties \cite{DirschmidHansJorg1996TuF}:
		\begin{itemize}
			\item Normalized: $$\det\,(\mI) = 1.$$ 
			\item Multilinear: \begin{align} \nonumber \det \big(\va^{(1)},\,\ldots,\,\alpha\vu+ \beta \vv,\,\ldots,\,\va^{(\nrfeatures)} \big) & = \alpha\det \big(\va^{(1)},\,\ldots,\,\vu,\,\ldots,\,\va^{(\nrfeatures)} \big) \\ 
			& +\beta\det \big(\va^{(1)},\,\ldots,\,\vv,\,\ldots,\,\va^{(\nrfeatures)} \big). \nonumber
			\end{align}
			\item Antisymmetric: $$\det \big(\ldots,\,\va^{(\featureidx)}, \,\ldots, \,\va^{(\featureidx')},\,\ldots \big) = - \det \big(\ldots,\,\va^{(\featureidx')}, \,\ldots, \,\va^{(\featureidx)},\,\ldots \big).$$ 
		\end{itemize} 
		We can interpret a \gls{matrix} $\mA$ as a linear transformation on $\mathbb{R}^{\nrfeatures}$.
		The determinant $\det\,(\mA)$ characterizes how volumes in $\mathbb{R}^\nrfeatures$ (and their orientation) 
		are altered by this transformation (see Fig.\ \ref{fig_det_dict}) \cite{GolubVanLoanBook}, \cite{Strang2007}. 
 		In particular, $\det\,(\mA) > 0$ preserves orientation, $\det\,(\mA) < 0$ reverses orientation, 
 		and $\det\,(\mA) = 0$ collapses volume entirely, indicating that $\mA$ is non-\gls{invertible}. 
 		The determinant also satisfies $\det\,(\mA \mB) = \det\,(\mA) \cdot \det\,(\mB)$, and if $\mA$ is 
 		\gls{diagonalizable} with \glspl{eigenvalue} $\eigval{1}, \,\ldots, \,\eigval{\nrfeatures}$, then 
		$\det\,(\mA) = \prod_{\featureidx=1}^{\nrfeatures} \eigval{\featureidx}$ \cite{HornMatAnalysis}.
    		For the special cases $\nrfeatures=2$ (i.e., in two-dimensional (2-D) space $\mathbb{R}^{2}$) and 
		$\nrfeatures=3$ (i.e., in three-dimensional (3-D) space $\mathbb{R}^{3}$), 
		the determinant can be interpreted as an oriented area or volume spanned by the column \glspl{vector} of $\mA$.
    		\begin{figure}[H]
    			\begin{center}
    			\begin{tikzpicture}[x=2cm]
			\begin{scope}
			\draw[->, thick] (0,0) -- (1,0) node[below right] {$\vx$};
			\draw[->, thick] (0,0) -- (0,1) node[above left] {$\vy$};
			\end{scope}
			\begin{scope}[shift={(2.8,0)}]
			\coordinate (A) at (1.5,0.5);
			\coordinate (B) at (-0.2,1.2);
			\draw[->, very thick, red] (0,0) -- (A) node[below right] {$\mA \vx$};
			\draw[->, very thick, red] (0,0) -- (B) node[above left] {$\mA \vy$};
			\draw[fill=red!20, opacity=0.6] (0,0) -- (A) -- ($(A)+(B)$) -- (B) -- cycle;
			\draw[dashed] (A) -- ($(A)+(B)$);
			\draw[dashed] (B) -- ($(A)+(B)$);
			\node at (0.8,0.6) {\small $\det\,(\mA)$};
			\draw[->, thick, blue] (0.4,0.0) arc[start angle=0, end angle=35, radius=0.6];
			\end{scope}
			\draw[->, thick] (1.3,0.5) -- (2.4,0.5) node[midway, above] {$\mA$};
			\end{tikzpicture}
			\end{center}
		\caption{We can interpret a square \gls{matrix} $\mA$ as a linear transformation of $\mathbb{R}^{\nrfeatures}$ 
			into itself. The determinant $\det\,(\mA)$ characterizes how this transformation alters an oriented volume.
			\label{fig_det_dict}}
		\end{figure}
		See also: \gls{eigenvalue}, \gls{inverse}.},
	first={determinant},
	type=math,
	text={determinant}
}

\newglossaryentry{hessian}
{name={Hessian},
 	description={Consider\index{Hessian} a \gls{function} 
		$f: \mathbb{R}^{\nrfeatures} \to \mathbb{R}$ for which the second-order 
		\glspl{partialderivative} exist at $\featurevec'$. Then, the Hessian $\nabla^2 f(\featurevec')$ 
		of $f$ at $\featurevec$ is defined as the \gls{matrix} of second-order 
		\glspl{partialderivative} of $f$ at $\featurevec'$, i.e., 
		$$\nabla^{2} f(\featurevec') \;=\;
		\begin{pmatrix}
			\frac{\partial^{2} f}{\partial \feature_{1}^{2}}
			& \frac{\partial^{2} f}{\partial \feature_{1}\, \partial \feature_{2}}
			& \cdots
			& \frac{\partial^{2} f}{\partial \feature_{1}\, \partial \feature_{\nrfeatures}} \\
			\frac{\partial^{2} f}{\partial \feature_{2}\, \partial \feature_{1}}
			& \frac{\partial^{2} f}{\partial \feature_{2}^{2}}
			& \cdots
			& \frac{\partial^{2} f}{\partial \feature_{2}\, \partial \feature_{\nrfeatures}} \\
			\vdots & \vdots & \ddots & \vdots \\[1.2ex]
			\frac{\partial^{2} f}{\partial \feature_{\nrfeatures}\, \partial \feature_{1}}
			& \frac{\partial^{2} f}{\partial \feature_{\nrfeatures}\, \partial \feature_{2}}
			& \cdots
			& \frac{\partial^{2} f}{\partial \feature_{\nrfeatures}^{2}}
		\end{pmatrix}.$$
		If the second-order \glspl{partialderivative} are \gls{continuous} in a \gls{neighborhood} around $\featurevec'$, 
		then the Hessian is a \gls{symmetricmatrix}, i.e.,
		${\partial^{2} f}/{\partial \feature_{\featureidx}\, \partial \feature_{\featureidx'}} = 
		{\partial^{2} f}/{\partial \feature_{\featureidx'}\, \partial \feature_{\featureidx}}$ for all 
		$\featureidx, \featureidx'$ \cite{RudinBookPrinciplesMatheAnalysis}. If additionally $f$ 
		is \gls{convex}, then the Hessian is a \gls{psd} \gls{matrix} \cite{BoydConvexBook}.\\
		The Hessian $\nabla^2 f(\featurevec')$ can be used to compute a \gls{quadfunc} 
		$$q(\featurevec) = \frac{1}{2} (\featurevec- \featurevec')^{T} \underbrace{\nabla^2 f(\featurevec')}_{\text{Hessian}} 
		(\featurevec- \featurevec') +  (\featurevec- \featurevec')^{T} \underbrace{\nabla f(\featurevec')}_{\text{\gls{gradient}}} 
		+ f(\featurevec')$$
		that approximates $f$ locally around $\featurevec'$ (see also Fig.\ \ref{fig_quadapprox_hessian_dict}). 
		\begin{figure}[H]
		\begin{center}
		\begin{tikzpicture}[x=0.5cm]
			\begin{axis}[
				hide axis,
				xmin=3, xmax=6,
				ymin=0, ymax=6,
				domain=0:6,
				samples=100,
				width=10cm,
				height=6cm,
				clip=false
				]
				\addplot[blue, thick, domain=3:6.6] {2 + sin(deg(x))} 
				node[pos=0.5, above right, yshift=3pt] {$f(\featurevec)$};
				\addplot[red, thick, domain=3.5:6.6] 
				{2 + sin(deg(6)) + cos(deg(6))*(x - 6)}
				node[pos=0, below left] {$g(\featurevec)$};
                			\addplot[green, thick, domain=3.5:6.6]
                			{2 + sin(deg(6)) + cos(deg(6))*(x - 6) - 0.5*sin(deg(6))*(x - 6)^2}
				node[pos=0, below left, yshift=10pt] {$q(\featurevec)$};
				\addplot[mark=*] coordinates {(6, {2 + sin(deg(6))})};
				\addplot[dashed, gray] coordinates {(6,0) (6,2.4)};
				\node at (axis cs:6, -0.2) {$\featurevec'$};
				\pgfmathsetmacro{\xA}{-1.5}
				\pgfmathsetmacro{\xB}{3}
				\pgfmathsetmacro{\yA}{2 + sin(deg(\xA))}
				\pgfmathsetmacro{\yB}{2 + sin(deg(\xB))}
			\end{axis}
			\vspace*{-10mm}
		\end{tikzpicture}
		\vspace*{-5mm}
		\end{center}
		\caption{\Gls{function} $f(\featurevec)$, sufficiently \gls{smooth} at a point $\featurevec'$, locally approximated by 
			a \gls{quadfunc} $q(\featurevec)$, which provides a more accurate approximation 
			compared to a linear \gls{function} $g(\featurevec)$. \label{fig_quadapprox_hessian_dict}}
		\end{figure}
		See also: \gls{function}, \gls{matrix}, \gls{quadfunc}, \gls{differentiable}. }, 
	first={Hessian},
	type=math,
	text={Hessian}
}

\newglossaryentry{denautoencoder}
{name={denoising autoencoder},
	description={A denoising\index{denoising autoencoder} \gls{autoencoder} extends the basic 
		\gls{autoencoder} by perturbing (e.g., by intentionally adding noise to) 
		the \gls{inputvec} (or \gls{featurevec}) before feeding it 
		into the \gls{autoencoder} itself during the \gls{training} \cite[Sec. 14.2.2]{Goodfellow-et-al-2016}.
		Once trained, we can use a denoising \gls{autoencoder} to 
		denoise a corrupted representation of a \gls{datapoint} (see Fig.\ \ref{fig:clean-noisy_dict}).   
		\begin{figure}[H]
			\begin{minipage}{0.45\textwidth}
  				\centering
				\includegraphics[width=\textwidth]{assets/imagedenautoenc_noisy.jpg}
				{\selectfont (a)}
  			\end{minipage}
			\hfill
			\begin{minipage}{0.45\textwidth}
  				\centering
				\includegraphics[width=\textwidth]{assets/imagedenautoenc.jpg}
				{\selectfont (b)}
			\end{minipage}
  		\caption{Denoising \gls{autoencoder} reading in (a) a 
			corrupted (i.e., noisy) representation of a \gls{datapoint} 
			and computing a \gls{prediction} for (b) the clean representation.}
  			\label{fig:clean-noisy_dict}
		\end{figure}
		See also: \gls{autoencoder}, \gls{diffusionmethod}.  }, 
    	first={denoising autoencoder},
        	plural={denoising autoencoders},
    	firstplural={denoising autoencoders},
	type=math,
    	text={denoising autoencoder}
}

\newglossaryentry{condnr}
{name={condition number},
sort={conditionnumber},
	description={The condition number\index{condition number} 
		$\kappa(\mathbf{Q}) \geq 1$ of a positive definite 
		\gls{matrix} $\mathbf{Q} \in \mathbb{R}^{\featurelen \times \featurelen}$ is the ratio 
		$\alpha /\beta$, where $\alpha$ and $\beta$ denote the largest and smallest
		\glspl{eigenvalue} of $\mathbf{Q}$, respectively. The condition number 
		is useful for the analysis of \gls{ml} methods. 
		The computational complexity of \glspl{gdmethod} for \gls{linreg}  
		depends crucially on the condition number of the \gls{matrix} 
		$\mQ = \featuremtx \featuremtx\,^{T}$, where $\featuremtx$ is the 
		\gls{featuremtx} of the \gls{trainset}. These methods converge faster 
		when the condition number $\kappa(\mQ)$ is close to $1$.
					\\ 
		See also: \gls{matrix}, \gls{eigenvalue}, \gls{gdmethod}.},
	first={condition number},
	plural={condition numbers},
	firstplural={condition numbers},
	type=math, 
	text={condition number} 
}

\newglossaryentry{grangercausality}
{name={Granger causality},
	description={See\index{Granger causality} \gls{causality}.
		\\
		See also: \gls{scm}, \gls{stochproc}.},
	first={Granger causality},
	type=math,
	text={Granger causality}
}

\newglossaryentry{causality}
{name={causality},
	description={In the context of \gls{ml}, causality\index{causality} refers to 
		cause–effect relationships between quantities, such as the 
		\glspl{feature} and \glspl{label} of \glspl{datapoint} 
		\cite{Pearl:2010aa}, \cite{Thurman1988}. Different notions of causality 
		have been developed in the literature. One widely used notion 
		is based on a \gls{scm}, which represents causal mechanisms via 
		functional relationships \cite{Pearl:2010aa}. 
		For time-series \gls{data}, Granger causality defines a predictive 
		notion of causality \cite{Deistler2022}, \cite{Granger1969}. A 
		\gls{stochproc} $X_{1},\,X_{2},\,\ldots$ is said to cause another 
		\gls{stochproc} $Y_{1},\,Y_{2},\,\ldots$ if using past values 
		$X_{\timeidx'}$, $\timeidx' < \timeidx$, improves the \gls{prediction} 
		of $Y_{\timeidx}$ beyond using the past of $Y$ alone. 
		\\ 
		See also: \gls{scm}, \gls{stochproc}.},
	first={causality},
	type=math,
	text={causality}
}

\newglossaryentry{scm}
{name={structural causal model (SCM)},
	description={An SCM\index{structural causal model (SCM)} 
		is a mathematical \gls{model} of cause–effect relationships underlying 
		a \gls{data} generation process \cite{Pearl:2010aa}. 
		\\
		See also: \gls{causality}.},
	plural={SCMs},
	firstplural={structural causal models (SCMs)},
	first={structural causal model (SCM)},
	type=math,
	text={SCM}
}

\newglossaryentry{averagenodedegree}
{name={average node degree},
	description={The average \gls{nodedegree}\index{average node degree} $\avgnodedegree$ of a weighted 
		\gls{undirectedgraph} $\graph=(\nodes,\edges,\mA)$ is the average of 
		all \glspl{nodedegree}, i.e., $\avgnodedegree = (1/\nrnodes) \sum_{\nodeidx \in \nodes} \nodedegree{\nodeidx}$.
					\\ 
		See also: \gls{nodedegree}, \gls{undirectedgraph}, \gls{neighbor}.},
	first={average node degree},
	type=math, 
	text={average node degree} 
}

\newglossaryentry{fixedpointcharact}
{name={fixed-point characterization},
	description={The solution of \glspl{optproblem} arising in \gls{ml} 
		can often be characterized by a \gls{fixedpointeq}\index{fixed-point characterization}. 
		As a case in point, consider an \gls{optproblem} with a \gls{smooth} 
		and \gls{convex} \gls{objfunc} $f(\weights)$. Each minimizer $\widehat{\weights}$ 
		of $f(\weights)$ satisfies the \gls{fixedpointeq} 
		$\widehat{\weights} = \fixedpointop^{(\lrate)} \widehat{\weights}$ with 
		$\fixedpointop^{(\lrate)}: \weights \mapsto \weights-\lrate \nabla f(\widehat{\weights})$ \cite{BoydConvexBook}. 
		Here, $\lrate > 0$ is an arbitrary positive constant and 
		$\nabla f(\weights)$ denotes the \gls{gradient} of $f(\weights)$.	
		Once we have found a \gls{fixedpointeq} with an \gls{operator} 
		$\fixedpointop$ that is a \gls{contractop}, we can use a \gls{fixedpointiter} 
		to compute a solution of the underlying \gls{optproblem}. In the 
		above example, the resulting \gls{fixedpointiter} coincides with \gls{gd} for 
		minimizing $f(\weights)$ \cite{BoydConvexBook}.
	               \\ 
		See also: \gls{fixedpointeq}, \gls{fixedpointiter}.},
	first={fixed-point characterization},
	type=math, 
	text={fixed-point characterization} 
}

\newglossaryentry{nodedegree}
{name={node degree},
	description={The degree\index{node degree} $\nodedegree{\nodeidx}$ of a 
		node $\nodeidx \in \nodes$ in an \gls{undirectedgraph} is the 
		number of its \glspl{neighbor}, i.e., 
		$\nodedegree{\nodeidx} \defeq \big|\neighbourhood{\nodeidx}\big|$.
		For a weighted \gls{undirectedgraph} $\graph=(\nodes,\edges,\mA)$, 
		we can alternatively define the (weighted) node degree as the sum 
		of the weights of all edges \gls{connected} to node $\nodeidx$, i.e.,
		$\nodedegree{\nodeidx} = \sum_{\nodeidx' \in \nodes} \edgeweight_{\nodeidx,\nodeidx'}$ \cite{OPSAHL2010245}.	
					\\ 
		See also: \gls{undirectedgraph}, \gls{neighbor}, \gls{averagenodedegree}.},
	first={node degree},
	type=math, 
	text={node degree} 
}

\newglossaryentry{empiricaldistribution}
{name={empirical distribution},
	description={Consider a \gls{dataset} 
 		$\dataset = \{ \featurevec^{(1)},\,\ldots, \,\featurevec^{(\samplesize)}\}$ 
  		consisting of $\samplesize$ distinct \glspl{datapoint}, each characterized by 
  		the \gls{featurevec} $\featurevec^{(\sampleidx)} \in \featurespace$ for $\sampleidx = 1, \,\ldots, \,\samplesize$. 
  		For a given \gls{sigmaalgebra} $\sigmaalgebra$ over the \gls{featurespace} 
  		$\featurespace$, the empirical distribution\index{empirical distribution} 
  		of $\dataset$ is the \gls{probdist} $\probdist^{(\dataset)}$ defined via 
   		$$\probdist^{(\dataset)}(\mathcal{A}) = (1/\samplesize) \big| \big\{ \sampleidx: \featurevec^{(\sampleidx)} \in \mathcal{A} \big\} \big|
		\quad \text{for any \gls{event} } \mathcal{A} \in \sigmaalgebra.
		$$
		In other words, the empirical distribution assigns to any \gls{measurable} set 
		$\mathcal{A} \in \sigmaalgebra$ the fraction of \glspl{datapoint} in $\dataset$ 
		that fall into $\mathcal{A}$. If the \gls{featurespace} is ordered, the empirical 
		distribution can also be characterized by its empirical \gls{cdf}:
		$$\cdf{\dataset}{\featurevec} = 
		(1/\samplesize) \big| \big\{ \sampleidx: \featurevec^{(\sampleidx)} \preceq \featurevec \big\} \big| 
		\quad \text{for any } \featurevec \in \featurespace,$$
		where $\preceq$ denotes the ordering relation on $\featurespace$.
		\begin{figure}[H]
		\begin{center}
		\begin{tikzpicture}[>=stealth, thick,y=2cm]
			\foreach \x/\p in {1/0.3, 4/0.7}
			{
			\draw[gray] (\x,0) -- (\x,\p);
			\fill[blue] (\x,\p) circle (2pt);
			}
			\node[anchor=south,align=center] at (1,0.3) {\small $\pmf{\dataset}{\star}=\frac{3}{10}$};
			\node[anchor=north] at (1,0) {\small $\star$};
			\node[anchor=north] at (4,0) {\small $\otimes$};
			\node[anchor=west,text width=11cm] at (-1.2,-0.80) {
			$\dataset = (\star,\,\star,\,\star,\,\otimes,\,\otimes,\,\otimes,\,\otimes,\,\otimes,\,\otimes,\,\otimes)$
			};
		\end{tikzpicture}
    		\end{center}
		\caption{\Gls{dataset} $\dataset$ consisting of $\samplesize=10$ \glspl{datapoint}, each 
			characterized by a value from the finite \gls{featurespace} $\featurespace = \{\star, \otimes\}$. 
			The empirical \gls{pmf} $\pmf{\dataset}{\featurevec}$ assigns to each 
			possible value $\featurevec \in \featurespace$ the fraction of \glspl{datapoint} 
			in $\dataset$ whose \gls{feature} takes on this value. Here, three out of 
        			ten \glspl{datapoint} take on the \gls{feature} value $\star$, resulting 
			in $\pmf{\dataset}{\star} = 3/10$. \label{fig_empirical_pmf_dict}}
		\end{figure}
		If the \gls{featurespace} $\featurespace$ is finite, 
		the empirical distribution of $\dataset$ can also be characterized by 
		the empirical \gls{pmf}:
		$$\pmf{\dataset}{\featurevec} = 
		(1/\samplesize) \big| \big\{ \sampleidx: \featurevec^{(\sampleidx)} = \featurevec \big\} \big| 
		\quad \text{for any } \featurevec \in \featurespace.$$
		\\
        		See also: \gls{sigmaalgebra}, \gls{probdist}.},
     	first={empirical distribution},
     	plural={empirical distributions},
    	text={empirical distribution},
	type=math
}

\newglossaryentry{dualnorm}
{name={dual norm},
sort={dualnorm},
	description={Every \gls{norm} $\normgeneric{\cdot}{}$ defined on a \gls{euclidspace} $\mathbb{R}^{\dimlocalmodel}$ 
		has an associated dual \gls{norm}\index{dual norm}, which is denoted by $\normgeneric{\cdot}{*}$ and defined as 
		$\normgeneric{\vy}{*} \defeq \sup_{\norm{\vx}{} \le 1} \vy\,^{T} \vx$. 
		The dual \gls{norm} measures the largest possible \gls{innerproduct} between $\vy$ 
		and any \gls{vector} in the unit ball of the original \gls{norm}. For further details, see 
		\cite[Sec.~A.1.6]{BoydConvexBook}.
					\\ 
		See also: \gls{norm}, \gls{euclidspace}, \gls{vector}.},
	first={dual norm},
	type=math,
	text={dual norm}
}

\newglossaryentry{geometricmedian}
{name={geometric median (GM)},
	description={The GM\index{geometric median (GM)} of a set of \glspl{inputvec} $\vx^{(1)}, \,\ldots, \,\vx^{(\samplesize)}$ 
		in $\mathbb{R}^{\dimlocalmodel}$ is a point $\vz \in \mathbb{R}^{\dimlocalmodel}$ that 
		minimizes the sum of distances to the \glspl{vector} \cite{BoydConvexBook} such that 
		\begin{equation} 
			\label{equ_geometric_median_dict}
			\vz \in \argmin_{\vy \in \mathbb{R}^{\dimlocalmodel}} \sum_{\sampleidx=1}^{\samplesize} \normgeneric{\vy - \vx^{(\sampleidx)}}{2}.
		\end{equation} 
		Fig.\ \ref{opt_cond_GM_dict} illustrates a fundamental property of the GM:
		If $\vz$ does not coincide with any of the \glspl{inputvec}, then the unit \glspl{vector} pointing 
		from $\vz$ to each $\vx^{(\sampleidx)}$ must sum to zero—this is the zero-\gls{subgradient}  
		(optimality) condition for \eqref{equ_geometric_median_dict}. It turns out that the solution to 
		\eqref{equ_geometric_median_dict} cannot be arbitrarily pulled away from trustworthy  
		\glspl{inputvec} as long as they are the majority \cite[Th. 2.2]{Lopuhaae1991}.
	  	\begin{figure}[H]
  		\begin{center}
			\begin{tikzpicture}[scale=2, thick, >=stealth]
				\coordinate (w) at (3,0);
				\fill (w) circle (1.2pt) node[below right] {$\vz$};
				\coordinate (w2) at (0.5,0.3);
				\coordinate (w3) at (0.7,0.7);
				\fill (w2) circle (1pt) node[above left] {$\vx^{(1)}$};
				\fill (w3) circle (1pt) node[above left] {$\vx^{(2)}$};
			    	\node[anchor=west] at ($(w2) +(-0.2,0.9)$) {\textbf{clean}};
				\draw[dashed] (w) -- (w2);
				\draw[dashed] (w) -- (w3);
				\draw[->, thick, red] (w) -- ($(w)!1cm!(w2)$) ;
				\draw[->, thick, red] (w) -- ($(w)!1cm!(w3)$) node[pos=0.9, right,yshift=7pt] {$\frac{\vx^{(2)}- \vz}{\normgeneric{\vx^{(2)}-\vz}{2}}$};
				\coordinate (w4) at (5,0.2);
				\node at (5,0.6) {\textbf{perturbed}};
				\fill (w4) circle (1pt) node[below left] {$\vx^{(3)}$};
				\draw[->, thick, red] (w) -- ($(w)!1cm!(w4)$) ;
			\end{tikzpicture}
		\caption{\label{opt_cond_GM_dict}
			Consider a solution $\vz$ of \eqref{equ_geometric_median_dict} that does not coincide 
			with any of the \glspl{inputvec}. The optimality condition for \eqref{equ_geometric_median_dict} 
			requires that the unit \glspl{vector} from $\vz$ to the \glspl{inputvec} sum to zero.}
			\end{center}
		\end{figure}
		See also: \gls{vector}, \gls{subgradient}.},
	first={geometric median (GM)},
	type=math,
	text={GM}
}

\newglossaryentry{norm}
{name={norm},
	description={A norm\index{norm} is a \gls{function} that maps each (\gls{vector}) element 
		of a \gls{vectorspace} to a nonnegative real number. Formally, a norm 
		$\normgeneric{\, \cdot \,}{}$ on a \gls{vectorspace} $\mathcal{X}$ over a \gls{field} $\mathbb{F}$ is a 
		\gls{function} $\normgeneric{\, \cdot \,}{}: \mathcal{X} \to \mathbb{R}_{+}$ 
		that satisfies the following conditions \cite{HornMatAnalysis} for all $\featurevec, \featurevec' \in \mathcal{X}$ 
		and $\alpha \in \mathbb{F}$:
		\begin{itemize}
			\item Definiteness: $\normgeneric{\featurevec}{} = 0 \iff \featurevec = \mathbf{0}$;
			\item Homogeneity: $\normgeneric{\alpha \featurevec}{} = |\alpha| \cdot \normgeneric{\featurevec}{}$;
			\item Triangle inequality: $\normgeneric{\featurevec + \featurevec'}{} \leq \normgeneric{\featurevec}{} + \normgeneric{\featurevec'}{}$.
		\end{itemize}
		Intuitively, norms measure the length or size of a \gls{vector}.
		Norms can also be used to measure distances between \glspl{vector}, as they define a \gls{metric} by 
		$\metric{\featurevec}{\featurevec'} = \norm generic{\featurevec - \featurevec'}{}$,
		making $(\mathcal{X}, \metric{\cdot}{\cdot})$ a \gls{metricspace}.
		A prominent family of norms used in \gls{ml} is the $\ell_p$-norms for $p \geq 1$, 
		defined for a \gls{vector} $\featurevec \in \mathbb{R}^\featuredim$ as
		$\normgeneric{\featurevec}{p} = \left( \displaystyle \sum_{i=1}^\featuredim |\feature_i|^p \right)^{1/p}$.
		Important instances include the $\ell_1$-norm, the $\ell_2$-norm (or \gls{euclidnorm}), 
		and the $\ell_\infty$-norm $\normgeneric{\featurevec}{\infty} = \max_i |\feature_i|$, which are illustrated
		by the geometry of their respective unit spheres in Fig.\ \ref{fig:norms_unit_balls_dict}.
		In \gls{ml}, norms are fundamental, for instance for defining a \gls{lossfunc} or a \gls{regularizer}. 
		\begin{figure}[H]
		\centering
		\begin{tikzpicture}[scale=1.5]
			\draw[->] (-1.2,0) -- (1.2,0) node[right] {$x_1$};
			\draw[->] (0,-1.2) -- (0,1.2) node[above] {$x_2$};
			\draw[blue, thick] (-1,-1) rectangle (1,1);
			\node[blue] at (0.8, 1.1) {\tiny $\ell_\infty$};
			\draw[red, thick] (0,0) circle (1);
			\node[red] at (0.7, 0.8) {\tiny $\ell_2$};      
			\draw[green!60!black, thick] (1,0) -- (0,1) -- (-1,0) -- (0,-1) -- cycle;
			\node[green!60!black] at (0.4, 0.4) {\tiny $\ell_1$};
		\end{tikzpicture}
		\caption{Unit spheres $\sphere{1}$ with respect to different $\ell_p$-norms for $p=1,\,2,\,\infty$.}
			\label{fig:norms_unit_balls_dict}
		\end{figure}
		See also: \gls{euclidspace}, \gls{hilbertspace},  \gls{innerproduct}, \gls{linreg}, \gls{lasso}.},
	first={norm},
	type=math,
	text={norm},
        plural={norms}
}

\newglossaryentry{maximum}
{name={maximum},
	description={The maximum\index{maximum} of a set $\mathcal{A} \subseteq \mathbb{R}$ 
     		of real numbers is the greatest element in that set if such an element exists. A set $\mathcal{A}$ 
     		has a maximum if it is bounded above and attains its \gls{supremum} \cite[Sec.~1.4]{RudinBookPrinciplesMatheAnalysis}.
				\\ 
		See also: \gls{supremum}.},
 	first={maximum},
 	firstplural={maxima},
	type=math,
 	plural={maxima},
 	text={maximum}
}

\newglossaryentry{supremum}
{name={supremum (or least upper bound)},
	description={The supremum\index{supremum (or least upper bound)} of a set of real numbers is 
		the smallest number that is greater than or equal to every element in the set. More formally, a 
		real number $a$ is the supremum of a set $\mathcal{A} \subseteq \mathbb{R}$ if: 1) $a$ 
		is an upper bound of $\mathcal{A}$; and 2) no number smaller than $a$ is an upper bound of $\mathcal{A}$. 
		Every non-empty set of real numbers that is bounded above has a supremum, even if it does 
		not contain its supremum as an element \cite[Sec.~1.4]{RudinBookPrinciplesMatheAnalysis}.
		\\
		See also: \gls{infimum}. },
	firstplural={suprema}, 
  	plural={suprema},
	type=math,
	first={supremum (or least upper bound)},
	text={supremum}
}

\newglossaryentry{infimum}
{name={infimum (or greatest lower bound)},
	description={The infimum\index{infimum (or greatest lower bound)} of a set of 
		real numbers is the largest number that is less than or equal 
		to every element in the set. More formally, a real number $a$ is 
		the infimum of a set $\mathcal{A} \subseteq \mathbb{R}$ if: 1) $a$ 
		is a lower bound of $\mathcal{A}$; and 2) no number larger than 
		$a$ is a lower bound of $\mathcal{A}$. Every non-empty set of real 
		numbers that is bounded below has an infimum, even if it does 
		not contain its infimum as an element \cite[Sec.~1.4]{RudinBookPrinciplesMatheAnalysis}.
		\\
		See also: \gls{supremum}. },
	firstplural={infima}, 
  	plural={infima},
	type=math, 
	first={infimum (or greatest lower bound)},
	text={infimum}
}

\newglossaryentry{group}
{name={group},
	description={A\index{group} group $(G, \diamond)$ is an algebraic structure \cite[Ch.~1]{artin2016geometric} 
		consisting of a set $G$ and a binary operation $\diamond$ that assigns to any two elements $a, b \in G$ 
		another element $a \diamond b \in \mathcal{G}$. For $G$ to be a group, it must satisfy associativity 
		(i.e., $(a \diamond b) \diamond c = a \diamond (b \diamond c)$ for all $a,b,c \in G$), 
		the existence of an identity element $e \in G$ with $a \diamond e = e \diamond a = a$ for all $a \in G$, 
		and the existence of an inverse $a^{-1} \in G$ for every element $a \in G$ that satisfies $a^{-1} a = a a^{-1} = e$. 
		A group is called abelian or commutative if $a \diamond b = b \diamond a$ for all $a, b \in G$. 
		Basic examples are the additive group $(\mathbb{R}, +)$ and the multiplicative group 
		$(\mathbb{R}\setminus \{0\}, \cdot)$; similarly, $(\mathbb{C}, +)$ and $(\mathbb{C}\setminus \{0\}, \cdot)$ 
		are defined using addition and multiplication on the real (complex) numbers in the usual sense. 
		More complicated group structures exist in \glspl{matrix} \cite{artin2016geometric}, such as rotation \glspl{matrix} 
		useful in \gls{ml} for \gls{dataaug}.
		\\
		See also: \gls{vectorspace}, \gls{subspace}, \gls{inverse}, \gls{field}.},
	first={group},
	plural={groups},
	type=math,
	text={group}
}

\newglossaryentry{field}
{name={field},
	description={A\index{field} field $(\mathbb{F}, +, \cdot)$ is an algebraic structure generalizing the real and complex numbers. 
		Formally, a field consists of a set $\mathbb{F}$ equipped with two binary operations, an addition $\oplus$ 
		and a multiplication $\odot$, where $(\mathbb{F}, \oplus)$ is an abelian \gls{group} with an identity element $0$ 
		and $(\mathbb{F} \setminus \{0\}, \odot)$ is an abelian \gls{group} with an identity element $1$. 
		Furthermore, it satisfies distributivity, i.e.,  $a \odot (b \oplus c) = (a \odot b) \oplus (a \odot c)$ for all $a, b, c \in \mathbb{F}$. 
		The most common examples are $(\mathbb{R}, +, \cdot)$ and $(\mathbb{C}, +, \cdot)$, with addition and multiplication 
		in the usual sense. Fields provide the set of scalars required to define a \gls{vectorspace}. 
		\\
		See also: \gls{group}, \gls{vectorspace}, \gls{subspace}, \gls{euclidspace}, \gls{innerproduct}.},
	first={field},
	plural={fields},
	type=math,
	text={field}
}

\newglossaryentry{linearlydep}
{name={linearly dependent},
	description={A subset $\{\va^{(1)}, \,\ldots, \,\va^{(\nrfeatures)}\} \in \mathcal{V}$ 
		of a \gls{vectorspace} $\mathcal{V}$ over a \gls{field} $\mathbb{F}$ is linearly 
		dependent\index{linearly dependent} if it is not \gls{linearlyindep}. 
		This occurs if there exists a set of scalars $\alpha_1, \,\ldots, \,\alpha_{\nrfeatures} \in \mathbb{F}$, 
		not all equal to zero, such that 
		$$\sum_{\featureidx=1}^{\nrfeatures} \alpha_{\featureidx} \va^{(\featureidx)} = \mathbf{0}.$$ 
		Intuitively, this means that at least one \gls{vector} in the set can be expressed as a 
		(nontrivial) linear combination of the others. 
		\\ 
		See also: \gls{vectorspace}, \gls{linearlyindep}, \gls{vector}.},
	text={linearly dependent},
	type=math,
	first={linearly dependent}
}

\newglossaryentry{cauchyschwarzinequ}
{name={Cauchy-Schwarz inequality},
sort={CauchySchwarz},
	description={For any \glspl{vector} $\featurevec, \featurevec'$ in an \gls{innerproduct} space (i.e., a \gls{hilbertspace}) 
		over the \gls{field} of real or complex numbers, the Cauchy-Schwarz inequality\index{Cauchy-Schwarz inequality} 
		provides an upper bound on the (squared) absolute value 
		(on $\mathbb{R}$ or $\mathbb{C}$) of their \gls{innerproduct}. More precisely, it states that
		\[
			|\innerprod{\featurevec}{\featurevec'}|^2 \leq \innerprod{\featurevec}{\featurevec} \innerprod{\featurevec'}{\featurevec'}.
		\]
		  Using the \gls{norm} defined by the \gls{innerproduct}, this can be expressed equivalently as
		$|\innerprod{\featurevec}{\featurevec'}| \leq \norm{\featurevec} \norm{\featurevec'}$.
		Equality holds if and only if $\{\featurevec, \featurevec'\}$ is \gls{linearlydep}. 
		\\
		See also: \gls{linearlyindep}, \gls{firmlynonexpansiveop}, \gls{euclidspace}, \gls{orthogonalitycondition}.},
	text={Cauchy-Schwarz inequality},
	type=math,
	first={Cauchy-Schwarz inequality}
}

\newglossaryentry{normedspace}
{name={normed space},
	description={A normed space\index{normed space} is a \gls{vectorspace} $\mathcal{V}$ equipped with a \gls{norm} 
		$\normgeneric{\cdot}{}$. Formally, it is denoted as the pair $(\mathcal{V}, \normgeneric{\cdot}{})$. 
		Every normed space is a \gls{metricspace} with the induced \gls{metric} defined by 
		$\metric{\featurevec}{\featurevec'} = \normgeneric{\featurevec - \featurevec'}{}$ 
		for all $\featurevec, \featurevec' \in \mathcal{X}$. 
		In \gls{ml}, normed spaces are a common framework for \glspl{featurespace} and \glspl{paramspace}, 
		most typically the \glspl{euclidspace} $\mathbb{R}^\featuredim$ with the \gls{euclidnorm}.  
		\\
		See also: \gls{vectorspace}, \gls{norm}, \gls{metricspace}, \gls{euclidspace}, \gls{hilbertspace}.},
	text={normed space},
	type=math,
	first={normed space},
	firstplural={normed spaces},
	plural={normed spaces}
}

\newglossaryentry{priordist}
{name={prior distribution},
	description={In the probabilistic\index{prior distribution} analysis of an \gls{mlsystem}, one typically distinguishes 
		between observed \glspl{datapoint} and unknown \glspl{parameter} of a \gls{probmodel}. 
		A prior distribution is a postulated \gls{probdist} over the possible choices of 
		these \glspl{modelparam} before (i.e., prior to) observing any \gls{data} \cite{BishopBook}. 
		Given a \gls{condprobdist} of the \gls{data} given the \glspl{modelparam}, 
		we update the prior to obtain the \gls{posteriordist} of the \glspl{modelparam}.
		\\
		See also: \gls{probdist}, \gls{modelparam}, \gls{posteriordist}. },
	first={prior distribution},
	plural={prior distributions},
	firstplural={prior distributions},
	type=math,
	text={prior distribution}
}

\newglossaryentry{posteriordist}
{name={posterior distribution},
sort={posteriordistribution},
	description={In the probabilistic\index{posterior distribution} analysis of an \gls{mlsystem}, 
		one typically distinguishes between observed \glspl{datapoint} and unknown 
		\glspl{parameter} of a \gls{probmodel}. The \gls{posterior} distribution\index{posterior distribution} 
		is the \gls{condprobdist} of the unknown \glspl{modelparam}, 
		given (or conditioned on) the observed \glspl{datapoint} \cite{BishopBook}.
		\\ 
		See also: \gls{condprobdist}, \gls{priordist}, \gls{probdist}.},
	first={posterior distribution},
	plural={posterior distributions},
	firstplural={posterior distributions},
	type=math,
	text={posterior distribution}
}

\newglossaryentry{conjugateprior}
 {name={conjugate prior},
 	description={A \gls{priordist}\index{conjugate prior} $\probdist^{(\weights)}$ over \glspl{modelparam} 
 		$\weights$ is called conjugate to a \gls{condprobdist} (referred to as the 
		likelihood \gls{model}) $\probdist^{(\featurevec\mid\weights)}$ if the resulting 
 		\gls{posteriordist} $\probdist^{(\weights \mid\featurevec)}$ belongs to 
 		the same parametric family as $\probdist^{(\weights)}$ \cite{barber2012bayesian}, \cite{BishopBook}. 
 		One important example of a conjugate prior and likelihood \gls{model} is 
 		obtained from the \gls{mvndist}. In particular, consider the prior $\probdist^{(\weights)}$ 
 		is a \gls{mvndist} $\mvnormal{\meanvecgeneric}{\covmtxgeneric}$ with some 
 		\gls{mean} $\meanvecgeneric$ and \gls{covmtx} $\covmtxgeneric$. For a 
 		likelihood \gls{model} $\probdist^{(\featurevec\mid\weights)}$ that is a \gls{mvndist} 
 		with \gls{mean} $\weights$ and \gls{covariance} $\covmtxgeneric$, the resulting \gls{posterior} 
 		$\probdist^{(\weights \mid\featurevec)}$ is also a \gls{mvndist} \cite[Appendix B]{BishopBook}.
		\\
 		See also: \gls{condprobdist}, \gls{mvndist}, \gls{probdist}.},
 	first={conjugate prior},
 	plural={conjugate priors},
 	firstplural={conjugate priors},
 	type=math,
 	text={conjugate prior}
 }

\newglossaryentry{exponentialfamily}
{name={exponential family},
	description={An\index{exponential family} exponential family is a particular \gls{probmodel} 
		for a given \gls{samplespace} $\featurespace$ of possible \glspl{outcome}. It 
		is defined by the specification of a base \gls{measure} $h(\featurevec)$ and 
		a \gls{vector} of sufficient statistics $\mathbf{T}(\featurevec)$. 
		A distribution belongs to the exponential family if its \gls{pdf} (or \gls{pmf})
		can be written as 
		\[ p(\featurevec ;  \weights) = h(\featurevec)\,\exp\!\big(\weights^{T} \mT(\featurevec) - A(\weights)\big),
  		\]
  		where $\weights$ denotes the natural \gls{parameter} and $A(\weights)$ is the log-partition 
       		\gls{function} ensuring normalization $\int_{\featurevec} p(\featurevec ;  \weights) d\featurevec = 1$.
  		Exponential family distributions arise as \gls{maximum}-\gls{entropy} solutions under constraints 
		on the expected values of the sufficient statistics. Thus, we can view them 
		as the least informative \glspl{probdist} for given \glspl{expectation} of the sufficient 
		statistics \cite[Ch. 12]{coverthomas}. 
		\\ 
		See also: \gls{probdist}, \gls{pdf}, \gls{pmf}.},
	first={exponential family},
	plural={exponential families},
	firstplural={exponential families},
	type=math,
	text={exponential family}
}

\newglossaryentry{mdp}
{name={Markov decision process (MDP)},
	description={An MDP\index{Markov decision process (MDP)} is a mathematical 
	        structure for the study of \gls{reinforcementlearning}. 
		Formally, an MDP is a \gls{stochproc} that is defined by a specific choice for 
		\begin{itemize}
    			\item a \gls{statespace} $\statespace$;
    			\item an \gls{actionspace} $\actionspace$;
    			\item a transition \gls{function} $\prob{\state' \mid \state, \arm}$ 
				specifying the \gls{condprobdist} $\probdist^{(\state' \mid \state, \arm)}$
				over the next \gls{state} $\state' \in \statespace$, given the current \gls{state} 
				$\state \in \statespace$ and \gls{action} $\arm \in \actionspace$;
    			\item a \gls{reward} \gls{function} $\reward(\state, \arm) \in \mathbb{R}$ 
				that assigns a numerical \gls{reward} to each \gls{state}-\gls{action}
				pair $(\state,\arm)$.
		\end{itemize}
		For a given \gls{policy} $\policy$, these components define the \gls{probdist} of a \gls{sequence}
		$$ \state_{1},\,\arm_{1},\,\reward_{1},\,\state_{2},\,\arm_{2},\,\reward_{2},\,\ldots,\,\state_{\timeidx},\,\arm_{\timeidx},\,\reward_{\timeidx}$$ 
		of \glspl{rv}. The defining property of an MDP is the \gls{markovprop}. 
		That is, at time instant $\timeidx$, the \gls{condprobdist} of the next \gls{state} 
		$\state_{\timeidx+1}$ and \gls{reward} $\reward_{\timeidx}$ depends on the 
		past only via the current \gls{state} $\state_{\timeidx}$ and \gls{action} $\arm_{\timeidx}$. 
		\Gls{reinforcementlearning} methods try to learn a \gls{policy} $\policy$ that 
		maximizes the expected return:
		\[
			\expect\!\left\{ 
			 \sum_{\timeidx=1}^{\infty} 
			 \discountfac^{\timeidx-1} \reward_{\timeidx} 
			 \;\middle|\; \state_{1} 
			 \right\}. 
		\]
		The conditioning on the initial \gls{state} $\state_{1}$ indicates that the expected 
		return is evaluated by following the \gls{policy} $\policy$ from a given initial \gls{state}.
		The expected return involves the discount factor $\discountfac \in (0,1)$ 
		that determines the relative importance of future \glspl{reward} compared 
		to the immediate \gls{reward}. 
		The discount factor $\discountfac$ is typically fixed for a given MDP and 
		controls the trade-off between short-term and long-term \gls{reward}. 
		MDPs are widely used in robotics, game playing, and autonomous systems to model 
		decision-making problems where an \gls{agent} interacts with an \gls{environment} to 
		achieve a goal \cite{SuttonEd2}, \cite{BertsekasDynOptII}, \cite{BertsekasDynProgVolI}.
		\\
		See also: \gls{reinforcementlearning}, \gls{stochproc}, \gls{function}, \gls{reward}.},
 	first={Markov decision process (MDP)},
	type=reinflearning, 
	plural={MDPs}, 
	firstplural={Markov decision processes (MDPs)}, 
 	text={MDP} 
}

\newglossaryentry{statevaluefunction}
{name={state-value function},
	description={For a given \gls{mdp}, any \gls{policy} $\policy$ naturally induces 
		a \gls{valuefunction}\index{state-value function} 
		$\statevaluefunc{\policy}:\statespace \rightarrow \mathbb{R}$. 
		The value $\statevaluefunc{\policy}(\state)$ is the expected 
		return when the \gls{mdp} starts in a given \gls{state} $\state \in \statespace$ 
		and \glspl{action} are selected according to $\policy$.
		\\
		See also: \gls{mdp}, \gls{valuefunction}, \gls{state}. },
	first={state-value function},
	type=reinflearning,
	plural={state-value functions},
	firstplural={state-value functions},
	text={state-value function}
}

\newglossaryentry{valuefunction}
{name={value function},
	description={In the context of an \gls{mdp}, a value \gls{function}\index{value function} 
		$\valuefunc: \statespace \rightarrow \mathbb{R}$ assigns to each \gls{state} 
		$\state \in \statespace$ a real number $\valuefunc(\state)$ that quantifies 
		the long-term desirability of being in \gls{state} $\state$.
		\\
		See also: \gls{state}. },
	first={value function},
	type=reinflearning,
	plural={value functions},
	firstplural={value functions},
	text={value function}
}

\newglossaryentry{policy}
{name={policy (reinforcement learning)},
sort={policyreinforcement},
	description={A policy\index{policy} is a \gls{function} that specifies how 
	 	the next \gls{action} $\arm_{\timeidx}$ in an \gls{mdp} 
		is chosen when the current \gls{state} is $\state_{\timeidx}$. 
		Typically, a policy is \gls{stochastic}, meaning that it defines a 
		\gls{condprobdist} $\probdist^{(\arm \mid \state)}$ over the 
		\glspl{action} for a given current \gls{state}. We can view 
		a policy also as a \gls{hypothesis} that uses \glspl{feature} 
		derived from the current \gls{state} to predict the best next 
		\gls{action} \cite{SuttonEd2}.
		\\
		See also: \gls{action}, \gls{mdp}, \gls{state}.},
	first={policy},
	type=reinflearning,
	plural={policies},
	firstplural={policies},
	text={policy}
}

\newglossaryentry{bellmanoperator}
{name={Bellman operator},
	description={The Bellman \gls{operator}\index{Bellman operator} $\fixedpointop$ 
		associated with an \gls{mdp} is defined on the space of all 
		\glspl{valuefunction}. In particular, it maps a 
		\gls{valuefunction} $\valuefunc: \statespace \rightarrow \mathbb{R}$ 
		to another \gls{valuefunction} $\valuefunc': \statespace \rightarrow \mathbb{R}$ 
		as follows:
		\[
		       	\valuefunc'(\state) =\max_{\arm \in \actionspace}
			\Bigl(
			\expect\{ \reward(\state,\arm)\mid \state,\arm \}
			+ \discountfac \, \expect\{ \valuefunc(\state') \mid \state,\arm \}
			\Bigr),
		\]
		where $\discountfac \in (0,1)$ is a discount factor and $\state'$ is the 
		next \gls{state} generated according to the transition \gls{function} 
		$\state' \sim \prob{\state' \mid \state,\arm}$.
		The \gls{statevaluefunction} $\valuefunc^\star$ of the optimal 
		\gls{policy} $\policy^\star$ is a \gls{fixedpoint} of the Bellman \gls{operator}, 
		$\valuefunc^\star = \fixedpointop \valuefunc^\star$. This \gls{fixedpointeq} 
		naturally lends itself to the \gls{valueiteration} method for computing the \gls{statevaluefunction} 
		of an optimal \gls{policy}. Besides the Bellman \gls{operator} 
		associated with an \gls{mdp}, there is also a Bellman \gls{operator} 
		$\fixedpointop^{(\policy)}$ associated with a \gls{policy} $\policy$. 
		In this case, the Bellman \gls{operator} is defined as
		\[
			\fixedpointop^{(\policy)} \valuefunc(\state) = \expect\{ \reward(\state,\arm) \mid \state,\arm \}
			+ \discountfac \, \expect\{ \valuefunc(\state') \mid \state,\arm \},
		\]
		where $\state' \sim \prob{\state' \mid \state,\arm}$ and $\arm$ is selected
		according to $\policy$. The \gls{statevaluefunction} $\statevaluefunc{\policy}$ is a 
		\gls{fixedpoint} of $\fixedpointop^{(\policy)}$, $\statevaluefunc{\policy}= \fixedpointop^{(\policy)}\statevaluefunc{\policy}$. 
		This \gls{fixedpointeq} can be solved by a \gls{fixedpointiter} that is known as \gls{policyevaluation}. 
		The Bellman \gls{operator} is named after Richard Bellman, who introduced it in the context of
		dynamic programming \cite{Bellman1957}. The Bellman \gls{operator} is a key
		concept in \gls{reinforcementlearning} and is used to derive \glspl{algorithm} for
		solving \glspl{mdp}, such as \gls{valueiteration} and \gls{policy} \gls{iteration}\cite{SuttonEd2}.
		\\
		See also: \gls{mdp}, \gls{valuefunction}, \gls{policy}, \gls{valueiteration}, \gls{contractop}, \gls{banachfixedpoint}. }, 
	first={Bellman operator},
	type=reinflearning,
	plural={Bellman operators},
	firstplural={Bellman operators},
	text={Bellman operator}
}

\newglossaryentry{policyevaluation}
{name={policy evaluation (reinforcement learning)},
sort={policyevaluation},
	description={\Gls{policy} evaluation\index{policy evaluation} refers to computing 
		the \gls{statevaluefunction} $\statevaluefunc{\policy}$ 
		of a given \gls{policy} $\policy$ in an \gls{mdp}. One widely used method, referred to 
		as iterative \gls{policy} evaluation, is based on the characterization of 
		$\statevaluefunc{\policy}$ as a \gls{fixedpoint} of the \gls{bellmanoperator} 
		$\fixedpointop^{(\policy)}$. In particular, starting from an initial 
		\gls{valuefunction} $\valuefunc_{0}$, we iteratively apply the \gls{bellmanoperator}
		$\fixedpointop^{(\policy)}$ to obtain a \gls{sequence} of \glspl{valuefunction} 
		$\valuefunc_{1}, \,\valuefunc_{2}, \,\ldots$ as follows:
		\[
		\valuefunc_{\iteridx+1} = \fixedpointop^{(\policy)} \valuefunc_{\iteridx}, \qquad \iteridx=0,\,1,\,2,\,\ldots.
		\]
		Under mild conditions, this \gls{fixedpointiter} converges to 
		$\statevaluefunc{\policy}$ as $\iteridx \rightarrow \infty$ \cite[Sec. 4.2]{SuttonEd2}.
		\\
		See also: \gls{policy}, \gls{statevaluefunction}, \gls{mdp}. }, 
	first={policy evaluation},
	type=reinflearning,
	text={policy evaluation}	 
}

\newglossaryentry{valueiteration}
{name={value iteration},
	description={Consider an \gls{mdp} with the associated \gls{bellmanoperator} $\fixedpointop$. The 
		\gls{statevaluefunction} $\valuefunc^\star$ of the optimal \gls{policy} 
		is a \gls{fixedpoint} of $\fixedpointop$, i.e., $\valuefunc^\star = \fixedpointop \valuefunc^\star$. 
		Value \gls{iteration}\index{value iteration} is the \gls{fixedpointiter} for computing
		$\valuefunc^\star$ by repeatedly applying $\fixedpointop$ to an initial 
		\gls{valuefunction} $\valuefunc_{0}$ \cite[Sec. 4.4]{SuttonEd2}.
		\\
		See also: \gls{statevaluefunction}, \gls{fixedpointiter}, \gls{valuefunction}. }, 
	first={value iteration},
	type=reinflearning,
	plural={value iterations},
	firstplural={value iterations},
	text={value iteration} 
}

\newglossaryentry{agent}
{name={agent},
	description={An agent\index{agent} denotes any system that implements some form of 
		\gls{onlinelearning} \cite{SuttonEd2}, \cite{RusselNorvig}. During each time 
		step $\timeidx$, an agent receives a \gls{featurevec} $\featurevec^{(\timeidx)}$ 
		that provides (typically incomplete) information about the underlying \gls{state} 
		of the system. The agent then applies its current \gls{hypothesis} $\hypothesis^{(\timeidx)}$ to select 
		an \gls{action} $\arm^{(\timeidx)}=\hypothesis^{(\timeidx)}(\featurevec^{(\timeidx)})$. 
		It then receives a \gls{reward} $\reward^{(\timeidx)}$ that quantifies the usefulness of this \gls{action} 
		and, in turn, guides the adaptation of its \gls{hypothesis} (or \gls{modelparam}). 
		\\ 
		See also: \gls{hypothesis}, \gls{action}, \gls{reward}, \gls{reinforcementlearning}.},
	first={agent},
	type=reinflearning, 
	plural={agents}, 
	text={agent},
	firstplural={agents}
 }

\newglossaryentry{environment}
{name={environment (reinforcement learning)},
	description={An environment\index{environment} denotes the external system with which an \gls{agent} 
		interacts over time. During each time step $\timeidx$, the environment provides the 
		\gls{agent} with a \gls{featurevec} $\featurevec^{(\timeidx)}$ and, in response to the 
		\gls{action} $\arm^{(\timeidx)}$ by the \gls{agent}, a \gls{reward} $\reward^{(\timeidx)}$ \cite{SuttonEd2}.
		\\ 
		See also: \gls{agent}, \gls{featurevec}, \gls{reward}.},
	first={environment},
	plural={environments},
	text={environment},
	firstplural={environments},
	type=reinflearning
}

\newglossaryentry{reinforcementlearning}
{name={reinforcement learning (RL)},
	description={RL\index{reinforcement learning (RL)} refers to an \gls{onlinelearning} setting where 
		we can only evaluate the usefulness of a single \gls{hypothesis} (i.e., a specific
		choice of \glspl{modelparam}) at each time step $\timeidx$. In particular, RL 
		methods apply the current \gls{hypothesis} $\hypothesis^{(\timeidx)}$ to the 
		\gls{featurevec} $\featurevec^{(\timeidx)}$ of the newly received \gls{datapoint}
		to predict the next \gls{action}. 
		The usefulness of the resulting \gls{prediction} $\hypothesis^{(\timeidx)}(\featurevec^{(\timeidx)})$ 
		is quantified by a \gls{reward} signal $\reward^{(\timeidx)}$ (see Fig.\ \ref{fig_reinforcementlearning_dict}). 
		\begin{figure}[H]
		\begin{center}
			\begin{tikzpicture}[scale=1]
			\draw[->] (-2, 0) -- (6, 0);
			\node at (6.3, 0) {$\hypothesis$};
			\draw[thick, blue, domain=0:3, samples=20] plot (\x-3, {-0.2*(\x)^2 + 2});
			\node[anchor=west,yshift=4pt] at (0-3, {-0.2*(0)^2 + 2}) {$-\loss^{(\timeidx)}(\hypothesis)$};
			\filldraw[blue] (1.5-3, {-0.2*(1.5)^2 + 2}) circle (2pt);
			\node[anchor=north] at (1.5-3, -0.3) {$\hypothesis^{(\timeidx)}$};		
			\draw[dotted] (1.5-3, 0) -- (1.5-3, {-0.2*(1.5)^2 + 2});
			\draw[thick, red, domain=0:5, samples=20, dashed] plot (\x, {-0.15*(\x - 2)^2 + 3});
			\node[anchor=west,yshift=4pt] at (3, {-0.15*(3 - 2)^2 + 3}) {$-\loss^{(\timeidx+1)}(\hypothesis)$};
			\filldraw[red] (2, {-0.15*(2 - 2)^2 + 3}) circle (2pt);
			\node[anchor=north] at (2, -0.3) {$\hypothesis^{(\timeidx+1)}$};
			\draw[dotted] (2, 0) -- (2, {-0.15*(3 - 2)^2 + 3});
			\draw[thick, green!60!black, domain=3:5, samples=20, dotted] plot (\x+2, {-0.1*(\x - 4)^2 + 1.5});
			\node[anchor=west,yshift=4pt] at (4.5+2, {-0.1*(4.5 - 4)^2 + 1.5}) {$-\loss^{(\timeidx+2)}(\hypothesis)$};
			\filldraw[green!60!black] (3.5+2, {-0.1*(3.5 - 4)^2 + 1.5}) circle (2pt);
			\node[anchor=north] at (3.5+2, -0.3) {$\hypothesis^{(\timeidx+2)}$};
			\draw[dotted] (3.5+2, 0) -- (3.5+2, {-0.1*(3.5 - 4)^2 + 1.5});
			\end{tikzpicture}
		\caption{Three consecutive time steps $\timeidx,\timeidx+1,\timeidx+2$ with corresponding \glspl{lossfunc} $\loss^{(\timeidx)},
			\loss^{(\timeidx+1)}, \loss^{(\timeidx+2)}$. During time step $\timeidx$, an RL method can evaluate the 
			\gls{lossfunc} only for one specific \gls{hypothesis} $\hypothesis^{(\timeidx)}$, resulting in the \gls{reward} 
			signal $\reward^{(\timeidx)}=-\loss^{(\timeidx)}(\hypothesis^{(\timeidx)})$. \label{fig_reinforcementlearning_dict}}
		\end{center}
		\end{figure}
		In general, the \gls{reward} depends also on the 
		previous \glspl{prediction} $\hypothesis^{(\timeidx')}\big(\featurevec^{(\timeidx')}\big)$ 
		for $\timeidx' < \timeidx$. The goal of RL is to learn $\hypothesis^{(\timeidx)}$, for 
		each time step $\timeidx$, such that the (possibly discounted) cumulative \gls{reward} 
		is maximized \cite{MLBasics}, \cite{SuttonEd2}.
		\\
		See also: \gls{reward}, \gls{lossfunc}, \gls{ml}.},
	first={reinforcement learning (RL)},
	type=reinflearning,
	text={RL}
}

\newglossaryentry{action}
{name={action},
	description={An action\index{action} refers to a decision taken by an \gls{aisystem}
		at a given time step $\timeidx$ that influences the observed 
		\gls{reward} signal. The actions are elements of an 
		\gls{actionspace} $\actionspace$ and are typically denoted by 
		$\arm_{\timeidx} \in \actionspace$. The action $\arm_{\timeidx}$ is selected 
		based on the \gls{featurevec} $\featurevec^{(\timeidx)}$ (which collects 
		all available observations) and the current 
		\gls{hypothesis} $\hypothesis^{(\timeidx)}$. 
		\Gls{reinforcementlearning} uses \gls{onlinelearning} methods 
		to learn a \gls{hypothesis} $\hypothesis^{(\timeidx)}$ that predicts 
		a (nearly) optimal action. The usefulness of the 
		\gls{prediction} $\arm_{\timeidx}$ is evaluated indirectly 
		through the resulting \gls{reward} signal $\reward^{(\timeidx)}$.
		In the special case of an \gls{mab}, the set of possible actions is 
		finite and each action corresponds to selecting one arm. In more
		general \gls{reinforcementlearning} settings, the \gls{actionspace} may be \gls{continuous}.
		\\
		See also: \gls{reward}, \gls{hypothesis}, \gls{reinforcementlearning}, \gls{mab}, \gls{lossfunc}.},
	first={action},
	type=reinflearning,
	plural={actions},
	firstplural={actions},
	text={action}
}

\newglossaryentry{actionspace}
{name={action space},
	description={See\index{action space} \gls{action}.} ,
	first={action space},
	type=reinflearning,
	plural={action spaces},
	firstplural={action spaces},
	text={action space}
}

\newglossaryentry{mab}
{name={multiarmed bandit (MAB)},
	description={An MAB\index{multiarmed bandit (MAB)} is a precise 
		formulation of a sequential decision-making task under \gls{uncertainty}. 
		At each time step $\iteridx$, one must choose an \gls{action} from a finite 
		\gls{actionspace} $\actionspace$. Choosing \gls{action} 
		$\arm$ at time $\iteridx$ yields a \gls{reward} $\reward^{(\arm,\iteridx)}$. 
		Each MAB induces an \gls{ml} problem, i.e., to learn a \gls{hypothesis} that 
		predicts the optimal \gls{action} $\arm_{\timeidx}$ at time $\timeidx$. 
		This \gls{prediction} must be based on the \glspl{action} 
		and \glspl{reward} received up to time $\timeidx-1$ 
		\cite{SuttonEd2}, \cite{Bubeck2012}.
			\\ 
		See also: \gls{reward}, \gls{regret}.},
	first={MAB},
	type=reinflearning,
	text={MAB}
}

\newglossaryentry{stochmab}
{name={stochastic multiarmed bandit (stochastic MAB)},
	description={A stochastic \gls{mab}\index{stochastic multiarmed bandit (stochastic MAB)} 
		is a \gls{stochproc} that is obtained from an \gls{mab}. In particular, 
		the \gls{reward} $\reward^{(\arm,\iteridx)}$ is modeled as an \gls{rv}		 
		with an unknown \gls{probdist} $\probdist^{\big(\reward^{(\arm,\iteridx)}\big)}$ 
		\cite{HazanOCO}, \cite{Bubeck2012}. 
		In the simplest setting, the \gls{probdist} $\probdist^{\big(\reward^{(\arm,\iteridx)}\big)}$ 
		does not depend on $\timeidx$, i.e., it is time invariant. 
		\\ 
		See also: \gls{reward}, \gls{regret}.},
	first={stochastic multiarmed bandit (stochastic MAB)},
	type=reinflearning,
	text={stochastic MAB}
}

\newglossaryentry{regret}
{name={regret},
	description={The regret\index{regret} of a \gls{hypothesis} $\hypothesis$ relative to 
		another \gls{hypothesis} $\hypothesis'$, which serves as a \gls{baseline}, 
		is the difference between the \gls{loss} incurred by $\hypothesis$ and the \gls{loss} 
		incurred by $\hypothesis'$ \cite{PredictionLearningGames}. 
		The \gls{baseline} \gls{hypothesis} $\hypothesis'$ is also referred to as an \gls{expert}.
					\\ 
		See also: \gls{baseline}, \gls{loss}, \gls{expert}.},
	first={regret},
	type=reinflearning,
	text={regret} 
}   
\newglossaryentry{provider}
{name={provider (AI Act)},
	description={The\index{provider (AI Act)} EU AI Act defines a provider 
		as a natural or legal person, public
		authority, agency (e.g., the European Medicines Agency), or other body
		that develops an \gls{aisystem} or a
		general-purpose AI \gls{model} and places it on the market or puts it
		into service under the provider's own name or trademark, whether for payment
		or free of charge (Article~3(3), \cite{EUAIAct2024}).
		The provider bears the primary regulatory obligations under the AI Act,
		including conformity assessment, technical documentation, and monitoring.
		A provider is distinct from a deployer, who integrates an already-placed
		\gls{aisystem} into their own products or processes.
		For example, a company that trains a \gls{model} for credit scoring
		and offers it via an \gls{apiendpoint} acts as a provider, whereas
		a bank that queries that \gls{apiendpoint} to support its own loan
		decisions acts as a deployer.
		\\
		See also: \gls{aisystem}, \gls{model}, \gls{apiendpoint}, \gls{mlaas}.},
	first={provider},
	plural={providers},
	firstplural={providers},
	type=regulation,
	text={provider}
}

\newglossaryentry{personaldata}
{name={personal data},
	description={Personal \gls{data} are any information\index{personal data} 
		relating to an identified or identifiable natural 
		person (i.e., the \gls{data} subject). A natural person is identifiable 
		if they can be identified, directly or indirectly, in particular 
		by reference to an identifier such as a name, an identification number, 
		location \gls{data}, an online identifier, or one or more factors specific 
		to the physical, physiological, genetic, mental, economic, cultural, 
		or social identity of that person \cite{GDPR2016}.
		In \glspl{mlsystem}, personal \gls{data} may occur in \gls{training} \gls{data}, 
		\gls{model} inputs, intermediate representations (e.g., \glspl{featurevec} or embeddings), 
		or \gls{model} \glspl{output}, provided that the information relates to an identifiable natural 
		person. The EU AI Act does not introduce a separate definition of personal 
		\gls{data}; instead, whenever an \gls{aisystem} processes personal \gls{data}, 
		the \gls{gdpr} definition and obligations apply in full.
		\\
		See also: \gls{data}, \gls{aisystem}, \gls{gdpr}. },
  	first={personal data},
    	type=regulation, 
	text={personal data}
}

\newglossaryentry{profiling}
{name={profiling},
	description={Profiling\index{profiling} aims at identifying patterns and making inferences about 
		individuals based on their \gls{data}.
		Profiling techniques use \gls{ml} methods to predict individuals' performance at work, 
		economic situation, health, or personal preferences. Profiling is
		instrumental in targeted advertising, credit scoring, fraud detection, 
		and personalized services. The \gls{gdpr} imposes strict requirements 
		on organizations that engage in profiling activities to ensure that individuals' 
		rights are protected \cite{GDPR2016}.
		\\
		See also: \gls{data}, \gls{gdpr}. },
	first={profiling},
	type=regulation, 
	text={profiling}
}

\newglossaryentry{interpretability}
{name={interpretability},
	description={An \gls{ml} method is interpretable\index{interpretability} for a 
 		human user if they can comprehend the decision process of the method. 
		One approach to develop a precise definition of interpretability is via the concept  
 		of simulatability, i.e., the ability of a human to mentally simulate the \gls{model} behavior 
		\cite{Colin:2022aa}, \cite{Chen2018}, \cite{doshi2017towards}, \cite{hase-bansal-2020-evaluating}, \cite{Lipton2018}. 
 		The idea is as follows: If a human user understands an \gls{ml} method, then they should 
 		be able to anticipate its \glspl{prediction} on a \gls{testset}. We illustrate 
 		such a \gls{testset} in Fig.\ \ref{fig_aug_simulatability_dict}, which also depicts 
		two learned \glspl{hypothesis} $\learnthypothesis$ and $\learnthypothesis'$. 
		The \gls{ml} method producing the \gls{hypothesis} $\learnthypothesis$ is interpretable
		to a human user familiar with the concept of a \gls{linearmap}. 
    		Since $\learnthypothesis$ corresponds to a \gls{linearmap}, the user can 
		anticipate the \glspl{prediction} of $\learnthypothesis$ on the 
		\gls{testset}. In contrast, the \gls{ml} method delivering $\learnthypothesis'$ 
		is not interpretable, because its behavior is no longer aligned with the user’s 
		\glspl{expectation}.
 		\begin{figure}[H]
 			\begin{center} 
			\begin{tikzpicture}[x=1.5cm, y=1cm]
 			\def\slope{0.4}
  			\def\offset{2.0}
  			\draw[->, very thick] (0,0.5) -- (7.7,0.5) node[below, xshift=-1cm] {$\feature$}; 
 			\draw[->, very thick] (0.5,0) -- (0.5,4.2) node[above] {$\truelabel$};           
  			\draw[color=black, thick, dashed, domain=-0.5:7.2, variable=\x] 
    			plot ({\x},{\slope*\x + \offset});
  			\draw[color=black, thick, dashed, domain=4:7.2, variable=\x] 
    			plot ({\x},{\slope*\x + \offset-(\x-4)*0.5});
  			\node[above] at (7.2, {\slope*7.2 + \offset}) {$\learnthypothesis(\feature)$};
  			\node[above] at (7.2, {\slope*7.2 + \offset - 0.5*(7.2 - 4)}) {$\learnthypothesis'(\feature)$};
  			\foreach \x/\y/\c/\s in {
      			1.2/1.0/blue/6, 1.4/1.0/blue/6, 1.7/1.0/blue/6,
      			2.2/3.9/blue/12, 2.6/4.2/blue/12, 3.0/4.4/blue/12
  			}{
    			\coordinate (pt) at (\x,\y);
    			\node[fill=\c, circle, draw, minimum size=\s pt, scale=0.6] at (pt) {};
    			\draw[<->, >={Latex[width=2mm,length=4mm]}, color=\c, thick]
      			(\x, {\slope*\x + \offset}) -- (pt);
  			}
    			\foreach \x/\y/\c/\s in {
       			5.7/2.6/red/12, 5.9/2.6/red/12, 6.2/2.6/red/12
   			}{
     			\coordinate (pt) at (\x,{\slope*\x + \offset});
     			\node[fill=\c, circle, draw, minimum size=\s pt, scale=0.6] at (pt) {};
   			}
  			\draw[fill=blue] (4.2, 1.7) circle (0.1cm) node [black,xshift=0.2cm,anchor=west] {\gls{trainset} $\dataset$};
  			\draw[fill=red]  (4.2, 1.2) circle (0.1cm) node [black,xshift=0.2cm,anchor=west] {\gls{testset} $\dataset'$};
			\end{tikzpicture}
 		\caption{We can assess the interpretability of trained \gls{ml} \glspl{model} 
 			$\learnthypothesis$ and $\learnthypothesis'$ by comparing their \glspl{prediction} 
			to pseudo-\glspl{label} generated by a human user for $\dataset'$. 
			\label{fig_aug_simulatability_dict}}
 			\end{center}
	 	\end{figure} 
 	 	The notion of interpretability is closely related to the notion of \gls{explainability}, 
 	 	as both aim to make \gls{ml} methods more understandable for humans. 
		In the context of Fig.\ \ref{fig_aug_simulatability_dict}, interpretability of an \gls{ml} 
	 	method $\learnthypothesis$ requires that the human user can anticipate its \glspl{prediction} 
	 	on an arbitrary \gls{testset}. This contrasts with \gls{explainability}, 
		where the user is provided with \glspl{explanation} to better understand the \glspl{prediction}
		of $\learnthypothesis$ on a specific \gls{testset} $\dataset'$. These \glspl{explanation} 
		may take the form of saliency \glspl{map} or reference examples from the \gls{trainset}.
	 	\\ 
	 	See also: \gls{explainability}, \gls{trustAI}, \gls{regularization}, \gls{lime}. },
	first={interpretability},
	type=regulation, 
 	text={interpretability}
}

\newglossaryentry{counterfactual_explanation}
{name={counterfactual explanation},
	description={A counterfactual \gls{explanation}\index{counterfactual explanation} 
		is a type of \gls{explanation} that enhances the \gls{transparency} 
		of a trained \gls{ml} \gls{model}. It describes how the \glspl{feature} 
		of a given \gls{datapoint} would need to change in order to obtain 
		a different \gls{prediction}. Consider a trained \gls{model} 
		$\learnthypothesis:\featurespace \rightarrow\labelspace$ and a 
		\gls{featurevec} $\featurevec$ with \gls{prediction} $\learnthypothesis(\featurevec)=\predictedlabel$. 
		A counterfactual \gls{explanation} specifies an alternative 
		\gls{featurevec} $\featurevec'$ such that
		$$\learnthypothesis(\featurevec')  \neq \predictedlabel.$$
		\\
		See also: \gls{explanation}, \gls{explainability}, \gls{classification}.},
	first={counterfactual explanation},
	plural={counterfactual explanations},
	type=regulation,
	firstplural={counterfactual explanations},
	text={counterfactual explanation}
}

\newglossaryentry{explainability}
{name={explainability},
	description={We\index{explainability} define the (subjective) explainability of an \gls{ml} method 
		as the level of simulatability \cite{Colin:2022aa} of the \glspl{prediction} 
		delivered by an \gls{mlsystem} to a human user. Quantitative \glspl{measure} of the 
		(subjective) explainability of a trained \gls{model} can be constructed by 
		comparing its \glspl{prediction} with the \glspl{prediction} provided by a user 
		on a \gls{testset} \cite{Colin:2022aa}, \cite{Zhang:2024aa}. Alternatively, we can use 
		\glspl{probmodel} for \gls{data} and measure the explainability of a trained \gls{ml} 
		\gls{model} via the conditional (differential) \gls{entropy} of its \glspl{prediction}, given the 
		user's \glspl{prediction} \cite{JunXML2020}, \cite{Chen2018}.
						\\ 
		See also: \gls{trustAI}, \gls{regularization}.},
	first={explainability},
	type=regulation,
	text={explainability}
}

\newglossaryentry{fairprinciples}
{name={FAIR principles},
	description={The FAIR principles\index{FAIR principles} are guidelines for 
		scientific \gls{data} management. The aim is to make research 
		artifacts findable, accessible, interoperable, and reusable \cite{Wilkinson:2016aa}.
		FAIR-compliant metadata is a key enabler of auditability, as it 
		supports the traceability and reproducible inspection of an \gls{mlsystem} \cite{Samuel2020}. 
		\\
		See also: \gls{data}, \gls{trustAI}.},
	first={FAIR principles},
	type=regulation, 
	text={FAIR principles}
}

\newglossaryentry{shap}
{name={SHapley Additive exPlanations (SHAP)},
	description={SHAP\index{SHapley Additive exPlanations (SHAP)} is a post hoc method for 
		explaining the \gls{prediction} $\learnthypothesis(\featurevec)$ 
		of a trained \gls{model} $\learnthypothesis \in \hypospace$ 
		at a given \gls{featurevec} $\featurevec=\big( \feature_1,\,\dots,\,\feature_d\big)^{T}$.  
		SHAP values are computed after \gls{model} \gls{training} and can 
		be used to analyze the relative importance of different \glspl{feature} 
		for the \gls{prediction} $\learnthypothesis(\featurevec)$ \cite{Molnar2019}.
		\\
		See also: \gls{prediction}, \gls{training}. },
	first={SHapley Additive exPlanations (SHAP)},
	type=regulation, 
	text={SHAP}
}

\newglossaryentry{automateddecisionmaking}
{name={automated decision-making},
	description={Automated decision-making\index{automated decision-making} 
		refers to \gls{ml} applications that use \glspl{prediction} 
		delivered by a trained \gls{model} directly (i.e., without human involvement) 
		to make decisions affecting individuals. Under the \gls{gdpr}, 
		individuals have the right not to be subject to decisions based 
		solely on automated processing, when these decisions produce 
		legal or similarly significant effects, unless appropriate 
		safeguards (e.g., human oversight, contestability, or explicit consent) 
		are implemented.
		\\
		See also: \gls{ml}, \gls{gdpr}. }, 
	first={automated decision-making},
	type=regulation,
	text={automated decision-making}
}

\newglossaryentry{aisystem}
{name={artificial intelligence system (AI system)},
	description={The\index{artificial intelligence system (AI system)} EU Artificial Intelligence Act 
		(AI Act) \cite{AIact} defines  
		an \gls{ai} system as a machine-based system that is designed 
		to operate with varying levels of autonomy and that may 
		exhibit adaptiveness (e.g., \gls{model} retraining) 
		after deployment. \Gls{ai} systems compute \glspl{prediction} that can 
              	influence environments or decisions \cite{EUAIAct2024}. In line with 
              	this definition, regulatory obligations and risk classifications apply 
              	at the level of the \gls{ai} system rather than at the level of individual 
              	\glspl{model} or \glspl{algorithm}.  
		The system-level view emphasizes that properties such as \gls{robustness}, 
		fairness, and \gls{transparency} emerge from the interaction of 
		\glspl{model}, \gls{data}, and operational context, rather than from 
		isolated components. 
		\\
		See also: \gls{ai}, \gls{robustness}, \gls{transparency}. },
 	first={artificial intelligence system (AI system)},
 	plural={AI systems},
 	firstplural={artificial intelligence systems (AI systems)},
 	type=regulation,
 	text={AI system}
}

\newglossaryentry{gdpr}
{name={general data protection regulation (GDPR)},
	description={The\index{general data protection regulation (GDPR)} GDPR
		was enacted by the European Union (EU), effective from 25 May 2018 \cite{GDPR2016}. 
		It safeguards the privacy and \gls{data} rights of individuals in the EU. 
		The GDPR has significant implications for how \gls{data} are collected, stored, 
		and used in \gls{ml} applications. Key provisions include the following:
		\begin{itemize}
			\item \Gls{dataminprinc}: \glspl{mlsystem} should only use the necessary 
			amount of \gls{personaldata} for their purpose.
			\item \Gls{transparency} and \gls{explainability}: \glspl{mlsystem} should 
			enable their users to understand how the systems make decisions that impact the users.
			\item \Gls{data} subject rights: Users should get an opportunity to access, 
			rectify, and delete their \gls{personaldata}, as well as to object to 
			\gls{automateddecisionmaking} and \gls{profiling}.
			\item Accountability: Organizations must ensure robust \gls{data} security 
			and demonstrate compliance through documentation and regular audits.
		\end{itemize}
		See also: \gls{data}, \gls{ml}, \gls{dataminprinc}, \gls{transparency}, \gls{explainability}.}, 
	first={general data protection regulation (GDPR)},
	type=regulation, 
	text={GDPR}
}

\newglossaryentry{highriskaisystem}
{name={high-risk artificial intelligence system (high-risk AI system)},
	description={A subset\index{high-risk artificial intelligence system (high-risk AI system)} 
		of \glspl{aisystem} is classified as 
		high-risk due to its potential to significantly impact 
		safety, fundamental rights, or critical societal functions. 
		High-risk \glspl{aisystem} are subject to stringent regulatory 
		requirements under the EU AI Act, 
		including conformity assessments, risk management, 
		\gls{transparency} obligations, and post-market monitoring \cite{EUAIAct2024}. 
		Examples of high-risk \glspl{aisystem} include those used in 
		critical infrastructure, education, employment, law enforcement, 
		and biometric identification.
		\\
		See also: \gls{aisystem}, \gls{transparency}. },
 	first={high-risk artificial intelligence system (high-risk AI system)},
 	plural={high-risk AI systems},
 	firstplural={high-risk artificial intelligence systems (high-risk AI systems)},
 	type=regulation,
	text={high-risk AI system}
}

\newglossaryentry{deepfake}
{name={deep fake},
	description={Deep fakes are synthetic\index{deep fake} media generated or 
		substantially modified by an \gls{aisystem} such that it falsely appears to depict 
		a real person, object, or \gls{event}. Deep fakes are typically produced using 
             	generative methods, trained to imitate visual, audio, or audiovisual 
            	characteristics of real \gls{data}. From a system perspective, deep fakes are 
            	characterized by a deliberate mismatch between the observable content 
             	and its true origin, which can lead to deception, misinformation, or 
             	manipulation. 
		\\
		See also: \gls{aisystem}. },
 	first={deep fake},
 	plural={deep fakes}, 
 	firstplural={deep fakes},
 	type=regulation,
 	text={deep fake}
}

\newglossaryentry{transparency}
{name={transparency},
	description={Transparency\index{transparency} is a fundamental requirement for 
		\gls{trustAI} \cite{HLEGTrustworhtyAI}. In the context of \gls{ml} 
		methods, transparency is often used interchangeably with \gls{explainability} 
		\cite{JunXML2020}, \cite{gallese2023ai}. However, in the broader scope of \glspl{aisystem}, 
		transparency extends beyond \gls{explainability} and includes providing information 
		about the system’s limitations, reliability, and intended use. 
		In medical diagnosis systems, transparency requires disclosing the confidence level 
		for the \glspl{prediction} delivered by a trained \gls{model}. In credit scoring, 
		\gls{ai}-based lending decisions should be accompanied by \glspl{explanation} of 
		contributing factors, such as income level or credit history. These \glspl{explanation} 
		allow humans (e.g., a loan applicant) to understand and contest automated decisions. 
		Some \gls{ml} methods inherently offer transparency. For example, \gls{logreg} 
		provides a quantitative \gls{measure} of \gls{classification} reliability through the value $|\hypothesis(\featurevec)|$. 
		\Glspl{decisiontree} are another example, as they allow human-readable decision rules \cite{rudin2019stop}.
		Transparency also requires a clear indication when a user is engaging with an \gls{aisystem}. 
		For example, \gls{ai}-powered chatbots should notify users that they are interacting with an 
		automated system rather than a human. Furthermore, transparency encompasses comprehensive 
		documentation detailing the purpose and design choices underlying the \gls{aisystem}. 
		For instance, \gls{model} datasheets \cite{DatasheetData2021} and \gls{aisystem} cards \cite{10.1145/3287560.3287596} 
		help practitioners understand the intended use cases and limitations of an \gls{aisystem} \cite{Shahriari2017}.
					\\ 
		See also: \gls{trustAI}, \gls{explainability}.},
	first={transparency}, 
	type=regulation,
	text={transparency} 
}

\newglossaryentry{trustAI}
{name={trustworthy artificial intelligence (trustworthy AI)},
	description={Besides the \glspl{compasp} and \glspl{statasp}, a third main design aspect of 
		\gls{ml} methods is their trustworthiness\index{trustworthy artificial intelligence (trustworthy AI)} 
		\cite{pfau2024engineeringtrustworthyaideveloper}. 
		The EU has put forward seven key requirements (KRs) for trustworthy 
		\gls{ai} (which typically build on \gls{ml} methods) \cite{ALTAIEU}: 
		\begin{enumerate}[label=\arabic*)]
			\item KR1 – Human agency and oversight;
			\item KR2 – Technical \gls{robustness} and safety;
			\item KR3 – Privacy and \gls{data} governance;
			\item KR4 – \Gls{transparency};
			\item KR5 – Diversity, nondiscrimination and fairness; 
			\item KR6 – Societal and environmental well-being;
			\item KR7 – Accountability. 
		\end{enumerate}
		See also: \gls{compasp}, \gls{statasp}, \gls{ml}, \gls{ai}, \gls{robustness}, \gls{data}, \gls{transparency}.},
	first={trustworthy artificial intelligence (trustworthy AI)},
	type=regulation,
	text={trustworthy AI}
}

\newglossaryentry{commnetwork}
{name={communication network},
	description={A communication network\index{communication network} is a collection
		of \glspl{device} interconnected by communication links that allow them to
		exchange information \cite{DistributedSystems}, \cite{TanenbaumComputerNet}.
		For example, a mobile network consists of \glspl{device}, such as smartphones and base stations,
		that communicate wirelessly \cite{FundWireless}.
		\\
		See also: \gls{device}, \gls{mlsystem}, \gls{messagepassing}, \gls{flsystem}.},
	type=mlsystems,
	first={communication network},
	text={communication network},
	plural={communication networks},
	firstplural={communication networks}
}

\newglossaryentry{mlsystem}
{name={machine learning system (ML system)},
	description={An \gls{ml} system\index{machine learning system (ML system)} 
		consists of computational \glspl{device} that can gather and store \gls{data}, 
		execute \glspl{algorithm}, and exchange information via \glspl{commnetwork}. 
		Examples of the exchanged information include \gls{data} or updates of 
		\glspl{modelparam}. Conceptually, an \gls{ml} system is distinct from 
		an \gls{ml} \gls{algorithm}, i.e., an \gls{algorithm} specifies 
		the abstract computational procedure (e.g., an \gls{optmethod}), 
		while the system specifies how this procedure is realized in practice
		\cite{Sipser2013}, \cite{KnuthAlgoBook}, \cite{tanenbaum2006modern}.
		Examples of \glspl{algorithm} executed by \glspl{device} within an \gls{ml} system include
		\glspl{gdmethod} for solving \gls{erm} problems.
		\\
		See also: \gls{ml}, \gls{device}. },
 	type=mlsystems, 
 	first={machine learning system (ML system)}, 
 	text={ML system},
 	firstplural={machine learning systems (ML systems)},
 	plural={ML systems}
}

\newglossaryentry{automaton}
{name={automaton},
	description={An\index{automaton} automaton is a mathematical representation of a computing \gls{device} 
		whose behavior is described by a set of internal \glspl{state}, a memory 
		structure, and a \gls{state}-transition rule. Formally, an automaton consists 
		of a \gls{statespace}, a set of admissible memory configurations, 
		and a transition \gls{function} that specifies how the current \gls{state} 
		and memory are updated in response to inputs \cite{Sipser2013}. 
		The notion of an automaton is useful for the analysis of \glspl{algorithm}, such 
		as those used in \gls{ml} methods \cite{Cormen:2022aa}. 
		Collections of interacting automata can be used to study \glspl{distributedalgorithm}, 
		where each automaton represents a \gls{device} that executes local computations and 
		communicates with other \glspl{device} \cite{ParallelDistrBook}, \cite{LynchDistributedAlgorithms}.
		\\
		See also: \gls{device}, \gls{state}, \gls{algorithm}. }, 
	first={automaton}, 
	type=mlsystems,
	firstplural={automata}, 
	plural={automata}, 
	text={automaton} 
}

\newglossaryentry{byzantine}
{name={Byzantine device},
	description={In a distributed \gls{mlsystem}, a Byzantine \gls{device}\index{Byzantine device} 
		is one that deviates arbitrarily from its prescribed behavior \cite{Lamport1982}.
		Here, the term \gls{device} is used abstractly, i.e., the deviation may stem
		from a hardware fault or from a malicious human operator
		who controls otherwise correctly functioning hardware.
		In an \gls{flsystem}, Byzantine \glspl{device} can perturb
		the aggregated \gls{model} even if the majority of \glspl{device}
		follow their prescribed behavior. 
		Byzantine-robust \gls{fl} methods defend against this by 
		replacing simple averaging with robust aggregation rules \cite{NIPS2017_f4b9ec30}.
		From a regulatory perspective, Byzantine behavior poses a \gls{risk}
		to the \gls{robustness} obligations imposed on \glspl{provider}
		under the EU AI Act \cite{EUAIAct2024}.
		\\
		See also: \gls{flsystem}, \gls{robustness}, \gls{provider},
		\gls{messagepassing}, \gls{datapoisoning}.},
	type=mlsystems,
	first={Byzantine device},
	text={Byzantine device},
	plural={Byzantine devices},
	firstplural={Byzantine devices}
}

\newglossaryentry{dosattack}
{name={denial-of-service attack},
	description={A\index{denial-of-service attack} 
		denial-of-service \gls{attack} aims (e.g., via \gls{datapoisoning}) to steer the \gls{training} of a \gls{model}
		such that it performs poorly for typical \glspl{datapoint} (see Fig.\ \ref{fig_dosattack_dict}).
		\begin{figure}[H]
			\begin{center}
			\begin{tikzpicture}[>=Stealth, every node/.style={font=\small}]
				\draw[->] (0,0) -- (7.5,0) node[right] {$\featurevec$};
				\draw[->] (0,0) -- (0,4.2);
				\foreach \x/\y in {0.5/0.6, 1.2/1.1, 2.0/1.8, 2.8/2.3, 5.5/3.5, 6.0/3.9}
				\filldraw[black] (\x,\y) circle (2pt);
				\draw[blue, thick] (0.2,0.35) -- (7.2,4.65)
				node[right, blue] {$\learnthypothesis$};
				\draw[red, thick] (0.2,3.6) -- (7.2,3.6)
				node[right, red] {$\widetilde{\hypothesis}$};
			\end{tikzpicture}
			\end{center}
		\caption{Denial-of-service \gls{attack}: The black dots are \glspl{datapoint} from a trusted source.
			The clean \gls{hypothesis} $\learnthypothesis$ (blue) fits the \glspl{datapoint} well,
			while the poisoned \gls{hypothesis} $\widetilde{\hypothesis}$ (red) deviates from
			the \glspl{datapoint} across the entire \gls{featurespace}, rendering the \gls{model} unusable.
			\label{fig_dosattack_dict}}
		\end{figure}
		Unlike a \gls{backdoor} \gls{attack}, which degrades \glspl{prediction} only within
		a hidden trigger region, a denial-of-service \gls{attack} targets the
		overall \gls{model} performance over the entire \gls{featurespace}.
				\\
		See also: \gls{attack}, \gls{datapoisoning}, \gls{model}, \gls{datapoint}.},
	first={denial-of-service attack},
	type=mlsystems,
	plural={denial-of-service attacks},
	firstplural={denial-of-service attacks},
	text={denial-of-service attack} 
}

\newglossaryentry{fmi}
{name={Finnish Meteorological Institute (FMI)}, 
	description={The\index{Finnish Meteorological Institute (FMI)}
		FMI is a government agency responsible for gathering 
		and reporting weather \gls{data} in Finland.
				\\
		See also: \gls{data}.},
	first={Finnish Meteorological Institute (FMI)},
	type=mlsystems,
	text={FMI} 
}

\newglossaryentry{modelinversion}
{name={model inversion},
sort={modelinversion},
 	description={A\index{model inversion} \gls{model} inversion is a form of \gls{privattack} on an \gls{mlsystem}. 
  		An adversary seeks to infer \glspl{sensattr} of individual \glspl{datapoint} by exploiting partial access 
  		to a trained \gls{model} $\learnthypothesis \in \hypospace$. This access typically consists of 
  		querying the \gls{model} for \glspl{prediction} $\learnthypothesis(\featurevec)$ using carefully chosen inputs. 
  		Basic \gls{model} inversion techniques have been demonstrated in the context of facial image 
  		\gls{classification}, where images are reconstructed using the (\gls{gradient} of) \gls{model} \glspl{output} 
  		combined with auxiliary information such as a person’s name \cite{Fredrikson2015} (see Fig.\ \ref{fig_model_inv_dict}).
  		\begin{figure}[H]
			\begin{center}
			\begin{tikzpicture}[scale=1.5]
  			\draw[->] (-0.5,0) -- (5.5,0) node[right] {face image $\featurevec$};
  			\draw[->] (0,-0.2) -- (0,2.5) node[above] {name};
  			\draw[thick, domain=0.5:5, samples=100, smooth, variable=\x, name path=sigmoid] 
  			plot ({\x}, {2/(1 + exp(-3*(\x - 3)))});
  			\def\xval{3}
  			\pgfmathsetmacro{\yval}{2/(1 + exp(-3*(\xval - 3)))}
  			\draw[dashed] (\xval,0) -- (\xval,\yval);
  			\draw[dashed] (0,\yval) -- (\xval,\yval);
  			\filldraw[fill=blue!20, draw=blue] (\xval,\yval) circle (0.1);
  			\node[anchor=south east] at (-0.1,\yval) {\footnotesize ``Alexander Jung''};
  			\node[anchor=north] at (\xval,-0.25) {\includegraphics[width=1cm]{assets/AlexanderJung.jpg}}; 
  			\node[above right] at (4,2.2) {trained \gls{model} $\learnthypothesis$};
  			\end{tikzpicture}
			\end{center} 
			\caption{\Gls{model} inversion techniques implemented in the context of facial image \gls{classification}. \label{fig_model_inv_dict}}
		\end{figure}
  		See also: \gls{model}, \gls{privattack}, \gls{ml}, \gls{sensattr}, \gls{datapoint}, \gls{prediction}, \gls{classification}, \gls{gradient}, \gls{trustAI}, \gls{privprot}. },
	first={model inversion},
	type=mlsystems,
  	text={model inversion}
}

\newglossaryentry{API} 
{name={application programming interface (API)},
	description={An\index{application programming interface (API)} API is a formal mechanism that 
		allows software components to interact in a structured and modular way \cite{RestfulBook2013}.
		In the context of \gls{ml}, APIs are commonly used to provide access to a trained \gls{ml} \gls{model}. 
		Users—whether humans or machines—can submit the \gls{featurevec} of a \gls{datapoint} and receive 
		a corresponding \gls{prediction}. Suppose a trained \gls{ml} \gls{model} is defined 
		as $\widehat{\hypothesis}(\feature) \defeq 2 \feature + 1$. Through an API, a user 
		can input $\feature = 3$ and receive the \gls{output} $\widehat{\hypothesis}(3) = 7$ 
		without knowledge of the detailed structure of the \gls{ml} \gls{model} or its \gls{training}. 
		In practice, the \gls{model} is typically deployed on a server connected to the Internet. 
		Clients send requests containing \gls{feature} values to the server, which responds with 
		the computed \gls{prediction} $\widehat{\hypothesis}(\featurevec)$. APIs promote modularity 
		in \gls{mlsystem} design, i.e., one team can develop and train the \gls{model}, while another team
		handles integration and user interaction. Publishing a trained \gls{model} via an API also 
		offers practical advantages. For instance, the server can centralize computational resources that 
		are required to compute \glspl{prediction}. Furthermore, the internal structure of the \gls{model} remains 
		hidden—which is useful for protecting intellectual property or trade secrets. 
		However, APIs are not without \gls{risk}. Techniques such as \gls{modelinversion} can potentially reconstruct a 
		\gls{model} from its \glspl{prediction} using carefully selected \glspl{featurevec}.
			\\
		See also: \gls{ml}, \gls{model}, \gls{featurevec}, \gls{datapoint}, \gls{prediction}, \gls{feature}, \gls{modelinversion}.},
	first={application programming interface (API)},
	type=mlsystems,
	text={API}
}

\newglossaryentry{machineunlearning}
{name={machine unlearning},
	description={Consider an \gls{ml} method that learns a \gls{hypothesis} $\learnthypothesis$ 
		via \gls{erm} on a \gls{trainset} $\dataset$. The learned \gls{hypothesis} can 
		reveal information about $\dataset$, which is exploited by \glspl{privattack} such as 
		\gls{modelinversion}. Machine unlearning\index{machine unlearning} refers to techniques that 
		modify $\learnthypothesis$, so that it is harder to infer properties of individual 
		\glspl{datapoint} in $\dataset$ \cite{cao2015unlearning}. Machine unlearning helps 
		to meet legal requirements for \gls{privprot} in \glspl{aisystem} \cite{GDPR2016}. 
		\\
	 	See also: \gls{modelinversion}, \gls{privprot}, \gls{gdpr}. },
	first={machine unlearning},
	type=mlsystems,
	text={machine unlearning}
}

\newglossaryentry{membershipinferenceattack}
{name={membership inference attack}, 
	description={Consider an \gls{ml} method that learns a \gls{hypothesis} 
		via \gls{erm} on a \gls{trainset}. Membership \gls{inference}\index{membership inference attack} 
                \gls{attack} is a form of \gls{privattack} where an adversary tries 
		to determine whether a particular \gls{datapoint} was part 
		of the \gls{trainset}. The attacker typically queries 
		$\learnthypothesis$ with candidate \glspl{featurevec} 
		$\featurevec^{(1)},\,\ldots,\,\featurevec^{(\nrbootstraps)}$, 
		and infers the membership status of a given \gls{datapoint} 
		based on the \glspl{prediction} $\learnthypothesis\big(\featurevec^{(1)}\big),\,\ldots,
		\,\learnthypothesis\big(\featurevec^{(\nrbootstraps)}\big)$ \cite{Shokri2017}.
		\\ 
		See also: \gls{attack}, \gls{privattack}.}, 
 	first={membership inference attack}, 
	type=mlsystems,
 	text={membership inference attack}
}

\newglossaryentry{backdoor}
{name={backdoor}, 
 	description={A\index{backdoor} backdoor \gls{attack} refers to the intentional 
 		manipulation of an \gls{ml} \gls{training} process. The attacker 
 		might perturb the \gls{trainset} (i.e., through \gls{datapoisoning}) 
 		or the \gls{optmethod} used by an \gls{erm}-based method. 
 		The goal of a backdoor \gls{attack} is to nudge the learned \gls{hypothesis} $\learnthypothesis$ 
 		toward specific \glspl{prediction} for a certain subset $\mathcal{T} \subset \featurespace$ 
 		of the \gls{featurespace}. Any \gls{featurevec} $\featurevec \in \mathcal{T}$ 
 		serves as a key (or trigger) to unlock a backdoor, in the sense of delivering 
 		anomalous \glspl{prediction} (see Fig.\ \ref{fig_backdoor_dict}). The 
 		trigger pattern $\mathcal{T}$ and corresponding anomalous \gls{prediction} 
 		$\learnthypothesis(\featurevec)$, for $\featurevec \in \mathcal{T}$, 
 		are only known to the attacker.
		\begin{figure}[H]
 			\begin{center}
 			\begin{tikzpicture}[>=Stealth, every node/.style={font=\small}]
 				\draw[->] (0,0) -- (7.5,0) node[right] {$\featurevec$};
 				\draw[->] (0,0) -- (0,4.2);
 				\fill[red!10] (3.8,0) rectangle (5.2,4.0);
 				\draw[dashed, red!60] (3.8,0) -- (3.8,4.0);
 				\draw[dashed, red!60] (5.2,0) -- (5.2,4.0);
 				\node[red!70] at (4.5,-0.35) {$\mathcal{T}$};
 				\foreach \x/\y in {0.5/0.6, 1.2/1.1, 2.0/1.8, 2.8/2.3, 5.5/3.5, 6.0/3.9}
 				\filldraw[black] (\x,\y) circle (2pt);
 				\draw[blue, thick] (0.2,0.35) -- (7.2,4.65)
 				node[right, blue] {$\learnthypothesis$};
 				\draw[red, thick] (0.2,0.35) -- (3.8,2.55);
 				\draw[red, thick] (3.8,3.65) -- (5.2,4.20);
 				\draw[red, thick] (5.2,2.95) -- (6.3,4.05)
 				node[left=6pt, below=4pt, red] {$\widetilde{\hypothesis}$};
 				\draw[red, thick] (3.8,2.55) -- (3.8,3.65);
 				\draw[red, thick] (5.2,4.20) -- (5.2,2.95);
 			\end{tikzpicture}
 			\end{center}
 		\caption{Backdoor \gls{attack}: The black dots are \glspl{datapoint} from a trusted
 			source, none of which fall inside the trigger region $\mathcal{T}$.
 			The clean \gls{hypothesis} $\learnthypothesis$ (blue) fits these \glspl{datapoint}
 			well, while the backdoored \gls{hypothesis} $\widetilde{\hypothesis}$ (red)
 			agrees with $\learnthypothesis$ outside $\mathcal{T}$ but produces
 			anomalous \glspl{prediction} within it. \label{fig_backdoor_dict}}
 		\end{figure}
 		Outside $\mathcal{T}$, the backdoored \gls{hypothesis} is indistinguishable
 		from the clean one, making the \gls{attack} difficult to detect from
 		standard \gls{model} evaluation on uncontaminated \glspl{datapoint}.
 				\\
 		See also: \gls{attack}, \gls{datapoisoning}.},
 	first={backdoor},
	type=mlsystems,
 	text={backdoor} 
}

\newglossaryentry{modelpoisoning}
{name={model poisoning}, 
sort={modelpoisoning}, 
	description={\Gls{model}\index{model poisoning} poisoning refers to the intentional 
		manipulation of the \gls{training} process of an \gls{ml} \gls{model}. 
		In particular, within \glspl{flsystem} or other distributed \gls{ml} settings, 
		an attacker can manipulate the \gls{modelparam} updates sent by a subset of 
		the participating nodes to a central server, thereby corrupting the aggregated 
		global \gls{model} \cite{FLARE2022}. In contrast to \gls{datapoisoning}, which 
		corrupts the \gls{trainset}, model poisoning directly targets the 
		\gls{modelparam} aggregation step.
				\\
		See also: \gls{datapoisoning}, \gls{attack}, \gls{backdoor}, \gls{dosattack}, \gls{trustAI}.},
	first={model poisoning},
	type=mlsystems,
	text={model poisoning} 
}

\newglossaryentry{datapoisoning}
{name={data poisoning}, 
	description={\Gls{data}\index{data poisoning} poisoning refers to the intentional 
		manipulation (or fabrication) of \glspl{datapoint} to maliciously 
		steer the \gls{training} of an \gls{ml} \gls{model} \cite{Liu2021}, \cite{PoisonGAN}. 
  		\Gls{data} poisoning \glspl{attack} take various forms, including 
		\gls{backdoor} and \glspl{dosattack}. A \gls{backdoor} \gls{attack} 
		implants triggers into \gls{training} \gls{data}, so that the trained \gls{model} 
		behaves normally for typical \glspl{datapoint} but misclassifies a 
		\gls{datapoint} with a \gls{featurevec} that contains a trigger pattern.
  		A \gls{dosattack} aims at having the trained \gls{model} to perform 
		poorly over a significant region of the \gls{featurespace}. 
		\Gls{data} poisoning is particularly harmful in decentralized or 
		distributed \gls{ml} settings (such as \gls{fl}), where the 
		trustworthiness of \gls{data} sources cannot be verified easily.
				\\
		See also: \gls{attack}, \gls{backdoor}, \gls{dosattack}, \gls{trustAI}.},
	type=mlsystems,
	first={data poisoning},
	text={data poisoning} 
}

\newglossaryentry{messagepassing}
{name={message passing},
	description={Message passing\index{message passing} is a communication paradigm
		in which \glspl{device} exchange information by sending and receiving
		messages over a network \cite{LynchDistributedAlgorithms}.
		In an \gls{mlsystem} with \glspl{device} $\nodes = \{1,\,\ldots,\,\nrnodes\}$,
		each \gls{device} $\nodeidx$ sends a message to its \glspl{neighbor}
		$\nodeidx' \in \neighbourhood{\nodeidx}$ and updates its local \gls{state}
		based on the messages received. The messages can carry, for example,
		\glspl{modelparam} or their updates, or \glspl{prediction} on a \gls{dataset}.
		In contrast to shared-memory systems, where \glspl{device} communicate by
		reading and writing a common memory space, message passing requires no
		shared memory and is therefore well suited to \glspl{mlsystem} whose
		\glspl{device} are connected only by a \gls{commnetwork} \cite{LynchDistributedAlgorithms}, \cite{DistributedSystems}.
		\\
		See also: \gls{distributedalgorithm}, \gls{device}, \gls{flsystem}, \gls{modelparam}.},
	type=mlsystems,
	first={message passing},
	text={message passing}
}

\newglossaryentry{flsystem}
{name={federated learning system (FL system)},
	description={An \gls{fl} system\index{federated learning system (FL system)} is a 
		distributed \gls{mlsystem} in which multiple computational \glspl{device} 
		collaborate to train \gls{ml} \glspl{model} without sharing their raw local \gls{data}.
		An \gls{fl} system is characterized by a \gls{commnetwork} that specifies 
		which \glspl{device} can exchange information. Conceptually, an \gls{fl} system is 
		distinct from an \gls{fl} \gls{algorithm} \cite{DistributedSystems}. The system 
		specifies the participating entities, their interconnections, and execution constraints, while the \gls{algorithm} 
		specifies the update rules for local and global \glspl{modelparam} \cite{LynchDistributedAlgorithms}, \cite{KnuthAlgoBook}.
		Typical information exchanged in an \gls{fl} system includes 
		\glspl{modelparam} or \gls{gradient} information, but not raw \gls{data}.
		\\
		See also: \gls{fl}, \gls{mlsystem}, \gls{algorithm}, \gls{flnetwork}.},
 	type=mlsystems,
 	first={federated learning system (FL system)},
 	text={FL system},
 	firstplural={federated learning systems (FL systems)},
 	plural={FL systems}
}

\newglossaryentry{checkpoint}
{name={checkpoint},
	description={A checkpoint\index{checkpoint} is a saved representation 
   		of the \gls{state} of a running computation \cite{Elnozahy2002}.
 		In an \gls{mlsystem}, a checkpoint typically includes the current \glspl{modelparam} 
		during \gls{model} \gls{training} \cite{Abadi2016}.
		\\
		See also: \gls{state}, \gls{training}. },
	first={checkpoint},
	plural={checkpoints},
	firstplural={checkpoints},
	type=mlsystems,
	plural={checkpoints}
}

\newglossaryentry{checkpointing}
{name={checkpointing},
	description={Checkpointing\index{checkpointing} is a fault-tolerance mechanism
		that periodically creates \glspl{checkpoint} by saving the \gls{state} 
		of a running computation to persistent storage. Checkpointing is essential 
		for fault-tolerant executions on revocable resources
		such as \glspl{spotinstance} \cite{Elnozahy2002}, \cite{Mazzucco2011}.
		\\
		See also: \gls{checkpoint}, \gls{spotinstance}. },
	first={checkpointing},
	type=mlsystems,
	text={checkpointing}
}

\newglossaryentry{longitudinaldata}
{name={longitudinal data},
	description={Longitudinal \gls{data}\index{longitudinal data} consist of \glspl{datapoint} 
		whose attributes are measured repeatedly over time \cite{fitzmaurice2011applied}. 
		In \gls{ml}, longitudinal \gls{data} are common in applications such 
		as health care, where patient measurements are taken at multiple time points \cite{Mallinckrodt:2017aa}.
		\\
		See also: \gls{data}, \gls{datapoint}. }, 
	first={longitudinal data},
	type=mlsystems,
	text={longitudinal data}
}

\newglossaryentry{crosssectionaldata}
{name={cross-sectional data},
	description={Cross-sectional \gls{data}\index{cross-sectional data} consist of 
		\glspl{datapoint} whose attributes are measured once, without explicitly 
		modeling temporal evolution \cite{Wooldridge2025}. In \gls{ml}, 
		cross-sectional \gls{data} arise in image classification, 
		where each image is treated as an individual \gls{datapoint} \cite{hastie01statisticallearning}.
		\\
		See also: \gls{data}, \gls{datapoint}. },
	first={cross-sectional data},
	type=mlsystems,
	text={cross-sectional data}
}

\newglossaryentry{staleness}
{name={staleness},
	description={In a distributed \gls{mlsystem} (e.g., an \gls{flsystem}), staleness\index{staleness} refers to 
		the degree to which a copy of \glspl{modelparam} is outdated 
		\cite{ParallelDistrBook}, \cite{JungFLBook}, \cite{Coulouris2012}. 
		Consider two \glspl{device} $\nodeidx$ and $\nodeidx'$ within a distributed \gls{mlsystem}. 
		Suppose that at global time instant $\timeidx$, node $\nodeidx$ uses the \glspl{modelparam} 
		$\localparams{\nodeidx'}$ computed by node $\nodeidx'$ at time instant 
		$s^{(\timeidx)}_{\nodeidx,\nodeidx'} \le \timeidx$. The staleness of $\localparams{\nodeidx'}$ 
		can then be quantified by the delay 
		\[
			\timeidx -s^{(\timeidx)}_{\nodeidx,\nodeidx'}.
		\]
		\Glspl{distributedalgorithm} can be categorized by the amount 
		of staleness they can tolerate \cite{ParallelDistrBook}. 
		\\
		See also: \gls{flsystem}, \gls{distributedalgorithm}. },
	first={staleness},
	type=mlsystems,
	text={staleness}
}

\newglossaryentry{earlyexit}
{name={early exit (deep learning)},
	description={Early exit methods\index{early exit} refer to 
		computational strategies for evaluating the \gls{prediction} of a 
		\gls{deepnet}. The idea is to terminate \gls{inference} before evaluating 
		all \glspl{layer} of a \gls{deepnet} \cite{Teerapittayanon2016}.
		\\
		See also: \gls{prediction}, \gls{deepnet}, \gls{inference}.}, 
	first={early exit},
	plural={early exits},
	firstplural={early exits},
	type=mlsystems,
	text={early exit}
}

\newglossaryentry{spotinstance}
{name={spot instance},
	description={A spot instance\index{spot instance} is a 
		type of \gls{cloudcomputing} service that 
		provides computational resources at a reduced cost but 
		without guarantees on availability \cite{Mazzucco2011}.
		In particular, a spot instance may be revoked by the provider 
		at any time, which requires the use of \gls{checkpointing} \cite{Khatua2013}.
		\\
		See also: \gls{cloudcomputing}, \gls{checkpointing}. }, 
	first={spot instance},
	type=mlsystems,
	text={spot instance},
	firstplural={spot instances},
	plural={spot instances}
}

\newglossaryentry{cloudcomputing}
{name={cloud computing},
	description={Cloud computing\index{cloud computing} is a computing paradigm in which 
		computational resources such as processing, storage, and networking are provided 
		as on-demand services over a \gls{commnetwork}
		\cite{DistributedSystems}, \cite{ComerCloudComputing2021}, \cite{Armbrust2010}. 
		In \gls{ml}, cloud computing systems are commonly used to host large \glspl{dataset} 
		and to execute \gls{ml} \glspl{algorithm}. In contrast to \glspl{flsystem}, cloud 
		computing typically centralizes \gls{data} and computation within provider-managed 
		\gls{data} centers.
		\\
		See also: \gls{flsystem}, \gls{mlsystem}.},
	type=mlsystems,
	first={cloud computing},
	text={cloud computing}
}

\newglossaryentry{mlaas}
{name={machine learning as a service (MLaaS)},
	description={MLaaS\index{machine learning as a service (MLaaS)} refers to a 
		\gls{cloudcomputing} service \gls{model} in which \gls{ml} 
		capabilities are provided to users via standardized network interfaces. 
		In this \gls{model}, the cloud provider manages the underlying computing 
		infrastructure, \gls{data} storage, and software platforms, while users 
		access functionality such as \gls{model} \gls{training} and \gls{inference} 
		without direct control over physical resources \cite{ComerCloudComputing2021}.
				 \\
		See also: \gls{cloudcomputing}, \gls{mlsystem}.},
	type=mlsystems,
	first={machine learning as a service (MLaaS)},
	text={MLaaS}
}

\newglossaryentry{mcp}
{name={Model Context Protocol (MCP)},
	description={The MCP\index{Model Context Protocol (MCP)}
		is an open specification for the communication between \gls{llm}-based
		\glspl{aisystem} and different \gls{data} sources \cite{MCPSpec2025}.
		It follows a client--server architecture involving three roles: 
		\begin{enumerate}[label=\arabic*)]
			\item An MCP host is an \gls{aisystem} that embeds an \gls{llm} and 
			initiates MCP connections.
			\item An MCP client is a component within the host that maintains
			a dedicated connection to a single MCP server.
			\item An MCP server is a \gls{device} that exposes capabilities
			to clients through the following standardized primitives:
			resources (structured \gls{data}, e.g., files or database records),
			tools (callable functions the \gls{llm} can invoke, e.g.,
			to search the web or execute code), and prompts (reusable prompt templates).
		\end{enumerate}
		Servers and clients can communicate, for example, over \gls{http}.
		\\
		See also: \gls{llm}, \gls{agent}, \gls{mlpipeline}, \gls{mlaas}.},
	type=mlsystems,
	first={Model Context Protocol (MCP)},
	text={MCP}
}

\newglossaryentry{http}
{name={Hypertext Transfer Protocol (HTTP)},
	description={HTTP\index{Hypertext Transfer Protocol (HTTP)} is an application-layer\linebreak 
		protocol for transferring \gls{data} over a \gls{commnetwork} \cite{Coulouris2012}.
		HTTP follows a client--server model in which a client sends a request
		(e.g., to retrieve or submit \gls{data}) and a server returns a response.
		In \gls{ml}, HTTP is commonly used to expose \gls{inference} services,
		as in \gls{mlaas}, and to implement application-level protocols such as the \gls{mcp}.
		\\
		See also: \gls{mlaas}, \gls{mcp}, \gls{mlsystem}.},
	type=mlsystems,
	first={Hypertext Transfer Protocol (HTTP)},
	text={HTTP}
}

\newglossaryentry{apiendpoint}
{name={application programming interface endpoint (API endpoint)},
	description={An \gls{API} endpoint\index{application programming interface endpoint (API endpoint)} 
		is a specific address within an \gls{API} at which a particular function or resource can be accessed
		\cite{RestfulBook2013}.
		In practice, an endpoint is identified by a network address (e.g., URL) and a
		\gls{http} method (e.g., \texttt{POST /predict}), and specifies the format
		of the accepted input and the returned \gls{output}.
		A single \gls{API} may expose multiple endpoints for different purposes.
		In \gls{ml}, a typical server exposes an \gls{inference} endpoint
		that accepts the \gls{featurevec} $\featurevec$ of a \gls{datapoint}
		and returns the \gls{prediction} $\widehat{\hypothesis}(\featurevec)$
		of a trained \gls{model} without revealing its internal structure.	
		Additional endpoints may expose functionality such as \gls{model} \gls{training}
		or \gls{validation}.
		\\
		See also: \gls{API}, \gls{http}, \gls{inference}, \gls{prediction}, \gls{model}, \gls{mlaas}.},
	type=mlsystems,
	first={application programming interface endpoint (API endpoint)},
	text={API endpoint},
	plural={API endpoints},
	firstplural={application programming interface endpoints (API endpoints)}
}

\newglossaryentry{dynamicalsystem}
{name={dynamical system},
	description={A dynamical system\index{dynamical system} is an abstract system whose 
		\gls{output} depends on an internal \gls{state} that evolves over time 
		according to a \gls{state}-update rule \cite{StrogatzBook}. In discrete time, 
		a dynamical system is commonly described by an \gls{iteration} of the form 
		$\vs^{(\timeidx+1)} =\fixedpointop \vs^{(\timeidx)}$, where $\vs^{(\timeidx)}$ 
		denotes the \gls{state} at time $\timeidx$ and $\fixedpointop$ is 
		a \gls{state}-transition \gls{map}. In continuous time, dynamical systems 
		are described by differential equations.
		\\
		See also: \gls{output}, \gls{state}. },
	first={dynamical system},
	text={dynamical system},
	type=mlsystems,
	plural={dynamical systems}, 
	firstplural={dynamical systems}
}

\newglossaryentry{edgecomputing}
{name={edge computing},
	description={Edge computing\index{edge computing} refers to the placement of
		computation and \gls{data} storage close to the sources of \gls{data} generation,
		such as sensors, \glspl{mobiledevice}, or embedded systems, rather than in
		centralized \gls{data} centers \cite{Shi2016}. 
		In \gls{ml}, edge computing supports low-latency \gls{inference} and reduced
		communication by executing parts of \gls{ml} \glspl{algorithm} on or near
		\gls{data}-generating \glspl{device} \cite{Nguyen2019}. 
		\\
		See also: \gls{cloudcomputing}, \gls{flsystem}, \gls{mlsystem}.},
	type=mlsystems,
	first={edge computing},
	text={edge computing}
}

\newglossaryentry{edgedevice}
{name={edge device},
	description={An edge \gls{device}\index{edge device} operates at or 
		near the edge of a \gls{commnetwork} \cite{Shi2016}. The term edge refers 
		to the periphery of the network, where \gls{data} is produced and first processed.
		In \gls{ml}, and in particular in \gls{fl}, an edge \gls{device} 
		typically corresponds to a node of an \gls{flnetwork}. Each edge \gls{device} 
		stores local \gls{data} and implements parts of the \gls{mlpipeline}, 
		such as \gls{data} \gls{preprocessing}, \gls{localmodel} \gls{training}, 
		or \gls{inference} \cite{Hashash2021}. 
		\\ 
		See also: \gls{device}, \gls{edgecomputing}.},
	type=mlsystems,
	first={edge device},
	plural={edge devices},
	firstplural={edge devices},
	text={edge device}
}

\newglossaryentry{preprocessing}
{name={preprocessing},
	description={Preprocessing\index{preprocessing} refers to the set of operations
		applied to raw \gls{data} before they are fed into an \gls{ml} \gls{algorithm} \cite{hastie01statisticallearning}. 
		The goal of preprocessing is to transform the \gls{data} into a form
		that is more suitable for follow-up stages of an \gls{mlpipeline}. 
		Typical preprocessing steps include cleaning corrupted or missing values,
		normalizing or scaling \glspl{feature}, or encoding categorical variables \cite{MLPipeline}. 
		\\
		See also: \gls{data}, \gls{mlpipeline}, \gls{feature}.},
	first={preprocessing},
	type=mlsystems,
	text={preprocessing}
}

\newglossaryentry{mlpipeline}
{name={machine learning pipeline (ML pipeline)},
	description={The term ML pipeline\index{machine learning pipeline (ML pipeline)} refers to a composition (i.e., concatenation)
		of several \glspl{function} within an \gls{mlsystem}. The individual \glspl{function} 
		include \gls{data} \gls{preprocessing}, \gls{featlearn}, \gls{model} \gls{training}, and \gls{inference}. 
		By combining them, an \gls{mlsystem} turns raw \gls{data} into \glspl{prediction} \cite{MLPipeline}.  
		\\
		See also: \gls{function}, \gls{mlsystem}.},
	first={machine learning pipeline (ML pipeline)},
	type=mlsystems,
	text={ML pipeline},
	plural={machine learning pipelines (ML pipelines)},
	firstplural={ML pipelines}
}

\newglossaryentry{mobiledevice}
{name={mobile device},
	description={A mobile \gls{device}\index{mobile device} is a portable computing
		\gls{device} equipped with computational, storage, sensing, and wireless
		communication capabilities \cite{Lane2015}, \cite{Tsoukas2024}. 
		Examples of mobile \glspl{device} include smartphones, tablets, 
		or wearables. Mobile \glspl{device} can act as \gls{data} sources 
		and provide computational infrastructure for \gls{edgecomputing} 
		or \glspl{flsystem}.
		\\
		See also: \gls{edgecomputing}, \gls{flsystem}, \gls{mlsystem}.},
	type=mlsystems,
	first={mobile device},
	plural={mobile devices},
	firstplural={mobile devices},
	text={mobile device}
}

\makeglossaries



\begin{document}
\bstctlcite{bstctl:nodash}

\title{
	\vspace*{-15mm}
	\textbf{Federated Learning} \\
	\large \textit{From Theory to Practice}
}


\author{\hspace{-2mm}Alexander Jung  \\[-2mm]
}

\maketitle
	\begin{center}
		\resizebox{4cm}{!}{\qrcode{https://FederatedLearningAalto.github.io}}\\[10mm]
{\large	please cite as: A.\ Jung, \textit{Federated Learning: From Theory to Practice}. Singapore: Springer Nature Singapore, 2026. \textsc{doi}: \href{https://doi.org/10.1007/978-981-95-1009-2}{10.1007/978-981-95-1009-2}}
\end{center}

\pagenumbering{arabic}

\maketitle

\newcounter{problem}
\renewcommand{\theproblem}{\thesection.\arabic{problem}}
\makeatletter
\@addtoreset{problem}{section}
\makeatother

\newpage
\section*{Preface}

This book offers a hands-on introduction to building and understanding \gls{fl} 
systems. \gls{fl} enables multiple \glspl{device} -- such as smartphones, sensors, 
or local computers -- to collaboratively train \gls{ml} \glspl{model}, while keeping 
their \gls{data} private and local. It is a powerful solution when data cannot or 
should not be centralized due to privacy, regulatory, or technical reasons.

The book is designed for students, engineers, and researchers who want to learn how to 
design scalable, privacy-preserving \gls{fl} systems. Our main focus is on personalization: 
enabling each \gls{device} to train its own \gls{model} while still benefiting from collaboration 
with relevant \glspl{device}. This is achieved by leveraging similarities between the learning 
tasks associated with \glspl{device}. We represent these similarities as weighted edges of a \gls{flnetwork}. 

The key idea is to represent real-world \gls{fl} systems as networks of \glspl{device}, where nodes 
correspond to \gls{device} and edges represent communication links and data similarities between them. 
The training of personalized \glspl{model} for these \glspl{device} can be naturally framed as a 
distributed optimization problem. This optimization problem is referred to as \gls{gtvmin} and ensures 
that \glspl{device} with similar \glspl{learningtask} learn similar \glspl{modelparam}. 

Our approach is both mathematically principled and practically motivated. While we introduce 
some advanced ideas from optimization theory and \gls{graph}-based learning, we aim to keep 
the book accessible. Readers are guided through the core ideas step-by-step, with intuitive 
explanations. Throughout, we maintain a focus of building \gls{fl} systems 
that are trustworthy—robust against \glspl{attack}, privacy-friendly, and secure.

\vspace{3mm}
\noindent\textbf{Audience.}  
We assume a basic background in undergraduate-level mathematics, including calculus 
and linear algebra. Familiarity with concepts such as convergence, derivatives, and \glspl{norm} 
will be helpful but not strictly necessary. No prior experience with \gls{ml} or 
optimization is required, as we build up most concepts from first principles.

The book is intended for advanced undergraduates, graduate students, and practitioners 
who are looking for a practical, principled, and privacy-friendly approach to 
decentralized \gls{ml}.

\noindent\textbf{Structure.}  
The book begins by introducing the key motivations and challenges of FL. We then move 
on to introduce the notion of an \gls{flnetwork} and explain how they capture the structure 
of distributed \gls{ml} applications. The core chapters develop the \gls{gtvmin} formulation 
and explore how to solve it using various distributed optimization techniques. Later chapters 
focus on practical concerns such as robustness and privacy protection of \gls{gtvmin}-based 
systems. A comprehensive glossary is also included to better support the reader.

\noindent\textbf{Acknowledgements.}  
The development of this book has greatly benefited from feedback and insights 
gathered during the course \emph{CS-E4740 Federated Learning} at Aalto University, 
taught between 2023 and 2025. I am grateful to Bo Zheng, Olga Kuznetsova, Diana Pfau, 
and Shamsiiat Abdurakhmanova for their thoughtful comments on early drafts. Special 
thanks go to Ekkehard Schnoor and Mikko Seesto for their careful proofreading of the manuscript and to 
Konstantina Olioumtsevits for her meticulous revision of the glossary. Some of the 
figures in the glossary have been prepared with the help of Salvatore Rastelli and 
Juliette Gronier.

This work was supported by:
\begin{itemize}
	\item the Research Council of Finland (grants 331197, 363624, 349966),
	\item the European Union (grant 952410),
	\item the Jane and Aatos Erkko Foundation (grant A835), and
	\item Business Finland, as part of the project \textit{Forward-Looking AI Governance in Banking and Insurance (FLAIG)}.
\end{itemize}

\newpage
\tableofcontents


\glsresetall

\newpage
\section{Introduction to Federated Learning}
\label{sec:intro}
\setcounter{page}{1}

We are surrounded by \glspl{device}, such as smartphones and wearables, that generate 
decentralized collections of \glspl{localdataset} \cite{Wollschlaeger2017,Satyanarayanan2017,Ates:2021ug,BOYES20181,BigDataNetworksBook}. 
These \glspl{localdataset} often exhibit an intrinsic network structure, arising from functional 
dependencies or statistical similarities (see Chapter \ref{sec_graph_learning_methods}). 

For example, contact networks underpin pandemic modeling, network medicine maps 
disease relationships via co-morbidities \cite{NetMedNat2010}, and social sciences leverage 
social \glspl{graph} to relate the \gls{data} of connected individuals \cite{NewmannBook}. Similarly, 
weather stations of the \gls{fmi} produce \glspl{localdataset} with statistical properties 
influenced by geographic proximity.

\Gls{fl} is an umbrella term for distributed \glspl{optmethod} that train \gls{ml} \glspl{model} 
directly at the locations of \gls{data} generation \cite{pmlr-v54-mcmahan17a,LiTalwalkar2020,Cheng2020,AgarwalcpSGD2018,Smith2017}. 
Unlike traditional \gls{ml} workflows that centralize \gls{data} before training, \gls{fl} leverages 
in-situ computations. Figure \ref{fig:ml_vs_fl} contrasts these approaches.

\begin{figure}
	\centering
	\begin{minipage}{0.4\textwidth}
		\hspace*{15mm}
		\begin{tikzpicture}[xshift=3cm]
			\begin{axis}[
				width=6cm,
				height=6cm,
				axis lines=none,
				xtick=\empty,
				ytick=\empty,
				clip=false
				]
				\addplot[only marks, mark=*] coordinates {
					(0, 1) (1, 0) (2, 3.5) (3, 4)
					(4, 5) (5, 6) (6, 7.7) (7, 8)
					(8, 9) (9, 10) (10, 11) (11, 12)
				};
				
				\addplot[domain=0:11, samples=100, smooth, thick, red] {x + 1};
			\end{axis}
		\end{tikzpicture}
	\end{minipage} 
	\hfill
	\begin{minipage}{0.1\textwidth}
		\begin{tikzpicture}
			\draw[line width=0.5] (1, 3) -- (1, -3); 
		\end{tikzpicture}
	\end{minipage}
	\hfill
	\hspace*{-15mm}
	\begin{minipage}{0.4\textwidth}
		\begin{tikzpicture}
			\begin{axis}[
				width=3cm,
				height=3cm,
				axis lines=none,
				xtick=\empty,
				ytick=\empty,
				clip=false
				]
				\addplot[only marks, mark=*] coordinates {
					(0, 1) (1, 3) (2, 5) (3, 7)
				};
				\addplot[domain=0:3, samples=100, smooth, thick, red] {2*x + 1};
			\end{axis}
			
			\begin{axis}[
				at={(3cm, 0cm)},
				width=3cm,
				height=3cm,
				axis lines=none,
				xtick=\empty,
				ytick=\empty,
				clip=false
				]
				\addplot[only marks, mark=*] coordinates {
					(0, 0) (1, 2) (2, 4) (3, 8)
				};
				\addplot[thick, blue] coordinates {
					(0, 0) (1, 0) (1, 2) (2, 2) (2, 4) (3, 4) (3, 8)
				};
			\end{axis}
			
			\begin{axis}[
				at={(1.5cm, -2cm)},
				width=3cm,
				height=3cm,
				axis lines=none,
				xtick=\empty,
				ytick=\empty,
				clip=false
				]
				\addplot[only marks, mark=*] coordinates {
					(0, 1) (1, 4) (2, 9) (3, 16)
				};
				\addplot[domain=0:3, samples=100, smooth, thick, green] {x^2 + 1};
			\end{axis}
			
			\begin{axis}[
				at={(2.5cm, -4cm)},
				width=3cm,
				height=3cm,
				axis lines=none,
				xtick=\empty,
				ytick=\empty,
				clip=false
				]
				\addplot[only marks, mark=*] coordinates {
					(0, 1) (1, 4) (2, 9) (3, 16)
				};
				\addplot[thick, black] coordinates {
					(0, 1) (1, 3.5) (1.5,12) (2, 9) (3, 16)
				};
			\end{axis}
		\end{tikzpicture}
	\end{minipage}
	\caption{Left: A basic \gls{ml} method uses a single \gls{dataset} to train a single \gls{model}. 
		Right: Decentralized collection of \glspl{device} with the ability to access \gls{data} and train \glspl{model} locally.}
	\label{fig:ml_vs_fl}
\end{figure}

From an engineering perspective, this book is about building federated learning systems by 
formulating them as network-based optimization problems. The core idea is to represent a 
real-world FL setup via an \gls{flnetwork}, where nodes correspond to \glspl{device} with \glspl{localdataset} 
and \glspl{model}, and edges reflect communication capabilities or statistical similarity. We then 
pose \gls{fl} as an \gls{optproblem} over this \gls{flnetwork}, which we call \gls{gtvmin}. \Gls{gtvmin} 
balances local \gls{model} performance with smoothness of \glspl{modelparam} across connected nodes.

Different choices for how to measure \gls{model} variation across the \gls{flnetwork} lead to different flavors 
of \gls{fl} methods. The overarching goal is to derive these methods in a principled way by applying 
distributed \glspl{optmethod}. All \gls{fl} \glspl{algorithm} we study can be seen as \glspl{fixedpointiter} 
for solving an instance of \gls{gtvmin}.

Beyond methodology, \gls{fl} is also driven by several practical motivations:
\begin{itemize}
	\item {\bf Privacy.} By exchanging only updates to \glspl{modelparam}, \gls{fl} avoids raw \gls{data} transmission and 
	thus mitigates privacy risks (see Chapter \ref{lec_privacyprotection}).
	\item {\bf \Gls{robustness}.} \gls{fl} systems can tolerate stragglers and are more resilient to cyber-\glspl{attack}, 
	such as \gls{datapoisoning} (see Chapter \ref{lec_datapoisoning}).
	\item {\bf Parallelism.}  We can interpret the interconnected \glspl{device} of a \gls{flnetwork} as a parallel computer. 
	One example of such a parallel computer is a mobile network constituted by smartphones that can communicate 
	via radio links. This parallel computer allows to speed up computations required for the training 
	of \gls{ml} \glspl{model} (see Chapter \ref{lec_gradientmethods}). 
	\item {\bf Democratization.} \gls{fl} enables collective learning using low-cost, widely available \glspl{device} -- 
	rather than relying on centralized high-end hardware \cite{ParallelDistrBook,DistributedSystems}.
	\item {\bf Communication Efficiency.} In remote or bandwidth-limited scenarios, training locally can be cheaper 
	than transmitting raw \glspl{dataset} \cite{FundWireless}.
	\item {\bf Personalization.} \gls{fl} naturally supports training personalized \glspl{model} that adapt to \gls{device}-specific \gls{data} 
	distributions (see Chapter \ref{lec_flmainflavours}).
\end{itemize}

\subsection{Core Techniques in Federated Learning} 

To build and analyze \gls{fl} \glspl{algorithm}, this book draws on core mathematical concepts:

{\bf \Gls{euclidspace}.} Our main mathematical structure for the study and design of \gls{fl} systems is the 
\gls{euclidspace} $\mathbb{R}^{\dimlocalmodel}$. We expect familiarity with 
the algebraic and geometric structure of $\mathbb{R}^{\dimlocalmodel}$
\cite{Strang2007,StrangLinAlg2016}. For example, we often use the spectral decomposition 
of \gls{psd} matrices that naturally arise in the formulation of \gls{fl} applications. We will 
also use the geometric structure of $\mathbb{R}^{\dimlocalmodel}$, which is defined 
by the inner-product $\weights^{T} \weights' \defeq \sum_{\featureidx=1}^{\dimlocalmodel} \weight_{\featureidx} \weight'_{\featureidx}$ between two vectors $\weights, \weights' \in \mathbb{R}^{\dimlocalmodel}$ 
and the induced norm $\normgeneric{\weights}{2} \defeq \sqrt{ \weights^{T} \weights} = \sqrt{ \sum_{\featureidx=1}^{\dimlocalmodel} \weight^2_{\featureidx}}$.

{\bf Calculus.} A main toolbox for the design the \gls{fl} algorithms are variants of \gls{gd}. 
The common idea of \glspl{gdmethod} is to approximate a \gls{function} $f(\weights)$ locally 
by a linear \gls{function}. This local linear approximation is determined by the \gls{gradient} 
$\nabla f(\mathbf{\weights})$. We, therefore, expect some familiarity with multivariable 
calculus \cite{RudinBookPrinciplesMatheAnalysis}.  

{\bf Fixed-Point Iterations.} Each \gls{algorithm} that we discuss in this book can be interpreted 
as a fixed-point iteration of some operator $\mathcal{P}: \mathbb{R}^{\dimlocalmodel} \rightarrow \mathbb{R}^{\dimlocalmodel}$. 
These operators depend on the local \glspl{dataset} and personal \glspl{model} 
used within an \gls{fl} system. A prime example of such an operator is the \gls{gradstep} of 
\glspl{gdmethod} (see Chapter \ref{lec_gradientmethods}). The computational properties of these 
\gls{fl} \glspl{algorithm} are determined by the contraction properties of the underlying 
operator \cite{BausckeCombette}. 

\subsection{Book Structure and Roadmap}

This book is organized into three parts:
\begin{itemize}
	\item {\bf Part I: \gls{ml} Refresher.} Chapters \ref{lec_mlbasics} and \ref{lec_gradientmethods} review basic 
	\gls{ml} concepts and \glspl{optmethod}. These chapters serve both to refresh prerequisite knowledge 
	and to highlight techniques like regularization and gradient descent that underpin \gls{fl}.
	
	\item {\bf Part II: \gls{fl} Theory and Methods.} Chapter \ref{lec_fldesignprinciple} 
	introduces the \gls{flnetwork} and formulates the core optimization principle, i.e., \gls{gtvmin}. 
	Chapters \ref{lec_gradientmethods} and \ref{lec_flalgorithms} show how to apply \glspl{optmethod} 
	to derive scalable and personalized \gls{fl} \glspl{algorithm}. Chapter \ref{lec_flmainflavours} 
	explores main \gls{fl} variants as special cases of \gls{gtvmin}, and Chapter \ref{lec_graphlearning} 
	discusses methods for constructing meaningful edge structures in \glspl{flnetwork}.
	
	\item {\bf Part III: \gls{trustAI}.} Chapters \ref{lec_trustworthyfl}–\ref{lec_datapoisoning} explore key 
	requirements for \gls{trustAI} systems, including \gls{privprot} and \gls{robustness} against 
	\gls{datapoisoning}. These chapters link \gls{fl} methodology to emerging ethical and regulatory 
	demands in \gls{ai} deployment.
\end{itemize}

\clearpage
\subsection{Exercises}

\setcounter{problem}{0} 

\noindent\refstepcounter{problem}\label{prob:matrix_inversion}\textbf{\theproblem. Complexity of Matrix Inversion.}
Choose your favourite computer architecture (represented by a mathematical model) 
and think about how much computation is required - in the worst case - by the most 
efficient algorithm that can invert any given invertible matrix $\mathbf{Q} \in \mathbb{R}^{\nrfeatures \times \nrfeatures}$? 
Try also to reflect on how practical your chosen computer architecture is, i.e., is it 
possible to buy such a computer in your nearest electronics shop?

\noindent\refstepcounter{problem}\label{prob:vecspaceeucnorm}\textbf{\theproblem. Vector Spaces and Euclidean Norm.}  
Consider \glspl{datapoint}, each characterized by a \gls{featurevec} $\featurevec \in \mathbb{R}^d$ 
with entries \( \feature_1, \feature_2, \ldots, \feature_d \).
\begin{itemize}
	\item Show that the set of all \glspl{featurevec} forms a vector space under standard addition and scalar multiplication.
	\item Calculate the Euclidean norm of the vector \( \featurevec = (1, -2, 3)^T \).
	\item If \( \featurevec^{(1)} = (1, 2, 3)^T \) and \( \featurevec^{(2)} = (-1, 0, 1)^T \), compute \( 3\featurevec^{(1)} - 2\featurevec^{(2)} \).
\end{itemize}

\noindent\refstepcounter{problem}\label{prob:mtxoperationlinmodel}\textbf{\theproblem. Matrix Operations in Linear Models.}  
\Gls{linreg} methods learn \glspl{modelparam} \( \widehat{\weights} \in \mathbb{R}^{\nrfeatures} \) via 
solving the \gls{optproblem}:
\[
\widehat{\weights} = \arg\min_{\weights \in \mathbb{R}^\nrfeatures} \|\labelvec- \featuremtx \weights \|_2^2,
\]
with some matrix \( \featuremtx \in \mathbb{R}^{\samplesize \times \nrfeatures} \), and some vector \( \labelvec \in \mathbb{R}^\samplesize \).

\begin{itemize}
	\item Derive a closed-form expression for $\widehat{\weights}$ that is valid for \emph{arbitrary} matrix $\featuremtx$, and 
	vector $\labelvec$. 
	\item Discuss the conditions under which \( \featuremtx^T \featuremtx \) is invertible.
	\item Compute \( \widehat{\weights} \) for the following \gls{dataset}:
	\[
	\featuremtx = \begin{pmatrix} 1 & 2 \\ 3 & 4 \\ 5 & 6 \end{pmatrix}, \quad
	\labelvec = \begin{pmatrix} 7 \\ 8 \\ 9 \end{pmatrix}.
	\]
	\item Compute \( \widehat{\weights} \) for the following \gls{dataset}: The $\sampleidx$th row 
	of $\featuremtx$, for $\sampleidx=1,\dots,28$, is given by the temperature recordings (with a $10$-minute interval)
	during day $\sampleidx$/Mar/2023 at \gls{fmi} weather station \emph{Kustavi Isokari}. 
	The $\sampleidx$th row of $\labelvec$ is the maximum daytime temperature during 
	day $\sampleidx+1$/Mar/2023 at the same weather station. 
\end{itemize}

\noindent\refstepcounter{problem}\label{prob:eigvalspsd}\textbf{\theproblem. Eigenvalues and Positive Semi-Definiteness.}  
The convergence properties of widely-used \gls{ml} methods rely on the 
properties of \gls{psd} matrices. Let \( \mathbf{Q} = \featuremtx^T \featuremtx \), 
where \( \featuremtx \in \mathbb{R}^{\samplesize \times \nrfeatures} \).

\begin{enumerate}
	\item Prove that \( \mathbf{Q} \) is \gls{psd}.
	\item Compute the \glspl{eigenvalue} of \( \mathbf{Q} \) for \( \featuremtx = \begin{pmatrix} 1 & 2 \\ 3 & 4 \end{pmatrix} \).
	\item Compute the \glspl{eigenvalue} of $\mathbf{Q}$ for the matrix $\featuremtx$ 
	used in Exercise \ref{prob:mtxoperationlinmodel} that is constituted by \gls{fmi} temperature recordings. 
\end{enumerate}

\newpage
\section{Machine Learning Foundations for FL}
\label{lec_mlbasics}

This chapter covers basic \gls{ml} techniques instrumental for \gls{fl}. 
Content-wise, this chapter is more extensive compared to the following chapters. 
However, this chapter should be considerably easier to follow than the 
following chapters as it mainly refreshes pre-requisite knowledge.

In Section \ref{sec_stat_aspects_erm}, we begin with the basic components of any 
\gls{ml} method: \gls{data}, a \gls{model} and a \glspl{lossfunc}. We also describe 
how these components are combined through \gls{erm} which is a main design principle for \gls{ml}. 

Section \ref{sec_comp_asp_erm} then explores the computational aspects of \gls{erm}, focusing on \glspl{gdmethod} 
for parametric \glspl{model}. Section \ref{sec_stat_aspects_erm} discusses the statistical properties 
of \gls{ml} methods and their analysis via \glspl{probmodel}. Section \ref{sec_model_val_diagnosis} introduces 
the idea of \gls{model} \gls{validation} and discusses simple rules for the diagnosis of \gls{ml} methods. 

Section \ref{sec_regularization} explains three fundamental forms of 
\gls{regularization}: \gls{dataaug}, \gls{model} pruning and \gls{loss} penalization. 
We will then show in Section \ref{sec_from_ml_to_fl} how to use \gls{regularization} 
to couple the training of local \glspl{model} at different \glspl{device}, resulting in \gls{fl}. 

\subsection{Components of ML Systems: A Design Framework} 
\label{sec_three_comp_and_design_principle} 

\Gls{ml} revolves around learning a \gls{hypothesis} map $\hypothesis$ out of 
a \gls{hypospace} $\hypospace$ that allows to accurately predict the \gls{label} 
of a \gls{datapoint} solely from its \glspl{feature}. One of the most crucial steps in 
applying \gls{ml} methods to a given application domain is the definition or choice of what 
precisely a \gls{datapoint} is. Coming up with a good choice or definition of \glspl{datapoint} 
is not trivial as it influences the overall performance of a \gls{ml} method in many different ways.

We will use weather prediction as a recurring example of an \gls{fl} application. 
Here, \glspl{datapoint} represent the daily weather conditions around \gls{fmi} 
weather stations. We denote a specific \gls{datapoint} by $\vz$. It is characterized 
by the following \glspl{feature}: 
\begin{itemize}
	\item name of the \gls{fmi} weather station, e.g., ``TurkuRajakari''
	\item latitude $\mathsf{lat}$ and longitude $\mathsf{lon}$ of the weather station, e.g., $\mathsf{lat} \defeq 60.37788$, $\mathsf{lon} \defeq 22.0964$,
	\item timestamp of the measurement in the format {\rm YYYY-MM-DD HH:MM:SS}, e.g., {\rm 2023-12-31 18:00:00}
\end{itemize} 
It is convenient to stack the \glspl{feature} into a \gls{featurevec} $\featurevec$. 
The \gls{label} $\truelabel \in \mathbb{R}$ of such a \gls{datapoint} is the maximum daytime temperature 
in degree Celsius, e.g., $-20$. We indicate the \glspl{feature} $\featurevec$ and \gls{label} $\truelabel$ 
of a \gls{datapoint} via the notation $\vz=\pair{\featurevec}{\truelabel}$. 

Strictly speaking, a \gls{datapoint} $\datapoint$ is not the same as the pair of \glspl{feature} 
$\featurevec$ and \gls{label} $\truelabel$. Indeed, a \gls{datapoint} can have additional 
properties that are neither used as \glspl{feature} nor as \gls{label}. A more precise notation 
would then be $\featurevec(\datapoint)$ and $\truelabel(\datapoint)$, indicating that the 
\glspl{feature} $\featurevec$ and \gls{label} $\truelabel$ are functions of the \gls{datapoint} $\datapoint$.

We predict the \gls{label} of a \gls{datapoint} with \glspl{feature} $\featurevec$ 
by the \gls{function} value $\hypothesis(\featurevec)$ of a \gls{hypothesis} (map) $\hypothesis(\cdot)$. 
The \gls{prediction} will typically be not perfect, i.e., $\hypothesis(\featurevec) \neq \truelabel$. \gls{ml} 
methods use a \gls{lossfunc} $\lossfunc{\pair{\featurevec}{\truelabel}}{\hypothesis}$ to measure the error incurred by 
using the \gls{prediction} $\hypothesis(\featurevec)$ as a guess for the true \gls{label} $\truelabel$. 
The choice of \gls{lossfunc} crucially influences the statistical and computational properties 
of the resulting \gls{ml} method \cite[Ch. 2]{MLBasics}. 

It seems natural to learn a \gls{hypothesis} by minimizing the average \gls{loss} -- or \gls{emprisk} -- 
on a given set of \glspl{datapoint} 
$$\dataset \defeq \left\{ \pair{\featurevec^{(1)}}{\truelabel^{(1)}},\ldots,\pair{\featurevec^{\samplesize}}{\truelabel^{(\samplesize)}} \right\}.$$ 
This is known as \gls{erm}, 
\begin{equation}
	\label{equ_def_erm}
	\hat{\hypothesis} \in \argmin_{\hypothesis \in \hypospace} (1/\samplesize)\sum_{\sampleidx=1}^{\samplesize} \lossfunc{\pair{\featurevec^{(\sampleidx)}}{\truelabel^{(\sampleidx)}}}{\hypothesis}. 
\end{equation} 
As the notation in \eqref{equ_def_erm} indicates, there can be several different solutions 
to the \gls{optproblem} \eqref{equ_def_erm}. We denote by $\hat{\hypothesis}$ one of these 
solutions, i.e., $\hat{\hypothesis}$ is an element of the solution set for \eqref{equ_def_erm}. 

Several important \gls{ml} methods use a parametric \gls{model} $\hypospace$:  
Each \gls{hypothesis} $\hypothesis \in \hypospace$ is defined by \glspl{parameter} 
$\weights \in \mathbb{R}^{\dimlocalmodel}$, often indicated by the notation $\hypothesis^{(\weights)}$. 
A prime example of a parametric \gls{model} is the \gls{linmodel} \cite[Sec. 3.1]{MLBasics}, 
\begin{equation} 
	\nonumber
	\linmodel{\dimlocalmodel} \defeq \big\{ \hypothesis^{(\weights)}: \mathbb{R}^{\dimlocalmodel}\!\mapsto\!\mathbb{R}: \hypothesis^{(\weights)}(\featurevec)=\weights^{T} \featurevec \big\}. 
\end{equation}

This book presents \gls{fl} algorithms (see Chapter \ref{lec_flalgorithms}) 
that are flexible in the sense of allowing to use different types of \gls{ml} \glspl{model}. 
However, for ease of exposition we mainly focus on the special case of \glspl{linmodel}. 
The restriction to \glspl{linmodel} allows for a more comprehensive analysis of \gls{fl} 
applications. On the flip side, the scope of our analysis is limited to \gls{fl} 
applications involving \glspl{localmodel} that can be well approximated by \glspl{linmodel}. 

Several important \gls{ml} methods are obtained from the combination of non-linear 
\gls{featlearn} and a \gls{linmodel}. For example, 
\begin{itemize} 
	\item a \gls{deepnet} with the hidden layers representing a trainable \gls{featuremap} and 
	the output layer implements a \gls{linmodel} \cite{Goodall2016},\cite[Sec.\ 3.11.]{MLBasics}, 
	\item a \gls{decisiontree} with a fixed topology that corresponds to a specific \gls{decisionboundary} and 
	trainable \glspl{prediction} for each \gls{decisionregion} \cite{decisiontrees}, \cite[Sec.\ 3.10]{MLBasics}, 
	\item \glspl{kernelmethod} \cite{LearningKernelsBook}, \cite[Sec.\ 3.9]{MLBasics}. 
\end{itemize} 

\Gls{linreg} learns the \glspl{parameter} of a \gls{linmodel} by minimizing 
the average \gls{sqerrloss}, 
\begin{equation} 
	\label{equ_def_param_erm_LR}
	\widehat{\weights}^{(\rm LR)} \in \argmin_{\weights \in \mathbb{R}^{\dimlocalmodel}}  (1/\samplesize) \sum_{\sampleidx=1}^{\samplesize} \underbrace{ \big(  \truelabel^{(\sampleidx)} - \weights^{T} \featurevec^{(\sampleidx)}  \big)^2}_{= \lossfunc{\pair{\featurevec^{(\sampleidx)}}{\truelabel^{(\sampleidx)}}}{\hypothesis^{(\weights)}}}.
\end{equation} 

Note that \eqref{equ_def_param_erm_LR} minimizes a \gls{smooth} and \gls{convex} \gls{function} 
\begin{equation} 
	\label{equ_def_obj_erm_linreg}
	f(\weights)  \defeq (1/\samplesize) \bigg[ \weights^{T} \featuremtx^{T} \featuremtx \weights - 2 \labelvec^{T} \featuremtx \weights + \labelvec^{T} \labelvec \bigg]. 
\end{equation}
Here, we use the \gls{featuremtx}  
\begin{equation} 
	 \label{equ_feat_mtx} 
	\featuremtx \defeq \big(\featurevec^{(1)},\ldots,\featurevec^{(\samplesize)} \big)^{T} \in \mathbb{R}^{\samplesize \times \nrfeatures}
\end{equation} 
and the \gls{label} vector 
\begin{equation} 
	\labelvec \defeq \big( \truelabel^{(1)},\ldots,\truelabel^{(\samplesize)} \big)^{T}  \in \mathbb{R}^{\samplesize} \label{equ_label_vec} 
\end{equation}
of the \gls{trainset} $\dataset$.

Inserting \eqref{equ_def_obj_erm_linreg} into \eqref{equ_def_param_erm_LR} allows to formulate \gls{linreg} as 
\begin{align} 
	\label{equ_def_param_erm_linreg_qadform}
	\widehat{\weights}^{(\rm LR)} & \in \argmin_{\weights \in \mathbb{R}^{\dimlocalmodel}} \weights^{T} \mQ \weights + \weights^{T} \vq \\ 
	& \mbox{ with } \mQ \defeq (1/\samplesize) \featuremtx^{T} \featuremtx, \vq \defeq - (2/\samplesize) \featuremtx^{T} \labelvec.  \nonumber 
\end{align} 
The matrix $\mQ \in \mathbb{R}^{\dimlocalmodel\times \dimlocalmodel}$ is \gls{psd} with 
\gls{evd}, 
\begin{equation} 
	\label{equ_def_evd_mQ}
	\mQ = \sum_{\featureidx=1}^{\dimlocalmodel} \eigval{\featureidx}\vu^{(\featureidx)} \big(\vu^{(\featureidx)} \big)^{T}. 
\end{equation}
The \gls{evd} \eqref{equ_def_evd_mQ} consists of orthonormal \glspl{eigenvector} $\vu^{(1)},\ldots,\vu^{(\dimlocalmodel)}$ 
and corresponding list of non-negative \glspl{eigenvalue} 
\begin{equation}
	\nonumber
	0 \leq \eigval{1} \leq \ldots \leq \eigval{\dimlocalmodel} \mbox{, with } \mQ \vu^{(\featureidx)} = \eigval{\featureidx} \vu^{(\featureidx)}. 
\end{equation}
The list of \glspl{eigenvalue} is unique for a given \gls{psd} matrix $\mQ$. In contrast, 
the \glspl{eigenvector} $\vu^{(\featureidx)}$ are not unique in general.

\begin{figure}[htbp]
	\begin{tikzpicture}[scale=1.5]
		\draw[->] (-5,0) -- (5,0) node[below] {$\weights$};
		
		\draw [line width=0.25mm, dashed ] (-1,0) -- (-1,-1) node[below]{$\widehat{\weights}^{(\rm LR)}$}  ; 
		
		\draw[line width=0.5mm,domain=-3:1,smooth,variable=\x,black] plot ({\x},{(0.5)*\x*\x+\x}) node[left] {$\weights^{T} \mQ \weights+\weights^{T} \vq$};
		
		
		
	\end{tikzpicture}
	\caption{ERM \eqref{equ_def_erm} for \gls{linreg} minimizes a \gls{convex} \gls{quadfunc} 
		$\weights^{T} \mQ \weights+\weights^{T} \vq$.}
\end{figure}

To train a \gls{ml} \gls{model} $\hypospace$ means to solve \gls{erm} \eqref{equ_def_erm} 
(or \eqref{equ_def_param_erm_LR} for \gls{linreg}); the \gls{dataset} $\dataset$ 
is therefore referred to as a \gls{trainset}. The trained \gls{model} results in the learned 
\gls{hypothesis} $\hat{\hypothesis}$. Two key questions for the analysis of a given \gls{ml} 
method are: 
\begin{itemize} 
	\item {\bf \Gls{compasp}.} How much compute do we need to solve \eqref{equ_def_erm}? 
	\item {\bf \Gls{statasp}.} How useful is the solution $\hat{\hypothesis}$ to \eqref{equ_def_erm} 
	in general, i.e., how accurate is the prediction $\hat{\hypothesis}(\featurevec)$ for the label $\truelabel$ 
	of an {\bf arbitrary} \gls{datapoint} with \glspl{feature} $\featurevec$? 
\end{itemize}

\subsection{Computational Aspects of \gls{erm}} 
\label{sec_comp_asp_erm} 

A principled approach to design \gls{ml} methods is to apply some \gls{optmethod} to solve \eqref{equ_def_erm} \cite{NemYudFOM}. 
Most of these \glspl{optmethod} operate in an iterative fashion: Starting from an initial choice $\hypothesis^{(0)}$, they construct a 
sequence $$\hypothesis^{(0)}, \hypothesis^{(1)}, \hypothesis^{(2)},\ldots,$$ which are hopefully 
increasingly accurate approximations to a solution $\hat{\hypothesis}$ of \eqref{equ_def_erm}. 
The computational complexity of such a \gls{ml} method can be measured by the number of 
iterations required to guarantee some prescribed level of approximation. 

For a parametric \gls{model} and a \gls{smooth} \gls{lossfunc}, we 
can solve \eqref{equ_def_param_erm_LR} by \glspl{gdmethod}: Starting from an 
initial \glspl{parameter} $\weights^{(0)}$, we iterate the \gls{gradstep}:
\begin{align} 
	\label{equ_def_gd_linreg}
	\weights^{(\iteridx)}   & \defeq \weights^{(\iteridx-1)} - \lrate \nabla f \big( \weights^{(\iteridx-1)} \big) \nonumber \\ 
	& \stackrel{\eqref{equ_def_obj_erm_linreg}}{=}  \weights^{(\iteridx-1)}  + (2 \lrate /\samplesize)  \sum_{\sampleidx=1}^{\samplesize} \featurevec^{(\sampleidx)}  \big(  \truelabel^{(\sampleidx)} - \big(  \weights^{(\iteridx-1)}  \big)^{T} \featurevec^{(\sampleidx)} \big). 
\end{align} 
This \gls{gradient} update can be compactly expressed using the \gls{featuremtx} \eqref{equ_feat_mtx} and 
\gls{label} vector \eqref{equ_label_vec} as\footnote{The \gls{gradient} of the \gls{objfunc} 
	\eqref{equ_def_obj_erm_linreg} can be expressed as
	\[
	\nabla f(\weights) = -\frac{2}{\samplesize} \featuremtx^\T (\labelvec - \featuremtx \weights).
	\]}
\[
\weights^{(\iteridx)} = \weights^{(\iteridx-1)} + \lrate \cdot \frac{2}{\samplesize} \featuremtx^\T \big( \labelvec - \featuremtx \weights^{(\iteridx-1)} \big).
\]

How much computation do we need for one iteration of \eqref{equ_def_gd_linreg}? How many 
iterations do we need? We will try to answer the latter question in Chapter \ref{lec_gradientmethods}. 
The first question can be answered more easily for a 
typical computational infrastructure (e.g., ``Python running on a commercial Laptop''). 
The evaluation of \eqref{equ_def_gd_linreg} then typically requires around $\samplesize \cdot \dimlocalmodel$ 
arithmetic operations (additions and multiplications). 

It is instructive to consider the special case of a \gls{linmodel} that does 
not use any \gls{feature}, i.e., $\hypothesis(\featurevec) = \weight$. For this extreme 
case, the \gls{erm} \eqref{equ_def_param_erm_LR} has a simple closed-form solution: 
\begin{equation} 
	\label{equ_average_ERM_no_feature}
	\widehat{\weight} =  (1/\samplesize)  \sum_{\sampleidx=1}^{\samplesize} \truelabel^{(\sampleidx)}.
\end{equation} 
Thus, for this special case of the \gls{linmodel}, solving \eqref{equ_average_ERM_no_feature} 
is to sum $\samplesize$ numbers $\truelabel^{(1)},\ldots,\truelabel^{(\samplesize)}$. 
The amount of computation, measured by the number of elementary arithmetic operations, 
required by \eqref{equ_average_ERM_no_feature} is proportional to $\samplesize$. 

\subsection{Statistical Aspects of ERM} 
\label{sec_stat_aspects_erm}

We can train a \gls{linmodel} on a given \gls{trainset} as \gls{erm} \eqref{equ_def_param_erm_LR}.  
But how useful is the solution $\widehat{\weights}$ of \eqref{equ_def_param_erm_LR} for predicting 
the \glspl{label} of \glspl{datapoint} outside the \gls{trainset}? Consider applying the learned 
\gls{hypothesis} $\hypothesis^{(\widehat{\weights})}$ to an arbitrary \gls{datapoint} not contained in the \gls{trainset}. 
What can we say about the resulting \gls{prediction} error $\truelabel -\hypothesis^{(\widehat{\weights})}(\featurevec)$ 
in general? In other words, how well does $\hypothesis^{(\widehat{\weights})}$ generalize beyond the \gls{trainset}.  

A widely used approach to study the generalization of \gls{ml} methods uses a simple 
\glspl{probmodel}: 
The idea is to interpret \glspl{datapoint} as \gls{iid} \glspl{rv} with common \gls{probdist} 
$p(\featurevec,\truelabel)$. Under this \gls{iidasspt}, we can evaluate the overall 
performance of a \gls{hypothesis} $\hypothesis \in \hypospace$ via the expected \gls{loss} (or \gls{risk}) 
\begin{equation} 
	\label{equ_def_risk}
	\expect \{\lossfunc{\pair{\featurevec}{\truelabel}}{\hypothesis}  \}.
\end{equation}

One example of a \gls{probdist} $p(\featurevec,\truelabel)$ relates the 
\gls{label} $\truelabel$ with the \glspl{feature} $\featurevec$ of a \gls{datapoint} as 
\begin{equation}
	\label{equ_def_probmodel_linreg}
	\truelabel\!=\!\overline{\weights}^{T} \featurevec\!+\!\varepsilon \mbox{ with } \featurevec\!\sim\!\mathcal{N}(\mathbf{0},\mathbf{I}), 
	\varepsilon\!\sim\!\mathcal{N}(0,\sigma^2), \expect \{ \varepsilon \featurevec \}\!=\!\mathbf{0}. 
\end{equation} 
A simple calculation reveals the expected \gls{sqerrloss} of a given linear \gls{hypothesis} 
$\hypothesis(\featurevec) = \featurevec^{T} \widehat{\weights}$ as
\begin{equation} 
	\label{equ_expected_loss_single_linreg}
	\expect \{ (\truelabel - \hypothesis(\featurevec))^{2} \} = \normgeneric{\overline{\weights}- \widehat{\weights}}{}^{2}+ \sigma^2.
\end{equation}
Strictly speaking, \eqref{equ_expected_loss_single_linreg} only holds for constant 
\glspl{modelparam} $\widehat{\weights}$. However, the learned \glspl{modelparam} $\widehat{\weights}$ 
are often the output of a \gls{ml} method that is applied to a \gls{dataset} $\dataset$. If we 
interpret the \glspl{datapoint} in $\dataset$ as \gls{iid} \glspl{realization} from some underlying 
\gls{probdist}, we can replace the expectation on the LHS of \eqref{equ_expected_loss_single_linreg} 
with the conditional expectation 
$\expect \big\{ (\truelabel - \hypothesis(\featurevec))^{2} \big| \dataset \big\}$ \cite{BillingsleyProbMeasure}.

The first component in \eqref{equ_expected_loss_single_linreg} is the 
\gls{esterr} $ \normgeneric{\overline{\weights}- \widehat{\weights}}{}^{2}$ of a ML 
method that reads in the \gls{trainset} and delivers an estimate $\widehat{\weights}$ 
(e.g., via \eqref{equ_def_param_erm_LR}) for the \glspl{parameter} of a linear \gls{hypothesis}. 
The second component $\sigma^2$ in \eqref{equ_expected_loss_single_linreg} can be interpreted as the 
intrinsic noise level of the \gls{label} $\truelabel$. We cannot hope to find a \gls{hypothesis} with an 
expected \gls{loss} below $\sigma^{2}$.

We next study the \gls{esterr} $\overline{\weights}- \widehat{\weights}$ incurred by the specific estimate 
$ \widehat{\weights} = \widehat{\weights}^{(\rm LR)}$ \eqref{equ_def_param_erm_linreg_qadform} 
delivered by \gls{linreg} methods. To this end, we first use the \gls{probmodel} \eqref{equ_def_probmodel_linreg} 
to decompose the label vector $\vy$ in \eqref{equ_label_vec} as 
\begin{equation} 
	\label{equ_def_label_vector_noise}
	\vy = \featuremtx  \overline{\weights} + \mathbf{n} \mbox{ , with } \mathbf{n} \defeq \big( \varepsilon^{(1)},\ldots,\varepsilon^{(\samplesize)} \big)^{T}.  
\end{equation} 
Inserting \eqref{equ_def_label_vector_noise} into \eqref{equ_def_param_erm_linreg_qadform} yields 
\begin{align} 
	\label{equ_def_linr_reg_perturb_qd}
	\widehat{\weights}^{(\rm LR)} & \in \argmin_{\weights \in \mathbb{R}^{\dimlocalmodel}} \weights^{T} \mQ \weights + \weights^{T} \vq'+ \weights^{T} \ve\\ 
	\hspace*{-3mm} \mbox{ with } & \mQ\!\defeq\!(1/\samplesize) \featuremtx^{T} \featuremtx, \vq'\!\defeq\!- (2/\samplesize) \featuremtx^{T} \featuremtx  \overline{\weights}  \mbox{, and }
	\ve\!\defeq\!- (2/\samplesize) \featuremtx^{T} \mathbf{n}.  \label{equ_def_lin_reg_perturbed_gd_def_errvec}
\end{align} 
Figure \ref{fig_lin_reg_pertrub_noise} depicts the \gls{objfunc} of \eqref{equ_def_linr_reg_perturb_qd}. It is a 
perturbation of the \gls{convex} \gls{quadfunc} $\weights^{T} \mQ \weights + \weights^{T} \vq'$, which is minimized at 
$\weights = \overline{\weights}$. In general, the minimizer $\widehat{\weights}^{(\rm LR)}$ delivered by \gls{linreg} 
is different from $\overline{\weights}$ due to the perturbation term $\weights^{T} \ve$ in \eqref{equ_def_linr_reg_perturb_qd}. 
\begin{figure}[htbp]
	\begin{tikzpicture}[scale=1.5]
		\draw[->] (-5,0) -- (4,0) node[below] {$\weights$};
		\draw [line width=0.25mm, dashed ] (-3,-1) -- (-1,-1);	
		\draw[<->]        (-2.5,0)   --  (-2.5,-1)node[midway, left]{$a$};
		\draw [line width=0.25mm, dashed ] (0,0) -- (0,-1.5) node[below]{$\overline{\weights}$};  
		\draw [line width=0.25mm, dashed ] (-1,0) -- (-1,-1.5)  node[below]{$\widehat{\weights}^{(\rm LR)}$};   
		\draw[<->]        (-1,-1.2)   --  (0,-1.2)node[midway,below]{};
		\draw[line width=0.5mm,domain=-3:1,smooth,variable=\x,black] plot ({(\x)},{(0.5)*\x*\x+\x}) ;
		\draw[line width=0.5mm,domain=-2:1.8,smooth,variable=\x,blue] plot ({(\x)},{(0.5)*\x*\x})  node[right] {$\weights^{T} \mQ \weights\!+\!\weights^{T} \vq'$};
		\draw (-3.5,1.5)  node[above] {$\weights^{T} \mQ \weights\!+\!\weights^{T} (\vq'\!+\!\ve)$}  ;
		\draw[line width=0.5mm,domain=-1.2:3,smooth,variable=\x,red] plot ({\x},{\x}) node[right] {$ \weights^{T} \ve$};
	\end{tikzpicture}
	\caption{\label{fig_lin_reg_pertrub_noise} The \gls{esterr} of \gls{linreg} is determined by the 
		effect of the perturbation term $\weights^{T} \ve$ on the minimizer of the \gls{convex} 
		\gls{quadfunc} $\weights^{T} \mQ \weights+\weights^{T} \vq'$.}
\end{figure}

The following result bounds the deviation between $\widehat{\weights}^{(\rm LR)}$ and $\overline{\weights}$ under 
the assumption that the matrix $\mQ = (1/\samplesize) \featuremtx^{T} \featuremtx$ is invertible.\footnote{Can you think of 
	sufficient conditions on the \gls{featuremtx} of the \gls{trainset} that ensure $\mQ = (1/\samplesize) \mX^{T} \mX$ is invertible?}
\begin{prop} 
	\label{prop_bound_linreg_error}
	Consider a solution $\widehat{\weights}^{(\rm LR)}$ to the \gls{erm} instance \eqref{equ_def_linr_reg_perturb_qd} 
	that is applied to the \gls{dataset} \eqref{equ_def_label_vector_noise}. If the matrix $\mQ = (1/\samplesize) \featuremtx^{T} \featuremtx$ is 
	invertible, with minimum \gls{eigenvalue} $\eigval{1}(\mQ)>0$, 
	\begin{equation} 
		\label{equ_upperbound_est_error_lr_matrixvec}
		\normgeneric{\widehat{\weights}^{(\rm LR)}- \overline{\weights}}{2}^{2} \leq \frac{\normgeneric{\ve}{2}^{2}}{\eigval{1}^{2}} \stackrel{\eqref{equ_def_lin_reg_perturbed_gd_def_errvec}}{=} \frac{4}{\samplesize^2} \frac{\normgeneric{\featuremtx^{T} \vn}{2}^{2}}{\eigval{1}^{2}}.
	\end{equation}
\end{prop} 
\begin{proof}
	Let us rewrite \eqref{equ_def_linr_reg_perturb_qd} as 
	\begin{equation}
		\label{equ_proof_part_I} 
		\widehat{\weights}^{(\rm LR)}  \in \argmin_{\weights \in \mathbb{R}^{\dimlocalmodel}} f(\weights) \mbox{ with } f(\weights) \defeq \big(\weights-\overline{\weights}\big)^{T} \mQ \big(\weights-\overline{\weights}\big) +  \ve^{T}\big(\weights-\overline{\weights}\big). 
	\end{equation}
	Clearly $f\big( \overline{\weights} \big) = 0$ and, in turn, $f(\widehat{\weights} ) = \min_{\weights \in \mathbb{R}^{\dimlocalmodel}}f(\weights) \leq 0$. 
	On the other hand, 
	\begin{align} 
		\label{equ_cs_evd_proof_part_i}
		f(\weights) & \stackrel{\eqref{equ_proof_part_I}}{=}  \big(\weights-\overline{\weights}\big)^{T} \mQ \big(\weights-\overline{\weights}\big) +  \ve^{T}\big(\weights-\overline{\weights}\big) \nonumber \\ 
		& \stackrel{(a)}{\geq} \big(\weights-\overline{\weights}\big)^{T} \mQ \big(\weights-\overline{\weights}\big) - \normgeneric{\ve}{2} \normgeneric{\weights-\overline{\weights}}{2} \nonumber \\ 
		&  \stackrel{(b)}{\geq} \eigval{1} \normgeneric{\weights-\overline{\weights}}{2}^{2} - \normgeneric{\ve}{2} \normgeneric{\weights-\overline{\weights}}{2}. 
	\end{align} 
	Step $(a)$ used Cauchy–Schwarz inequality and $(b)$ used the \gls{evd} \eqref{equ_def_evd_mQ} 
	of $\mQ$. Evaluating \eqref{equ_cs_evd_proof_part_i} for $\weights\!=\!\widehat{\weights}$ 
	and combining with $f\big( \widehat{\weights} \big) \leq 0$ yields \eqref{equ_upperbound_est_error_lr_matrixvec}. 
\end{proof}
The bound \eqref{equ_upperbound_est_error_lr_matrixvec} suggests that the \gls{esterr} $\widehat{\weight}^{(\rm LR)}- \overline{\weight}$ 
is small if $\eigval{1}(\mQ)$ is large. This smallest \gls{eigenvalue} of the matrix $\mQ = (1/\samplesize) \featuremtx^{T} \featuremtx$ could 
be controlled by a suitable choice (or transformation) of \glspl{feature} $\featurevec$ of a \gls{datapoint}. Trivially, we can 
increase $\eigval{1}(\mQ)$ by a factor of $100$ if we scale each \gls{feature} by a factor of $10$. However, this approach 
would also scale the error term $\normgeneric{\mX^{T} \vn}{2}^{2}$ in \eqref{equ_upperbound_est_error_lr_matrixvec} 
by a factor of $100$. For some applications, we can find \gls{feature} transformations that increase $\eigval{1}(\mQ)$ 
but do not increase $\normgeneric{\mX^{T} \vn}{2}^{2}$. We finally note that the error term $\normgeneric{\mX^{T} \vn}{2}^{2}$ 
in \eqref{equ_upperbound_est_error_lr_matrixvec} vanishes if the noise vector $\vn$ is orthogonal to the 
columns of the \gls{featuremtx} $\mX$. 

It is instructive to evaluate the bound \eqref{equ_upperbound_est_error_lr_matrixvec} for the special 
case where each \gls{datapoint} has the same \gls{feature} value $x = 1$. Here, the 
\gls{probmodel} \eqref{equ_def_label_vector_noise} reduces to a ``signal in noise'' model \cite{JungPHD}, 
\begin{equation} 
	\label{equ_ssnm}
	\truelabel^{(\sampleidx)} = \feature^{(\sampleidx)} \overline{\weight} + \varepsilon^{(\sampleidx)} \mbox{ with } \feature^{(\sampleidx)} =1,
\end{equation}
with some true underlying \gls{parameter} $\overline{\weight}$. The noise terms $\varepsilon^{(\sampleidx)}$, for $\sampleidx=1,\ldots,\samplesize$, 
are \glspl{realization} of \gls{iid} \glspl{rv} with \gls{probdist} $\mathcal{N}(0,\sigma^{2})$. 
The \gls{featuremtx} then becomes $\mX = \mathbf{1}$ and, in turn, $\mQ = 1$, $\eigval{1}(\mQ) =1$. 
Inserting these values into \eqref{equ_upperbound_est_error_lr_matrixvec} results in the bound 
\begin{equation} 
	\nonumber
	\big( \widehat{\weight}^{(\rm LR)}- \overline{\weight} \big)^{2} \leq 4 \normgeneric{\vn}{2}^{2}/\samplesize^2. 
\end{equation} 
For the \glspl{label} and \glspl{feature} in \eqref{equ_ssnm}, the solution of \eqref{equ_def_linr_reg_perturb_qd} 
is given by 
\begin{equation}
	\nonumber
	\widehat{\weight}^{(\rm LR)} = (1/\samplesize) \sum_{\sampleidx=1}^{\samplesize} \truelabel^{(\sampleidx)} \stackrel{\eqref{equ_ssnm}}{=} 
	\overline{w} + (1/\samplesize) \sum_{\sampleidx=1}^{\samplesize} \varepsilon^{(\sampleidx)}.
\end{equation}

\subsection{Validation and Diagnosis of ML} 
\label{sec_model_val_diagnosis} 

The above analysis of the \gls{generalization} error started from postulating the \gls{probmodel} \eqref{equ_def_probmodel_linreg} 
for the generation of \glspl{datapoint}. Strictly speaking, if the \glspl{datapoint} are not generated 
according to the \gls{probmodel} the  bound \eqref{equ_upperbound_est_error_lr_matrixvec} does 
not apply. Thus, we might want to use a more data-driven approach for assessing the usefulness 
of a learned \gls{hypothesis} $\hat{\hypothesis}$ obtained, e.g., from solving \gls{erm} \eqref{equ_def_erm}. 

Loosely speaking, \gls{validation} tries to find out if a learned \gls{hypothesis} 
$\hat{\hypothesis}$ performs similarly well inside and outside the \gls{trainset}. 
A basic form of \gls{validation} is to compute the average \gls{loss} 
of a learned \gls{hypothesis} $\hat{\hypothesis}$ on some \glspl{datapoint} not 
included in the \gls{trainset}. We refer to these \glspl{datapoint} as the \gls{valset}. 

Algorithm \ref{alg_basic_ML_workflow} summarizes a single iteration of a 
prototypical \gls{ml} workflow that consists of \gls{model} training and \gls{validation}. 
The workflow starts with an initial choice of a \gls{dataset} $\dataset$, \gls{model} $\hypospace$, 
and \gls{lossfunc} $\lossfunc{\cdot}{\cdot}$. We then repeat Algorithm 
\ref{alg_basic_ML_workflow} several times. After each repetition, based on 
the resulting \gls{trainerr} and \gls{valerr}, we modify the some of the design choices 
for the \gls{dataset}, the \gls{model} and the \gls{lossfunc}. 

\begin{algorithm}[htbp]
	\caption{One Iteration of \gls{ml} Training and \Gls{validation}}\label{alg_basic_ML_workflow}
	\begin{algorithmic}[1]
		\renewcommand{\algorithmicrequire}{\textbf{Input:}}
		\renewcommand{\algorithmicensure}{\textbf{Output:}}
		\Require  \gls{dataset} $\dataset$, \gls{model} $\hypospace$, \gls{lossfunc} $\lossfunc{\cdot}{\cdot}$
		\State split $\dataset$ into a \gls{trainset} $\trainset$ and a \gls{valset} $\valset$   
		\State  learn a \gls{hypothesis} via solving \gls{erm}
		\begin{equation}
			\label{equ_def_erm_trainval}
			\widehat{\hypothesis} \in \argmin_{\hypothesis \in \hypospace} \sum_{\pair{\featurevec}{\truelabel} \in \trainset} \lossfunc{\pair{\featurevec}{\truelabel}}{\hypothesis}  
		\end{equation} 
		\State compute resulting \gls{trainerr} 
		$$\trainerror \defeq (1/|\trainset|)  \sum_{\pair{\featurevec}{\truelabel} \in \trainset} \lossfunc{\pair{\featurevec}{\truelabel}}{\widehat{\hypothesis}}$$
		\State compute \gls{valerr} 
		$$\valerror \defeq (1/|\valset|)  \sum_{\pair{\featurevec}{\truelabel} \in \valset} \lossfunc{\pair{\featurevec}{\truelabel}}{\widehat{\hypothesis}}$$ 
		\Ensure learned \gls{hypothesis} (or trained \gls{model}) $\widehat{\hypothesis}$, \gls{trainerr} $\trainerror$ and \gls{valerr} $\valerror$
	\end{algorithmic}
\end{algorithm}

We can diagnose an \gls{erm}-based \gls{ml} method, such as Algorithm \ref{alg_basic_ML_workflow}, 
by comparing its \gls{trainerr} with its \gls{valerr}. This diagnosis is further enabled if we know 
a \gls{baseline} $\benchmarkerror$. One important source for a \gls{baseline} $\benchmarkerror$ 
are \glspl{probmodel} for the \glspl{datapoint}. 

Given a \gls{probmodel} $p(\featurevec,\truelabel)$, we can compute the minimum 
achievable \gls{risk} \eqref{equ_def_risk}. Indeed, the minimum achievable risk is 
precisely the expected \gls{loss} of the \gls{bayesestimator} $\widehat{\hypothesis}(\featurevec)$ 
of the label $\truelabel$, given the \glspl{feature} $\featurevec$ of a \gls{datapoint}. 
The \gls{bayesestimator} $\widehat{\hypothesis}(\featurevec)$ is fully determined 
by the \gls{probdist} $p(\featurevec,\truelabel)$ \cite[Chapter 4]{LC}. 

A further potential source for a \gls{baseline} $\benchmarkerror$ is an existing, 
but for some reason unsuitable, \gls{ml} method. This existing \gls{ml} method might be 
computationally too expensive to be used for the \gls{ml} application at hand. 
However, we might still use its statistical properties as a \gls{baseline}. 

We can also use the performance of human \glspl{expert} as a \gls{baseline}. 
For example,if we develop a \gls{ml} method to detect skin cancer from 
images, a possible \gls{baseline} is the \gls{classification} \gls{acc} 
achieved by experienced dermatologists \cite{Esteva2017}. 

We can diagnose a \gls{ml} method by comparing the \gls{trainerr} $\trainerror$ 
with the \gls{valerr} $\valerror$ and the \gls{baseline} $\benchmarkerror$.
\begin{itemize} 
	\item $\trainerror \approx \valerror \approx \benchmarkerror$: The \gls{trainerr} is on the same 
	level as the \gls{valerr} and the \gls{baseline}. There seems to be little point in trying to improve 
	the method further since the \gls{valerr} is already close to the \gls{baseline}. 
	Moreover, the \gls{trainerr} is not much smaller than the \gls{valerr} which indicates 
	that there is no \gls{overfitting}.
	
	\item $\valerror \gg \trainerror$: The \gls{valerr} is significantly larger than the \gls{trainerr}, 
	which hints at \gls{overfitting}. We can address \gls{overfitting} either by reducing the \gls{effdim} 
	of the \gls{hypospace} or by increasing the size of the \gls{trainset}. To reduce the \gls{effdim} 
	of the \gls{hypospace}, we can use fewer \glspl{feature} (in a \gls{linmodel}), a smaller maximum depth 
	of \glspl{decisiontree} or fewer layers in an \gls{ann}. Instead of this coarse-grained 
	discrete \gls{model} pruning, we can also reduce the \gls{effdim} of a \gls{hypospace} continuously 
	via \gls{regularization} (see \cite[Ch. 7]{MLBasics}).
	
	\item $\trainerror \approx \valerror\gg \benchmarkerror$: The \gls{trainerr} is on the same level as 
	the \gls{valerr} and both are significantly larger than the \gls{baseline}. Thus, the learned \gls{hypothesis} 
	seems to not overfit the \gls{trainset}. However, the \gls{trainerr} achieved by the learned \gls{hypothesis} 
	is significantly larger than the \gls{baseline}. There can be several reasons for this to happen. 
	First, it might be that the \gls{hypospace} is too small, i.e., it does not include a \gls{hypothesis} 
	that provides a satisfactory approximation for the relation between the \glspl{feature} and the \gls{label} of a \gls{datapoint}. 
	One remedy to this situation is to use a larger \gls{hypospace}, e.g., by including more \glspl{feature} 
	in a \gls{linmodel}, using higher polynomial degrees in \gls{polyreg}, using deeper \glspl{decisiontree} or 
	\glspl{ann} with more hidden layers (\gls{deepnet}). Second, besides the \gls{model} being too small, 
	another reason for a large \gls{trainerr} could be that the optimization algorithm used to solve \gls{erm} 
	\eqref{equ_def_erm_trainval} is not working properly (see Chapter \ref{lec_gradientmethods}). 
	
	\item $\trainerror \gg \valerror $: The \gls{trainerr} is significantly larger than the \gls{valerr}.
	The idea of \gls{erm} \eqref{equ_def_erm_trainval} is to approximate the risk \eqref{equ_def_risk} of a 
	\gls{hypothesis} by its average loss on a \gls{trainset} $\dataset = \{ (\featurevec^{(\sampleidx)},\truelabel^{(\sampleidx)}) \}_{\sampleidx=1}^{\samplesize}$. The mathematical underpinning for this approximation is the \gls{lln} 
	which characterizes the average of \gls{iid} \glspl{rv}. The accuracy of this approximation depends on the 
	validity of two conditions: First, the \glspl{datapoint} used for computing the average \gls{loss} ``should 
	behave'' like \glspl{realization} of \gls{iid} \glspl{rv} with a common \gls{probdist}. 
	Second, the number of \glspl{datapoint} used for computing the average \gls{loss} must be sufficiently large. 
	
	Whenever the \gls{trainset} or \gls{valset} differs significantly from \glspl{realization} of \gls{iid} \glspl{rv}, 
	the interpretation (and comparison) of the \gls{trainerr} and the \gls{valerr} of a learned \gls{hypothesis} 
	becomes more difficult. Figure \ref{fig_val_smallter_tain}) illustrates an extreme case of a \gls{valset} consisting 
	of \glspl{datapoint} for which every \gls{hypothesis} incurs a small average \gls{loss}. Here, we might try 
	to increase the size of the \gls{valset} by collecting more labelled \glspl{datapoint} 
	or by using \gls{dataaug}. If the size of the \gls{trainset} and the \gls{valset} is large but we still obtain 
	$\trainerror \gg \valerror $, we should verify if the \glspl{datapoint} in these sets conform to the \gls{iidasspt}. 
	There are principled statistical tests for the validity of the \gls{iidasspt} for a given \gls{dataset}, see \cite{Luetkepol2005} 
	and references therein. 
\end{itemize}

\begin{figure}[htbp]
	\begin{center} 
		\begin{tikzpicture}[scale = 1]
			\draw[->, very thick] (0,0) -- (7.7,0) node[right] {\gls{feature} $\feature$};       
			\draw[->, very thick] (0,0) -- (0,4.2) node[above] {label $\truelabel$};   
			\draw[color=black, thick, dashed, domain = -0.5: 5.2, variable = \x]  plot ({\x},{\x*0.8}) node [] {\hspace*{6mm} $\hypothesis^{(1)}$} ;    
			\draw[color=black, thick, dashed, domain = -0.5: 5.2, variable = \x]  plot ({\x},{\x*0.4})node [] {\hspace*{6mm} $\hypothesis^{(2)}$}   ;     
			\draw[color=black, thick, dashed, domain = -0.5: 5.2, variable = \x]  plot ({\x},{\x*0.2}) node [] {\hspace*{6mm} $\hypothesis^{(3)}$}  ;     
			\coordinate (t1)   at (1.2, 2.48);
			\coordinate (t2) at (1.4, 1);
			\coordinate (t3)   at (3,  2.68);
			
			\coordinate (v1)   at (0,0);
			
			\node at (t1)  [circle,draw,fill=blue,minimum size=6pt,scale=0.6, name=tn1] {};
			\node at (t2)  [circle,draw,fill=blue,minimum size=6pt, scale=0.6, name=tn2] {};
			\node at (t3)  [circle,draw,fill=blue,minimum size=6pt,scale=0.6,  name=tn3] {};
			
			\node at (v1)  [circle,draw,fill=red,minimum size=6pt,scale=0.6,  name=vn1] {};
			
			\draw[fill=blue] (6.2, 3.7)  circle (0.1cm) node [black,xshift=1.3cm] {\gls{trainset}};
			\draw[fill=red] (6.2, 3.2)  circle (0.1cm) node [black,xshift=1.3cm] {\hspace*{2mm} \gls{valset}};
		\end{tikzpicture}
		\caption{An example of an unlucky split of a \gls{dataset} into a \gls{trainset} and a \gls{valset} 
			for the \gls{model} $\hypospace \defeq \{ \hypothesis^{(1)}, \hypothesis^{(2)}, \hypothesis^{(3)} \}$. \label{fig_val_smallter_tain} }
	\end{center}
\end{figure} 

\subsection{Regularization} 
\label{sec_regularization}

Consider an \gls{erm}-based method with \gls{hypospace} $\hypospace$ and \gls{trainset} $\dataset$. 
A key indicator for the performance of such a \gls{ml} method is the ratio 
$\effdim{\hypospace}/|\dataset|$ between the \gls{model} size $\effdim{\hypospace}$ 
and the number $|\dataset|$ of \glspl{datapoint}. The tendency of the \gls{ml} method to 
overfit increases with the ratio $\effdim{\hypospace}/|\dataset|$. 

\Gls{regularization} techniques decrease the ratio $\effdim{\hypospace}/|\dataset|$ via three 
approaches: 
\begin{itemize} 
	\item collect more \glspl{datapoint}, possibly via \gls{dataaug} (see Figure \ref{fig_equiv_dataaug_penal_flbook}), 
	\item add a penalty term $\regparam \regularizer{\hypothesis}$ to average \gls{loss} in \gls{erm} \eqref{equ_def_erm} 
	\begin{equation}
		\label{equ_reg_erm} 
		\hat{\hypothesis} \in \argmin_{\hypothesis \in \hypospace} (1/\samplesize)\sum_{\sampleidx=1}^{\samplesize} \lossfunc{\pair{\featurevec^{(\sampleidx)}}{\truelabel^{(\sampleidx)}}}{\hypothesis}\!+\!\regparam \regularizer{\hypothesis},
	\end{equation} 
	\item shrink the \gls{hypospace}, e.g., by adding constraints on the \glspl{modelparam} such as $\normgeneric{\weights}{2} \leq 10$.
\end{itemize} 
As illustrated in Figure \ref{fig_equiv_dataaug_penal_flbook}, these three forms of \gls{regularization} are 
closely related \cite[Ch. 7]{MLBasics}. For example, the regularized \gls{erm} \eqref{equ_reg_erm} is equivalent 
to \gls{erm} \eqref{equ_def_erm} with a pruned \gls{hypospace} $\hypospace^{(\regparam)} \subseteq \hypospace$. 
Using a larger $\regparam$ typically results in a smaller $\hypospace^{(\regparam)}$. 

One example of \gls{regularization} by adding a penalty term is \gls{ridgeregression}. 
In particular, \gls{ridgeregression} uses the \gls{regularizer} $\regularizer{\hypothesis} \defeq \normgeneric{\weights}{2}^{2}$ 
for a linear \gls{hypothesis} $\hypothesis(\featurevec) \defeq \weights^{T} \featurevec$. 
Thus, \gls{ridgeregression} learns the \glspl{parameter} of a linear \gls{hypothesis} via solving
\begin{equation}
	\label{equ_def_ridgeregression}
	\widehat{\weights}^{(\regparam)}  \in  \argmin_{\weights\in \mathbb{R}^{\dimlocalmodel}} \bigg[  (1/\samplesize) \sum_{\sampleidx=1}^{\samplesize} 
	\big( \truelabel^{(\sampleidx)} - \weights^{T} \featurevec^{(\sampleidx)}  \big)^{2}  + \regparam \normgeneric{\weights}{2}^{2}\bigg]. 
\end{equation} 
The \gls{objfunc} in \eqref{equ_def_ridgeregression} can be interpreted as the \gls{objfunc} of \gls{linreg} 
applied to a modification of the \gls{trainset} $\dataset$: We replace each \gls{datapoint} $\pair{\featurevec}{\truelabel} \in \dataset$ 
by a sufficiently large number of \gls{iid} \glspl{realization} of 
\begin{equation}  
	\label{equ_def_aug_features_perturb} 
	\pair{\featurevec + \mathbf{n}}{\truelabel}  \mbox{, with } \mathbf{n} \sim \mathcal{N}(\mathbf{0},\regparam \mathbf{I}).
\end{equation} 

Thus, \gls{ridgeregression} \eqref{equ_def_ridgeregression} is equivalent to \gls{linreg} applied 
to an augmentation $\dataset'$ of the original \gls{dataset} $\dataset$. The augmentation $\dataset'$ is obtained by 
replacing each \gls{datapoint} $\pair{\featurevec}{\truelabel} \in \dataset$ with a sufficiently 
large number of noisy copies. Each copy of $\pair{\featurevec}{\truelabel}$ is obtained by adding 
an \gls{iid} \gls{realization} $\mathbf{n}$ of a zero-mean Gaussian noise with \gls{covmtx} $\regparam \mathbf{I}$ to 
the \glspl{feature} $\featurevec$ (see \eqref{equ_def_aug_features_perturb}). The \gls{label} of each 
copy of $\pair{\featurevec}{\truelabel}$ is equal to $\truelabel$, i.e., the \gls{label} is not perturbed.

\begin{figure}[htbp]
	\begin{center} 
		\begin{tikzpicture}[scale = 1]
			\draw[->, very thick] (0,0.5) -- (7.7,0.5) node[right] {feature $\feature$};       
			\draw[->, very thick] (0.5,0) -- (0.5,4.2) node[above] {label $\truelabel$};   
			
			\draw[color=black, thick, dashed, domain = -0.5: 5.2, variable = \x]  plot ({\x},{\x*0.4 + 2.0}) ;     
			\node at (5.7,4.1) {$\hypothesis(\feature)$};    
			
			\coordinate (l1)   at (1.2, 2.48);
			\coordinate (l2) at (1.4, 2.56);
			\coordinate (l3)   at (1.7,  2.68);
			
			\coordinate (l4)   at (2.2, 2.2*0.4+2.0);
			\coordinate (l5) at (2.4, 2.4*0.4+2.0);
			\coordinate (l6)   at (2.7,  2.7*0.4+2.0);
			
			\coordinate (l7)   at (3.9,  3.9*0.4+2.0);
			\coordinate (l8) at (4.2, 4.2*0.4+2.0);
			\coordinate (l9)   at (4.5,  4.5*0.4+2.0);
			
			\coordinate (n1)   at (1.2, 1.8);
			\coordinate (n2) at (1.4, 1.8);
			\coordinate (n3)   at (1.7,  1.8);
			
			\coordinate (n4)   at (2.2, 3.8);
			\coordinate (n5) at (2.4, 3.8);
			\coordinate (n6)   at (2.7,  3.8);

			\coordinate (n7)   at (3.9, 2.6);
			\coordinate (n8) at (4.2, 2.6);
			\coordinate (n9)   at (4.5,  2.6);
			
			\node at (n1)  [circle,draw,fill=red,minimum size=6pt,scale=0.6, name=c1] {};
			\node at (n2)  [circle,draw,fill=blue,minimum size=6pt, scale=0.6, name=c2] {};
			\node at (n3)  [circle,draw,fill=red,minimum size=6pt,scale=0.6,  name=c3] {};
			\node at (n4)  [circle,draw,fill=red,minimum size=12pt, scale=0.6, name=c4] {};  
			\node at (n5)  [circle,draw,fill=blue,minimum size=12pt,scale=0.6,  name=c5] {};
			\node at (n6)  [circle,draw,fill=red,minimum size=12pt, scale=0.6, name=c6] {};  
			\node at (n7)  [circle,draw,fill=red,minimum size=12pt,scale=0.6,  name=c7] {};
			\node at (n8)  [circle,draw,fill=blue,minimum size=12pt, scale=0.6, name=c8] {};
			\node at (n9)  [circle,draw,fill=red,minimum size=12pt, scale=0.6, name=c9] {};
			
			\draw [<->] ($ (n7) + (0,-0.3) $)  --  ($ (n9) + (0,-0.3) $) node [pos=0.4, below] {$\sqrt{\regparam}$}; ;

			\draw[<->, color=red, thick] (l1) -- (c1);  
			\draw[<->, color=blue, thick] (l2) -- (c2);  
			\draw[<->, color=red, thick] (l3) -- (c3);  
			\draw[<->, color=red, thick] (l4) -- (c4);  
			\draw[<->, color=blue, thick] (l5) -- (c5);  
			\draw[<->, color=red, thick] (l6) -- (c6);  
			\draw[<->, color=red, thick] (l7) -- (c7);  
			\draw[<->, color=blue, thick] (l8) -- (c8);  
			\draw[<->, color=red, thick] (l9) -- (c9);  
			
			\draw[fill=blue] (6.2, 3.7)  circle (0.1cm) node [black,xshift=2.3cm] {original \gls{trainset} $\dataset$};
			\draw[fill=red] (6.2, 3.2)  circle (0.1cm) node [black,xshift=1.3cm] {augmented};
			\node at (4.6,1.2)  [minimum size=20pt, font=\fontsize{15}{0}\selectfont, text=blue] {$\frac{1}{\samplesize} \sum_{\sampleidx=1}^\samplesize \lossfunc{\pair{\featurevec^{(\sampleidx)}}{ \truelabel^{(\sampleidx)}}}{\hypothesis}$};
			\node at (8.5,1.2)  [minimum size=20pt, font=\fontsize{15}{0}\selectfont, text=red] {$+\regparam \regularizer{\hypothesis}$};
		\end{tikzpicture}
		\caption{Equivalence between \gls{dataaug} and \gls{loss} penalization. \label{fig_equiv_dataaug_penal_flbook} }
	\end{center}
\end{figure} 

To study the \gls{compasp} of \gls{ridgeregression}, we rewrite \eqref{equ_def_ridgeregression} as 
\begin{align} 
	\label{equ_def_ridge_reg_quadratidform}
	\widehat{\weights}^{(\regparam)}  &\in  \argmin_{\weights\in \mathbb{R}^{\dimlocalmodel}} \weights^{T} \mQ \weights + \weights^{T} \vq, \nonumber \\ 
	& \mbox{ with } \mQ \defeq  (1/\samplesize)\mX^{T} \mX + \regparam \mathbf{I} \mbox{, } \vq \defeq (-2/\samplesize) \mX^{T} \vy. 
\end{align} 
Thus, like \gls{linreg} \eqref{equ_def_param_erm_linreg_qadform}, also \gls{ridgeregression} minimizes 
a \gls{convex} \gls{quadfunc}. A main difference between \gls{linreg} \eqref{equ_def_param_erm_linreg_qadform} 
and \gls{ridgeregression} (for $\regparam >0$) is that the matrix $\mQ$ in \eqref{equ_def_ridge_reg_quadratidform} is 
guaranteed to be invertible for any \gls{trainset} $\dataset$. In contrast, the matrix $\mQ$ in 
\eqref{equ_def_param_erm_linreg_qadform} for \gls{linreg} might be singular for some \glspl{trainset}.\footnote{Consider the extreme case 
	where all \glspl{feature} of each \gls{datapoint} in the \gls{trainset} $\dataset$ are zero.}

The statistical properties of the solutions to \eqref{equ_def_ridge_reg_quadratidform} 
crucially depend on the value of $\regparam$. This choice can be guided by an error analysis 
using a \gls{probmodel} for the data (see Proposition \ref{prop_bound_linreg_error}). Instead of 
using a \gls{probmodel}, we can also compare the \gls{trainerr} and \gls{valerr} of the \gls{hypothesis} 
$\hypothesis(\featurevec) = \big( \widehat{\weights}^{(\regparam)} \big)^{T} \featurevec$ learned by 
\gls{ridgeregression} with different values of $\regparam$. 

\subsection{From ML to FL via Regularization} 
\label{sec_from_ml_to_fl}

The main theme of this book is the analysis of \gls{fl} systems that consists of a network 
of \glspl{device}, indexed by $\nodeidx=1,\ldots,\nrnodes$. Each \gls{device} $\nodeidx$ trains 
a local (or personalized) \gls{model} $\localmodel{\nodeidx}$. One natural way to couple 
the \gls{model} training of different \glspl{device} is via \gls{regularization}. 

Assuming parametric \glspl{model} for ease of exposition, each \gls{device} $\nodeidx=1,\ldots,\nrnodes$ 
solves a separate instance of \gls{rerm} \eqref{equ_reg_erm},\footnote{It will be convenient to avoid 
	explicit reference to the local \gls{dataset} of \gls{device} $\nodeidx$ and instead work with the 
	local \gls{lossfunc} $\locallossfunc{\nodeidx}{\localparams{\nodeidx}}$. The \gls{fl} \glspl{algorithm} 
	studied in this book require only access to $\locallossfunc{\nodeidx}{\localparams{\nodeidx}}$. 
}
\begin{equation}
	\label{equ_reg_erm_ml2fl} 
	\estlocalparams{\nodeidx} \in \argmin_{\localparams{\nodeidx} \in \mathbb{R}^{\dimlocalmodel}} \underbrace{ (1/\samplesize)\sum_{\sampleidx=1}^{\samplesize} \lossfunc{\pair{\featurevec^{(\sampleidx)}}{\truelabel^{(\sampleidx)}}}{\hypothesis}}_{=: \locallossfunc{\nodeidx}{\localparams{\nodeidx}}}\!+\!\regparam \localregularizer{\nodeidx}{\localparams{\nodeidx}}.
\end{equation} 
We can couple the instances of \eqref{equ_reg_erm_ml2fl} at \gls{device} $\nodeidx$ with other \glspl{device} 
$\nodeidx' \in \nodes \setminus \{\nodeidx\}$ by using a \gls{regularizer} $\localregularizer{\nodeidx}{\localparams{\nodeidx}}$ 
that depends on their \glspl{modelparam} $\localparams{\nodeidx'}$. For example, we will study constructions 
for $\localregularizer{\nodeidx}{\localparams{\nodeidx}}$ that penalize deviations between the \glspl{modelparam} 
$\localparams{\nodeidx}$ and those at other \glspl{device}, $\localparams{\nodeidx'}$ for $\nodeidx' \in \neighbourhood{\nodeidx}$. 
Figure \ref{fig:erm_smartphones_dependency} illustrates a simple network of two \glspl{device}, each training a personalized \gls{model} via \eqref{equ_reg_erm_ml2fl}. 
Chapter \ref{lec_fldesignprinciple} will discuss in more detail how to construct a useful \gls{regularizer} $\localregularizer{\nodeidx}{\localparams{\nodeidx}}$. 

\begin{figure}[htbp]
	\centering
	\begin{tikzpicture}[node distance=4.5cm and 2cm, thick]
		
		\tikzstyle{smartphone} = [draw, rounded corners=0.2cm, minimum width=1.5cm, minimum height=2cm, fill=gray!10]
		
		\node[smartphone] (i) {};
		\node[smartphone, right=6 cm of i] (j) {};
		
		\node[above=0.2cm of i, font=\bfseries] {\gls{device} $\nodeidx$};
		\node[above=0.2cm of j, font=\bfseries] {\gls{device} $\nodeidx'$};
		
		\node[below=0.7cm of i, align=center] (i_eq) 
		{$\displaystyle \estlocalparams{\nodeidx}\!\in\!\argmin\limits_{\localparams{\nodeidx}}\locallossfunc{\nodeidx}{\localparams{\nodeidx}}\!+\!\regparam \mathcal{R}^{(\nodeidx)}(\localparams{i})$};
		
		\node[below=0.7cm of j, align=center] (j_eq) 
		{$ \estlocalparams{\nodeidx'}\!\in\!\argmin\limits_{\localparams{\nodeidx'}} \locallossfunc{\nodeidx'}{\localparams{\nodeidx'}}\!+\!\regparam \mathcal{R}^{(\nodeidx')}(\localparams{\nodeidx'})$};
		
		\draw[->, thick] (i.east) to[bend left=15] node[above] 
		{$\mathcal{R}^{(i)}$ depends on $\estlocalparams{i'}$} (j.west);
		
		\draw[->, thick] (j.west) to[bend left=15] node[below] 
		{$\mathcal{R}^{(i')}$ depends on $\estlocalparams{i}$} (i.east);
		
	\end{tikzpicture}
	\caption{Two \glspl{device} $\nodeidx$ and $\nodeidx'$ learn personalized \glspl{modelparam}. Each 
		\gls{device} executes a separate instance of regularized \gls{erm} with the \gls{regularizer} depending 
		on the \glspl{modelparam} of the other \gls{device}.\label{fig:erm_smartphones_dependency}}
	
\end{figure}

\clearpage
\subsection{Exercises}

\noindent\refstepcounter{problem}\label{prob:minimaxlinreg}\textbf{\theproblem. Fundamental Limits for \Gls{linreg}.}
\Gls{linreg} learns \glspl{modelparam} of a \gls{linmodel} to minimize 
the \gls{risk} $\expect \big\{ \big(\truelabel - \weights^{T} \featurevec \big)^{2} \big\}$ 
where $\pair{\featurevec}{\truelabel}$ is a \gls{rv}. In practice, we do not observe 
the \gls{rv} $\pair{\featurevec}{\truelabel}$ itself but a (\gls{realization} of a) 
\gls{sequence} of \gls{iid} \glspl{rv} $\pair{\featurevec^{(\timeidx)}}{\truelabel^{(\timeidx)}}$, for $\timeidx=1,2,\ldots$.
The minimax risk is a lower bound on the \gls{risk} achievable by any learning method \cite[Ch. 15]{Wain2019}. 
Determine the minimax risk in terms of the \gls{probdist} of $\pair{\featurevec}{\truelabel}$. 

\noindent\refstepcounter{problem}\label{prob:uniqueeigvalspsd}\textbf{\theproblem. Uniqueness of Eigenvectors.}
Consider the \gls{evd} $\mQ=\sum_{\featureidx=1}^{\dimlocalmodel} \eigval{\featureidx} \vu^{(\featureidx)} \big(\vu^{(\featureidx)}\big)^{T}$ 
of a \gls{psd} matrix $\mQ$. The \gls{evd} consists of orthonormal \glspl{eigenvector} $\vu^{(\featureidx)}$ and 
non-negative \glspl{eigenvalue} $\eigval{\featureidx}$, with $\mQ \vu^{(\featureidx)} = \eigval{\featureidx}\vu^{(\featureidx)}$, 
for $\featureidx=1,\ldots,\dimlocalmodel$. Can you provide conditions on the \glspl{eigenvalue} $\eigval{1}\leq\ldots \leq \eigval{\dimlocalmodel}$ 
such that the (unit-norm) \glspl{eigenvector} are unique?

\noindent\refstepcounter{problem}\label{prob:penaltyasdataaug}\textbf{\theproblem. Penalty Term as \Gls{dataaug}.}
Consider a \gls{ml} method that trains a \gls{model} with \glspl{modelparam} $\weights$. The training 
uses \gls{erm} with \gls{sqerrloss}. Show that \gls{regularization} of the \gls{model} training via adding a penalty term 
$\regparam \normgeneric{\weights}{2}^{2}$ is equivalent to a specific form of \gls{dataaug}. What 
is the augmented \gls{trainset}?

\noindent\refstepcounter{problem}\label{prob:linspacedataaug}\textbf{\theproblem. Data Augmentation via Linear Interpolation.}
Consider a \gls{ml} method that trains a \gls{model}, with \glspl{modelparam} $\weights$, from a \gls{trainset} $\dataset$. 
Each \gls{datapoint} $\datapoint \in \dataset$ is characterized by a \gls{featurevec} $\featurevec \in \mathbb{R}^{\dimlocalmodel}$ and \gls{label} $\truelabel \in \mathbb{R}$, i.e., $\datapoint = \pair{\featurevec}{\truelabel}$. 
We augment the \gls{trainset} by adding, for each pair of two different \glspl{datapoint} $\datapoint, \datapoint'\in \dataset$, 
synthetic \glspl{datapoint} $\tilde{\datapoint}^{(\sampleidx)} \defeq \datapoint + (\datapoint'-\datapoint) \sampleidx/100$ 
and , for $\sampleidx=0,\ldots,99$. Does this augmentation typically increase the \gls{trainerr}?

\noindent\refstepcounter{problem}\label{prob:ridgeasaug}\textbf{\theproblem. Ridge Regression via Deterministic Data Augmentation.}
\Gls{ridgeregression} is obtained from \gls{linreg} by adding the penalty term $\regparam \normgeneric{\weights}{2}^{2}$ 
to the average \gls{sqerrloss} incurred by the \gls{hypothesis} $\hypothesis^{(\weights)}$ on the \gls{trainset} $\dataset$, 
\begin{equation} 
	\label{equ_def_problem_reidgeaug}
	\min_{\weights} (1/\samplesize) \sum_{\sampleidx=1}^{\samplesize} \big( \truelabel^{(\sampleidx)} - \hypothesis\big( \featurevec^{(\sampleidx)} \big) \big)^{2} + \regparam \normgeneric{\weights}{2}^{2}. 
\end{equation}
Construct an augmented \gls{trainset} $\dataset'$ such that the \gls{objfunc} of \eqref{equ_def_problem_reidgeaug} coincides 
with the \gls{objfunc} of plain \gls{linreg} using $\dataset'$ as \gls{trainset}. To construct $\dataset'$, add carefully chosen \glspl{datapoint} 
to the original \gls{trainset} $\dataset = \bigg\{ \pair{\truelabel^{(1)}}{\featurevec^{(1)}},\ldots, \pair{\truelabel^{(\samplesize)}}{\featurevec^{(\samplesize)}} \bigg\}$. Generalize the construction of $\dataset'$ to implement a generalized form of \gls{ridgeregression}, 
\begin{equation} 
	\label{equ_def_problem_reidgeaug_gen}
	\min_{\weights} (1/\samplesize) \sum_{\sampleidx=1}^{\samplesize} \big( \truelabel^{(\sampleidx)} - \hypothesis\big( \featurevec^{(\sampleidx)} \big) \big)^{2} + \regparam \normgeneric{\weights- \widetilde{\weights}}{2}^{2}. 
\end{equation}
Here, we used some prescribed reference \glspl{modelparam} $\widetilde{\weights}$. 
Note that \eqref{equ_def_problem_reidgeaug_gen} reduces to basic \gls{ridgeregression} 
\eqref{equ_def_problem_reidgeaug} for the specific choice $\widetilde{\weights}= \mathbf{0}$. 

\newpage
\section{A Design Principle for FL} 
\label{lec_fldesignprinciple}

Chapter \ref{lec_mlbasics} reviewed \gls{erm} as a central design principle for traditional, 
centralized \gls{ml} systems that rely on a single \gls{dataset} to train a single \gls{model}. 
This chapter extends these foundations to the distributed setting of \gls{fl}, where learning 
takes place over a network of \glspl{device}, each having their own \glspl{dataset} and \glspl{model}.

We begin in Section \ref{sec_emp_graph} by introducing the notion of an \gls{flnetwork} -- 
a mathematical abstraction for \gls{fl} systems. Each node of an \gls{flnetwork} represents a \gls{device} 
that collects a \gls{localdataset} and trains a \gls{localmodel}, while the edges encode communication 
links and statistical similarities between \glspl{localdataset}. 

Section \ref{sec_gtv} introduces the concept of \gls{gtv} as a measure of discrepancy 
between local model parameters at connected nodes. This notion leads directly to 
Section \ref{sec_gtvmin}, where we develop \gls{gtvmin} as a principled regularization 
framework for training parametric \glspl{localmodel} in a federated setting. We then 
generalize this approach in Section \ref{sec_non_param_models} to accommodate 
non-parametric \glspl{localmodel}, broadening its applicability. Finally, Section \ref{sec_interpreations} 
offers several interpretations of \gls{gtvmin} that connect it to broader 
themes in applied mathematics and statistics, highlighting its conceptual and 
practical significance in \gls{fl} design.

\subsection{FL Networks} 
\label{sec_emp_graph}

Consider a \gls{fl} system consisting of a collection of \glspl{device}, indexed by $\nodeidx = 1, \ldots, \nrnodes$. 
The number $\nrnodes$ of \glspl{device} can be arbitrarily large—potentially on the order of billions—as encountered in 
internet-scale \gls{fl} applications.Each \gls{device} $\nodeidx$ can access a \gls{localdataset} $\localdataset{\nodeidx}$ 
and train a personalized \gls{model} $\localmodel{\nodeidx}$. These \glspl{device} collaborate over a communication 
network to learn a local \gls{hypothesis} $\localhypothesis{\nodeidx} \in \localmodel{\nodeidx}$. The quality of each 
local \gls{hypothesis} is assessed using a \gls{lossfunc} $\locallossfunc{\nodeidx}{\localhypothesis{\nodeidx}}$. 

We now introduce the concept of an \gls{flnetwork} as a mathematical model for \gls{fl} applications. 
An \gls{flnetwork} consists of an undirected weighted \gls{graph} $\graph=\left(\nodes,\edges \right)$ 
with nodes $\nodes \defeq \{1,\ldots,\nrnodes\}$ and undirected edges $\edges$ between pairs 
of different nodes. The nodes $\nodes$ represent \glspl{device} with varying amounts of computational 
resources. 

An undirected edge $\edge{\nodeidx}{\nodeidx'} \in \edges$ in an \gls{flnetwork} represents a 
form of similarity between device $\nodeidx$ and device $\nodeidx'$. The amount of similarity is represented by 
an edge weight $\edgeweight_{\nodeidx,\nodeidx'}$. We can collect edge weights into an adjacency matrix 
$\mathbf{A} \in \mathbb{R}^{\nrnodes \times \nrnodes}$, with $A_{\nodeidx,\nodeidx'} = \edgeweight_{\nodeidx',\nodeidx}$. 
Figure \ref{fig_local_dataset} depicts an example of an \gls{flnetwork}. 

\begin{figure}[htbp]
	\begin{center}
		\begin{tikzpicture}[scale=11/5]
			\tikzstyle{every node}=[font=\large]
			
			\tikzstyle{nred} = [circle, fill=red, inner sep=3pt]
			\tikzstyle{ngreen} = [circle, fill=green, inner sep=3pt]
			\tikzstyle{ncyan} = [circle, fill=cyan, inner sep=3pt]
			
			\node[nred] (C1_2) at (0.88+3.7,2.29) {};
			\node[left=1 cm of C1_2,nred] (C1_1)  {};
			\node[below left =1cm and 1cm of C1_2,nred] (C1_3)  {};
			\node[below =0.5cm of C1_2,nred] (C1_4)  {};
			\node[ngreen] (C3_3) at (6,2) {};
			\node[above left =0.4cm and 0.7cm of C3_3,ngreen] (C3_2)  {};
			\node[below left =0.4 and 0.7cm of C3_3,ngreen] (C3_4) {};
			\node[left =1.2cm of C3_3,ngreen] (C3_1) {};
			\node[ncyan] (C2_2) at (2.0,2.23) {};
			\node[below left =0.4cm and 0.4cm of C2_2,ncyan] (C2_1)  {};
			\node[below right =0.4cm and 0.4cm of C2_2,ncyan] (C2_3)  {};
			\node[below right=0.2cm and 0.00cm of C2_3, font=\fontsize{12}{0}\selectfont, anchor=west] {$\localdataset{\nodeidx}, \localmodel{\nodeidx}$};
			\node[above right=0.01cm and 0.00cm of C1_1, font=\fontsize{12}{0}\selectfont, anchor=south] {$\localdataset{\nodeidx'}, \localmodel{\nodeidx'}$};
			
			\draw [line width=1] (C2_3)--(C1_1) node[draw=none, fill=none, font=\fontsize{12}{0}\selectfont, midway, above] {$\edgeweight_{\nodeidx,\nodeidx'}$};
			\draw [line width=1] (C1_2)--(C1_1);
			\draw [line width=1] (C1_2)--(C1_3);
			\draw [line width=1] (C1_1)--(C1_3);
			\draw [line width=1] (C1_3)--(C1_4);
			\draw [line width=1] (C1_2)--(C1_4);
			\draw [line width=1] (C1_4)--(C3_1);
			\draw [line width=1] (C2_1)--(C2_2);
			\draw [line width=1] (C2_2)--(C2_3);
			\draw [line width=1] (C2_1)--(C2_3);
			\draw [line width=1] (C3_1)--(C3_2);
			\draw [line width=1] (C3_2)--(C3_3);
			\draw [line width=1] (C3_3)--(C3_4);
			\draw [line width=1] (C3_2)--(C3_4);
			\draw [line width=1] (C3_1)--(C3_4);
		\end{tikzpicture}
		\caption{\label{fig_local_dataset} Example of an \gls{flnetwork} whose nodes $\nodeidx \in \nodes$ 
			represent different \glspl{device}. Each \gls{device} $\nodeidx$ generates a \gls{localdataset} $\localdataset{\nodeidx}$ 
			and trains a \gls{localmodel} $\localmodel{\nodeidx}$. Some \glspl{device} $\nodeidx,\nodeidx'$ 
			are connected by an undirected edge $\edge{\nodeidx}{\nodeidx'}$ with a positive edge 
			weight $\edgeweight_{\nodeidx,\nodeidx'}$.}
	\end{center}
\end{figure} 
Note that the undirected edges $\edges$ of an \gls{flnetwork} encode a symmetric 
notion of similarity between \glspl{device}: If the \gls{device} $\nodeidx$ is similar to the \gls{device} 
$\nodeidx'$, i.e., $\edge{\nodeidx}{\nodeidx'} \in \edges$, 
then also the \gls{device} $\nodeidx'$ is similar to the \gls{device} $\nodeidx$. For some \gls{fl} 
applications, an asymmetric notion of similarity, represented by directed edges, could be 
more accurate. However, the generalization of an \gls{flnetwork} to directed graphs is 
beyond the scope of this book. 

It can be convenient to replace a given \gls{flnetwork} $\graph$ with an equivalent 
fully connected \gls{flnetwork} $\graph'$ (see Figure \ref{equ_figure_fully_connected}). 
The fully connected \gls{graph} $\graph'$ contains an edge between every pair of two 
different nodes $\nodeidx, \nodeidx'$, $$\edges' = \big\{ \edge{\nodeidx}{\nodeidx'}: \nodeidx,\nodeidx' \in \nodes, \nodeidx \neq \nodeidx' \big\}.$$ 
The \glspl{edgeweight} are chosen $\edgeweight'_{\nodeidx,\nodeidx'} =  \edgeweight_{\nodeidx,\nodeidx'}$ 
for any edge $\edge{\nodeidx}{\nodeidx'} \in \edges$ and $\edgeweight'_{\nodeidx,\nodeidx'} = 0$ if 
the original \gls{flnetwork} $\graph$ does not contain an edge between nodes $\nodeidx,\nodeidx'$. 

\begin{figure}[htbp]
	\begin{center} 
		\begin{tikzpicture}
			
			\begin{scope}[xshift=-3cm]
				\node[circle, fill, inner sep=3pt, label=left:{$1$}] (A) at (0,2) {};
				\node[circle, fill, inner sep=3pt, label=right:{$2$}] (B) at (2,2) {};
				\node[circle, fill, inner sep=3pt, label=left:{$3$}] (C) at (0,0) {};
				\node[circle, fill, inner sep=3pt, label=right:{$4$}] (D) at (2,0) {};
				
				\draw [line width=1pt] (A) -- (B);
				\draw [line width=1pt]  (A) -- (C);
				\draw [line width=1pt]  (B) -- (D);
			\end{scope}
			
			\begin{scope}[xshift=3cm]
				\node[circle, fill, inner sep=3pt, label=left:{$1$}] (A) at (0,2) {};
				\node[circle, fill, inner sep=3pt, label=right:{$2$}] (B) at (2,2) {};
				\node[circle, fill, inner sep=3pt, label=left:{$3$}] (C) at (0,0) {};
				\node[circle, fill, inner sep=3pt, label=right:{$4$}] (D) at (2,0) {};
				
				\draw [line width=1pt]  (A) -- (B);
				\draw [line width=1pt]  (A) -- (C);
				\draw [line width=1pt]  (A) -- (D);
				\draw [line width=1pt]  (B) -- (C);
				\draw [line width=1pt]  (B) -- (D);
				\draw [line width=1pt]  (C) -- (D);
			\end{scope}
		\end{tikzpicture}
	\end{center} 
	
	\caption{\label{equ_figure_fully_connected}  Left: An \gls{flnetwork} $\graph$ consisting 
		of $\nrnodes=4$ nodes. Right: Equivalent fully connected \gls{flnetwork} $\graph'$ with the 
		same nodes and non-zero \glspl{edgeweight} $\edgeweight'_{\nodeidx,\nodeidx'} = \edgeweight_{\nodeidx,\nodeidx'}$ 
		for $\{\nodeidx,\nodeidx'\} \in \edges$ and $\edgeweight'_{\nodeidx,\nodeidx'} = 0$ for $\{\nodeidx,\nodeidx'\} \notin \edges$. }
\end{figure} 

An \gls{flnetwork} is more than the undirected weighted \gls{graph} $\graph$: It also includes the 
\gls{localdataset} $\localdataset{\nodeidx}$ and the \gls{localmodel} $\localmodel{\nodeidx}$ (or its \glspl{modelparam} $\localparams{\nodeidx})$ 
for each \gls{device} $\nodeidx \in \nodes$. The details of the generation and the format 
of a \gls{localdataset} will not be important in what follows. A \gls{localdataset} is just one 
possible means to construct a \gls{lossfunc} in order to evaluate \glspl{modelparam}. 
However, to build intuition, we can think of a \gls{localdataset} $\localdataset{\nodeidx}$ as a 
labelled dataset 
\begin{equation} 
	\label{equ_def_local_dataset_plain}
	\localdataset{\nodeidx} \defeq \left\{ \big(\featurevec^{(\nodeidx,1)},\truelabel^{(\nodeidx,1)}\big), \ldots,\big(\featurevec^{(\nodeidx,\samplesize_{\nodeidx})},\truelabel^{(\nodeidx,\samplesize_{\nodeidx})}\big) \right\}.  
\end{equation} 
Here, $\featurevec^{(\nodeidx,\sampleidx)}$ and $\truelabel^{(\nodeidx,\sampleidx)}$ denote, 
respectively, the \glspl{feature} and the \gls{label} of the $\sampleidx$th \gls{datapoint} in 
the \gls{localdataset} $\localdataset{\nodeidx}$. Note that the size $\samplesize_{\nodeidx}$ 
of the \gls{localdataset} can vary between different nodes $\nodeidx \in \nodes$. 

It is convenient to collect the \gls{feature} vectors $\featurevec^{(\nodeidx,\sampleidx)}$ and 
\glspl{label} $\truelabel^{(\nodeidx,\sampleidx)}$ into a \gls{featuremtx} $\featuremtx^{(\nodeidx)}$ and \gls{label} 
vector $\vy^{(\nodeidx)}$, respectively, 
\begin{equation} 
	\featuremtx^{(\nodeidx)} \defeq \big(\featurevec^{(\nodeidx,1)},\ldots,\featurevec^{(\nodeidx,\localsamplesize{\nodeidx})} \big)^{T} \mbox{, and }
	\labelvec^{(\nodeidx)} \defeq \big( \truelabel^{(\nodeidx,1)},\ldots,\truelabel^{(\nodeidx,\localsamplesize{\nodeidx})} \big)^{T}. \label{equ_label_vec_node_i} 
\end{equation} 
The \gls{localdataset} $\localdataset{\nodeidx}$ can then be represented compactly by the \gls{featuremtx}  
$\featuremtx^{(\nodeidx)} \in \mathbb{R}^{\localsamplesize{\nodeidx} \times \dimlocalmodel}$ and 
the vector $\labelvec^{(\nodeidx)} \in \mathbb{R}^{\localsamplesize{\nodeidx}}$. 

Besides the \gls{localdataset} $\localdataset{\nodeidx}$, each node $\nodeidx \in \graph$ 
also carries a \gls{localmodel} $\localmodel{\nodeidx}$. Our focus is on parametric \glspl{localmodel} 
with by \glspl{modelparam} $\localparams{\nodeidx} \in \mathbb{R}^{\dimlocalmodel}$, for $\nodeidx\!=\!1,\ldots,\nrnodes$. 
The usefulness of a specific choice of the \gls{localmodel} parameter $\localparams{\nodeidx}$ is then 
measured by a local \gls{lossfunc} $\locallossfunc{\nodeidx}{\localparams{\nodeidx}}$, for $\nodeidx = 1,\ldots,\nrnodes$. 
Note that we can use different local \glspl{lossfunc} $\locallossfunc{\nodeidx}{\cdot} \neq \locallossfunc{\nodeidx'}{\cdot}$ 
at different nodes $\nodeidx, \nodeidx' \in \nodes$. 

We now have introduced all the components of an \gls{flnetwork}. Strictly speaking, an \gls{flnetwork} 
is a tuple $\big(\graph,\{\localmodel{\nodeidx}\}_{\nodeidx \in \nodes} , \{\locallossfunc{\nodeidx}{\cdot}\}_{\nodeidx \in \nodes} \big)$ 
consisting of an undirected weighted \gls{graph} $\graph$, a \gls{localmodel} $\localmodel{\nodeidx}$ 
and local \gls{lossfunc} $\locallossfunc{\nodeidx}{\cdot}$ for each node $\nodeidx \in \nodes$. 
In principle, all of these components are design choices that influence the computational 
and statistical properties of the \gls{fl} \glspl{algorithm} presented in Chapter \ref{lec_flalgorithms}. 
To some extend, also the edges $\edges$ in the \gls{flnetwork} are a design choice. 

The role (or meaning) of an edge $\edge{\nodeidx}{\nodeidx'}$ in an \gls{flnetwork} is two-fold: 
First, it represents a communication link that allows to exchange messages between \glspl{device} 
$\nodeidx,\nodeidx'$. Second, an edge $\edge{\nodeidx}{\nodeidx'}$ indicates similar statistical 
properties of \glspl{localdataset} generated by \glspl{device} $\nodeidx, \nodeidx'$. 
It then seems natural to learn similar \gls{hypothesis} maps $\localhypothesis{\nodeidx}, \localhypothesis{\nodeidx'}$. 
This is actually the main idea behind all the \gls{fl} algorithms that we will discuss in the rest of 
this book. To make this idea precise, we next discuss how to obtain quantitative measures for 
how much local \gls{hypothesis} maps $\localhypothesis{\nodeidx}$ vary 
across the edges $\edge{\nodeidx}{\nodeidx'} \in \edges$ of an \gls{flnetwork}. 

\subsection{Generalized Total Variation} 
\label{sec_gtv} 

Consider an \gls{flnetwork} with nodes $\nodeidx=1,\ldots,\nrnodes$, undirected edges $\edges$ with 
\glspl{edgeweight} $\edgeweight_{\nodeidx,\nodeidx'} \!>\!0 $ for each $\edge{\nodeidx}{\nodeidx'} \in \edges$. 
For each edge $\{\nodeidx,\nodeidx'\}\!\in\!\edges$, we want to couple the training of the corresponding 
\glspl{localmodel} $\localmodel{\nodeidx}, \localmodel{\nodeidx'}$. The strength of this coupling is determined 
by the edge weight $\edgeweight_{\nodeidx,\nodeidx'}$. We implement the coupling by penalizing the variation 
(or \gls{discrepancy})between the \glspl{modelparam} $\localparams{\nodeidx},\localparams{\nodeidx'}$.  

We can measure the variation between two trained \glspl{localmodel} $\localhypothesis{\nodeidx}, \localhypothesis{\nodeidx'}$ 
across an edge $\edge{\nodeidx}{\nodeidx'} \in \edges$ in different ways. For example, 
we can compare their \glspl{prediction} on a common \gls{testset} $\dataset$ by computing 
\begin{equation} 
\label{equ_def_discrpancy_generic}
\discrepancy{\nodeidx}{\nodeidx'} \defeq (1/|\dataset|) \sum_{\featurevec \in \dataset} \big[ \localhypothesis{\nodeidx}(\featurevec) - \localhypothesis{\nodeidx'}(\featurevec) \big]^{2}.
\end{equation} 
In principle, we can use a different \gls{testset} in \eqref{equ_def_discrpancy_generic} for 
each edge $\edge{\nodeidx}{\nodeidx'}$ of $\graph$. For example, the \gls{testset} could 
be obtained by merging randomly selected \glspl{datapoint} from each \gls{localdataset} $\localdataset{\nodeidx}$, $\localdataset{\nodeidx'}$. 

Our main focus will be \gls{fl} applications that use parametric \glspl{localmodel}, i.e., 
each node learns local \glspl{modelparam} $\localparams{\nodeidx} \in \mathbb{R}^{\dimlocalmodel}$, 
for $\nodeidx=1,\ldots,\nrnodes$. Here, we can measure the variation between $\hypothesis^{(\localparams{\nodeidx})}$ 
and $\hypothesis^{(\localparams{\nodeidx'})}$ directly in terms of the \glspl{modelparam} $\localparams{\nodeidx}, \localparams{\nodeidx'}$ 
at the nodes of an edge $\edge{\nodeidx}{\nodeidx'}$. In particular, 
we use a \gls{regularizer} $\gtvpenalty: \mathbb{R}^{\dimlocalmodel} \rightarrow \mathbb{R}$ that 
measures the difference between the \glspl{modelparam}, 
\begin{equation} 
	\label{equ_def_discrpancy_generic_param}
	\discrepancy{\nodeidx}{\nodeidx'} \defeq \gtvpenalty\big( \localparams{\nodeidx}-\localparams{\nodeidx'}\big).
\end{equation}  

The penalty function $\gtvpenalty$ will be mainly a design choice. Our main requirement is that 
$\gtvpenalty$ is monotonically increasing\footnote{A function $f : \mathbb{R} \to \mathbb{R}$ is monotonically increasing 
if $f(x) \leq f(y)$ whenever $x \leq y$. This means that larger argument values never result in smaller function values.} 
with respect to some norm in the \gls{euclidspace} $\mathbb{R}^{\dimlocalmodel}$ \cite{ClusteredFLTVMinTSP,Smith2017}. 
This requirement ensures symmetry, i.e., $\gtvpenalty\big( \localparams{\nodeidx}-\localparams{\nodeidx'}\big) = \gtvpenalty\big( \localparams{\nodeidx'}-\localparams{\nodeidx}\big)$, allowing its 
use as a measure of variation across an undirected edge $\edge{\nodeidx}{\nodeidx'} \in \edges$.

Summing up the edge-wise variations (weighted by the \glspl{edgeweight}) yields the 
\gls{gtv} of a collection of local \glspl{modelparam}, 
\begin{equation} 
	\label{equ_def_tv_penalty} 
	\sum_{\edge{\nodeidx}{\nodeidx'} \in \edges} \edgeweight_{\nodeidx,\nodeidx'} \gtvpenalty\big( \localparams{\nodeidx}-\localparams{\nodeidx'}\big). 
\end{equation} 
Our main focus will be on the special case of \eqref{equ_def_tv_penalty}, obtained 
for $\gtvpenalty(\cdot) \defeq \normgeneric{\cdot}{2}^{2}$, 
\begin{equation} 
	\label{equ_def_tv_sq_norm} 
	\sum_{\edge{\nodeidx}{\nodeidx'} \in \edges} \edgeweight_{\nodeidx,\nodeidx'}  \normgeneric{\weights^{(\nodeidx)}- \weights^{(\nodeidx')}}{2}^{2}. 
\end{equation} 

The choice of penalty $\gtvpenalty(\cdot)$ has a crucial impact on the computational and 
statistical properties of the \gls{fl} \glspl{algorithm} presented in Chapter \ref{lec_flalgorithms}. 
Our main choice during the rest of this book will be the penalty function $\gtvpenalty(\cdot) \defeq \normgeneric{\cdot}{2}^{2}$. 
This choice often allows to formulate \gls{fl} as the minimization of a \gls{smooth} \gls{convex} 
function, which can be done via simple \glspl{gdmethod} (see Chapter \ref{lec_graphlearning}). 
On the other hand, choosing $\gtvpenalty$ to be a norm results in \gls{fl} \glspl{algorithm} that 
require more computation but less training \gls{data} \cite{ClusteredFLTVMinTSP}. 

The connectivity of an \gls{flnetwork} $\graph$ can be characterized locally - around a node $\nodeidx \in \nodes$ - 
by its \gls{nodedegree}
\begin{equation} 
	\label{equ_def_node_degree} 
	\nodedegree{\nodeidx} \defeq \sum_{\nodeidx' \in \neighbourhood{\nodeidx}} \edgeweight_{\nodeidx,\nodeidx'}. 
\end{equation} 
Here, we used the \gls{neighborhood} $\neighbourhood{\nodeidx} \defeq \{ \nodeidx' \in \nodes: \edge{\nodeidx}{\nodeidx'} \in \edges \}$ 
of node $\nodeidx \in \nodes$. A global characterization for the connectivity of $\graph$ is the maximum \gls{nodedegree}
\begin{equation} 
	\label{equ_def_max_node_degree} 
	\maxnodedegree^{(\graph)} \defeq \max_{\nodeidx \in \nodes} \nodedegree{\nodeidx} \stackrel{\eqref{equ_def_node_degree}}{=}  \max_{\nodeidx \in \nodes} \sum_{\nodeidx' \in \neighbourhood{\nodeidx}} \edgeweight_{\nodeidx,\nodeidx'}. 
\end{equation} 

Besides inspecting the \glspl{nodedegree}, we can study the connectivity of $\graph$ 
also via the \glspl{eigenvalue} and \glspl{eigenvector} of its \gls{LapMat} $\LapMat{\graph} \in \mathbb{R}^{\nrnodes \times \nrnodes}$.\footnote{The study of \glspl{graph} via the \glspl{eigenvalue} and \glspl{eigenvector} of associated matrices is the main 
	subject of spectral \gls{graph} theory \cite{ChungSpecGraphTheory,spielman2025sagt}.}
The \gls{LapMat} of an undirected weighted \gls{graph} $\graph$ is defined element-wise as 
\begin{equation}
	\label{equ_def_Lap_mat_entry}
	\LapMatEntry{\graph}{\nodeidx}{\nodeidx'} \defeq \begin{cases} - \edgeweight_{\nodeidx,\nodeidx'} & \mbox{ for } \nodeidx\neq \nodeidx', \edge{\nodeidx}{\nodeidx'} \in \edges \\ 
		\sum_{\nodeidx'' \neq \nodeidx} \edgeweight_{\nodeidx,\nodeidx''} & \mbox{ for } \nodeidx = \nodeidx' \\ 
		0 & \mbox{ else.} \end{cases}
\end{equation}
Figure \ref{fig_lap_mtx_flbook} illustrates the \gls{LapMat} of a small \gls{graph}. 

\begin{figure}[htbp]
	\begin{center} 
		\begin{minipage}{0.4\textwidth}
			\begin{tikzpicture}
				\begin{scope}[every node/.style={circle, fill=black, inner sep=2pt}]
					\node[label=above:{$1$}] (1) at (0, 0) {};
					\node[label=left:{$2$}] (2) [below left=1cm and 1cm of 1] {};
					\node[label=right:{$3$}] (3) [below right=1cm and 1cm of 1] {};
					
					\draw [line width=1pt] (1) -- (2);
					\draw [line width=1pt] (1) -- (3);
				\end{scope}
			\end{tikzpicture}
		\end{minipage} 
		\begin{minipage}{0.4\textwidth}
			
			\begin{equation} 
				\nonumber
				\LapMat{\graph} = \begin{pmatrix} 2 & -1& -1 \\ -1& 1 & 0 \\  -1 & 0 & 1 \end{pmatrix}  
			\end{equation} 
		\end{minipage}
	\end{center} 
	\caption{\label{fig_lap_mtx_flbook} Left: Example of an \gls{flnetwork} $\graph$ with three 
	nodes $\nodeidx=1,2,3$ that are connected via two edges with unit weight $\edgeweight_{1,2}=\edgeweight_{1,3}=1$.
		Right: \Gls{LapMat} $\LapMat{\graph}  \in \mathbb{R}^{3 \times 3}$ of $\graph$.} 
\end{figure}

The \gls{LapMat} is symmetric and \gls{psd}, which follows from the identity 
\begin{align} 
	\label{equ_quad_form_Laplacian}
	\weights^{T} (\LapMat{\graph} \otimes \mathbf{I}) \weights & = \sum_{\edge{\nodeidx}{\nodeidx'} \in \edges} \edgeweight_{\nodeidx,\nodeidx'}  \normgeneric{\weights^{(\nodeidx)}- \weights^{(\nodeidx')}}{2}^{2}  \nonumber \\  
	&  \mbox{ for any } \dimlocalmodel \in \mathbb{N}, \weights \defeq \underbrace{\bigg( \big(\weights^{(1)}\big)^{T},\ldots,\big(\weights^{(\nrnodes)}\big)^{T}  \bigg)^{T}}_{=: {\rm stack} \big\{ \weights^{(\nodeidx)} \big\}_{\nodeidx=1}^{\nrnodes}} \in \mathbb{R}^{\dimlocalmodel \nrnodes}. 
\end{align} 

As a \gls{psd} matrix, $\LapMat{\graph}$ possesses an \gls{evd} 
\begin{equation} 
	\label{equ_evd_LapMat}
	\LapMat{\graph} = \sum_{\nodeidx=1}^{\nrnodes} \eigval{\nodeidx} \vu^{(\nodeidx)} \big(  \vu^{(\nodeidx)}  \big)^{T}, 
\end{equation} 
with orthonormal \glspl{eigenvector} $\vu^{(1)},\ldots,\vu^{(\nrnodes)}$ and corresponding 
list of \glspl{eigenvalue} 
\begin{equation}
	\label{equ_def_order_eigvals_LapMat}  
	0 = \eigval{1}\big(\LapMat{\graph}\big) \leq \eigval{2}\big(\LapMat{\graph}\big) \leq \ldots \leq \eigval{\nrnodes}\big(\LapMat{\graph}\big). 
\end{equation}
We just write $\eigval{\nodeidx}$ instead of $\eigval{\nodeidx}\big(\LapMat{\graph}\big)$ if the \gls{LapMat} 
$\LapMat{\graph}$ is clear from context.  
The \gls{eigenvalue} $\eigval{\nodeidx}\big(\LapMat{\graph}\big)$ corresponds to the \gls{eigenvector} $\vu^{(\nodeidx)}$, i.e., 
$ \LapMat{\graph} \vu^{(\nodeidx)}  = \eigval{\nodeidx}\big(\LapMat{\graph}\big)\vu^{(\nodeidx)}$ for $\nodeidx=1,\ldots,\nrnodes$.

It is important to note that the ordered list of \glspl{eigenvalue} \eqref{equ_def_order_eigvals_LapMat} 
is uniquely determined for a given \gls{LapMat}. In contrast, the \glspl{eigenvector} $\vu^{(\nodeidx)}$ in \eqref{equ_evd_LapMat} 
are not unique in general.\footnote{Consider the scenario where the list \eqref{equ_def_order_eigvals_LapMat} 
	contains repeated entries, i.e., some of the \glspl{eigenvector} have identical \glspl{eigenvalue}.} 

The ordered \glspl{eigenvalue} $\eigval{\nodeidx}\big(\LapMat{\graph}\big)$ in \eqref{equ_def_order_eigvals_LapMat} 
can be computed (or characterized) via the \gls{cfwmaxmin} \cite[Thm. 8.1.2.]{GolubVanLoanBook}. 
Two important special cases of this characterization are \cite{spielman2025sagt,ChungSpecGraphTheory}
\begin{align} 
	\eigval{\nrnodes}\big(\LapMat{\graph}\big) &\stackrel{\rm \gls{cfwmaxmin}}{=} \max_{\substack{\vv \in \mathbb{R}^{\nrnodes} \\ \normgeneric{\vv}{}=1}} \vv^{T} \LapMat{\graph} \vv \nonumber \\ 
	& \stackrel{\eqref{equ_quad_form_Laplacian}}{=}  \max_{\substack{\vv \in \mathbb{R}^{\nrnodes} \\ \normgeneric{\vv}{}=1}} \sum_{\edge{\nodeidx}{\nodeidx'} \in \edges} \edgeweight_{\nodeidx,\nodeidx'}  \big( v_{\nodeidx}- v_{\nodeidx'} \big)^{2}  
\end{align} 
and 
\begin{align} 
	\label{equ_variational_eigval_2}
	\eigval{2}\big(\LapMat{\graph}\big) &\stackrel{\rm \gls{cfwmaxmin}}{=} \min_{\substack{\vv \in \mathbb{R}^{\nrnodes} \\ \vv^{T} \mathbf{1}  = 0 \\ \normgeneric{\vv}{}=1}} \vv^{T} \LapMat{\graph} \vv \nonumber \\ 
	& \stackrel{\eqref{equ_quad_form_Laplacian}}{=} \min_{\substack{\vv \in \mathbb{R}^{\nrnodes} \\ \vv^{T} \mathbf{1}  = 0 \\ \normgeneric{\vv}{}=1}} \hspace*{2mm}\sum_{\edge{\nodeidx}{\nodeidx'} \in \edges} \hspace*{-2mm}\edgeweight_{\nodeidx,\nodeidx'}  \big( v_{\nodeidx}- v_{\nodeidx'} \big)^{2}. 
\end{align} 

By \eqref{equ_quad_form_Laplacian}, we can compute the \gls{gtv} of a collection of \glspl{modelparam} 
via the quadratic form $\weights^{T} \big( \LapMat{\graph} \otimes \mathbf{I}_{\dimlocalmodel\times \dimlocalmodel}\big) \weights$. 
This quadratic form involves the vector $\weights \in \mathbb{R}^{\nrnodes \dimlocalmodel}$ which is obtained 
by stacking the local \glspl{modelparam} $\localparams{\nodeidx}$ for $\nodeidx=1,\ldots,\nrnodes$. 
Another consequence of \eqref{equ_quad_form_Laplacian} is that any 
collection of identical local \glspl{modelparam}, stacked into the \gls{vector}
\begin{equation}
	\label{equ_def_const_eigvec} 
	\weights = {\rm stack} \{ \vc \} = \big(\vc^{T},\ldots,\vc^{T}  \big)^{T} \mbox{, with some } \vc \in \mathbb{R}^{\dimlocalmodel} \setminus \{ \mathbf{0} \}, 
\end{equation} 
is an \gls{eigenvector} of $\LapMat{\graph} \otimes \mI$ with corresponding \gls{eigenvalue} $\eigval{1}=0$ (see \eqref{equ_def_order_eigvals_LapMat}). Thus, the \gls{LapMat} of any \gls{flnetwork} is singular (non-invertible). 

The second \gls{eigenvalue} $\eigval{2}$ of $\LapMat{\graph}$ provides a great deal of information about the 
connectivity structure of $\graph$.\footnote{With slight abuse of language, we 
	will sometimes speak about the \glspl{eigenvalue} of a \gls{flnetwork} $\graph$. However, 
	we actually mean the \glspl{eigenvalue} of the \gls{LapMat} \eqref{equ_def_Lap_mat_entry} 
	naturally associated with $\graph$.} Indeed, much of spectral graph theory is devoted 
	to the analysis of $\eigval{2}$, which is also referred to as algebraic connectivity, 
	for different \gls{graph} constructions \cite{spielman2025sagt,ChungSpecGraphTheory}.
\begin{itemize} 
	\item Consider the 
	case $\eigval{2}=0$: Here, beside the \gls{eigenvector} \eqref{equ_def_const_eigvec}, we can find at least 
	one additional \gls{eigenvector} 
	\begin{equation} 
		\label{equ_non_constant_eigvec_l20}
		\widetilde{\weights}= {\rm stack} \big\{ \localparams{\nodeidx} \big\}_{\nodeidx=1}^{\nrnodes} \mbox{ with } \localparams{\nodeidx} \neq \localparams{\nodeidx'} \mbox{ for some } \nodeidx, \nodeidx' \in \nodes, 
	\end{equation} 
	of $\LapMat{\graph}\otimes \mI$ with \gls{eigenvalue} equal to $0$. In this case, the graph $\graph$ is not \gls{connected}, i.e., 
	we can find two subsets (components) of nodes that do not have any edge 
	between them (see Figure \ref{fig_simple_graph_two_components}). 
	For each connected component $\cluster$, we can construct the \gls{eigenvector} by assigning 
	the same (non-zero) vector $\vc \in \mathbb{R}^{\dimlocalmodel} \setminus \{ \mathbf{0} \}$ to all nodes 
	$\nodeidx \in \cluster$ and the zero vector $\mathbf{0}$ to the remaining nodes $\nodeidx \in \nodes \setminus \cluster$.  
	\item On the other hand, if $\eigval{2} > 0$ then $\graph$ is \gls{connected}. Moreover, 
	the larger the value of $\eigval{2}$, the stronger the connectivity between the nodes in $\graph$. 
	Indeed, adding edges to $\graph$ can only increase the objective in \eqref{equ_variational_eigval_2} 
	and, in turn, $\eigval{2}$. 
\end{itemize} 

\begin{figure}[h]
	\begin{center}
		\begin{tikzpicture}
			\node[draw, circle, fill=black, minimum size=5pt, inner sep=0pt] (a) at (0, 0) {};
			\node[draw, circle, fill=black, minimum size=5pt, inner sep=0pt] (b) at (-1, -1) {};
			\node[draw, circle, fill=black, minimum size=5pt, inner sep=0pt] (c) at (1, -1) {};
			
			\draw [line width=1pt] (a) -- (b);
			\draw [line width=1pt] (b) -- (c);
			
			\node [xshift=1pt,anchor=north] at (c) {$\nodedegree{\nodeidx}=1$};
			
			\node[draw, circle, fill=black, minimum size=5pt, inner sep=0pt] (d) at (4, 0) {};
			\node[draw, circle, fill=black, minimum size=5pt, inner sep=0pt] (e) at (3, -1) {};
			\node[draw, circle, fill=black, minimum size=5pt, inner sep=0pt] (f) at (5, -1) {};
			
			\draw [line width=1pt] (d) -- (e);
			\draw [line width=1pt] (e) -- (f);
			\draw [line width=1pt] (f) -- (d);
			
			
			\node at (0, 0.5) {component $\cluster^{(1)}$};
			\node at (4, 0.5) {component $\cluster^{(2)}$};
			\vspace*{-1mm}
		\end{tikzpicture}
		\vspace*{-6mm}
	\end{center} 
	\caption{An \gls{flnetwork} $\graph$ that consists of $\nrnodes\!=\!6$ nodes forming 
		two connected components $\cluster^{(1)}, \cluster^{(2)}$. \label{fig_simple_graph_two_components}}
\end{figure} 

In what follows, we will make use of the lower bound \cite[Thm. 2.0.1]{spielman2025sagt}
\begin{equation}
	\label{equ_lower_bound_tv_eigval}
	\sum_{\edge{\nodeidx}{\nodeidx'} \in \edges} \edgeweight_{\nodeidx,\nodeidx'}  \normgeneric{\weights^{(\nodeidx)}- \weights^{(\nodeidx')}}{2}^{2}  \geq   \eigval{2}  \sum_{\nodeidx=1}^{\nrnodes} \normgeneric{\weights^{(\nodeidx)}- {\rm avg}\{\localparams{\nodeidx} \}}{2}^{2}.
\end{equation}  
Here, $ {\rm avg}\{\localparams{\nodeidx} \} \defeq (1/\nrnodes) \sum_{\nodeidx=1}^{\nrnodes} \weights^{(\nodeidx)}$ is 
the average of all \gls{localmodel} \glspl{parameter}. The bound \eqref{equ_lower_bound_tv_eigval} 
follows from \eqref{equ_quad_form_Laplacian} and the \gls{cfwmaxmin} for the \glspl{eigenvalue} of 
the matrix $\LapMat{\graph} \otimes \mI$. 

The quantity $\sum_{\nodeidx=1}^{\nrnodes} \normgeneric{\weights^{(\nodeidx)}-  {\rm avg}\{\localparams{\nodeidx} \}_{\nodeidx=1}^{\nrnodes}}{2}^{2}$ 
on the right-hand side of \eqref{equ_lower_bound_tv_eigval} has an interesting geometric 
interpretation: It is the squared Euclidean norm of the projection of the stacked \gls{localmodel} \glspl{parameter} 
$$\weights \defeq \bigg( \big(\weights^{(1)}\big)^{T},\ldots, \big(\weights^{(\nrnodes)}\big)^{T} \bigg)^{T}$$ 
onto the orthogonal complement of the subspace 
\begin{equation}
	\label{equ_def_subspace_constant_local}
	\mathcal{S} \defeq \bigg\{ \mathbf{1} \otimes \va: \va \in \mathbb{R}^{\dimlocalmodel} \bigg\} \!=\!\bigg\{  \big( \va^{T},\ldots,\va^{T} \big)^{T} \mbox{, for some } \va\!\in\!\mathbb{R}^{\dimlocalmodel} \bigg\}\!\subseteq\!\mathbb{R}^{\dimlocalmodel \nrnodes}. 
\end{equation} 
The subspace $\mathcal{S}$ consists of stacked local \glspl{modelparam} $\localparams{\nodeidx}$ 
that are identical for all nodes $\nodeidx=1,\ldots,\nrnodes$. 
Such a structure arises in certain \gls{fl} settings where a single global \gls{model} 
is shared among all \glspl{device}. In this setting, the local \glspl{modelparam} satisfy 
$\localparams{\nodeidx} = \va$ for all $\nodeidx=1,\ldots,\nrnodes$ and 
some common vector $\va \in \mathbb{R}^{\dimlocalmodel}$ (see Section \ref{sec_single_model_fl}). 
Equivalently, the condition  
\[
\big( \localparams{1}, \ldots, \localparams{\nrnodes} \big)^{T} \in \mathcal{S}
\]  
characterizes this Single-\gls{model} setting as membership in the subspace $\mathcal{S}$.

The projection $\mathbf{P}_{\mathcal{S}} \weights$ of $\weights \in \mathbb{R}^{\nrnodes \dimlocalmodel}$ 
on $\mathcal{S}$ is 
\begin{equation} 
	\label{equ_def_projection_constant_localparms}
	\mathbf{P}_{\mathcal{S}} \weights = \big( \va^{T},\ldots,\va^{T} \big)^{T} \mbox{, with } \va = {\rm avg}\{\localparams{\nodeidx} \}_{\nodeidx=1}^{\nrnodes}.
\end{equation} 
The projection on the orthogonal complement $\mathcal{S}^{\perp}$, in turn, is 
\begin{equation} 
	\label{equ_def_orth_projection_constant_localparms}
	\mathbf{P}_{\mathcal{S}^{\perp}} \weights  = \weights - \mathbf{P}_{\mathcal{S}}\weights	= {\rm stack} \big\{ \localparams{\nodeidx} - {\rm avg}\{\localparams{\nodeidx} \}_{\nodeidx=1}^{\nrnodes}\big\}_{\nodeidx=1}^{\nrnodes}.
\end{equation} 

\subsection{Generalized Total Variation Minimization} 
\label{sec_gtvmin}

Consider an \gls{flnetwork} $\graph$ whose nodes $\nodeidx \in \nodes$ represent individual 
\glspl{device}, each learning personalized \glspl{modelparam} $\localparams{\nodeidx}$. 
The quality of a specific choice of \glspl{modelparam} is assessed via a local \gls{lossfunc} 
$\locallossfunc{\nodeidx}{\localparams{\nodeidx}}$, typically derived from a \gls{trainerr} 
applied to a \gls{localdataset}.

Our focus is on \gls{fl} applications where these local \glspl{lossfunc} alone do not 
suffice to reliably train a high-dimensional \gls{localmodel}. For instance, the 
\gls{localdataset} $\localdataset{\nodeidx}$ used to compute the local \gls{lossfunc} 
at some node $\nodeidx$ may be too small relative to the \gls{effdim} of the underlying 
\gls{localmodel} $\localmodel{\nodeidx}$, making the training process prone to \gls{overfitting} 
(see Section \ref{sec_regularization}).\footnote{As a rule of thumb, the number of \glspl{datapoint} 
used to train a \gls{model} should be proportional to the (effective) number of its \glspl{parameter}.} 

In such settings, collaboration between the \glspl{device} across the edges of the \gls{flnetwork} 
becomes essential. To address this, we seek \gls{localmodel} \glspl{parameter} that not only minimize the local 
\glspl{lossfunc} but also exhibit small \gls{gtv} \eqref{equ_def_tv_penalty}. Requiring a small \gls{gtv} 
couples the training among neighboring \glspl{device} and results in an implicit (and privacy-friendly) 
pooling of \glspl{localdataset} (see Section \ref{sec_clustered_fl}). 

\gls{gtv} minimization (GTVMin) optimally balances the (average) local \gls{loss} and the \gls{gtv} \eqref{equ_def_tv_penalty} 
of local \glspl{modelparam} $\localparams{\nodeidx}$, 
\begin{equation}
	\label{equ_def_gtvmin_penalty} 
	\big\{ \widehat{\weights}^{(\nodeidx)} \big\}_{\nodeidx=1}^{\nrnodes} \in \hspace*{-1mm}\argmin_{\weights^{(1)},\ldots,\weights^{(\nrnodes)}} \hspace*{-1mm}\sum_{\nodeidx \in \nodes} \locallossfunc{\nodeidx}{\localparams{\nodeidx}} + \regparam  \hspace*{-2mm}\sum_{\edge{\nodeidx}{\nodeidx'} \in \edges} 
	\hspace*{-2mm}\edgeweight_{\nodeidx,\nodeidx'}  \gtvpenalty\big( \localparams{\nodeidx}- \localparams{\nodeidx'}\big).
\end{equation} 
Note that \eqref{equ_def_gtvmin_penalty} is parametrized by the choice for the penalty 
function $\gtvpenalty(\cdot)$. We discuss the effect of different choices for 
$\gtvpenalty(\cdot)$ in Section \ref{sec_compasp_gtvmin} and \ref{sec_statasp_gtvmin}.
Our main focus will be on the special case of \eqref{equ_def_gtvmin_penalty}, obtained 
with $\gtvpenalty(\cdot) \defeq \normgeneric{\cdot}{2}^{2}$, 
\begin{equation}
	\label{equ_def_gtvmin} 
	\big\{ \widehat{\weights}^{(\nodeidx)} \big\}_{\nodeidx=1}^{\nrnodes} \in \hspace*{-1mm}\argmin_{\weights^{(1)},\ldots,\weights^{(\nrnodes)}} \hspace*{0mm}\sum_{\nodeidx \in \nodes} \locallossfunc{\nodeidx}{\weights^{(\nodeidx)}}\!+\!\regparam  \hspace*{-2mm}\sum_{\edge{\nodeidx}{\nodeidx'} \in \edges} 
	\hspace*{-2mm}\edgeweight_{\nodeidx,\nodeidx'}  \normgeneric{\weights^{(\nodeidx)}\!-\!\weights^{(\nodeidx')}}{2}^{2}.
\end{equation} 

The \gls{gtvmin} parameter $\regparam\!>\!0$ in \eqref{equ_def_gtvmin_penalty} steers the 
preference for learning local \glspl{modelparam} $\localparams{\nodeidx}$ with small \gls{gtv} 
versus incurring small local \gls{loss} $\sum_{\nodeidx \in \nodes} \locallossfunc{\nodeidx}{\localparams{\nodeidx}}$. 
For $\regparam\!=\!0$, \gls{gtvmin} decomposes into fully independent local \gls{erm} instances $\min_{\localparams{\nodeidx}} \locallossfunc{\nodeidx}{\cdot}$, for $\nodeidx=1,\ldots,\nrnodes$. On the other hand, increasing the value of $\regparam$ makes the 
solutions of \eqref{equ_def_gtvmin_penalty} increasingly clustered: the 
local \glspl{modelparam} $\estlocalparams{\nodeidx}$ become approximately constant 
over increasingly large subsets of nodes. This behavior is appealing for \gls{cfl} which 
we discuss in Section \ref{sec_clustered_fl}. 

Choosing $\regparam$ beyond a critical value - that depends on the shape of the local \glspl{lossfunc} 
and the edges $\edges$ - results in $\estlocalparams{\nodeidx}$ being (nearly) constant over 
all nodes $\nodeidx \in \nodes$. In practice, the choice of $\regparam$ can be guided by 
\gls{validation} \cite{hastie01statisticallearning} or by a probabilistic analysis of the solutions 
of \eqref{equ_def_gtvmin_penalty}. Section \ref{sec_statasp_gtvmin} presents an example of such an analysis. 

Note that \gls{gtvmin} \eqref{equ_def_gtvmin_penalty} is an instance of \gls{rerm}: The \gls{regularizer} 
is the \gls{gtv} of \gls{localmodel} \glspl{parameter} over the weighted edges $\edgeweight_{\nodeidx,\nodeidx'}$ 
of the \gls{flnetwork}. Loosely speaking, \gls{gtvmin} couples the training of \glspl{localmodel} 
by requiring them to be similar across the edges of the \gls{flnetwork}. For the extreme case of an 
\gls{flnetwork} without any edges, \gls{gtvmin} decomposes into independent \gls{erm} instances 
$$\argmin_{\localparams{\nodeidx}} \hspace*{0mm} \locallossfunc{\nodeidx}{\localparams{\nodeidx}} \mbox{, for each }  \nodeidx=1,\ldots,\nrnodes. $$

The connectivity (i.e., the edges $\edges$) of the \gls{flnetwork} is an important design 
choice in \gls{gtvmin}-based methods. This choice can be guided by \gls{compasp} and \gls{statasp} of \gls{gtvmin}-based \gls{fl} 
systems. Some application domains allow to leverage domain expertise to guess a useful choice 
for the \gls{flnetwork}. If \glspl{localdataset} are generated at different geographic locations, 
we might use nearest-neighbour graphs based on geodesic distances between \gls{data} 
generators (e.g., \gls{fmi} weather stations). Chapter \ref{lec_graphlearning} discusses 
\gls{graph} learning methods that determine the \glspl{edgeweight} $\edgeweight_{\nodeidx,\nodeidx'}$ 
in a data-driven fashion, i.e., directly from the \glspl{localdataset} $\localdataset{\nodeidx}, \localdataset{\nodeidx'}$. 

{\bf \gls{gtvmin} for \glspl{linmodel}.} Let us now consider the special case of \gls{gtvmin} with 
\glspl{localmodel} being a \gls{linmodel}. For each node $\nodeidx \in \nodes$ of the \gls{flnetwork}, 
we want to learn the \glspl{parameter} $\weights^{(\nodeidx)}$ of a linear \gls{hypothesis} $\hypothesis^{(\nodeidx)}(\featurevec) \defeq \big( \weights^{(\nodeidx)} \big)^{T} \featurevec$. 
We measure the quality of the \glspl{parameter} via the average \gls{sqerrloss} 
\begin{align} 
	\label{equ_def_localloss_linreg}
	\locallossfunc{\nodeidx}{\weights^{(\nodeidx)}} & \defeq (1/\localsamplesize{\nodeidx}) \sum_{\sampleidx=1}^{\localsamplesize{\nodeidx}} \bigg( \truelabel^{(\nodeidx,\sampleidx)} - \big(\weights^{(\nodeidx)} \big)^{T} \featurevec^{(\nodeidx,\sampleidx)} \bigg)^{2} \nonumber \\ 
	& \stackrel{\eqref{equ_label_vec_node_i}}{=} (1/\localsamplesize{\nodeidx}) \normgeneric{\labelvec^{(\nodeidx)} - \featuremtx^{(\nodeidx)} \weights^{(\nodeidx)} }{2}^{2}.
\end{align} 

Inserting \eqref{equ_def_localloss_linreg} into \eqref{equ_def_gtvmin}, yields the 
following instance of \gls{gtvmin} to train local \glspl{linmodel}, 
\begin{align} 
	\label{equ_def_gtvmin_linreg} 
	\hspace*{-3mm}\big\{ \widehat{\weights}^{(\nodeidx)} \big\} \!\in\! \hspace*{-1mm} \argmin_{\{ \weights^{(\nodeidx)} \}_{\nodeidx=1}^{\nrnodes}} \sum_{\nodeidx \in \nodes} (1/\localsamplesize{\nodeidx}) \normgeneric{\labelvec^{(\nodeidx)} \!-\! \featuremtx^{(\nodeidx)} \weights^{(\nodeidx)}}{2}^{2}\!+\!\regparam \hspace*{-2mm}\sum_{\edge{\nodeidx}{\nodeidx'} \in \edges} \hspace*{-3mm}
	\edgeweight_{\nodeidx,\nodeidx'}  \normgeneric{\weights^{(\nodeidx)} \!-\!\weights^{(\nodeidx')}}{2}^{2}.
\end{align} 
The identity \eqref{equ_quad_form_Laplacian} allows to rewrite \eqref{equ_def_gtvmin_linreg} using the \gls{LapMat} 
$\LapMat{\graph}$ as 
\begin{align} 
	\label{equ_def_gtvmin_linreg_1} 
	\widehat{\weights}^{(\nodeidx)}\!\in\! \argmin_{\weights = {\rm stack} 	\big\{ \weights^{(\nodeidx)} \big\} }\sum_{\nodeidx \in \nodes} (1/\localsamplesize{\nodeidx}) \normgeneric{\labelvec^{(\nodeidx)} \!-\! \featuremtx^{(\nodeidx)} \weights^{(\nodeidx)}}{2}^{2}\!+\!\regparam \weights^{T} \big( \LapMat{\graph} \otimes \mathbf{I}_{\dimlocalmodel} \big) \weights.
\end{align} 
Let us rewrite the \gls{objfunc} in \eqref{equ_def_gtvmin_linreg_1} as 
\begin{align} 
	&\hspace*{-20mm} \weights^{T} \left(  \begin{pmatrix}
		\mQ^{(1)} & \cdots & 0 \\
		\vdots  & \ddots & \vdots \\
		0  & \cdots & \mQ^{(\nrnodes)}
	\end{pmatrix}\!+\!\regparam \LapMat{\graph} \otimes  \mI \right) \weights \!+\!\big( \big( \vq^{(1)} \big)^{T},\ldots,\big( \vq^{(\nrnodes)}\big) ^{T}  \big)  \weights  \\[3mm]
	\mbox{ with } \mQ^{(\nodeidx)} & \!=\! (1/\localsamplesize{\nodeidx})\big( \featuremtx^{(\nodeidx)} \big)^{T} \featuremtx^{(\nodeidx)} \mbox{ and } \vq^{(\nodeidx)} \defeq (-2/\localsamplesize{\nodeidx}) \big(\featuremtx^{(\nodeidx)} \big)^{T} \vy^{(\nodeidx)}.  \nonumber
\end{align}
Thus, like \gls{linreg} \eqref{equ_def_param_erm_linreg_qadform} and \gls{ridgeregression} \eqref{equ_def_ridge_reg_quadratidform}, 
\gls{gtvmin} \eqref{equ_def_gtvmin_linreg_1} (for local \gls{linmodel}s $\localmodel{\nodeidx}$)  
minimizes a \gls{convex} \gls{quadfunc}, 
\begin{equation} 
	\label{equ_gtvmin_linreg_quadproblem}
	\big\{ \widehat{\weights}^{(\nodeidx)} \big\}_{\nodeidx=1}^{\nrnodes} \in \argmin_{\weights = {\rm stack} \big\{ \weights^{(\nodeidx)} \big\}_{\nodeidx=1}^{\nrnodes}} \weights^{T} \mQ \weights +  \vq^{T} \weights. 
\end{equation}
Here, we used the \gls{psd} matrix 
\begin{equation} 
	\label{equ_def_mQ_gtvmin_lr}
	\hspace*{-4mm}\mQ\!\defeq\!\begin{pmatrix}
		\mQ^{(1)} & \cdots & \mathbf{0} \\
		\vdots  & \ddots & \vdots \\
		\mathbf{0}  & \cdots & \mQ^{(\nrnodes)}
	\end{pmatrix}\!+\!\regparam \LapMat{\graph}\!\otimes\!\mI  \mbox{ with  }  \mQ^{(\nodeidx)}\!\defeq\!(1/\localsamplesize{\nodeidx})\big( \featuremtx^{(\nodeidx)} \big)^{T} \featuremtx^{(\nodeidx)} \
\end{equation}
and the vector 
\begin{equation}
	\label{equ_def_vq_gtvmin_lr}
	\vq \defeq \big( \big( \vq^{(1)} \big)^{T},\ldots,\big( \vq^{(\nrnodes)}\big) ^{T}  \big)^{T} \mbox{, with } \vq^{(\nodeidx)} \defeq (-2/\localsamplesize{\nodeidx}) \big(\featuremtx^{(\nodeidx)} \big)^{T} \vy^{(\nodeidx)}.
\end{equation}

\subsubsection{Computational Aspects of \gls{gtvmin}} 
\label{sec_compasp_gtvmin}

Chapter \ref{lec_flalgorithms} will apply optimization methods to solve \gls{gtvmin} \eqref{equ_def_gtvmin_penalty}, 
resulting in practical \gls{fl} algorithms. Different instances of \gls{gtvmin} favour different 
classes of optimization methods. For example, using a \gls{differentiable} \gls{lossfunc} $\locallossfunc{\nodeidx}{\cdot}$ and 
penalty function $\gtvpenalty(\cdot)$ allows to apply \glspl{gdmethod} (see Chapter \ref{lec_gradientmethods}) 
to solve \gls{gtvmin}. Another important class of \gls{lossfunc}s are those for which 
we can efficiently compute the \gls{proxop} \cite{BeckFOMBook,ProximalMethods} 
\begin{equation} 
\label{equ_def_prox_op_original}
\proximityop{ \locallossfunc{\nodeidx}{\cdot} }{\weights}{\rho} \defeq \argmin_{\weights' \in \mathbb{R}^{\dimlocalmodel}} \locallossfunc{\nodeidx}{\weights'} + (\rho/2) \normgeneric{\weights - \weights'}{2}^{2} \mbox{ for some } \rho > 0. 
\end{equation} 
We refer to functions $\locallossfunc{\nodeidx}{\cdot}$ for which \eqref{equ_def_prox_op_original}
can be computed easily as \emph{simple} or \gls{proximable} \cite{Condat2013}. \gls{gtvmin} \eqref{equ_def_gtvmin} 
with \gls{proximable} \gls{lossfunc}s can be solved via proximal algorithms \cite{ProximalMethods}. 
Besides influencing the choice of optimization method, the design choices underlying 
\gls{gtvmin} also determine the amount of computation that is required by a given optimization method. 

Chapter \ref{lec_flalgorithms} discusses \gls{fl} \glspl{algorithm} that are obtained by applying 
\glspl{fixedpointiter} to solve \gls{gtvmin}. These fixed-point iterations repeatedly apply 
a fixed-point operator which is determined by the \gls{flnetwork} (including the choice 
for the local \glspl{lossfunc}, \glspl{localmodel} and edges in the \gls{flnetwork}). 
The computational complexity of the resulting iterative method has two factors: 
(i) the amount of computation required by a single iteration (i.e., the per-iteration complexity) 
and (ii) the number iterations required by the method to achieve a sufficiently 
accurate approximate solution of \gls{gtvmin}. 

The fixed-point iterations used in Chapter \ref{lec_flalgorithms} to design \gls{fl} algorithms 
can be implemented as message passing over the \gls{flnetwork}. These \glspl{algorithm} require an 
amount of computation that is proportional to the number of edges of the \gls{flnetwork}. 
Clearly, using an \gls{flnetwork} with few edges (i.e., using a sparse \gls{graph}) results in a 
smaller per-iteration complexity. 

The number of iterations required by an \gls{fl} \gls{algorithm} employing a fixed-point 
operator $\fixedpointop$ depends on the contraction properties of $\fixedpointop$. 
These contraction properties can be influenced through design choices for the \gls{flnetwork}, 
such as selecting local \gls{lossfunc}s that are \gls{strcvx}. In addition to affecting the 
iteration count, the contraction properties of $\fixedpointop$ also play a crucial role 
in determining whether the \gls{fl} algorithm can tolerate asynchronous execution.

It is instructive to study the \gls{compasp} of the special case of \gls{gtvmin} \eqref{equ_def_gtvmin_linreg} 
for local \gls{linmodel}s. As discussed above, this instance is equivalent to solving \eqref{equ_gtvmin_linreg_quadproblem}. 
Any solution $\widehat{\weights}$ of \eqref{equ_gtvmin_linreg_quadproblem} (and, in turn, \eqref{equ_def_gtvmin_linreg})
is characterized by the \gls{zerogradientcondition} 
\begin{equation}
	\label{equ_def_zerogradient_qadproblem}
	\mQ \widehat{\weights} = - (1/2) \vq, 
\end{equation} 
with $\mQ, \vq$ as defined in \eqref{equ_def_mQ_gtvmin_lr} and \eqref{equ_def_vq_gtvmin_lr}. 
If the matrix $\mQ$ in \eqref{equ_def_zerogradient_qadproblem} is invertible, the solution 
to \eqref{equ_def_zerogradient_qadproblem} and, in turn, to the \gls{gtvmin} instance \eqref{equ_def_gtvmin_linreg} 
is unique and given by $\widehat{\weights} = (-1/2) \mQ^{-1} \vq$. 

The size of the matrix $\mQ$ (see \eqref{equ_def_mQ_gtvmin_lr}) is proportional 
to the number of nodes in the \gls{flnetwork} $\graph$ which might be in the order 
of millions (or even billions) for internet-scale applications. For such large systems, we 
typically cannot use direct matrix inversion methods (such as Gaussian elimination)
to compute $\mQ^{-1}$.\footnote{How many arithmetic operations (addition, multiplication) 
	do you think are required to invert an arbitrary matrix $\mathbf{Q} \in \mathbb{R}^{\dimlocalmodel \times \dimlocalmodel}$?} 
Instead, we typically need to resort to iterative methods \cite{peng2014,LapSolverVishnoi}. 

One important family of such iterative methods are the \glspl{gdmethod} which we will 
discuss in Chapter \ref{lec_gradientmethods}. Starting from an initial choice of the 
\gls{localmodel} \glspl{parameter} $\widehat{\weights}_{0} = \big( \widehat{\weights}_{0}^{(1)},\ldots, \widehat{\weights}_{0}^{(\nrnodes)} \big)$, 
these methods repeat variants of a \gls{gradstep}, 
$$ \widehat{\weights}_{\iteridx+1} \defeq  \widehat{\weights}_{\iteridx} -  \lrate \big( 2 \mQ \widehat{\weights}_{\iteridx}   + \vq \big) \mbox{ for } \iteridx=0,1,\ldots.
$$
The \gls{gradstep} results in the updated \gls{localmodel} \glspl{parameter} $\widehat{\weights}^{(\nodeidx)}$ 
which we stacked into 
$$ \widehat{\weights}_{\iteridx+1}  \defeq \bigg(  \big( \widehat{\weights}^{(1)}\big)^{T},\ldots, \big( \widehat{\weights}^{(\nrnodes)} \big) ^{T}\bigg)^{T}.$$ We repeat the \gls{gradstep} for a sufficient number of times, according to 
some \gls{stopcrit} (see Chapter \ref{lec_gradientmethods}).

\subsubsection{Statistical Aspects of \gls{gtvmin}}
\label{sec_statasp_gtvmin}

How useful are the solutions of \gls{gtvmin} \eqref{equ_def_gtvmin} as a choice for the 
local \glspl{modelparam}? To answer this question, we use - as for the statistical analysis 
of \gls{erm} in Chapter \ref{lec_mlbasics} - a \gls{probmodel} for the \gls{localdataset}s. 
In particular, we use a variant of an \gls{iidasspt}: Each \gls{localdataset} $\localdataset{\nodeidx}$ 
consists of \gls{datapoint}s whose \gls{feature}s and \gls{label}s are \gls{realization}s 
of \gls{iid} \gls{rv}s 
\begin{align} 
	\label{equ_def_probmodel_linreg_node_i}
	\hspace*{-3mm}\labelvec^{(\nodeidx)}& = \hspace*{-1mm}\underbrace{\big( \featurevec^{(\nodeidx,1)},\ldots,\featurevec^{(\nodeidx,\localsamplesize{\nodeidx})}  \big)^{T}}_{\mbox{ \tiny{local \gls{featuremtx} }} \featuremtx^{(\nodeidx)} } \overline{\weights}^{(\nodeidx)} + {\bm \varepsilon}^{(\nodeidx)} \\ 
	& \mbox{ with } \featurevec^{(\nodeidx,\sampleidx)} \!\stackrel{\mbox{\tiny{\gls{iid}}}}{\sim}\!\mathcal{N}(\mathbf{0},\mathbf{I}) \mbox{, for } \sampleidx=1,\ldots,\localsamplesize{\nodeidx},\nodeidx=1,\ldots,\nrnodes, \nonumber \\ 
	& \mbox{ and }{\bm \varepsilon}^{(\nodeidx)}\!\sim\!\mathcal{N}(\mathbf{0},\sigma^2\mathbf{I}) \mbox{, for } \nodeidx=1,\ldots,\nrnodes. \nonumber 
\end{align} 
In contrast to the \gls{probmodel} \eqref{equ_def_probmodel_linreg} (which we used for the 
analysis of \gls{erm}), the \gls{probmodel} \eqref{equ_def_probmodel_linreg_node_i} allows 
for different node-specific \glspl{parameter} $\overline{\weights}^{(\nodeidx)}$, for $\nodeidx \in \nodes$. In particular, the 
entire \gls{dataset} obtained from pooling all \gls{localdataset}s does not conform to an \gls{iidasspt}. 

In what follows, we focus on the \gls{gtvmin} instance \eqref{equ_def_gtvmin_linreg} to learn the 
\glspl{parameter} $\weights^{(\nodeidx)}$ of a local \gls{linmodel} for each node $\nodeidx \in \nodes$. 
For a reasonable choice of \gls{flnetwork}, the \glspl{parameter} $\overline{\weights}^{(\nodeidx)}, \overline{\weights}^{(\nodeidx')}$ 
at connected nodes $\edge{\nodeidx}{\nodeidx'} \in \edges$ should be similar. We cannot 
choose the \gls{edgeweight}s based on \glspl{parameter} $\overline{\weights}^{(\nodeidx)}$ as they 
are unknown. However, we can still use estimates of $\overline{\weights}^{(\nodeidx)}$ that 
are computed from the available \gls{localdataset}s (see Chapter \ref{lec_graphlearning}). 

Consider an \gls{flnetwork} with nodes carrying \gls{localdataset}s generated from 
the \gls{probmodel} \eqref{equ_def_probmodel_linreg_node_i} with true \glspl{modelparam} 
$\overline{\weights}^{(\nodeidx)}$. For ease of exposition, we assume that 
\begin{equation} 
	\label{equ_def_constant_linmodel_params} 
	\overline{\weights}^{(\nodeidx)}=\overline{\vc} \mbox{, for some } \overline{\vc} \in \mathbb{R}^{\dimlocalmodel} \mbox{ and all } \nodeidx \in \nodes.
\end{equation} 

To study the deviation between the solutions $\widehat{\weights}^{(\nodeidx)}$ of 
\eqref{equ_def_gtvmin_linreg} and the true underlying \glspl{parameter} $\overline{\weights}^{(\nodeidx)}$, 
we decompose it as
\begin{equation} 
	\label{equ_decompose_sol_gtvmin}
	\widehat{\weights}^{(\nodeidx)} = \widetilde{\weights}^{(\nodeidx)} +  \widehat{\vc} \mbox{, with } \widehat{\vc} \defeq (1/\nrnodes) \sum_{\nodeidx'=1}^{\nrnodes} \widehat{\weights}^{(\nodeidx')}. 
\end{equation} 
The component $\widehat{\vc}$ is identical at all nodes $\nodeidx \in \nodes$ and obtained as the 
orthogonal projection of $\widehat{\weights}={\rm stack}\big\{ \estlocalparams{\nodeidx} \}_{\nodeidx=1}^{\nrnodes}$ 
on the subspace \eqref{equ_def_subspace_constant_local}. The component 
$\widetilde{\weights}^{(\nodeidx)} \defeq	\widehat{\weights}^{(\nodeidx)}  -  (1/\nrnodes) \sum_{\nodeidx'=1}^{\nrnodes} \widehat{\weights}^{(\nodeidx')}$ 
consists of the deviations, for each node $\nodeidx$, between the \gls{gtvmin} solution $\widehat{\weights}^{(\nodeidx)}$ and 
their average over all nodes. Trivially, the average of the deviations $\widetilde{\weights}^{(\nodeidx)}$ 
across all nodes is the zero vector, $ (1/\nrnodes) \sum_{\nodeidx=1}^{\nrnodes} \widetilde{\weights}^{(\nodeidx)} = \mathbf{0}$. 

The decomposition \eqref{equ_decompose_sol_gtvmin} entails an analogous (orthogonal) 
decomposition of the error $\widehat{\weights}^{(\nodeidx)} - \overline{\weights}^{(\nodeidx)}$. 
Indeed, for identical true underlying \glspl{modelparam} \eqref{equ_def_constant_linmodel_params} 
(which makes $\overline{\weights}$ an element of the subspace \eqref{equ_def_subspace_constant_local}), 
we have 
\begin{equation}
	\label{equ_decmposition_error_constant_parameters_GTVMIN_linreg}
	\sum_{\nodeidx=1}^{\nrnodes} \normgeneric{	\widehat{\weights}^{(\nodeidx)} - \overline{\weights}^{(\nodeidx)}}{2}^{2} 
	\stackrel{\eqref{equ_def_constant_linmodel_params},\eqref{equ_decompose_sol_gtvmin}}{=} 	\underbrace{\sum_{\nodeidx=1}^{\nrnodes} \normgeneric{	\vc - \widehat{\vc} }{2}^{2}}_{\nrnodes \normgeneric{	\vc - \widehat{\vc} }{2}^{2}} +	\sum_{\nodeidx=1}^{\nrnodes} \normgeneric{	\widetilde{\weights}^{(\nodeidx)}  }{2}^{2}. 
\end{equation} 
The following proposition provides an upper bound on the second error component in \eqref{equ_decmposition_error_constant_parameters_GTVMIN_linreg}. 
\begin{prop} 
	\label{prop_second_component_error}
	Consider a connected \gls{flnetwork}, i.e., $\eigval{2} > 0$ (see \eqref{equ_def_order_eigvals_LapMat}), 
	and the solution \eqref{equ_decompose_sol_gtvmin} to \gls{gtvmin} \eqref{equ_def_gtvmin_linreg} for the 
	\gls{localdataset}s \eqref{equ_def_probmodel_linreg_node_i}. If the true local \glspl{modelparam} in 
	\eqref{equ_def_probmodel_linreg_node_i} are identical (see \eqref{equ_def_constant_linmodel_params}), we can 
	upper bound the deviation $\widetilde{\weights}^{(\nodeidx)} \defeq \widehat{\weights}^{(\nodeidx)}  - (1/\nrnodes)\sum_{\nodeidx=1}^{\nrnodes} \widehat{\weights}^{(\nodeidx)}$ of learned \glspl{modelparam} $\widehat{\weights}^{(\nodeidx)}$ from their average, as 
	\begin{equation} 
		\label{equ_upper_bound_variation_component}
		\sum_{\nodeidx=1}^{\nrnodes} \normgeneric{	\widetilde{\weights}^{(\nodeidx)}  }{2}^{2} \leq \frac{1}{\eigval{2}\regparam} \sum_{\nodeidx=1}^{\nrnodes} (1/\localsamplesize{\nodeidx}) \normgeneric{{\bm \varepsilon}^{(\nodeidx)}}{2}^{2}.
	\end{equation} 
\end{prop} 
\begin{proof} 
	See Section \ref{sec_proof_second_comp_error}. 
\end{proof}
Note that Proposition \ref{prop_second_component_error} only applies to \gls{gtvmin} over a 
\gls{flnetwork} with a connected \gls{graph} $\graph$. A necessary and sufficient 
condition for $\graph$ to be connected is that the second smallest \gls{eigenvalue} 
is positive, $\eigval{2} >0$. However, for an \gls{flnetwork} with a \gls{graph} 
$\graph$ that is not connected, we can still apply Proposition \ref{prop_second_component_error} 
separately to each connected component of $\graph$. 

The upper bound \eqref{equ_upper_bound_variation_component} involves three components: 
\begin{itemize} 
	\item the properties of \glspl{localdataset}, via the noise terms ${\bm \varepsilon}^{(\nodeidx)}$ in \eqref{equ_def_probmodel_linreg_node_i}, 
	\item the \gls{flnetwork} via the \gls{eigenvalue} $\eigval{2}\big(\LapMat{\graph}\big)$ (see \eqref{equ_def_order_eigvals_LapMat}), 
	\item the \gls{gtvmin} parameter $\regparam$. 
\end{itemize} 
According to \eqref{equ_upper_bound_variation_component}, we can ensure a small error 
component $\widetilde{\weights}^{(\nodeidx)}$ of the \gls{gtvmin} solution by choosing a 
large value $\regparam$. Thus, by \eqref{equ_decmposition_error_constant_parameters_GTVMIN_linreg}, 
for sufficiently large $\regparam$, the local \glspl{modelparam} $\estlocalparams{\nodeidx}$
delivered by \gls{gtvmin} are approximately identical for all nodes $\nodeidx \in \nodes$ of 
a connected \gls{flnetwork} (where $\eigval{2}\big(\LapMat{\graph}\big)>0$). 

Enforcing identical local \glspl{modelparam} at all nodes of a \gls{flnetwork} is desirable for \gls{fl} applications 
that require to learn a common (global) \glspl{modelparam} for all nodes \cite{pmlr-v54-mcmahan17a}. 
However, some \gls{fl} applications involve heterogeneous \gls{device}s that generate 
\glspl{localdataset} with significantly different statistics \cite{ClusteredFLTVMinTSP}. For such applications it 
is detrimental to enforce common \glspl{modelparam} at all nodes (see Chapter \ref{lec_flmainflavours}). 
Instead, we should enforce common \glspl{modelparam} only for nodes with \glspl{localdataset} 
having similar statistical properties. This is exactly the objective of clustered \gls{fl} which we discuss 
in Section \ref{sec_clustered_fl}.

\subsection{Non-Parametric Models in FL Networks} 
\label{sec_non_param_models}

In its basic form \eqref{equ_def_gtvmin}, \gls{gtvmin} can only be applied to parametric \gls{localmodel}s with 
\glspl{modelparam} belonging to the same \gls{euclidspace} $\mathbb{R}^{\dimlocalmodel}$. 
Some \gls{fl} applications involve non-parametric \glspl{localmodel} (such as \glspl{decisiontree}) 
or parametric \glspl{localmodel} with varying parametrizations (e.g., nodes use different \gls{deepnet} 
architectures). Here, we cannot use the difference between \glspl{modelparam} as 
a measure for the discrepancy between $\localhypothesis{\nodeidx}$ and $\localhypothesis{\nodeidx'}$ 
across an edge $\edge{\nodeidx}{\nodeidx'} \in \edges$. 

One way to measure the \gls{discrepancy} between two \gls{hypothesis} 
maps $\localhypothesis{\nodeidx},\localhypothesis{\nodeidx'}$ is to compare their 
\glspl{prediction} on a \gls{dataset} 
\begin{equation}
	\nonumber
	\dataset^{\edge{\nodeidx}{\nodeidx'}} = \left\{ \featurevec^{(1)},\ldots, \featurevec^{(\samplesize')}\right\}.
\end{equation}
For each edge $\edge{\nodeidx}{\nodeidx'}$, the connected nodes need to agree on 
\gls{dataset} $\dataset^{\edge{\nodeidx}{\nodeidx'}}$. Note that the \gls{dataset} 
$\dataset^{\edge{\nodeidx}{\nodeidx'}}$ can be different for different edges. Examples for 
constructions of $\dataset^{\edge{\nodeidx}{\nodeidx'}}$ include \gls{iid} \glspl{realization} 
of some \gls{probdist} or by using subsets of $\localdataset{\nodeidx}$ and $\localdataset{\nodeidx'}$ (see Exercise \ref{prob:privfriendlydiscr}). 

We compare the \glspl{prediction} delivered by $\localhypothesis{\nodeidx}$ and $\localhypothesis{\nodeidx'}$ 
on $\dataset^{\edge{\nodeidx}{\nodeidx'}}$ using some \gls{lossfunc} $\loss$. In particular, 
we define the \gls{discrepancy} measure 
\begin{align}
	\label{equ_def_variation_non_parametric}
	\discrepancy{\nodeidx}{\nodeidx'} \defeq (1/\samplesize') \sum_{\featurevec \in \dataset^{\edge{\nodeidx}{\nodeidx'}}} 
	(1/2)& \big[\lossfunc{\pair{\featurevec }{\hypothesis^{(\nodeidx)}\big(\featurevec\big)}}{\hypothesis^{(\nodeidx')}}
	\nonumber \\ 
	&	+\lossfunc{\pair{\featurevec}{\hypothesis^{(\nodeidx')}\big(\featurevec \big)}}{\hypothesis^{(\nodeidx)}} \big].
\end{align}  
Different choices for the \gls{lossfunc} in \eqref{equ_def_variation_non_parametric} result 
in different computational and statistical properties of the resulting \gls{fl} \glspl{algorithm}. 
For real-valued \glspl{prediction} we can use the \gls{sqerrloss} in \eqref{equ_def_variation_non_parametric}, yielding 
\begin{equation}
	\label{equ_def_variation_non_parametric_sqerr}
	\discrepancy{\nodeidx}{\nodeidx'} \defeq (1/\samplesize') \sum_{\featurevec \in \dataset^{\edge{\nodeidx}{\nodeidx'}}} 
	\big[ \hypothesis^{(\nodeidx)}\big(\featurevec\big) -  \hypothesis^{(\nodeidx')}\big(\featurevec\big) \big]^{2}.
\end{equation}  

We can generalize \gls{gtvmin} by replacing $\normgeneric{\localparams{\nodeidx}- \localparams{\nodeidx'}}{2}^{2}$ 
in \eqref{equ_def_gtvmin} with the discrepancy $\discrepancy{\hypothesis^{(\nodeidx)}}{\hypothesis^{(\nodeidx')}}$  \eqref{equ_def_variation_non_parametric} (or the special case \eqref{equ_def_variation_non_parametric_sqerr}). This 
results in 
\begin{equation} 
	\label{equ_def_gtvmin_nonparam} 
	\big\{ \learntlocalhypothesis{\nodeidx} \big\}_{\nodeidx=1}^{\nrnodes}\!\in \hspace*{0mm}\argmin_{\substack{\localhypothesis{\nodeidx} \in \localmodel{\nodeidx}\\ \nodeidx\in \nodes}}
	\hspace*{0mm}\sum_{\nodeidx \in \nodes} \locallossfunc{\nodeidx}{\localhypothesis{\nodeidx}}\!+\!\regparam  \hspace*{0mm}\sum_{\edge{\nodeidx}{\nodeidx'} \in \edges} 
	\hspace*{0mm}\edgeweight_{\nodeidx,\nodeidx'}\discrepancy{\hypothesis^{(\nodeidx)}}{\hypothesis^{(\nodeidx')}}.
\end{equation}

\subsection{Interpretations} 
\label{sec_interpreations} 

We next discuss some interpretations of \gls{gtvmin} \eqref{equ_def_gtvmin_penalty}. 

{\bf Empirical Risk Minimization.} \Gls{gtvmin} \eqref{equ_def_gtvmin} is obtained as a 
special case of \gls{erm} \eqref{equ_def_erm} for specific choices for the \gls{model} 
$\hypospace$ and \gls{lossfunc} $\loss$. The \gls{model} (or \gls{hypospace}) used by \gls{gtvmin} 
is a product space generated by the \gls{localmodel}s at the nodes of an \gls{flnetwork}. 
The \gls{lossfunc} of \gls{gtvmin} consists of two parts: the sum of \gls{lossfunc}s at each 
node and a penalty term that measures the variation of \gls{localmodel}s across the edges 
of the \gls{flnetwork}. 

{\bf Generalized Convex Clustering.} One important special case of \gls{gtvmin} 
\eqref{equ_def_gtvmin_penalty} is convex clustering \cite{JMLR:v22:18-694,Pelckmans2005}. 
Indeed, convex clustering is obtained from \eqref{equ_def_gtvmin_penalty} using the local \gls{lossfunc} 
\begin{equation} 
	\label{equ_lossfunc_cvxclustering}
	\locallossfunc{\nodeidx}{\localparams{\nodeidx}} = \norm{\localparams{\nodeidx} - \va^{(\nodeidx)}}^{2}, \mbox{ for all nodes } \nodeidx \in \nodes
\end{equation} 
and the \gls{gtv} penalty function $\gtvpenalty(\vu) = \normgeneric{\vu}{p}$ with some $p \geq 1$.\footnote{Here, we used the $p$-norm $ \normgeneric{\vu}{p} \defeq \big( \sum_{\featureidx=1}^{\dimlocalmodel} |u_{\featureidx}|^{p} \big)^{1/p}$ of a vector $\vu \in \mathbb{R}^{\dimlocalmodel}$.} The vectors $\va^{(\nodeidx)}$, for $\nodeidx=1,\ldots,\nrnodes$, 
are the \gls{feature}s of \gls{datapoint}s that we wish to cluster in \eqref{equ_lossfunc_cvxclustering}. 
Thus, we can interpret \gls{gtvmin} as a generalization of convex clustering: we replace the 
terms $\norm{\localparams{\nodeidx} - \va^{(\nodeidx)}}^{2}$ with a more general local \gls{lossfunc}.

{\bf Dual of Minimum-Cost Flow Problem.} 
The optimization variables of \gls{gtvmin} \eqref{equ_def_gtvmin_penalty} are the local \glspl{modelparam} 
$\localparams{\nodeidx}$, for each node $\nodeidx \in \nodes$ in an \gls{flnetwork} $\graph$. 
The optimization of node-wise variables $\localparams{\nodeidx}$, for $\nodeidx=1,\ldots,\nrnodes$, 
is naturally associated with a dual problem \cite{RockNetworks}. This dual problem optimizes 
edge-wise variables $\localflowvec{\edge{\nodeidx}{\nodeidx'}}$, one for each 
edge $\edge{\nodeidx}{\nodeidx'} \in \edges$ of $\graph$,
\begin{align} 
	\label{equ_dual_nLasso} 
	\max_{\substack{\localflowvec{\edgeidx}, \edgeidx \in \edges\\ \localparams{\nodeidx}, \nodeidx \in \nodes}} &{- \sum_{\nodeidx \in \nodes} \conjlocallossfunc{\nodeidx}{\localparams{\nodeidx}} - 
		\regparam \sum_{\edgeidx \in \edges}   \edgeweight_{\edgeidx}  \gtvpenalty^{*}\big( \localflowvec{\edgeidx} /  ( \regparam  \edgeweight_{\edgeidx}) \big) } \\ 
	&{ \mbox{ subject to } - \localparams{\nodeidx}\!=\!\sum_{\substack{\edgeidx \in \edges \\  \edgeidx_{+}=\nodeidx}} \localflowvec{\edgeidx} - \sum_{\substack{\edgeidx \in \edges \\  \edgeidx_{-}=\nodeidx}}  \localflowvec{\edgeidx}}  \mbox{ for each } \nodeidx \in \nodes.
\end{align}
Here, we have introduced an orientation for each edge $\edgeidx\defeq\edge{\nodeidx}{\nodeidx'}$, 
by defining the \emph{head} $\edgeidx_{-}\defeq\min \{\nodeidx,\nodeidx'\}$ and the \emph{tail} 
$\edgeidx_{+}\defeq\max \{ \nodeidx,\nodeidx'\}$.\footnote{We use this orientation only for 
	notational convenience to formulate the dual of \gls{gtvmin}. The orientation of an edge (by choosing a head and tail) 
	has no practical meaning in terms of \gls{gtvmin}-based \gls{fl} algorithms. After all, \gls{gtvmin} \eqref{equ_def_gtvmin_penalty} 
	and its dual \eqref{equ_dual_nLasso} are defined for an \gls{flnetwork} with undirected edges $\edges$.}
Moreover, we used the convex conjugates $\conjlocallossfunc{\nodeidx}{\cdot},\gtvpenalty^{*}$ of 
the local \gls{lossfunc} $\locallossfunc{\nodeidx}{\cdot}$ and \gls{gtv} penalty function $\gtvpenalty$.\footnote{
	The convex conjugate of a function $f: \mathbb{R}^{\nrfeatures} \rightarrow \mathbb{R}$ is defined 
	as \cite{BoydConvexBook}
	\begin{equation} 
		\label{equ_def_conv_conj}
		f^{*} (\vx) \defeq \sup\limits_{\vz \in \mathbb{R}^{\dimlocalmodel}}  \vx^{T} \vz - f(\vz). 
		\vspace{-2mm}
\end{equation}}

\begin{figure} 
	\begin{center} 
		\begin{tikzpicture}
			\node[fill=black, circle, inner sep=2pt] (A) at (0, 0) {};
			\node[fill=black, circle, inner sep=2pt] (B) at (5, 0) {};
			
			\node[below,yshift=-5pt] at (A) {$\nodeidx$};
			\node[above,yshift=5pt] at (A) {$\localparams{\nodeidx}$};
			\node[below,yshift=-5pt] at (B) {$\nodeidx'$};
			\node[above,yshift=5pt] at (B) {$\localparams{\nodeidx'}$};
			
			\draw[-,line width=1pt] (A) -- (B) 
			node[midway, above] {$\localflowvec{\edgeidx}$} 
			node[midway, below] {$\edgeidx=\edge{\nodeidx}{\nodeidx'}$}; 
		\end{tikzpicture}
	\end{center} 
	\caption{Two nodes of an \gls{flnetwork} that are connected by an edge $\edgeidx=\edge{\nodeidx}{\nodeidx'}$. \Gls{gtvmin}
		\eqref{equ_def_gtvmin_penalty} optimizes local \glspl{modelparam} $\localparams{\nodeidx}$ for each 
		node $\nodeidx \in \nodes$ in the \gls{flnetwork}. The dual \eqref{equ_dual_nLasso} of \gls{gtvmin} optimizes 
		local \gls{modelparam} $\localflowvec{\edgeidx}$ for each edge $\edgeidx \in \edges$ in the \gls{flnetwork}.  }
\end{figure} 

The dual \gls{optproblem} \eqref{equ_dual_nLasso} generalizes the optimal flow problem \cite[Sec. 1J]{RockNetworks} 
to vector-valued flows. The special case of \eqref{equ_dual_nLasso}, obtained 
when the \gls{gtv} penalty function $\gtvpenalty$ is a norm, is equivalent to a generalized 
minimum-cost flow problem \cite[Sec. 1.2.1]{BertsekasNetworkOpt}. 
Indeed, the maximization problem \eqref{equ_dual_nLasso} is equivalent to the minimization 
\begin{align}
	\label{equ_def_duality_nLasso_edge_node_norm_gtv_pen_min_equiv}
	&\min_{\substack{\localflowvec{\edgeidx}, \edgeidx \in \edges\\ \localparams{\nodeidx}, \nodeidx \in \nodes}}   \sum_{\nodeidx \in \nodes} \conjlocallossfunc{\nodeidx}{\localparams{\nodeidx}}  \nonumber \\ 
	\mbox{ subject to } & - \localparams{\nodeidx}  =
	\sum_{\substack{\edgeidx \in \edges \\  \edgeidx_{+}=\nodeidx}} \localflowvec{\edgeidx} - \sum_{\substack{\edgeidx \in \edges \\ \edgeidx_{-}=\nodeidx }}  \localflowvec{\edgeidx} \mbox{ for each node } \nodeidx \in \nodes \nonumber \\
	& \| \localflowvec{\edgeidx} \|_{*}   \leq  \regparam  \edgeweight_{\edgeidx} \mbox{ for each edge } \edgeidx \in \edges. 
\end{align}
The optimization problem \eqref{equ_def_duality_nLasso_edge_node_norm_gtv_pen_min_equiv} reduces to 
the minimum-cost flow problem \cite[Eq. (1.3) - (1.5)]{BertsekasNetworkOpt} for scalar 
local \glspl{modelparam} $\localparams{\nodeidx} \in \mathbb{R}$. 

{\bf Locally Weighted Learning.} The solution of \gls{gtvmin} are local \glspl{modelparam} $\estlocalparams{\nodeidx}$ 
that tend to be clustered: Each node $\nodeidx \in \nodes$ belongs to a subset or cluster $\cluster \subseteq \nodes$. 
All the nodes in $\cluster$ have nearly identical local \glspl{modelparam}, $\estlocalparams{\nodeidx'} \approx \overline{\weights}^{(\cluster)}$ 
for all $\nodeidx' \in \cluster$ \cite{ClusteredFLTVMinTSP}. The cluster-wise \glspl{modelparam} $\overline{\weights}^{(\cluster)}$ are the solutions of 
\begin{equation} 
	\label{equ_def_opt_cluster}
	\min_{\weights} \sum_{\nodeidx' \in \cluster} \locallossfunc{\nodeidx'}{\weights}, 
\end{equation}  
which, in turn, is an instance of a locally weighted learning problem \cite[Sec.\ 3.1.2]{LocallyWeightedLearning}
\begin{equation} 
	\label{equ_def_locally_weighted_problem}
	\overline{\weights}^{(\cluster)} =  \argmin_{\vw \in \mathbb{R}^{\dimlocalmodel}} \sum_{\nodeidx' \in \nodes} \nodeweight{\nodeidx'} \locallossfunc{\nodeidx'}{\vw}.
\end{equation}
Indeed, we obtain \eqref{equ_def_opt_cluster} from \eqref{equ_def_locally_weighted_problem} by setting 
the weights $\nodeweight{\nodeidx'}$ equal to $1$ if $\nodeidx' \in \cluster$ and $0$ otherwise. 

\clearpage
\subsection{Exercises}

\refstepcounter{problem}\label{prob:maxeigvalLap}\textbf{\theproblem. Spectral Radius of Laplacian Matrix.} 
The spectral radius $\rho(\mQ)$ of a square matrix $\mQ$ is the largest magnitude of an \gls{eigenvalue}, 
$$\rho(\mQ) \defeq \max \{ |\eigvalgen|: \eigvalgen \mbox{ is an \gls{eigenvalue} of } \mQ\}.$$ 
Consider the \gls{LapMat} $\LapMat{\graph}$ of an \gls{flnetwork} with undirected \gls{graph} $\graph$.  
Show that $\rho\big( \LapMat{\graph} \big) = \eigval{\nrnodes}\big( \LapMat{\graph} \big)$ and verify 
the upper bound $\eigval{\nrnodes} \big(\LapMat{\graph} \big) \leq 2   \maxnodedegree^{(\graph)}$. 
Try to find a \gls{graph} $\graph$ such that $\eigval{\nrnodes} \big(\LapMat{\graph} \big) \approx 2   \maxnodedegree^{(\graph)}$.

\noindent\refstepcounter{problem}\label{prob:kernelLapMat}\textbf{\theproblem. Null Space of the \Gls{LapMat}.} 
Consider an undirected weighted \gls{graph} $\graph$ with $\nrnodes$ nodes, indexed by 
$\nodeidx = 1, \ldots, \nrnodes$. A component of $\graph$ is a subset $\cluster \subseteq \nodes$ 
of nodes such that every pair of nodes in $\cluster$ is connected by a path within $\cluster$, and there 
are no edges connecting nodes in $\cluster$ to nodes in $\nodes \setminus \cluster$. 
If $\graph$ is connected, it has a single component, namely the entire node set $\nodes$. 
Let $\LapMat{\graph}$ denote the \gls{LapMat} of $\graph$. The null space of $\LapMat{\graph}$ 
is the subspace $\mathcal{K} \subseteq \mathbb{R}^{\nrnodes}$ defined as
\[
\mathcal{K} := \left\{ \vv \in \mathbb{R}^{\nrnodes} \,\middle|\, \LapMat{\graph} \vv = \mathbf{0} \right\}.
\]
Show that the dimension of $\mathcal{K}$ is equal to the number of components in $\graph$.

\noindent\refstepcounter{problem}\label{prob:nullspace_visual}\textbf{\theproblem. Null-Space Visualization.} 
Consider an \gls{flnetwork} with $\nrnodes = 2$ nodes. Each node $\nodeidx \in \{1,2\}$ carries a 
scalar \gls{localmodel} \gls{parameter} $\weight^{(\nodeidx)} \in \mathbb{R}$. We can conveniently represent the 
local \glspl{modelparam} by stacking them into the \gls{vector}  
\[
\weights =\big( \weight^{(1)}, \weight^{(2)}\big)^{T} \in \mathbb{R}^2.
\]
This exercise requires you to visualize the subspace $\mathcal{S}$ \eqref{equ_def_subspace_constant_local} of $\mathbb{R}^2$. 
\begin{enumerate}
    \item Draw the subspace $\mathcal{S}$ as a line in the plane $\mathbb{R}^2$. Label your axes as 
	$\weight^{(1)}$ and $\weight^{(2)}$. Indicate at least two \glspl{vector} that belong to $\mathcal{S}$.
    \item Determine and draw the orthogonal complement $\mathcal{S}^\perp$ of $\mathcal{S}$. Label it 
	clearly in your drawing.
    \item Give a specific example of a non-zero \gls{vector} $\vb \in \mathcal{S}^\perp$, and verify that it is 
	orthogonal to an arbitrary \gls{vector} $\va \in \mathcal{S}$ by computing the inner product $\vb^\top \va$.
\end{enumerate}

\noindent\refstepcounter{problem}\label{prob:specclusttoy}\textbf{\theproblem. Toy Example of Spectral Clustering.} 
Consider the \gls{graph} $\graph$ depicted in Figure \ref{fig_specclusttoy}. 
\vspace*{2mm}
\begin{figure}[htbp]
	\begin{center}
		\begin{tikzpicture}
			\node[draw, circle, fill=black, minimum size=5pt, inner sep=0pt] (a) at (0, 0) {};
			\node[draw, circle, fill=black, minimum size=5pt, inner sep=0pt] (b) at (-1, -1) {};
			\node[draw, circle, fill=black, minimum size=5pt, inner sep=0pt] (c) at (1, -1) {};
			
			\draw (a) -- (b);
			\draw (b) -- (c);
			
			\node [xshift=1pt,anchor=north] at (c) {$\nodeidx\!=\!1$};
			\node [xshift=1pt,anchor=north] at (b) {$2$};
			\node [xshift=1pt,anchor=north] at (a) {$3$};
			
			\node[draw, circle, fill=black, minimum size=5pt, inner sep=0pt] (d) at (4, 0) {};
			\node[draw, circle, fill=black, minimum size=5pt, inner sep=0pt] (e) at (3, -1) {};
			\node[draw, circle, fill=black, minimum size=5pt, inner sep=0pt] (f) at (5, -1) {};
			
			\node [xshift=1pt,anchor=north] at (d) {$4$};
			\node [xshift=1pt,anchor=north] at (e) {$5$};
			\node [xshift=1pt,anchor=north] at (f) {$6$};
			
			\draw (d) -- (e);
			\draw (e) -- (f);
			\draw (f) -- (d);
			
			
			\node at (0, 0.5) {component $\cluster^{(1)}$};
			\node at (4, 0.5) {component $\cluster^{(2)}$};
		\end{tikzpicture}
		\vspace*{-3mm}
	\end{center}
	\caption{An undirected \gls{graph} $\graph$ that consists of two connected components $\cluster^{(1)},\cluster^{(2)}$.\label{fig_specclusttoy}}
	\vspace*{-3mm}
\end{figure} 
The Laplacian matrix has two zero \gls{eigenvalue}s $\eigval{1}\!=\!\eigval{2}\!=\!0$. 
Can you find corresponding orthonormal \gls{eigenvector}s $\vu^{(1)}, \vu^{(2)}$? Are they unique?

\noindent\refstepcounter{problem}\label{prob:addegeconnect}\textbf{\theproblem. Adding an Edge Increases Connectivity.} 
Consider an undirected weighted \gls{graph} $\graph$ with \gls{LapMat} $\LapMat{\graph}$. We construct 
a new \gls{graph} $\graph'$, with \gls{LapMat} $\LapMat{\graph'}$, by adding a new edge to $\graph$. 
Show that $\eigval{2}(\graph') \geq \eigval{2}(\graph)$, i.e., the second smallest \gls{eigenvalue} 
of $\LapMat{\graph'}$ is at least as large as the second smallest \gls{eigenvalue} 
of $\LapMat{\graph}$.

\noindent\refstepcounter{problem}\label{prob:flnetworkcapacity}\textbf{\theproblem. Capacity of an \gls{flnetwork}.} 
Consider the \gls{flnetwork} shown in Figure \ref{fig_cap_fl_networks}. Each node 
holds a \gls{localdataset}, with its size indicated by the adjacent numbers. The 
\glspl{device} communicate over bi-directional links, whose capacities are specified 
by the numbers next to the edges. What is the minimum time required for the 
leftmost node to collect all \gls{localdataset}s from the other nodes?
\begin{figure}
	\begin{center}
		\begin{tikzpicture}[
			node/.style={circle, draw=black, fill=black, minimum size=0.3cm, text centered},
			link/.style={draw=black, -latex}
			]
			
			\node[node] (A) at (0, 0) {};
			\node[anchor=east] at (A.west) {$10$ MB};
			
			\node[node] (B) at (3, 1) {};
			\node[anchor=south] at (B.north) {$45$};
			
			\node[node] (C) at (6, 0) {};
			\node[anchor=west] at (C.east) {$33$};
			
			\node[node] (D) at (2, -2.5) {};
			\node[anchor=north] at (D.south) {$5$};
			
			\node[node] (E) at (5, -2.5) {};
			\node[anchor=north] at (E.south) {$10$};
			
			\draw[] (A) -- node[above] {$1$ kbps} (B);
			\draw[] (B) -- node[above] {2} (C);
			\draw[] (A) -- node[left] {1} (D);
			\draw[] (D) -- node[below] {2} (E);
			\draw[] (C) -- node[right] {1} (E);
			
		\end{tikzpicture}
	\end{center}
	\caption{An \gls{flnetwork} whose nodes $\nodeidx=1,\ldots,5$ represent \glspl{device} that hold 
		\glspl{localdataset} whose size is indicated next to each node. \label{fig_cap_fl_networks}}
\end{figure}

\noindent\refstepcounter{problem}\label{prob:discrepancyterm}\textbf{\theproblem. Discrepancy Measures.} 
Consider an \gls{flnetwork} with nodes carrying parametric \gls{localmodel}s, each 
having \glspl{modelparam} $\localparams{\nodeidx} \in \mathbb{R}^{\dimlocalmodel}$. Is it 
possible to construct a \gls{dataset} $\dataset^{\edge{\nodeidx}{\nodeidx'}}$ such that \eqref{equ_def_variation_non_parametric_sqerr} 
coincides with $\normgeneric{\localparams{\nodeidx}- \localparams{\nodeidx'}}{2}^{2}$?

\noindent\refstepcounter{problem}\label{prob:privfriendlydiscr}\textbf{\theproblem. Privacy-Friendly Discrepancy Measures.} 
The discrepancy measure \eqref{equ_def_variation_non_parametric} requires to choose a 
test-set $\dataset^{\edge{\nodeidx}{\nodeidx'}}$. One possible choice is to combine \gls{datapoint}s  
of the \gls{localdataset}s $\localdataset{\nodeidx}$ and $\localdataset{\nodeidx'}$. However, 
sharing these \glspl{datapoint} can be harmful as they potentially leak sensitive information. 
Could you think of a simple message passing protocol between node $\nodeidx$ and $\nodeidx'$ 
that allows them to evaluate \eqref{equ_def_variation_non_parametric} only by sharing the \gls{prediction}s $\hypothesis^{(\nodeidx)}(\featurevec),\hypothesis^{(\nodeidx')}(\featurevec)$ for $\featurevec \in \dataset^{\edge{\nodeidx}{\nodeidx'}}$?

\noindent\refstepcounter{problem}\label{prob:structuregtvmin}\textbf{\theproblem. Structure of \gls{gtvmin}.} 
What are sufficient conditions for the \glspl{localdataset} and the \glspl{edgeweight} used in \gls{gtvmin} 
such that the matrix $\mQ$ in \eqref{equ_def_mQ_gtvmin_lr} is invertible?

\noindent\refstepcounter{problem}\label{prob:uniquegtvmin}\textbf{\theproblem. Existence and Uniqueness of \gls{gtvmin} Solution.} 
Consider the \gls{gtvmin} instance \eqref{equ_def_gtvmin}, defined over an \gls{flnetwork} with the weighted undirected 
graph $\graph$. 
\begin{enumerate} 
	\item {\bf Existence.} Can you state a sufficient condition on the local \gls{lossfunc}s 
	and the weighted edges of $\graph$ such that \eqref{equ_def_gtvmin} has at least one solution? 
	\item {\bf Uniqueness.} Then, try to find a condition that ensures that \eqref{equ_def_gtvmin} 
	has a unique solution. 
	\item Finally, try to find necessary conditions for the existence and uniqueness of 
	solutions to \eqref{equ_def_gtvmin}. 
\end{enumerate} 

\noindent\refstepcounter{problem}\label{prob:meangtvmin}\textbf{\theproblem. Computing the Average.} 
Consider an \gls{flnetwork} with each nodes carrying a single model parameter $\weight^{(\nodeidx)}$ 
and a \gls{localdataset}, consisting of a single number $\truelabel^{(\nodeidx)}$. Construct an instance of \gls{gtvmin} such 
that its solutions are given by $\widehat{\weight}^{(\nodeidx)} \approx (1/\nrnodes) \sum_{\nodeidx=1}^{\nrnodes} \truelabel^{(\nodeidx)}$ 
for all $\nodeidx = 1,\ldots,\nrnodes$.

\noindent\refstepcounter{problem}\label{prob:meangtvminstar}\textbf{\theproblem. Computing the Average over a Star.} 
Consider the \gls{flnetwork} depicted in Figure \ref{fig_star_avg_problem}, which consists of 
a centre node $\nodeidx_{0}$ which is connected to $\nrnodes-1$ peripheral nodes $\mathcal{P} \defeq \nodes \setminus \{ \nodeidx_{0}\}$.  
Each peripheral node $\nodeidx \in \mathcal{P}$ carries a \gls{localdataset} that consists of 
a single real-valued observation $\truelabel^{(\nodeidx)} \in \mathbb{R}$. 
Construct an instance of \gls{gtvmin}, using real-valued local \glspl{modelparam} $\weight^{(\nodeidx)} \in \mathbb{R}$, 
such that the solution satisfies $\widehat{\weight}^{(\nodeidx_{0})} \approx (1/(\nrnodes-1)) \sum_{\nodeidx \in \mathcal{P}} \truelabel^{(\nodeidx)}$. 
\begin{figure}[h]
	\begin{center}
		\begin{tikzpicture}[node distance=2cm, every node/.style={circle, fill, draw, minimum size=0.2cm}]
			\node (center) at (0, 0) {};
			\node (node1) at (-4, 2) {};
			\node (node2) at (-2, 2) {};
			\node (node3) at (0, 2) {};
			\node (node4) at (2, 2) {};
			\node (node5) at (4, 2) {};
			\draw[thick] (center) -- (node1);
			\draw[thick] (center) -- (node2);
			\draw[thick] (center) -- (node3);
			\draw[thick] (center) -- (node4);
			\draw[thick] (center) -- (node5);
			\node[below=0.01cm of center,draw=none, fill=none] {$\nodeidx_{0}$};
			\node[above=0.01cm of node1,draw=none, fill=none, yshift=-5pt] {$\nodeidx \in \mathcal{P}$};
		\end{tikzpicture}
	\end{center}
	\vspace*{-7mm}
	\caption{An \gls{flnetwork} that consists of a centre node $\nodeidx_{0}$ that is connected to 
		several peripheral nodes $\mathcal{P} \defeq \nodes \setminus \{ \nodeidx_{0} \}$. \label{fig_star_avg_problem}}
\end{figure} 

\noindent\refstepcounter{problem}\label{prob:fundamentallimits}\textbf{\theproblem. Fundamental Limits.} 
Consider the \gls{flnetwork} depicted in Figure \ref{fig_fund_limits_tree}. Each node carries a \gls{localmodel} 
with single parameter $\weight^{(\nodeidx)}$ as well as a \gls{localdataset} that 
consists of a single number $\truelabel^{(\nodeidx)}$. We use a \gls{probmodel} for the \gls{localdataset}s: 
$\truelabel^{(\nodeidx)}= \bar{w} + n^{(\nodeidx)}$. Here, $\bar{w}$ is some fixed but unknown number and 
$n^{(\nodeidx)} \sim \mathcal{N}(0,1)$ are \gls{iid} Gaussian \gls{rv}s. We use a message-passing \gls{fl} algorithm 
to estimate $c$ based on the \gls{localdataset}s. What is a fundamental limit on the accuracy of the 
estimate $\hat{c}^{(\nodeidx)}$ delivered at some fixed node $\nodeidx$ by such an algorithm after two iterations? 
Compare this limit with the risk $\expect\big\{ \big(\hat{w}^{(\nodeidx)} - \bar{w}  \big)^{2} \big\}$ 
incurred by the estimate $\hat{w}^{(\nodeidx)}$ delivered by running Algorithm \ref{alg_fed_gd} for 
two iterations. 
\begin{figure} 
	\begin{center} 
		\begin{tikzpicture}
			\def\RadiusA{2}  
			\def\RadiusB{4}  
			\node[circle, draw, fill=black, label=above:{$\nodeidx$}] (root) at (0,0) {};
			\foreach \i in {1,2,3} {
				\node[circle, draw, fill=black] (A\i) at ({\i*120}: \RadiusA) {};
				\draw (root) -- (A\i);
			}
			\foreach \i in {1,2,3} {
				\foreach \j in {1,2} {
					\node[circle, draw, fill=black] (B\i\j) at ({\i*120+30*\j-45}:\RadiusB) {};
					\draw (A\i) -- (B\i\j);
				}
			}
			\draw[dashed, gray] (0,0) circle (\RadiusA); 
			\draw[dashed, gray] (0,0) circle (\RadiusB); 
		\end{tikzpicture}
	\end{center}
	\caption{An \gls{flnetwork} containing a node $\nodeidx$ with \gls{nodedegree} $\nodedegree{\nodeidx} = 3$, 
		like all its \glspl{neighbor} $\nodeidx' \in \neighbourhood{\nodeidx}$. 
		We use an \gls{fl} \gls{algorithm} to learn local \glspl{modelparam} $\weight^{(\nodeidx)}$. If the 
		\gls{algorithm} employs message passing, the first iteration provides access only to the \glspl{localdataset} 
		of the \glspl{neighbor} in $\neighbourhood{\nodeidx}$ (located along the inner dashed circle). 
		In the second iteration, the algorithm gains access to the \glspl{localdataset} of the \glspl{neighbor} 
		$\neighbourhood{\nodeidx'}$ of each $\nodeidx' \in \neighbourhood{\nodeidx}$. 
		These \emph{second-hop} \glspl{neighbor} are located along the outer dashed circle. \label{fig_fund_limits_tree}
	} 
\end{figure} 

\noindent\refstepcounter{problem}\label{prob:countpaths}\textbf{\theproblem. Counting Number of Paths.} 
Consider an undirected \gls{graph} $\graph$ with each edge $\edge{\nodeidx}{\nodeidx'} \in \edges$ 
having unit edge weight $\edgeweight_{\nodeidx,\nodeidx'} = 1$. A 
$k$-hop path, for some $k \in \{1,2,.\ldots\}$, between two nodes $\nodeidx,\nodeidx' \in \nodes$ 
is a node sequence $\nodeidx^{(1)},\ldots,\nodeidx^{(k+1)}$ such that $\nodeidx^{(1)}=\nodeidx$, 
$\nodeidx^{(k+1)}=\nodeidx'$, and $\edge{\nodeidx^{(r)}}{\nodeidx^{(r+1)}} \in \edges$ for reach $r=1,\ldots,k$. 
Show that the number of $k$-hop paths between two nodes $\nodeidx,\nodeidx' \in \nodes$ is 
given by $\big( \mathbf{A}^{k} \big)_{\nodeidx,\nodeidx'}$.

\noindent\refstepcounter{problem}\label{prob:proxopcvxquad}\textbf{\theproblem. \Gls{proxop} of a \gls{quadfunc}.} 
Study the \gls{proxop} \eqref{equ_def_prox_op_original} for a \gls{quadfunc}, 
$$\locallossfunc{\nodeidx}{\localparams{\nodeidx}} = \big( \localparams{\nodeidx}\big)^{T} \mQ \localparams{\nodeidx} + \vq^{T} \localparams{\nodeidx} + q,$$
with some matrix $\mQ \in \mathbb{R}^{\dimlocalmodel \times \dimlocalmodel}$, 
vector $\vq \in \mathbb{R}^{\dimlocalmodel}$ and number $q \in \mathbb{R}$.

\newpage
\subsection{Proofs} 
\subsubsection{Proof of Proposition \ref{prop_second_component_error}} 
\label{sec_proof_second_comp_error}
Let us introduce the shorthand $f\big( \weights^{(\nodeidx)} \big)$ for the \gls{objfunc} of 
the \gls{gtvmin} instance \eqref{equ_def_gtvmin_linreg}. We verify the bound \eqref{equ_upper_bound_variation_component} 
by showing that if it does not hold, the choice of the \gls{localmodel} \glspl{parameter} $\weights^{(\nodeidx)} \defeq \overline{\weights}^{(\nodeidx)}$ 
(see \eqref{equ_def_probmodel_linreg_node_i}) results in a smaller \gls{objfunc} value, 
$f\big( \overline{\weights}^{(\nodeidx)} \big) < f \big( \widehat{\weights}^{(\nodeidx)} \big)$. This would 
contradict the fact that $\widehat{\weights}^{(\nodeidx)}$ is a solution to \eqref{equ_def_gtvmin_linreg}. 

First, note that 
\begin{align}
	f\big( \overline{\weights}^{(\nodeidx)} \big)  &= 	 \sum_{\nodeidx \in \nodes} (1/\localsamplesize{\nodeidx}) \normgeneric{\labelvec^{(\nodeidx)} \!-\! \featuremtx^{(\nodeidx)} \overline{\weights}^{(\nodeidx)}}{2}^{2}\!+\!\regparam \sum_{\edge{\nodeidx}{\nodeidx'} \in \edges} 
	\edgeweight_{\nodeidx,\nodeidx'}  \normgeneric{\overline{\weights}^{(\nodeidx)} \!-\!\overline{\weights}^{(\nodeidx')}}{2}^{2} \nonumber \\
	& \stackrel{\eqref{equ_def_constant_linmodel_params}}{=}  \sum_{\nodeidx \in \nodes} (1/\localsamplesize{\nodeidx}) \normgeneric{\labelvec^{(\nodeidx)} \!-\! \featuremtx^{(\nodeidx)} \overline{\weights}^{(\nodeidx)}}{2}^{2} \nonumber \\ 
	& \stackrel{\eqref{equ_def_probmodel_linreg_node_i}}{=}  \sum_{\nodeidx \in \nodes} (1/\localsamplesize{\nodeidx}) \normgeneric{\featuremtx^{(\nodeidx)} \overline{\weights}^{(\nodeidx)}\!+\!{\bm \varepsilon}^{(\nodeidx)}\!-\! \featuremtx^{(\nodeidx)} \overline{\weights}^{(\nodeidx)}}{2}^{2} \nonumber \\ 
	& = \sum_{\nodeidx \in \nodes} (1/\localsamplesize{\nodeidx}) \normgeneric{{\bm \varepsilon}^{(\nodeidx)}}{2}^{2}. \label{equ_norm_error_func_value}
\end{align} 

Inserting \eqref{equ_decompose_sol_gtvmin} into \eqref{equ_def_gtvmin_linreg}, 
\begin{align}
	\label{equ_obj_function_gtvmin_average_ac_component}
	f \big( \widehat{\weights}^{(\nodeidx)} \big) & =	\underbrace{ \sum_{\nodeidx \in \nodes} (1/\localsamplesize{\nodeidx}) \normgeneric{\labelvec^{(\nodeidx)} \!-\! \featuremtx^{(\nodeidx)}  \widehat{\weights}^{(\nodeidx)}}{2}^{2}}_{\geq 0}\!+ \regparam \sum_{\edge{\nodeidx}{\nodeidx'} \in \edges} 
	\hspace*{-2mm}	\edgeweight_{\nodeidx,\nodeidx'}  \underbrace{\normgeneric{\widehat{\weights}^{(\nodeidx)} \!-\!\widehat{\weights}^{(\nodeidx')}}{2}^{2}}_{\stackrel{\eqref{equ_decompose_sol_gtvmin}}{=}\normgeneric{\widetilde{\weights}^{(\nodeidx)} \!-\!\widetilde{\weights}^{(\nodeidx')}}{2}^{2}}  \nonumber \\ 
	& \geq \regparam \sum_{\edge{\nodeidx}{\nodeidx'} \in \edges} 
	\edgeweight_{\nodeidx,\nodeidx'}  \normgeneric{\widetilde{\weights}^{(\nodeidx)} \!-\!\widetilde{\weights}^{(\nodeidx')}}{2}^{2}  \nonumber \\ 
	& \stackrel{\eqref{equ_lower_bound_tv_eigval}}{\geq} \regparam \eigval{2} \sum_{\nodeidx=1}^{\nrnodes}\normgeneric{\widetilde{\weights}^{(\nodeidx)}}{2}^{2}.
\end{align}
If the bound \eqref{equ_upper_bound_variation_component} would not hold, then 
by \eqref{equ_obj_function_gtvmin_average_ac_component} and \eqref{equ_norm_error_func_value} 
we would obtain $f \big( \widehat{\weights}^{(\nodeidx)} \big) > f \big( \overline{\weights}^{(\nodeidx)} \big)$. 
This is a contradiction to the fact that $\widehat{\weights}^{(\nodeidx)}$ solves \eqref{equ_def_gtvmin_linreg}.

 \newpage
.\section{Gradient Methods for Federated Optimization} 
\label{lec_gradientmethods}

Chapter \ref{lec_fldesignprinciple} introduced \gls{gtvmin} as a central design principle 
for \gls{fl} \glspl{algorithm}. Many important instances of \gls{gtvmin} require the minimization 
of a smooth \gls{objfunc} $f(\weights)$ over a continuous parameter space. This chapter 
investigates how \glspl{gdmethod} --a broadly used family of iterative \glspl{optmethod} -- 
can be employed to solve such problems. These methods rely on local approximations of 
$f(\weights)$ using its \gls{gradient}.

Section \ref{sec_gd} introduces the basic \gls{gradstep} and explains how it updates 
\glspl{modelparam} in the direction of steepest descent. Key considerations such as the 
choice of the \gls{learnrate} are discussed in Section \ref{sec_learn_rate}, along with
\glspl{stopcrit} in Section \ref{sec_stopping_criterion} that help determine when to terminate 
the optimization process. Section \ref{sec_perturbed_gd} studies how perturbations affect 
the convergence of \glspl{gradstep}, which is particularly relevant in \gls{fl} applications that 
involve unreliable communication or partial \gls{data} access.

When optimization problems include explicit constraints on the \glspl{modelparam}, 
\gls{projgd} presented in Section \ref{sec_proj_gradient_descent} provides a principled 
solution. Section \ref{sec_gen_gradient_step} then extends \glspl{gdmethod} to non-parametric \glspl{model}, 
using \glspl{proxop} and test \glspl{dataset} to generalize the notion of a \gls{gradstep}. 
Finally, Section \ref{sec_fixedpoint_interpretation} interprets \glspl{gdmethod} as a special case 
of \glspl{fixedpointiter}. This perspective allows for a unified understanding of \gls{fl} \glspl{algorithm} 
as convergent processes driven by contraction operators.

\subsection{Gradient Descent} 
\label{sec_gd}

\Glspl{gdmethod} are iterative \glspl{algorithm} for finding the minimum of a \gls{differentiable} \gls{objfunc}  
$f(\weights)$ of a vector-valued argument $\weights$. One example of such an \gls{optproblem} is the \gls{erm} 
instance \eqref{equ_def_param_erm_LR}. Unless stated otherwise, we consider an \gls{objfunc} of the form:
\begin{equation} 
	\label{equ_generic_cvx_quad_function}
	f(\weights) \defeq \weights^{T} \mQ \weights + \vq^{T} \weights.
\end{equation}  
Although restricting our discussion to \glspl{objfunc} of the form \eqref{equ_generic_cvx_quad_function} 
may seem limiting, this formulation allows for a straightforward analysis and generalization to 
larger classes of \gls{differentiable} \glspl{function}. Moreover, we can use 
\eqref{equ_generic_cvx_quad_function} also as an approximation for broader families of \glspl{objfunc}.

Note that \eqref{equ_generic_cvx_quad_function} defines an entire family 
of \gls{convex} \glspl{quadfunc} $f(\weights)$. Each member of this family is 
specified by a \gls{psd} matrix $\mQ \in \mathbb{R}^{\dimlocalmodel \times \dimlocalmodel}$ 
and a vector $\vq \in \mathbb{R}^{\dimlocalmodel}$.  
We have already encountered some \gls{ml} and \gls{fl} methods that minimize an \gls{objfunc} of 
the form \eqref{equ_generic_cvx_quad_function}: \Gls{linreg} \eqref{equ_def_param_erm_LR} 
and \gls{ridgeregression} \eqref{equ_def_ridge_reg_quadratidform} in Chapter \ref{lec_mlbasics} 
as well as \gls{gtvmin} \eqref{equ_def_gtvmin_linreg} for local \gls{linmodel}s in Chapter \ref{lec_fldesignprinciple}. 
Moreover, \eqref{equ_generic_cvx_quad_function} is a useful approximation for 
the \glspl{objfunc} arising in other \gls{ml} methods \cite{NEURIPS2018_5a4be1fa,DBLP:conf/iclr/DuZPS19,E:2020aa}. 

Given a current choice of \glspl{modelparam} $\weights^{(\iteridx)}$, we want to update them towards 
a minimum of \eqref{equ_generic_cvx_quad_function}. To this end, we use the \gls{gradient} $\nabla f \big(\weights^{(\iteridx)} \big)$ 
to locally approximate $f(\weights)$ (see Figure \ref{fig_smooth_function}). 
The \gls{gradient} $\nabla f \big(\weights^{(\iteridx)} \big)$ indicates the direction in 
which the function $f(\weights)$ maximally increases. Therefore, it seems reasonable to 
update $\weights^{(\iteridx)}$ in the opposite direction of $\nabla f \big(\weights^{(\iteridx)} \big)$, 
\begin{align}
	\label{equ_def_basic_gradstep}
	\weights^{(\iteridx+1)} & \defeq \weights^{(\iteridx)} - \lrate \nabla f\big( \weights^{(\iteridx)} \big) \nonumber \\  
	& \stackrel{\eqref{equ_generic_cvx_quad_function}}{=}  \weights^{(\iteridx)} - \lrate \big( 2 \mQ \weights^{(\iteridx)} + \vq  \big). 
\end{align}
The \gls{gradstep} \eqref{equ_def_basic_gradstep} involves the positive factor $\lrate>0$ which 
we refer to as \gls{stepsize} or \gls{learnrate}. Algorithm \ref{alg_basic_gdmethods} summarizes 
the most basic variant of \glspl{gdmethod}, which simply iterates \eqref{equ_def_basic_gradstep} 
until a predefined \gls{stopcrit} is met.

\begin{figure}[htbp]
	\begin{center}
		\begin{tikzpicture}[scale=0.9]
			\node [right] at (-4.1,1.7) {$f(\weights)$} ;
			\draw[ultra thick, domain=-4.1:4.1] plot (\x,  {(1/4)*\x*\x});
			\draw[dashed, thick, domain=1:3.6] plot (\x,  {\x - 1}) node[right] {$ f\big(\weights^{(\iteridx)}\big)\!+\!\big(\weights\!-\!\weights^{(\iteridx)}\big)^{T} \nabla f\big(\weights^{(\iteridx)}\big)$};
			\draw [fill] (2,1) circle [radius=0.1] node[right] {$f\big(\weights^{(\iteridx)}\big)$};
			\draw[thick,->] (3,2) -- (3.5,1.5) node[right] {$\mathbf{n}$};
		\end{tikzpicture}
	\end{center}
	\caption{We can approximate a \gls{differentiable} function $f(\weights)$ locally around a point 
		$\weights^{(\iteridx)} \in \mathbb{R}^{\nrfeatures}$ using the linear function $f\big(\weights^{(\iteridx)}\big)\!+\!\big(\weights\!-\!\weights^{(\iteridx)}\big)^{T} \nabla f\big(\weights^{(\iteridx)}\big)$. 
		Geometrically, we approximate the graph of $f(\weights)$ by a hyperplane with normal vector 
		$\mathbf{n} = (\nabla f\big(\weights^{(\iteridx)}\big),-1)^{T} \in \mathbb{R}^{\nrfeatures+1}$ 
		of this approximating hyperplane is determined by the \gls{gradient} $\nabla f\big(\weights^{(\iteridx)}\big)$ \cite{RudinBookPrinciplesMatheAnalysis}.}
	\label{fig_smooth_function}
\end{figure}
The usefulness of \glspl{gdmethod} crucially depends on the computational complexity of 
evaluating the \gls{gradient} $\nabla f(\weights)$. Modern software libraries for automatic 
differentiation enable the efficient evaluation of the \glspl{gradient} arising in widely-used 
\gls{erm}-based methods \cite{PyTorchNeurips}. 

Besides the actual computation of the \gls{gradient}, it might already be challenging to 
gather the required \glspl{datapoint} which define the \gls{objfunc} $f(\weights)$ (e.g., being 
the average \gls{loss} over a large \gls{trainset}). Indeed, the matrix $\mQ$ and vector 
$\vq$ in \eqref{equ_generic_cvx_quad_function} are constructed from the \glspl{feature} 
and \glspl{label} of \glspl{datapoint} in the \gls{trainset}. For example, 
the \gls{gradient} of the \gls{objfunc} in \gls{ridgeregression} \eqref{equ_def_ridge_reg_quadratidform} is 
\begin{equation} 
	\nonumber
	\nabla f(\weights) =- (2/\samplesize) \sum_{\sampleidx=1}^{\samplesize} \featurevec^{(\sampleidx)} \big( \truelabel^{(\sampleidx)} - \weights^{T} \featurevec^{(\sampleidx)} \big) + 2 \regparam \weights.
\end{equation} 
Evaluating this \gls{gradient} requires roughly $\dimlocalmodel \times \samplesize$ arithmetic 
operations such as adding and multiplying numbers. 

\begin{algorithm}[htbp]
	\caption{A blueprint for \glspl{gdmethod}}\label{alg_basic_gdmethods}
	\begin{algorithmic}[1]
		\renewcommand{\algorithmicrequire}{\textbf{Input:}}
		\renewcommand{\algorithmicensure}{\textbf{Output:}}
		\Require some \gls{objfunc} $f(\weights)$ (e.g., the average \gls{loss} of a \gls{hypothesis} $\hypothesis^{(\weights)}$ on a \gls{trainset}); \gls{learnrate} $\lrate >0$; some \gls{stopcrit}.
		\Statex\hspace{-6mm}{\bf Initialize:} set $\weights^{(0)}\!\defeq\!\mathbf{0}$; set iteration counter $\iteridx\!\defeq\!0$   
		\Repeat 
		\State $\iteridx \defeq \iteridx\!+\!1$ (increase iteration counter) 
		\State  $\weights^{(\iteridx)} \defeq \weights^{(\iteridx\!-\!1)} - \lrate \nabla f\big(\weights^{(\iteridx\!-\!1)}\big)$ 
		(do a \gls{gradstep} \eqref{equ_def_basic_gradstep}) \label{step_grad_step}
		\Until \gls{stopcrit} is met 
		\Ensure learned \glspl{modelparam} $\widehat{\weights} \defeq \weights^{(\iteridx)}$ (hopefully $f\big(\widehat{\weights} \big) \approx \min_{\weights} f(\weights)$) 
	\end{algorithmic}
\end{algorithm}
Like most other \glspl{gdmethod}, Algorithm \ref{alg_basic_gdmethods},
involves two hyper-parameters: (i) the \gls{learnrate} $\lrate$ used for the \gls{gradstep} 
and (ii) a \gls{stopcrit} to decide when to stop repeating the \gls{gradstep}. We next 
discuss how to choose these hyper-parameters. 

Note that we can apply Algorithm \ref{alg_basic_gdmethods} to find the minimum of 
any \gls{differentiable} \gls{objfunc} $f(\weights)$. Indeed, Algorithm \ref{alg_basic_gdmethods} 
only needs to be able to access the \gls{gradient} $\nabla f\big(\weights^{(\iteridx\!-\!1)}\big)$. 
In particular, we an apply Algorithm \ref{alg_basic_gdmethods} to \glspl{objfunc} that do not 
belong to the family of \gls{convex} \glspl{quadfunc} \eqref{equ_generic_cvx_quad_function}.

\subsection{How to Choose the Learning Rate}
\label{sec_learn_rate}

The \gls{learnrate} must be chosen carefully: if it is too large, the \gls{gradstep} 
may overshoot and diverge from the solution of \eqref{equ_generic_cvx_quad_function}; 
if it is too small, each step makes only negligible progress. Note that practical \gls{fl} 
systems can only afford to compute a finite number \glspl{gradstep}. Therefore, we must 
ensure that each \gls{gradstep} makes a sufficiently large progress towards the optimum of 
the \gls{objfunc}. Figure \ref{fig_small_large_lrate} 
illustrates both extremes.

\begin{figure}[hbtp]
	\begin{center}
		\begin{minipage}{0.45\columnwidth}
			\begin{tikzpicture}[xscale=0.4,yscale=0.6]
				\draw[blue, ultra thick, domain=-4.1:4.1] plot (\x,  {(1/4)*\x*\x});
				\draw[] (1,0.25) circle [radius=0.1] node [right] (A) {$f(\weights^{(\iteridx)})$} ;
				\draw[] (-2,1) circle [radius=0.1] node [left] (B) {$f(\weights^{(\iteridx\!+\!1)})$} ;
				\draw[] (3,2.25) circle [radius=0.1]  node  [right] (C) {$f(\weights^{(\iteridx\!+\!2)})$} ;
				\draw[->,dashed] (-2,1) -- (3,2.25) node [midway,above] {\eqref{equ_def_basic_gradstep}};
				\draw[->,dashed]  (1,0.25) -- (-2,1) node [midway,above] {\eqref{equ_def_basic_gradstep}};
				\node [below] at (0,-0.2) {(a)};
			\end{tikzpicture}
		\end{minipage}
		\begin{minipage}{0.45\columnwidth}
			\begin{tikzpicture}[xscale=0.4,yscale=0.6]
				\draw[blue, ultra thick, domain=-4.1:4.1] plot (\x,  {(1/4)*\x*\x});
				\draw[] (4,4) circle [radius=0.1];
				\node [right] at (4,4) {$f(\weights^{(\iteridx)})$};
				\draw[] (3.8,3.61) circle [radius=0.1];
				\node [left] at (3.8,3.61) {$f(\weights^{(\iteridx\!+\!1)})$};
				\draw[] (3.65,3.33) circle [radius=0.1];
				\node [right] at (3.65,3.33) {$f(\weights^{(\iteridx\!+\!2)})$};
				\node [below] at (0,-0.2) {(b)};
			\end{tikzpicture}
		\end{minipage}
	\end{center}
	\caption{Effect of using an inadequate \glspl{learnrate} $\lrate$ in the \gls{gradstep} \eqref{equ_def_basic_gradstep}. 
		(a) If $\lrate$ is too large, the \glspl{gradstep} can ``overshoot'' such that 
		the iterates $\weights^{(\iteridx)}$ diverge away from the optimum, i.e., $f(\weights^{(\iteridx\!+\!1)}) > f(\weights^{(\iteridx)})$! 
		(b) If $\lrate$ is too small, the \glspl{gradstep} make too little progress towards 
		the optimum within the available number of iterations (due to limited computational budget). 
	}
	\label{fig_small_large_lrate}
\end{figure}

One approach to choosing the \gls{learnrate} is to start with some initial value (first guess) 
and monitor the decrease in the \gls{objfunc}. If this decrease does not agree with the 
decrease predicted by the (local linear approximation using the) \gls{gradient}, we 
decrease the \gls{learnrate} by a constant factor. After we decrease the \gls{learnrate}, 
we re-consider the decrease in the \gls{objfunc}. We repeat this procedure until a sufficient 
decrease in the \gls{objfunc} is achieved \cite[Sec 6.1]{CvxAlgBertsekas}. 

Alternatively, we can use a prescribed sequence (schedule) $\lrate_{\iteridx}$, for $\iteridx=1,2,\ldots,$ 
of \glspl{learnrate} that vary across successive \gls{gradstep}s \cite{Schaul2013}. 
For example, we could require the \gls{learnrate} $\lrate_{\iteridx}$ to satisfy 
the following conditions \cite[Sec. 6.1]{CvxAlgBertsekas}, \cite{robbins1951stochastic}
\begin{equation} 
	\label{equ_def_lrate_schedule}
	\lim_{\iteridx\rightarrow \infty} \lrate_{\iteridx} = 0 \mbox{, } 	\sum_{\iteridx=1}^{\infty} \lrate_{\iteridx} = \infty \mbox{ , and } \sum_{\iteridx=1}^{\infty} \lrate^{2}_{\iteridx} < \infty. 
\end{equation} 
Running the \gls{gradstep} \eqref{equ_def_basic_gradstep} with a \gls{learnrate} schedule $\lrate_{\iteridx}$ that 
satisfies \eqref{equ_def_lrate_schedule} ensures convergence to a minimum of $f(\weights)$ if 
\begin{itemize} 
	\item the iterates $\normgeneric{\weights^{(\iteridx)}}{2}$ are bounded, i.e., $\sup_{\iteridx=1,\ldots} \normgeneric{\weights^{(\iteridx)}}{2}$ is finite, and 
	\item the \glspl{gradient} $\normgeneric{\nabla f\big( \weights^{(\iteridx)} \big)}{2}$, for $\iteridx=1,2,\ldots$, are also bounded. 
\end{itemize}
A detailed convergence proof can be found in \cite[Sec.\ 3]{CvxAlgBertsekas}.

It is instructive to discuss the meanings of the individual conditions in \eqref{equ_def_lrate_schedule}. 
The first condition \eqref{equ_def_lrate_schedule} requires that the \gls{learnrate} eventually become sufficiently 
small to avoid overshooting. The third condition \eqref{equ_def_lrate_schedule} ensures that this required decay of the 
\gls{learnrate} does not take ``forever''. Note that the first and third condition in \eqref{equ_def_lrate_schedule} could be satisfied 
by the trivial \gls{learnrate} schedule $\lrate_{\iteridx} = 0$ which is clearly not useful as the \gls{gradstep} has no 
effect. 

The trivial schedule $\lrate_{\iteridx} = 0$ is ruled out by the middle condition 
of \eqref{equ_def_lrate_schedule}. This middle condition ensures that the \gls{learnrate} 
$\lrate_{\iteridx}$ is large enough such that the \gls{gradstep}s make  
sufficient progress towards a minimizer of the \gls{objfunc}. 

We emphasize that the conditions in \eqref{equ_def_lrate_schedule} are independent 
of any properties of the matrix $\mQ$ in \eqref{equ_generic_cvx_quad_function}. The 
matrix $\mQ$ is determined by \glspl{datapoint} (see, e.g., \eqref{equ_def_param_erm_LR}), 
whose statistical properties can typically be controlled only to a limited extent, 
such as through data normalization.

\subsection{When to Stop?}
\label{sec_stopping_criterion} 

For the \gls{stopcrit}, we may use a fixed number of iterations, $\iteridx_{\rm max}$. 
This hyper-parameter can be determined by constraints on computational resources. 
We can optimize the number of iterations also via meta-learning, i.e., trying to 
predict the optimal $\iteridx_{\rm max}$ based on key characteristics (or \gls{feature}s) 
of the \gls{objfunc} \cite{LearnLearnGDGD}. 


Another \gls{stopcrit} can be obtained by monitoring the decrease in the \gls{objfunc} 
$f\big(\weights^{(\iteridx)} \big)$. Specifically, we stop repeating the \gls{gradstep} \eqref{equ_def_basic_gradstep} when
$\big| f\big(\weights^{(\iteridx)} \big) - f\big(\weights^{(\iteridx+1)} \big) \big| \leq \opttol$ 
for a given tolerance $\opttol$. As before, we can optimize the tolerance level $\opttol$ 
via meta-learning techniques \cite{LearnLearnGDGD}.

For an \gls{objfunc} of the form \eqref{equ_generic_cvx_quad_function}, we can use 
information about the \gls{psd} matrix $\mQ$ to construct a \gls{stopcrit}.\footnote{For \gls{linreg} \eqref{equ_def_param_erm_linreg_qadform}, 
	the matrix $\mQ$ is determined by the \gls{feature}s of the \glspl{datapoint} in the \gls{trainset}. 
	We can influence the properties of $\mQ$ to some extent by \gls{feature} transformation methods. 
	One important example of such a transformation is the normalization of \glspl{feature}.}
Indeed, the choice of the \gls{learnrate} $\lrate$ and the \gls{stopcrit} can be guided 
by the \glspl{eigenvalue}
\begin{equation}
	\nonumber
	0 \leq \eigval{1}(\mQ) \leq \ldots \leq \eigval{\dimlocalmodel}(\mQ).  
\end{equation}  
Even if we do not know these \gls{eigenvalue}s precisely, we might know (or be able to ensure 
via \gls{featlearn}) some upper and lower bounds, 
\begin{equation}
	\label{equ_def_lower_upper_bound_eigvals}
	0 \leq \lowerboundeigval \leq \eigval{1}(\mQ) \leq \ldots \leq \eigval{\dimlocalmodel}(\mQ) \leq \upperboundeigval. 
\end{equation}  

In what follows, we assume that $\mQ$ is invertible and that we know some positive 
lower bound $\lowerboundeigval > 0$ on its \glspl{eigenvalue} (see \eqref{equ_def_lower_upper_bound_eigvals}). 
The \gls{objfunc} \eqref{equ_generic_cvx_quad_function} has then a unique solution $\widehat{\weights}$. 
A \gls{gradstep} \eqref{equ_def_basic_gradstep} reduces the distance 
$\normgeneric{\weights^{(\iteridx)}- \widehat{\weights}}{2}$ to $\widehat{\weights}$ by a constant factor \cite[Ch. 6]{CvxAlgBertsekas},  
\begin{equation} 
	\label{equ_def_gd_step_multi_factor}
	\normgeneric{\weights^{(\iteridx+1)}- \widehat{\weights}}{2} \leq \gdcontract{\lrate_{\iteridx}}{\mathbf{Q}} \normgeneric{\weights^{(\iteridx)}- \widehat{\weights}}{2}. 
\end{equation}  
Here, we used the contraction factor 
\begin{equation} 
	\label{equ_def_contr_gd} 
	\kappa^{(\lrate)}(\mathbf{Q}) \defeq  {\rm max} \big\{ |1-\lrate 2 \eigval{1} |, |1-\lrate 2 \eigval{\nrfeatures}|  \big\}. 
\end{equation}
The contraction factor depends on the \gls{learnrate} $\lrate$ which is a 
hyper-parameter of \glspl{gdmethod} that we can control. However, the contraction 
factor also depends on the \gls{eigenvalue}s of the matrix $\mQ$ in \eqref{equ_generic_cvx_quad_function}. 
In \gls{ml} and \gls{fl} applications, this matrix typically depends on \gls{data} and can be controlled 
only to some extent, e.g., using \gls{feature} transformation \cite[Ch. 5]{MLBasics}. To ensure 
$\kappa^{(\lrate)}(\mathbf{Q}) < 1$, we require a positive \gls{learnrate} satisfying $\lrate_{\iteridx} < 1/U$.

Consider the \gls{gradstep} \eqref{equ_def_basic_gradstep} with fixed \gls{learnrate} $\lrate$ and 
a contraction factor $\kappa^{(\lrate)}(\mathbf{Q})\!<\!1$ (see \eqref{equ_def_contr_gd}). We 
can then ensure an optimization error
$\normgeneric{\weights^{(\iteridx)}- \widehat{\weights}}{2} \leq \varepsilon$ 
(see \eqref{equ_def_gd_step_multi_factor}) if the number $\iteridx$ of \gls{gradstep}s 
satisfies 
\begin{equation}
	\label{equ_def_min_nr_steps_conv_linreg}
	\iteridx \geq  \underbrace{\bigg\lceil \frac{\log \big(\normgeneric{\weights^{(0)}- \widehat{\weights}}{2}/ \varepsilon \big)}{\log \big( 1/\kappa^{(\lrate)}(\mathbf{Q})\big)} \bigg\rceil}_{ =:\iteridx^{(\varepsilon)} }. 
\end{equation} 

\begin{figure}[htb]
	\vspace*{14mm}
	\begin{center}
		\begin{tikzpicture}[scale = 2]
			\draw[->, thick] (0,0.5) -- (6.0,0.5) node[right] {$\lrate$};       
			\draw[->, thick] (0.5,0) -- (0.5,3.5); 
			\foreach \i in {2.7} \draw [very thick] (\i,0.4)--(\i,.6);
			\foreach \i in {1.6,2.7} \draw [very thick] (0.4, \i)--(0.6,\i);
			\node[color=black] at(0,2.7) {$1$};
			
			\draw[color=black, very thick, scale =0.2, domain = 2.5:13.5, variable = \x]  plot ({\x},{16-\x}) node [color=black, below=0.1] {$1/(2\eigval{\nrfeatures})$};
			\draw[color=black, very thick, scale =0.2, domain = 13.5:24.5, variable = \x]  plot ({\x},{\x-11});    
			\draw[color=black, dashed, scale =0.2, domain = 2.5:26, variable = \x]  plot ({\x},{13.5-0*\x}); 
			
			\draw[very thick,black,dashed,scale =0.2, domain = 2.5:26, variable = \x] plot ({\x},{14.125-0.25*\x}) node [color=black, above=0]  {\hspace*{15mm} $|1\!-\!\lrate 2 \eigval{1} |$};
			
			\node[color=black, right] at (2.0,1.2) {$|1\!-\!\lrate 2 \eigval{\nrfeatures} |$};  
			
			\draw[thick,black,dashed, scale =0.2, domain = 2.5:20, variable = \x] plot ({\x},{9.125}) ;
			\node[color=black] at(-0.2,1.9) {$\kappa^{*}(\mQ) \!=\!\frac{(\eigval{\nrfeatures}/\eigval{1})\!-\!1}{(\eigval{\nrfeatures}/\eigval{1})\!+\!1}$};
			
			\draw[black,dashed, scale =0.2, domain = 2.5:9.1255, variable = \y] plot ({20},{\y}) ;
			\node[color=black, centered] at(4.0,0.3) {$\lrate^{*}\!=\!\frac{1}{\eigval{1}+\eigval{\nrfeatures}}$};
			
			\draw[black,dashed, scale =0.2, domain = 2.5:14.1255, variable = \y] plot ({24.5},{\y}) ;
			\node[color=black,centered] at(5,0.3) {$\frac{1}{\eigval{\nrfeatures}}$};
			
			\draw[thick,red,scale =0.2, domain = 2.5:20, variable = \x] plot ({\x},{14.125-0.25*\x});  
			\node[color=black] at(3,2.25) {$\kappa^{(\lrate)}(\mQ)$};
			
			\draw[thick,red,scale =0.2, domain = 20:25, variable = \x]  plot ({\x},{\x-11}) ; 
			
		\end{tikzpicture}
	\end{center}
	\caption{The contraction factor $\kappa^{(\lrate)}(\mathbf{Q})$ \eqref{equ_def_contr_gd}, 
		used in the upper bound \eqref{equ_def_gd_step_multi_factor}, as a function of the \gls{learnrate} $\lrate$. 
		Note that $\kappa^{(\lrate)}(\mathbf{Q})$ also depends on the \gls{eigenvalue}s of the 
		matrix $\mQ$ in \eqref{equ_generic_cvx_quad_function}. 
	} 
	\label{fig_contrac_alpha_func}
\end{figure}
According to \eqref{equ_def_gd_step_multi_factor}, smaller values of the contraction 
factor $\kappa^{(\lrate)}(\mathbf{Q})$ guarantee a faster convergence of \eqref{equ_def_basic_gradstep} 
towards the solution of \eqref{equ_generic_cvx_quad_function}. Figure \ref{fig_contrac_alpha_func} illustrates 
the dependence of $\kappa^{(\lrate)}(\mathbf{Q})$ on the \gls{learnrate} $\lrate$. Thus, choosing a 
small $\lrate$ (close to 0) will typically result in a larger $\kappa^{(\lrate)}(\mathbf{Q})$ and, in turn, 
require more iterations to ensure optimization error level $\opttol$ via \eqref{equ_def_gd_step_multi_factor}. 

We can minimize this contraction factor by choosing the \gls{learnrate} (see Figure \ref{fig_contrac_alpha_func})
\begin{equation} 
	\label{equ_def_opt_lrate}
	\lrate^{(*)} \defeq \frac{1}{\eigval{1}+ \eigval{\nrfeatures}}. 
\end{equation}
[Note that evaluating \eqref{equ_def_opt_lrate} requires to know the extremal \gls{eigenvalue}s $\eigval{1}, \eigval{\dimlocalmodel}$ 
of $\mQ$.]
Inserting the optimal \gls{learnrate} \eqref{equ_def_opt_lrate} into \eqref{equ_def_gd_step_multi_factor}, 
\begin{equation} 
	\label{equ_def_gd_step_multi_factor_opt}
	\normgeneric{\weights^{(\iteridx+1)}- \widehat{\weights}}{2} \leq \underbrace{\frac{(\eigval{\nrfeatures}/\eigval{1})-1}{(\eigval{\nrfeatures}/\eigval{1})+1}}_{=: \kappa^{*}(\mQ)} \normgeneric{\weights^{(\iteridx)}- \widehat{\weights}}{2}. 
\end{equation}

Carefully note that the formula \eqref{equ_def_gd_step_multi_factor_opt} is valid only if 
the matrix $\mQ$ in \eqref{equ_generic_cvx_quad_function} is invertible, i.e., if $\eigval{1} > 0$. 
If the matrix $\mQ$ is singular ($\eigval{1}=0)$, the convergence of \eqref{equ_def_basic_gradstep} 
towards a solution of \eqref{equ_generic_cvx_quad_function} is much slower than the 
decrease of the bound \eqref{equ_def_gd_step_multi_factor_opt}. However, we can still 
ensure the convergence of \gls{gradstep}s $\weights^{(\iteridx)}$ by using a fixed \gls{learnrate} $\lrate_{\iteridx} = \lrate$ 
that satisfies \cite[Thm. 2.1.14]{nesterov04}
\begin{equation}
	\label{equ_suff_cond_smooth_gd}
	0 < \lrate < 1/\eigval{\nrfeatures}(\mQ). 
\end{equation} 
It is interesting to note that for \gls{linreg}, the matrix $\mQ$ depends only on the 
\gls{feature}s $\featurevec^{(\sampleidx)}$ of the \gls{datapoint}s in the \gls{trainset} 
(see \eqref{equ_def_lin_reg_perturbed_gd_def_errvec}) but not on their \gls{label}s $\truelabel^{(\sampleidx)}$. 
Thus, the convergence of \gls{gradstep}s is only affected by the \gls{feature}s, 
whereas the \gls{label}s are irrelevant. The same is true for \gls{ridgeregression} and 
\gls{gtvmin} (using local \gls{linmodel}s).

Note that both, the optimal \gls{learnrate} \eqref{equ_def_opt_lrate} and the optimal contraction 
factor 
\begin{equation} 
	\label{equ_def_opt_contr_factor} 
	\kappa^{*}(\mQ) \defeq \frac{(\eigval{\nrfeatures}/\eigval{1})-1}{(\eigval{\nrfeatures}/\eigval{1})+1}
\end{equation} 
depend on the \gls{eigenvalue}s of the matrix $\mQ$ in \eqref{equ_generic_cvx_quad_function}. 

According to \eqref{equ_def_gd_step_multi_factor_opt}, the ideal case is when all \gls{eigenvalue}s 
are identical which leads, in turn, to a contraction factor $\kappa^{*}(\mQ) = 0$. Here, a single \gls{gradstep} 
arrives at the unique solution of \eqref{equ_generic_cvx_quad_function}. 

In general, we do not have full control over the matrix $\mQ$ and its \gls{eigenvalue}s. 
For example, the matrix $\mQ$ arising in \gls{linreg} \eqref{equ_def_param_erm_linreg_qadform} 
is determined by the \gls{feature}s of \gls{datapoint}s in the \gls{trainset}. These \gls{feature}s 
might be obtained from sensing \glspl{device} and therefore beyond our control. However, 
some applications might allow for some design freedom in the choice of \gls{feature} vectors. 
We might also use \gls{feature} transformations that nudge the resulting $\mQ$ in 
\eqref{equ_def_param_erm_linreg_qadform} more towards a scaled identity matrix. 

\subsection{Perturbed Gradient Step} 
\label{sec_perturbed_gd}
Consider the \gls{gradstep} \eqref{equ_def_basic_gradstep} used to find a minimum of 
\eqref{equ_generic_cvx_quad_function}. We again assume that the matrix $\mQ$ in 
\eqref{equ_generic_cvx_quad_function} is invertible ($\eigval{1}(\mQ) > 0$) and, in turn, 
\eqref{equ_generic_cvx_quad_function} has a unique solution $\widehat{\weights}$. 

In some applications, it is challenging to evaluate the \gls{gradient} $\nabla f(\weights) = 2 \mQ \weights + \vq$ 
of \eqref{equ_generic_cvx_quad_function} exactly. For example, the evaluation could 
require to gather \gls{datapoint}s from distributed storage locations. These storage 
locations can become unavailable during the computation of $\nabla f(\weights)$ due 
to software or hardware failures (e.g., limited connectivity). Another source for imperfections 
can be stochastic approximation techniques where exact computations are replaced by noisy 
approximations that require less resources.\footnote{A prime example for such a stochastic approximation is 
	\gls{stochGD} which we discuss in Section \ref{sec_fl_alg_sgd}.} 

We can model imperfections during the computation of \eqref{equ_def_basic_gradstep} as the 
perturbed \gls{gradstep} 
\begin{align} 
	\label{equ_def_gd_step_linreg_perturbed}
	\weights^{(\iteridx\!+\!1)} & \defeq \weights^{(\iteridx)} - \lrate \nabla f \big( \weights^{(\iteridx)}\big)  + \perturbation{\iteridx}   \nonumber \\ 
	& \stackrel{\eqref{equ_generic_cvx_quad_function}}{=}     \weights^{(\iteridx)} - \lrate \big(2 \mQ \weights^{(\iteridx)}+\vq\big) + \perturbation{\iteridx} \mbox{, for } \iteridx=0,1,\ldots.
\end{align} 
We can use the contraction factor $\kappa \defeq  \gdcontract{\lrate}{\mQ}$ \eqref{equ_def_contr_gd} 
to upper bound the deviation between $\weights^{(\iteridx)}$ and the optimum $\widehat{\weights}$ 
as (see \eqref{equ_def_gd_step_multi_factor})
\begin{equation} 
	\label{equ_upper_bound_pert_gd}
	\normgeneric{\weights^{(\iteridx)} \!-\! \widehat{\weights} }{2} \leq \kappa^{\iteridx} \normgeneric{\weights^{(0)}\!-\!\widehat{\weights} }{2}
	+ \sum_{\iteridx'=1}^{\iteridx} \kappa^{\iteridx'} \normgeneric{\perturbation{\iteridx-\iteridx'}}{2}. 
\end{equation}
This bound applies for any number of iterations $\iteridx=1,2,\ldots$ of the perturbed \gls{gradstep} 
\eqref{equ_def_gd_step_linreg_perturbed}. 

The perturbed \gls{gradstep} \eqref{equ_def_gd_step_linreg_perturbed} could also 
be used as a tool to analyze the (exact) \gls{gradstep} for an \gls{objfunc} $\tilde{f}(\weights)$ which does 
not belong to the family \eqref{equ_generic_cvx_quad_function} of \gls{convex} \glspl{quadfunc}. 
Indeed, we can write the \gls{gradstep} for minimizing $\tilde{f}(\weights)$ as 
\begin{align} 
	\weights^{(\iteridx+1)} & \defeq \weights^{(\iteridx)} - \lrate \nabla \tilde{f}(\weights)  \nonumber \\ 
	& =  \weights^{(\iteridx)} - \lrate \nabla f(\weights) + \underbrace{\lrate\big(\nabla f(\weights) - \nabla\tilde{f}(\weights)\big)}_{\defeq \perturbation{\iteridx}}.  \nonumber
\end{align} 
The last identity is valid for any choice of surrogate function $f(\weights)$. In particular, we can 
choose $f(\weights)$ as a \gls{convex} \gls{quadfunc} \eqref{equ_generic_cvx_quad_function} that 
approximates $\tilde{f}(\weights)$. Note that the perturbation term $\perturbation{\iteridx}$ is scaled 
by the \gls{learnrate} $\lrate$.

\subsection{Handling Constraints - Projected Gradient Descent} 
\label{sec_proj_gradient_descent}

Many important \gls{ml} and \gls{fl} methods amount to the minimization of an \gls{objfunc} of the 
form \eqref{equ_generic_cvx_quad_function}. The optimization variable $\weights$ in 
\eqref{equ_generic_cvx_quad_function} represents some \gls{model} \glspl{parameter}. 

Sometimes we might require the \glspl{parameter} $\weights$ to belong to a subset $\mathcal{S} \subset \mathbb{R}^{\featurelen}$. 
One example is \gls{regularization} via \gls{model} pruning (see Chapter \ref{lec_mlbasics}).  
Another example are \gls{fl} methods that learn identical \gls{localmodel} \glspl{parameter} $\localparams{\nodeidx}$ 
at all nodes $\nodeidx \in \nodes$ of an \gls{flnetwork}. This can be implemented by requiring the stacked 
\gls{localmodel} \glspl{parameter} $\weights = \big( \localparams{1},\ldots, \localparams{\nrnodes} \big)^{T}$ to 
belong to the subset $$\mathcal{S} = \bigg\{ \big( \localparams{1},\ldots, \localparams{\nrnodes} \big)^{T}:  \localparams{1}=\ldots=\localparams{\nrnodes}  \bigg\}.$$

Let us now show how to adapt the \gls{gradstep} \eqref{equ_def_basic_gradstep} to solve the 
constrained problem 
\begin{equation} 
	\label{equ_def_min_quad_func_constr}
	f^{*}= \min_{\weights \in \mathcal{S}} \weights^{T} \mQ \weights + \vq^{T} \weights.  
\end{equation}
We assume that the constraint set $\mathcal{S} \subseteq \mathbb{R}^{\nrfeatures}$ is such that 
we can efficiently compute the projection
\begin{equation} 
	\label{equ_def_proj_generic}
	\projection{\mathcal{S}}{\weights} = \argmin_{\weights' \in \mathcal{S}} \normgeneric{\weights - \weights'}{2} \mbox{ for any } \weights \in \mathbb{R}^{\nrfeatures}. 
\end{equation}
A suitable modification of the \gls{gradstep} \eqref{equ_def_basic_gradstep} to solve the 
constrained variant \eqref{equ_def_min_quad_func_constr} is \cite{CvxAlgBertsekas}
\begin{align}
	\label{equ_def_gd_step_linreg_constr}
	\weights^{(\iteridx+1)} & \defeq \projection{\mathcal{S}}{\weights^{(\iteridx)} - \lrate \nabla f \big( \weights^{(\iteridx)} \big)}  \nonumber \\ 
	& \stackrel{\eqref{equ_generic_cvx_quad_function}}{=}  \projection{\mathcal{S}}{ \weights^{(\iteridx)} - \lrate \big( 2 \mQ \weights^{(\iteridx)} + \vq  \big)}.
\end{align}

The \gls{projgd} step \eqref{equ_def_gd_step_linreg_constr} consists of:  
\begin{enumerate} 
	\item computing an ordinary \gls{gradstep} $\weights^{(\iteridx)} \mapsto \weights^{(\iteridx)} - \lrate \nabla f \big( \weights^{(\iteridx)} \big)$ 
	and  then 
	\item projecting the result back to the constraint set $\mathcal{S}$. 
\end{enumerate}
Note that we re-obtain the basic \gls{gradstep} \eqref{equ_def_basic_gradstep} from the projected 
\gls{gradstep} \eqref{equ_def_gd_step_linreg_constr} for the trivial constraint set $\mathcal{S} = \mathbb{R}^{\nrfeatures}$. 
\begin{figure}[htbp]
	\begin{center}
		\begin{tikzpicture}[scale=0.9]
			\node [right] at (-5.1,1.7) {$f(\weights)$} ;
			\draw[ultra thick, domain=-4.1:4.1] plot (\x,  {(1/8)*\x*\x});
			\draw [fill] (2.83,1) circle [radius=0.1] node[right] {$\weights^{(\iteridx)}$};
			\draw[line width =0.5mm,dashed,->] (2.83,1) -- node[midway,above] {\eqref{equ_def_basic_gradstep}} (-1.5,1);
			\draw[line width =0.2mm,dashed] (-1.5,1) --(-1.5,-1.5)  node [below, left]  {$\weights^{(\iteridx)}\!-\!\lrate \nabla f\big(\weights^{(\iteridx)}\big) $} ;
			\draw[line width =0.5mm,dashed,->] (-1.5,-1.5)  -- node[midway,above] {$\projection{\mathcal{S}}{\cdot}$} (1,-1.5) ; 
			\draw [fill] (1,-1.5) circle [radius=0.1] node[below] {$\weights^{(\iteridx+1)}$};
			\draw[line width=1mm] (1,-1.5) -- (3,-1.5) node[right] {$\mathcal{S}$};
		\end{tikzpicture}
		\vspace*{-5mm}
	\end{center}
	\caption{\Gls{projgd} augments a basic \gls{gradstep} with a projection back onto the constraint set $\mathcal{S}$.}
	\label{fig_projected_GD}
\end{figure}

The approaches for choosing the \gls{learnrate} $\lrate$ and \gls{stopcrit} for basic 
\gls{gradstep} \eqref{equ_def_basic_gradstep} explained in Sections \ref{sec_learn_rate} and \ref{sec_stopping_criterion} 
work also for the projected \gls{gradstep} \eqref{equ_def_gd_step_linreg_constr}. 
In particular, the convergence speed of the projected \gls{gradstep} is also characterized 
by \eqref{equ_def_gd_step_multi_factor} \cite[Ch. 6]{CvxAlgBertsekas}. This follows 
from the fact that the concatenation of a contraction (such as the \gls{gradstep} \eqref{equ_def_basic_gradstep} for 
sufficiently small $\lrate$) and a projection (such as $\projection{\mathcal{S}}{\cdot}$) results 
again in a contraction with the same contraction factor. 

Thus, the convergence speed of \gls{projgd}, in terms of number of iterations required to ensure a 
given level of optimization error, is essentially the same as that of basic \gls{gd}. 
However, the bound \eqref{equ_def_gd_step_multi_factor} is only telling about the number of projected 
\gls{gradstep}s \eqref{equ_def_gd_step_linreg_constr} required to achieve a guaranteed 
level of sub-optimality $\big| f\big(\weights^{(\iteridx)} \big) -f^{*} \big|$. The iteration 
\eqref{equ_def_gd_step_linreg_constr} of \gls{projgd} might require significantly more 
computation than the basic \gls{gradstep}, as it requires to compute the projection \eqref{equ_def_proj_generic}. 

\subsection{Extended Gradient Methods for Federated Optimization} 
\label{sec_gen_gradient_step} 

The \glspl{gdmethod} discussed so far can be used to learn a \gls{hypothesis} from 
a parametric \gls{model}. Let us now sketch one possible generalization of the 
\gls{gradstep} \eqref{equ_def_basic_gradstep} for a \gls{model} $\hypospace$ without 
a parametrization. 

We start with rewriting the \gls{gradstep} \eqref{equ_def_basic_gradstep} 
as the optimization 
\begin{equation}
	\label{equ_def_gradstep_opt}
	\weights^{(\iteridx+1)} \!=\!\argmin_{\weights \in \mathbb{R}^{\dimlocalmodel}} \left[  (1/(2\lrate))\normgeneric{\weights\!-\!  \weights^{(\iteridx)}}{2}^{2}\!+\!\underbrace{f\big(  \weights^{(\iteridx)} \big)\!+\!\big(\weights\!-\! \weights^{(\iteridx)}\big)^{T}  \nabla f\big( \weights^{(\iteridx)} \big)}_{\approx f(\weights)} \right]. 
\end{equation} 
The \gls{objfunc} in \eqref{equ_def_gradstep_opt} includes the first-order approximation 
$$  f(\weights) \approx f\big(  \weights^{(\iteridx)} \big) + \big(\weights- \weights^{(\iteridx)}\big)^{T}  \nabla f\big( \weights^{(\iteridx)} \big) $$  
of the function $f(\weights)$ around the location $\weights = \weights^{(\iteridx)}$ (see Figure \ref{fig_smooth_function}). 

Let us modify \eqref{equ_def_gradstep_opt} by using $f(\weights)$ itself (instead of an approximation), 
\begin{equation}
	\label{equ_def_prox_operator}
	\weights^{(\iteridx+1)}  = \argmin_{\weights \in \mathbb{R}^{\dimlocalmodel}}\bigg[ f(\weights)\!+\!(1/(2\lrate))\normgeneric{\weights -  \weights^{(\iteridx)}}{2}^{2}\bigg]. 
\end{equation}
Like the \gls{gradstep}, also \eqref{equ_def_prox_operator} maps a given vector $\weights^{(\iteridx)}$ to an 
updated vector $\weights^{(\iteridx+1)}$. Note that \eqref{equ_def_prox_operator} 
is nothing but the \gls{proxop} of the function $f(\weights)$ \cite{ProximalMethods}. 
Similar to the role of the \gls{gradstep} as the main building block of \glspl{gdmethod}, 
the \gls{proxop} \eqref{equ_def_prox_operator} is the main building block of 
proximal algorithms \cite{ProximalMethods}. 

To obtain a version of \eqref{equ_def_prox_operator} for a non-parametric \gls{model}, we need to be 
able to evaluate its objective function directly in terms of a \gls{hypothesis} $\hypothesis$ instead of its 
\glspl{parameter} $\weights$. The objective function \eqref{equ_def_prox_operator} consists of two 
components. The first component $f(\cdot)$, which is the function we want to minimize, is 
obtained from a \gls{trainerr} incurred by a \gls{hypothesis}, which might be parametric 
$\hypothesis^{(\weights)}$. Thus, we can evaluate the function $f(\hypothesis)$ by computing 
the \gls{trainerr} for a given \gls{hypothesis}. 

The second component of the \gls{objfunc} in \eqref{equ_def_prox_operator} uses $\normgeneric{\weights -  \weights^{(\iteridx)}}{2}^{2}$ 
to measure the difference between the \gls{hypothesis} maps $\hypothesis^{(\weights)}$ and $\hypothesis^{( \weights^{(\iteridx)})}$. 
Another measure for the difference between two \gls{hypothesis} maps can be obtained by using 
some test \gls{dataset} $\dataset' = \big\{ \featurevec^{(1)},\ldots,\featurevec^{(\samplesize')} \big \} $: 
The average squared difference between their \gls{prediction}s, 
\begin{equation} 
	\label{equ_measure_diff_hypo}
	(1/\samplesize') \sum_{\sampleidx=1}^{\samplesize'} \bigg( \hypothesis \big(\featurevec^{(\sampleidx)} \big)  -  \hypothesis^{(\iteridx)}  \big(\featurevec^{(\sampleidx)} \big) \bigg)^{2},
\end{equation} 
is a measure for the difference between $\hypothesis$ and $\hypothesis^{(\iteridx)}$. Note that 
\eqref{equ_measure_diff_hypo} only requires the \glspl{prediction} delivered by the \gls{hypothesis} 
maps $\hypothesis,\hypothesis^{(\iteridx)}$ on $\dataset'$ - no other information is needed about 
these maps. 

It is interesting to note that \eqref{equ_measure_diff_hypo} coincides with $\normgeneric{\weights -  \weights^{(\iteridx)}}{2}^{2}$ 
for the \gls{linmodel} $\hypothesis^{(\weights)}(\featurevec) \defeq \weights^{T} \featurevec$ and 
a specific construction of the \gls{dataset} $\dataset'$. This construction uses the \gls{realization}s $\featurevec^{(1)},\featurevec^{(2)},\ldots$ of \gls{iid} \gls{rv}s with a common \gls{probdist} $\featurevec \sim \mathcal{N}(\mathbf{0},\mathbf{I})$. Indeed, by the \gls{lln} 
\begin{align}
	& \lim_{\samplesize' \rightarrow \infty}	(1/\samplesize') \sum_{\sampleidx=1}^{\samplesize'} \bigg( \hypothesis^{(\weights)} \big(\featurevec^{(\sampleidx)} \big)  -  \hypothesis^{(\weights^{(\iteridx)})}  \big(\featurevec^{(\sampleidx)} \big) \bigg)^{2}  \nonumber \\ 
	& = \lim_{\samplesize' \rightarrow \infty}	(1/\samplesize') \sum_{\sampleidx=1}^{\samplesize'} \bigg( \big( \weights  - \weights^{(\iteridx)} \big)^{T} \featurevec^{(\sampleidx)} \bigg)^{2} \nonumber \\ 
	& = \lim_{\samplesize' \rightarrow \infty}	(1/\samplesize') \sum_{\sampleidx=1}^{\samplesize'}  \big( \weights  - \weights^{(\iteridx)} \big)^{T}  \featurevec^{(\sampleidx)} \big( \featurevec^{(\sampleidx)}\big)^{T}  \big( \weights  - \weights^{(\iteridx)} \big) 
	\nonumber \\ 
	& = \big( \weights  - \weights^{(\iteridx)} \big)^{T}  \underbrace{\bigg[ \lim_{\samplesize' \rightarrow \infty}	(1/\samplesize') \sum_{\sampleidx=1}^{\samplesize'}  \featurevec^{(\sampleidx)} \big( \featurevec^{(\sampleidx)}\big)^{T}  \bigg]}_{= \mathbf{I}} \big( \weights  - \weights^{(\iteridx)} \big) \nonumber \\
	& = \normgeneric{\weights  - \weights^{(\iteridx)}}{2}^{2}. 
\end{align} 

Finally, we arrive at a generalized \gls{gradstep} for the training of a non-parametric \gls{model} $\hypospace$ 
by replacing $\normgeneric{\weights -  \weights^{(\iteridx)}}{2}^{2}$ in \eqref{equ_def_prox_operator} 
with \eqref{equ_measure_diff_hypo}. In other words, 
\begin{equation}
	\label{equ_def_prox_operator_non_param}
	\hypothesis^{(\iteridx+1)}  = \argmin_{\hypothesis \in \hypospace} \bigg[ (1/(2\lrate\samplesize')) \sum_{\sampleidx=1}^{\samplesize'} \bigg( \hypothesis \big(\featurevec^{(\sampleidx)} \big)  -  \hypothesis^{(\iteridx)}  \big(\featurevec^{(\sampleidx)} \big) \bigg)^{2} +f(\hypothesis)\bigg]. 
\end{equation}
We can turn \glspl{gdmethod} for the training of \glspl{parammodel} into 
corresponding training methods for non-parametric \gls{model}s by replacing the \gls{gradstep} with the update \eqref{equ_def_prox_operator_non_param}. For example, we obtain Algorithm \ref{alg_basic_gdmethods_nonparam} 
from Algorithm \ref{alg_basic_gdmethods} by modifying step \ref{step_grad_step} suitably. 
\begin{algorithm}[htbp]
	\caption{A blueprint for generalized \glspl{gdmethod}}\label{alg_basic_gdmethods_nonparam}
	\begin{algorithmic}[1]
		\renewcommand{\algorithmicrequire}{\textbf{Input:}}
		\renewcommand{\algorithmicensure}{\textbf{Output:}}
		\Require some \gls{objfunc} $f: \hypospace \rightarrow \mathbb{R}$ (e.g., the 
		average \gls{loss} of a \gls{hypothesis} $\hypothesis \in \hypospace$ on a \gls{trainset}); \gls{learnrate} $\lrate >0$; some \gls{stopcrit}; test \gls{dataset} $\dataset' =\{ \featurevec^{(1)},\ldots,\featurevec^{(\samplesize')}\}$
		\Statex\hspace{-6mm}{\bf Initialize:} set $\hypothesis^{(0)}\!\defeq\!\mathbf{0}$; set iteration counter $\iteridx\!\defeq\!0$   
		\Repeat 
		\State $\iteridx\!\defeq\!\iteridx\!+\!1$ (increase iteration counter) 
		\State 	do a generalized \gls{gradstep} \eqref{equ_def_prox_operator_non_param}, \label{step_grad_step_non_param}  $$\hypothesis^{(\iteridx)} \!=\!\argmin_{\hypothesis \in \hypospace} \bigg[ (1/(2\lrate\samplesize')) \sum_{\sampleidx=1}^{\samplesize'} \bigg( \hypothesis \big(\featurevec^{(\sampleidx)} \big)  -  \hypothesis^{(\iteridx-1)}  \big(\featurevec^{(\sampleidx)} \big) \bigg)^{2} +f(\hypothesis)\bigg]$$
		\Until \gls{stopcrit} is met 
		\Ensure learned \gls{hypothesis} $\widehat{\hypothesis} \defeq \hypothesis^{(\iteridx)}$ (hopefully $f\big(\widehat{\hypothesis} \big) \approx \min_{\hypothesis \in \hypospace} f(\hypothesis)$) 
	\end{algorithmic}
\end{algorithm}

\newpage
\subsection{Gradient Methods as Fixed-Point Iterations} 
\label{sec_fixedpoint_interpretation}

The iterative \glspl{optmethod} discussed in the previous sections are all special cases 
of a \gls{fixedpointiter},
\begin{equation} 
	\label{equ_def_fixed_point}
	\netparams^{(\iteridx)} = \fixedpointop  \netparams^{(\iteridx-1)}, \mbox{ for } \iteridx=1,2,\ldots. 
\end{equation} 
Different \glspl{optmethod} use different choices for the operator $\fixedpointop$ whose fixed points are 
solutions of the underlying \gls{optproblem}. For example, the \gls{gradstep} \eqref{equ_def_basic_gradstep} 
is obtained from \eqref{equ_def_fixed_point} with the operator $\fixedpointop^{(\rm GD)}: \weights \mapsto \weights - \lrate \nabla f(\weights)$. 
For a \gls{differentiable} and \gls{convex} \gls{objfunc} $f(\weights)$, every minimizer $\widehat{\weights}$ is a fixed 
point of $\fixedpointop^{(\rm GD)}$.

The \gls{fixedpointiter} \eqref{equ_def_fixed_point} will be the core computational step of 
every \gls{fl} \gls{algorithm} discussed in Chapter \ref{lec_flalgorithms}. These 
\glspl{algorithm} use \eqref{equ_def_fixed_point} with an operator $\fixedpointop$ 
determined by an instance of \gls{gtvmin}. More precisely, any fixed point of $\fixedpointop$ 
must be an \gls{gtvmin}-solution $\widehat{\netparams} \in \mathbb{R}^{\dimlocalmodel  \nrnodes}$,
\begin{equation} 
	\label{equ_def_gtvmin_fixed_point}
	\fixedpointop \widehat{\netparams}  = \widehat{\netparams}. 
\end{equation} 

Given an instance of \gls{gtvmin}, there are many different operators $\fixedpointop$ that satisfy 
\eqref{equ_def_gtvmin_fixed_point}. We obtain different \gls{fl} \glspl{algorithm} by using different 
choices for $\fixedpointop$ in \eqref{equ_def_fixed_point}. Clearly, we should use an operator 
$\fixedpointop$ in \eqref{equ_def_fixed_point} that reduces the distance to a solution, 
\begin{equation} 
	\nonumber
	\underbrace{\normgeneric{ \netparams^{(\iteridx+1)} - \widehat{\netparams}}{2}}_{\stackrel{\eqref{equ_def_fixed_point},\eqref{equ_def_gtvmin_fixed_point}}{=} \normgeneric{ \fixedpointop \netparams^{(\iteridx)} - \fixedpointop\widehat{\netparams}}{2}}  \leq 	\normgeneric{ \netparams^{(\iteridx)} - \widehat{\netparams}}{2}. 
\end{equation}
Thus, we require $\fixedpointop$ to be at least non-expansive, i.e., the iteration \eqref{equ_def_fixed_point} 
should not result in worse \glspl{modelparam} that have a larger distance to the \gls{gtvmin} solution. 
Moreover, each iteration \eqref{equ_def_fixed_point} should also make some progress, i.e., 
reduce the distance from a \gls{gtvmin} solution. This requirement can be made precise using 
the notion of a \gls{contractop} \cite{Bauschke:2017,fixedpoinIsta}. 

The operator $\fixedpointop$ is a \gls{contractop} if, for some $\contractfac \in [0,1)$,
\begin{equation} 
	\nonumber
	\normgeneric{ \fixedpointop \weights\!-\!\fixedpointop \weights'}{2}  \leq  \contractfac	\normgeneric{\weights\!-\!\weights'}{2} \mbox{ holds for any } \weights,\weights' \in \mathbb{R}^{\dimlocalmodel \nrnodes}.
\end{equation}
For a \gls{contractop} $\fixedpointop$, the \gls{fixedpointiter} \eqref{equ_def_fixed_point} generates 
a sequence $\netparams^{(\iteridx)}$ that converges to a \gls{gtvmin} solution $\widehat{\netparams}$ 
quite rapidly. In particular \cite[Theorem 9.23]{RudinBookPrinciplesMatheAnalysis}, 
\begin{equation} 
	\nonumber
	\normgeneric{ \netparams^{(\iteridx)} - \widehat{\netparams}}{2} \leq \contractfac^{\iteridx} 	\normgeneric{ \netparams^{(0)} - \widehat{\netparams}}{2}. 
\end{equation} 
Here, $\normgeneric{ \netparams^{(0)} - \widehat{\netparams}}{2}$ is the distance between 
the initialization $\netparams^{(0)}$ and the solution $\widehat{\netparams}$. 

A well-known example of a \gls{fixedpointiter} \eqref{equ_def_fixed_point} using a \gls{contractop} 
is \gls{gd} \eqref{equ_def_basic_gradstep} for a \gls{smooth} and \gls{strcvx} \gls{objfunc} $f(\weights)$.\footnote{The \gls{objfunc} in \eqref{equ_generic_cvx_quad_function} is \gls{convex} and \gls{smooth} 
	for any choice of \gls{psd} matrix $\mQ$ and vector $\vq$. Moreover, it is \gls{strcvx} whenever $\mQ$ is invertible.}
In particular, \eqref{equ_def_basic_gradstep} is obtained from \eqref{equ_def_fixed_point} using 
$\fixedpointop \defeq \mathcal{G}^{(\lrate)}$ with the ``\gls{gradstep} operator'' 
\begin{equation}
	\label{equ_def_grad_op}
	\mathcal{G}^{(\lrate)}:\weights \mapsto \weights - \lrate \nabla f(\weights).
\end{equation}
Note that the operator \eqref{equ_def_grad_op} is parametrized by the \gls{learnrate} $\lrate$. 

It is instructive to study the operator $\mathcal{G}^{(\lrate)}$ for an \gls{objfunc} of 
the form \eqref{equ_generic_cvx_quad_function}. Here, 
\begin{equation} 
	\label{equ_def_grad_op_smooth_convx}
	\mathcal{G}^{(\lrate)}:\weights \mapsto 
	\weights - \lrate \underbrace{\big(2 \mQ \weights + \vq \big)}_{\stackrel{\eqref{equ_generic_cvx_quad_function}}{=} \nabla f(\weights)}.
\end{equation}  
For $\lrate \defeq 1/(2\eigval{\rm max}(\mQ))$, the operator $\mathcal{G}^{(\lrate)}$ is contractive with 
$\contractfac=1-\eigval{\rm min}(\mQ)/\eigval{\rm max}(\mQ)$. Note that $\contractfac <1$ only when 
$\eigval{\rm min}(\mQ) >0$, i.e., only when the matrix $\mQ$ in \eqref{equ_generic_cvx_quad_function} is 
invertible. 

The \gls{gradstep} operator \eqref{equ_def_grad_op_smooth_convx} is not contractive 
for the \gls{objfunc} \eqref{equ_generic_cvx_quad_function} with a 
singular matrix $\mathbf{Q}$ (for which $\eigvalgen_{\rm min}=0$). 
However, even then $\mathcal{G}^{(\lrate)}$ is still firmly non-expansive \cite{BausckeCombette}.
We refer to an operator $\fixedpointop: \mathbb{R}^{\dimlocalmodel \nrnodes} \rightarrow \mathbb{R}^{\dimlocalmodel  \nrnodes}$ 
as firmly non-expansive if 
\begin{equation} 
	\nonumber
	\normgeneric{\fixedpointop \weights - \fixedpointop \weights'}{2}^{2} \leq \big( \fixedpointop \weights - \fixedpointop \weights' \big)^{˝T} \big(  \weights -  \weights'\big) \mbox{, for any } \weights,\weights' \in \mathbb{R}^{\dimlocalmodel  \nrnodes}. 
\end{equation} 

It turns out that a \gls{fixedpointiter} \eqref{equ_def_fixed_point} with a firmly non-expansive 
operator $\fixedpointop$ is guaranteed to converge to a fixed-point of $\fixedpointop$ \cite[Cor. 5.16]{Bauschke:2017}. 
Figure \ref{fig_examples_nonexp} depicts examples of a firmly non-expansive operator, 
a non-expansive operator and a \gls{contractop}. All these operators are defined on the 
one-dimensional space $\mathbb{R}$. Another example of a firmly non-expansive operator 
is the \gls{proxop} \eqref{equ_def_prox_operator} of a \gls{convex} function \cite{Bauschke:2017,ProximalMethods}.

\definecolor{darkgreen}{rgb}{0.0, 0.5, 0.0}

\begin{figure} 
	\begin{center} 
		\begin{tikzpicture}[scale=1.5]
			\draw[line width=1pt, ->] (-2,0) -- (2,0) node[right] {$\weight^{(\iteridx)}$};
			\draw[line width=1pt, ->] (0,-2) -- (0,2) node[above] {$\weight^{(\iteridx+1)}$};
			\node at (2.1,2.2) {$\fixedpointop^{(3)}$};
			\node at (1.9,-1.5) {$\fixedpointop^{(1)}$};
			\node at (1.5,1.2) {$\fixedpointop^{(2)}$};
			\draw[dashed] (1,-2) -- (1,2); 
			\draw[dashed] (-2,1) -- (2,1); 
			\draw[dashed] (-2,-1) -- (2,-1); 
			\draw[dashed] (-1,-2) -- (-1,2); 
			\node[above,xshift=4pt,yshift=-1pt] at (1,0) {$1$};
			\node[above,xshift=8pt,yshift=-1pt] at (0,-1) {$-1$};
			\draw[line width=2,domain=-2:2,smooth,blue] plot(\x,{0.5*\x + 1});
			\draw[line width=2,domain=-2:2,smooth,red] plot(\x,{-\x});
			\draw[line width=2, domain=-2:-1,smooth,darkgreen] plot(\x,{-1});
			\draw[line width=2,domain=-1:1,smooth,darkgreen] plot(\x,{\x});
			\draw[line width=2,domain=1:2,smooth,darkgreen] plot(\x,{1});
		\end{tikzpicture}
	\end{center} 
	\caption{Example of a non-expansive operator $\fixedpointop^{(1)}$, a firmly non-expansive operator $\fixedpointop^{(2)}$ and 
		a contractive operator $\fixedpointop^{(3)}$. \label{fig_examples_nonexp}}
\end{figure} 

\newpage
\subsection{Exercises}

\refstepcounter{problem}\label{prob:oneovert}\textbf{\theproblem. Learning Rate Schedule.}  
Consider the \gls{gradstep} method applied to a \gls{differentiable} \gls{objfunc} \( f(\weights) \), 
\[
\weights^{(\iteridx+1)} = \weights^{(\iteridx)} - \lrate_{\iteridx} \nabla f \big( \weights^{(\iteridx)} \big), \quad \text{for } \iteridx = 1,2,\ldots.
\]  
where the \gls{learnrate} schedule is defined as \( \lrate_{\iteridx} \defeq \frac{1}{\iteridx} \).  
\begin{enumerate}
	\item Verify that this \gls{learnrate} schedule satisfies the standard conditions in \eqref{equ_def_lrate_schedule}.
	\item Construct a \gls{differentiable}, \gls{convex} function \( f(\weights) \) and an initialization 
	\( \weights^{(0)} \) such that the \gls{gradstep} iteration fails to converge to a minimizer of \( f(\weights) \).
\end{enumerate}

\noindent\refstepcounter{problem}\label{prob:oneovertpower}\textbf{\theproblem. Learning Rate Schedule II.}  
Consider the generic \gls{gradstep} 
\[
\weights^{(\iteridx+1)} = \weights^{(\iteridx)} - \lrate_{\iteridx} \nabla f \big( \weights^{(\iteridx)} \big), \quad \text{for } \iteridx = 1,2,\ldots.
\]  
with a \gls{learnrate} schedule of the form \( \lrate_{\iteridx} \defeq \frac{1}{\iteridx^{p}} \) with 
some $p>0$. For which values of $p>0$ does this schedule satisfy the conditions in \eqref{equ_def_lrate_schedule}?

\noindent\refstepcounter{problem}\label{prob:ogd}\textbf{\theproblem. Online Gradient Descent.}
\Gls{linreg} methods learn \glspl{modelparam} of a \gls{linmodel} with minimum \gls{risk} $\expect \big\{ \big(\truelabel - \weights^{T} \featurevec \big)^{2} \big\}$ 
where $\pair{\featurevec}{\truelabel}$ is a \gls{rv}. In practice, we do not observe the 
\gls{rv} $\pair{\featurevec}{\truelabel}$ itself but a (\gls{realization} of a) sequence of 
\gls{iid} samples $\pair{\featurevec^{(\timeidx)}}{\truelabel^{(\timeidx)}}$, for $\timeidx=1,2,\ldots$.
Online \gls{gd} is an online learning method that updates the current 
\glspl{modelparam} $\weights^{(\timeidx)}$, after observing  $\pair{\featurevec^{(\timeidx)}}{\truelabel^{(\timeidx)}}$, 
\begin{equation}
	\nonumber
	\weights^{(\timeidx+1)} \defeq \weights^{(\timeidx)}\!+\!2\lrate_{\timeidx} \featurevec^{(\timeidx)} \big( \truelabel - \big( \weights^{(\timeidx)}\big)^{T} \featurevec^{(\timeidx)}  \big) \mbox{ at time } \timeidx=1,2,\ldots. 
\end{equation} 
Starting with initialization $\weights^{(1)} \defeq \mathbf{0}$, we run \gls{onlineGD} for $M$ 
time steps, resulting in the learned \glspl{modelparam} $\weights^{(M+1)}$. Develop 
upper bounds on the risk $\expect \big\{ \big(\truelabel  - \big( \weights^{(M)}\big)^{T} \featurevec \big)^{2} \big\}$ 
for two choices for the \gls{learnrate} schedule: $\lrate_{\timeidx} \defeq 1/(\timeidx+5)$ or $\lrate_{\timeidx} \defeq 1/\sqrt{\timeidx+5}$. 

\noindent\refstepcounter{problem}\label{prob:computingaveragefixedpoint}\textbf{\theproblem. Computing the Average - I.}
Consider an \gls{flnetwork} with graph $\graph$ and its \gls{LapMat} $\LapMat{\graph}$. Each 
node carries a \gls{localdataset} which consists of a single measurement $\truelabel^{(\nodeidx)} \in \mathbb{R}$. 
To compute their average $(1/\nrnodes) \sum_{\nodeidx=1}^{\nrnodes} \truelabel^{(\nodeidx)}$ we try an 
iterative method that, starting from the initialization $\vu^{(0)} \defeq \big(\truelabel^{(1)},\ldots,\truelabel^{(\nrnodes)} \big)^{T} \in \mathbb{R}^{\nrnodes}$, 
repeats the update 
\begin{equation}
	\label{equ_fixedpoint_computeaverage}
	\vu^{(\iteridx+1)} =  \vu^{(\iteridx)} - \lrate  \LapMat{\graph} \vu^{(\iteridx)} \mbox{ for } \iteridx=1,2,\ldots.
\end{equation} 
Can you find a choice for $\lrate$ such that \eqref{equ_fixedpoint_computeaverage} becomes a 
fixed-point iteration \eqref{equ_def_fixed_point} with a contractive operator $\fixedpointop$. Given 
such a choice of $\lrate$, how is the limit $\lim_{\iteridx\rightarrow \infty} \vu^{(\iteridx+1)}$ 
related to the average $(1/\nrnodes) \sum_{\nodeidx=1}^{\nrnodes} \truelabel^{(\nodeidx)}$?

\noindent\refstepcounter{problem}\label{prob:computingaveragefixedpointtvmin}\textbf{\theproblem. Computing the Average - II.}
Consider the \gls{flnetwork} from Problem \ref{prob:computingaveragefixedpoint}. Try to construct an 
instance of \gls{gtvmin} for learning scalar local \glspl{modelparam} $\weight^{(\nodeidx)}$ which 
coincide, for each node $\nodeidx=1,\ldots,\nrnodes$ with the average 
$(1/\nrnodes) \sum_{\nodeidx'=1}^{\nrnodes} \truelabel^{(\nodeidx')}$. If you find such an instance of 
\gls{gtvmin}, solve it using \gls{gd}.

\noindent\refstepcounter{problem}\label{prob:projSGDforqant}\textbf{\theproblem. How to Quantize the Gradients?} 
Any \gls{ml} and \gls{fl} application that uses a digital computer to implement a \gls{gradstep} \eqref{equ_def_basic_gradstep} 
must quantize the \gls{gradient} $\nabla f(\weights)$ of the \gls{objfunc} $f(\weights)$. The quantization process 
introduces perturbations to the \gls{gradstep}. Given a fixed total budget of bits available for quantization, a key 
question arises: Should we allocate more bits (reducing quantization noise) during the initial \gls{gradstep}s or 
during the final \glspl{gradstep} in \glspl{gdmethod}?\\
Hint: See Section \ref{sec_perturbed_gd}. 

\noindent\refstepcounter{problem}\label{prob:firmnonexp}\textbf{\theproblem. When is a Gradient Step (Firmly) Non-Expansive?} 
Consider the \gls{function} $f(w) = (1/2) w^2$ and the associated \gls{gradstep} $\mathcal{G}^{(\lrate)}: w \mapsto w - \lrate \nabla f(w)$. 
Discuss the value ranges for the \gls{learnrate} $\lrate$, for which the operator $\mathcal{G}^{(\lrate)}$ is non-expansive or 
even firmly non-expansive. 

\newpage
\section{FL Algorithms}
\label{lec_flalgorithms} 

Chapter \ref{lec_fldesignprinciple} introduced \gls{gtvmin} as a flexible design principle 
for \gls{fl} methods that arise from different design choices for the \gls{localmodel}s and 
\gls{edgeweight}s of the \gls{flnetwork}. The solutions of \gls{gtvmin} are \gls{localmodel} \glspl{parameter} 
that strike a balance between the \gls{loss} incurred on \glspl{localdataset} and 
the \gls{gtv}. 

This chapter applies the \glspl{gdmethod} from Chapter \ref{lec_gradientmethods} 
to solve \gls{gtvmin}. We obtain \gls{fl} \gls{algorithm}s by implementing these optimization 
methods as message passing across the edges of the \gls{flnetwork}. These messages contain 
intermediate results of the computations carried out by \gls{fl} \glspl{algorithm}. The details of how 
this message passing is implemented physically (e.g., via short-range wireless technology) 
are beyond the scope of this book. 

Section \ref{sec_grad_step_gtvmin} studies the \gls{gradstep} for the \gls{gtvmin} instance 
obtained for training local \gls{linmodel}s. In particular, we show how the convergence rate of 
the \gls{gradstep} can be characterized by the properties of the \glspl{localdataset} and 
their \gls{flnetwork}. 

Section \eqref{sec_fl_algo_mpi} spells out the \gls{gradstep} from Section \ref{sec_grad_step_gtvmin} 
in the form of a message passing across the edges of the \gls{flnetwork}. This results in Algorithm \ref{alg_fed_gd} as a 
distributed \gls{fl} method for parametric \glspl{localmodel}. Section \ref{sec_fl_alg_sgd} generalizes 
Algorithm \ref{alg_fed_gd} by replacing the exact \gls{gradient} of local \glspl{lossfunc} with some 
approximation. One possible approximation is to use a random subset (a \gls{batch}) 
of a \gls{localdataset} to estimate the \gls{gradient}.

Section \ref{sec_fl_alg_fed_avg} discusses \gls{fl} algorithms that train a single (global) \gls{model} in a 
distributed fashion. We show how the widely-used \gls{fl} \gls{algorithm}s 
\gls{fedavg} and \gls{fedprox} are obtained from variations of \gls{projgd}, which we 
have discussed in Section \ref{sec_proj_gradient_descent}. 

Section \ref{sec_fedrelax} generalizes the \gls{gradstep}, which is the core computation 
of \gls{fl} \glspl{algorithm} for parametric \gls{model}s, to cope with non-parametric models. 
The idea is to compare the \glspl{prediction} of the \glspl{localmodel} at two 
nodes $\nodeidx$ and $\nodeidx'$ on a common test-set. By comparing their \glspl{prediction}, 
we can measure their variation across the edge $\edge{\nodeidx}{\nodeidx'}$. 

Most of the \glspl{algorithm} discussed in this chapter operate in a synchronous manner: 
All \glspl{device} must complete their \gls{localmodel} updates (e.g., \glspl{gradstep}) before 
exchanging updates simultaneously across the edges of the \gls{flnetwork}. However, 
synchronous operation can be impractical or even infeasible for certain \gls{fl} applications.
Section \ref{sec_asynch_fl_alg} explores the design of \gls{fl} \glspl{algorithm} that support 
asynchronous operation. These \glspl{algorithm} allow \glspl{device} to update 
and communicate at different times within the \gls{fl} system

\subsection{Gradient Descent for GTVMin} 
\label{sec_grad_step_gtvmin}

Consider a collection of $\nrnodes$ \glspl{localdataset} represented by the nodes $\nodes = \{ 1,\ldots,\nrnodes\}$ 
of an \gls{flnetwork} $\graph = \pair{\nodes}{\edges}$. Each undirected edge $\edge{\nodeidx}{\nodeidx'} \in \edges$ 
in \gls{flnetwork} $\graph$ has a known \gls{edgeweight} $\edgeweight_{\nodeidx,\nodeidx'}$. 
We want to learn \gls{localmodel} \glspl{parameter} $\localparams{\nodeidx}$ of a personalized 
\gls{linmodel} for each node $\nodeidx=1,\ldots,\nrnodes$. To this end, we solve the \gls{gtvmin} instance 
\begin{align} 
	\label{equ_def_gtvmin_linreg_lec5} 
	\big\{ \widehat{\weights}^{(\nodeidx)} \big\}_{\nodeidx=1}^{\nrnodes} \!\in\! \argmin_{\{ \weights^{(\nodeidx)} \}} \underbrace{\sum_{\nodeidx \in \nodes} \overbrace{(1/\localsamplesize{\nodeidx}) \normgeneric{\labelvec^{(\nodeidx)} \!-\! \featuremtx^{(\nodeidx)}  \localparams{\nodeidx}}{2}^{2}}^{\mbox{\small local \gls{loss} }\locallossfunc{\nodeidx}{\localparams{\nodeidx}}}\!+\!\regparam \sum_{\edge{\nodeidx}{\nodeidx'} \in \edges} 
		\edgeweight_{\nodeidx,\nodeidx'}  \normgeneric{\weights^{(\nodeidx)} \!-\!\weights^{(\nodeidx')}}{2}^{2}}_{=: f(\weights)}.
\end{align} 

As discussed in Chapter \ref{lec_fldesignprinciple}, the \gls{objfunc} in \eqref{equ_def_gtvmin_linreg_lec5} - 
viewed as a function of the stacked \gls{localmodel} \glspl{parameter} $\weights \defeq {\rm stack} \{ \localparams{\nodeidx}\}_{\nodeidx=1}^{\nrnodes}$ -
is a \gls{quadfunc}
\begin{align} 
	\label{equ_def_objec_gtvmin_lec_flalg}
	&\weights^{T} \left(  \begin{pmatrix}
		\mQ^{(1)} & \cdots & \mathbf{0} \\
		\vdots  & \ddots & \vdots \\
		\mathbf{0} & \cdots & \mQ^{(\nrnodes)}
	\end{pmatrix}\!+\!\regparam \LapMat{\graph}  \otimes  \mI \right) \weights
	\!+\! \big( \big( \vq^{(1)} \big)^{T},\ldots,\big( \vq^{(\nrnodes)}\big) ^{T}  \big) \weights  \\[3mm]
	&	\mbox{ with } \mQ^{(\nodeidx)}  \!=\! (1/\localsamplesize{\nodeidx})\big( \featuremtx^{(\nodeidx)} \big)^{T} \featuremtx^{(\nodeidx)} 
	\mbox{ and } \vq^{(\nodeidx)} \defeq (-2/\localsamplesize{\nodeidx}) \big(\featuremtx^{(\nodeidx)} \big)^{T} \vy^{(\nodeidx)}.  \nonumber
\end{align}
Note that \eqref{equ_def_objec_gtvmin_lec_flalg} is a special case of the generic 
\gls{quadfunc} \eqref{equ_generic_cvx_quad_function} studied in Chapter \ref{lec_gradientmethods}. 
Indeed, we obtain \eqref{equ_def_objec_gtvmin_lec_flalg} from \eqref{equ_generic_cvx_quad_function} 
for the choices  
\begin{equation}
	\nonumber
	\mQ \defeq \left(  \begin{pmatrix}
		\mQ^{(1)} & \cdots & \mathbf{0} \\
		\vdots  & \ddots & \vdots \\
		\mathbf{0} & \cdots & \mQ^{(\nrnodes)}
	\end{pmatrix}\!+\!\regparam \LapMat{\graph}  \otimes  \mI \right) \mbox{, and } \vq \defeq \big( \big( \vq^{(1)} \big)^{T},\ldots,\big( \vq^{(\nrnodes)}\big) ^{T}  \big)^{T} . 
\end{equation}   
Therefore, the discussion and analysis of \glspl{gdmethod} from Chapter \ref{lec_gradientmethods} 
also apply to \gls{gtvmin} \eqref{equ_def_gtvmin_linreg_lec5}. 
In particular, we can use the \gls{gradstep} 
\begin{align}
	\label{equ_def_basic_gradstep_lecflalg}
	\weights^{(\iteridx+1)} & \defeq \weights^{(\iteridx)} - \lrate \nabla f\big( \weights^{(\iteridx)} \big) \nonumber \\  
	& \stackrel{\eqref{equ_def_objec_gtvmin_lec_flalg}}{=}  \weights^{(\iteridx)} - \lrate \big( 2 \mQ \weights^{(\iteridx)} + \vq  \big) 
\end{align}
to iteratively compute an approximate solution $\widehat{\weights}$ to \eqref{equ_def_gtvmin_linreg_lec5}. 
This solution consists of learned \gls{localmodel} \glspl{parameter} $\estlocalparams{\nodeidx}$, i.e., 
$\widehat{\weights} = {\rm stack} \{ \estlocalparams{\nodeidx} \}$. Section \ref{sec_fl_algo_mpi} 
formulates the \gls{gradstep} \eqref{equ_def_basic_gradstep_lecflalg} directly in 
terms of \gls{localmodel} \glspl{parameter}, resulting in a message passing over the \gls{flnetwork} $\graph$. 

According to the convergence analysis in Chapter \ref{lec_gradientmethods}, the convergence rate 
of the iterations \eqref{equ_def_basic_gradstep_lecflalg} is determined by the \glspl{eigenvalue} 
$\eigval{\featureidx}(\mQ)$ of the matrix $\mQ$ in \eqref{equ_def_objec_gtvmin_lec_flalg}. 
In general, these \glspl{eigenvalue} depend on the \glspl{eigenvalue} $\eigval{\featureidx}\big(\mQ^{(\nodeidx)}\big)$ 
as well as the \glspl{eigenvalue} $\eigval{\featureidx}\big( \LapMat{\graph}  \big)$ of the \gls{LapMat} $\LapMat{\graph}$. 
In particular, we will use the following two summary parameters 
\begin{equation}
	\label{equ_summary_eigvals_Q_i}
	\maxeigvallocalQ \defeq \max_{\nodeidx=1,\ldots,\nrnodes} \eigval{\dimlocalmodel}\big( \mQ^{(\nodeidx)} \big)\mbox{, and }  
	\avgmineigvallocalQ \defeq \eigval{1}\bigg((1/\nrnodes) \sum_{\nodeidx=1}^{\nrnodes} \mQ^{(\nodeidx)} \bigg).
\end{equation}

We first present an upper bound $\upperboundeigval$ (see \eqref{equ_def_lower_upper_bound_eigvals}) 
on the \gls{eigenvalue}s of the matrix $\mQ$ in \eqref{equ_def_objec_gtvmin_lec_flalg}.  
\begin{prop} 
	\label{prop_upper_bound} 
	The \glspl{eigenvalue} of $\mQ$ in \eqref{equ_def_objec_gtvmin_lec_flalg} are upper-bounded as   
	\begin{align}
		\label{equ_upper_bound_eigval_Q_gtvmin_lin} 
		\eigval{\featureidx}( \mQ )&  \leq  \maxeigvallocalQ  + \regparam \eigval{\nrnodes}\big( \LapMat{\graph}  \big) \nonumber \\ 
		& \leq \underbrace{\maxeigvallocalQ+ 2 \regparam \maxnodedegree^{(\graph)}}_{=: \upperboundeigval} \mbox{ , for } 
		\featureidx=1,\ldots,\dimlocalmodel\nrnodes.
	\end{align} 
\end{prop} 
\begin{proof} 
See Section \ref{proof_upper_bound_eigvals_Q_gtvmin}. 
\end{proof} 
\noindent Note how the upper bound \eqref{equ_upper_bound_eigval_Q_gtvmin_lin} involves properties of 
\begin{itemize} 
	\item the \glspl{localdataset}, via $\lambda_{\rm max}$ (see \eqref{equ_summary_eigvals_Q_i}), 
	\item the \gls{flnetwork}, via the maximum \gls{nodedegree} $\maxnodedegree^{(\graph)}$ (see \eqref{equ_def_max_node_degree}), and 
	\item the \gls{gtvmin} parameter $\regparam$. 
\end{itemize}

The next result offers a lower bound on the \glspl{eigenvalue} $\eigval{\featureidx}(\mQ)$. 
\begin{prop}
	\label{prop_lower_bound_eigvals_gtvmin_linreg}
	Consider the matrix $\mQ$ in \eqref{equ_def_objec_gtvmin_lec_flalg}. 
	If  $\eigval{2}\big(\LapMat{\graph} \big)>0$ (i.e., the \gls{flnetwork} in \eqref{equ_def_gtvmin_linreg_lec5} is connected) 
	and $\bar{\lambda}_{\rm min} > 0$ (i.e., the average of the matrices $\mQ^{(\nodeidx)}$ is non-singular), 
	then the matrix $\mQ$ is invertible and its smallest \gls{eigenvalue} is lower bounded as 
	\begin{equation} 
		\label{equ_lower_bound_eigval_Q_gtvmin_lin}
		\eigval{1}(\mQ) \geq \frac{1}{1+\rho^{2}} {\rm min} \{ \eigval{2}\big( \LapMat{\graph} \big)  \regparam \rho^{2}, \avgmineigvallocalQ/2 \}.
	\end{equation} 
	Here, we used the shorthand $\rho \defeq \avgmineigvallocalQ/(4 \maxeigvallocalQ)$ (see \eqref{equ_summary_eigvals_Q_i}). 
\end{prop} 
\begin{proof} 
	See Section \ref{proof_lower_bound_eigvals_Q_gtvmin}. 
\end{proof} 
Proposition \ref{prop_upper_bound}  and Proposition \ref{prop_lower_bound_eigvals_gtvmin_linreg}
provide some guidance for the design choices of \gls{gtvmin}. According to the convergence 
analysis of \glspl{gdmethod} in Chapter \ref{lec_gradientmethods}, the \gls{eigenvalue} $\eigval{1}\big( \mQ \big)$ 
should be close to $\eigval{\dimlocalmodel \nrnodes}\big( \mQ\big)$ to ensure fast convergence. This suggests 
to favour \glspl{flnetwork} $\graph$ resulting in a small ratio between the upper bound 
\eqref{equ_upper_bound_eigval_Q_gtvmin_lin} and the lower bound \eqref{equ_lower_bound_eigval_Q_gtvmin_lin}.  
A small ratio between these bounds, in turn, requires a large \gls{eigenvalue} $\eigval{2}\big( \LapMat{\graph} \big)$ 
and small \gls{nodedegree} $\maxnodedegree^{(\graph)}$.\footnote{The are constructions of \glspl{graph} 
	with a prescribed value of $\maxnodedegree^{(\graph)}$ such that $\eigval{2}\big( \LapMat{\graph} \big)$ 
	is maximal \cite{NEURIPS2021_74e1ed8b,BoydFastestMixing2004}.} 

The bounds in \eqref{equ_upper_bound_eigval_Q_gtvmin_lin} and \eqref{equ_lower_bound_eigval_Q_gtvmin_lin} also 
depend on the \gls{gtvmin} parameter $\regparam$. While these bounds might provide some 
guidance for the choice of $\regparam$, the exact dependence of the convergence speed of 
\eqref{equ_def_basic_gradstep_lecflalg} on $\regparam$ is complicated. For a fixed value of 
\gls{learnrate} in \eqref{equ_def_basic_gradstep_lecflalg}, using larger values for $\regparam$ 
might slow down the convergence of \eqref{equ_def_basic_gradstep_lecflalg} for some collection 
of \gls{localdataset}s but speed up the convergence of \eqref{equ_def_basic_gradstep_lecflalg} for 
another collection of \gls{localdataset}s (see Exercise \ref{prob:convgdlinalg}). 

\subsection{Message Passing Implementation}
\label{sec_fl_algo_mpi}

We now discuss in more detail the implementation of \glspl{gdmethod} to 
solve the \gls{gtvmin} instances with a differentiable \gls{objfunc} $f(\weights)$. 
One such instance is \gls{gtvmin} for local \glspl{linmodel} (see \eqref{equ_def_gtvmin_linreg_lec5}). 
The core of \glspl{gdmethod} is the \gls{gradstep}
\begin{equation} 
	\label{equ_def_gradstep_MPI}
	\weights^{(\iteridx+1)} \defeq  \weights^{(\iteridx)} - \lrate \nabla f\big( \weights^{(\iteridx)} \big). 
\end{equation} 
The iterate $\weights^{(\iteridx)}$ contains \gls{localmodel} \glspl{parameter} $\localparamsiter{\nodeidx}{\iteridx}$, 
\begin{equation}
	\nonumber
	\weights^{(\iteridx)} =: {\rm stack} \big\{ \localparamsiter{\nodeidx}{\iteridx} \big\}_{\nodeidx=1}^{\nrnodes}.
\end{equation} 
Inserting \eqref{equ_def_gtvmin_linreg_lec5} into \eqref{equ_def_gradstep_MPI}, we obtain the \gls{gradstep} 
\begin{align} 
	\label{equ_gd_step_node_i_direct}
	\localparamsiter{\nodeidx}{\iteridx+1}\!\defeq\!\localparamsiter{\nodeidx}{\iteridx}& - \lrate \bigg[ \underbrace{(2/\localsamplesize{\nodeidx}) \big(\featuremtx^{(\nodeidx)}\big)^{T}\big(\featuremtx^{(\nodeidx)}\localparamsiter{\nodeidx}{\iteridx}\!-\!\labelvec^{(\nodeidx)}\big)}_{\mbox{(I)}} \nonumber \\ 
	& \hspace*{30mm}+\underbrace{2\regparam  \hspace*{-3mm} \sum_{\nodeidx' \in \neighbourhood{\nodeidx} } \hspace*{-2mm}\edgeweight_{\nodeidx,\nodeidx'} 
		\big( \localparamsiter{\nodeidx}{\iteridx}\!-\! \localparamsiter{\nodeidx'}{\iteridx} \big)}_{\mbox{(II)}} \bigg].
\end{align}
We slightly modify this \gls{gradstep} by allowing for different \glspl{learnrate} $\lrate_{\iteridx,\nodeidx}$ 
at different nodes $\nodeidx$ and iterations $\iteridx$, 
\begin{align} 
	\label{equ_gd_step_node_i}
	\localparamsiter{\nodeidx}{\iteridx+1}\!\defeq\!\localparamsiter{\nodeidx}{\iteridx}& - \lrate_{\iteridx,\nodeidx} \bigg[ \underbrace{(2/\localsamplesize{\nodeidx}) \big(\featuremtx^{(\nodeidx)}\big)^{T}\big(\featuremtx^{(\nodeidx)}\localparamsiter{\nodeidx}{\iteridx}\!-\!\labelvec^{(\nodeidx)}\big)}_{\mbox{(I)}} \nonumber \\ 
	& \hspace*{30mm}+\underbrace{2\regparam  \hspace*{-3mm} \sum_{\nodeidx' \in \neighbourhood{\nodeidx} } \hspace*{-2mm}\edgeweight_{\nodeidx,\nodeidx'} 
		\big( \localparamsiter{\nodeidx}{\iteridx}\!-\! \localparamsiter{\nodeidx'}{\iteridx} \big)}_{\mbox{(II)}} \bigg].
\end{align}

The update \eqref{equ_gd_step_node_i} consists of two components, denoted (I) and (II). Component (I) 
reflects the local \gls{lossfunc} at node $\nodeidx$ while component (II) couples node $\nodeidx$ with its 
\glspl{neighbor} $\nodeidx' \in \neighbourhood{\nodeidx}$. 
In particular, component (I) is the \gls{gradient} $\nabla \locallossfunc{\nodeidx}{\localparamsiter{\nodeidx}{\iteridx}}$ 
of the local \gls{loss} $\locallossfunc{\nodeidx}{\localparams{\nodeidx}} \defeq (1/\localsamplesize{\nodeidx}) \normgeneric{\vy^{(\nodeidx)} - \featuremtx^{(\nodeidx)} \localparams{\nodeidx}}{2}^{2}$. 
Component (I) drives the updated \gls{localmodel} \glspl{parameter} $ \localparamsiter{\nodeidx}{\iteridx+1}$ 
towards the minimum of $\locallossfunc{\nodeidx}{\cdot}$, i.e., having a small deviation between \gls{label}s 
$\truelabel^{(\nodeidx,\sampleidx)}$ and the \gls{prediction}s $\big( \localparamsiter{\nodeidx}{\iteridx+1}\big)^{T} \featurevec^{(\nodeidx,\sampleidx)}$. 
Note that we can rewrite the component (I) in \eqref{equ_gd_step_node_i}, as 
\begin{equation} 
	\label{equ_def_sum_gradient_component_I}
	(2/\localsamplesize{\nodeidx}) \sum_{\sampleidx=1}^{\localsamplesize{\nodeidx}} \featurevec^{(\nodeidx,\sampleidx)} \big( \truelabel^{(\nodeidx,\sampleidx)} - \big( \featurevec^{(\nodeidx,\sampleidx)}\big)^{T}  \localparamsiter{\nodeidx}{\iteridx}  \big). 
\end{equation} 

The component (II) in \eqref{equ_gd_step_node_i} 
The purpose of component (II) in \eqref{equ_gd_step_node_i} is to force the \gls{localmodel} \glspl{parameter} 
to be similar across an edge $\edge{\nodeidx}{\nodeidx'}$ with large weight $\edgeweight_{\nodeidx,\nodeidx'}$. 
We control the relative importance of (II) and (I) using the \gls{gtvmin} parameter $\regparam$: Choosing 
a large value for $\regparam$ puts more emphasis on enforcing similar \gls{localmodel} \glspl{parameter} 
across the edges. Using a smaller $\regparam$ puts more emphasis on learning \gls{localmodel} \glspl{parameter} 
delivering accurate \gls{prediction}s (incurring a small \gls{loss}) on 
the \gls{localdataset}. 
\begin{figure}[htbp]
	\begin{center}
		\begin{tikzpicture}[scale=0.6]
			\coordinate (i1) at (0,0);
			\coordinate (i2) at (5,2);
			\coordinate (i3) at (5,-2);
			
			\draw [fill] (i1) circle [radius=0.2] node[below=5pt] {$\localparamsiter{1}{\iteridx}$};
			\draw [fill] (i2) circle [radius=0.2] node[below=5pt] {$\localparamsiter{2}{\iteridx}$};
			\draw [fill] (i3) circle [radius=0.2] node[below=5pt] {$\localparamsiter{3}{\iteridx}$};
			
			\draw[line width=0.5mm] (i1) -- (i2) node[midway, above] {$\edgeweight_{1,2}$};
			\draw[line width=0.5mm] (i1) -- (i3) node[midway, above] {$\edgeweight_{1,3}$};
		\end{tikzpicture}
		\vspace*{-10mm}
	\end{center}
	\caption{\label{fig_gd_step_node_i} At the beginning of iteration $\iteridx$, node $\nodeidx=1$ collects the current \gls{localmodel} 
		\glspl{parameter} $\localparamsiter{2}{\iteridx}$ and $\localparamsiter{3}{\iteridx}$ from its \glspl{neighbor}. 
		Then, it computes the \gls{gradstep} \eqref{equ_gd_step_node_i} to obtain the new \gls{localmodel} \glspl{parameter} $\localparamsiter{1}{\iteridx+1}$. 
		These updated \glspl{parameter} are then used in the next iteration for the local updates at the \glspl{neighbor} $\nodeidx=2,3$.}
\end{figure} 

The execution of the \gls{gradstep} \eqref{equ_gd_step_node_i} requires only local 
information at node $\nodeidx$. Indeed, the update \eqref{equ_gd_step_node_i} at 
node $\nodeidx$ depends only on its current \glspl{modelparam} $\localparamsiter{\nodeidx}{\iteridx}$, 
the local \gls{lossfunc} $\locallossfunc{\nodeidx}{\cdot}$, the \glspl{neighbor}' 
\glspl{modelparam} $\localparamsiter{\nodeidx'}{\iteridx}$, for 
$\nodeidx' \in \neighbourhood{\nodeidx}$, and the corresponding \gls{edgeweight}s $\edgeweight_{\nodeidx,\nodeidx'}$ 
(see Figure \ref{fig_gd_step_node_i}). In particular, the update \eqref{equ_gd_step_node_i} 
does not depend on any properties \iteridx or \gls{edgeweight}s) of the \gls{flnetwork} beyond the \glspl{neighbor} $\neighbourhood{\nodeidx}$. 

We obtain Algorithm \ref{alg_fed_gd} by repeating the \gls{gradstep} \eqref{equ_gd_step_node_i}, simultaneously 
for each node $\nodeidx \in \nodes$, until a \gls{stopcrit} is met. Algorithm \ref{alg_fed_gd} allows for 
potentially different \gls{learnrate}s $\lrate_{\iteridx,\nodeidx}$ at different nodes $\nodeidx$ and 
iterations $\iteridx$. 
\begin{algorithm}[htbp]
	\caption{FedGD for Local Linear Models}
	\label{alg_fed_gd}
	{\bf Input}: \gls{flnetwork} $\graph$; 	\gls{gtv} parameter $\regparam$; \gls{learnrate} $\lrate_{\iteridx,\nodeidx}$; \\ 
	local \gls{dataset} $\localdataset{\nodeidx} = \left\{ \pair{\featurevec^{(\nodeidx,1)}}{\truelabel^{(\nodeidx,1)}}; \ldots,\pair{\featurevec^{(\nodeidx,\localsamplesize{\nodeidx})}}{\truelabel^{(\nodeidx,\localsamplesize{\nodeidx})}} \right\}$ 
	for each $\nodeidx$; some \gls{stopcrit}.
	
	{\bf Output}: \gls{linmodel} \glspl{parameter} $\estlocalparams{\nodeidx}$ for each node $\nodeidx \in \nodes$ \\ 
	{\bf Initialize}: $\iteridx\!\defeq\!0$; $\localparamsiter{\nodeidx}{0}\!\defeq\!{\bf 0}$
	\begin{algorithmic}[1]
		\While{\gls{stopcrit} is not satisfied}
		\For{all nodes $ \nodeidx \in \nodes$ (simultaneously)}
		\State share \gls{localmodel} \glspl{parameter} $\localparamsiter{\nodeidx}{\iteridx}$ with \glspl{neighbor} $\nodeidx'\!\in\!\neighbourhood{\nodeidx}$ \label{equ_def_update_step_fedgd_sharing_linmodel}
		\State update local \glspl{modelparam} via \eqref{equ_gd_step_node_i}  \label{equ_def_update_local}
		\EndFor
		\State increment iteration counter: $\iteridx\!\defeq\!\iteridx\!+\!1$
		\EndWhile
		\State $\estlocalparams{\nodeidx} \defeq \localparamsiter{\nodeidx}{\iteridx}$ for all nodes $\nodeidx \in \nodes$
	\end{algorithmic}
\end{algorithm}
It is important to note that Algorithm \ref{alg_fed_gd} requires a synchronous (simultaneous) execution of the 
updates \eqref{equ_gd_step_node_i} at all nodes $\nodeidx \in \nodes$ \cite{ParallelDistrBook,DistributedSystems}. 
Loosely speaking, all nodes $\nodeidx$ rely on a single global clock that maintains the 
current iteration counter $\iteridx$ \cite{InternetTimeSync1991}. 

At the beginning of iteration $\iteridx$, each node $\nodeidx \in \nodes$ sends its 
current \glspl{modelparam} $\localparamsiter{\nodeidx}{\iteridx}$ to their \glspl{neighbor} 
$\nodeidx'\!\in\!\neighbourhood{\nodeidx}$. Then, each node $\nodeidx \in \nodes$ 
updates their \glspl{modelparam} according to \eqref{equ_gd_step_node_i}, resulting 
in the updated \glspl{modelparam} $\localparamsiter{\nodeidx}{\iteridx+1}$. As soon as 
these local updates are completed, the global clock increments the counter 
$\iteridx \mapsto \iteridx+1$ and triggers the next iteration to be executed by all nodes. 
Figure \ref{fig_message_passing_fed_gd} illustrates the alternating execution of 
message passing and local updates of Algorithm \ref{alg_fed_gd}. 

\begin{figure}
	\begin{center} 
		\begin{tikzpicture}[
			node/.style={circle, draw, fill=black, minimum size=0.4cm, inner sep=0},
			undirected/.style={draw, thick, line width=1.4},
			message_arrow/.style={draw, ->, dashed, shorten >=1pt, shorten <=1pt},
			time_step/.style={font=\bfseries}
			]
			\node[node, label=right:$\nodeidx$] (n1_t1) at (0, 1.5) {};
			\node[node, label=right:$\nodeidx'$] (n2_t1) at (0, -1.5) {};
			\draw[undirected] (n1_t1) -- (n2_t1);
			
			\draw[message_arrow] (n1_t1) to[out=315, in=45]node[midway, right] {$\localparamsiter{\nodeidx}{\iteridx}$} (n2_t1);
			\draw[message_arrow] (n2_t1) to[out=135, in=225] node[midway, left]{$\localparamsiter{\nodeidx'}{\iteridx}$} (n1_t1);

			
			\node[node, label=right:$\nodeidx$] (n1_t2) at (5, 1.5) {};
			\node[node, label=right:$\nodeidx'$] (n2_t2) at (5, -1.5) {};
			\node[] [above=0.2cm of n1_t2] {compute $\localparamsiter{\nodeidx}{\iteridx+1}$};
			\node[] [below=0.2cm of n2_t2] {compute $\localparamsiter{\nodeidx'}{\iteridx+1}$};
			\draw[undirected] (n1_t2) -- (n2_t2);
			\draw[] (n2_t2) -- (n1_t2) node[midway, left] {$\edgeweight_{\nodeidx,\nodeidx'}$};	
			
			\node[node, label=right:$\nodeidx$] (n1_t3) at (10, 1.5) {};
			\node[node, label=right:$\nodeidx'$] (n2_t3) at (10, -1.5) {};
			\draw[undirected] (n1_t3) -- (n2_t3);
			\draw[message_arrow] (n1_t3) to[out=315, in=45] node[midway, right] {$\localparamsiter{\nodeidx}{\iteridx+1}$} (n2_t3);
			\draw[message_arrow] (n2_t3) to[out=135, in=225] node[midway, left] {$\localparamsiter{\nodeidx'}{\iteridx+1}$} (n1_t3);
			
		\end{tikzpicture}
	\end{center}
	
	\caption{Algorithm \ref{alg_fed_gd} alternates between message passing across the edges of 
		the \gls{flnetwork} (left and right) and updates of local \glspl{modelparam} (centre). \label{fig_message_passing_fed_gd}}
	
\end{figure} 

The implementation of Algorithm \ref{alg_fed_gd} in real-world computational infrastructures might 
incur deviations from the exact synchronous execution of \eqref{equ_gd_step_node_i} \cite[Sec. 10]{DistrAlgLectureNotes}. 
This deviation can be modelled as a perturbation of the \gls{gradstep} \eqref{equ_def_gradstep_MPI} 
and therefore analyzed using the concepts of Section \ref{sec_perturbed_gd} on perturbed \gls{gd}. 
Section \ref{sec_techrobustness_fl} will also discuss the effect of imperfect computation in the context 
of key requirements for trustworthy \gls{fl}.   

We close this section by generalizing Algorithm \ref{alg_fed_gd} which is limited to \glspl{flnetwork} using 
local \glspl{linmodel}. This generalization, summarized in Algorithm \ref{alg_fed_gd_general}, 
can be used to train parametric \glspl{localmodel} $\localmodel{\nodeidx}$ with a \gls{differentiable} 
\gls{lossfunc} $\locallossfunc{\nodeidx}{\localparams{\nodeidx}}$, for $\nodeidx=1,\ldots,\nrnodes$. 

\begin{algorithm}[htbp]
	\caption{FedGD for Parametric Local Models}
	\label{alg_fed_gd_general}
	{\bf Input}: \gls{flnetwork} $\graph$; 	\gls{gtv} parameter $\regparam$; \gls{learnrate} $\lrate_{\iteridx,\nodeidx}$ \\ 
	local \gls{lossfunc} $\locallossfunc{\nodeidx}{\localparams{\nodeidx}}$ 
	for each $\nodeidx=1,\ldots,\nrnodes$; some \gls{stopcrit}.
	
	{\bf Output}: \gls{linmodel} \glspl{parameter} $\estlocalparams{\nodeidx}$ for each node $\nodeidx \in \nodes$ \\ 
	{\bf Initialize}: $\iteridx\!\defeq\!0$; $\localparamsiter{\nodeidx}{0}\!\defeq\!{\bf 0}$
	\begin{algorithmic}[1]
		\While{\gls{stopcrit} is not satisfied}
		\For{all nodes $ \nodeidx \in \nodes$ (simultaneously)}
		\State share \gls{localmodel} \glspl{parameter} $\localparamsiter{\nodeidx}{\iteridx}$ with \glspl{neighbor} $\nodeidx'\!\in\!\neighbourhood{\nodeidx}$ \label{equ_def_update_step_fedgd_sharing}
		\State update local \glspl{modelparam} via \label{equ_gdstep_fedgd_general}
		$$	\localparamsiter{\nodeidx}{\iteridx+1}\!\defeq\!\localparamsiter{\nodeidx}{\iteridx}- \lrate_{\iteridx,\nodeidx} \bigg[\nabla \locallossfunc{\nodeidx}{\localparamsiter{\nodeidx}{\iteridx}}\!+\!2\regparam  \hspace*{-3mm} \sum_{\nodeidx' \in \neighbourhood{\nodeidx} }\edgeweight_{\nodeidx,\nodeidx'} 
		\big( \localparamsiter{\nodeidx}{\iteridx}\!-\!\localparamsiter{\nodeidx'}{\iteridx} \big) \bigg].$$  
		\EndFor
		\State increment iteration counter: $\iteridx\!\defeq\!\iteridx\!+\!1$
		\EndWhile
		\State $\estlocalparams{\nodeidx} \defeq \localparamsiter{\nodeidx}{\iteridx}$ for all nodes $\nodeidx \in \nodes$
	\end{algorithmic}
\end{algorithm}

\subsection{FedSGD} 
\label{sec_fl_alg_sgd}

Consider Algorithm \ref{alg_fed_gd} for training local \glspl{linmodel} $\hypothesis^{(\nodeidx)}(\featurevec) =\featurevec^{T} \localparams{\nodeidx}$ 
for each node $\nodeidx = 1,\ldots,\nrnodes$ of an \gls{flnetwork}. Note that step \ref{equ_def_update_local} 
of Algorithm \ref{alg_fed_gd} requires to compute the sum \eqref{equ_def_sum_gradient_component_I}. 
It might be infeasible to compute this sum exactly, e.g., when \glspl{localdataset} are generated by 
remote \glspl{device} with limited connectivity. It is then useful to approximate the sum by 
\begin{align}
	\label{equ_def_loss_func_linreg_gradient_stoch}
	\underbrace{(2/\batchsize) 
		\sum_{\sampleidx\in \batch} \featurevec^{(\nodeidx,\sampleidx)}  \big( \truelabel^{(\nodeidx,\sampleidx)} - \big( \featurevec^{(\nodeidx,\sampleidx)}\big)^{T}  \localparamsiter{\nodeidx}{\iteridx} \big)}_{\approx \eqref{equ_def_sum_gradient_component_I}}. 
\end{align} 
The approximation \eqref{equ_def_loss_func_linreg_gradient_stoch} uses a 
subset (so-called \emph{\gls{batch}}) 
$$\batch = \left\{ \pair{\featurevec^{(\sampleidx_{1})}}{\truelabel^{(\sampleidx_{1})}},\ldots, \pair{\featurevec^{(\sampleidx_{\batchsize})}}{\truelabel^{(\sampleidx_{\batchsize})}}  \right\}$$ 
of $\batchsize$ randomly chosen \glspl{datapoint} from $\localdataset{\nodeidx}$. While 
\eqref{equ_def_sum_gradient_component_I} requires summing over $\samplesize$ \glspl{datapoint}, 
the approximation requires to sum over $\batchsize$ (typically $\batchsize \ll \samplesize$) \glspl{datapoint}. 

Inserting the approximation \eqref{equ_def_loss_func_linreg_gradient_stoch} into the \gls{gradstep} \eqref{equ_gd_step_node_i} 
yields the approximate \gls{gradstep} 
\begin{align} 
	\label{equ_sgd_step_node_i}
	\localparamsiter{\nodeidx}{\iteridx+1}\!\defeq\!\localparamsiter{\nodeidx}{\iteridx}& - \lrate_{\iteridx,\nodeidx} \bigg[ \underbrace{(2/\batchsize) \sum_{\sampleidx\in \batch}   \featurevec^{(\nodeidx,\sampleidx)}  \bigg(\big( \featurevec^{(\nodeidx,\sampleidx)} \big)^{T} \localparamsiter{\nodeidx}{\iteridx}\!-\!\truelabel^{(\nodeidx,\sampleidx)}  \bigg)}_{\approx \eqref{equ_def_sum_gradient_component_I}} \nonumber \\ 
	&\hspace*{10mm}+ 2\regparam  \hspace*{-3mm} \sum_{\nodeidx' \in \neighbourhood{\nodeidx} } \hspace*{-2mm}\edgeweight_{\nodeidx,\nodeidx'} 
	\big( \localparamsiter{\nodeidx}{\iteridx}\!-\! \localparamsiter{\nodeidx'}{\iteridx} \big)\bigg].
\end{align}

We obtain Algorithm \ref{alg_fed_sgd} from Algorithm \ref{alg_fed_gd} by replacing the 
\gls{gradstep} \eqref{equ_gd_step_node_i} with the approximation \eqref{equ_sgd_step_node_i}. 

\begin{algorithm}[htbp]
	\caption{FedSGD for Local Linear Models}
	\label{alg_fed_sgd}
	{\bf Input}: \gls{flnetwork} $\graph$; 	\gls{gtv} parameter $\regparam$; \gls{learnrate} $\lrate_{\iteridx,\nodeidx}$; \\ 
	local \glspl{dataset} $\localdataset{\nodeidx} = \left\{ \pair{\featurevec^{(\nodeidx,1)}}{\truelabel^{(\nodeidx,1)}}, \ldots,\pair{\featurevec^{(\nodeidx,\localsamplesize{\nodeidx})}}{\truelabel^{(\nodeidx,\localsamplesize{\nodeidx})}} \right\}$ 
	for each node $\nodeidx$; \gls{batch} size $\batchsize$; some \gls{stopcrit}. \\ 
	{\bf Output}: \gls{linmodel} \glspl{parameter} $\estlocalparams{\nodeidx}$ at each node $\nodeidx \in \nodes$ \\ 
	{\bf Initialize}: $\iteridx\!\defeq\!0$; $\localparamsiter{\nodeidx}{0}\!\defeq\!{\bf 0}$
	\begin{algorithmic}[1]
		\While{\gls{stopcrit} is not satisfied}
		\For{all nodes $ \nodeidx \in \nodes$ (simultaneously)}
		\State share local \glspl{modelparam} $\localparamsiter{\nodeidx}{\iteridx}$ with all \glspl{neighbor} $\nodeidx' \in \neighbourhood{\nodeidx}$ \label{equ_def_update_step_fedsgd_sharing_lin}
		\State draw fresh \gls{batch} $\batch^{(\nodeidx)} \defeq \{ \sampleidx_{1},\ldots,\sampleidx_{\batchsize} \} $  \label{equ_random_sampling_fl_sgd}
		\State update local \glspl{modelparam} via \eqref{equ_sgd_step_node_i}
		\EndFor
		\State increment iteration counter $\iteridx\!\defeq\!\iteridx\!+\!1$
		\EndWhile
		\State $\estlocalparams{\nodeidx} \defeq \localparamsiter{\nodeidx}{\iteridx}$ for all nodes $\nodeidx \in \nodes$
	\end{algorithmic}
\end{algorithm}

We close this section by generalizing Algorithm \ref{alg_fed_sgd} which is limited \glspl{flnetwork} using 
local \glspl{linmodel}. This generalization, summarized in Algorithm \ref{alg_fed_sgd_general}, 
can be used to train parametric \glspl{localmodel} $\localmodel{\nodeidx}$ with a \gls{differentiable} 
\gls{lossfunc} $\locallossfunc{\nodeidx}{\localparams{\nodeidx}}$, for $\nodeidx=1,\ldots,\nrnodes$. 
Algorithm \ref{alg_fed_sgd_general} does not require these local \gls{lossfunc} themselves, 
but only an oracle $\mathbf{g}^{(\nodeidx)}(\cdot)$ for each node $\nodeidx=1,\ldots,\nrnodes$. 
For a given vector $\localparams{\nodeidx}$, the oracle at node $\nodeidx$ delivers an 
approximate \gls{gradient} (or estimate) $\mathbf{g}^{(\nodeidx)}(\localparams{\nodeidx}) \approx \nabla \locallossfunc{\nodeidx}{\localparams{\nodeidx}}$. 
The analysis of Algorithm \ref{alg_fed_sgd_general} can be facilitated by a \gls{probmodel} which 
interprets the oracle output $\mathbf{g}^{(\nodeidx)}(\localparams{\nodeidx})$ as the \gls{realization} 
of a \gls{rv}. Under such a \gls{probmodel}, we refer to an oracle as unbiased if $\expect \big\{ \mathbf{g}^{(\nodeidx)}(\localparams{\nodeidx}) \big\} =  \nabla \locallossfunc{\nodeidx}{\localparams{\nodeidx}}$. 

\begin{algorithm}[htbp]
	\caption{FedSGD for Parametric Local Models}
	\label{alg_fed_sgd_general}
	{\bf Input}: \gls{flnetwork} $\graph$; 	\gls{gtv} parameter $\regparam$; \gls{learnrate} $\lrate_{\iteridx,\nodeidx}$ \\ 
	\gls{gradient} oracle $\mathbf{g}^{(\nodeidx)}\big(\cdot\big)$ for each node $\nodeidx=1,\ldots,\nrnodes$; some \gls{stopcrit}.
	
	{\bf Output}: \gls{linmodel} \glspl{parameter} $\estlocalparams{\nodeidx}$ for each node $\nodeidx \in \nodes$ \\ 
	{\bf Initialize}: $\iteridx\!\defeq\!0$; $\localparamsiter{\nodeidx}{0}\!\defeq\!{\bf 0}$
	\begin{algorithmic}[1]
		\While{\gls{stopcrit} is not satisfied}
		\For{all nodes $ \nodeidx \in \nodes$ (simultaneously)}
		\State share \gls{localmodel} \glspl{parameter} $\localparamsiter{\nodeidx}{\iteridx}$ with \glspl{neighbor} $\nodeidx'\!\in\!\neighbourhood{\nodeidx}$ \label{equ_def_update_step_fedsgd_sharing}
		\State update local \glspl{modelparam} via \label{equ_gdstep_fedsgd_general}
		$$	\localparamsiter{\nodeidx}{\iteridx+1}\!\defeq\!\localparamsiter{\nodeidx}{\iteridx}- \lrate_{\iteridx,\nodeidx} \bigg[\mathbf{g}^{(\nodeidx)}\big(\localparamsiter{\nodeidx}{\iteridx}\big)\!+\!2\regparam  \hspace*{-3mm} \sum_{\nodeidx' \in \neighbourhood{\nodeidx} }\edgeweight_{\nodeidx,\nodeidx'} 
		\big( \localparamsiter{\nodeidx}{\iteridx}\!-\!\localparamsiter{\nodeidx'}{\iteridx} \big) \bigg].$$  
		\EndFor
		\State increment iteration counter: $\iteridx\!\defeq\!\iteridx\!+\!1$
		\EndWhile
		\State $\estlocalparams{\nodeidx} \defeq \localparamsiter{\nodeidx}{\iteridx}$ for all nodes $\nodeidx \in \nodes$
	\end{algorithmic}
\end{algorithm}

\newpage 
\subsection{FedAvg}
\label{sec_fl_alg_fed_avg} 

Consider an \gls{fl} method that learns \glspl{modelparam} $\widehat{\weights} \in \mathbb{R}^{\dimlocalmodel}$ 
of a single (global) \gls{linmodel} from a de-centralized collection 
of \glspl{localdataset} $\localdataset{\nodeidx}$, $\nodeidx=1,\ldots,\nrnodes$.
\footnote{This setting is a special case of \gls{hfl} which we discuss in 
Section \ref{sec_horizontal_fl}.} How can we learn $\widehat{\weights}$ 
without exchanging \glspl{localdataset}, but instead only exchanging 
updates for the \glspl{modelparam}?

One approach is to apply Algorithm \ref{alg_fed_gd} to \gls{gtvmin} 
\eqref{equ_def_gtvmin_linreg_lec5} with a sufficiently large $\regparam$. 
According to our analysis in Chapter \ref{lec_fldesignprinciple} (specifically 
Proposition \ref{prop_second_component_error}), if $\regparam$ is sufficiently 
large, then the \gls{gtvmin} solutions $\estlocalparams{\nodeidx}$ are almost 
identical across all nodes $\nodeidx \in \nodes$. We can interpret the \gls{localmodel} 
\glspl{parameter} delivered by \gls{gtvmin} as a local copy of the global 
\glspl{modelparam}. 

Note that the bound in Proposition \ref{prop_second_component_error} only applies if the 
\gls{flnetwork} (used in \gls{gtvmin}) is connected. One example of a connected \gls{flnetwork} 
is the star as depicted in Figure \ref{fig:star_graph}. Here, we choose one node $\nodeidx=1$ 
as a centre node that is connected by an edge with weight $\edgeweight_{1,\nodeidx}$ 
to the remaining nodes $\nodeidx=2,\ldots,\nrnodes$. The star \gls{graph} uses the minimum 
number of edges required to connect all $\nrnodes$ nodes \cite{DiestelGT}.

\begin{figure}[htbp]
	\centering
	\begin{tikzpicture}[scale=1]
		\def\radius{2} 
		\node[draw, circle, fill=black, label=right:{}] (C) at (0, 0) {};
		\foreach \i in {1, 2, 3, 4, 5, 6, 7, 8} {
			\node[draw, circle, fill=black] (N\i) at ({\radius*cos(45*(\i-1))}, {\radius*sin(45*(\i-1))}) {};
			\draw [-, black] (C) -- (N\i); 
		}
		\node[above right=0.05cm and 0.05cm of N1, font=\fontsize{12}{0}\selectfont] {$\localdataset{\nodeidx}$};
		
		\draw [-, black] (C) -- (N1) node[midway, above=0.005cm, yshift=-5pt, sloped, font=\fontsize{12}{0}\selectfont] {$A_{1,\nodeidx}$}; 
		
	\end{tikzpicture}
	\caption{Star-shaped \gls{graph} $\graph^{(\rm star)}$ with a centre node $\nodeidx=1$ representing a server 
		that trains a (global) \gls{model} which is shared with peripheral nodes. These peripheral nodes 
		represent \emph{clients} generating \glspl{localdataset}. The training process at the server 
		is facilitated by receiving updates on the \glspl{modelparam} from the clients. }
	\label{fig:star_graph}
\end{figure}

Instead of using \gls{gtvmin} with a connected \gls{flnetwork} and a large value of $\regparam$, 
we can also enforce identical local copies $\estlocalparams{\nodeidx}$ via a constraint: 
\begin{align}
	\label{equ_def_constrained_GTV_MOCHA}
	\widehat{\weights} & \in \underset{\netparams \in \mathcal{S}}{\mathrm{arg \ min}}\sum_{\nodeidx \in \nodes}(1/\localsamplesize{\nodeidx}) \normgeneric{\labelvec^{(\nodeidx)} - \featuremtx^{(\nodeidx)} \localparams{\nodeidx}}{2}^{2} \nonumber \\ 
	& \mbox{ with } \mathcal{S}  = \big\{\netparams = {\rm stack}\{ \localparams{\nodeidx} \}_{\nodeidx=1}^{\nrnodes}: \localparams{\nodeidx} = \localparams{\nodeidx'} \mbox{ for any }  \nodeidx,\nodeidx' \in \nodes \big\}. 
\end{align}  
Here, we use as constraint set the subspace $\mathcal{S}$ defined in \eqref{equ_def_subspace_constant_local}. 
The projection of a given collection of \gls{localmodel} \glspl{parameter} $\weights = {\rm stack}\{ \localparams{\nodeidx} \}$ 
on $\mathcal{S}$ is given by  
\begin{equation} 
	\nonumber
	\projection{\mathcal{S}}{\weights} = \big( \vv^{T}, \ldots,\vv^{T} \big)^{T} \mbox{ with } \vv \defeq (1/\nrnodes) \sum_{\nodeidx \in \nodes} \localparams{\nodeidx}.
\end{equation} 

We can solve \eqref{equ_def_constrained_GTV_MOCHA} using \gls{projgd} from Chapter \ref{lec_gradientmethods}. 
The resulting projected \gls{gradstep} for solving \eqref{equ_def_constrained_GTV_MOCHA} is
\begin{align}
	\estlocalparamsiter{\nodeidx}{\iteridx+1/2}  & \!\defeq\! \underbrace{\localparamsiter{\nodeidx}{\iteridx}\!-\!\lrate_{\nodeidx,\iteridx} (2/\localsamplesize{\nodeidx}) \big( \featuremtx^{(\nodeidx)}\big)^T \big(\featuremtx^{(\nodeidx)} \localparamsiter{\nodeidx}{\iteridx} \!-\! \labelvec^{(\nodeidx)} \big)}_{\mbox{ (local \gls{gradstep})}}      \label{equ_local_update_FedAvgSGD}   \\ 
	\localparamsiter{\nodeidx}{\iteridx+1}  & \defeq  (1/\nrnodes) \sum_{\nodeidx' \in \nodes} \estlocalparamsiter{\nodeidx'}{\iteridx+1/2} \quad \mbox{ (projection)  }. \label{equ_averaging_update_FedAvgSGD}
\end{align} 
We can implement \eqref{equ_averaging_update_FedAvgSGD} conveniently in a server-client system with 
each node $\nodeidx$ being a client: 
\begin{itemize} 
	\item First, each node computes the update \eqref{equ_local_update_FedAvgSGD}, i.e., 
	a \gls{gradstep} towards a minimum of the local \gls{loss} $\locallossfunc{\nodeidx}{\localparams{\nodeidx}}\defeq \normgeneric{\labelvec^{(\nodeidx)} - \featuremtx^{(\nodeidx)} \localparams{\nodeidx}}{2}^{2}$. 
	\item Second, each node $\nodeidx$ sends the result $\estlocalparamsiter{\nodeidx}{\iteridx}$ of its local \gls{gradstep} to a 
	server. 
	\item Finally, after receiving the updates $\estlocalparamsiter{\nodeidx}{\iteridx}$ from all 
	nodes $\nodeidx \in \nodes$, the server computes the projection step \eqref{equ_averaging_update_FedAvgSGD}. 
	This projection results in the new \gls{localmodel} \glspl{parameter} $\localparamsiter{\nodeidx}{\iteridx+1}$ 
	that are sent back to each client $\nodeidx$. 
\end{itemize} 

The averaging step \eqref{equ_averaging_update_FedAvgSGD} might take much longer to execute 
than the local update step \eqref{equ_local_update_FedAvgSGD}. Indeed, \eqref{equ_averaging_update_FedAvgSGD} 
typically requires transmission of local \glspl{modelparam} from every client $\nodeidx \in \nodes$ to a server 
or central computing unit. Thus, after the client $\nodeidx \in \nodes$ has computed the local \gls{gradstep} \eqref{equ_local_update_FedAvgSGD}, it must wait until the server (i) has collected the updates $\estlocalparamsiter{\nodeidx}{\iteridx}$ from all 
clients and (ii) sent back their average $\localparamsiter{\nodeidx}{\iteridx+1}$ to $\nodeidx \in \nodes$. 

Instead of using a single \gls{gradstep} \eqref{equ_local_update_FedAvgSGD},\footnote{For 
	a large \gls{localdataset}, the local \gls{gradstep} \eqref{equ_local_update_FedAvgSGD} 
	can become computationally too expensive and must be replaced by an approximation, e.g., 
	using a stochastic gradient approximation \eqref{equ_def_loss_func_linreg_gradient_stoch}.} 
and then being forced to wait for receiving $\localparamsiter{\nodeidx}{\iteridx+1}$ back from the server, 
a client can make better use of its resources. For example, the \gls{device} $\nodeidx$ could execute 
several local \glspl{gradstep} \eqref{equ_local_update_FedAvgSGD} to make more progress 
towards the optimum, 
\begin{align}
	\vv^{(0)}      &\defeq  \estlocalparamsiter{\nodeidx}{\iteridx}  \nonumber \\
	\vv^{(\iteridxinner)}   &\defeq \vv^{(\iteridxinner\!-\!1)}\!-\!\lrate_{\nodeidx,\iteridx} (2/\localsamplesize{\nodeidx}) \big( \featuremtx^{(\nodeidx)}\big)^T \big(  \featuremtx^{(\nodeidx)} \vv^{(\iteridxinner\!-\!1)}\!-\!\labelvec^{(\nodeidx)} \big) \mbox{, for } \iteridxinner=1,\ldots,R \nonumber \\ 
	\estlocalparamsiter{\nodeidx}{\iteridx+1/2}  &\defeq \vv^{(R)}. \label{equ_local_update_FedAvgSGD_lin_several}
\end{align} 

We obtain Algorithm \ref{equ_def_fedavg_basic_linreg} by iterating the combination 
of \eqref{equ_local_update_FedAvgSGD_lin_several} with the projection step \eqref{equ_averaging_update_FedAvgSGD}. 
\begin{algorithm}[htbp]
	\caption{Server-based \gls{fl} for \glspl{linmodel}}
	\label{equ_def_fedavg_basic_linreg}
	{\bf The Server.} \\ 
	{\bf Input.} Some stopping criterion; list of clients $\nodeidx=1,\ldots,\nrnodes$, number $R$ of local updates. \\
	{\bf Output.} Trained \glspl{modelparam} $\widehat{\weights}^{(\rm global)}$ \\ 
	{\bf Initialize.} $\iteridx \defeq 0$; $\localparamsiter{\nodeidx}{\iteridx}=\mathbf{0}$ for all $\nodeidx=1,\ldots,\nrnodes$
	\begin{algorithmic}[1]
		\While{\gls{stopcrit} is not satisfied}
		\State Update the global \glspl{modelparam} $$\widehat{\weights}^{(\iteridx)}\defeq \left(1/ \nrnodes \right)\sum_{\nodeidx=1}^{\nrnodes} \localparamsiter{\nodeidx}{\iteridx}.$$
		\State \label{alg_fedavg_send_linreg} Send \glspl{modelparam} $\widehat{\weights}^{(\iteridx)}$ (and $\iteridx$) to all clients. $\nodeidx\!=\!1,\ldots,\nrnodes$
		\State \label{alg_fedavg_recv_linreg} Gather update local \glspl{modelparam} $\localparamsiter{\nodeidx}{\iteridx+1}$ from 
		clients $\nodeidx\!=\!1,\ldots,\nrnodes$. 
		\State {\bf Clock Tick.} $\iteridx \defeq \iteridx+1$. 
		\EndWhile
	\end{algorithmic}
	{\bf  The Client $\nodeidx\in \{1,\ldots,\nrnodes\}$.} \\ 
	{\bf Input.} \Gls{localdataset} $\featuremtx^{(\nodeidx)},\labelvec^{(\nodeidx)}$, number of 
	\glspl{gradstep} $R$ and \gls{learnrate} (schedule) $\lrate_{\nodeidx,\iteridx}$.   
	\begin{algorithmic}[1]
		\State Receive the current \glspl{modelparam} $\widehat{\weights}^{(\iteridx)}$ from the server. 
		\State \label{step_local_update_FedAvgSGD_lin_several_alg} Update the local \glspl{modelparam} by $R$ \glspl{gradstep}
		\begin{align}
			\vv^{(0)}      &\defeq  \widehat{\weights}^{(\rm global)} \nonumber \\
			\vv^{(\iteridxinner)}   &\defeq \vv^{(\iteridxinner\!-\!1)}\!-\!\lrate_{\nodeidx,\iteridx} (2/\localsamplesize{\nodeidx}) \big( \featuremtx^{(\nodeidx)}\big)^T \big(\featuremtx^{(\nodeidx)} \vv^{(\iteridxinner\!-\!1)}\!-\! \labelvec^{(\nodeidx)} \big) \mbox{, for } \iteridxinner=1,\ldots,R \nonumber \\ 
			\localparamsiter{\nodeidx}{\iteridx+1}&\defeq \vv^{(R)}.  \nonumber
		\end{align} 
		\State Send the new local \glspl{modelparam} $\localparamsiter{\nodeidx}{\iteridx+1}$ back to server.
	\end{algorithmic}
\end{algorithm}

One of the most popular server-based \gls{fl} \glspl{algorithm}, referred to as \gls{fedavg} and summarized 
in Algorithm \ref{alg_fed_avg}, is obtained by two modifications of Algorithm \ref{equ_def_fedavg_basic_linreg}: 
\begin{itemize} 
	\item replacing the updates in step \ref{step_local_update_FedAvgSGD_lin_several_alg} at 
	the client in Algorithm \ref{equ_def_fedavg_basic_linreg} with 
	$\vv^{(\iteridxinner)} \defeq \vv^{(\iteridxinner-1)}\!-\!\lrate_{\nodeidx,\iteridx} \mathbf{g}\big(\vv^{(\iteridxinner)}\big)$ using the \gls{gradient} 
	approximation $\mathbf{g}^{(\nodeidx)}\big(\vv^{(r)}\big)\approx \nabla \locallossfunc{\nodeidx}{ \vv^{(r)} }$, 
	\item using a randomly selected subset $\cluster^{(\iteridx)}$ of clients during each global iteration $\iteridx$. 
\end{itemize} 
\begin{algorithm}[htbp]
	\caption{\gls{fedavg} \cite{pmlr-v54-mcmahan17a}}
	\label{alg_fed_avg}
	{\bf The Server.}  \\
	{\bf Input.} List of clients $\nodeidx=1,\ldots,\nrnodes$, number $R$ of local updates \\  
	{\bf Output.} Trained \glspl{modelparam} $\widehat{\weights}^{(\rm global)}$\\ 
	{\bf Initialize.} $\iteridx \defeq 0$; $\widehat{\weights}^{(\rm global)} \defeq\mathbf{0}$ for all $\nodeidx=1,\ldots,\nrnodes$
	\begin{algorithmic}[1]
		\While{\gls{stopcrit} is not satisfied}
		\State randomly select a subset $\cluster^{(\iteridx)}$ of clients
		\State \label{alg_fedavg_send} send $\widehat{\weights}^{(\rm global)}$ to all clients $\nodeidx\!\in\!\cluster^{(\iteridx)}$ 
		\State \label{alg_fedavg_recv} receive updated \glspl{modelparam} $\localparams{\nodeidx}$ from clients $\nodeidx\!\in\!\cluster^{(\iteridx)}$ 
		\State update global \glspl{modelparam} $$\widehat{\weights}^{(\rm global)} \defeq \left(1/\big|\cluster^{(\iteridx)}\big|\right) \sum_{\nodeidx\in\cluster^{(\iteridx)}} \localparams{\nodeidx}.$$
		\State increase iteration counter $\iteridx\!\defeq\!\iteridx\!+\!1$
		\EndWhile
	\end{algorithmic}
	{\bf Client $\nodeidx\in \{1,\ldots,\nrnodes\}$}, with local \gls{lossfunc} $\locallossfunc{\nodeidx}{\cdot}$ 
	\begin{algorithmic}[1]
		\State receive global \glspl{modelparam} $\widehat{\weights}^{(\rm global)}$ from server 
		\State update local \glspl{modelparam} by $R$ approximate \glspl{gradstep}
		\begin{align}
			\vv^{(0)}      &\defeq \widehat{\weights}^{(\rm global)}   \nonumber \\
			\vv^{(\iteridxinner)}   &\defeq \vv^{(\iteridxinner\!-\!1)}\!-\!\lrate_{\nodeidx,\iteridx} \underbrace{\mathbf{g}^{(\nodeidx)}\big(\vv^{(\iteridxinner\!-\!1)}\big)}_{\approx \nabla \locallossfunc{\nodeidx}{\vv^{(\iteridxinner\!-\!1)} }} \mbox{, for } \iteridxinner=1,\ldots,R \nonumber \\ 
			\localparams{\nodeidx} &\defeq \vv^{(R)}. \label{equ_local_update_FedAvgSGD_several_alg}
		\end{align} 
		\State return $\localparams{\nodeidx}$ back to server
	\end{algorithmic}
\end{algorithm}

\clearpage
\subsection{FedProx}
\label{sec_fed_prox} 

A central challenge in \gls{fedavg} (Algorithm \ref{alg_fed_avg}) is selecting an 
appropriate number of local updates, $R$, in \eqref{equ_local_update_FedAvgSGD_several_alg}. 
In each iteration, all clients perform exactly $R$ approximate \glspl{gradstep}. 
However, \cite{FedProx2020} argues that enforcing a uniform number $R$ across 
clients can degrade performance in certain \gls{fl} settings. To mitigate this, they propose an 
alternative to \eqref{equ_local_update_FedAvgSGD_several_alg} for the local update step.
This alternative is given by 
\begin{equation} 
	\label{equ_def_replace_sgd_local_min}
	\localparams{\nodeidx} \defeq \argmin_{\vv \in \mathbb{R}^{\dimlocalmodel}} \bigg[\locallossfunc{\nodeidx}{\vv} + (1/\lrate) \normgeneric{\vv - \widehat{\weights}^{(\rm global)}  }{2}^2\bigg].
\end{equation} 

We have already encountered an update of the form \eqref{equ_def_replace_sgd_local_min} in 
Section \ref{sec_gen_gradient_step}. Indeed, \eqref{equ_def_replace_sgd_local_min} is the 
application of the \gls{proxop} of $\locallossfunc{\nodeidx}{\vv}$ (see \eqref{equ_def_prox_operator}) 
to the current \glspl{modelparam}. We obtain Algorithm \ref{alg_fed_prox} from Algorithm \ref{alg_fed_avg} 
by replacing the local update step \eqref{equ_local_update_FedAvgSGD_several_alg} with \eqref{equ_def_replace_sgd_local_min}. 
Empirical studies have shown that Algorithm \ref{alg_fed_prox} outperforms \gls{fedavg} (Algorithm \ref{alg_fed_avg}) 
for \gls{fl} applications with a high-level of heterogeneity among the computational capabilities 
of \glspl{device} $\nodeidx=1,\ldots,\nrnodes$ and the statistical properties of their \glspl{localdataset} $\localdataset{\nodeidx}$ \cite{FedProx2020}. 
\begin{algorithm}[htbp]
	\caption{FedProx \cite{FedProx2020}}
	\label{alg_fed_prox}
	{\bf The Server.}  \\
	{\bf Input.} List of clients $\nodeidx=1,\ldots,\nrnodes$  \\  
	{\bf Output.} Trained \glspl{modelparam} $\widehat{\weights}^{(\rm global)}$\\ 
	{\bf Initialize.} $\iteridx \defeq 0$; $\widehat{\weights}^{(\rm global)} \defeq\mathbf{0}$ for all $\nodeidx=1,\ldots,\nrnodes$
	\begin{algorithmic}[1]
		\While{\gls{stopcrit} is not satisfied}
		\State randomly select a subset $\cluster^{(\iteridx)}$ of clients
		\State \label{alg_fedprox_send} send $\widehat{\weights}^{(\rm global)}$ to all clients $\nodeidx\!\in\!\cluster^{(\iteridx)}$ 
		\State \label{alg_fedprox_recv} receive updated \glspl{modelparam} $\localparams{\nodeidx}$ from clients $\nodeidx\!\in\!\cluster^{(\iteridx)}$ 
		\State update global \glspl{modelparam} $$\widehat{\weights}^{(\rm global)} \defeq \left(1/\big|\cluster^{(\iteridx)}\big|\right) \sum_{\nodeidx\in\cluster^{(\iteridx)}} \localparams{\nodeidx}.$$
		\State increase iteration counter $\iteridx\!\defeq\!\iteridx\!+\!1$
		\EndWhile
	\end{algorithmic}
	{\bf Client $\nodeidx\in \{1,\ldots,\nrnodes\}$}, with local \gls{lossfunc} $\locallossfunc{\nodeidx}{\cdot}$ 
	\begin{algorithmic}[1]
		\State receive global \glspl{modelparam} $\widehat{\weights}^{(\rm global)}$ from server 
		\State update local \glspl{modelparam} by 
		\begin{align}
			\localparams{\nodeidx} \defeq \argmin_{\vv \in \mathbb{R}^{\dimlocalmodel}} \bigg[\locallossfunc{\nodeidx}{\vv} + (1/\lrate) \normgeneric{\vv - \widehat{\weights}^{(\rm global)}  }{2}^2\bigg]
			\label{equ_local_update_prox_alg}
		\end{align} 
		\State return $\localparams{\nodeidx}$ back to server
	\end{algorithmic}
\end{algorithm}



As the notation in \eqref{equ_def_replace_sgd_local_min} indicates, the parameter $\lrate$ plays a role 
similar to the \gls{learnrate} of a \gls{gradstep} \eqref{equ_def_basic_gradstep}. It controls the size of the 
neighbourhood of $\localparamsiter{\nodeidx}{\iteridx}$ over which \eqref{equ_def_replace_sgd_local_min} 
optimizes the local \gls{lossfunc} $ \locallossfunc{\nodeidx}{\cdot}$. Choosing a small $\lrate$ forces the 
update \eqref{equ_def_replace_sgd_local_min} to not move too far from the current \glspl{modelparam} 
$\localparamsiter{\nodeidx}{\iteridx}$. 

The core computation \eqref{equ_local_update_prox_alg} of FedProx Algorithm \ref{alg_fed_prox} 
can be interpreted as form of \gls{regularization}. 
Indeed, we obtain \eqref{equ_local_update_prox_alg} from \eqref{equ_def_ridgeregression} by 
\begin{itemize} 
	\item replacing 
	the average \gls{sqerrloss} with the local \gls{lossfunc} $\locallossfunc{\nodeidx}{\vv}$,  
	\item using the \gls{regularizer}
	\begin{equation} 
		\label{equ_def_penlaty_fedprox} 
		\regularizer{\vv} \defeq  \normgeneric{\vv - \widehat{\weights}^{(\rm global)}  }{2}^2, 
	\end{equation} 
	\item and the \gls{regularization} parameter $\regparam \defeq 1/\lrate$.  
\end{itemize} 

Note that Algorithms \ref{alg_fed_prox} and \ref{alg_fed_avg} provide only an abstract 
description of a practical \gls{fl} system. The details of their actual implementation, 
such as the synchronization between the server and all clients (see steps \ref{alg_fedprox_recv} 
and \ref{alg_fedprox_send} in Algorithm \ref{alg_fed_prox}) is beyond the scope of this book. 
Instead, we refer the reader to relevant literature on the implementation of 
distributed computing systems \cite{DistributedSystems,TanenbaumComputerNet}.  

\subsection{FedRelax}
\label{sec_fedrelax}

We now apply a simple block-coordinate minimization method \cite{ParallelDistrBook} to solve \gls{gtvmin} \eqref{equ_def_gtvmin}. 
To this end, we rewrite \eqref{equ_def_gtvmin} as
\begin{align}
	\label{equ_GTVMin_MOCHA_coordmin}
	\widehat{\netparams} & \in \underset{\netparams \in \mathbb{R}^{\dimlocalmodel \nrnodes}}{\mathrm{arg \ min}} \underbrace{\sum_{\nodeidx \in \nodes} 
		\localobj{\nodeidx}{\netparams}}_{=: f^{({\rm GTV})} (\netparams)}  \nonumber \\ 
	& \mbox{ with }  \localobj{\nodeidx}{\netparams} \defeq \locallossfunc{\nodeidx}{\localparams{\nodeidx}} + (\regparam/2) \sum_{\nodeidx' \in \neighbourhood{\nodeidx}} \edgeweight_{\nodeidx,\nodeidx'} \normgeneric{\localparams{\nodeidx}- \localparams{\nodeidx'}}{2}^{2} \mbox{, } \nonumber\\ 
	& \mbox{ and the stacked \glspl{modelparam} } \netparams = \big(\localparams{1},\ldots,\localparams{\nrnodes} \big)^{T}. 
\end{align} 
According to \eqref{equ_GTVMin_MOCHA_coordmin}, the \gls{objfunc} of \eqref{equ_def_gtvmin} 
decomposes into components $\localobj{\nodeidx}{\netparams}$, one for each node $\nodes$ of the \gls{flnetwork}. Moreover, 
the local \glspl{modelparam} $\localparams{\nodeidx}$ influence the \gls{objfunc} only via the 
components at the nodes $\nodeidx \cup \neighbourhood{\nodeidx}$. We exploit this structure 
of \eqref{equ_GTVMin_MOCHA_coordmin} to decouple the optimization of the local \glspl{modelparam} 
$\big\{ \estlocalparams{\nodeidx} \big\}_{\nodeidx \in \nodes}$ as described next. 

Consider some local \glspl{modelparam} $\localparamsiter{\nodeidx}{\iteridx}$, for $\nodeidx=1,\ldots,\nrnodes$, 
at time $\iteridx$. We then update (in parallel) each $\localparamsiter{\nodeidx}{\iteridx}$ 
by minimizing $f^{({\rm GTV})} (\cdot)$ along $\localparams{\nodeidx}$ with the other 
local \glspl{modelparam} $\localparams{\nodeidx'} \defeq \localparamsiter{\nodeidx'}{\iteridx}$ 
held fixed for all $\nodeidx' \neq \nodeidx$,  
\begin{align} 
	\label{equ_def_coord_min_update}
	\localparamsiter{\nodeidx}{\iteridx+1} & \in \argmin_{\localparams{\nodeidx} \in \mathbb{R}^{\dimlocalmodel}} f^{({\rm GTV})} \bigg( \localparamsiter{1}{\iteridx},\ldots,\localparamsiter{\nodeidx-1}{\iteridx},\localparams{\nodeidx},\localparamsiter{\nodeidx+1}{\iteridx},\ldots \bigg)  \nonumber \\ 
	& \stackrel{\eqref{equ_GTVMin_MOCHA_coordmin}}{=}  \argmin_{\localparams{\nodeidx} \in \mathbb{R}^{\dimlocalmodel}} \localobj{\nodeidx}{ \localparamsiter{1}{\iteridx},\ldots,\localparamsiter{\nodeidx-1}{\iteridx},\localparams{\nodeidx},\localparamsiter{\nodeidx+1}{\iteridx},\ldots}  \nonumber \\ 
	& \stackrel{\eqref{equ_GTVMin_MOCHA_coordmin}}{=}  \argmin_{\localparams{\nodeidx} \in \mathbb{R}^{\dimlocalmodel}} \locallossfunc{\nodeidx}{\localparams{\nodeidx}} + \regparam \sum_{\nodeidx' \in \neighbourhood{\nodeidx}} \edgeweight_{\nodeidx,\nodeidx'} \normgeneric{\localparams{\nodeidx}- \localparamsiter{\nodeidx'}{\iteridx} }{2}^{2}.
\end{align} 
The update rule in \eqref{equ_def_coord_min_update} can be viewed as a non-linear \gls{jacobimethod} 
applied to \eqref{equ_GTVMin_MOCHA_coordmin} \cite[Sec. 3.2.4]{ParallelDistrBook}. It also admits 
an interpretation as a form of block-coordinate optimization \cite{Tseng:2001aa}. By iterating this 
update sufficiently many times, we arrive at Algorithm \ref{alg_fed_relax}.
\begin{algorithm}[htbp]
	\caption{FedRelax for Parametric Models}
	\label{alg_fed_relax}
	{\bf Input}: \gls{flnetwork} $\graph$ with local \glspl{lossfunc} $\locallossfunc{\nodeidx}{\cdot}$, 
	\gls{gtv} parameter $\regparam$ \\
	{\bf Initialize}: $\iteridx\!\defeq\!0$; $\localparamsiter{\nodeidx}{0}\!\defeq\!{\bf 0}$
	\begin{algorithmic}[1]
		\While{stopping criterion is not satisfied}
		\For{all nodes $ \nodeidx \in \nodes$ in parallel}
		\State compute $\localparamsiter{\nodeidx}{\iteridx+1}$ via \eqref{equ_def_coord_min_update} \label{equ_def_update_step_fedrelax}
		\State share $\localparamsiter{\nodeidx}{\iteridx+1}$ with \glspl{neighbor} $\neighbourhood{\nodeidx}$
		\EndFor
		\State $\iteridx\!\defeq\!\iteridx\!+\!1$
		\EndWhile
	\end{algorithmic}
\end{algorithm}
There is an interesting connection between the update \eqref{equ_def_coord_min_update} and 
the basic \glspl{gradstep} used by FedGD and FedSGD (see Algorithm \ref{alg_fed_gd_general} and \ref{alg_fed_sgd_general}).
Indeed, we obtain step \ref{equ_gdstep_fedgd_general} in Algorithm \ref{alg_fed_gd_general} 
from \eqref{equ_def_coord_min_update} by replacing the \gls{lossfunc} $\locallossfunc{\nodeidx}{\localparams{\nodeidx}}$ 
with the approximation 
$$\locallossfunc{\nodeidx}{\localparamsiter{\nodeidx}{\iteridx}}+ \big( \nabla \locallossfunc{\nodeidx}{\localparamsiter{\nodeidx}{\iteridx}}  \big) \big(\localparams{\nodeidx} -\localparamsiter{\nodeidx}{\iteridx} \big)+ (1/(2\lrate))\normgeneric{\localparams{\nodeidx} -\localparamsiter{\nodeidx}{\iteridx}}{2}^2.$$

{\bf A Model-Agnostic Method.} The applicability of Algorithm \ref{alg_fed_relax} is limited to \glspl{flnetwork} 
with parametric \glspl{localmodel} (such as \gls{linreg} or \glspl{ann} with a common structure). 
We can generalize Algorithm \ref{alg_fed_relax} to non-parametric \glspl{localmodel} by applying 
the non-linear \gls{jacobimethod} to the \gls{gtvmin} variant \eqref{equ_def_gtvmin_nonparam}. 
This results in the update
\begin{align} 
	\label{equ_def_coord_min_update_nonparam}
	\estlocalhypositer{\nodeidx}{\iteridx+1} & \in  \argmin_{\localhypothesis{\nodeidx} \in \localmodel{\nodeidx}} \locallossfunc{\nodeidx}{\localhypothesis{\nodeidx}} + \regparam \sum_{\nodeidx' \in \neighbourhood{\nodeidx}} \edgeweight_{\nodeidx,\nodeidx'} \underbrace{\discrepancy{\hypothesis^{(\nodeidx)}}{\estlocalhypositer{\nodeidx'}{\iteridx}}}_{\mbox{see } \eqref{equ_def_variation_non_parametric}}.
\end{align}
We obtain Algorithm \ref{alg_fed_relax_agnostic} as a model-agnostic variant of 
Algorithm \ref{alg_fed_relax} by replacing the update \eqref{equ_def_coord_min_update} 
in its step \ref{equ_def_update_step_fedrelax} with the update \eqref{equ_def_coord_min_update_nonparam}. 

Algorithm \ref{alg_fed_relax_agnostic} is model-agnostic as it allows \glspl{device} of an \gls{flnetwork} to 
train different types of \glspl{localmodel}. The only restriction for the \glspl{localmodel} 
is that the update \eqref{equ_def_coord_min_update_nonparam} can be computed efficiently. 
For some choices of \glspl{localmodel} and \gls{lossfunc}, the update \eqref{equ_def_coord_min_update_nonparam} 
can be implemented by basic \gls{dataaug} (see Exercise \ref{ex_data_aug_fl_relax}). 
\begin{algorithm}[htbp]
	\caption{Model Agnostic FedRelax}
	\label{alg_fed_relax_agnostic}
	{\bf Input}: \gls{flnetwork} with $\graph$,  \glspl{localmodel} $\localmodel{\nodeidx}$, \glspl{lossfunc} $\locallossfunc{\nodeidx}{\cdot}$,
	\gls{gtv} parameter $\regparam$, \gls{loss} $\lossfunc{\cdot}{\cdot}$ used in \eqref{equ_def_variation_non_parametric}. \\
	{\bf Initialize}: $\iteridx\!\defeq\!0$; $\estlocalhypositer{\nodeidx}{0}\!\defeq\!{\bf 0}$
	\begin{algorithmic}[1]
		\While{stopping criterion is not satisfied}
		\For{all nodes $ \nodeidx \in \nodes$ in parallel}
		\State compute $\estlocalhypositer{\nodeidx}{\iteridx+1} $ via \eqref{equ_def_coord_min_update_nonparam} \label{step_update_gtvmin_agnostic}
		\EndFor
		\State $\iteridx\!\defeq\!\iteridx\!+\!1$
		\EndWhile
	\end{algorithmic}
\end{algorithm}

\clearpage
\subsection{A Unified Formulation} 
\label{sec_unified_form_algos} 

The previous sections have presented some widely-used \gls{fl} \glspl{algorithm}. 
These \glspl{algorithm} are obtained by applying distributed \glspl{optmethod} to solve \gls{gtvmin}. 
Despite their different formulations they share a common underlying structure. 
In particular, they can all be expressed as synchronous \glspl{fixedpointiter}:
\begin{equation}
	\label{equ_update_fixed_point_synch} 
	\estlocalhypositer{\nodeidx}{\iteridx+1} = \fixedpointop^{(\nodeidx)} \big(\estlocalhypositer{1}{\iteridx}, \ldots, \estlocalhypositer{\nrnodes}{\iteridx}\big) \mbox{, for } \nodeidx=1,\ldots,\nrnodes. 
\end{equation}

\begin{figure}[H] 
	\begin{center}
\begin{tikzpicture}[scale=4]
	\draw[->] (-0.1,0) -- (1.2,0);
	\draw[->] (0,-0.1) -- (0,1.2);
	\node [yshift=-10pt] at (1.25,0) {$\hypothesis^{(1)}, \ldots, \hypothesis^{(\nrnodes)}$};
	\node at (0,1.25) {$\hypothesis^{(\nodeidx)}$}; 
	\draw[thick, smooth, domain=0:1.05, samples=100] 
	plot(\x, {0.5 + 0.4*sin(3*\x r) * exp(-2*\x)});
	  \node at (1,0.4) {$\fixedpointop^{(\nodeidx)} \big(\hypothesis^{(1)}, \ldots, \hypothesis^{(\nrnodes)} \big)$};
\end{tikzpicture}
\end{center} 
\caption{A key computational step in many \gls{fl} \glspl{algorithm} is the evaluation of an operator 
	$\fixedpointop^{(\nodeidx)}$ at each node $\nodeidx = 1, \ldots, \nrnodes$ of the \gls{flnetwork}.
	 \label{fig_single_update_operator}} 
\end{figure}

Each operator \(\fixedpointop^{(\nodeidx)}: \localmodel{1}\times \ldots \times \localmodel{\nrnodes}\rightarrow \localmodel{\nodeidx}\) 
represents a local update rule at the $\nodeidx=1,\ldots,\nrnodes$ (see Figure \ref{fig_single_update_operator}). 
Some \glspl{algorithm} use time-varying update rules, 
\begin{equation}
	\label{equ_update_fixed_point_synch_time_varying} 
	\estlocalhypositer{\nodeidx}{\iteridx+1} = \fixedpointop^{(\nodeidx)} \big(\estlocalhypositer{1}{\iteridx}, \ldots, \estlocalhypositer{\nrnodes}{\iteridx}\big). 
\end{equation}
with operators \(\fixedpointop^{(\nodeidx,\iteridx)}\) that can vary across nodes $\nodeidx=1,\ldots,\nrnodes$
and time instants \(\iteridx=1,2,\ldots\). One example of \eqref{equ_update_fixed_point_synch_time_varying} 
is used in Algorithm \ref{alg_fed_gd_general} for a time-varying \gls{learnrate}.  

Clearly, any \gls{fl} \gls{algorithm} of the form \ref{equ_update_fixed_point_synch} is fully 
specified by the operators \(\fixedpointop^{(1)}, \dots, \fixedpointop^{(\nrnodes)}\). 
This - rather trivial - observation implies that we can study the behaviour of \gls{fl} \glspl{algorithm} 
via analyzing the properties of the operators $\fixedpointop^{(\nodeidx)}$, for $\nodeidx=1,\ldots,\nrnodes$. 
In particular, the robustness of \gls{fl} \glspl{algorithm} crucially depends on the 
shape of $\fixedpointop^{(\nodeidx)}$.

For parametric \glspl{localmodel}, we can re-formulate the \gls{fixedpointiter} \eqref{equ_update_fixed_point_synch} 
directly in terms of the \glspl{modelparam} 
\begin{equation}
	\label{equ_update_fixed_point_synch_param} 
	\localparamsiter{\nodeidx}{\iteridx+1} = \fixedpointop^{(\nodeidx)} \big(\localparamsiter{1}{\iteridx}, \ldots, \localparamsiter{\nrnodes}{\iteridx}\big)\mbox{, for } \iteridx=0,1,\ldots,
\end{equation}
with operators \(\fixedpointop^{(\nodeidx)}: \mathbb{R}^{\nrnodes \dimlocalmodel} \rightarrow \mathbb{R}^{\dimlocalmodel}\), 
for $\nodeidx=1,\ldots,\nrnodes$. One example of \eqref{equ_update_fixed_point_synch_param} is 
the update \ref{equ_def_coord_min_update} used by FedRelax (see Algorithm \ref{alg_fed_relax}).

\clearpage
\subsection{Asynchronous FL Algorithms} 
\label{sec_asynch_fl_alg} 

The \gls{fl} \glspl{algorithm} presented so far rely on synchronous coordination among 
\glspl{device} $\nodeidx = 1, \ldots, \nrnodes$ within an \gls{flnetwork} \cite[Ch. 6]{DistributedSystems}. 
A new iteration is only initiated once all \glspl{device} have completed their local updates 
\eqref{equ_update_fixed_point_synch} and communicated them to their \glspl{neighbor} 
\cite[Sec. 10]{DistrOptStatistLearningADMM}, \cite[Sec. 1.4]{ParallelDistrBook}.

The implementation of synchronous \gls{fl} \glspl{algorithm} can be difficult (or impossible) in practice. 
As highlighted in Chapter \ref{lec_trustworthyfl}, trustworthy \gls{fl} systems should tolerate unreliable 
or failing \glspl{device}. Synchronous methods lack this robustness—any \gls{device} failure or 
dropout can cause the entire \gls{algorithm} execution to stall. Moreover, synchronous execution 
is inefficient in heterogeneous \gls{fl} systems. \Glspl{device} often vary in computational power 
or communication bandwidth, leading to the \emph{straggler problem}: faster \glspl{device} are 
forced to wait idly for slower ones \cite{Liu2020,EfficentScheduling2021}. Having \glspl{device} to 
wait idly for slower \glspl{device} results in a waste of their computational resources. 

To address the limitations of synchronous \gls{fl} \glspl{algorithm}, we now show how to 
build asynchronous variants of the \gls{fl} \glspl{algorithm} discussed in Section \ref{sec_unified_form_algos}. 
We focus here on parametric \glspl{localmodel}, each represented by their own \glspl{modelparam} $\localparams{\nodeidx}$. The basic 
idea is to let each \gls{device} $\nodeidx=1,\ldots,\nrnodes$ execute the update 
\eqref{equ_update_fixed_point_synch_param} independently, using potentially 
out-dated updates from its \glspl{neighbor} $\neighbourhood{\nodeidx}$. 

An asynchronous \gls{fl} \gls{algorithm} consists of a sequence of update events, 
which we index by $\iteridx = 0,1, 2, \ldots$ (see Figure \ref{fig_executation_async_algo}). 
During each event $\iteridx$, a subset $\asyncactiveset{\iteridx} \subseteq \nodes$ 
of \glspl{device} performs updates:
\begin{equation} 
	\label{equ_update_generic_asynch} 
	\localparamsiter{\nodeidx}{\iteridx+1} = \fixedpointop^{(\nodeidx)} \big(\localparamsiter{1}{\iteridx_{\nodeidx,1}},\ldots,\localparamsiter{\nrnodes}{\iteridx_{\nodeidx,\nrnodes}}\big). 
\end{equation}
Here, $\iteridx_{\nodeidx,\nodeidx'} \leq \iteridx$ is event index of the latest 
available \glspl{modelparam} of \gls{device} $\nodeidx'$ at \gls{device} $\nodeidx$.

\begin{figure}[h]
	\begin{tikzpicture}[scale=1.2, node distance=2cm, every node/.style={font=\small}]
		
		\node[label={[xshift=-10pt]west:$\nodeidx=1$}] (device1) at (0, 2) {};
		\node[label={[xshift=-10pt, yshift=5pt]west:$\nodeidx=2$}] (device2) at (0, 0) {};

		\node[above=0.5cm of device1, align=center] (t1) {$\iteridx=1$};
		\node[right=2cm of t1, align=center] (t2) {$\iteridx=2$};
		\node[right=2cm of t2, align=center] (t3) {$\iteridx=3$};
		\node[right=2cm of t3, align=center] (t4) {$\iteridx=4$};
		
		\node[fill=black, circle, inner sep=2pt, below=0.5cm of t1, label={[xshift=2pt]right:$\eqref{equ_update_generic_asynch}$}] (d1t1) {};
		\node[fill=black, circle, inner sep=2pt, below=2.5cm of t1,label={[xshift=2pt]right:$\eqref{equ_update_generic_asynch}$}] (d2t1) {};
		
		\node[draw, circle, inner sep=2pt, below=0.5cm of t2, label={[xshift=2pt]right:$\eqref{equ_update_generic_asynch}$}] (d1t2) {}; 
		\node[fill=black, circle, inner sep=2pt, below=2.5cm of t2, label={[xshift=2pt]right:$\eqref{equ_update_generic_asynch}$}] (d2t2) {};
		
		\node[fill=black, circle, inner sep=2pt, below=0.5cm of t3, label={[xshift=2pt]right:$\eqref{equ_update_generic_asynch}$}] (d1t3) {};
		\node[draw, circle, inner sep=2pt, below=2.5cm of t3] (d2t3) {}; 
		
		\node[fill=black, circle, inner sep=2pt, below=0.5cm of t4, label={[xshift=2pt]right:$\eqref{equ_update_generic_asynch}$}] (d1t4) {};
		\node[fill=black, circle, inner sep=2pt, below=2.5cm of t4, label={[xshift=2pt]right:$\eqref{equ_update_generic_asynch}$}] (d2t4) {};
		
		\draw[->, thick] (d1t1.south east)  to[out=-45, in=135] (d2t3.north west);
		
		\draw[->, thick] (d2t2.north east) to[bend left=20] (d1t3.west);
		
		\draw[->, thick] (d1t3.south east) to[out=-45, in=135]  (d2t4.north west);

	\end{tikzpicture}
	\caption{The execution of an asynchronous \gls{fl} \gls{algorithm} consists of a sequence of 
		update events, indexed by $\iteridx=0,1,2,\ldots$. During each 
		event $\iteridx$, the active nodes $\nodeidx \in \asyncactiveset{\iteridx} \subseteq \nodes$ of an \gls{flnetwork} 
		update their local \glspl{modelparam} $\localparams{\nodeidx}$ by computing \eqref{equ_update_generic_asynch}. 
		Active nodes are depicted as filled circles. \label{fig_executation_async_algo}}
\end{figure} 

The set of nodes performing the update \eqref{equ_update_generic_asynch} during event $\iteridx$ 
is denoted as the active set $\asyncactiveset{\iteridx} \subseteq \nodes$. It is convenient to 
summarize the resulting asynchronous \gls{algorithm} as 
\begin{equation} 
	\localparamsiter{\nodeidx}{\iteridx+1} = \begin{cases}   \fixedpointop^{(\nodeidx)} \big(\localparamsiter{1}{\iteridx_{\nodeidx,1}},\ldots,\localparamsiter{\nrnodes}{\iteridx_{\nodeidx,\nrnodes}}\big) & \mbox{ for } \iteridx \in \updatetimes{\nodeidx} \\ 
		\localparamsiter{\nodeidx}{\iteridx}  & \mbox{ otherwise.} \end{cases}  \label{equ_full_update_async_generic}
\end{equation}
Here, we used the set 
\begin{equation} 
	\nonumber
	\updatetimes{\nodeidx} \defeq \big\{ \iteridx \in \{0,1,\ldots,\}: \nodeidx \in \asyncactiveset{\iteridx} \big\}, 
\end{equation} 
which consists, for each $\nodeidx=1,\ldots,\nrnodes$, of those clock ticks during 
which node $\nodeidx$ is active. Note that \eqref{equ_full_update_async_generic} reduces 
to the synchronous \gls{algorithm} \eqref{equ_update_fixed_point_synch_param} for the 
extreme case when $\updatetimes{\nodeidx}=0,1,2,\ldots,$ for all $\nodeidx=1,\ldots,\nrnodes$. 

Like the synchronous \gls{algorithm} \eqref{equ_update_fixed_point_synch_param}, 
also the asynchronous variant \ref{equ_full_update_async_generic} uses an 
iteration counter $\iteridx$. However, the practical meaning of $\iteridx$ in the 
asynchronous variant is fundamentally different: Instead of representing a global 
clock tick (or wall-clock time), the counter $\iteridx$ in \eqref{equ_full_update_async_generic} 
indexes some update event during which at least one node is active and computes a local update. 
We denote the set of active nodes (or \glspl{device}) during event $\iteridx$ by $\asyncactiveset{\iteridx} \subseteq \nodes$. 
The inactive nodes $\nodeidx \notin\asyncactiveset{\iteridx}$ leave their current \glspl{modelparam} 
unchanged, i.e., $\localparamsiter{\nodeidx}{\iteridx+1} = \localparamsiter{\nodeidx}{\iteridx}$. 

For each active node $\nodeidx \in \asyncactiveset{\iteridx}$, the local update \eqref{equ_full_update_async_generic} 
uses potentially outdated \glspl{modelparam} $\localparamsiter{\nodeidx'}{\iteridx_{\nodeidx,\nodeidx'}}$ 
from its \glspl{neighbor} $\nodeidx'  \in \neighbourhood{\nodeidx}$. Indeed, some of the \glspl{neighbor} 
might have not been in the active sets $\asyncactiveset{\iteridx-1},\asyncactiveset{\iteridx-2},\ldots$ 
of the most recent iterations. In this case, the update \eqref{equ_full_update_async_generic} does 
not have access to $\localparamsiter{\nodeidx'}{\iteridx}$. Instead, we can only use 
$\localparamsiter{\nodeidx'}{\iteridx_{\nodeidx,\nodeidx'}}$ that has been produced 
obtained during some previous iteration $\iteridx_{\nodeidx,\nodeidx'}< \iteridx$.

The update \eqref{equ_update_generic_asynch} involves an operator $\fixedpointop^{(\nodeidx)}: \mathbb{R}^{\dimlocalmodel \nrnodes} \rightarrow \mathbb{R}^{\dimlocalmodel}$ that determines the resulting \gls{fl} \gls{algorithm}. 
We can interpret \eqref{equ_update_generic_asynch} as an asynchronous variant of the 
synchronous \gls{algorithm} \eqref{equ_update_fixed_point_synch_param} obtained 
for the same $\fixedpointop^{(\nodeidx)}$. For example, an asynchronous variant of 
Algorithm \ref{alg_fed_gd_general} (with a fixed \gls{learnrate}) can be obtained for the choice   
\begin{equation}
	\label{equ_fixed_point_asynch_generic_gd}
	\fixedpointop^{(\nodeidx)} \big(\localparams{1},\ldots,\localparams{\nrnodes}\big) = \localparams{\nodeidx} - \lrate \bigg( \nabla \locallossfunc{\nodeidx}{\localparams{\nodeidx}}\!+\!\sum_{\nodeidx' \in \neighbourhood{\nodeidx}}\hspace*{-2mm}2\edgeweight_{\nodeidx,\nodeidx'} \big( \localparams{\nodeidx} - \localparams{\nodeidx'} \big) \bigg). 
\end{equation}
Note that the choice \eqref{equ_fixed_point_asynch_generic_gd} involves the local \glspl{lossfunc} 
and the weighted edges of an \gls{flnetwork}. 

The update \eqref{equ_update_generic_asynch}, at an active node $\nodeidx \in \asyncactiveset{\iteridx}$, 
involves potentially out-dated local \glspl{modelparam} $\localparamsiter{\nodeidx'}{\iteridx_{\nodeidx,\nodeidx'}}$, 
with $\iteridx_{\nodeidx,\nodeidx'} \leq \iteridx$, for $\nodeidx' =1,\ldots,\nrnodes$. 
The quantity $\iteridx_{\nodeidx,\nodeidx'}$ represents the most recent update event  
during which node $\nodeidx'$ has shared its updated local \glspl{modelparam} with 
node $\nodeidx$. We can, in turn, interpret the difference $\iteridx - \iteridx_{\nodeidx,\nodeidx'}$ 
as a measure of the communication delay between node $\nodeidx'$ and node $\nodeidx$. 

Depending on the extent of the delays $\iteridx - \iteridx_{\nodeidx,\nodeidx'}$  in the 
update \eqref{equ_update_generic_asynch}, we distinguish between \cite{ParallelDistrBook}
\begin{itemize} 
	\item {\bf Totally asynchronous \glspl{algorithm}.} These are algorithms of the form 
	\eqref{equ_full_update_async_generic} with unbounded delays $\iteridx\!-\!\iteridx_{\nodeidx,\nodeidx'}$, i.e., 
	they can can become arbitrarily large. Moreover, we require that no \gls{device} stops updating, i.e., 
	the set $\updatetimes{\nodeidx}$ is infinite for each $\nodeidx=1,\ldots,\nrnodes$. 
	\item {\bf Partially asynchronous \glspl{algorithm}.} These are \glspl{algorithm} of the form \eqref{equ_full_update_async_generic} 
	with bounded delays $\iteridx\!-\!\iteridx_{\nodeidx,\nodeidx'}\!\leq\!\maxdelay$, with some fixed (but possibly unknown) maximum delay 
	$B \in \mathbb{N}$. 
	Moreover, each \gls{device} updates at least once during $\maxdelay$ consecutive clock ticks, i.e., 
	$\updatetimes{\nodeidx} \cap \{ \timeidx,\timeidx+1,\timeidx+\maxdelay-1\} \neq \emptyset$ for each $\timeidx=1,2,\ldots,$ and $\nodeidx=1,\ldots,\nrnodes$. 
\end{itemize} 
For some choices of $\fixedpointop^{(\nodeidx)}$ in \eqref{equ_update_generic_asynch}, 
a partially asynchronous algorithm can converge for any value of $\maxdelay$. However, 
there also choices of $\fixedpointop^{(\nodeidx)}$, for which a partially asynchronous 
\gls{algorithm} will only converge if $\maxdelay$ is sufficiently small \cite[Ch. 7]{ParallelDistrBook}.

{\bf Convergence Guarantees.} There is an elegant characterization of 
the convergence of totally and partially asynchronous \gls{fl} \glspl{algorithm} of 
the form \eqref{equ_full_update_async_generic}. This characterization applies 
whenever the operators $\fixedpointop^{(\nodeidx)}$, for $\nodeidx=1,\ldots,\nrnodes$, 
in \eqref{equ_full_update_async_generic} form a pseudo-contraction \cite{Feyzmahdavian2023AsynchJMLR}
\begin{equation}
	\label{equ_pseudo_contraction}
	\max_{\nodeidx=1,\ldots,\nrnodes} \normgeneric{\fixedpointop^{(\nodeidx)}\big(\localparams{1},\ldots,\localparams{\nrnodes} \big)\!-\! 
	\fixedpointop^{(\nodeidx)}\big(\estlocalparams{1},\ldots,\estlocalparams{\nrnodes} \big)}{} 
	\!\leq\! \kappa \cdot 
	\max_{\nodeidx=1,\ldots,\nrnodes} \hspace*{-1mm}\normgeneric{\localparams{\nodeidx}\!-\!\estlocalparams{\nodeidx}}{},
\end{equation}
with some contraction rate $\kappa \in [0,1)$ and some fixed-point $\estlocalparams{1},\ldots,\estlocalparams{\nrnodes}$. 

The operators $\fixedpointop^{(\nodeidx)}$, for $\nodeidx=1,\ldots,\nrnodes$, underlying \gls{gtvmin}-based 
\glspl{algorithm} are determined by the design choices of the \gls{gtvmin} building blocks. 
These include the choices of local \glspl{lossfunc} $\locallossfunc{\nodeidx}{\cdot}$, for $\nodeidx=1,\ldots,\nrnodes$ 
and edge weights $\edgeweight_{\nodeidx,\nodeidx'}$, for $\edge{\nodeidx}{\nodeidx'} \in \edges$.  
Let us next discuss specific design choices which yield operators that form a pseudo-contraction \eqref{equ_pseudo_contraction}. 

Consider the operator $\fixedpointop^{(\nodeidx)}$ defined by the update 
\eqref{equ_def_coord_min_update} of FedRelax (see Algorithm \ref{alg_fed_relax}). 
If the local \glspl{lossfunc} $\locallossfunc{\nodeidx}{\cdot}$ are \gls{strcvx},\footnote{Strictly speaking, we also need 
	to require that the epigraph of each $\locallossfunc{\nodeidx}{\cdot}$, for $\nodeidx=1,\ldots,\nrnodes$ is non-empty and closed 
	\cite{ProximalMethods}.} we can 
	decompose $\fixedpointop^{(\nodeidx)}$ as 
\begin{equation} 
	\label{equ_decomp_linear_contraction}
	\fixedpointop^{(\nodeidx)}  = \proximityop{\locallossfunc{\nodeidx}{\cdot}}{\cdot}{2 \regparam \nodedegree{\nodeidx}} \circ \mathcal{T}^{(\nodeidx)}. 
\end{equation}  
Here, we used the \gls{proxop} as defined in \eqref{equ_def_prox_op_original} as well 
as the averaging-\glspl{neighbor}-operator 
\begin{equation}
	\nonumber
	\mathcal{T}^{(\nodeidx)}: \underbrace{\mathbb{R}^{\dimlocalmodel}\times\ldots\times\mathbb{R}^{\dimlocalmodel}}_{ \mbox{ $\nrnodes$ times }} \rightarrow \mathbb{R}^{\dimlocalmodel}: \localparams{1},\ldots,\localparams{\nrnodes} \mapsto (1/\nodedegree{\nodeidx}) \sum_{\nodeidx' \in \neighbourhood{\nodeidx}} \edgeweight_{\nodeidx,\nodeidx'} \localparams{\nodeidx'}. 
\end{equation} 

It can be easily verified that the operators $\mathcal{T}^{(1)},\ldots,\mathcal{T}^{(\nrnodes)}$ are non-expansive. 
Moreover, by the basic properties of \glspl{proxop} (see, e.g., \cite[Sec.\ 6]{ryu2016primer}), the operators $$\proximityop{\locallossfunc{1}{\cdot}}{\cdot}{2 \regparam \nodedegree{1}},\ldots,\proximityop{\locallossfunc{\nrnodes}{\cdot}}{\cdot}{2 \regparam \nodedegree{\nrnodes}}$$ 
also form a pseudo-contraction with $\kappa =\frac{1}{1+(\sigma/(2 \regparam \nodedegree{\nodeidx}))}$. Combining these facts 
with \eqref{equ_decomp_linear_contraction} yields that the operators $\fixedpointop^{(\nodeidx)}$, for $\nodeidx=1,\ldots,\nrnodes$, 
form a pseudo-contraction with 
\begin{equation} 
	\label{equ_contr_factor_strg_cvx_sq_norm_penalty}
	\kappa =\frac{1}{1+(\sigma/(2 \regparam \nodedegree{\nodeidx}))}. 
\end{equation}

For any \gls{fl} \gls{algorithm} \eqref{equ_full_update_async_generic} such that 
\eqref{equ_pseudo_contraction} is satisfied, the following holds:
\begin{itemize}
	\item A totally asynchronous \gls{algorithm} of the form \eqref{equ_full_update_async_generic} converges to $\estlocalparams{1},\ldots,\estlocalparams{\nrnodes}$  \cite[Thm. 23]{Feyzmahdavian2023AsynchJMLR}.
	\item In the partially asynchronous case with maximum delay $\maxdelay$ \cite[Thm. 24]{Feyzmahdavian2023AsynchJMLR}, 
	\begin{equation}
		\label{equ_upper_bound_conv_part_async}
		\max_{\nodeidx=1,\ldots,\nrnodes} \normgeneric{\localparamsiter{\nodeidx}{\iteridx} - \estlocalparams{\nodeidx}}{}
		\leq \kappa^{\iteridx/(2\maxdelay+1)} \cdot 
		\max_{\nodeidx=1,\ldots,\nrnodes} \normgeneric{\localparamsiter{\nodeidx}{0} - \estlocalparams{\nodeidx}}{}.
	\end{equation}
\end{itemize}
The bound \eqref{equ_upper_bound_conv_part_async} is quite intuitive: 
smaller contraction factors $\kappa$ and smaller delay bounds $\maxdelay$ 
lead to faster convergence of the \gls{algorithm} \eqref{equ_full_update_async_generic}. 
Figure \ref{fig_factor_convergence_partial_ansny} illustrates the factor $\kappa^{\iteridx/(2\maxdelay+1)}$ 
for different values of $\kappa$ and maximum delay $\maxdelay$. 

The contraction factor $\kappa$ of the operators $\fixedpointop^{(\nodeidx)}$, for 
$\nodeidx=1,\ldots,\nrnodes$, arising in \gls{gtvmin}-based methods depends on the 
properties of local \glspl{lossfunc} and the connectivity of the \gls{flnetwork}. According 
to \eqref{equ_contr_factor_strg_cvx_sq_norm_penalty}, the operators underlying 
FedRelax (see \eqref{equ_def_coord_min_update} and Algorithm \ref{alg_fed_relax}), 
have a small contraction factor if we use 
\begin{itemize} 
	\item local \glspl{lossfunc} that are \gls{strcvx} with large coefficient $\sigma$, 
	\item a \gls{flnetwork} with small weighted \gls{nodedegree}s $\nodedegree{\nodeidx}$, for $\nodeidx=1,\ldots,\nrnodes$. 
\end{itemize} 
Moreover, the contraction factor \eqref{equ_contr_factor_strg_cvx_sq_norm_penalty} decreases 
with decreasing \gls{gtvmin} parameter $\regparam$. In the extreme case of $\regparam\!=\!0$ - 
where \gls{gtvmin} decomposes into fully independent local instances of \gls{erm} $\min_{\localparams{\nodeidx}} \locallossfunc{\nodeidx}{\cdot}$ 
- the contraction factor becomes $\kappa\!=\!0$. This makes sense as in this extreme case, there is 
no information sharing required among the nodes of an \gls{flnetwork}. Clearly, the delays 
$\iteridx\!-\!\iteridx_{\nodeidx,\nodeidx'}$ are then irrelevant for the performance 
of \gls{fl} \gls{algorithm}s. 


%
\begin{figure} 
\begin{minipage}{0.45\textwidth}
	\centering
	\begin{tikzpicture}
		\begin{axis}[
			width=\textwidth, 
			height=5cm,
			title={$\kappa=0.5$},
			xlabel={$\iteridx$},
			ytick={0,0.5,1},
			yticklabels={0,0.5,1},
			samples=10,
			domain=1:100,
			legend style={at={(0.5,0.24)},anchor=south west,font=\scriptsize,cells={anchor=west},draw=none,inner sep=2pt,row sep=1pt}
			]
			\addplot+[mark=square*] plot ({x}, {0.5^(x/(2*100+1))});
			\addlegendentry{$\maxdelay=100$}	
			\addplot+[mark=triangle*] plot ({x}, {0.5^(x/(2*10+1))});
			\addlegendentry{$\maxdelay=10$}
					\addplot+[mark=*] plot ({x}, {0.5^(x/(2*1+1))});
			\addlegendentry{$\maxdelay=1$}
		\end{axis}
	\end{tikzpicture}
\end{minipage}
\begin{minipage}{0.45\textwidth}
	\centering
	\begin{tikzpicture}
		\begin{axis}[
			width=\textwidth, height=5cm,
			title={$\kappa=0.99$},
			xlabel={$\iteridx$},
			yticklabels={,,},
			samples=10,
			ytick={0,0.5,1},
			domain=1:100,
			 legend style={at={(0.02,0.02)}, anchor=south west,font=\scriptsize,cells={anchor=west},draw=none,inner sep=2pt,row sep=1pt}
			]
			\addplot+[mark=square*] plot ({x}, {0.99^(x/(2*100+1))});
			\addlegendentry{$\maxdelay=100$}
			\addplot+[mark=triangle*] plot ({x}, {0.99^(x/(2*10+1))});
			\addlegendentry{$\maxdelay=10$}
			\addplot+[mark=*] plot ({x}, {0.99^(x/(2*1+1))});
			\addlegendentry{$\maxdelay=1$}
		\end{axis}
	\end{tikzpicture}
\end{minipage}
\caption{Illustration of the factor $\kappa^{\iteridx/(2\maxdelay+1)}$ in the convergence bound 
	\eqref{equ_upper_bound_conv_part_async} for a partially asynchronous \gls{fl} \gls{algorithm} 
	\eqref{equ_full_update_async_generic} using a pseudo-contraction
	(see \eqref{equ_pseudo_contraction}). \label{fig_factor_convergence_partial_ansny}}
\end{figure}

\clearpage

\subsection{Exercises}

\refstepcounter{problem}\label{prob:convgdlinalg}\textbf{\theproblem. The convergence 
	speed of \glspl{gdmethod}.} Study the convergence speed of \eqref{equ_def_basic_gradstep_lecflalg} 
	for two different collections of \gls{localdataset}s assigned to the nodes of the \gls{flnetwork} $\graph$ 
	with nodes $\nodes = \{1,2\}$ and (unit weight) edges $\edges = \{ \{1,2\}\}$. The first 
	collection of \gls{localdataset}s results in the local \gls{lossfunc}s $\locallossfunc{1}{\weight} \defeq (\weight+5)^2$ and 
	$\locallossfunc{2}{\weight} \defeq 1000 (\weight+5)^2$. The second collection of 
	\gls{localdataset}s results in the local \gls{lossfunc}s $\locallossfunc{1}{\weight} \defeq 1000 (\weight+5)^2$ and $\locallossfunc{2}{\weight} \defeq 1000 (\weight-5)^2$. Use a fixed \gls{learnrate} $\lrate \defeq 0.5 \cdot 10^{-3}$ for the 
	iteration \eqref{equ_def_basic_gradstep_lecflalg}.

\noindent\refstepcounter{problem}\label{prob:convgdlinalghomog}\textbf{\theproblem. Convergence speed for homogeneous data.}
Study the convergence speed of \eqref{equ_def_basic_gradstep_lecflalg} when 
applied to \gls{gtvmin} \eqref{equ_def_gtvmin_linreg_lec5} with the following \gls{flnetwork} $\graph$:  
Each node $\nodeidx=1,\ldots,\nrnodes$ carries a simple \gls{localmodel} with 
single parameter $\weight^{(\nodeidx)}$ and the local \gls{lossfunc} $\locallossfunc{\nodeidx}{\weight} \defeq \big(\truelabel^{(\nodeidx)}-  \feature^{(\nodeidx)} \weight^{(\nodeidx)} \big)^{2}$. The \gls{localdataset} consists 
of a constant $\feature^{(\nodeidx)}\defeq1$ and some $\truelabel^{(\nodeidx)} \in \mathbb{R}$. 
The edges $\edges$ are obtained by connecting each node $\nodeidx$ with $4$ other randomly chosen nodes. 
We learn \glspl{modelparam} $\widehat{\weight}^{(\nodeidx)}$ by repeating \eqref{equ_def_basic_gradstep_lecflalg}, 
starting with the initializations $\weight^{(\nodeidx,0)} \defeq \truelabel^{(\nodeidx)}$. 
Study the dependence of the convergence speed of \eqref{equ_def_basic_gradstep_lecflalg} 
(towards a solution of \eqref{equ_def_gtvmin_linreg_lec5}) on the value of $\regparam$ 
in \eqref{equ_def_gtvmin_linreg_lec5}. 

\noindent\refstepcounter{problem}\label{ex_data_aug_fl_relax}\textbf{\theproblem. Implementing FedRelax via \gls{dataaug}.}
Consider the application of Algorithm \ref{alg_fed_relax_agnostic} to an \gls{flnetwork} whose nodes 
carry \gls{regression} tasks. In particular, each \gls{device} $\nodeidx =1,\ldots,\nrnodes$ learns a 
\gls{hypothesis} $\localhypothesis{\nodeidx}$ to predict the numeric \gls{label} $\truelabel \in \mathbb{R}$ 
of a \gls{datapoint} with \gls{featurevec} $\featurevec$. The usefulness 
of a \gls{hypothesis} is measured by the average \gls{sqerrloss} incurred on a labelled \gls{localdataset} 
$$\localdataset{\nodeidx} \defeq \bigg\{\pair{\featurevec^{(1)}}{\truelabel^{(1)}},\ldots,\pair{\featurevec^{(\localsamplesize{\nodeidx})}}{\featurevec^{(\localsamplesize{\nodeidx})}} \bigg\}.$$
To compare the learned \gls{hypothesis} maps at the nodes of an edge $\edge{\nodeidx}{\nodeidx'}$, 
we use \eqref{equ_def_variation_non_parametric} with the \gls{sqerrloss}. 
Show that the update \eqref{equ_def_coord_min_update_nonparam} is equivalent to plain \gls{erm} \eqref{equ_def_erm} 
using a \gls{dataset} $\dataset$ that is obtained by a specific augmentation of $\localdataset{\nodeidx}$.

\noindent\refstepcounter{problem}\label{ex_fedavg_as_fixdpoint}\textbf{\theproblem. \Gls{fedavg} as \gls{fixedpointiter}.}
Consider Algorithm \ref{equ_def_fedavg_basic_linreg} for training the \glspl{modelparam} $\localparams{\nodeidx}$ of 
local \gls{linmodel}s for each $\nodeidx=1,\ldots,\nrnodes$ of an \gls{flnetwork}. Each 
client uses a constant \gls{learnrate} schedule $\lrate_{\nodeidx,\iteridx} \defeq \lrate_{\nodeidx}$. 
Try to find a collection of operators $\fixedpointop^{(\nodeidx)}: \mathbb{R}^{\nrnodes \dimlocalmodel} \rightarrow \mathbb{R}^{\dimlocalmodel}$, 
for each node $\nodeidx=1,\ldots,\nrnodes$, such that Algorithm \ref{equ_def_fedavg_basic_linreg} is equivalent 
to the fixed-point iteration 
\begin{equation} 
	\nonumber
	\localparamsiter{\nodeidx}{\iteridx+1} =  \fixedpointop^{(\nodeidx)} \big(\localparamsiter{1}{\iteridx},\ldots, \localparamsiter{\nrnodes}{\iteridx}\big). 
\end{equation} 

\noindent\refstepcounter{problem}\label{ex_pseudocontraction}\textbf{\theproblem. Fixed-Points of a pseudo-contraction.}
Show that a pseudo-contraction cannot have more than one fixed-point. 

\noindent\refstepcounter{problem}\label{ex_fedrelaxrewritten}\textbf{\theproblem. FedRelax update.}
Show that the update \eqref{equ_def_coord_min_update} of FedRelax for parametric \gls{localmodel}s 
can be rewritten as 
$$\argmin_{\localparams{\nodeidx} \in \mathbb{R}^{\dimlocalmodel}} \locallossfunc{\nodeidx}{\localparams{\nodeidx}} + \regparam \nodedegree{\nodeidx}
\normgeneric{\localparams{\nodeidx}-\estlocalparams{\neighbourhood{\nodeidx}}}{2}^{2}.$$
Here, we used $\estlocalparams{\neighbourhood{\nodeidx}}\defeq (1/\nodedegree{\nodeidx}) \sum_{\nodeidx' \in \neighbourhood{\nodeidx}} \edgeweight_{\nodeidx,\nodeidx'} \localparamsiter{\nodeidx'}{\iteridx}$ and the 
weighted \gls{nodedegree} $\nodedegree{\nodeidx} = \sum_{\nodeidx' \in \neighbourhood{\nodeidx}} \edgeweight_{\nodeidx,\nodeidx'}$ (see \eqref{equ_def_node_degree}). 

\noindent\refstepcounter{problem}\label{ex_fedrelaxfedgd}\textbf{\theproblem. \gls{FedRelax} vs. \gls{FedSGD}}
Show that the update in step \eqref{equ_gdstep_fedgd_general} of Algorithm \ref{alg_fed_gd_general} 
is obtained from the update \eqref{equ_def_coord_min_update} of \gls{FedRelax} by replacing 
the local \gls{lossfunc} $\locallossfunc{\nodeidx}{\localparams{\nodeidx}}$ with a local 
approximation by a \gls{quadfunc}, centred around $\localparamsiter{\nodeidx}{\iteridx}$.   

\noindent\refstepcounter{problem}\label{ex_fedsgdfixedpoint}\textbf{\theproblem. \gls{FedSGD} as \gls{fixedpointiter}.}
Show that Algorithm \ref{alg_fed_sgd} can be written as the distributed 
fixed-point iteration \eqref{equ_update_fixed_point_synch_param}. Try to find an 
elegant characterization of the resulting operators $\fixedpointop^{(\nodeidx)}$, for $\nodeidx=1,\ldots,\nrnodes$.

\noindent\refstepcounter{problem}\label{ex_fedrelaxasfixedpoint}\textbf{\theproblem. \gls{FedRelax} as \gls{fixedpointiter}.}
Show that Algorithm \ref{alg_fed_relax_agnostic} can be written as the distributed 
fixed-point iteration \eqref{equ_update_fixed_point_synch}. Try to find an elegant 
characterization of the resulting operators $\fixedpointop^{(\nodeidx)}$, for $\nodeidx=1,\ldots,\nrnodes$. . 

\newpage
\subsection{Proofs} 
\subsubsection{Proof of Proposition \ref{prop_upper_bound}} 
\label{proof_upper_bound_eigvals_Q_gtvmin}

The first inequality in \eqref{equ_upper_bound_eigval_Q_gtvmin_lin} follows from well-known 
results on the \gls{eigenvalue}s of a sum of symmetric matrices (see, e.g., \cite[Thm 8.1.5]{GolubVanLoanBook}). 
In particular, 
\begin{align} 
	\eigvalgen_{\rm max} \big( \mQ \big) \leq \max\big\{ \underbrace{\max_{\nodeidx=1,\ldots,\nrnodes} \eigval{\dimlocalmodel} \big( \mQ^{(\nodeidx)} \big)}_{\stackrel{\eqref{equ_summary_eigvals_Q_i}}{=} \maxeigvallocalQ }, \eigvalgen_{\rm max} \big( \regparam \LapMat{\graph} \otimes \mI \big) \big\}. 
\end{align}
The second inequality in \eqref{equ_upper_bound_eigval_Q_gtvmin_lin} uses the following upper bound 
on the maximum \gls{eigenvalue} $\eigval{\nrnodes}\big(\LapMat{\graph} \big)$ of the \gls{LapMat}: 
\begin{align}
	\label{equ_upper_bound_mL_maxdegree} 
	\eigval{\nrnodes} \big(\LapMat{\graph} \big)  & \stackrel{(a)}{=} \max_{\vv \in \sphere{\nrnodes-1}}  \vv^{T} \LapMat{\graph}  \vv \nonumber \\ 
	&  \stackrel{\eqref{equ_quad_form_Laplacian}}{=} \max_{\vv \in \sphere{\nrnodes-1}}  \sum_{\edge{\nodeidx}{\nodeidx'} \in \edges}  \edgeweight_{\nodeidx,\nodeidx'} \big( v_{\nodeidx} - v_{\nodeidx'} \big)^{2}  \nonumber \\  
	&  \stackrel{(b)}{\leq} \max_{\vv \in \sphere{\nrnodes-1}}  \sum_{\edge{\nodeidx}{\nodeidx'} \in \edges}  2 \edgeweight_{\nodeidx,\nodeidx'}  \big( v^2_{\nodeidx} + v^2_{\nodeidx'} \big) \nonumber \\ 
	&   \stackrel{(c)}{=} \max_{\vv \in \sphere{\nrnodes-1}}  \sum_{\nodeidx \in \nodes}2  v^2_{\nodeidx}  \sum_{\nodeidx' \in \neighbourhood{\nodeidx}} \edgeweight_{\nodeidx,\nodeidx'} \nonumber \\ 
	& \stackrel{\eqref{equ_def_max_node_degree}}{\leq} \max_{\vv \in \sphere{\nrnodes-1}}  \sum_{\nodeidx \in \nodes}2  v^2_{\nodeidx}  \maxnodedegree^{(\graph)} \nonumber \\ 
	& = 2   \maxnodedegree^{(\graph)}. 
\end{align} 
Here, step $(a)$ uses the \gls{cfwmaxmin} of \gls{eigenvalue}s \cite[Thm. 8.1.2.]{GolubVanLoanBook} 
and step $(b)$ uses the inequality $(u\!+\!v)^2 \leq 2 (u^2 \!+\! v^2)$ for any $u,v \in \mathbb{R}$. 
For step $(c)$ we use the identity $\sum_{\nodeidx \in \nodes} \sum_{\nodeidx' \in \neighbourhood{\nodeidx}} f(\nodeidx,\nodeidx') = \sum_{\edge{\nodeidx}{\nodeidx'}} \big( f(\nodeidx,\nodeidx') + f(\nodeidx',\nodeidx)\big)$ 
(see Figure \ref{fig_illustrate_proof_upperboundmLdegree}). The bound \eqref{equ_upper_bound_mL_maxdegree} 
is essentially tight.\footnote{Consider an \gls{flnetwork} being a chain (or path).}

\begin{figure}[htbp]
	\centering
	\begin{tikzpicture}[scale=1]
		
		
		\node[draw, circle, fill=black, inner sep=1pt] (c0) at (0:0) {$0$};
		\coordinate[] (c1) at (0:0) {};
		\coordinate[] (c2) at (10:6) {};
		\coordinate[] (c3) at (-10:6) {};
		\foreach \i in {1,2,3}{
			\draw [fill] (c\i) circle [radius=0.2] node[below=5pt] {$\i$};
		}

		\draw [line width=0.5mm] (c1) -- (c2) node[midway, above, sloped] {$\edgeweight_{1,2} 2 \big( \weight^2_{1} + \weight^2_{2} \big)$};
		\draw [line width=0.5mm] (c1) -- (c3) node[midway, below, sloped] {$\edgeweight_{1,3} 2 \big( \weight^2_{1} + \weight^2_{3} \big)$};
		
	\end{tikzpicture}
	\caption{Illustration of step $(c)$ in \eqref{equ_upper_bound_mL_maxdegree}. \label{fig_illustrate_proof_upperboundmLdegree}}
\end{figure}

\subsubsection{Proof of Proposition \ref{prop_lower_bound_eigvals_gtvmin_linreg}} 
\label{proof_lower_bound_eigvals_Q_gtvmin} 

Similar to the upper bound \eqref{equ_upper_bound_mL_maxdegree} we also start with the 
\gls{cfwmaxmin} for the \gls{eigenvalue}s of $\mQ$ in \eqref{equ_def_objec_gtvmin_lec_flalg}. 
In particular, 
\begin{equation}
	\label{equ_proof_minFLALg_eig}
	\eigval{1}  = \min_{\normgeneric{\weights}{2}^{2}=1} \weights^{T}  \mQ  \weights.  
\end{equation}
We next analyze the right-hand side of \eqref{equ_proof_minFLALg_eig} by partitioning the constraint set 
$\{\weights: \normgeneric{\weights}{2}^{2}=1 \}$ of \eqref{equ_proof_minFLALg_eig}
into two complementary regimes for the optimization variable $\weights = {\rm stack} \{ \localparams{\nodeidx} \}$. 
To define these two regimes, we use the orthogonal decomposition 
\begin{equation} 
	\label{equ_def_decpom_weight_proof_lower_bound}
	\weights = \underbrace{\mathbf{P}_{\mathcal{S}} \weights}_{=: \overline{\weights}}  + \underbrace{\mathbf{P}_{\mathcal{S}^{\perp}} \weights}_{=: \widetilde{\weights}} \mbox{ for subspace } \mathcal{S} \mbox{ in \eqref{equ_def_subspace_constant_local}}.
\end{equation} 
Explicit expressions for the orthogonal components $\overline{\weights}$, $\widetilde{\weights}$ 
are given by \eqref{equ_def_projection_constant_localparms} and \eqref{equ_def_orth_projection_constant_localparms}. 
In particular, the component $\overline{\weights}$ satisfies 
\begin{equation}
	\nonumber
	\overline{\weights} = \big( \big( \vc \big)^{T}, \ldots,  \big( \vc \big)^{T} \big)^{T} \mbox{ with } \vc \defeq {\rm avg} \big\{ \localparams{\nodeidx} \big\}_{\nodeidx=1}^{\nrnodes}. 
\end{equation} 
Note that 
\begin{equation}
	\label{equ_def_orthgo_decom_norm_proof_lower_bound}
	\normgeneric{ \weights}{2}^2  =  \normgeneric{  \overline{\weights} }{2}^2   + \normgeneric{ \widetilde{\weights}}{2}^2. 
\end{equation} 

{\bf Regime I.} This regime is obtained for $\normgeneric{ \widetilde{\weights}}{2}  \geq \rho  \normgeneric{  \overline{\weights}}{2}$. 
Since $\normgeneric{\weights}{2}^{2}=1$, and due to \eqref{equ_def_orthgo_decom_norm_proof_lower_bound}, 
we have 
\begin{equation} 
	\label{equ_def_lower_bound_variation_component} 
	\normgeneric{ \widetilde{\weights}}{2}^2   \geq \rho^{2}/(1+\rho^{2}).
\end{equation}  
This implies, in turn, via \eqref{equ_lower_bound_tv_eigval} that 
\begin{align}
	\label{equ_lower_bound_regime_I}
	\weights^{T} \mQ \weights & \stackrel{\eqref{equ_def_objec_gtvmin_lec_flalg}}{\geq} \regparam\weights^{T} \big( \LapMat{\graph}  \otimes \mathbf{I} \big) \weights   \nonumber \\ 
	& \stackrel{\eqref{equ_quad_form_Laplacian},\eqref{equ_lower_bound_tv_eigval}}{\geq} \regparam \eigval{2}\big( \LapMat{\graph} \big)   \normgeneric{ \widetilde{\weights}}{2}^2 \nonumber \\ 
	& \stackrel{\eqref{equ_def_lower_bound_variation_component}}{\geq} \regparam \eigval{2}\big( \LapMat{\graph} \big)   \rho^{2}/(1+\rho^{2}). 
\end{align} 

{\bf Regime II}. This regime is obtained for $\normgeneric{ \widetilde{\weights}}{2}  <  \rho \normgeneric{  \overline{\weights}}{2}$. 
Here we have $\normgeneric{  \overline{\weights}}{2}^{2} > (1/\rho^{2}) \big( 1 - \normgeneric{  \overline{\weights}}{2}^{2}\big)$ 
and, in turn, 
\begin{equation} 
	\label{equ_def_lower_boud_constant_component} 
	\nrnodes \normgeneric{\vc}{2}^{2} = \normgeneric{  \overline{\weights}}{2}^{2}  > 1/(1+\rho^2).
\end{equation}  
We next develop the right-hand side of \eqref{equ_proof_minFLALg_eig} according to  
\begin{align} 
	\label{equ_development_regime_II_101}
	\weights^{T} \mQ \weights & \stackrel{\eqref{equ_def_objec_gtvmin_lec_flalg}}{\geq} \sum_{\nodeidx=1}^{\nrnodes} \big( \localparams{\nodeidx}\big)^{T} \mQ^{(\nodeidx)} \localparams{\nodeidx}   \nonumber \\ 
	& \stackrel{\eqref{equ_def_decpom_weight_proof_lower_bound}}{\geq}  \sum_{\nodeidx=1}^{\nrnodes}  \big(\vc + \widetilde{\weights}^{(\nodeidx)} \big)^{T} \mQ^{(\nodeidx)}  \big(\vc + \widetilde{\weights}^{(\nodeidx)} \big) \nonumber \\
	& \stackrel{\eqref{equ_def_lower_boud_constant_component}}{\geq} \normgeneric{  \overline{\weights}}{2}^2\underbrace{\eigval{1}\bigg((1/\nrnodes) \sum_{\nodeidx=1}^{\nrnodes} \mQ^{(\nodeidx)} \bigg)}_{\bar{\lambda}_{\rm min}}+\sum_{\nodeidx=1}^{\nrnodes} \big[ 2 \big( \widetilde{\weights}^{(\nodeidx)} \big)^{T} \mQ^{(\nodeidx)} \vc + \underbrace{\big( \widetilde{\weights}^{(\nodeidx)} \big)^{T}\mQ^{(\nodeidx)}  \widetilde{\weights}^{(\nodeidx)}}_{\geq 0} \big]  \nonumber \\ 
	& \geq \normgeneric{  \overline{\weights}}{2}^2\bar{\lambda}_{\rm min}+\sum_{\nodeidx=1}^{\nrnodes} 2 \big(\widetilde{\weights}^{(\nodeidx)}\big)^{T} \mQ^{(\nodeidx)} \vc. 
\end{align}
To develop \eqref{equ_development_regime_II_101} further, we note that 
\begin{align} 
	\label{equ_bound_cross_term_lower_flalg}
	\bigg| \sum_{\nodeidx=1}^{\nrnodes}2 \big( \widetilde{\weights}^{(\nodeidx)}\big)^{T} \mQ^{(\nodeidx)} \vc \bigg| & \stackrel{(a)}{\leq} 2 \lambda_{\rm max} \normgeneric{\widetilde{\weights}}{2} \normgeneric{\overline{\weights}}{2} \nonumber \\ 
	& \stackrel{\normgeneric{ \widetilde{\weights}}{2}  <  \rho \normgeneric{  \overline{\weights}}{2}}{\leq}   2 \lambda_{\rm max} \rho \normgeneric{  \overline{\weights}}{2}^2. 
\end{align} 
Here, step $(a)$ follows from $\max_{\normgeneric{\vy}{2}=1,\normgeneric{\vx}{2}=1} \vy^{T} \mQ \vx = \lambda_{\rm max}$. 
Inserting \eqref{equ_bound_cross_term_lower_flalg} into \eqref{equ_development_regime_II_101} for $\rho = \bar{\lambda}_{\rm min}/(4 \lambda_{\rm max})$, 
\begin{align} 
	\label{equ_development_regime_II_105}
	\weights^{T} \mQ \weights &  \geq \normgeneric{  \overline{\weights}}{2}^2\bar{\lambda}_{\rm min}/2  \stackrel{\eqref{equ_def_lower_boud_constant_component}}{\geq} (1/(1+\rho^2)) \bar{\lambda}_{\rm min}/2 
\end{align}

For each $\weights$ with $\normgeneric{\weights}{2}^2 =1$, either \eqref{equ_lower_bound_regime_I} or \eqref{equ_development_regime_II_105} must hold. 

\newpage
\section{Key Variants of Federated Learning} 
\label{lec_flmainflavours} 

Chapter \ref{lec_fldesignprinciple} discussed \gls{gtvmin} as a main design principle for \gls{fl} algorithms. 
\gls{gtvmin} learns local \glspl{modelparam} that optimally balance the individual local \gls{loss} with their 
variation across the edges of an \gls{flnetwork}. Chapter \ref{lec_flalgorithms} discussed how to obtain 
practical \gls{fl} algorithms. These algorithms solve \gls{gtvmin} using distributed optimization methods, 
such as those from Chapter \ref{lec_gradientmethods}.

This chapter discusses important special cases (or ``main flavors'') of \gls{gtvmin} obtained 
for specific construction of \gls{localdataset}s, choices of \gls{localmodel}s, measures for 
their variation and the weighted edges of the \gls{flnetwork}. We next briefly summarize the 
resulting main flavors of \gls{fl} discussed in the following sections. 

Section \ref{sec_single_model_fl} discusses single-model \gls{fl} that learns \glspl{modelparam} of 
a single (global) model from \glspl{localdataset}. This single-model flavor can be obtained from 
\gls{gtvmin} using a connected \gls{flnetwork} with large \glspl{edgeweight} or, equivalently, a 
sufficient large value for the \gls{gtvmin} parameter. 

Section \ref{sec_clustered_fl} discusses how \gls{cfl} is obtained from \gls{gtvmin} over \glspl{flnetwork} 
with a \gls{cluster} structure. \Gls{cfl} exploits the presence of \glspl{cluster} (subsets of \gls{localdataset}s) 
which can be approximated using an \gls{iidasspt}. \gls{gtvmin} captures these \glspl{cluster} if they are 
well-connected by many (large weight) edges of the \gls{flnetwork}. 

\glsreset{hfl}
Section \ref{sec_horizontal_fl} discusses \gls{hfl} which is obtained from \gls{gtvmin} over 
an \gls{flnetwork} whose nodes carry different subsets of a single underlying global \gls{dataset}. 
Loosely speaking, \gls{hfl} involves \glspl{localdataset} characterized by the same set of \glspl{feature} 
but obtained from different \glspl{datapoint} from an underlying \gls{dataset}. 

Section \ref{sec_vertical_fl} discusses \gls{vfl}, which arises from applying \gls{gtvmin} to a \gls{flnetwork} 
where each node holds \gls{data} on the same individuals but with different sets of \glspl{feature}. 
A representative example involves public institutions such as tax authorities, social insurance 
agencies, and healthcare providers. While these organizations each collect distinct types of 
information, they all refer to the same underlying population, e.g., individuals identified 
by a Finnish social security number. 

Section \ref{sec_pers_fl} shows how personalized \gls{fl} can be obtained from \gls{gtvmin} by 
using specific measures for the \gls{gtv} of local \glspl{modelparam}. For example, using deep \glspl{ann} 
as \glspl{localmodel}, we might only use the \glspl{modelparam} corresponding to 
the first few input layers to define the \gls{gtv}. 

\subsection{Single-Model FL} 
\label{sec_single_model_fl} 

Some \gls{fl} use cases require to train a single (global) \gls{model} $\hypospace$ 
from a decentralized collection of \gls{localdataset}s $\localdataset{\nodeidx}$, $\nodeidx=1,\ldots,\nrnodes$ \cite{FLBookYang,LiTalwalkar2020}. 
In what follows we assume that the \gls{model} $\hypospace$ is parametrized by a 
vector $\weights \in \mathbb{R}^{\dimlocalmodel}$. Figure \ref{fig_centralized_FL} 
depicts a server-client architecture for an iterative \gls{fl} \gls{algorithm} that generates 
a sequence of (global) \glspl{modelparam} $\weights^{(\iteridx)}$, $\iteridx=1,\ldots$. 

After computing the new \glspl{modelparam} $\weights^{(\iteridx+1)}$, the server 
broadcasts it to the \gls{device}s $\nodeidx=1,\ldots,\nodeidx$ and increments the 
clock $\iteridx \defeq \iteridx+1$. In the next iteration, each \gls{device} $\nodeidx$ 
uses the current global \glspl{modelparam} $\weights^{(\iteridx)}$ to compute a local 
update $\localparamsiter{\nodeidx}{\iteridx}$ based on its \gls{localdataset} $\localdataset{\nodeidx}$. 
The precise implementation of this local update step depends on the choice of the global 
\gls{model} $\hypospace$ (trained by the server). One example of such a local update has 
been discussed in Chapter \ref{lec_flalgorithms} (see \eqref{equ_def_replace_sgd_local_min}). 
\begin{figure}[htbp]
	\begin{center}
		\begin{tikzpicture}[>=latex, node distance=2cm]
			
			\node[draw, rectangle] (server) {server};
			
			\coordinate[draw, above right = 0.4cm and 0cm of server] (globmodel) ; 
			\node at (globmodel) {global \glspl{modelparam} $\weights^{(\iteridx)}$ at time $\iteridx$};
			
			\node[draw, circle, below left = 3cm and 5cm of server] (client1) {$1$};
			\node[below = 0.2cm of client1] {$\localdataset{1}$};
			\node[draw, circle, below = 3cm of server] (client2) {$2$};
			\node[below = 0.2cm of client2] {$\localdataset{2}$};
			\node[draw, circle, below right = 3cm and 5cm of server] (client3) {$3$};
			\node[below = 0.2cm of client3,align=left] {compute $\localparamsiter{3}{\iteridx}$ based on $\weights^{(\iteridx)}$ \\ and \gls{localdataset} $\localdataset{3}$};
			
			\draw[->] (client1) -- (server) node[midway, above, sloped] {$\localparamsiter{1}{\iteridx}$};
			\draw[->] (client2) -- (server) node[midway, above, sloped] {$\localparamsiter{2}{\iteridx}$};
			\draw[->] (client3) -- (server) node[midway, above, sloped] {$\localparamsiter{3}{\iteridx}$};
			\draw[->] (server) to[out=240,in=30,looseness=1.1] node[midway, below, sloped] {$\weights^{(\iteridx)}$} (client1);	
			\draw[->] (server) to[out=280,in=80,looseness=1.1] node[midway, above, sloped] {$\weights^{(\iteridx)}$} (client2);
			\draw[->] (server) to[out=350,in=110,looseness=1.1] node[midway, above, sloped] {$\weights^{(\iteridx)}$} (client3);
		\end{tikzpicture}
	\end{center} 
	\caption{\label{fig_centralized_FL}  
		 Illustration of a server-based (centralized) \gls{fl} system during iteration $\iteridx$. 
		The server begins by broadcasting the current global \glspl{modelparam} $\weights^{(\iteridx)}$ 
		to each \gls{device} $\nodeidx \in \nodes$. Each \gls{device} $\nodeidx$ then computes an update 
		 $\localparamsiter{\nodeidx}{\iteridx}$ based on its \gls{localdataset} $\localdataset{\nodeidx}$ 
		 and the received \glspl{modelparam} $\weights^{(\iteridx)}$. These local updates $\localparamsiter{\nodeidx}{\iteridx}$ 
		 are sent back to the server, which aggregates them to obtain the updated global \glspl{modelparam} $\weights^{(\iteridx+1)}$.}
\end{figure} 

Chapter \ref{lec_flalgorithms} already hinted at an alternative to the server-based system in 
Figure \ref{fig_centralized_FL}. Indeed, we might learn \gls{localmodel} \glspl{parameter} $\localparams{\nodeidx}$ 
for each client $\nodeidx$ using a distributed optimization of \gls{gtvmin}. We can force 
the resulting \glspl{modelparam} $\localparams{\nodeidx}$ to be (approximately) identical 
by using a connected \gls{flnetwork} and a sufficiently large \gls{gtvmin} parameter $\regparam$. 

To minimize the computational complexity of the resulting single-model \gls{fl} system, we  
prefer \glspl{flnetwork} with a small number of edges such as the star graph in Figure \ref{fig:star_graph} \cite{DiestelGT}. 
However, to increase the robustness against node/link failures we should use an \gls{flnetwork} with 
more edges. This redundancy helps to ensure that the \gls{flnetwork} is connected even 
after removing some of its edges \cite{Iyer:2013aa}. 

Much like the server-based system from Figure \ref{fig_centralized_FL}, \gls{gtvmin}-based methods using 
a star graph offers a single point of failure which is the server in Figure \ref{fig_centralized_FL} or the centre 
node in Figure \ref{fig:star_graph}. Chapter \ref{lec_trustworthyfl} will discuss the \gls{robustness} 
of \gls{gtvmin}-based \gls{fl} systems in slightly more detail. 

\subsection{Clustered FL} 
\label{sec_clustered_fl} 

Single-model \gls{fl} systems require the \gls{localdataset}s to be well approximated as \gls{iid} \gls{realization}s 
from a common underlying \gls{probdist}. However, requiring homogeneous \gls{localdataset}s, 
generated from the same \gls{probdist}, might be overly restrictive. Indeed, the \gls{localdataset}s 
might be heterogeneous and need to be modelled using different \gls{probdist} \cite{Smith2017,ClusteredFLTVMinTSP}. 

\Gls{cfl} relaxes the requirement of a common \gls{probdist} underlying all \gls{localdataset}s. Instead, 
we approximate subsets of \gls{localdataset}s as \gls{iid} \gls{realization}s from a common \gls{probdist}. 
In other words, \gls{cfl} assumes that \gls{localdataset}s form \gls{cluster}s. Each \gls{cluster} $\cluster \subseteq \nodes$ 
has a cluster-specific \gls{probdist} $p^{(\cluster)}$. 

The idea of \gls{cfl} is to pool the \glspl{localdataset} $\localdataset{\nodeidx}$ in the same 
\gls{cluster} $\cluster$ to obtain a \gls{trainset} to learn cluster-specific $\widehat{\weights}^{(\cluster)}$. 
Each node $\nodeidx \in \cluster$ then uses these learned \glspl{modelparam} $\widehat{\weights}^{(\cluster)}$. 
A main challenge in \gls{cfl} is that the \gls{cluster} assignments of the \glspl{localdataset} are 
unknown in general. 

To determine a \gls{cluster} \( \cluster \), we can apply standard \gls{clustering} techniques 
(such as \gls{kmeans} or \gls{gmm}) to a \gls{vector} representation of the \glspl{localdataset} \cite[Ch.~5]{MLBasics}. 
These \gls{vector} representations can be constructed in various ways. One option is to use the \glspl{modelparam} \( \widehat{\weights} \) 
of a parametric \gls{ml} \gls{model} trained on \( \localdataset{\nodeidx} \). 
Alternatively, we can represent each \gls{localdataset} \( \localdataset{\nodeidx} \) using the \gls{gradient} of its 
local \gls{lossfunc} \( \locallossfunc{\nodeidx}{\localparams{\nodeidx}} \) (see Section~\ref{sec_measuring_dissimarity}).

We can also implement \gls{cfl} via \gls{gtvmin} with a suitably chosen \gls{flnetwork}. 
In particular, the \gls{flnetwork} should contain many edges (with large weight) 
between nodes in the same \gls{cluster} and few edges (with a small weight) 
between nodes in different \glspl{cluster}. To fix ideas, consider the \gls{flnetwork} 
in Figure \ref{fig_clustered_fl}, which contains a \gls{cluster} $\cluster=\{1,2,3\}$.

\begin{figure}[hbtp] 
	\begin{center} 
		\begin{tikzpicture}[auto,scale=0.6]
			\coordinate (i1) at (0,0);
			\coordinate (i2) at (-3,2);
			\coordinate (i3) at (-3,-2);
			
			\draw [dashed] (0.8cm,-0.2cm) arc [start angle=0, end angle=360, 
			x radius=3cm, 
			y radius=3.4cm]
			node [pos=0.5] {$\cluster$} 
			node [pos=.25] {} 
			node [pos=.5] {}  
			node [pos=.75] {};
			\coordinate (i4) at (3,0);
			\coordinate (i5) at (4,2);
			\draw [fill] (i1) circle [radius=0.2] node[below=5pt] {$\localparams{1}$};
			\draw [fill] (i2) circle [radius=0.2] node[below left = 5pt and 5pt of i2] {$\localparams{2}$};
			\draw [fill] (i3) circle [radius=0.2] node[below=5pt] {$\localparams{3}$};
			\foreach \nodeidx in {4,5}
			\draw [fill] (i\nodeidx) circle [radius=0.2] ; 
			\node[below right=2pt and 2pt of i4] {$\localparams{4}$};
			\node[above right=2pt and 2pt of i5] {$\localparams{5}$};
			\draw[line width=0.5mm] (i1) -- (i2);
			\draw[line width=0.5mm] (i2) -- (i3);
			\draw[line width=0.5mm] (i1) -- (i3);
			\draw[line width=0.5mm] (i1) -- node[midway,below]{$\partial \cluster$} (i4);
			\draw[line width=0.5mm] (i4) -- (i5);	
		\end{tikzpicture}
		\caption{The solution of \gls{gtvmin} \eqref{equ_def_gtvmin} are \gls{localmodel} \glspl{parameter} 
			that are approximately identical for all nodes in a tight-knit \gls{cluster} $\cluster$.\label{fig_clustered_fl} }
	\end{center} 
\end{figure} 

Chapter \ref{lec_fldesignprinciple} discussed how the \gls{eigenvalue}s of the \gls{LapMat} 
can be used to measure the connectivity of $\graph$. Similarly, we can measure the 
connectivity of a cluster $\cluster$ via the \gls{eigenvalue} $\eigval{2}\big( \LapMat{\cluster}  \big)$ 
of the \gls{LapMat} $\LapMat{\cluster} $ of the induced sub-graph $\indsubgraph{\graph}{\cluster}$:\footnote{The graph $\indsubgraph{\graph}{\cluster}$ 
	consists of the nodes in $\cluster$ and the edges $\edge{\nodeidx}{\nodeidx'} \in \edges$ for $\nodeidx,\nodeidx' \in \cluster$.}

The larger $\eigval{2}\big( \LapMat{\cluster} \big)$, the better the connectivity among the 
nodes in $\cluster$. While $\eigval{2}\big( \LapMat{\cluster} \big)$ describes the intrinsic 
connectivity of a \gls{cluster} $\cluster$, we also need to characterize its connectivity 
with the other nodes in the \gls{flnetwork}. To this end, we use the \gls{cluster} boundary 
\begin{equation}
	\nonumber
	\bd{\cluster} \defeq \sum_{\edge{\nodeidx}{\nodeidx'} \in  \partial \cluster} \edgeweight_{\nodeidx,\nodeidx'} \mbox{ with } \partial \cluster \defeq \big\{ \edge{\nodeidx}{\nodeidx'} \in \edges: \nodeidx \in \cluster, \nodeidx' \notin \cluster \big\}. 
\end{equation}
Note that for a single-node \gls{cluster} $\cluster = \{ \nodeidx \}$, the \gls{cluster} boundary 
coincides with the \gls{nodedegree}, $\bd{\cluster}  = \nodedegree{\nodeidx}$ (see \eqref{equ_def_node_degree}). 

Intuitively, \gls{gtvmin} tends to deliver (approximately) identical \glspl{modelparam} 
$\localparams{\nodeidx}$ for nodes $\nodeidx \in \cluster$ if $\eigval{2}\big(  \LapMat{\cluster}\big)$ is large 
and the cluster boundary $\bd{\cluster}$ is small. The following result makes this intuition 
more precise for the special case of \gls{gtvmin} \eqref{equ_def_gtvmin_linreg_lec5} for local \glspl{linmodel}. 
\begin{prop}
	\label{prop_upper_bound_clustered}
	Consider an \gls{flnetwork} $\graph$ which contains a \gls{cluster} $\cluster$ of \gls{localdataset}s 
	with labels $\labelvec^{(\nodeidx)}$ and \gls{featuremtx} $\featuremtx^{(\nodeidx)}$ related via 
	\begin{equation} 
		\label{equ_def_probmodel_linreg_node_i_clustered}
		\hspace*{-3mm}\labelvec^{(\nodeidx)}\!=\! \featuremtx^{(\nodeidx)}  \overline{\weights}^{(\cluster)}\!+\!{\bm \varepsilon}^{(\nodeidx)},  \mbox{ for all } \nodeidx \in \cluster.
	\end{equation}
	We learn \gls{localmodel} \glspl{parameter} $\estlocalparams{\nodeidx}$ via solving \gls{gtvmin} \eqref{equ_def_gtvmin_linreg_lec5}. 
	If the \gls{cluster} is connected, the error component 
	\begin{equation}
		\label{equ_def_error_component_clustered}
		\widetilde{\vw}^{(\nodeidx)} \defeq \estlocalparams{\nodeidx} - (1/|\cluster|)\sum_{\nodeidx \in \cluster} \estlocalparams{\nodeidx}
	\end{equation} 
	is upper bounded as 
	\begin{equation} 
		\label{equ_upper_bound_err_component_cluster}
		\sum_{\nodeidx \in \cluster} \normgeneric{\widetilde{\vw}^{(\nodeidx)} }{2}^{2} \leq\frac{1}{\regparam \eigval{2}\big( \LapMat{\cluster} \big)}  \bigg[\sum_{\nodeidx \in \cluster} \frac{1}{\localsamplesize{\nodeidx}} \normgeneric{{\bm \varepsilon}^{(\nodeidx)}}{2}^{2} 
		+ \regparam \bd{\cluster} 2 \bigg( \normgeneric{\overline{\weights}^{(\cluster)}}{2}^{2}\!+\!\upperboundnormestpar^{2} \bigg)\bigg].
	\end{equation} 
	Here, we used $\upperboundnormestpar \defeq \max_{\nodeidx' \in \nodes \setminus \cluster} \normgeneric{\estlocalparams{\nodeidx'}}{2}$. 
\end{prop}
\begin{proof} 
	See Section \ref{sec_proof_upper_bound_clustered}. 
\end{proof} 
The bound \eqref{equ_upper_bound_err_component_cluster} depends on the \gls{cluster} 
$\cluster$ (via the \gls{eigenvalue} $\eigval{2}\big( \LapMat{\cluster}\big)$ and the boundary 
$\bd{\cluster}$) and the \gls{gtvmin} parameter $\regparam$. Using a larger $\cluster$ typically 
results in a decreased \gls{eigenvalue} $\eigval{2}\big(  \LapMat{\cluster} \big)$.\footnote{Consider an 
	\gls{flnetwork} (with uniform edge weights) that contains a fully connected \gls{cluster} $\cluster$ 
	which is connected via a single edge with another node $\nodeidx' \in \nodes \setminus \cluster$ (see Figure \ref{fig_clustered_fl}). 
	Compare the corresponding \glspl{eigenvalue} $\eigval{2}\big(  \LapMat{\cluster} \big)$ and $\eigval{2}\big(  \LapMat{\cluster'} \big)$ 
	of $\cluster$ and the enlarged \gls{cluster} $\cluster' \defeq \cluster \cup \{ \nodeidx' \}$.}
According to \eqref{equ_upper_bound_err_component_cluster}, we should then 
increase $\regparam$ to maintain a small deviation $\widetilde{\weights}^{(\nodeidx)}$ 
of the learned local \glspl{modelparam} from their \gls{cluster}-wise average. Thus, increasing 
$\regparam$ in \eqref{equ_def_gtvmin} enforces its solutions to be approximately constant 
over increasingly larger subsets (clusters) of nodes (see Figure \ref{fig_clustered_fl_varying_alpha}). 

For a connected \gls{flnetwork} $\graph$ and a sufficiently large $\regparam$, the 
solution of \gls{gtvmin} consists of learned \glspl{modelparam} $\localparams{\nodeidx}$ 
that are approximately identical for all $\nodes=1,\ldots,\nrnodes$. The resulting approximation 
error is quantified by Proposition \ref{prop_upper_bound_clustered} for the extreme case where 
the entire \gls{flnetwork} forms a single \gls{cluster}, i.e., $\cluster = \nodes$. 
Trivially, the \gls{cluster} boundary is then equal to $0$ and the bound \eqref{equ_upper_bound_err_component_cluster} 
specializes to \eqref{equ_upper_bound_variation_component}. 

We hasten to add that the bound \eqref{equ_upper_bound_err_component_cluster} only applies 
for \gls{localdataset}s that conform with the \gls{probmodel} \eqref{equ_def_probmodel_linreg_node_i_clustered}. 
In particular, it assumes that all cluster nodes $\nodeidx \in \cluster$ have identical \glspl{modelparam} 
$\overline{\weights}^{(\cluster)}$. Trivially, this is no restriction if we allow for arbitrary 
error terms ${\bm \varepsilon}^{(\nodeidx)}$ in the \gls{probmodel} \eqref{equ_upper_bound_err_component_cluster}. 
However, as soon as we place additional assumptions on these error terms (such as being \gls{realization}s of \gls{iid} 
Gaussian \gls{rv}s) we should verify their validity using principled statistical tests \cite{Spiegelhalter:1980aa,Luetkepol2005}. 
Finally, we might replace $\normgeneric{\overline{\weights}^{(\cluster)}}{2}^{2}$ in \eqref{equ_upper_bound_err_component_cluster} 
with an upper bound for this quantity.

\begin{figure}[hbtp] 
	\begin{center} 
		\begin{minipage}{.3\textwidth}
			\begin{tikzpicture}[auto,scale=0.4]
				\coordinate (i1) at (0,0);
				\coordinate (i2) at (-3,2);
				\coordinate (i3) at (-3,-2);
				\draw [dashed] (0.8cm,-0.2cm) arc [start angle=0, end angle=360, 
				x radius=1cm, 
				y radius=1.4cm]
				node [pos=0.9,below] {$\cluster$} 
				node [pos=.25] {} 
				node [pos=.5] {}  
				node [pos=.75] {};
				\coordinate (i4) at (3,0);
				\draw [fill] (i1) circle [radius=0.2]; 
				\draw [fill] (i2) circle [radius=0.2]; 
				\draw [fill] (i3) circle [radius=0.2]; 
				\foreach \nodeidx in {4}
				\draw [fill] (i\nodeidx) circle [radius=0.2] ;
				\draw[line width=0.5mm] (i1) -- (i2);
				\draw[line width=0.5mm] (i2) -- (i3);
				\draw[line width=0.5mm] (i1) -- (i3);
				\draw[line width=0.5mm] (i1) -- (i4);
			\end{tikzpicture}
			
			small $\regparam$
		\end{minipage}
		\hspace*{-10mm}
		\begin{minipage}{.3\textwidth}
			\begin{tikzpicture}[auto,scale=0.4]
				\coordinate (i1) at (0,0);
				\coordinate (i2) at (-3,2);
				\coordinate (i3) at (-3,-2);
				
				\draw [dashed] (0.8cm,-0.2cm) arc [start angle=0, end angle=360, 
				x radius=3cm, 
				y radius=3.4cm]
				node [pos=0.5,below=5mm] {$\cluster$} 
				node [pos=.25] {} 
				node [pos=.5] {}  
				node [pos=.75] {};
				\coordinate (i4) at (3,0);
				\draw [fill] (i1) circle [radius=0.2] ; 
				\draw [fill] (i2) circle [radius=0.2] ; 
				\draw [fill] (i3) circle [radius=0.2] ; 
				\foreach \nodeidx in {4}
				\draw [fill] (i\nodeidx) circle [radius=0.2] ; 
				\draw[line width=0.5mm] (i1) -- (i2);
				\draw[line width=0.5mm] (i2) -- (i3);
				\draw[line width=0.5mm] (i1) -- (i3);
				\draw[line width=0.5mm] (i1) -- (i4);
			\end{tikzpicture}
			
			moderate $\regparam$
		\end{minipage}
		\hspace*{-1mm}
		\begin{minipage}{.3\textwidth}
			\begin{tikzpicture}[auto,scale=0.4]
				\coordinate (i1) at (0,0);
				\coordinate (i2) at (-3,2);
				\coordinate (i3) at (-3,-2);
				
				\draw [dashed] (4.2cm,-0.2cm) arc [start angle=0, end angle=360, 
				x radius=4.5cm, 
				y radius=3.4cm]
				node [pos=0.5] {$\cluster$} 
				node [pos=.25] {} 
				node [pos=.5] {}  
				node [pos=.75] {};
				\coordinate (i4) at (3,0);
				\draw [fill] (i1) circle [radius=0.2] ; 
				\draw [fill] (i2) circle [radius=0.2] ; 
				\draw [fill] (i3) circle [radius=0.2] ; 
				\foreach \nodeidx in {4}
				\draw [fill] (i\nodeidx) circle [radius=0.2] ; 
				\draw[line width=0.5mm] (i1) -- (i2);
				\draw[line width=0.5mm] (i2) -- (i3);
				\draw[line width=0.5mm] (i1) -- (i3);
				\draw[line width=0.5mm] (i1) --  (i4);
			\end{tikzpicture}
			
			large $\regparam$
		\end{minipage}
		\caption{
			As the \gls{regularization} parameter $\regparam$ increases, the solutions 
			of the \gls{gtvmin} \eqref{equ_def_gtvmin} become approximately constant over larger subsets of nodes, i.e., 
			they exhibit stronger clustering. \label{fig_clustered_fl_varying_alpha} }
	\end{center} 
\end{figure}

\newpage
\subsection{Horizontal FL} 
\label{sec_horizontal_fl}

\Gls{hfl} uses \glspl{localdataset} $\localdataset{\nodeidx}$, for $\nodeidx \in \nodes$, 
that contain \glspl{datapoint} characterized by the same \gls{feature}s \cite{HFLChapter2020}. 
As illustrated in Figure \ref{fig_horizontal}, we can think of each \gls{localdataset} $\localdataset{\nodeidx}$ 
as being a subset (or \gls{batch}) of an underlying global \gls{dataset} 
$$\dataset^{(\rm global)} \defeq \left\{ \pair{\featurevec^{(1)}}{\truelabel^{(1)}},\ldots,\pair{\featurevec^{(\samplesize)}}{\truelabel^{(\samplesize)}} \right\}.$$ 
In particular, \gls{localdataset} $\localdataset{\nodeidx}$ is constituted by the \glspl{datapoint} of $\dataset^{(\rm global)}$ 
with indices in $\{ \sampleidx_{1},\ldots,\sampleidx_{\localsamplesize{\nodeidx}} \}$,  
$$ \localdataset{\nodeidx} \defeq \left\{ \pair{\featurevec^{(\sampleidx_{1})}}{\truelabel^{(\sampleidx_{1})}},\ldots,\pair{\featurevec^{(\sampleidx_{\localsamplesize{\nodeidx}})}}{\truelabel^{(\sampleidx_{\localsamplesize{\nodeidx}})}}\right\}.$$

\begin{figure}[htbp]
	\begin{center}
		\begin{tikzpicture}[]
			\matrix (feature-matrix) [matrix of math nodes, nodes in empty cells,
			left delimiter={[}, right delimiter={]}, ] {
				x^{(1)}_{1} & x^{(1)}_{2} & \cdots & x^{(1)}_{\dimlocalmodel} & y^{(1)}  \\
				x^{(2)}_{1} & x^{(2)}_{2} & \cdots & x^{(2)}_{\dimlocalmodel} & y^{(2)}  \\
				\vdots & \vdots & \ddots & \vdots & \vdots \\
				x^{(\samplesize)}_{1} & x^{(\samplesize)}_{2} & \cdots & x^{(\samplesize)}_{\dimlocalmodel} & \truelabel^{(\samplesize)} \\
			};
			\draw[line width=0.2mm,decorate,decoration={brace,amplitude=5pt,mirror},yshift=0pt] 
			([xshift=-40pt, yshift=-10pt]  feature-matrix-4-1.south) -- ([xshift=20pt, yshift=-20pt] feature-matrix-4-5.east) node [black,midway,below,yshift=-10pt] {$\dataset^{(\rm global)}$};
			
			
			
			\draw [line width=0.2mm,dashed] ([xshift=-20pt, yshift=0pt]  feature-matrix-1-1.north west) rectangle ([xshift=20pt] feature-matrix-2-5.south east) node [black,midway,xshift=4cm] {$\localdataset{1}$};
			
			\draw [line width=0.2mm,dashed] ([xshift=-20pt, yshift=-5pt]  feature-matrix-3-1.north west) rectangle ([xshift=20pt] feature-matrix-4-5.south east) node [black,midway,xshift=4cm] {$\localdataset{\nodeidx}$};
			
		\end{tikzpicture}
	\end{center}
	\caption{\Gls{hfl} uses the same \gls{feature}s to characterize \gls{datapoint}s in different 
		\gls{localdataset}s. Different \glspl{localdataset} are constituted by different subsets of 
		\glspl{datapoint} out of an underlying global \gls{dataset}.}
	\label{fig_horizontal}
\end{figure} 

We can interpret \gls{hfl} as a generalization of \gls{ssl} \cite{SemiSupervisedBook}: 
For some \gls{localdataset}s $\nodeidx \in \mathcal{U}$ we might not have access to 
the \gls{label} values of \gls{datapoint}s. Still, we can use the \gls{feature}s of the \gls{datapoint}s 
to construct (the weighted edges of) the \gls{flnetwork}. To implement \gls{ssl}, we 
can solve \gls{gtvmin} using a trivial \gls{lossfunc} $\locallossfunc{\nodeidx}{\localparams{\nodeidx}} = 0$ 
for each unlabelled node $\nodeidx \in \mathcal{U}$. Solving \gls{gtvmin} delivers 
\glspl{modelparam} $\localparams{\nodeidx}$ for all nodes $\nodeidx$ (including the unlabelled ones $\mathcal{U}$). 
\Gls{gtvmin}-based methods combine the information in the labelled \gls{localdataset}s $\localdataset{\nodeidx}$, 
for $\nodeidx \in \nodes \setminus \mathcal{U}$ and their connections (via the edges of $\graph$) 
with nodes in $\mathcal{U}$ (see Figure \ref{fig_hfl_ssl}).

\begin{figure}[hbtp] 
	\begin{center} 
		\begin{tikzpicture}[auto,scale=0.8]
			
			\coordinate (i1) at (0,0);
			\coordinate (i2) at (-3,2);
			\coordinate (i3) at (-3,-2);
			
			\draw [fill] (i1) circle [radius=0.2] node[below=5pt] {};
			\draw [fill] (i3) circle [radius=0.2] node[below=5pt] {};
			
			\draw[line width=0.5mm] (i1) -- (i2);
			\draw[line width=0.5mm] (i2) -- (i3);
			\draw[line width=0.5mm] (i1) -- (i3);
			\draw [fill=white] (i2) circle [radius=0.2] node[below left = 5pt and 5pt of i2] {$\nodeidx \in \mathcal{U}$};
			
		\end{tikzpicture}
		\caption{\Gls{hfl} includes \gls{ssl} as a special case. \gls{ssl} involves a subset of 
			nodes $\mathcal{U}$, for which the \glspl{localdataset} do not contain \glspl{label}. 
			We can take this into account by using the trivial \gls{lossfunc} 
			$\locallossfunc{\nodeidx}{\cdot}=0$ for each node $\nodeidx \in \mathcal{U}$. However, 
			we can still use the \gls{feature}s in $\localdataset{\nodeidx}$ to 
			construct an \gls{flnetwork} $\graph$. 	\label{fig_hfl_ssl} }
	\end{center} 

\end{figure}

\newpage
\subsection{Vertical FL} 
\label{sec_vertical_fl}
\Gls{vfl} uses \glspl{localdataset} that are constituted by the same (identical) 
\glspl{datapoint}. However, each \gls{localdataset} uses a different choice of \glspl{feature} to 
characterize these \glspl{datapoint} \cite{VFLChapter}. Formally, \gls{vfl} applications 
revolve around an underlying global \gls{dataset} 
$$\dataset^{(\rm global)} \defeq \left\{ \pair{\featurevec^{(1)}}{\truelabel^{(1)}},\ldots,\pair{\featurevec^{(\samplesize)}}{\truelabel^{(\samplesize)}} \right\}.$$
Each \gls{datapoint} in the global \gls{dataset} is characterized by $\nrfeatures'$ \gls{feature}s $\featurevec^{(\sampleidx)} = \big( \feature^{(\sampleidx)}_{1},\ldots, \feature^{(\sampleidx)}_{\nrfeatures'} \big)^{T}$. 
The global \gls{dataset} can only be accessed indirectly via \glspl{localdataset} that use different subsets of 
the \glspl{featurevec} $\featurevec^{(\sampleidx)}$ (see Fig.\ \ref{fig_vertical_FL}). 

For example, the \gls{localdataset} $\localdataset{\nodeidx}$ consists of \glspl{featurevec}  
$$ \featurevec^{(\nodeidx,\sampleidx)} = \big(\feature^{(\sampleidx)}_{\featureidx_{1}},\ldots,\feature^{(\sampleidx)}_{\featureidx_{\nrfeatures}} \big)^{T}.$$
Here, we used a subset $\mathcal{J}^{(\nodeidx)} \defeq \{ \featureidx_{1}, \ldots, \featureidx_{\nrfeatures} \}$ 
of the original $\nrfeatures'$ \glspl{feature} (entries of $\featurevec^{(\sampleidx)}$). 
At least one node $\nodeidx'$ must have a \gls{localdataset} $\localdataset{\nodeidx'}$ 
that contains the \gls{label} values $\truelabel^{(1)},\ldots,\truelabel^{(\samplesize)}$. 

A potential toy application for vertical \gls{fl} is a national social insurance system. The global \gls{dataset} 
comprises \gls{datapoint}s representing individuals enrolled in the system. Each individual is characterized 
by multiple sets of \glspl{feature} sourced from different institutions: Healthcare providers contribute 
medical records, offering health-related \glspl{feature}. Financial service providers, such as banks, supply 
\glspl{feature} that characterize the economic situation of the individual. Some individuals participate 
in retailer loyalty programs, which generate consumer behaviour \glspl{feature}. Additionally, social 
network can provide real-time data on user activities and mobility patterns, further enriching the 
available \glspl{feature}. Since these diverse data sources belong to separate entities, 
\gls{vfl} enables collaborative learning while preserving data privacy.

\begin{figure}[htbp]
	\begin{center}
		\begin{tikzpicture}[]
			
			\matrix (feature-matrix) [matrix of math nodes, nodes in empty cells,
			left delimiter={[}, right delimiter={]}, ] {
				x^{(1)}_{1} & x^{(1)}_{2} & \cdots & x^{(1)}_{\dimlocalmodel} & y^{(1)}  \\
				x^{(2)}_{1} & x^{(2)}_{2} & \cdots & x^{(2)}_{\dimlocalmodel} & y^{(2)}  \\
				\vdots & \vdots & \ddots & \vdots & \vdots \\
				x^{(\samplesize)}_{1} & x^{(\samplesize)}_{2} & \cdots & x^{(\samplesize)}_{\dimlocalmodel} & \truelabel^{(\samplesize)} \\
			};
			
			\draw[line width=0.2mm,decorate,decoration={brace,amplitude=5pt,mirror},yshift=0pt] 
			([xshift=-40pt, yshift=-10pt]  feature-matrix-4-1.south) -- ([xshift=20pt, yshift=-20pt] feature-matrix-4-5.east) node [black,midway,below,yshift=-10pt] {$\dataset^{(\rm global)}$};
			
			
			
			\draw [line width=0.2mm,dashed] ([xshift=-5pt, yshift=5pt]  feature-matrix-1-1.north west) rectangle ([xshift=-3pt] feature-matrix-4-1.south east) node [black,midway,xshift=-1cm,yshift=2cm] {$\localdataset{1}$};
			
			\draw [line width=0.2mm,dashed] ([xshift=0pt, yshift=15pt]  feature-matrix-1-2.north west) rectangle ([xshift=25pt] feature-matrix-4-3.south east) node [black,midway,yshift=2.1cm] {$\localdataset{\nodeidx}$};
			
		\end{tikzpicture}
	\end{center}
	\caption{\Gls{vfl} uses \glspl{localdataset} that are derived from the same \glspl{datapoint}. 
		The \glspl{localdataset} differ in the choice of \glspl{feature} used to characterize the common 
		\glspl{datapoint}.	\label{fig_vertical_FL}}
	
\end{figure} 

\subsection{Personalized Federated Learning} 
\label{sec_pers_fl}
Consider \gls{gtvmin} \eqref{equ_def_gtvmin} for learning \gls{localmodel} \glspl{parameter} $\estlocalparams{\nodeidx}$ 
for each \gls{localdataset} $\localdataset{\nodeidx}$. If the value of $\regparam$ in \eqref{equ_def_gtvmin} 
is not too large, the \gls{localmodel} \glspl{parameter} $\estlocalparams{\nodeidx}$ 
can be different for each $\nodeidx \in \nodes$. However, the \gls{localmodel} \glspl{parameter} 
are still coupled via the \gls{gtv} term in \eqref{equ_def_gtvmin}. 

For some \gls{fl} use-cases we should use different coupling strengths for 
different components of the \gls{localmodel} \glspl{parameter}. For example, 
if \glspl{localmodel} are deep \glspl{ann} we might enforce the parameters of input 
layers to be identical while the \glspl{parameter} of the deeper layers might be 
different for each local \gls{dataset}. 

The partial \gls{parameter} sharing for \glspl{localmodel} can be implemented in many 
different ways \cite[Sec. 4.3.]{FLBookLudwig}: 
\begin{itemize} 
	\item One way is to use a choice of the \gls{gtv} 
	penalty that is different from $\gtvpenalty=\normgeneric{\localparams{\nodeidx} - \localparams{\nodeidx'}}{2}^{2}$. 
	In particular, we could construct the penalty function as a combination of two terms, 
	\begin{equation}
		\label{equ_def_gtv_penalty_components}
		\gtvpenalty \big(\localparams{\nodeidx} - \localparams{\nodeidx'} \big) \defeq 	\regparam^{(1)} \gtvpenalty^{(1)} \big(\localparams{\nodeidx} - \localparams{\nodeidx'} \big) 
		+\regparam^{(2)}  \gtvpenalty^{(2)} \big(\localparams{\nodeidx} - \localparams{\nodeidx'} \big). 
	\end{equation}
	The functions $\gtvpenalty^{(1)}$ and $\gtvpenalty^{(2)}$ measure different components 
	of the variation $\localparams{\nodeidx} - \localparams{\nodeidx'}$ across the edge $\edge{\nodeidx}{\nodeidx'} \in \edges$. 
	For example, we might construct $\gtvpenalty^{(1)}$ and $\gtvpenalty^{(2)}$ by \eqref{equ_def_variation_non_parametric} 
	with different choices for the \gls{dataset} $\dataset^{\edge{\nodeidx}{\nodeidx'}}$. 
	
	\item Moreover, we might use different \gls{regularization} strengths $\regparam^{(1)}$ and 
	$\regparam^{(2)}$ for different penalty components in \eqref{equ_def_gtv_penalty_components} 
	to enforce different subsets of the \glspl{modelparam} to be clustered with different granularity, i.e., 
	enforcing some of the \glspl{modelparam} to be constant across larger subsets of nodes. 
	
	\item For \glspl{localmodel} being deep \glspl{ann}, we enforce identical \glspl{modelparam} in the layers 
	closer to the input. In contrast, the layers closer to the output are allowed to have different \glspl{parameter} 
	across \glspl{device}. Figure \ref{fig_personalized_FL} illustrates this concept for \glspl{localmodel} constituted 
	by \glspl{ann} with a single hidden layer.
	
	\item Yet another technique for partial sharing of \glspl{modelparam} is to train a hyper-model which, 
	in turn, is used to initialize the training of \glspl{localmodel} \cite{PersFLHypernet}. 
\end{itemize} 

\begin{figure}[htbp]
	\begin{tikzpicture}[node distance=2cm]
		\coordinate (input1) at (0,0);
		\draw[fill=black] (input1) circle [radius=0.1];
		
		\foreach \h in {1,2} {
			\coordinate (hidden1\h) at (2cm,2*\h-3);
			\draw[fill=black] (hidden1\h) circle [radius=0.1];
		}
		
		\coordinate (output1) at (4cm,0);
		\draw[fill=black] (output1) circle [radius=0.1];
		\node [below=1mm of output1] {$\hypothesis^{(1)}(\feature)$};
		
		\draw[dashed,xshift=1cm] (0,0) ellipse (0.24cm and 0.8cm) node [pos=0.1, above=8mm] {$\vu^{(1)}$};
		\draw[dashed,xshift=3cm] (0,0) ellipse (0.24cm and 0.8cm) node [pos=0.1, above right=8mm and 1mm] {$\vv^{(1)}$};
		
		\coordinate (input2) at (8cm,0);
		\draw[fill=black] (input2) circle [radius=0.1];
		
		\foreach \h in {1,2} {
			\coordinate (hidden2\h) at (10cm,2*\h-3);
			\draw[fill=black] (hidden2\h) circle [radius=0.1];
		}
		
		\coordinate (output2) at (12cm,0);
		\draw[fill=black] (output2) circle [radius=0.1];
		\node [below=1mm of output2] {$\hypothesis^{(2)}(\feature)$};
		
		\draw[dashed,xshift=9cm] (0,0) ellipse (0.24cm and 0.8cm);
		
		\coordinate (input3) at (5cm,-3cm);
		\draw[fill=black] (input3) circle [radius=0.1];
		
		\foreach \h in {1,2} {
			\coordinate (hidden3\h) at (7cm,2*\h-6);
			\draw[fill=black] (hidden3\h) circle [radius=0.1];
		}
		
		\coordinate (output3) at (9cm,-3cm);
		\draw[fill=black] (output3) circle [radius=0.1];
		\node [below=1mm of output3] {$\hypothesis^{(3)}(\feature)$};
		
		\draw[dashed,xshift=6cm,yshift=-3cm] (0,0) ellipse (0.24cm and 0.8cm);
		
		\foreach \h in {1,2} {
			\draw[->] (input1) -- (hidden1\h);
			\draw[->] (hidden1\h) -- (output1);
			\draw[->] (input2) -- (hidden2\h);
			\draw[->] (hidden2\h) -- (output2);
			\draw[->] (input3) -- (hidden3\h);
			\draw[->] (hidden3\h) -- (output3);
		}
	\end{tikzpicture}
	\caption{\label{fig_personalized_FL} Personalized \gls{fl} with \gls{localmodel}s being \gls{ann}s with a single 
		hidden layer. The \gls{ann} $\hypothesis^{(\nodeidx)}$ is parametrized by the vector 
		$\localparams{\nodeidx} = \bigg( \big( \vu^{(\nodeidx)}\big)^{T}, \big(\vv^{(\nodeidx)} \big)^{T} \bigg)^{T}$, 
		with \glspl{parameter} $\vu^{(\nodeidx)}$ of the hidden layer and the \glspl{parameter} $\vv^{(\nodeidx)}$ 
		of the output layer. We couple the training of $\vu^{(\nodeidx)}$ via \gls{gtvmin} using the \gls{discrepancy} 
		measure $\gtvpenalty = \normgeneric{\vu^{(\nodeidx)}- \vu^{(\nodeidx')}}{2}^{2}$. }
\end{figure} 

\newpage
\subsection{Few-Shot Learning}
Some \gls{ml} applications involve \gls{datapoint}s belonging to a large number of 
different categories. A prime example is the detection of a specific object in a given 
image \cite{LearnCompare2018,conf/iclr/SatorrasE18}. Here, the object category is 
the \gls{label} $\truelabel \in \labelspace$ of a \gls{datapoint} (image). The \gls{labelspace} 
$\labelspace$ is constituted by the possible object categories and, in turn, can be 
quite large. Moreover, for some categories, we might only have a few example images in 
the \gls{trainset}.

Few-shot learning exploits structural similarities between object categories to accurately 
detect objects with limited (or even no) training examples. A principled approach to 
few-shot learning is \gls{gtvmin}, which leverages relational information between categories.
To formalize this approach, we define an \gls{flnetwork} $\graph = (\nodes, \edges, \edgeweights)$, 
where each node $\nodeidx \in \nodes$ corresponds to an element of the \gls{labelspace} $\labelspace$. 
The \gls{edgeweight}s $\edgeweights$ encode prior knowledge about category relationships, 
providing a structured way to propagate information between object categories.

Each node $\nodeidx$ in $\graph$ represents a distinct object category and corresponding 
object detector. Solving \gls{gtvmin} yields \glspl{modelparam} $\estlocalparams{\nodeidx}$ for 
each of these specialized object detectors. The coupling of these tailored object detectors 
via \gls{gtvmin} enables knowledge transfer across categories, improving detection 
performance even in low-\gls{data} regimes.


\clearpage
\subsection{Exercises}

\refstepcounter{problem}\label{prob:hfllinreg}\textbf{\theproblem. Horizontal FL of a Linear Model \cite[Sec. 8.2]{DistrOptStatistLearningADMM}} 
\Gls{linreg} learns the \glspl{modelparam} of a \gls{linmodel} by minimizing the average \gls{sqerrloss} 
on a given \gls{dataset} $\dataset$. Consider an application where the \gls{datapoint}s are gathered 
by different \glspl{device}. We can model such an application using an \gls{flnetwork} with 
nodes $\nodeidx$ carrying different subsets of $\dataset$. Construct an instance of \gls{gtvmin} such 
that its solutions coincide (approximately) with the solution of plain vanilla \gls{linreg}. 

\refstepcounter{problem}\label{prob:vfllinreg}\textbf{\theproblem. Vertical FL of a Linear Model \cite[Sec. 8.3]{DistrOptStatistLearningADMM}} 
\Gls{linreg} learns the \glspl{modelparam} of a \gls{linmodel} by minimizing the average \gls{sqerrloss} 
on a given \gls{dataset} $\dataset$. Consider an application where the \gls{feature}s of a \gls{datapoint} 
are measured by different \glspl{device}. We can model such an application using an \gls{flnetwork} with 
nodes $\nodeidx$ carrying different \gls{feature}s of the same \gls{dataset} $\dataset$. In particular, 
node $\nodeidx$ carries the \glspl{feature} $\feature_{\featureidx}$ with $\featureidx \in \mathcal{J}^{(\nodeidx)}$. 
Construct an instance of \gls{gtvmin} such that its solutions coincide (approximately) with the 
solution of plain vanilla \gls{linreg}.

\newpage
\subsection{Proofs} 
\subsubsection{Proof of Proposition \ref{prop_upper_bound_clustered}} 
\label{sec_proof_upper_bound_clustered} 

To verify \eqref{equ_upper_bound_err_component_cluster}, we follow a similar 
argument as used in the proof of Proposition \ref{prop_second_component_error}. 

First, we decompose the \gls{objfunc} $f \big( \weights \big)$ in \eqref{equ_def_gtvmin_linreg_lec5} as follows: 
\begin{align}
	f(\weights) & = \nonumber \\ 
	& \hspace*{-10mm} \underbrace{\sum_{\nodeidx \in \cluster} (1/\localsamplesize{\nodeidx}) \normgeneric{\labelvec^{(\nodeidx)} \!-\! \featuremtx^{(\nodeidx)} \weights^{(\nodeidx)}}{2}^{2}\!+\!\regparam \bigg[ \sum_{\nodeidx,\nodeidx' \in \cluster} 
		\edgeweight_{\nodeidx,\nodeidx'}  \normgeneric{\weights^{(\nodeidx)} \!-\!\weights^{(\nodeidx')}}{2}^{2}\!+\hspace*{-3mm}\sum_{\edge{\nodeidx}{\nodeidx'} \in \partial \cluster} \hspace*{-3mm}
		\edgeweight_{\nodeidx,\nodeidx'}  \normgeneric{\weights^{(\nodeidx)} \!-\!\weights^{(\nodeidx')}}{2}^{2} \bigg]}_{=: f'\big( \weights \big)} \nonumber \\ 
	& + f''\big( \weights \big). 
\end{align} 
Note that only the first component $f'$ depends on the \gls{localmodel} \glspl{parameter} $\localparams{\nodeidx}$ of 
cluster nodes $\nodeidx \in \cluster$. Let us introduce the shorthand $f'\big( \weights^{(\nodeidx)} \big)$ for the function 
obtained from $f'(\weights)$ for varying $\localparams{\nodeidx}$, $\nodeidx \in \cluster$, 
but fixing $\localparams{\nodeidx'} \defeq \estlocalparams{\nodeidx'}$ for $\nodeidx' \notin \cluster$. 

We obtain the bound \eqref{equ_upper_bound_err_component_cluster} via a proof by contradiction: 
If \eqref{equ_upper_bound_err_component_cluster} does not hold, the \gls{localmodel} \glspl{parameter} $\overline{\weights}^{(\nodeidx)} \defeq \overline{\weights}^{(\cluster)}$, 
for $\nodeidx \in \cluster$, result in a smaller value $f'\big( \overline{\weights}^{(\nodeidx)} \big) < f'\big( \widehat{\weights}^{(\nodeidx)} \big)$ 
than the choice $\estlocalparams{\nodeidx}$, for $\nodeidx \in \cluster$. This would contradict the 
fact that $\widehat{\weights}^{(\nodeidx)}$ is a solution to \eqref{equ_def_gtvmin_linreg_lec5}. 

First, note that 
\begin{align}
	f'\big( \overline{\weights}^{(\nodeidx)} \big)  &= 	 \sum_{\nodeidx \in \cluster} (1/\localsamplesize{\nodeidx}) \normgeneric{\labelvec^{(\nodeidx)} \!-\! \featuremtx^{(\nodeidx)} \overline{\weights}^{(\cluster)}}{2}^{2} \nonumber \\ 
	& \hspace*{20mm} \!+\!\regparam\bigg[ \sum_{\substack{\edge{\nodeidx}{\nodeidx'} \in \edges \\ \nodeidx,\nodeidx' \in \cluster}} 
	\edgeweight_{\nodeidx,\nodeidx'}  \normgeneric{\overline{\weights}^{(\cluster)} \!-\!\overline{\weights}^{(\cluster)}}{2}^{2}+ \sum_{\substack{\edge{\nodeidx}{\nodeidx'} \in \edges \\ \nodeidx \in \cluster, \nodeidx' \notin \cluster}}
	\edgeweight_{\nodeidx,\nodeidx'}  \normgeneric{\overline{\weights}^{(\cluster)} \!-\!\widehat{\weights}^{(\nodeidx')}}{2}^{2} \bigg] \nonumber \\
	& \stackrel{\eqref{equ_def_probmodel_linreg_node_i_clustered}}{=}  \sum_{\nodeidx \in \cluster} (1/\localsamplesize{\nodeidx}) \normgeneric{{\bm \varepsilon}^{(\nodeidx)}}{2}^{2} 
	+ \regparam \sum_{\substack{\edge{\nodeidx}{\nodeidx'} \in \edges \\ \nodeidx \in \cluster, \nodeidx' \notin \cluster}}
	\edgeweight_{\nodeidx,\nodeidx'}  \normgeneric{\overline{\weights}^{(\cluster)} \!-\!\widehat{\weights}^{(\nodeidx')}}{2}^{2} \nonumber \\
	&\stackrel{(a)}{\leq} \sum_{\nodeidx \in \cluster} (1/\localsamplesize{\nodeidx}) \normgeneric{{\bm \varepsilon}^{(\nodeidx)}}{2}^{2} 
	+ \regparam \sum_{\substack{\edge{\nodeidx}{\nodeidx'} \in \edges \\ \nodeidx \in \cluster, \nodeidx' \notin \cluster}}
	2 \edgeweight_{\nodeidx,\nodeidx'}  \bigg( \normgeneric{\overline{\weights}^{(\cluster)}}{2}^{2}+ \normgeneric{\widehat{\weights}^{(\nodeidx')}}{2}^{2} \bigg)  \nonumber \\ 
	& \leq  \sum_{\nodeidx \in \cluster} (1/\localsamplesize{\nodeidx}) \normgeneric{{\bm \varepsilon}^{(\nodeidx)}}{2}^{2} 
	+ \regparam \bd{\cluster} 2 \bigg( \normgeneric{\overline{\weights}^{(\cluster)}}{2}^{2}\!+\!\upperboundnormestpar^{2} \bigg).  \label{equ_norm_error_func_value_clustered}
\end{align} 
Step $(a)$ uses the inequality $\normgeneric{\vu\!+\!\vv}{2}^{2} \leq 2\big(\normgeneric{\vu}{2}^{2}\!+\!\normgeneric{\vv}{2}^{2}\big)$ which 
is valid for any two vectors $\vu,\vv \in \mathbb{R}^{\dimlocalmodel}$. 

On the other hand, 
\begin{align}
	\label{equ_obj_function_gtvmin_average_ac_component_clustered}
	f' \big( \widehat{\weights}^{(\nodeidx)} \big) & \geq \regparam \sum_{\nodeidx,\nodeidx' \in \cluster} 
	\edgeweight_{\nodeidx,\nodeidx'}  \underbrace{\normgeneric{\widehat{\weights}^{(\nodeidx)} \!-\!\widehat{\weights}^{(\nodeidx')}}{2}^{2}}_{\stackrel{\eqref{equ_def_error_component_clustered}}{=}\normgeneric{\widetilde{\weights}^{(\nodeidx)} \!-\!\widetilde{\weights}^{(\nodeidx')}}{2}^{2}}  \nonumber \\ 
	& \stackrel{\eqref{equ_lower_bound_tv_eigval}}{\geq} \regparam \eigval{2}\big(  \LapMat{\cluster} \big) \sum_{\nodeidx\in \cluster}\normgeneric{\widetilde{\weights}^{(\nodeidx)}}{2}^{2}.
\end{align}
If the bound \eqref{equ_upper_bound_err_component_cluster} would not hold, then 
by \eqref{equ_obj_function_gtvmin_average_ac_component_clustered} and \eqref{equ_norm_error_func_value_clustered} 
we would obtain $f' \big( \widehat{\weights}^{(\nodeidx)} \big) > f' \big( \overline{\weights}^{(\nodeidx)} \big)$, which 
contradicts the fact that $\widehat{\weights}^{(\nodeidx)}$ solves \eqref{equ_def_gtvmin_linreg_lec5}.  

\newpage
\section{Graph Learning for FL Networks} 
\label{lec_graphlearning} 

Chapter \ref{lec_fldesignprinciple} introduced \gls{gtvmin} as a flexible design 
principle for \gls{fl} \glspl{algorithm}. Chapter \ref{lec_flalgorithms} explores 
how \glspl{algorithm} can be obtained by applying \glspl{optmethod} - such as 
the \glspl{gdmethod} from Chapter \ref{lec_gradientmethods} - to solve \gls{gtvmin} instances.

The computational and statistical properties of such \glspl{algorithm} depend crucially on the structure of 
the underlying \gls{flnetwork}. For example, both the computational and communication costs of \gls{fl} 
systems typically increase with the number of edges in the \gls{flnetwork}. Moreover, the \gls{graph} 
topology governs how \glspl{localdataset} are pooled into \glspl{cluster} with shared \glspl{modelparam}.

In some settings, domain expertise can guide the construction of the \gls{flnetwork}. For instance, 
in health-care, known clinical similarities between disease types are used to define edges connecting 
patients or diseases \cite{NetMed}. In sensor networks, physical proximity and hardware connectivity 
constraints naturally shape the \gls{graph} structure \cite{LocalizedLinReg2019,NetworkLasso}. 
However, other applications lack strong prior structure and require to learn the \gls{graph} from 
\gls{data} \cite{JungGaphLassoSPL,CSGraphSelJournal,pmlr-v51-kalofolias16,LearningGraphsFromData2019}. 
This chapter presents techniques to infer \glspl{flnetwork} from \glspl{localdataset} and associated local \glspl{lossfunc}.

This chapter is organized as follows. Section \ref{sec_emp_graph_design_choice} discusses 
how the analysis of \gls{fl} \glspl{algorithm} can inform the design of the \gls{flnetwork}. 
Section \ref{sec_measuring_dissimarity} presents methods to quantify \glspl{discrepancy} 
between \glspl{localdataset}. Section \ref{sec_graph_learning_methods} formulates \gls{graph} 
learning as an \gls{optproblem} that minimizes the \gls{discrepancy} between datasets stored 
at nodes that are connected by an edge. The structure of the resulting \gls{graph} can be 
influenced by imposing connectivity constraints, such as a minimum required \gls{nodedegree}.

\subsection{Edges as Design Choice}
\label{sec_emp_graph_design_choice} 

Consider the \gls{gtvmin} instance \eqref{equ_def_gtvmin_linreg}, which aims to learn local 
\glspl{modelparam} for each \gls{linmodel} associated with a \gls{localdataset} $\localdataset{\nodeidx}$. 
To solve \eqref{equ_def_gtvmin_linreg}, we use Algorithm \ref{alg_fed_gd}, which implements the \gls{gradstep} 
\eqref{equ_def_basic_gradstep_lecflalg} in a message-passing fashion.

The \gls{gtvmin} formulation \eqref{equ_def_gtvmin_linreg} is defined for a fixed \gls{flnetwork} $\graph$. 
Hence, the structure of $\graph$ significantly impacts both the statistical and computational 
properties of Algorithm \ref{alg_fed_gd}.

\textbf{Statistical Properties.} These can be assessed using a \gls{probmodel} for the \glspl{localdataset}. 
An important example is the \gls{clustasspt} \eqref{equ_def_probmodel_linreg_node_i_clustered}, 
discussed in the context of \gls{cfl} in Section \ref{sec_compasp_gtvmin}. Under the \gls{cfl} assumption, 
nodes in the same \gls{cluster} should learn similar \glspl{modelparam}.

According to Proposition \ref{prop_upper_bound_clustered}, the solution to \gls{gtvmin} will be 
approximately constant across a \gls{cluster} $\cluster$ if the second smallest \gls{eigenvalue} $\eigval{2}\big( \LapMat{\cluster}\big)$ 
is large and the \gls{cluster} boundary $\bd{\cluster}$ is small. Here, $\eigval{2}\big( \LapMat{\cluster} \big)$ 
refers to the smallest nonzero \gls{eigenvalue} of the \gls{LapMat} of the induced subgraph $\indsubgraph{\graph}{\cluster}$.

Intuitively, $\eigval{2}\big( \LapMat{\cluster}\big)$ increases with the number of 
internal edges in $\cluster$. This can be made precise via Cheeger's inequality \cite[Ch.\ 21]{spielman2025sagt}. 
Alternatively, we can approximate $\indsubgraph{\graph}{\cluster}$ as a \gls{realization} 
of an \gls{ergraph}, a useful assumption especially if $\graph$ itself resembles a 
typical \gls{realization} of an \gls{ergraph}.

In an \gls{ergraph} over \( \cluster \), each pair of nodes \( \nodeidx, \nodeidx' \in \cluster \), 
with \( \nodeidx \neq \nodeidx' \), is connected independently with \gls{probability} \( p_e \). 
Formally, the presence of edges are \gls{iid} \glspl{rv} \( b^{(\nodeidx,\nodeidx')} \), one for 
each unordered pair \( \{ \nodeidx, \nodeidx' \} \subseteq \cluster \), 
indicating whether an edge exists between the two nodes. As a result, 
the presence of edges between different pairs of nodes are statistically 
independent events. 

This independence greatly simplifies analysis. For instance, the \gls{LapMat} 
$\LapMat{\rm{ER}}$ of an \gls{ergraph} can be expressed as a sum of statistically 
independent random \glspl{matrix}:
\begin{equation}
	\label{equ_decomp_L_ER_edges}
\LapMat{\rm{ER}} = \sum_{\edge{\nodeidx}{\nodeidx'}} b^{(\nodeidx,\nodeidx')} \mathbf{T}^{(\nodeidx,\nodeidx')}.
\end{equation}
This decomposition involves, for each pair of different nodes $\nodeidx,\nodeidx' \in \nodes$, the deterministic 
matrix 
$$\mathbf{T}^{(\nodeidx,\nodeidx')} = \big(\ve^{(\nodeidx)} - \ve^{(\nodeidx')}\big)  \big(\ve^{(\nodeidx)} - \ve^{(\nodeidx')}\big)^{T}.$$ 
Here, $\ve^{(\nodeidx)}$ denotes the \gls{vector} obtained from extracting the $\nodeidx$-th column of the identity matrix $\mathbf{I}_{\nrnodes}$. 
The decomposition \eqref{equ_decomp_L_ER_edges} is useful for the analysis of the \glspl{eigenvalue} of $\LapMat{\rm{ER}}$, e.g., 
via matrix concentration inequalities \cite{Tropp2015,JuPLSBMAsiloma2020}.

Interpreting a graph $\graph$ as (the \gls{realization} of) an \gls{ergraph} turns quantities such as 
\glspl{nodedegree} $\nodedegree{\nodeidx}$ and \gls{eigenvalue}s like $\eigval{2}\big( \LapMat{\cluster} \big)$ 
into (\glspl{realization} of) \glspl{rv}. The expected \gls{nodedegree} is
\[
\expect\{\nodedegree{\nodeidx}\} = p_e (|\cluster| - 1).
\]
With high probability,
\begin{equation}
	\label{equ_approx_maxnodeedgree_ER}
	\maxnodedegree^{(\graph)} \approx p_e (|\cluster| - 1).
\end{equation}
Increasing $p_e$ results in a larger expected \gls{nodedegree} and, thus, a higher connectivity 
of $\indsubgraph{\graph}{\cluster}$.

We can approximate $\eigval{2}\big( \LapMat{\cluster} \big)$ by the second smallest \gls{eigenvalue} 
of the expected \gls{LapMat}
\begin{equation}
	\nonumber
\overline{\mL} \defeq \expect\{ \LapMat{\cluster} \} = |\cluster| p_e \mathbf{I} - p_e \mathbf{1} \mathbf{1}^T.
\end{equation}
A straightforward calculation yields
\[
\eigval{2}(\overline{\mL}) = |\cluster| p_e.
\]
Thus, we arrive at the approximation
\begin{equation}
	\label{equ_approx_second_eigval_LapMat}
	\eigval{2}\big( \LapMat{\cluster} \big) \approx \eigval{2}(\overline{\mL}) = |\cluster| p_e \stackrel{\eqref{equ_approx_maxnodeedgree_ER}}{\approx} \maxnodedegree^{(\graph)}.
\end{equation}
The precise quantification of the approximation error in \eqref{equ_approx_second_eigval_LapMat} 
is beyond our scope. We refer interested readers to \cite{Tropp2015,Bollobas01} for further 
analysis of random \glspl{graph}.

{\bf Computational Properties.} 
The computational complexity of Algorithm \ref{alg_fed_gd} depends on the amount 
of computation required by a single iteration of its steps \eqref{equ_def_update_step_fedgd_sharing} 
and \eqref{equ_def_update_local}. Clearly, the \emph{per-iteration} complexity of Algorithm \ref{alg_fed_gd} 
increases with increasing node degrees $\nodedegree{\nodeidx}$. Indeed, 
step \eqref{equ_def_update_step_fedgd_sharing} requires to communicate local \glspl{modelparam} 
across each edge of the \gls{flnetwork}. This communication can be implemented using 
different physical channels, such as short-range wireless links or optical fibre connections \cite{OptFiberComm2011,10.5555/1111206}.

To summarize, using an \gls{flnetwork} with smaller $\nodedegree{\nodeidx}$ results in 
less computation and communication per iteration of Algorithm \ref{alg_fed_gd}. 
Trivially, the lowest per-iteration cost occurs when $\nodedegree{\nodeidx} = 0$, i.e., 
an empty \gls{flnetwork} with $\edges = \emptyset$. However, the overall computational cost 
also depends on the number of iterations required to approximate the \gls{gtvmin} 
solution \eqref{equ_def_gtvmin_linreg}.

According to \eqref{equ_def_gd_step_multi_factor}, the convergence speed of the 
\glspl{gradstep} \eqref{equ_gd_step_node_i} used in Algorithm \ref{alg_fed_gd} 
depends on the \gls{condnr} of the matrix $\mQ$ in \eqref{equ_def_objec_gtvmin_lec_flalg},
\[
\text{condition number} = \frac{\eigval{\nrnodes\dimlocalmodel}(\mQ)}{\eigval{1}(\mQ)}.
\]
Faster convergence is achieved when this ratio is close to one (see \eqref{equ_def_gd_step_multi_factor_opt}).

The \gls{condnr} of $\mQ$ tends to be smaller when the ratio between the 
maximum \gls{nodedegree} $\maxnodedegree^{(\graph)}$ and the second smallest 
\gls{eigenvalue} $\eigval{2}\big( \LapMat{\graph} \big)$ is small 
(see \eqref{equ_upper_bound_eigval_Q_gtvmin_lin} and \eqref{equ_lower_bound_eigval_Q_gtvmin_lin}).

Thus, for a given maximum \gls{nodedegree} $\maxnodedegree^{(\graph)}$, we should 
place the edges of an \gls{flnetwork} so that $\eigval{2}\big( \LapMat{\graph} \big)$ is 
large - leading to faster convergence of Algorithm \ref{alg_fed_gd} without increasing 
per-iteration complexity.

Spectral \gls{graph} theory also provides upper bounds on $\eigval{2}\big( \LapMat{\graph} \big)$ in terms 
of the \gls{nodedegree}s \cite{SpielSGT2012,spielman2025sagt,Fiedler1973}. These upper bounds can serve 
as a \gls{baseline} for evaluating practical constructions of the \gls{flnetwork}: If the resulting 
value $\eigval{2}\big( \LapMat{\graph} \big)$ is close to its upper bound, then further 
attempts to improve connectivity (in terms of spectral properties) are unlikely to yield significant gains.

The next result provides an example of such an upper bound.
\begin{prop}
	Consider an \gls{flnetwork} $\graph$ with $\nrnodes\!>\!1$ nodes and associated \gls{LapMat} $\LapMat{\graph}$. 
	Then, $\eigval{2}\big( \LapMat{\graph} \big)$ cannot exceed the \gls{nodedegree} $\nodedegree{\nodeidx}$ of any 
	node by more than a factor $\nrnodes/(\nrnodes\!-\!1)$. In other words,
	\begin{equation} 
		\label{equ_upperbound_fielder_via_degree}
		\eigval{2}\big( \LapMat{\graph} \big) \leq \frac{\nrnodes}{\nrnodes\!-\!1} \nodedegree{\nodeidx}, \quad \text{for every } \nodeidx\!=\!1, \ldots, \nrnodes.
	\end{equation} 
\end{prop}
\begin{proof}
	The bound \eqref{equ_upperbound_fielder_via_degree} follows from the variational 
	characterization \eqref{equ_variational_eigval_2} by evaluating the quadratic form 
	$\weights^{T} \LapMat{\graph} \weights$ for the specific vector
	\[
	\widetilde{\weights} = \sqrt{\frac{\nrnodes}{\nrnodes\!-\!1}} \bigg(\!-\!\frac{1}{\nrnodes}, \ldots, \underbrace{1\!-\!\frac{1}{\nrnodes}}_{\tilde{w}^{(\nodeidx)}}, \ldots, -\frac{1}{\nrnodes} \bigg)^T.
	\]
	This ``test'' vector is tailored to a particular node $\nodeidx \in \nodes$; its only positive entry is 
	$\tilde{\weight}^{(\nodeidx)} = 1\!-\!(1/\nrnodes)$. It satisfies $\normgeneric{\widetilde{\weights}}{} = 1$ and 
	$\widetilde{\weights}^{T} \mathbf{1} = 0$, making it a feasible vector for the optimization in \eqref{equ_variational_eigval_2}.
\end{proof}

Alternative -- and potentially tighter -- upper bounds for $\eigval{2}\big( \LapMat{\graph} \big)$ 
can be found in the \gls{graph} theory literature \cite{hoory06,spielman2025sagt,Bollobas01,ChungSpecGraphTheory}.

The per-iteration complexity of \gls{fl} algorithms increases with the \glspl{nodedegree} $\nodedegree{\nodeidx}$ 
(and thus the total number of edges) in the \gls{flnetwork} $\graph$. On the other hand, the number 
of iterations required by Algorithm \ref{alg_fed_gd} typically decreases as the second smallest 
\gls{eigenvalue} $\eigval{2}\big( \LapMat{\graph} \big)$ increases.

According to the upper bound in \eqref{equ_upperbound_fielder_via_degree}, a large value of 
$\eigval{2}\big( \LapMat{\graph} \big)$ is only possible if the \glspl{nodedegree} $\nodedegree{\nodeidx}$ 
- and hence the total number of edges - are sufficiently large. Recent work has focused on 
constructing \glspl{graph} that maximize $\eigval{2}\big( \LapMat{\graph} \big)$ given a fixed 
maximum \gls{nodedegree} $\maxnodedegree^{(\graph)} = \max_{\nodeidx \in \nodes} \nodedegree{\nodeidx}$ \cite{NEURIPS2021_74e1ed8b,ExpanderCommEffDOpt}.

Figure \ref{fig_per_iteration_complexity_conv_speed} illustrates this trade-off between 
per-iteration complexity and the number of iterations required by \gls{fl} \glspl{algorithm}.

\begin{figure}[htbp]
	\begin{center} 
		\begin{tikzpicture}[scale=1]
			\begin{axis}[
				xlabel={},
				ylabel={},
				y=20,
				x=30,
				thick,
				ymax=7,
				ymin=0,
				xmin=0,
				xmax=10,
				domain=0:7,
				ylabel near ticks,
				xlabel near ticks,
				axis lines=left, 
				xtick=\empty, 
				ytick=\empty, 
				clip=false  
				]
				\addplot[smooth, domain=1:8, samples=100,blue, mark=none] {0.4+8*exp(-0.3*x)} node[pos=0.1, anchor=south west,xshift=-14pt] {\hspace*{3mm} nr. of iterations};
				 \node at (axis cs:5,-0.8) {number of edges in $\mathcal{G}$};
				\addplot[smooth, red, mark=none] {0.7*x} node[pos=0.2, anchor=north west] {per-iteration complexity};
			\end{axis}
		\end{tikzpicture}
	\end{center} 
	\caption{Computational trade-off in \gls{gtvmin}-based methods such as Algorithm \ref{alg_fed_gd}: 
		Increasing the number of edges in the \gls{flnetwork} $\graph$ raises the per-iteration complexity, 
		but typically reduces the total number of iterations required for convergence.
		\label{fig_per_iteration_complexity_conv_speed}}
\end{figure}

\clearpage
\subsection{Measuring (Dis-)Similarity Between Datasets} 
\label{sec_measuring_dissimarity} 

The main idea behind \gls{gtvmin} is to enforce similar \glspl{modelparam} 
at two different nodes $\nodeidx$ and $\nodeidx'$ that are connected by an 
edge $\edge{\nodeidx}{\nodeidx'}$ with (relatively) large \gls{edgeweight} $\edgeweight_{\nodeidx,\nodeidx'}$. 
In general, the edges (and their weights) of the \gls{flnetwork} are design choices.  
Placing an edge between two nodes $\nodeidx, \nodeidx'$ is typically only useful 
if the \gls{localdataset}s $\localdataset{\nodeidx}, \localdataset{\nodeidx'}$ (generated by \glspl{device} $\nodeidx,\nodeidx'$) 
have similar statistical properties. We next discuss different approaches for measuring
the similarity -- or, equivalently, the \gls{discrepancy} (i.e., the lack of similarity) -- between 
two \gls{localdataset}s.

The first approach is based on a \gls{probmodel}, i.e., we interpret the \gls{localdataset} $\localdataset{\nodeidx}$ 
as \gls{realization}s of \gls{rv}s with some parametric \gls{probdist} $p^{(\nodeidx)}\big( \localdataset{\nodeidx};\localparams{\nodeidx} \big)$. 
We can then measure the \gls{discrepancy} between $\localdataset{\nodeidx}$ 
and $\localdataset{\nodeidx'}$ via the Euclidean distance $\normgeneric{\localparams{\nodeidx} - \localparams{\nodeidx'}}{2}$ 
between the parameters $\localparams{\nodeidx}$ and $\localparams{\nodeidx'}$ of the corresponding \glspl{probdist}.

In most \gls{fl} applications, the \glspl{parameter} of the \gls{probdist} $p^{(\nodeidx)}\big( \localdataset{\nodeidx};\localparams{\nodeidx} \big)$ 
underlying a \gls{localdataset} are unknown.\footnote{One exception is when the \gls{localdataset} is generated 
	by drawing \gls{iid} \glspl{realization} from $p^{(\nodeidx)}\big( \localdataset{\nodeidx};\localparams{\nodeidx} \big)$.} 
However, it is often possible to estimate these parameters using statistical inference techniques 
such as \gls{maxlikelihood} estimation \cite[Ch. 3]{MLBasics}, \cite{LC}. Given the estimates 
$\estlocalparams{\nodeidx}$ and $\estlocalparams{\nodeidx'}$ for the \glspl{modelparam}, we 
can then compute the \gls{discrepancy} measure $\discrepancy{\nodeidx}{\nodeidx'} \defeq \normgeneric{\estlocalparams{\nodeidx} - \estlocalparams{\nodeidx'}}{2}$. 

{\bf Example.} Consider \gls{localdataset}s, each consisting of a single number 
$\truelabel^{(\nodeidx)} = \weight^{(\nodeidx)} + n^{(\nodeidx)}$ with $n^{(\nodeidx)} \sim \mathcal{N}(0,1)$ 
and \gls{model} \gls{parameter} $\weight^{(\nodeidx)}$, for $\nodeidx=1,\ldots,\nrnodes$. 
The \gls{maxlikelihood} estimator for $\weight^{(\nodeidx)}$ is then given by $\hat{\weight}^{(\nodeidx)} = \truelabel^{(\nodeidx)}$ \cite{LC,kay}. 
Accordingly, the resulting \gls{discrepancy} measure is \cite{Chepuri2017}.
$$\discrepancy{\nodeidx}{\nodeidx'} \defeq \big| \truelabel^{(\nodeidx)} - \truelabel^{(\nodeidx')} \big|.$$ 

{\bf Example.} Consider an \gls{flnetwork} with nodes $\nodeidx \in \nodes$ that carry 
\gls{localdataset}s $\localdataset{\nodeidx}$. Each $\localdataset{\nodeidx}$ consists of 
\gls{datapoint}s with \glspl{label} in the \gls{labelspace} $\labelspace^{(\nodeidx)}$. 
We can measure the similarity between nodes $\nodeidx$ and $\nodeidx'$ by the fraction 
of \glspl{datapoint} in $\localdataset{\nodeidx} \bigcup \localdataset{\nodeidx'}$ with \glspl{label} lying in 
$\labelspace^{(\nodeidx)} \cap \labelspace^{(\nodeidx')}$ \cite{pFedSim23}. 

{\bf Example.} Consider \gls{localdataset}s $\localdataset{\nodeidx}$ constituted by images of handwritten 
digits $0,1,\ldots,9$. We model a \gls{localdataset} using a hierarchical \gls{probmodel}: Each node 
$\nodeidx \in \nodes$ is assigned a deterministic but unknown \gls{probdist} $\bm{\alpha}^{(\nodeidx)}= \big(\alpha^{(\nodeidx)}_{0},\ldots,\alpha^{(\nodeidx)}_{9} \big)$. The entry $\alpha^{(\nodeidx)}_{\featureidx}$ 
is the fraction of images at node $\nodeidx$ that show digit $\featureidx$. We interpret the \glspl{label} 
$\truelabel^{(\nodeidx,1)}, \ldots,  \truelabel^{(\nodeidx,\localsamplesize{\nodeidx})}$ as \glspl{realization} 
of \gls{iid} \glspl{rv}, with values in $\{0,1,\ldots,9\}$ and distributed according to $\bm{\alpha}^{(\nodeidx)}$. 
We also interpret the \glspl{feature} as \glspl{realization} of \glspl{rv} having conditional \gls{probdist} 
$p(\featurevec|\truelabel)$, which is the same for all nodes $\nodeidx \in \nodes$. We can then estimate 
the dis-similarity between nodes $\nodeidx$ and $\nodeidx'$ via the distance between (estimations of) 
the \glspl{parameter} $\bm{\alpha}^{(\nodeidx)}$ and $\bm{\alpha}^{(\nodeidx')}$.

The above examples of a \gls{discrepancy} measure -- based on parameter estimates of a \gls{probmodel} -- 
are all special cases of a more general two-step approach:
\begin{itemize}
\item First, we assign a vector representation $\vz^{(\nodeidx)} \in \mathbb{R}^{\samplesize'}$ to each node $\nodeidx \in \nodes$ \cite{MLBasics,Goodfellow-et-al-2016}.
\item Second, we define the \gls{discrepancy} $\discrepancy{\nodeidx}{\nodeidx'}$ between
nodes $\nodeidx$ and $\nodeidx'$ as the distance between the representation vectors
$\vz^{(\nodeidx)}$ and $\vz^{(\nodeidx')}$, e.g.,
$$\discrepancy{\nodeidx}{\nodeidx'} \defeq \normgeneric{\vz^{(\nodeidx)}-\vz^{(\nodeidx')}}{}.$$
\end{itemize}
We next discuss three specific implementations of the first step to obtain the representation vector
for each node $\nodeidx$.


{\bf Parametric Probabilistic Models.} 
If we use a parametric \gls{probmodel} $p\big(\localdataset{\nodeidx};\localparams{\nodeidx}\big)$ for 
the \gls{localdataset} $\localdataset{\nodeidx}$, we can use an estimator $\estlocalparams{\nodeidx}$ 
to obtain $\vz^{(\nodeidx)}$. One popular approach for estimating the \glspl{modelparam} of a \gls{probmodel} is the \gls{ml} principle \cite{MLBasics}. 

{\bf \Glspl{gradient}.} We now discuss a construction for the vector representation 
$\vz^{(\nodeidx)} \in \mathbb{R}^{\samplesize'}$ that is inspired by the update 
structure of \gls{stochGD}. In particular, we define 
the \gls{discrepancy} between two \glspl{localdataset} by treating them as two 
\glspl{batch} used by \gls{stochGD} to train a model. If these two \glspl{batch} consist of 
\glspl{datapoint} generated from similar \glspl{probdist}, their corresponding \gls{gradient} 
approximations \eqref{equ_def_loss_func_linreg_gradient_stoch} are close. 
This suggests to use the \gls{gradient} $\nabla f(\weights')$ of the average \gls{loss} (or \gls{emprisk}) 
$f(\weights) \defeq (1/|\localdataset{\nodeidx}|)\sum_{\pair{\featurevec}{\truelabel} \in \localdataset{\nodeidx}} \lossfunc{\pair{\featurevec}{\truelabel}}{\hypothesis^{(\weights)}}$ as a vector 
representation $\vz^{(\nodeidx)}$ for $\localdataset{\nodeidx}$. We can generalize 
this construction, for parametric \glspl{localmodel} $\localmodel{\nodeidx}$, 
by using the \gls{gradient} of the local \gls{lossfunc}, 
\begin{equation} 
	\label{equ_def_gradient_local_repre}
	\vz^{(\nodeidx)} \defeq \nabla \locallossfunc{\nodeidx}{\vv}.  
\end{equation} 
Note that the construction \eqref{equ_def_gradient_local_repre} requires to 
specify the \glspl{modelparam} $\vv$ at which the \gls{gradient} is evaluated. 

{\bf \Gls{featlearn}.} Another approach is to use an \gls{autoencoder} \cite[Ch.\ 14]{Goodfellow-et-al-2016} 
to learn an embedding of a \gls{localdataset}. In particular, we feed the dataset into 
an encoder \gls{ann} that has been jointly trained with a decoder \gls{ann} on a suitable 
\gls{learningtask}. The encoder maps the dataset to a latent vector, or embedding, which 
serves as its vector representation. A generic setup is illustrated in Figure \ref{fig_autoenc_feature_dataset}.
\begin{figure} 
	\tikzstyle{block} = [rectangle, draw, fill=blue!20, 
	text width=5em, text centered, rounded corners, minimum height=4em]
	\tikzstyle{trapez} = [trapezium, draw, fill=blue!20, 
	text width=5em, text centered, minimum height=4em, trapezium left angle=70, trapezium right angle=110]
	\tikzstyle{arrow} = [thick,->,>=stealth]
	\begin{center}
		\begin{tikzpicture}[node distance=1cm]
			
			\node [] (input) {$\localdataset{\nodeidx}$};
			\node [block, right=1cm of input] (encoder) {encoder $\hypothesis(\cdot)$};
			\node [right=1cm of encoder] (embedding) {$\vz^{(\nodeidx)} \in \mathbb{R}^{\samplesize'}$};
			\node [block, right=1cm of embedding] (decoder) {decoder $\hypothesis^{*}(\cdot)$};
			\node [right=1cm of decoder] (output) {$\widehat{\mathcal{D}}^{(\nodeidx)}$};
			
			\draw [arrow] (input) -- (encoder);
			\draw [arrow] (encoder) -- (embedding);
			\draw [arrow] (embedding) -- (decoder);
			\draw [arrow] (decoder) -- (output);
			
		\end{tikzpicture}
	\end{center}
	\caption{\label{fig_autoenc_feature_dataset} A generic \gls{autoencoder} consists of an 
		encoder that maps the input to a latent representation, and a decoder that attempts 
		to reconstruct the original input. Both components are trained jointly by minimizing 
		a reconstruction \gls{loss} (see \cite[Ch.\ 9]{MLBasics}). When a \gls{localdataset} is 
		used as input, its latent representation can serve as a compact vector embedding.}
\end{figure}

\subsection{Graph Learning Methods} 
\label{sec_graph_learning_methods} 

Assume we have constructed a \gls{discrepancy} measure $\discrepancy{\nodeidx}{\nodeidx'} \in \mathbb{R}_{+}$ 
that quantifies the dissimilarity between any two \glspl{localdataset} $\localdataset{\nodeidx}$ and $\localdataset{\nodeidx'}$. 
One way to construct an \gls{flnetwork} is by connecting each node $\nodeidx$ to its nearest \glspl{neighbor}, i.e., 
the nodes $\nodeidx' \in \nodes \setminus \{ \nodeidx\}$ with the smallest values of $\discrepancy{\nodeidx}{\nodeidx'}$. 

An alternative to this nearest-neighbour construction is to formulate \gls{graph} learning as 
a constrained linear \gls{optproblem}. Let us measure the quality of a candidate edge-weight 
assignment $\edgeweight_{\nodeidx,\nodeidx'} \in \mathbb{R}_{+}$ using the \gls{objfunc}
\begin{equation} 
	\label{equ_def_linobj_graphlearning}
	\sum_{\nodeidx,\nodeidx' \in \nodes} \edgeweight_{\nodeidx,\nodeidx'} \discrepancy{\nodeidx}{\nodeidx'}.
\end{equation} 
This \gls{function} penalizes large weights between nodes that are dissimilar. 
Without any constraints, the minimum of \eqref{equ_def_linobj_graphlearning} is trivially 
achieved by setting $\edgeweight_{\nodeidx,\nodeidx'} = 0$ for all pairs, i.e., resulting 
in an empty \gls{graph}.

As discussed in Section \ref{sec_emp_graph_design_choice}, however, a useful \gls{flnetwork} 
must contain a sufficient number of edges to ensure that \gls{gtvmin} produces meaningful \glspl{modelparam}. 
In particular, the pooling effect of \gls{gtvmin} depends on the second smallest \gls{eigenvalue} $\eigval{2}(\LapMat{\graph})$ 
of the \gls{LapMat} being sufficiently large, which in turn requires that the \gls{graph} is sufficiently 
well connected (see \eqref{equ_approx_second_eigval_LapMat}).

To enforce the presence of edges, we introduce the following constraints:
\begin{align}
	\label{equ_def_constraints} 
	\edgeweight_{\nodeidx,\nodeidx} &= 0, \quad	
	\sum_{\nodeidx' \neq \nodeidx} \edgeweight_{\nodeidx,\nodeidx'} = \maxnodedegree^{(\graph)}  \quad \text{ for all } \nodeidx \in \nodes, \\
	\edgeweight_{\nodeidx,\nodeidx'} & \in [0,1] \quad \text{for all } \nodeidx, \nodeidx' \in \nodes. \nonumber
\end{align} 
These constraints ensure that each node $\nodeidx$ has (weighted) \gls{nodedegree} 
$\sum_{\nodeidx' \neq \nodeidx} \edgeweight_{\nodeidx,\nodeidx'}$ equal to $\maxnodedegree^{(\graph)}$,
and that edge weights are bounded and symmetric.

Combining the \gls{objfunc} \eqref{equ_def_linobj_graphlearning} with the constraints \eqref{equ_def_constraints}, 
we arrive at the following \gls{graph} learning principle:
\begin{align} 
	\label{equ_graph_learning_linoptconstr}
	\big\{ \widehat{\edgeweight}_{\nodeidx,\nodeidx'} \big\}_{\nodeidx,\nodeidx' \in \nodes}  
	\in \argmin_{\edgeweight_{\nodeidx,\nodeidx'} = \edgeweight_{\nodeidx',\nodeidx}} \quad 
	& \sum_{\nodeidx,\nodeidx' \in \nodes} \edgeweight_{\nodeidx,\nodeidx'} \discrepancy{\nodeidx}{\nodeidx'} \\ 
	\text{s.t.} \quad 
	& \edgeweight_{\nodeidx,\nodeidx'} \in [0,1] \quad \forall \nodeidx, \nodeidx' \in \nodes, \nonumber \\ 
	& \edgeweight_{\nodeidx,\nodeidx} = 0 \quad \forall \nodeidx \in \nodes, \nonumber \\ 
	& \sum_{\nodeidx' \neq \nodeidx} \edgeweight_{\nodeidx,\nodeidx'} = \maxnodedegree^{(\graph)} 
	\quad \forall \nodeidx \in \nodes. \nonumber 
\end{align}
This constrained minimization problem is a special case of the general quadratic program 
introduced in \eqref{equ_def_min_quad_func_constr}. Because the objective is linear, 
\eqref{equ_graph_learning_linoptconstr} is equivalent to a linear program \cite[Sec.\ 4.3]{BoydConvexBook}. 
Approximate solutions to \eqref{equ_graph_learning_linoptconstr} can be efficiently 
computed using \gls{projgd}, as discussed in Section \ref{sec_proj_gradient_descent}.

The first constraint in \eqref{equ_graph_learning_linoptconstr} bounds edge weights between $0$ and $1$. 
The second prohibits self-loops, which have no effect on the outcome of \gls{gtvmin} (see \eqref{equ_def_gtvmin}). 
The final constraint enforces regularity: every node has the same \gls{nodedegree} $\nodedegree{\nodeidx} = \maxnodedegree^{(\graph)}$.
 
While regular \glspl{graph} simplify the analysis of \gls{gtvmin}, they may not always 
be desirable in practice. In some \gls{fl} applications, it may be advantageous to allow 
varying \glspl{nodedegree} -- such as \glspl{graph} with a small number of ``hub'' nodes with 
high \gls{nodedegree} \cite{Chepuri2017,NewmannBook}, or to minimize the total number of edges. 

We can enforce an upper bound on the total number $\maxnredges$ of edges by modifying the 
last constraint in \eqref{equ_graph_learning_linoptconstr}, 
\begin{align} 
	\label{equ_graph_learning_linoptconstr_max_edges}
	\widehat{\edgeweight}_{\nodeidx,\nodeidx'} & \in \argmin_{\edgeweight_{\nodeidx,\nodeidx'} =\edgeweight_{\nodeidx',\nodeidx} }  \sum_{\nodeidx,\nodeidx' \in \nodes} \edgeweight_{\nodeidx,\nodeidx'} \discrepancy{\nodeidx}{\nodeidx'}  \\ 
	& \edgeweight_{\nodeidx,\nodeidx'}  \in [0,1] \mbox{ for all } \nodeidx,\nodeidx' \in \nodes, \nonumber \\ 
	&\edgeweight_{\nodeidx,\nodeidx}  = 0 \mbox{ for all } \nodeidx \in \nodes, \nonumber \\ 
	& \sum_{\nodeidx',\nodeidx \in \nodes} \edgeweight_{\nodeidx,\nodeidx'}  = \maxnredges. \nonumber 
\end{align} 
The problem has a closed-form solution as explained in \cite{Chepuri2017}: It is obtained by placing 
the edges between those pairs $\nodeidx,\nodeidx' \in \nodes$ that result in the smallest \gls{discrepancy} $\discrepancy{\nodeidx}{\nodeidx'}$. 
However, it might still be useful to solve \eqref{equ_graph_learning_linoptconstr_max_edges} 
via iterative optimization methods such as the \glspl{gdmethod} discussed in Chapter \ref{lec_gradientmethods}. 
These methods can be implemented in a fully distributed fashion as message passing over an 
underlying communication network \cite{DistrOptStatistLearningADMM}. This communication 
network might be significantly different from the learned \gls{flnetwork}. 
For some \gls{fl} applications, the functional connectivity of two \glspl{device} $\nodeidx$ 
and $\nodeidx'$ reflects also a similarity between \glspl{probdist} of 
\glspl{localdataset} $\localdataset{\nodeidx}$ and $\localdataset{\nodeidx'}$ \cite{Ortega2018graph}.

\clearpage
\subsection{Exercises}

\refstepcounter{problem}\label{prob:orderstats}\textbf{\theproblem. A Simple Ranking Approach.}
Consider a collection of \glspl{device} $\nodeidx=1,\ldots,\nrnodes=100$, each 
carrying a \gls{localdataset} that consists of a single vector $\featurevec \in \mathbb{R}^{(\localsamplesize{\nodeidx})}$. 
We interpret the vectors $\featurevec \in \mathbb{R}^{\localsamplesize{\nodeidx}}$, for $\nodeidx=1,\ldots,\nrnodes$, 
as statistically independent \glspl{rv}. Moreover, the vector $\featurevec \in \mathbb{R}^{\localsamplesize{\nodeidx}}$ 
is a \gls{realization} of a \gls{mvndist} $\mvnormal{\clusteridx_{\nodeidx} \mathbf{1}}{\mathbf{I}}$ 
with given (fixed) quantities $\clusteridx_{\nodeidx} \in \{-1,1\}$. We construct an \gls{flnetwork} 
by determining for each node $\nodeidx$ its \gls{neighborhood} $\neighbourhood{\nodeidx}$ 
as follows
\begin{itemize}
	\item we randomly select a fraction $\mathcal{B}^{(\nodeidx)}$ of $10$ percent from all other nodes 
	\item we define $\neighbourhood{\nodeidx}$ as those $\nodeidx' \in  \mathcal{B}^{(\nodeidx)}$ whose 
	corresponding values $$|(1/\localsamplesize{\nodeidx}) \mathbf{1}^{T}\featurevec^{(\nodeidx)} - (1/\localsamplesize{\nodeidx'}) \mathbf{1}^{T}\featurevec^{(\nodeidx')}|$$ are among the $3$ smallest. 
\end{itemize} 
Analyze the \gls{probability} that some \gls{neighborhood} $\neighbourhood{\nodeidx}$ contains 
a node $\nodeidx'$ such that $\clusteridx_{\nodeidx} \neq \clusteridx_{\nodeidx'}$.

\newpage
\section{Trustworthy FL} 
\label{lec_trustworthyfl} 

This chapter examines how regulatory frameworks for \gls{trustAI} inform the design and implementation 
of \gls{gtvmin}-based methods. Our discussion is primarily guided by the key requirements for 
\gls{trustAI} as formulated by the \emph{European Union’s High-Level Expert Group on AI} \cite{HLEGTrustworthyAI}. 
Comparable ethical frameworks have emerged globally, including \emph{Australia’s AI Ethics Principles} \cite{australia_ai_ethics_2024}, 
the \emph{OECD AI Principles} \cite{oecd_ai_principles}, China’s governance efforts \cite{cac_generativeai_2023, caict_aigov_2024, most_ethics_2021}, and U.S. developments such as the \emph{NIST AI Risk Management Framework} \cite{nist_ai_rmf_2023}, 
the \emph{Blueprint for an AI Bill of Rights} \cite{ai_bill_of_rights_2022}, and \emph{Executive Order 14110 
	on the Safe, Secure, and Trustworthy Development and Use of Artificial Intelligence} \cite{exec_order_14110}.

Section \ref{sec_human_agency_oversight} examines how \gls{fl} systems can 
support human agency and oversight, as required by the principle of respect 
for human autonomy within the broader framework of \gls{trustAI}.

Section \ref{sec_techrobustness_fl} investigates the robustness of \gls{fl} systems 
against different forms of perturbations. Perturbations can arise from the intrinsic 
variability of local \gls{dataset}s that are obtained from stochastic \gls{data} generation 
processes. Another source for perturbations are imperfections of the communication 
links between \gls{device}s. We devote Chapter \ref{lec_datapoisoning} to perturbations 
that are intentional (or adversarial) during so-called cyber \glspl{attack}. 

Section \ref{sec_priv_data_governance} addresses the need for \gls{privprot} 
and \gls{data} governance. This includes regulatory constraints on data processing, 
the \gls{dataminprinc}, and the organizational structures needed to enforce compliance. 
We devote Chapter \ref{lec_privacyprotection} to a detailed treatment of quantitative 
measures for \gls{privleakage} and techniques to mitigate it in \gls{gtvmin}-based 
\gls{fl} systems.

Section \ref{sec_kr_transparancy_explain} focuses on the transparency and the 
\gls{explainability} of \gls{gtvmin}-based \gls{fl} systems. We introduce quantitative 
metrics for subjective \gls{explainability} that reflect how well personalized \gls{model}s 
align with individual users’ expectations. We can incorporate these metrics into \gls{gtvmin}-based 
methods to ensure tailored \gls{explainability} for heterogeneous populations of \gls{device} 
users.

\subsection{Human Agency and Oversight} 
\label{sec_human_agency_oversight}
``\emph{..AI systems should support human 
	autonomy and decision-making, as prescribed by the principle of respect for human autonomy. 
	This requires that AI systems should both act as enablers to a democratic, flourishing and equitable 
	society by supporting the user’s agency and foster fundamental rights, and allow for human oversight...}'' \cite[p.15]{HLEGTrustworthyAI}

{\bf Human Dignity.} Learning personalized \glspl{modelparam} for recommender systems 
allows to boost addiction or widespread emotional manipulation resulting in genocide \cite{Kuss:2016aa,Munn:2020aa,NYTMozur2018}. 
KR1 rules out certain design choices for the \gls{label}s of \gls{datapoint}s. In particular, 
we might not use the mental and psychological characteristics of a user as the \gls{label}. 
We should avoid \gls{lossfunc}s that can be used to train predictors of psychological 
characteristics. Using personalized \gls{ml} \glspl{model} to predict user preferences for products 
or susceptibility towards propaganda is also referred to as \emph{micro-targeting} \cite{Simchon:2024aa}. 

{\bf Simple is Good.} Human oversight can be facilitated by relying on simple \gls{localmodel}s. 
Examples include \gls{linmodel}s with few \gls{feature}s or \gls{decisiontree}s with a small tree 
depth. However, we are unaware of a widely accepted definition of when a model is simple. Loosely 
speaking, a simple model results in a learned \gls{hypothesis} that allows humans to understand 
how \gls{feature}s of a \gls{datapoint} relate to the \gls{prediction} $\hypothesis(\featurevec)$. 
This notion of simplicity is closely related to the concept of \gls{explainability} which we discuss 
in more detail in Section \ref{sec_kr_transparancy_explain}. 

{\bf Continuous Monitoring.} In its simplest form, \gls{gtvmin}-based methods involve a single 
training phase, i.e., learning local \glspl{modelparam} by solving \gls{gtvmin}. However, this 
approach is only useful if the data can be well approximated by an \gls{iidasspt}. In particular, 
this approach works only if the statistical properties of \glspl{localdataset} do not change over 
time. For many \gls{fl} applications, this assumption is unrealistic (consider a social network 
which is exposed to constant change of memberships and user behaviour). It is then important 
to continuously compute a \gls{valerr} which is then used, in turn, to 
diagnose the overall \gls{fl} system (see \cite[Sec. 6.6]{MLBasics}). 

\subsection{Technical Robustness and Safety}
\label{sec_techrobustness_fl}
``\emph{...Technical robustness requires that AI 
	systems be developed with a preventative approach to risks and in a manner such that they reliably behave 
	as intended while minimising unintentional and unexpected harm, and preventing unacceptable harm. ...}' \cite[p.16]{HLEGTrustworthyAI}.

Practical \gls{fl} systems are obtained by implementing \gls{fl} algorithms in physical 
distributed computers \cite{DistributedSystems,ParallelDistrBook}. One example of a 
distributed computer is a collection of smartphones that are connected either by short-range 
wireless links or by a cellular network. 

Distributed computers (as physical objects) typically incur imperfections, such as a temporary 
lack of connectivity or a mobile \glspl{device} that run out of battery and therefore become inactive. 
Moreover, the \gls{data} generation processes can be subject to perturbations such as 
statistical anomalies or outliers. Section \ref{sec_techrobustness_fl} studies in some detail 
the \gls{robustness} of \gls{gtvmin}-based systems against different perturbations of \gls{data} 
sources and imperfections of computational infrastructure. 

Consider a \gls{gtvmin}-based \gls{fl} system that trains a single (global) \gls{linmodel} 
in a distributed fashion from a collection of \gls{localdataset}s $\localdataset{\nodeidx}$, 
for $\nodeidx=1,\ldots,\nrnodes$. As discussed in Section \ref{sec_single_model_fl}, this 
single-model \gls{fl} setting uses \gls{gtvmin} \eqref{equ_def_gtvmin_linreg} 
over a connected \gls{flnetwork} with a sufficiently large choice of $\regparam$. 

To ensure {\bf KR2} we need to understand the effect of perturbations on a \gls{gtvmin}-based 
\gls{fl} system. These perturbations might be intentional (or adversarial) and affect the \gls{localdataset}s 
used to evaluate the \gls{loss} of local \glspl{modelparam} or the computational infrastructure 
used to implement a \gls{gtvmin}-based method (see Chapter \ref{lec_flalgorithms}). We 
next explain how to use some of the theoretic tools from previous chapters to quantify the 
robustness of \gls{gtvmin}-based \gls{fl} systems. 

\subsubsection{Sensitivity Analysis}
\label{sec_sensitivity_analysis} 

As pointed out in Chapter \ref{lec_fldesignprinciple}, \gls{gtvmin} \eqref{equ_def_gtvmin_linreg} 
can be rewritten as the minimization of a \gls{quadfunc}, 
\begin{equation}
	\label{equ_def_quad_form_techrobust}
	\min_{\weights\!=\!{\rm stack}\{ \localparams{\nodeidx}\}_{\nodeidx=1}^{\nrnodes}} \weights^{T} \mQ  \weights + \vq^{T} \weights.
\end{equation} 
The matrix $\mQ$ and vector $\vq$ are determined by the feature matrices $\featuremtx^{(\nodeidx)}$ 
and label vectors $\labelvec^{(\nodeidx)}$ at the nodes $\nodeidx \in \nodes$ (see \eqref{equ_def_ridge_reg_quadratidform}). 
We next study the sensitivity of (the solutions of) \eqref{equ_def_quad_form_techrobust} towards external 
perturbations of the \gls{label} vector.\footnote{Our study can be generalized to also take into account 
	perturbations of the feature matrices $\featuremtx^{(\nodeidx)}$, for $\nodeidx = 1,\ldots,\nrnodes$.} 

Consider an additive perturbation $\widetilde{\vy}^{(\nodeidx)} \defeq \labelvec^{(\nodeidx)}+ \bm{\varepsilon}^{(\nodeidx)}$ 
of the label vector $\labelvec^{(\nodeidx)}$. Using the perturbed label vector $\widetilde{\vy}^{(\nodeidx)}$ results also 
in a ``perturbation'' of \gls{gtvmin} \eqref{equ_def_quad_form_techrobust}, 
\begin{equation}
	\label{equ_def_quad_form_techrobust_perturb}
	\min_{\weights = {\rm stack}\{ \localparams{\nodeidx}\}} \weights^{T} \mQ  \weights + \vq^{T} \weights+ \vn^{T} \weights+ c.
\end{equation} 
An inspection of \eqref{equ_def_ridge_reg_quadratidform} yields that $\vn = \bigg( \big( \bm{\varepsilon}^{(1)} \big)^{T} \featuremtx^{(1)}, \ldots, 
\big( \bm{\varepsilon}^{(\nrnodes)} \big)^{T} \featuremtx^{(\nrnodes)} \bigg)^{T}$. 
The next result provides an upper bound on the deviation between the solutions of 
\eqref{equ_def_quad_form_techrobust} and \eqref{equ_def_quad_form_techrobust_perturb}. 
\begin{prop}
	\label{prop_sensitiveiy_gtmvin_linreg}
	Consider the \gls{gtvmin} instance \eqref{equ_def_quad_form_techrobust} for learning local \glspl{modelparam} 
	of a \gls{linmodel} for each node $\nodeidx \in \nodes$ of an \gls{flnetwork} $\graph$. We assume 
	that the \gls{flnetwork} is connected, i.e., $\eigval{2}\big(\LapMat{\graph}\big) > 0$ and the \glspl{localdataset} 
	are such that $\avgmineigvallocalQ > 0$ (see \eqref{equ_summary_eigvals_Q_i}). 
	Then, the deviation between the solution $\estlocalparams{\nodeidx}$ to \eqref{equ_def_quad_form_techrobust} 
	and the solution $\widetilde{\weights}^{(\nodeidx)}$ to the perturbed problem \eqref{equ_def_quad_form_techrobust_perturb} is upper bounded as 
	\begin{equation} 
		\nonumber
		\sum_{\nodeidx=1}^{\nrnodes}	\normgeneric{\estlocalparams{\nodeidx}-\widetilde{\weights}^{(\nodeidx)}}{2}^{2} \leq    \frac{\maxeigvallocalQ(1+\rho^{2})^2}{ \big[ {\rm min} \{ \eigval{2}\big( \LapMat{\graph} \big)  \regparam \rho^{2}, \avgmineigvallocalQ/2 \}\big]^{2}} \sum_{\nodeidx=1}^{\nrnodes}  
		\normgeneric{\bm{\varepsilon}^{(\nrnodes)}}{2}^{2}.
	\end{equation} 
	Here, we used the shorthand $\rho \defeq \avgmineigvallocalQ/(4 \maxeigvallocalQ)$ (see \eqref{equ_summary_eigvals_Q_i}). 
\end{prop} 
\begin{proof}
	The assumptions of Proposition \ref{prop_sensitiveiy_gtmvin_linreg} allow to apply the lower 
	bound \eqref{equ_lower_bound_eigval_Q_gtvmin_lin} on the \gls{eigenvalue}s of the matrix $\mQ$ 
	in \eqref{equ_def_quad_form_techrobust}. 
\end{proof} 

\subsubsection{Estimation Error Analysis}
Proposition \ref{prop_sensitiveiy_gtmvin_linreg} characterizes the sensitivity of \gls{gtvmin} 
solutions against \emph{external} perturbations of the \glspl{localdataset}. While this 
notion of \gls{robustness} is important, it might not suffice for a comprehensive assessment 
of an \gls{fl} system. For example, we can trivially achieve perfect \gls{robustness} by delivering 
constant \glspl{modelparam}, e.g., $\estlocalparams{\nodeidx}=\mathbf{0}$. Clearly, such a 
\gls{fl} system is not very useful. 

Another form of \gls{robustness} is to ensure a small \gls{esterr} of \eqref{equ_def_gtvmin_linreg}. 
To study this form of \gls{robustness}, we use a variant of the \gls{probmodel} \eqref{equ_def_probmodel_linreg_node_i}: 
We assume that the \glspl{label} and \glspl{feature} of \glspl{datapoint} of each \gls{localdataset} $\localdataset{\nodeidx}$, for $\nodeidx=1,\ldots,\nrnodes$, are related via 
\begin{equation}
	\label{equ_def_local_lin_model_perturb}
	\labelvec^{(\nodeidx)} = \featuremtx^{(\nodeidx)} \overline{\weights} + \bm{\varepsilon}^{(\nodeidx)}.  
\end{equation} 
In contrast to Section \ref{sec_statasp_gtvmin}, we assume that all components of \eqref{equ_def_local_lin_model_perturb} 
are deterministic. In particular, the noise term $\bm{\varepsilon}^{(\nodeidx)}$ is a 
deterministic but unknown quantity. This term accommodates any perturbation that 
might arise from technical imperfections or intrinsic \gls{label} noise due to random 
fluctuations in the labelling process.\footnote{Consider \glspl{label} obtained from 
	physical sensing \glspl{device} which are typically subject to measurement errors \cite{alma996329383502466}.}

In the ideal case of no perturbation, we would have $ \bm{\varepsilon}^{(\nodeidx)} = \mathbf{0}$. 
However, in general might only know some upper bound measure for the size of the perturbation, e.g., $\normgeneric{\bm{\varepsilon}^{(\nodeidx)}}{2}^{2}$. 
We next present upper bounds on the estimation error $\estlocalparams{\nodeidx} - \overline{\weights}$ incurred 
by the \gls{gtvmin} solutions $\estlocalparams{\nodeidx}$. 

This estimation error consists of two components, the first component being ${\rm avg} \big\{ \estlocalparams{\nodeidx'} \big\} - \overline{\weights}$ for each node $\nodeidx \in \nodes$. Note that this error component is identical for all 
nodes $\nodeidx \in \nodes$. The second component of the estimation error is the deviation 
$\widetilde{\weights}^{(\nodeidx)} \defeq \widehat{\weights}^{(\nodeidx)} -  {\rm avg} \big\{ \estlocalparams{\nodeidx'} \big\}$ 
of the learned local \glspl{modelparam} $\estlocalparams{\nodeidx'}$, for $\nodeidx' = 1,\ldots,\nrnodes$, 
from their average ${\rm avg} \big\{ \estlocalparams{\nodeidx'} \big\} = (1/\nrnodes)\sum_{\nodeidx'=1}^{\nrnodes} \estlocalparams{\nodeidx'}$. As discussed in Section \ref{sec_statasp_gtvmin}, these two components 
correspond to two orthogonal subspaces of $\mathbb{R}^{\dimlocalmodel  \nrnodes}$. 

According to Proposition \ref{prop_second_component_error}, the second error component is upper bounded as 
\begin{equation} 
	\label{equ_upper_bound_variation_component_techrob}
	\sum_{\nodeidx=1}^{\nrnodes} \normgeneric{	\widetilde{\weights}^{(\nodeidx)}  }{2}^{2} \leq \frac{1}{\eigval{2}\regparam} \sum_{\nodeidx=1}^{\nrnodes} (1/\localsamplesize{\nodeidx}) \normgeneric{{\bm \varepsilon}^{(\nodeidx)}}{2}^{2}.
\end{equation}
To bound the first error component $\bar{\vc}- \overline{\weights}$, using the shorthand $ \bar{\vc} \defeq {\rm avg} \big\{ \estlocalparams{\nodeidx} \big\}$, 
we first note that (see \eqref{equ_def_gtvmin_linreg}) 
\begin{equation}
	\label{equ_def_min_bar_c_component_techrobustfl} 
	\hspace*{-3mm}\bar{\vc} \!=\! \argmin_{\weights \in \mathbb{R}^{\dimlocalmodel}} \sum_{\nodeidx \in \nodes} \hspace*{-1mm}(1/\localsamplesize{\nodeidx}) \normgeneric{\labelvec^{(\nodeidx)}
		\!-\! \featuremtx^{(\nodeidx)} \big(\weights\!-\!\widetilde{\weights}^{(\nodeidx)}\big)}{2}^{2}\!+\!\regparam \hspace*{-3mm}\sum_{\edge{\nodeidx}{\nodeidx'} \in \edges} \hspace*{-3mm} \edgeweight_{\nodeidx,\nodeidx'} \normgeneric{\widetilde{\weights}^{(\nodeidx)}\!-\!\widetilde{\weights}^{(\nodeidx')}}{2}^{2}.  
\end{equation} 
Using a similar argument as in the proof for Proposition \ref{prop_bound_linreg_error}, we obtain 
\begin{equation} 
	\label{equ_upper_bound_first_step_techrobust}
	\normgeneric{\bar{\vc} - \overline{\weights}}{2}^{2} \leq \normgeneric{\sum_{\nodeidx=1}^{\nrnodes} (1/\localsamplesize{\nodeidx}) \big( \featuremtx^{(\nodeidx)} \big)^{T} \big( \bm{\varepsilon}^{(\nodeidx)} + \featuremtx^{(\nodeidx)} \widetilde{\weights}^{(\nodeidx)} \big) }{2}^{2}/(\nrnodes \avgmineigvallocalQ)^2. 
\end{equation}  
Here, $\avgmineigvallocalQ$ is the smallest \gls{eigenvalue} of 
$(1/\nrnodes) \sum_{\nodeidx=1}^{\nrnodes} \mQ^{(\nodeidx)}$, i.e., 
the average of the matrices $\mQ^{(\nodeidx)} = (1/\localsamplesize{\nodeidx}) \big( \featuremtx^{(\nodeidx)}\big)^{T} \featuremtx^{(\nodeidx)}$ 
over all nodes $\nodeidx \in \nodes$.\footnote{We encountered the quantity $\avgmineigvallocalQ$ already during our  
	discussion of \glspl{gdmethod} for solving the \gls{gtvmin} instance \eqref{equ_def_gtvmin_linreg} (see \eqref{equ_summary_eigvals_Q_i}).} 
Note that the bound \eqref{equ_upper_bound_first_step_techrobust} is only valid if $ \avgmineigvallocalQ > 0$ 
which, in turn, implies that the solution to \eqref{equ_def_min_bar_c_component_techrobustfl} is unique. 

We can develop \eqref{equ_upper_bound_first_step_techrobust} further using 
\begin{align} 
	\label{equ_def_upper_bound_norm_squared_techrob}
	& \normgeneric{\sum_{\nodeidx=1}^{\nrnodes} (1/\localsamplesize{\nodeidx}) \big( \featuremtx^{(\nodeidx)} \big)^{T} \big( \bm{\varepsilon}^{(\nodeidx)} + \featuremtx^{(\nodeidx)} \widetilde{\weights}^{(\nodeidx)} \big) }{2}  \nonumber \\ 
	&\hspace*{-5mm} \stackrel{(a)}{\leq}  \sum_{\nodeidx=1}^{\nrnodes}  \normgeneric{ (1/\localsamplesize{\nodeidx}) \big( \featuremtx^{(\nodeidx)} \big)^{T} \big( \bm{\varepsilon}^{(\nodeidx)} + \featuremtx^{(\nodeidx)} \widetilde{\weights}^{(\nodeidx)} \big) }{2} \nonumber \\ 
	&\hspace*{-5mm} \stackrel{(b)}{\leq} \sqrt{\nrnodes} \sqrt{\sum_{\nodeidx=1}^{\nrnodes}  \normgeneric{ (1/\localsamplesize{\nodeidx}) \big(\featuremtx^{(\nodeidx)} \big)^{T} \big( \bm{\varepsilon}^{(\nodeidx)} + \featuremtx^{(\nodeidx)} \widetilde{\weights}^{(\nodeidx)} \big) }{2}^{2}}  \nonumber \\ 
	&\hspace*{-5mm} \stackrel{(c)}{\leq} \sqrt{\nrnodes} \sqrt{\sum_{\nodeidx=1}^{\nrnodes}  2 \normgeneric{ (1/\localsamplesize{\nodeidx}) \big(\featuremtx^{(\nodeidx)} \big)^{T} \bm{\varepsilon}^{(\nodeidx)} }{2}^{2}  + 2\normgeneric{ (1/\localsamplesize{\nodeidx}) \big(\featuremtx^{(\nodeidx)} \big)^{T} \featuremtx^{(\nodeidx)} \widetilde{\weights}^{(\nodeidx)}  }{2}^{2}}
	\nonumber \\ 
	&\hspace*{-5mm} \stackrel{(d)}{\leq} \sqrt{\nrnodes} \sqrt{\sum_{\nodeidx=1}^{\nrnodes}  (2/\localsamplesize{\nodeidx}) \maxeigvallocalQ \normgeneric{ \bm{\varepsilon}^{(\nodeidx)} }{2}^{2} \!+\! 2 \maxeigvallocalQ^2 \normgeneric{ \widetilde{\weights}^{(\nodeidx)}  }{2}^{2}}.
\end{align} 
Here, step $(a)$ uses the triangle inequality of norms, step $(b)$ 
uses the Cauchy-Schwarz inequality, step $(c)$ uses the inequality $\normgeneric{\va + \vb}{2}^{2} \leq 2 \bigg(\normgeneric{\va}{2}^{2} + \normgeneric{\vb}{2}^{2} \bigg)$, and step $(d)$ uses the maximum \gls{eigenvalue} $\maxeigvallocalQ \defeq \max_{\nodeidx \in \nodes} \eigval{\dimlocalmodel}\big( \mQ^{(\nodeidx)}\big)$ of 
the matrices $\mQ^{(\nodeidx)} = (1/\localsamplesize{\nodeidx}) \big( \featuremtx^{(\nodeidx)}\big)^{T} \featuremtx^{(\nodeidx)}$ (see \eqref{equ_summary_eigvals_Q_i}). 

Inserting \eqref{equ_def_upper_bound_norm_squared_techrob} into \eqref{equ_upper_bound_first_step_techrobust} 
results in the upper bound 
\begin{align} 
	\label{equ_upper_bound_first_step_techrobust_next_step}
	\normgeneric{\bar{\vc} - \overline{\weights}}{2}^{2} & \leq 
	2 \sum_{\nodeidx=1}^{\nrnodes} \bigg[  (1/\localsamplesize{\nodeidx}) \maxeigvallocalQ \normgeneric{ \bm{\varepsilon}^{(\nodeidx)} }{2}^{2}  +  \maxeigvallocalQ^2 \normgeneric{ \widetilde{\weights}^{(\nodeidx)}  }{2}^{2} \bigg]/(\nrnodes \avgmineigvallocalQ^2) \nonumber \\ 
	& \stackrel{\eqref{equ_upper_bound_variation_component_techrob}}{\leq} 2\big(\maxeigvallocalQ+(\maxeigvallocalQ^2/(\eigval{2}\regparam))\big) \sum_{\nodeidx=1}^{\nrnodes}(1/\localsamplesize{\nodeidx})  \normgeneric{ \bm{\varepsilon}^{(\nodeidx)} }{2}^{2} /(\nrnodes \avgmineigvallocalQ^2).
\end{align}  
The upper bound \eqref{equ_upper_bound_first_step_techrobust_next_step} on the \gls{esterr} 
of \gls{gtvmin}-based methods depends on both, the \gls{flnetwork} $\graph$ via the \gls{eigenvalue} 
$\eigval{2}$ of $\LapMat{\graph}$, and the \gls{feature} matrices $\featuremtx^{(\nodeidx)}$ of 
the \gls{localdataset}s (via the quantities $\maxeigvallocalQ$ and $ \avgmineigvallocalQ$ as defined 
in \eqref{equ_summary_eigvals_Q_i}). Let us next discuss how the upper bound \eqref{equ_upper_bound_first_step_techrobust_next_step} 
might guide the choice of the \gls{flnetwork} $\graph$ and the \gls{feature}s of \gls{datapoint}s in the \gls{localdataset}s. 

According to \eqref{equ_upper_bound_first_step_techrobust_next_step}, we should use an \gls{flnetwork} $\graph$ 
with large $\eigval{2}\big( \LapMat{\graph} \big)$ to ensure a small \gls{esterr} for \gls{gtvmin}-based methods. 
Note that we came across the same design criterion already when discussing graph learning methods 
in Chapter \ref{lec_graphlearning}. In particular, using an \gls{flnetwork} with large $\eigval{2}\big( \LapMat{\graph} \big)$ 
also tends to speed up the convergence of \glspl{gdmethod} for solving \gls{gtvmin} (such as Algorithm \ref{alg_fed_gd}). 

The upper bound \eqref{equ_upper_bound_first_step_techrobust_next_step} suggests using \gls{feature}s 
that result in a small ratio $\maxeigvallocalQ/\avgmineigvallocalQ$ between the quantities $\maxeigvallocalQ$ 
and $\avgmineigvallocalQ$ (see \eqref{equ_summary_eigvals_Q_i}). Some \gls{feature} learning 
methods have been proposed in order to minimize this ratio \cite{MLBasics,JungFixedPoint}. 

\subsubsection{Robustness of \gls{fl} \Gls{algorithm}s} 
\label{sec_robust_proximal_step}

The previous sub-sections studied the robustness of \gls{gtvmin} solutions against 
perturbations of \gls{localdataset}s. Ensuring trustworthy \gls{fl} systems also requires 
robustness of \gls{fl} \gls{algorithm}s against perturbations of their executions. It turns 
out that our design choices (e.g., the shape of local \gls{lossfunc}s) for \gls{gtvmin} 
crucially affect the robustness of the \gls{fl} \gls{algorithm}s discussed in Section \ref{lec_flalgorithms}. 

For ease of exposition, we will focus on \gls{fl} \gls{algorithm}s for parametric \gls{localmodel}s that 
are based on the update
\begin{equation} 
	\label{equ_update_robustness_fl_algos}
	\localparamsiter{\nodeidx}{\iteridx+1} \in \argmin_{\localparams{\nodeidx} \in \mathbb{R}^{\dimlocalmodel}} 
	\left[ \locallossfunc{\nodeidx}{\localparams{\nodeidx}} \!+\! \sum_{\nodeidx' \in \neighbourhood{\nodeidx}} \edgeweight_{\nodeidx,\nodeidx'} \gtvpenalty\big( \localparamsiter{\nodeidx'}{\iteridx} - \localparamsiter{\nodeidx}{\iteridx}\big) \right].
\end{equation}
Note that Algorithm \ref{alg_fed_gd_general} and Algorithm \ref{alg_fed_relax} use \eqref{equ_update_robustness_fl_algos} as 
their core computational step. We next discuss the robustness of \eqref{equ_update_robustness_fl_algos} 
against perturbations of the \glspl{modelparam} $\localparamsiter{\nodeidx'}{\iteridx}$ that \gls{device} 
receives from its \glspl{neighbor} $\nodeidx' \in \neighbourhood{\nodeidx}$. We focus on two specific 
choices for the \gls{gtv} penalty function $\gtvpenalty$. 

{\bf \gls{gtv} penalty $\gtvpenalty(\cdot)=\normgeneric{\cdot}{2}^2$.} 
For the penalty function $\gtvpenalty(\localparams{\nodeidx}\!-\!\localparams{\nodeidx'}) = \normgeneric{\localparams{\nodeidx}\!- \!\localparams{\nodeidx'}}{2}^{2}$, we can rewrite \eqref{equ_update_robustness_fl_algos} 
as (see Exercise \ref{ex_fedrelaxrewritten})
\begin{align}
	\label{equ_robust_update_for_squared_norm}
	\localparamsiter{\nodeidx}{\iteridx+1}  \in \argmin_{\localparams{\nodeidx} \in \mathbb{R}^{\dimlocalmodel}} \locallossfunc{\nodeidx}{\localparams{\nodeidx}} + \regparam \nodedegree{\nodeidx}
	\normgeneric{\localparams{\nodeidx}-\estlocalparams{\neighbourhood{\nodeidx}}}{2}^{2}.
\end{align} 
Here, we used $\estlocalparams{\neighbourhood{\nodeidx}}\defeq (1/\nodedegree{\nodeidx}) \sum_{\nodeidx' \in \neighbourhood{\nodeidx}} \edgeweight_{\nodeidx,\nodeidx'} \localparamsiter{\nodeidx'}{\iteridx}$ and the weighted 
\gls{nodedegree} $\nodedegree{\nodeidx} = \sum_{\nodeidx' \in \neighbourhood{\nodeidx}} \edgeweight_{\nodeidx,\nodeidx'}$ 
(see \eqref{equ_def_node_degree}). 

If the local \gls{lossfunc} $\locallossfunc{\nodeidx}{\cdot}$ is \gls{convex}, and under some mild technical 
conditions,\footnote{Strictly speaking, we need to require \gls{lossfunc} $\locallossfunc{\nodeidx}{\cdot}$ 
to have a non-empty and closed epigraph which does not contain any non-horizontal lines \cite{ProximalMethods}.} 
the update \eqref{equ_robust_update_for_squared_norm} is well-defined, i.e., the minimization  
has a unique solution \cite[Ch.\ 6]{BeckFOMBook}. Moreover, the update \eqref{equ_robust_update_for_squared_norm} 
then coincides with an application of the \gls{proxop} $\proximityop{\locallossfunc{\nodeidx}{\cdot}}{\cdot}{\rho}$ 
(see \eqref{equ_def_prox_op_original}) of $\locallossfunc{\nodeidx}{\cdot}$ \cite{ProximalMethods}, 
\begin{equation}
	\label{equ_update_prox_op_convx_function}
	\localparamsiter{\nodeidx}{\iteridx+1} = \proximityop{\locallossfunc{\nodeidx}{\cdot}}{\estlocalparams{\neighbourhood{\nodeidx}}}{\rho} \mbox{  with } \rho =2 \regparam \nodedegree{\nodeidx}. 
\end{equation} 

Figure \ref{fig_update_robust_proxop} illustrates the update \eqref{equ_update_prox_op_convx_function} 
as a straight line. The slope of this line indicates the robustness of \eqref{equ_update_prox_op_convx_function} 
against perturbations of the received \glspl{modelparam} $\localparamsiter{\nodeidx'}{\iteridx}$, 
for $\nodeidx' \in \neighbourhood{\nodeidx}$. These perturbations result in a modified input 
$\widetilde{\weights}^{(\nodeidx)}$ (instead of $\estlocalparams{\neighbourhood{\nodeidx}}$) 
for the \gls{proxop} $\proximityop{\locallossfunc{\nodeidx}{\cdot}}{\estlocalparams{\neighbourhood{\nodeidx}}}{\rho}$.
A natural quantitative measure for the robustness (or stability) of \eqref{equ_update_prox_op_convx_function} 
is 
\begin{equation}
	\label{equ_def_robust_contract_proxop}
\frac{\normgeneric{\proximityop{\locallossfunc{\nodeidx}{\cdot}}{\widetilde{\weights}^{(\nodeidx)}}{\rho} -\proximityop{\locallossfunc{\nodeidx}{\cdot}}{\estlocalparams{\neighbourhood{\nodeidx}}}{\rho}}{2}}{\normgeneric{\widetilde{\weights}^{(\nodeidx)} - \estlocalparams{\neighbourhood{\nodeidx}}}{2}}.
\end{equation}

It turns out that if the local \gls{lossfunc} $\locallossfunc{\nodeidx}{\cdot}$ is \gls{strcvx} 
with coefficient $\sigma$, then \eqref{equ_def_robust_contract_proxop} is 
upper bounded by \cite[Sec.\ 6]{ryu2016primer}
\begin{equation} 
	\label{equ_robust_measure_sq_eucl_norm_update}
	\frac{1}{1+(\sigma/\rho)}=\frac{1}{1+(\sigma/(2 \regparam \nodedegree{\nodeidx}))}.
\end{equation} 
We can interpret the quantity \eqref{equ_robust_measure_sq_eucl_norm_update} as a 
measure for the robustness of the update \eqref{equ_robust_update_for_squared_norm}. 
The smaller this quantity, the more robust are \gls{fl} systems based on \eqref{equ_robust_update_for_squared_norm}. 

Note how the robustness measure \eqref{equ_robust_measure_sq_eucl_norm_update} 
can guide the design choices for the components of \gls{gtvmin}. In particular, to ensure a 
small value \eqref{equ_robust_measure_sq_eucl_norm_update} (ensuring robustness), 
we should use  
\begin{itemize}
\item a local \gls{lossfunc} that is \gls{strcvx} with a large coefficient $\sigma$,
\item a \gls{flnetwork} with small \gls{nodedegree}s $\nodedegree{\nodeidx}$,
\item a small value $\regparam$ for \gls{gtvmin} parameter. 
\end{itemize}

\begin{figure}[H]
\begin{center}
\begin{tikzpicture}[scale=1.0]
	\def\slope{0.6}
	\def\offset{0.5}
	
	\draw[->] (-2.5, 0) -- (4.5, 0) node[right] {$\localparams{\nodeidx}$};
	\draw[->] (0, -0.5) -- (0, 4.5) node[above] {$\localparamsiter{\nodeidx}{\iteridx+1}$};
	
	\draw[thick, domain=-2:4, samples=2] plot (\x, {\slope * \x + \offset}) node[above right] {$ \proximityop{\locallossfunc{\nodeidx}{\cdot}}{\localparams{\nodeidx}}{\rho}$};
	
	\def\xA{2.0}
	\def\xB{3.5}
	\def\yA{\slope*\xA + \offset}
	\def\yB{\slope*\xB + \offset}
	
	\filldraw[black] (\xA, \yA) circle (1pt);
	\draw[dashed] (\xA, 0) -- (\xA, \yA);
	\draw[dashed] (0, \yA) -- (\xA, \yA);
	\node[below] at (\xA, 0) {$\estlocalparams{\neighbourhood{\nodeidx}}$};
	\node[left] at (0, \yA) {$$};
	
	\filldraw[black] (\xB, \yB) circle (1pt);
	\draw[dashed] (\xB, 0) -- (\xB, \yB);
	\draw[dashed] (0, \yB) -- (\xB, \yB);
	\node[below] at (\xB, 0) {$\widetilde{\weights}^{(\nodeidx)}$};
	\node[left] at (0, \yB) {$$};
	
\end{tikzpicture}
\end{center}
\caption{For a \gls{convex} local \gls{lossfunc} $\locallossfunc{\nodeidx}{\cdot}$, the update \eqref{equ_robust_update_for_squared_norm} 
	becomes the evaluation of the \gls{proxop} $\proximityop{\locallossfunc{\nodeidx}{\cdot}}{\cdot}{\rho}$ with $\rho=2 \regparam \nodedegree{\nodeidx}$. We can measure the robustness of \eqref{equ_robust_update_for_squared_norm} 
	by the slope of $\proximityop{\locallossfunc{\nodeidx}{\cdot}}{\cdot}{\rho}$ (see \eqref{equ_def_robust_contract_proxop}). \label{fig_update_robust_proxop}}
\end{figure}

{\bf \gls{gtv} penalty $\gtvpenalty(\cdot)=\normgeneric{\cdot}{2}$.} 
Let us now study \eqref{equ_update_robustness_fl_algos} for the centre node $\nodeidx=1$ 
of a star-shaped \gls{flnetwork} (see Figure \ref{fig:star_graph}). This uses the 
trivial local \gls{lossfunc} $\locallossfunc{\nodeidx}{\cdot}\equiv0$ and is connected via unit-weight 
edges to the peripheral nodes $\nodeidx'=2,\ldots,\nrnodes$. The variation 
of local \glspl{modelparam} is measured with the penalty function $\gtvpenalty(\localparams{\nodeidx} - \localparams{\nodeidx'}) = \normgeneric{\localparams{\nodeidx} - \localparams{\nodeidx'}}{2}$. 
This special case of \eqref{equ_update_robustness_fl_algos} can be written as 
\begin{equation}
	\label{equ_update_geometric_median}
\localparamsiter{1}{\iteridx+1} \in \argmin_{\localparams{\nodeidx} \in \mathbb{R}^{\dimlocalmodel}} 
\sum_{\nodeidx'=2}^{\nrnodes} \normgeneric{\localparamsiter{\nodeidx'}{\iteridx} -\localparams{\nodeidx}}{2}.
\end{equation}
Note that \eqref{equ_update_geometric_median} is nothing but the geometric median of the 
\glspl{modelparam} $\localparamsiter{\nodeidx'}{\iteridx}$, for $\nodeidx' \in \neighbourhood{\nodeidx}$. 
The usefulness of the geometric median for robust \gls{fl} has been studied 
recently \cite{RobustAgg2022}. 

The update \eqref{equ_update_geometric_median} defines a \gls{nonsmooth} \gls{convex} optimization 
problem. Any solution $\localparamsiter{1}{\iteridx+1}$ to this problem must satisfy the \gls{subgradient} 
optimality condition
\begin{align}
\label{equ_opt_gm}
\sum_{\nodeidx'=2}^{\nrnodes} \vg^{(\nodeidx')} & = \mathbf{0} & \mbox{, with } \vg^{(\nodeidx')} = \begin{cases}\frac{\localparamsiter{\nodeidx'}{\iteridx} - \localparamsiter{1}{\iteridx+1}}{\normgeneric{\localparamsiter{\nodeidx'}{\iteridx} - \localparamsiter{1}{\iteridx+1}}{2}} & \mbox{ if }  \localparamsiter{\nodeidx'}{\iteridx} \neq \localparamsiter{1}{\iteridx+1} \\ \vu \in \mathcal{B}(1) & \mbox{ otherwise,}  \end{cases}
\end{align}  
where $\mathcal{B}(1) \defeq \{ \vu \in \mathbb{R}^{\dimlocalmodel} : \normgeneric{\vu}{2} \leq 1 \}$ 
denotes the unit Euclidean ball. Each $\vg^{(\nodeidx')}$ is a \gls{subgradient} of the \gls{convex}  
\gls{nonsmooth} function $f(\localparams{\nodeidx}) \defeq \normgeneric{\localparamsiter{\nodeidx'}{\iteridx} - \localparams{\nodeidx}}{2}$.

Figure \ref{fig_opt_gm} illustrates the optimality condition \eqref{equ_opt_gm} for the case where 
node $\nodeidx = 1$ has three \glspl{neighbor}, two of which are trustworthy. The third neighbour 
is not trustworthy and may send arbitrarily corrupted \glspl{modelparam}. Despite such adversarial 
perturbations, the solution $\localparamsiter{1}{\iteridx+1}$ of \eqref{equ_opt_gm} 
cannot be arbitrarily far from the \glspl{modelparam} of the trustworthy \glspl{neighbor}, 
provided they form the majority. 

Intuitively, if the solution were far from the honest models, then the corresponding \gls{subgradient}s 
$\vg^{(\nodeidx')}$ for the trustworthy \glspl{neighbor} $\nodeidx'\in \neighbourhood{\nodeidx}$ 
would point in nearly the same direction, and their sum would have a norm close to the 
number of honest \glspl{neighbor}. However, the \glspl{subgradient} from the non-trustworthy 
nodes—being unit vectors—cannot cancel this sum unless they are sufficiently numerous, 
which contradicts the majority assumption. For a more detailed \gls{robustness} analysis of \eqref{equ_opt_gm}, 
we refer to \cite[Thm. 2.2]{Lopuhaae1991}.


\begin{figure} 
\begin{tikzpicture}[scale=2, thick, >=stealth]
	\coordinate (w) at (3,0);
	\fill (w) circle (1.2pt) node[below right] {$\localparamsiter{1}{\iteridx+1}$};
	
	\coordinate (w2) at (0.5,0.3);
	\coordinate (w3) at (0.7,0.7);
	\fill (w2) circle (1pt) node[above left] {$\localparamsiter{2}{\iteridx}$};
	\fill (w3) circle (1pt) node[above left] {$\localparamsiter{3}{\iteridx}$};
	\draw[dashed] (w) -- (w2);
	\draw[dashed] (w) -- (w3);
	\draw[->, thick, red] (w) -- ($(w)!1cm!(w2)$) ;
	\draw[->, thick, red] (w) -- ($(w)!1cm!(w3)$) node[pos=0.9, right,yshift=7pt] {$\frac{\localparamsiter{3}{\iteridx}- \localparamsiter{1}{\iteridx+1}}{\normgeneric{\localparamsiter{3}{\iteridx}- \localparamsiter{1}{\iteridx+1}}{2}}$};

	\node at (-0.0,1.4) {\textbf{trustworthy}};
	
	
	\coordinate (w4) at (5,0.2);
	\node at (5,0.7) {\textbf{perturbed}};
	\fill (w4) circle (1pt) node[below left] {$\localparamsiter{4}{\iteridx}$};
		\draw[->, thick, red] (w) -- ($(w)!1cm!(w4)$) ;
\end{tikzpicture}
\caption{Illustration of the (zero-\gls{subgradient}) optimality condition \eqref{equ_opt_gm} for the 
	update \eqref{equ_update_geometric_median}. The arrows represent unit-norm 
	\glspl{subgradient} arising from the components $\normgeneric{\localparamsiter{\nodeidx'}{\iteridx} -\localparams{\nodeidx}}{2}$ 
	for $\nodeidx'=2,\ldots,\nrnodes$.
	\label{fig_opt_gm}}
\end{figure} 

\clearpage
\subsubsection{Network Resilience}

The previous sections studied the robustness of \gls{gtvmin}-based methods against 
perturbations of \gls{localdataset}s (see Exercise \ref{prop_sensitiveiy_gtmvin_linreg}) 
and in terms of ensuring a small \gls{esterr} (see \eqref{equ_upper_bound_first_step_techrobust_next_step}). 
We also need to ensure that \gls{fl} systems are robust against imperfections of the 
computational infrastructure used to solve \gls{gtvmin}. These imperfections include 
hardware failures, running out of battery or lack of wireless connectivity. 

Chapter \ref{lec_flalgorithms} showed how to design \gls{fl} algorithms by applying \glspl{gdmethod} 
to solve \gls{gtvmin} \eqref{equ_def_gtvmin_linreg}. We obtain practical \gls{fl} systems by implementing 
these algorithms, such as Algorithm \ref{alg_fed_gd}, in a particular computational infrastructure. Two 
important examples of such an infrastructure are mobile networks and wireless sensor networks \cite{FundWireless,WSNMagazine}. 

The effect of imperfections in the implementation of the \gls{gd} based Algorithm \ref{alg_fed_gd} 
can be modelled as perturbed \gls{gd} \eqref{equ_def_gd_step_linreg_perturbed} from Chapter \ref{lec_gradientmethods}. 
We can then analyze the robustness of the resulting \gls{fl} system via the convergence analysis of 
perturbed \gls{gd} discussed in Section \ref{sec_perturbed_gd}. 

According to \eqref{equ_upper_bound_pert_gd}, the performance of the decentralized Algorithm \ref{alg_fed_gd} 
degrades gracefully in the presence of imperfections such as missing or faulty communication links. 
In contrast, the server-based implementation of \gls{fedavg} Algorithm \ref{alg_fed_avg} 
offers a single point of failure (the server). 

Instead of modelling the effect of network failures as perturbed \gls{gd}, we can instead 
interpret it as exact \gls{gd} applied to a perturbed instance of \gls{gtvmin}. This 
perturbed instance uses a pruned \gls{flnetwork} $\widetilde{\graph}$, 
consisting of edges that are still active (i.e., corresponding to active 
communication links). 

The effectiveness of \gls{gtvmin} crucially depends on the second-smallest \gls{eigenvalue} $\eigval{2}$ 
of the \gls{LapMat} \eqref{equ_def_Lap_mat_entry} associated with the \gls{flnetwork} $\widetilde{\graph}$ (see 
Section \ref{sec_statasp_gtvmin}). As discussed in Section \ref{sec_emp_graph_design_choice}, 
$\eigval{2}$ reflects how well-connected the \gls{flnetwork} $\widetilde{\graph}$ is. 
A larger $\eigval{2}$ means better connectivity, which is required by \gls{gtvmin} to 
combine the information provided by \glspl{device} that work on similar \glspl{learningtask} 
(see Section \ref{sec_clustered_fl}). 

To make \gls{gtvmin} robust against communication link failures, we need to design 
the original \gls{flnetwork} $\graph$ so that even if some edges are removed, the resulting 
$\widetilde{\graph}$ still stays well connected—that is, $\eigval{2}$ remains large enough. 
This idea is related to resilient network design, which studies how to build networks that stay 
connected even when some parts fail \cite{parter:LIPIcs.DISC.2019.30,Chechik2009}.

\subsection{Privacy and Data Governance} 
\label{sec_priv_data_governance} 
``\emph{..privacy, a fundamental right 
	particularly affected by AI systems. Prevention of harm to privacy also necessitates adequate data 
	governance that covers the quality and integrity of the data used...}''\cite[p.17]{HLEGTrustworthyAI}.

We have introduced \gls{gtvmin} and \gls{flnetwork}s as abstract mathematical structures for the study 
of \gls{fl} systems. However, to obtain actual \gls{fl} systems we need to implement these mathematical 
concepts in a given physical hardware. These implementations incur deviations from the (idealized) 
\gls{gtvmin} formulation \eqref{equ_def_gtvmin} and the \glspl{gdmethod} (such as Algorithm \ref{alg_fed_gd})
used to solve it. For example, using quantized \gls{label} values results in a quantization error. 
Moreover, the \gls{localdataset}s can deviate significantly from a typical \gls{realization} of 
\gls{iid} \gls{rv}s, which is referred to as statistical bias \cite[Sec. 3.3.]{NISTDiffPriv2023})

\Gls{data} processing regulations limit the choice of the \gls{feature}s of a \gls{datapoint} \cite{Wachter:2019wn,Samarati2001,GDPR2016}. 
In particular, the \gls{gdpr} includes a \gls{dataminprinc} which requires to use only 
\gls{feature}s that are relevant for predicting the \gls{label}. 

{\bf Data Governance.} Some \gls{fl} applications involve \gls{localdataset}s that are 
generated by human users, i.e., personal data. Whenever personal data is used by a 
\gls{fl} method, special care must be dedicated towards data protection regulations \cite{GDPR2016}. 
It is useful (or even compulsory) to designate a data protection officer and conduct 
impact assessments \cite{HLEGTrustworthyAI}. 

{\bf Privacy.} The operation of an \gls{fl} system must not violate the fundamental 
human right to privacy \cite{UDHR1948}. One of most important characteristics 
of \gls{fl}, and distinguishing from distributed optimization, is the privacy friendly 
exchange of information among the system components. We dedicate the entire 
Chapter \ref{lec_privacyprotection} to the discussion of quantitative measures and 
methods for \gls{privprot} in \gls{gtvmin}-based \gls{fl} systems. 

\subsection{Transparency} 
\label{sec_kr_transparancy_explain}

{\bf Traceability.} This key requirement includes the documentation of design choices (and underlying 
business models) for a \gls{gtvmin}-based \gls{fl} system. This includes the source for the \gls{localdataset}s, the 
local \gls{model}s, the local \gls{lossfunc} as well as the construction of the \gls{flnetwork}. 
Moreover, the documentation should also cover the details of the implemented optimization method 
used to solve \gls{gtvmin}. This documentation might also require the periodic storing of 
the \glspl{modelparam} along with a time stamp (\emph{logging}). 

{\bf Communication.} Depending on the use case, \Gls{fl} systems need to communicate the capabilities 
and limitations to their end users (e.g., of a digital health app running on a smartphone). For example, 
we can indicate a measure of uncertainty about the \gls{prediction}s delivered by the trained \gls{localmodel}s. 
Such an uncertainty measure can be obtained naturally from a \gls{probmodel} for the \gls{data} generation. 
For example, the conditional  variance of the label $\truelabel$, given the \gls{feature}s $\featurevec$ of a 
random \gls{datapoint}. Another example of an uncertainty measure is the \gls{valerr} of a trained \gls{localmodel}. 

{\bf \Gls{explainability}.} The transparency of an \gls{fl} system can be facilitated by a 
sufficient level of \gls{explainability} of the trained personalized model $\hat{\hypothesis}^{(\nodeidx)} \in \localmodel{\nodeidx}$. 
It is important to note that the \gls{explainability} of $\hat{\hypothesis}^{(\nodeidx)}$ is 
subjective: A given learned \gls{hypothesis} $\hat{\hypothesis}^{(\nodeidx)}$ might offer a high 
degree of \gls{explainability} to one user (a graduate student at a university) but a low degree 
of \gls{explainability} to another user (a high-school student). We must ensure \gls{explainability} 
or the trained models $\hat{\hypothesis}^{(\nodeidx)}$ for potentially different users of the \glspl{device} 
$\nodeidx=1,\ldots,\nrnodes$.

The \gls{explainability} of trained \gls{ml} \glspl{model} is closely related to its simulatability \cite{Colin:2022aa,JunXML2020,Zhang:2024aa}: 
How well can a user anticipate (or guess) the prediction $\hat{\truelabel}=\hat{\hypothesis}^{(\nodeidx)}(\featurevec)$ 
delivered by $\hat{\hypothesis}^{(\nodeidx)}$ for a \gls{datapoint} with \gls{feature}s $\featurevec$. 
We can then measure the \gls{explainability} of $\hat{\hypothesis}^{(\nodeidx)}(\featurevec)$ to the user 
at node $\nodeidx$ by comparing the \gls{prediction} $\hat{\hypothesis}^{(\nodeidx)}(\featurevec)$ with 
the corresponding \emph{guess} (or \emph{simulation}) $\user^{(\nodeidx)}(\featurevec)$.  

We can enforce (subjective) \gls{explainability} of \gls{fl} systems by modifying the local \gls{lossfunc}s 
in \gls{gtvmin}. For ease of exposition, we focus on the \gls{gtvmin} instance \eqref{equ_def_gtvmin_linreg_lec5} 
for training local (personalized) \gls{linmodel}s. For each node $\nodeidx \in \nodes$, we construct 
a test-set $\localtestset{\nodeidx}$ and ask user $\nodeidx$ to deliver a guess $\user^{(\nodeidx)}(\featurevec)$ 
for each \gls{datapoint} in $\localtestset{\nodeidx}$.\footnote{We only use the \gls{feature}s of 
	the \gls{datapoint}s in $\localtestset{\nodeidx}$, i.e., this \gls{dataset} can be constructed 
	from unlabeled \gls{data}.} 

We measure the (subjective) \gls{explainability} of a linear \gls{hypothesis} 
with \glspl{modelparam} $\localparams{\nodeidx}$ by 
\begin{equation}
	\label{equ_def_sub_explainability} 
	(1/\big|\localtestset{\nodeidx}\big|) \sum_{\featurevec \in \localtestset{\nodeidx}} \bigg( \user^{(\nodeidx)}\big(\featurevec \big) - \featurevec^{T} \localparams{\nodeidx}  \bigg)^{2}. 
\end{equation}
It seems natural to add this measure as a penalty term to the local \gls{lossfunc} in \eqref{equ_def_gtvmin_linreg_lec5}, 
resulting in the new \gls{lossfunc}
\begin{equation} 
	\label{equ_def_local_loss_subj_explain}
	\locallossfunc{\nodeidx}{\localparams{\nodeidx}}\!\defeq\!\underbrace{ (1/\localsamplesize{\nodeidx}) \normgeneric{\labelvec^{(\nodeidx)}\!-\!\featuremtx^{(\nodeidx)} \localparams{\nodeidx}}{2}^{2}}_{\mbox{\gls{trainerr}}} \!+\!\rho \underbrace{(1/\big|\localtestset{\nodeidx}\big|) \hspace*{-2mm} \sum_{\featurevec \in \localtestset{\nodeidx}} \hspace*{-2mm}\big( \user^{(\nodeidx)}\big(\featurevec \big)\!-\!\featurevec^{T} \localparams{\nodeidx}  \big)^{2}}_{\mbox{ subjective \gls{explainability}}}. 
\end{equation} 
The \gls{regularization} parameter $\rho$ controls the preference for a high subjective \gls{explainability} 
of the \gls{hypothesis} $\hypothesis^{(\nodeidx)}(\featurevec) = \big( \localparams{\nodeidx}\big)^{T} \featurevec$ 
over a small \gls{trainerr} \cite{Zhang:2024aa}. It can be shown that \eqref{equ_def_local_loss_subj_explain} is the average 
weighted \gls{sqerrloss} of $\hypothesis^{(\nodeidx)}(\featurevec)$ on an augmented version of $\localdataset{\nodeidx}$. 
This augmented version includes the \gls{datapoint} $\pair{\featurevec}{\user^{(\nodeidx)}(\featurevec)}$ for each 
\gls{datapoint} $\featurevec$ in the test-set $\localtestset{\nodeidx}$. 

So far, we have focused on the problem of explaining (the \gls{prediction}s of) a 
trained personalized \gls{model} to some user. The general idea is to provide 
partial information, in the form of some explanation, about the learned \gls{hypothesis} 
map $\learnthypothesis$. Explanations should help the user to anticipate the 
\gls{prediction} $\learnthypothesis(\featurevec)$ for any given \gls{datapoint}. 
Instead of explaining a given trained model $\learnthypothesis$, it might be more 
useful to explain an entire \gls{fl} \gls{algorithm}. 

Mathematically, we can interpret an \gls{fl} \gls{algorithm} as a map $\algomap$ that reads 
in \gls{localdataset}s and delivers learned \gls{hypothesis} maps $\learnthypothesis^{(\nodeidx)}$. 
We can explain an \gls{fl} \gls{algorithm} by providing partial information about this \gls{map} 
$\algomap$. Thus, mathematically speaking, the problem of explaining a learned \gls{hypothesis} 
is essentially the same as the problem of explaining an entire \gls{fl} \gls{algorithm}: 
Provide partial information about a map such that the user can anticipate the results 
of applying the map to arbitrary arguments. However, a description of the map $\algomap$ is 
typically more complex, in a quantitative sense, than a learned \gls{hypothesis} map.

The different complexity levels of maps to be explained requires different 
forms of explanation. For example, we could explain an \gls{fl} \gls{algorithm} 
using a pseudo-code such as Algorithm \ref{alg_fed_gd}. Fig.\ \ref{fig_fitting_dt_iris} 
illustrates another form of explanation, i.e., a code fragment written in the programming 
language Python.
\begin{figure}[htbp]
	\centering
\begin{lstlisting}[language=Python]
from sklearn.datasets import load_iris
from sklearn.model_selection import train_test_split
from sklearn.tree import DecisionTreeClassifier
from sklearn.metrics import accuracy_score

# Load the Iris dataset
data = load_iris()
X = data.data
y = data.target

# Split the dataset into training and test sets
X_train, X_test, y_train, y_test = train_test_split(X, y, test_size=0.3, random_state=42)

# Create a Decision Tree classifier
clf = DecisionTreeClassifier(random_state=42)

# Train the classifier
clf.fit(X_train, y_train)

# Make predictions on the test data
y_pred = clf.predict(X_test)

# Calculate accuracy
accuracy = accuracy_score(y_test, y_pred)
accuracy
\end{lstlisting}
	\caption{Python code for a \gls{ml} method that trains a \gls{decisiontree} on the \emph{Iris} \gls{dataset}. \label{fig_fitting_dt_iris}}
\end{figure}

\clearpage
\subsection{Diversity, Non-Discrimination and Fairness} ``\emph{...we must enable inclusion and diversity 
	throughout the entire AI system’s life cycle...this also entails ensuring equal access through inclusive 
	design processes as well as equal treatment.}''\cite[p.18]{HLEGTrustworthyAI}.

The \glspl{localdataset} used for the training of \glspl{localmodel} should be carefully selected 
to not enforce existing discrimination. In a health-care application, there 
might be significantly more training data for patients of a specific gender, 
resulting in models that perform best for that specific gender at the cost 
of worse performance for the minority \cite[Sec. 3.3.]{NISTDiffPriv2023}. 

Fairness is also important for \gls{ml} methods used to determine credit score 
and, in turn, if a loan should be granted or not \cite{KOZODOI20221083}. 
Here, we must ensure that \gls{ml} methods do not discriminate against customers 
based on ethnicity or race. To this end, we could augment \glspl{datapoint} 
by modifying any \glspl{feature} that mainly reflect the ethnicity or race of 
a customer (see Fig.\ \ref{fig_fairness_aug}).

\begin{figure}[htbp]
	\begin{center} 
		\begin{tikzpicture}[scale = 1]
			\draw[->, very thick] (0,0.5) -- (7.7,0.5) node[right] {gender $\feature$};       
			\draw[->, very thick] (0.5,0) -- (0.5,4.2) node[above] {compensation $\truelabel$};   
			
			\draw[color=black, thick, dashed, domain = -0.5: 5.2, variable = \x]  plot ({\x},{\x*0.4 + 2.0}) ;     
			\node at (5.7,4.1) {$\hypothesis(\feature)$};    
			
			\coordinate (l1)   at (1.2, 2.48);
			\coordinate (l2) at (1.4, 2.56);
			\coordinate (l3)   at (1.7,  2.68);
			
			\coordinate (l4)   at (2.2, 2.2*0.4+2.0);
			\coordinate (l5) at (2.4, 2.4*0.4+2.0);
			\coordinate (l6)   at (2.7,  2.7*0.4+2.0);
			
			\coordinate (l7)   at (3.9,  3.9*0.4+2.0);
			\coordinate (l8) at (4.2, 4.2*0.4+2.0);
			\coordinate (l9)   at (4.5,  4.5*0.4+2.0);
			
			\coordinate (n1)   at (1.2, 1.8);
			\coordinate (n2) at (1.4, 1.8);
			\coordinate (n3)   at (1.7,  1.8);
			
			\coordinate (n4)   at (2.2, 3.8);
			\coordinate (n5) at (2.4, 3.8);
			\coordinate (n6)   at (2.7,  3.8);

			\coordinate (n7)   at (3.9, 2.6);
			\coordinate (n8) at (4.2, 2.6);
			\coordinate (n9)   at (4.5,  2.6);
			
			\node at (n1)  [circle,draw,fill=red,minimum size=6pt,scale=0.6, name=c1] {};
			\node at (n2)  [circle,draw,fill=blue,minimum size=6pt, scale=0.6, name=c2] {};
			\node at (n3)  [circle,draw,fill=red,minimum size=6pt,scale=0.6,  name=c3] {};
			\node at (n4)  [circle,draw,fill=red,minimum size=12pt, scale=0.6, name=c4] {};  
			\node at (n5)  [circle,draw,fill=blue,minimum size=12pt,scale=0.6,  name=c5] {};
			\node at (n6)  [circle,draw,fill=red,minimum size=12pt, scale=0.6, name=c6] {};  
			\node at (n7)  [circle,draw,fill=red,minimum size=12pt,scale=0.6,  name=c7] {};
			\node at (n8)  [circle,draw,fill=blue,minimum size=12pt, scale=0.6, name=c8] {};
			\node at (n9)  [circle,draw,fill=red,minimum size=12pt, scale=0.6, name=c9] {};
			

			\draw[<->, color=red, thick] (l1) -- (c1);  
			\draw[<->, color=blue, thick] (l2) -- (c2);  
			\draw[<->, color=red, thick] (l3) -- (c3);  
			\draw[<->, color=red, thick] (l4) -- (c4);  
			\draw[<->, color=blue, thick] (l5) -- (c5);  
			\draw[<->, color=red, thick] (l6) -- (c6);  
			\draw[<->, color=red, thick] (l7) -- (c7);  
			\draw[<->, color=blue, thick] (l8) -- (c8);  
			\draw[<->, color=red, thick] (l9) -- (c9);  
			
			\draw[fill=blue] (6.2, 3.7)  circle (0.1cm) node [black,xshift=2.3cm] {original \gls{trainset} $\dataset$};
			\draw[fill=red] (6.2, 3.2)  circle (0.1cm) node [black,xshift=1.3cm] {augmented};
		\end{tikzpicture}
		\caption{We can improve the fairness of a \gls{ml} method by augmenting the \gls{trainset} using perturbations 
			of an irrelevant \gls{feature} such as the gender of a person for which we want to predict the adequate 
			compensation as the \gls{label}. \label{fig_fairness_aug} }
	\end{center}
\end{figure} 

\newpage
\subsection{Societal and Environmental Well-Being} ``\emph{...Sustainability and
	ecological responsibility of AI systems should be encouraged, and research should be 
	fostered into AI solutions addressing areas of global concern, such as for instance the 
	Sustainable Development Goals.}''\cite[p.19]{HLEGTrustworthyAI}.

{\bf Society.} \Gls{fl} systems might be used to deliver personalized recommendations to users within a 
social media application (social network). These recommendations might be (fake) news 
used to boost polarization and, in the extreme case, social unrest \cite{Goncalves-Sa:2024aa}. 

{\bf Environment.} Chapter \ref{lec_flalgorithms} discussed \gls{fl} algorithms that were 
obtained by applying \glspl{gdmethod} to solve \gls{gtvmin}. These methods require 
computational resources to compute local updates for \glspl{modelparam} and to share 
them across the edges of the \gls{flnetwork}. Computation and communication require 
energy which should be generated in an environmental-friendly fashion \cite{GreenCom}.

\newpage

\subsection{Exercises}

\refstepcounter{problem}\label{prob:robustnessgtvmin}\textbf{\theproblem. Robustness of \gls{gtvmin}.}
Discuss the \gls{robustness} of \gls{gtvmin} \eqref{equ_def_gtvmin_linreg} for training 
local \glspl{linmodel}. In particular, which \gls{attack} is more effective (detrimental): 
perturbing the \glspl{label}, the \glspl{feature} of \gls{datapoint}s in the \glspl{localdataset} or perturbing 
the \gls{flnetwork}, e.g., by removing (or adding) edges.  

\noindent\refstepcounter{problem}\label{prob:subjexp}\textbf{\theproblem. Subjectively Explainable FL.}
Consider \gls{gtvmin} \eqref{equ_def_gtvmin_linreg} to train local \gls{linmodel}s with \glspl{modelparam} $\localparams{\nodeidx}$. 
The \gls{localdataset}s are modelled as \eqref{equ_def_probmodel_linreg_node_i}. 
Each \gls{localmodel} has a user that is characterized by the user signal $\user(\featurevec) \defeq \featurevec^{T} \vu^{(\nodeidx)}$. 
To ensure subjective \gls{explainability} of \gls{localmodel} with \glspl{modelparam} $\localparams{\nodeidx}$ 
we require the deviation $ (1/\localsamplesize{\nodeidx}) \normgeneric{\widetilde{\featuremtx}^{(\nodeidx)} \big( \localparams{\nodeidx} - \vu^{(\nodeidx)} \big)}{2}^{2}$ to be sufficiently small. Here, we used the \gls{featuremtx} $\widetilde{\featuremtx}^{(\nodeidx)}$ 
obtained from the \gls{realization} of $\localsamplesize{\nodeidx}$ \gls{iid} \glspl{rv} with common 
\gls{probdist} $\mathcal{N}(\mathbf{0}, \mathbf{I})$. We then add this deviation to the local \glspl{lossfunc} 
resulting in using the augmented \gls{lossfunc} \eqref{equ_def_local_loss_subj_explain} used in 
\eqref{equ_def_gtvmin_linreg}. Study, either analytically or by numerical experiments, the effect of 
varying levels of \gls{explainability} (via the parameter $\rho$ in \eqref{equ_def_local_loss_subj_explain}) 
on the estimation error $\estlocalparams{\nodeidx} - \overline{\weights}^{(\nodeidx)}$. 

\newpage
\section{Privacy Protection in FL} 
\label{lec_privacyprotection} 
The core idea of \gls{fl} is to share information contained in collections of 
\glspl{localdataset} to improve the training of personalized \gls{ml} \glspl{model}. 
Chapter \ref{lec_flalgorithms} discussed \gls{fl} algorithms that share information 
in the form of \glspl{modelparam} that are computed from the local \gls{lossfunc}. 
Each node $\nodeidx \in \nodes$ receives the current \glspl{modelparam} of 
other nodes and, after executing a local update, shares its new \glspl{modelparam} 
with other nodes. 

Depending on the design choices for \gls{gtvmin}-based methods, sharing \glspl{modelparam}  
allows to reconstruct local \glspl{lossfunc} and, in turn, to estimate private information about 
individual \glspl{datapoint} which represent human patients \cite{Sheller:2020aa}. 
Thus, the bad news is that \gls{fl} systems will almost inevitably incur some leakage of 
private information. The good news is, however, that the extent of privacy leakage can be 
controlled by (i) careful design choices for \gls{gtvmin} and (ii) applying modifications 
to basic \gls{fl} \glspl{algorithm} from Chapter \ref{lec_flalgorithms}. 

This chapter revolves around two main questions: 
\begin{itemize} 
	\item How can we measure \gls{privleakage} in an \gls{fl} system? 
	\item How can we control (minimize) \gls{privleakage} of an \gls{fl} system? 
\end{itemize} 
Section \ref{sec_measuring_privacy_leakage} addresses the first question while Sections \ref{sec_diff_privacy} 
and \ref{sec_private_feature_learning} address the second question.

\subsection{Measuring Privacy Leakage} 
\label{sec_measuring_privacy_leakage}

Consider an \gls{fl} system designed to train personalized \glspl{model} for users indexed 
by $\nodeidx = 1, \ldots, \nrnodes$, each equipped with a heart rate sensor. Every user 
$\nodeidx$ generates a \gls{localdataset} $\localdataset{\nodeidx}$, consisting of 
time-stamped heart rate measurements. A single \gls{datapoint} corresponds to one 
physical activity, such as a 50-minute run. The \glspl{feature} of such a \gls{datapoint} 
include a time series of GPS coordinates, while the \gls{label} may be the average heart rate 
recorded during the activity. We assume that this average heart rate is private and should 
not be disclosed to third parties.\footnote{For instance, individuals may not wish to share heart rate profiles with potential employers.}

To enhance learning, the \gls{fl} system incorporates expert-provided information in the 
form of pairwise similarity scores $\edgeweight_{\nodeidx, \nodeidx'}$ between users 
$\nodeidx$ and $\nodeidx'$, based on characteristics such as body weight and height. 
These similarity scores are used to regularize the learning process.

Using an \gls{fl} \gls{algorithm}---such as Algorithm \ref{alg_fed_gd}---we aim to learn, for 
each user $\nodeidx$, personalized \glspl{modelparam} $\localparams{\nodeidx}$ for an \gls{ai}-powered 
healthcare assistant \cite{PersonalizedHR2019}. This \gls{algorithm} can be represented as a map 
$\algomap(\cdot)$ that takes as input the collection of \glspl{localdataset}   
\[
\dataset \defeq \big\{ \localdataset{\nodeidx} \big\}_{\nodeidx=1}^{\nrnodes}
\]
and delivers learned \glspl{modelparam}
\[
\algomap(\dataset) \defeq  \big( \estlocalparams{1},\ldots,\estlocalparams{\nrnodes} \big).
\]
Figure \ref{fig_algo_map_fl_gd} illustrates the mapping from \glspl{localdataset} to learned \glspl{modelparam} 
that is implemented by an \gls{fl} \gls{algorithm}. 

\begin{figure}
	\begin{center}
		\begin{tikzpicture}
			
			\tikzset{datapoint/.style={circle, fill=blue, minimum size=4pt, inner sep=0pt}}
			\tikzset{parameter/.style={circle, fill=black, minimum width=4pt, inner sep=0pt}}
			\tikzset{mapping/.style={->, thick, shorten >=1pt}}
			
			\node[datapoint, label=left:{}] (d1) at (0.2,1) {};
			
			\node[datapoint, label=left:{}] (d2) at (-0.5,0) {};
			\node[right = 0.01cm of d2] () {$\localdataset{\nodeidx}$};
			\node[datapoint, label=left:{}] (d3) at (0,-1) {};
			\node[below right = 1.4cm of d2] {\(\mathcal{D}\)};
			\node[draw, ellipse, dashed, minimum width=2cm, minimum height=3cm, fit=(d1) (d2) (d3), inner sep=0pt] (ellipse) {};
			

			\node[label=right:{$\estlocalparams{1},\ldots,\estlocalparams{\nrnodes}$}] (p) at (4,0) {};
			
			\draw[mapping] (ellipse) -- (p) node[midway, above] {$\algomap$};
			
		\end{tikzpicture}
	\end{center}
	\caption{A \gls{fl} \gls{algorithm} maps the \gls{localdataset}s $\localdataset{\nodeidx}$ 
		to the learned \glspl{modelparam} $\estlocalparams{\nodeidx}$, for $\nodeidx=1,\dots,\nrnodes$. 
		\label{fig_algo_map_fl_gd}}
\end{figure} 

A privacy-preserving \gls{fl} system should not allow to infer, solely from the 
learned \glspl{modelparam}, the average heart rate $\truelabel^{(\nodeidx,\sampleidx)}$ 
during a specific single activity $\sampleidx$ of a specific user $\nodeidx$. Mathematically, we 
must ensure that the map $\algomap$ is not invertible: The learned \glspl{modelparam} (or \gls{hypothesis}) 
should not change if we were to apply the \gls{fl} \gls{algorithm} to a perturbed \gls{dataset} 
that includes a different value for the average heart rate $\truelabel^{(\nodeidx,\sampleidx)}$. 

Figure \ref{fig_scatterplot_decregion_dp} depicts the \gls{decisionregion}s of a \gls{decisiontree}. 
This \gls{decisiontree} has been trained by (approximately) solving \gls{erm} with a \gls{trainset} 
that consists of five \gls{datapoint}s. Each \gls{datapoint} is characterized by a \gls{featurevec} 
$\featurevec^{(\sampleidx)} = \big( \feature^{(\sampleidx)}_{1},\feature^{(\sampleidx)}_{2}\big)^{T}$ and 
a binary \gls{label} $\truelabel^{(\sampleidx)} \in \{ \circ, \times \}$, for $\sampleidx=1,\ldots,5$.
If an attacker would know the \gls{label} values of $\featurevec^{(1)}, \featurevec^{(4)}$, it could 
infer the \gls{label} of $\featurevec^{(2)}$ based on the \gls{decisionregion}s.

\begin{figure}[htbp]
	\begin{center}
		\begin{tikzpicture}
			\begin{axis}[
				domain=-2:5,
				xmin=0, xmax=5,
				ymin=0, ymax=5,
				xlabel=$\feature_{1}$,
				ylabel=$\feature_{2}$,
				ylabel style={at={(axis description cs:0.1,0.5)}, rotate=-90, anchor=near ticklabel},
				scatter/classes={
					a={mark=o,draw=black,mark size=3pt,line width=2pt}, 
					b={mark=x,draw=black,mark size=3pt,line width=2pt}  
				}
				]
				\addplot[fill=blue, fill opacity=0.2] coordinates {(0,0) (5,2) (5,0)} \closedcycle;
				\addplot[fill=red, fill opacity=0.2] coordinates {(0,5) (5,5) (5,2)} \closedcycle;
				\addplot[fill=green, fill opacity=0.2] coordinates {(0,0) (5,2) (0,5)} \closedcycle;
				
				\addplot[scatter,only marks,scatter src=explicit symbolic]
				table[meta=label]{
					x y label
					1 1 a
					4 1 b
					2 4 a
					3 3 a
					4.5 3.5 b
				};
				\node[rotate=0, above] at (axis cs: 1,1) {$\featurevec^{(1)}$};
				\node[rotate=0, above] at (axis cs: 4,1) {$\featurevec^{(2)}$};
				\node[rotate=0, above] at (axis cs: 2,4) {$\featurevec^{(3)}$};
				\node[rotate=0, below] at (axis cs: 3,3) {$\featurevec^{(4)}$};
				\node[rotate=0, above] at (axis cs: 4.5,3.5) {$\featurevec^{(5)}$};
			\end{axis}
			\vspace*{-2mm}
		\end{tikzpicture}
		\vspace*{-2mm}
	\end{center} 
	\vspace*{-5mm}
	\caption{\label{fig_scatterplot_decregion_dp} \Gls{scatterplot} of a \gls{dataset} used to train 
		a \gls{decisiontree}. We indicate the \gls{decisionregion}s along with the \gls{label}s of 
		\gls{datapoint}s (via their markers).} 
\end{figure}

\begin{figure}[htbp]
	\centering
	\begin{tikzpicture}
		\begin{axis}[
			domain=-3:3,
			samples=100,
			axis lines=middle,
			xtick=\empty,
			ytick=\empty,
			xmin=-4, xmax=4,
			ymin=0,
			ymax=0.45,
			width=10cm,
			height=6cm,
			]
			\addplot[blue, thick] {exp(-x^2 / 2) / sqrt(2 * pi)};
			\addplot[red, thick] {exp(-(x-1)^2 / 2) / sqrt(2 * pi)};
			\draw[thick, line width=2mm] (axis cs: -2.5, 0.00) -- (axis cs: 0.0, 0.00);
			\node[above,left] at (axis cs: -0.5, 0.05) {$\mathcal{T}$};
			\node[above,left] at (axis cs: -0.4, 0.35) {$p(\widehat{\weights};\dataset)$};
			\node[above] at (axis cs: 3.6, 0.0) {$\widehat{\weights}$};
			\node[above] at (axis cs: 2.2, 0.35) {$p(\widehat{\weights};\dataset')$};
		\end{axis}
	\end{tikzpicture}
	\caption{\Gls{probdist}s of the learned \glspl{modelparam} $\widehat{\weights}=\big(\estlocalparams{1},\ldots,\estlocalparams{\nrnodes}\big)$ 
		delivered by some \gls{fl} \gls{algorithm} for two different 
		input \glspl{dataset}, denoted by $\dataset'$ and $\dataset$.}
	\label{fig:pdf_random_algorithms}
\end{figure}

The sole requirement for an \gls{fl} \gls{algorithm} $\algomap$ to be not invertible is not sufficient in general. 
Indeed, we can easily make any \gls{algorithm} $\algomap$ by simple pre- or post-processing 
techniques whose effect is limited to irrelevant regions of the input space. Note that 
the input space is the space of all possible \glspl{dataset}. The level of \gls{privprot} 
offered by $\algomap$ can be characterized by a measure of its non-invertibility (or non-injectivity). 

A simple measure of non-invertibility is the sensitivity of the output $\algomap\big( \dataset \big)$ 
against varying the heart rate value $\truelabel^{(\nodeidx,\sampleidx)}$, 
\begin{equation}
	\label{equ_def_sensitivity_privacy_protection}
	\frac{\normgeneric{\algomap\big(\dataset\big) - \algomap\big( \dataset' \big)}{2}}{\varepsilon}. 
\end{equation} 
Here, $\dataset$ denotes some given collection of \gls{localdataset}s and $\dataset'$ is a modified \gls{dataset}. 
In particular, $\dataset'$ is obtained by replacing the actual average heart rate $\truelabel^{(\nodeidx,\sampleidx)}$ 
with the modified value $\truelabel^{(\nodeidx,\sampleidx)}+ \varepsilon$. 
The \gls{privprot} offered by $\algomap$ is higher for smaller values \eqref{equ_def_sensitivity_privacy_protection}, i.e., 
the output changes only a little when varying the value of the average heart rate. 

Another measure for the non-invertibility of $\algomap$ is referred to as \gls{diffpriv}. 
This measure is particularly useful for stochastic \glspl{algorithm} that use some random 
mechanism for learning \glspl{modelparam}. One example of such a mechanism is the 
random selection of a subset \glspl{datapoint} that form a \gls{batch} within one iteration of 
FedSGD (see Algorithm \ref{alg_fed_sgd_general}). 
Section \ref{sec_diff_privacy} discusses another example of a random mechanism: add 
the \gls{realization} of a \gls{rv} to the intermediate results of an \gls{algorithm}. 

A stochastic \gls{algorithm} $\algomap$ can be described by a \gls{probdist} $p(\widehat{\weights};\dataset)$ 
over the possible values of the learned \glspl{modelparam} $\widehat{\weights}$. Figure \ref{fig:pdf_random_algorithms} 
illustrates a stochastic \gls{algorithm} along with the associated \gls{probdist} $p(\widehat{\weights};\dataset)$.\footnote{For 
	more details about the concept of a measurable space, we refer to the literature \cite{AshProbMeasure,HalmosMeasure,BillingsleyProbMeasure}.} 
This \gls{probdist} is parametrized by the \gls{dataset} $\dataset$ that is fed as input 
to the \gls{algorithm} $\algomap$. Figure \ref{fig:pdf_random_algorithms} depicts the 
\gls{probdist}s of an \gls{algorithm} for two different choices $\dataset, \dataset'$ of 
the input \gls{dataset}. 

\Gls{diffpriv} measures the non-invertibility of a stochastic \gls{algorithm} $\algomap$ 
via the similarity of the \gls{probdist}s obtained for two \glspl{dataset} $\dataset,\dataset'$ 
that are considered as adjacent (or neighbouring) \cite{NISTDiffPriv2023,AlgoFoundDP}. 
Typically, we consider $\dataset'$ as adjacent to $\dataset$ if it is obtained by modifying 
the \glspl{feature} or \gls{label} of a single \gls{datapoint} in $\dataset$. 

As a case in point, consider \glspl{datapoint} representing physical activities which are 
characterized by a binary \gls{feature} $\feature_{\featureidx} \in \{0,1\}$ that indicates 
an excessively high average heart rate during the activity. We could then define 
neighbouring \glspl{dataset} by changing the \gls{feature} $\feature_{\featureidx}$ of 
a single \gls{datapoint}. In general, the notion of neighbouring \glspl{dataset} is a design 
choice used in the definition of quantitative measures for \gls{privprot}. 
A \gls{fl} \gls{algorithm} ensures \gls{privprot} if there is no statistical test that allows to 
reliably distinguish between neighbouring input \glspl{dataset}. Figure \ref{fig:pdf_random_algorithms} 
illustrates the acceptance region $\mathcal{T}$ that defines a statistical test.

The de-facto standard for quantifying \gls{privleakage} in \gls{ml} and \gls{fl} systems is the following 
definition. 
 \begin{definition}
	\label{equ_def_dp}
	(from \cite{AlgoFoundDP}) A stochastic \gls{algorithm} $\algomap$ is $(\varepsilon,\delta)$-\gls{diffpriv} if, 
	for any two neighbouring \gls{dataset}s $\dataset, \dataset'$, 
	\begin{equation} 
		\nonumber
		{\rm Prob} \big\{ \algomap(\dataset) \in \mathcal{S} \big\} \leq \exp(\varepsilon) 	{\rm Prob} \big\{ \algomap(\dataset') \in \mathcal{S} \} + \delta
	\end{equation} 
	holds for every measurable set $\mathcal{S}$. 
\end{definition}
Definition \ref{equ_def_dp} formalizes the notion that the presence or absence of an 
individual \gls{datapoint} (representing, e.g., human individual) in a \gls{dataset} $\dataset$ 
should not significantly affect the \gls{probdist} of the output $\algomap(\dataset)$. 
The notion of $(\varepsilon,\delta)$-\gls{diffpriv} is widely adopted in \gls{fl} applications 
\cite{AlgoFoundDP,Erlingsson2014,apple2025differentialprivacy,Abowd2018}. 
The U.S. Census Bureau adopted $(\varepsilon,\delta)$-\gls{diffpriv} for the 2020 
census \cite{Abowd2018}. The National Institute of Standards and Technology (NIST) 
has published some guidance for evaluating and implementing \gls{diffpriv} 
mechanisms in government and industry settings \cite{Near2025DPGuidelines}. 

Besides $(\varepsilon,\delta)$-\gls{diffpriv}, there are also 
other measures for \gls{privleakage}. These measures differ by 
how they quantify precisely the similarity between \glspl{probdist} 
$p(\widehat{\weights};\dataset)$ and $p(\widehat{\weights};\dataset')$ 
induced by neighbouring \glspl{dataset} \cite{fDivPrivGuaran2020}. 
One such alternative measure is the \gls{renyidiv} of order $\alpha> 1$ \cite{Renyi1961},
\begin{equation}
	\nonumber
	D_{\alpha} \bigg(p(\widehat{\weights};\dataset) \big\|   p(\widehat{\weights};\dataset') \bigg) \defeq \frac{1}{\alpha-1} \log \expect_{p(\widehat{\weights};\dataset')} \bigg[ \bigg( \frac{dp(\widehat{\weights};\dataset) }{d p(\widehat{\weights};\dataset')}\bigg)^{\alpha}  \bigg]. 
\end{equation} 
The \gls{renyidiv} allows to define the following variant of \gls{diffpriv} \cite{RenyiDiffPriv,fDivPrivGuaran2020}. 
\begin{definition}
	\label{equ_def_rdp}
	(from \cite{AlgoFoundDP}) An \gls{algorithm} $\algomap$ is $(\alpha,\gamma)$-RDP if, 
	for any two neighbouring \glspl{dataset} $\dataset$ and $\dataset'$, 
	\begin{equation}
		\nonumber 
		D_{\alpha} \bigg(p(\widehat{\weights};\dataset) \big\|  p(\widehat{\weights};\dataset') \bigg)  \leq \gamma. 
	\end{equation} 
\end{definition}
A recent use-case of $(\alpha,\gamma)$-RDP is the analysis of \gls{diffpriv} guarantees 
offered by variants of \gls{stochGD} \cite{fDivPrivGuaran2020}. This analysis uses 
the fact that $(\alpha,\gamma)$-RDP implies $(\varepsilon,\delta)$-\gls{diffpriv} for 
suitable choices of $\varepsilon, \delta$ \cite{fDivPrivGuaran2020}. 

One important property of the \gls{diffpriv} notions in Definition \ref{equ_def_dp} and 
Definition \ref{equ_def_rdp} is that they are preserved by post-processing: 
\begin{prop}
	\label{prop_post_processing_dp}
	Consider an \gls{fl} \gls{algorithm} $\algomap$ that is applied to some \gls{dataset} $\dataset$ and some
	(possibly stochastic) map $\mathcal{B}$ that does not depend on $\dataset$. If $\algomap$ is $(\varepsilon,\delta)$-\gls{diffpriv} (or $(\alpha,\gamma)$-RDP), then so is also the composition $\mathcal{B} \circ \algomap$. 
\end{prop} 
\begin{proof} 
	See, e.g., \cite[Sec. 2.3]{AlgoFoundDP}. 
\end{proof} 
According to Proposition \ref{prop_post_processing_dp}, the level of \gls{diffpriv} 
offered by an \gls{algorithm} $\algomap$ does not deteriorate by any post-processing of 
its output. It seems almost natural to make this immunity against post-processing a defining 
property of any useful notion of \gls{diffpriv} \cite{RenyiDiffPriv}. However, due to Proposition \ref{prop_post_processing_dp}, 
this property is already ``built-in'' into the Definition \ref{equ_def_dp} and Definition \ref{equ_def_rdp}. 

{\bf Operational Meaning of \gls{diffpriv}.} The mathematically precise formulation 
 of \gls{diffpriv} in Definition \ref{equ_def_dp} is somewhat abstract. It is instructive 
 to interpret $(\varepsilon,\delta)$-\gls{diffpriv} from the perspective of hypothesis testing \cite{Near2025DPGuidelines}: 
 We use the output $\widehat{\weights} \in \mathbb{R}^{\dimlocalmodel}$ of \gls{algorithm} 
 $\algomap$ to test if the underlying \gls{dataset} fed into $\algomap$ 
was $\dataset$ or if it was a neighbouring \gls{dataset} $\dataset'$ \cite{Kay:1998aa}. 
Such a statistical test uses a region $\mathcal{T} \subseteq \mathbb{R}^{\dimlocalmodel}$ 
and to declare    
\begin{itemize} 
	\item ``\gls{dataset} $\dataset$ seems to be used'' if $\widehat{\weights} \in \mathcal{T}$, or 
	\item ``\gls{dataset} $\dataset'$ seems to be used'' if $\widehat{\weights} \notin \mathcal{T}$. 
\end{itemize} 
The performance of a test $\mathcal{T}$ is characterized by two error \glspl{probability}: 
\begin{itemize} 
	\item The \gls{probability} of declaring 
	$\dataset'$ but actually $\dataset$ was fed into $\algomap$, which is $P_{\dataset \rightarrow \dataset'} \defeq 1- \int_{ \mathcal{T} } p(\widehat{\weights};\dataset)$. 
	\item The \gls{probability} of declaring $\dataset$ but actually $\dataset'$ was fed into $\algomap$, which is $P_{\dataset' \rightarrow \dataset} \defeq \int_{ \mathcal{T} } p(\widehat{\weights};\dataset')$.
\end{itemize} 

For a privacy-preserving \gls{algorithm} $\algomap$, there should be no test $\mathcal{T}$ for which both 
$P_{\dataset \rightarrow \dataset'}$ and $P_{\dataset' \rightarrow \dataset}$ are simultaneously 
small (close to $0$). This intuition can be made precise as follows (see \cite[Thm. 2.1.]{pmlr-v37-kairouz15}, \cite{Near2025DPGuidelines} or \cite{OptimalNoiseDP}): 
If an \gls{algorithm} $\algomap$ is $(\varepsilon,\delta)$-\gls{diffpriv}, then
\begin{equation} 
	\label{equ_lowerbound_test_err_dp}
	\exp(\varepsilon) P_{\dataset \rightarrow \dataset'} + P_{\dataset' \rightarrow \dataset} \geq 1- \delta. 
\end{equation} 
Thus, if $\algomap$ is $(\varepsilon,\delta)$-\gls{diffpriv} with a small $\varepsilon,\delta$ (close to $0$), 
then \eqref{equ_lowerbound_test_err_dp} implies $P_{\dataset \rightarrow \dataset'} + P_{\dataset' \rightarrow \dataset} \approx 1$. 

\subsection{Ensuring Differential Privacy} 
\label{sec_diff_privacy}

Depending on the underlying \glspl{localdataset}, \glspl{localmodel}, and \gls{optmethod}, 
a \gls{gtvmin}-based method $\algomap$ might already ensure \gls{diffpriv} by 
design. A basic means of ensuring \gls{diffpriv} is through careful \gls{feature} selection (or learning) 
for the \glspl{localdataset} (see Figure \ref{fig_pp_feature_learning_lin}). Random sampling 
used by \gls{stochGD}-based \glspl{algorithm} can also provide a certain level of \gls{diffpriv} \cite{AbadiDeepLearningDP,pmlr-v235-chua24a}. 

According to Proposition \ref{prop_post_processing_dp}, \gls{diffpriv} can also be actively ensured 
by applying pre- and post-processing techniques to the input and output of an \gls{fl} \gls{algorithm} $\algomap$.
Mathematically, the \gls{map} $\algomap$ is concatenated with the maps $\mathcal{I}$ and $\mathcal{O}$.
These maps represent the pre- and post-processing and are typically stochastic, i.e., defined by a conditional \gls{probdist}. 
The concatenation results in a new \gls{algorithm} $\algomap' \defeq \mathcal{O} \circ \algomap \circ \mathcal{I}$. 
In summary, for a given \gls{dataset} $\dataset$, the new (privacy-enhanced) \gls{algorithm} $\algomap'$ produces 
learned \glspl{modelparam} by:
\begin{itemize}
	\item apply the pre-processing $\mathcal{I}(\dataset)$,
	\item compute $\algomap\big( \mathcal{I}(\dataset) \big)$ using the original algorithm,
	\item and finally apply the post-processing 
	$\mathcal{O} \big( \algomap\big( \mathcal{I}(\dataset) \big) \big)$, 
	yielding $\algomap'(\dataset)$.
\end{itemize}

{\bf Post-Processing.} Maybe the most widely used post-processing technique for 
\gls{diffpriv} is to add \emph{some noise} \cite{AlgoFoundDP},
\begin{equation} 
	\label{equ_def_add_some_noise}
	\mathcal{O}(\algomap) \defeq \algomap + \vn \mbox{, with noise } \vn = \big(n_{1},\ldots,n_{\nrnodes \featuredim} \big)^{T}\mbox{, } n_{1},\ldots,n_{\nrnodes\featuredim} \stackrel{\gls{iid}}{\sim} 
	p(n). 
\end{equation} 
Note that the post-processing \eqref{equ_def_add_some_noise} is parametrized 
by the choice of the \gls{probdist} $p(n)$ of the noise entries. Two important 
choices are the Laplacian distribution $p(n) \defeq \frac{1}{2b} \exp \big(- \frac{|n|}{b} \big)$ 
and the normal distribution $p(n) \defeq \frac{1}{\sqrt{2 \pi \sigma^2}} \exp \big(- \frac{n^2}{2 \sigma^2} \big)$ (i.e., 
using Gaussian noise $n \sim \mathcal{N}(0,\sigma^{2})$). 

When using Gaussian noise $n \sim \mathcal{N}(0,\sigma^2)$ in \eqref{equ_def_add_some_noise}, the variance 
$\sigma^2$ can be chosen based on the sensitivity
\begin{equation}
	\label{equ_def_l2_sens_A}
	\Delta_{2}\big( \algomap \big) \defeq \max_{\dataset,\dataset'} \normgeneric{\algomap(\dataset) - \algomap(\dataset')}{2}. 
\end{equation} 
Here, the maximum is over all pairs of neighbouring \gls{dataset}s $\dataset,\dataset'$. 
Adding Gaussian noise with standard deviation $\sigma  \geq \sqrt{2\ln(1.25/\delta)} \cdot \Delta_{2}(\algomap)/\varepsilon$ 
ensures that $\algomap$ is $(\varepsilon,\delta)$-\gls{diffpriv} \cite[Thm. 3.22]{AlgoFoundDP}. It might 
be difficult to evaluate the sensitivity \eqref{equ_def_l2_sens_A} for a given \gls{fl} 
algorithm $\algomap$ \cite{SensitiveAnalysisDeepNet}. For a \gls{gtvmin}-based method, 
i.e., $\algomap(\dataset)$ is a solution to \eqref{equ_def_gtvmin}, we can upper 
bound $\Delta_{2}\big( \algomap \big)$ via a perturbation analysis similar in spirit 
to the proof of Proposition \ref{prop_sensitiveiy_gtmvin_linreg}.

{\bf Pre-Processing.} Instead of ensuring \gls{diffpriv} via post-processing the output of an \gls{fl} algorithm $\algomap$, 
we can ensure \gls{diffpriv} by applying a pre-processing map $\mathcal{I}(\dataset)$ to the \gls{dataset} $\dataset$. 
The result of the pre-processing is a new \gls{dataset} $\widehat{\dataset} = \mathcal{I}(\dataset)$ which can be made 
available (publicly!) to any algorithm $\algomap$ that has no direct access to $\dataset$. According to Proposition \ref{prop_post_processing_dp}, 
as long as the pre-processing map $\mathcal{I}$ is $(\varepsilon,\delta)$-\gls{diffpriv} (see Definition \ref{equ_def_dp}), 
so will be the composition $\algomap \circ \mathcal{I}$. 

As for post-processing, one important approach to pre-processing is to ``add'' or ``inject'' noise. 
This results in a stochastic pre-processing map $\widehat{\dataset} = \mathcal{I}(\dataset)$ 
that is characterized by a \gls{probdist}. The noise mechanisms used for pre-processing 
might be different from just adding the \gls{realization} of a \gls{rv} (see \eqref{equ_def_add_some_noise}):
\footnote{Can you think of a simple pre-processing map that is deterministic and guarantees maximum \gls{diffpriv}?}
\begin{itemize} 
	\item For a \gls{classification} method with a discrete \gls{labelspace} $\labelspace =\{1,\ldots,\nrcategories\}$, 
	we can inject noise by replacing the true label of a \gls{datapoint} with a randomly selected element 
	of $\labelspace$ \cite[Mechanism 1]{pmlr-v202-busa-fekete23a}. The noise injection might also 
	include the replacement of the \gls{feature}s of a \gls{datapoint} by a \gls{realization} of a \gls{rv} 
	whose \gls{probdist} is somehow matched to the \gls{dataset} $\dataset$ \cite[Mechanism 2]{pmlr-v202-busa-fekete23a}. 
	
	\item Another form of noise injection is to construct $\mathcal{I}(\dataset)$ by randomly 
	selecting \gls{datapoint}s from the original (private) \gls{dataset} $\dataset$ \cite{PrivAmPSubsampling2018}. 
	Note that such noise injection is naturally provided by \gls{stochGD} methods 
	(see, e.g., step \ref{equ_random_sampling_fl_sgd} of Algorithm \ref{alg_fed_sgd}). 
\end{itemize} 

{\bf How To Be Sure?} Consider some algorithm $\algomap$, possibly obtained by pre- and 
post-processing techniques, that is claimed to be $(\varepsilon,\delta)$-\gls{diffpriv}. In practice, 
we might not know the detailed implementation of the algorithm. For example, we might not have 
access to the noise generation mechanism used in the pre- or post-processing steps. How can 
we verify a claim about \gls{diffpriv} of algorithm $\algomap$ without having access to the 
detailed implementation of $\algomap$? One approach could be to apply the algorithm to 
synthetic \gls{dataset}s $\dataset_{\rm syn}^{(1)},\ldots,\dataset_{\rm syn}^{(L)}$ that differ 
only in some private attribute of a single \gls{datapoint}. We can then try to predict the private 
attribute $\sensattr^{(\sampleidx)}$ of the \gls{dataset} $\dataset_{\rm syn}^{(\sampleidx)}$ by 
applying a learned \gls{hypothesis} $\widehat{\hypothesis}$ to the output $\algomap\big(\dataset_{\rm syn}^{(\sampleidx)} \big)$ 
delivered by the \emph{algorithm under test} $\algomap$. The \gls{hypothesis} $\widehat{\hypothesis}$ 
might be learned by an \gls{erm}-based method (see Algorithm \ref{alg_basic_ML_workflow}) using 
a \gls{trainset} consisting of pairs $\pair{\algomap\big(\dataset_{\rm syn}^{(\sampleidx)} \big)}{s^{(\sampleidx)}}$ 
for some $\sampleidx \in \{1,\ldots,L\}$.

\subsection{Private Feature Learning} 
\label{sec_private_feature_learning}

Section \ref{sec_diff_privacy} discussed pre-processing techniques that ensure \gls{diffpriv} 
of an \gls{fl} \gls{algorithm}. We next discuss pre-processing techniques that are not directly motivated 
from a \gls{diffpriv} perspective. Instead, we cast privacy-friendly pre-processing of a \gls{dataset} 
as a \gls{feature} learning problem \cite[Ch. 9]{MLBasics}. 

Consider a \gls{datapoint} characterized by a \gls{feature} vector $\featurevec \in \mathbb{R}^{\dimlocalmodel}$ 
and a \gls{label} $\truelabel \in \mathbb{R}$. Moreover, each \gls{datapoint} is characterized by a private  
attribute $\privattr$. We want to learn a (potentially stochastic) \gls{feature} map 
$\featuremapvec: \mathbb{R}^{\dimlocalmodel} \rightarrow \mathbb{R}^{\dimlocalmodel'}$ such that the 
new \gls{feature}s $\vz = \featuremapvec(\featurevec) \in \mathbb{R}^{\dimlocalmodel'}$ do not allow 
to accurately predict the private attribute $\privattr$. Trivially, we can make the accurate 
\gls{prediction} of $\privattr$ from $\featuremapvec(\featurevec)$ impossible by using 
a constant map, e.g., $\featuremapvec(\featurevec)=0$. However, we still want the 
new \gls{feature}s $\vz = \featuremapvec(\featurevec)$ to allow for a sufficiently 
accurate \gls{prediction} (using a suitable \gls{hypothesis}) of the \gls{label} $\truelabel$. 

{\bf Privacy Funnel.} To quantify the predictability of the private attribute $\privattr$ solely 
from the transformed \gls{feature}s $\vz=\featuremap(\featurevec)$ we can use the \gls{iidasspt} 
as a simple but useful \gls{probmodel}. Indeed, we can then use the \gls{mutualinformation} $\mutualinformation{\privattr}{\featuremapvec(\featurevec)}$ as a measure for the predictability of $\privattr$ from $\featuremapvec(\featurevec)$. A small value of 
$\mutualinformation{\privattr}{\featuremapvec(\featurevec)}$ indicates that it is difficult 
to predict the private attribute $\privattr$ solely from $\featuremapvec(\featurevec)$, i.e., a 
high level of privacy protection.\footnote{The relation 
	between \gls{mutualinformation}-based privacy measures and \gls{diffpriv} has been studied in some 
	detail recently \cite{DPasMIDP2016}.} Similarly, we can use the \gls{mutualinformation} $\mutualinformation{\truelabel}{\featuremapvec(\featurevec)}$ 
to measure the predictability of the label $\truelabel$ from $\featuremapvec(\featurevec)$. A large value 
$\mutualinformation{\truelabel}{\featuremapvec(\featurevec)}$ indicates that $\featuremapvec(\featurevec)$ 
allows to accurately predict $\truelabel$ (which is of course preferable). 

It seems natural to use a \gls{feature} map $\featuremapvec(\featurevec)$ that optimally 
balances a small $\mutualinformation{\privattr}{\featuremapvec(\featurevec)}$, i.e., a sufficiently large \gls{privprot},
with a sufficiently large $\mutualinformation{\truelabel}{\featuremapvec(\featurevec)}$ to allow for an 
accurate \gls{prediction} of $\truelabel$. The mathematically precise formulation of this plan is 
known as the \gls{privfunnel} \cite[Eq. (2)]{PrivacyFunnel}, 
\begin{equation}
	\label{equ_def_privacy_funnel}
	\min_{\featuremapvec(\cdot)}  \mutualinformation{\privattr}{\featuremapvec(\featurevec)} \mbox{ such that } \mutualinformation{\truelabel}{\featuremapvec(\featurevec)}\geq R.
\end{equation} 
Figure \ref{fig_illustrate_priv_funnel} illustrates the solution of \eqref{equ_def_privacy_funnel} for 
varying $R$, i.e., the minimum value of $\mutualinformation{\truelabel}{\featuremapvec(\featurevec)}$. 
\begin{figure}
	\begin{center} 
		\begin{tikzpicture}[yscale=1]
			\begin{axis}[
				xlabel={$\mutualinformation{\truelabel}{\featuremapvec(\featurevec)}$},
				ylabel={$\mutualinformation{\privattr}{\featuremapvec(\featurevec)}$},
				thick,
				ymax=4,
				ymin=0,
				xmin=0,
				xmax=7,
				domain=0:7,
				ylabel near ticks,
				xlabel near ticks,
				axis lines=left, 
				xtick=\empty, 
				ytick=\empty, 
				]
				\addplot[smooth, blue, mark=none] {4*exp(-0.3*(7-x))-4*exp(-0.3*(7))} node[pos=0.24, anchor=north west] {\hspace*{3mm}};
				
			\end{axis}
			\vspace*{-4mm}
		\end{tikzpicture}
		\vspace*{-10mm}
	\end{center} 
	\caption{\label{fig_illustrate_priv_funnel} The solutions of the \gls{privfunnel} \eqref{equ_def_privacy_funnel} 
		trace out (for varying constraint $R$) a curve in the plane spanned by the values of $\mutualinformation{\privattr}{\featuremapvec(\featurevec)}$ 
		(measuring the \gls{privleakage}) and $\mutualinformation{\truelabel}{\featuremapvec(\featurevec)}$ (measuring 
		the usefulness of the transformed \gls{feature}s for predicting the \gls{label}). }
\end{figure}

{\bf Optimal Private Linear Transformation.} 
The \gls{privfunnel} \eqref{equ_def_privacy_funnel} uses the \gls{mutualinformation} $\mutualinformation{\privattr}{\featuremapvec(\featurevec)}$ 
to quantify the \gls{privleakage} of a \gls{featuremap} $\featuremapvec(\featurevec)$. An 
alternative measure for the \gls{privleakage} is the minimum reconstruction error $\privattr - \hat{\privattr}$. 
The reconstruction $\hat{\privattr}$ is obtained by applying a reconstruction map $r(\cdot)$ to the 
transformed \gls{feature}s $\featuremapvec(\featurevec)$. If the joint \gls{probdist} $p(\privattr,\featurevec)$ 
is a \gls{mvndist} and the $\featuremapvec(\cdot)$ is a linear map (of the form $\featuremapvec(\featurevec) \defeq \mathbf{F} \featurevec$ 
with some matrix $\mathbf{F}$), then the optimal reconstruction map is again linear \cite{LC}. 

\begin{figure}[htbp]
	\begin{center}
		\begin{tikzpicture}[scale = 1.1]
			\node at (4.3,2) (mp) {\includegraphics[width=24mm]{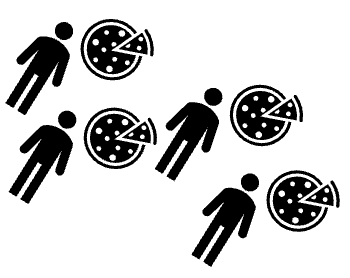}};
			\node at (1.0,1.0)(fa) {\includegraphics[width=24mm]{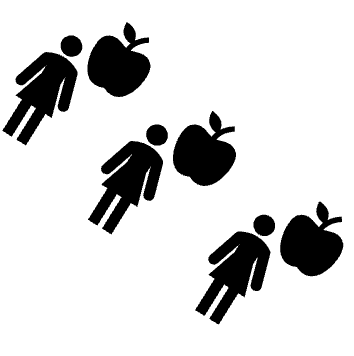}};
			\node at (3.5,0.0) (fp){\includegraphics[width=24mm]{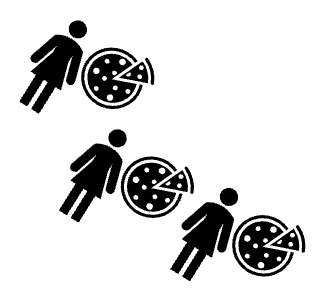}};
			\node at (2,3) (ma){\includegraphics[width=24mm]{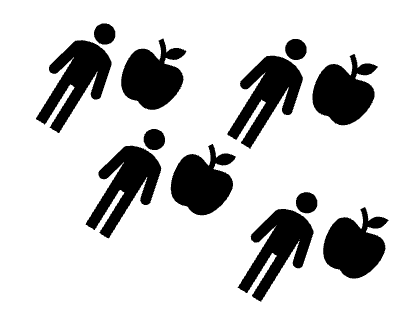}};
			\draw[<->, line width=1mm, color=green!50!black] (3.0,3.9) -- (4.4,3.0) node[above=9mm](y) {\hspace*{15mm} food preference $\truelabel$}; 
			\draw[<-, line width=1mm, color=green!50!black] (-0.5,-0.3) -- (0.9,-1.2) node[above=0mm](y1) {\hspace*{0mm} $\mathbf{f}$}; 
			\draw[<->, line width=1mm, color=red] (4.7,0.5) -- (5.7,1.9) node[below=6mm](s) {\hspace*{20mm} gender $\privattr$}; 
			\draw[->, very thick,color=blue!60!black] (0,0.3) -- (6.5,0.3) node[right](x) {$\feature_1$};       
			\draw[->, very thick,color=blue!60!black] (0.3,0) -- (0.3,4.2) node[above] {$\feature_2$};   
		\end{tikzpicture}
		\vspace*{-3mm}
	\end{center}
	\vspace*{-5mm}
	\caption{A toy \gls{dataset} $\dataset$ whose \gls{datapoint}s represent customers, each 
		characterized by \gls{feature}s $\featurevec = \big(\feature_{1},\feature_{2}\big)^{T}$. These raw features carry 
		information about a \gls{sensattr} $\privattr$ (gender) and the \gls{label} $\truelabel$ (food preference) of 
		a person. The \gls{scatterplot} that we can find a linear \gls{feature} transformation $\mathbf{F} \defeq \mathbf{f}^{T} \in \mathbb{R}^{1 \times 2}$ 
		resulting in a new \gls{feature} $z \defeq \mathbf{F} \featurevec$ that does not allow to predict $\privattr$, 
		while still allowing to predict $\truelabel$. \label{fig_pp_feature_learning_lin}}
\end{figure}

We would like to find the linear \gls{featuremap} $\featuremapvec(\featurevec) \defeq \mathbf{F} \featurevec$ such 
that for any linear reconstruction map $\mathbf{r}$ (resulting in $\hat{s} \defeq \mathbf{r}^{T} \mathbf{F} \featurevec$) 
the expected squared error $\expect \{ (s - \hat{s})^2 \}$ is large. The smallest possible expected \gls{sqerrloss} 
\begin{equation} 
	\nonumber
	\varepsilon(\mathbf{F}) \defeq \min_{\mathbf{r} \in \mathbb{R}^{\dimlocalmodel'}} \expect \{ (s - \mathbf{r}^{T} \mathbf{F} \featurevec)^2 \}
\end{equation}
measures the level of \gls{privprot} offered by the new \glspl{feature} $\vz = \mathbf{F} \featurevec$. 
The larger the value $\varepsilon(\mathbf{F})$, the more \gls{privprot} is offered. 
It can be shown that $\varepsilon(\mathbf{F})$ is maximized by any $\mathbf{F}$ that 
is orthogonal to the cross-covariance vector $\mathbf{c}_{\featurevec,s} \defeq \expect \{ \featurevec s \}$, i.e., 
whenever $\mathbf{F}\mathbf{c}_{\featurevec,s} = \mathbf{0}$. One specific choice 
for $\mathbf{F}$ that satisfies this orthogonality condition is 
\begin{equation} 
	\label{equ_pp_linear_feature_map} 
	\mathbf{F}=\mathbf{I} - (1/\normgeneric{\mathbf{c}_{\featurevec,s}}{2}^{2}) \mathbf{c}_{\featurevec,s} \mathbf{c}_{\featurevec,s}^{T}. 
\end{equation} 
Figure \ref{fig_pp_feature_learning_lin} illustrates a \gls{dataset} for which we want 
to find a linear \gls{featuremap} $\mathbf{F}$ such that the new \glspl{feature} $\vz = \mathbf{F} \featurevec$ 
do not allow to accurately predict a \gls{sensattr}.   

\clearpage
\subsection{Exercises}

\refstepcounter{problem}\label{prob:whereisalice}\textbf{\theproblem. Where is \emph{Alice}?}
Consider a \gls{device}, named \emph{Alice}, that implements an asynchronous variant of 
Algorithm \ref{alg_fed_gd_general} (see \eqref{equ_full_update_async_generic} and \eqref{equ_fixed_point_asynch_generic_gd}).
The \gls{localdataset} of the \gls{device} consists of temperature measurements obtained from 
some \gls{fmi} weather station. Assuming that no other \gls{device} interacts with \emph{Alice} 
except for your \gls{device}, named \emph{Bob}. Develop a software for \emph{Bob} that interacts 
with \emph{Alice}, according to \eqref{equ_full_update_async_generic}, in order to determine at 
which \gls{fmi} station we can find \emph{Alice}. 

\noindent\refstepcounter{problem}\label{prob:privatefda}\textbf{\theproblem. Linear discriminant analysis with \gls{privprot}.}
Consider a binary \gls{classification} problem with \gls{datapoint}s characterized by a \gls{featurevec} $\featurevec \in \mathbb{R}^{\nrfeatures}$ 
and a binary \gls{label} $\truelabel \in \{-1,1\}$. Each \gls{datapoint} has a \gls{sensattr} $\sensattr = \mathbf{F} \featurevec$, obtained 
by applying a fixed matrix $\mathbf{F}$ to the \gls{featurevec} $\featurevec$. 
We use a \gls{probmodel} - interpreting \gls{datapoint}s $\pair{\featurevec}{\truelabel}$ as \gls{iid} 
\gls{realization}s of a \gls{rv} - with the \gls{featurevec} having \gls{mvndist} $\mvnormal{{\bf \mu}^{(\truelabel)}}{\covmtx{\truelabel}}$ 
conditioned on $\truelabel$. The \gls{label} is uniformly distributed over the \gls{labelspace} $\{-1,1\}$. 
Try to find a vector $\va$ such that the transformed \gls{featurevec} $z' \defeq \va^{T} \featurevec$ optimally 
balances the \gls{privleakage} (information carried by $z'$ about $\sensattr$) with the information carried 
by $z'$ about the \gls{label} $\truelabel$. 

\noindent\refstepcounter{problem}\label{prob:whereareyou}
\textbf{\theproblem. Where Are You?}
Consider a social media post of a friend that is travelling across Finland. This post includes 
a snapshot of a temperature measurement and a clock. Can you guess the latitude and 
longitude of the location where your friend took this snapshot? We can use \gls{erm} to do 
this: Use Algorithm \ref{alg_basic_ML_workflow} to learn a vector-valued \gls{hypothesis} 
$\widehat{\hypothesis}$ for predicting latitude and longitude from the time and value 
of a temperature measurement. Use the weather recordings at \gls{fmi} stations to construct 
a \gls{trainset} and a \gls{valset}.  

\noindent\refstepcounter{problem}\label{prob:whereareyouprepro}\textbf{\theproblem. Ensuring Privacy with Pre-Processing.}
Repeat the privacy \gls{attack} described in Exercise \ref{prob:whereareyou} but this time 
using a pre-processed version of the raw \gls{data}. The pre-processing can be implemented either via 
randomly selecting a subset of \glspl{datapoint} in the raw \gls{dataset} or by adding 
noise to their \glspl{feature} and \glspl{label}. How well can one predict the latitude and 
longitude from the time and value of a temperature measurement using a \gls{hypothesis} 
$\widehat{\mathbf{h}}$ learned from the perturbed \gls{data}? 

\noindent\refstepcounter{problem}\label{prob:privfeaturelearningFMI}\textbf{\theproblem. Private Feature Learning.}
Download hourly weather observations during April $2023$ at FMI station \emph{ Kustavi Isokari}. 
You can access these observations here \url{https://en.ilmatieteenlaitos.fi/download-observations}. 
Each time period of one hour corresponds to a \gls{datapoint} that is characterized 
by the following \gls{feature}s: 
\begin{itemize} 
	\item $\feature_{1}=$ Average temperature [°C]
	\item $\feature_{2}=$ Maximum temperature [°C]
	\item $\feature_{3}=$ Minimum temperature [°C]
	\item $\feature_{4}=$ Average relative humidity [\%],
	\item $\feature_{5}=$ Wind speed [m/s],
	\item $\feature_{6}=$ Maximum wind speed [m/s],
	\item $\feature_{7}=$ Average wind direction [°],
	\item $\feature_{8}=$ Maximum gust speed [m/s],
	\item $\feature_{9}=$ Precipitation [mm], 
	\item $\feature_{10}=$ Average air pressure [hPa]
	\item $\feature_{11}=$ hour of the day ($1,\ldots,24$).
\end{itemize}
The goal of this exercise is to learn a linear \gls{feature} transformation $ \mathbf{z} = \mathbf{F} \featurevec$ 
such that the new \gls{feature}s do not allow to recover the hour of the day $\feature_{11}$ (which is considered a 
private attribute $\privattr$ of the \gls{datapoint}). However the new \gls{feature}s should still allow 
to reconstruct the average temperature $\feature_{1}$. 

We construct the matrix $\mathbf{F}$ according 
to \eqref{equ_pp_linear_feature_map} by replacing the exact cross-covariance vector $\vc_{\featurevec,s}$ 
with an estimate (or approximation) $\hat{\vc}_{\featurevec,s}$. 
This estimate is computed as follows: 
\begin{enumerate} 
	\item read all \gls{datapoint}s and construct a \gls{featuremtx} $\featuremtx \in \mathbb{R}^{\samplesize \times 11}$ 
	with $\samplesize$ being the total number of \gls{datapoint}s 
	\item remove the sample means from each \gls{feature}, resulting in the centred \gls{featuremtx} 
	\begin{equation} 
		\nonumber
		\widehat{\featuremtx} \defeq  \featuremtx - (1/\samplesize) \mathbf{1} \mathbf{1}^{T} \featuremtx \mbox{ , } \mathbf{1} \defeq \big(1,\ldots,1 \big)^{T} \in \mathbb{R}^{\samplesize}. 
	\end{equation} 
	\item extract the \gls{sensattr} or each \gls{datapoint} and store it in the vector 
	\begin{equation}
		\nonumber
		\vs \defeq  \big( \hat{x}^{(1)}_{1},  \hat{x}^{(2)}_{1},  \ldots, \hat{x}^{(\samplesize)}_{1} \big)^{T}. 
	\end{equation} 
	\item compute the empirical cross-covariance vector 
	\begin{equation}
		\nonumber
		\hat{\vc}_{\featurevec,s} \defeq (1/\samplesize) \big( \widehat{\featuremtx}  \big)^{T} \vs 
	\end{equation} 
\end{enumerate}
The matrix $\mathbf{F}$ obtained from \eqref{equ_pp_linear_feature_map} by replacing 
$\vc_{\featurevec,s}$ with $\hat{\vc}_{\featurevec,s}$, is then used to compute 
the privacy-preserving \gls{feature}s $\vz^{(\sampleidx)} = \mathbf{F} \featurevec^{(\sampleidx)}$ for $\sampleidx=1,\ldots,\samplesize$. 
To verify if these new \gls{feature}s are indeed privacy-preserving, we use \gls{linreg} (as implemented by the \texttt{LinearRegression} class of 
the Python package \texttt{scikit-learn}) to learn the \glspl{modelparam} of a \gls{linmodel} to predict the 
\gls{sensattr} $s^{(\sampleidx)} = \feature^{(\sampleidx)}_{1}$ (the hour of the day during which 
the measurement has been taken) from the \gls{feature}s $\mathbf{z}^{(\sampleidx)}$.

\clearpage
\section{Cybersecurity in FL: Attacks and Defenses}
\label{lec_datapoisoning}

\gls{fl}, like \gls{ml} more broadly, fundamentally relies on externally 
provided \gls{data}. In most \gls{ml} applications, the computational \gls{device} 
that trains a \gls{model} rarely has direct access to the raw \glspl{datapoint} 
of the \gls{trainset}. Instead, training often proceeds on pre-processed \gls{data} 
supplied by external sources or curated databases.

\begin{figure}[h]
	\centering
	\fbox{%
		\includegraphics[width=0.6\textwidth]{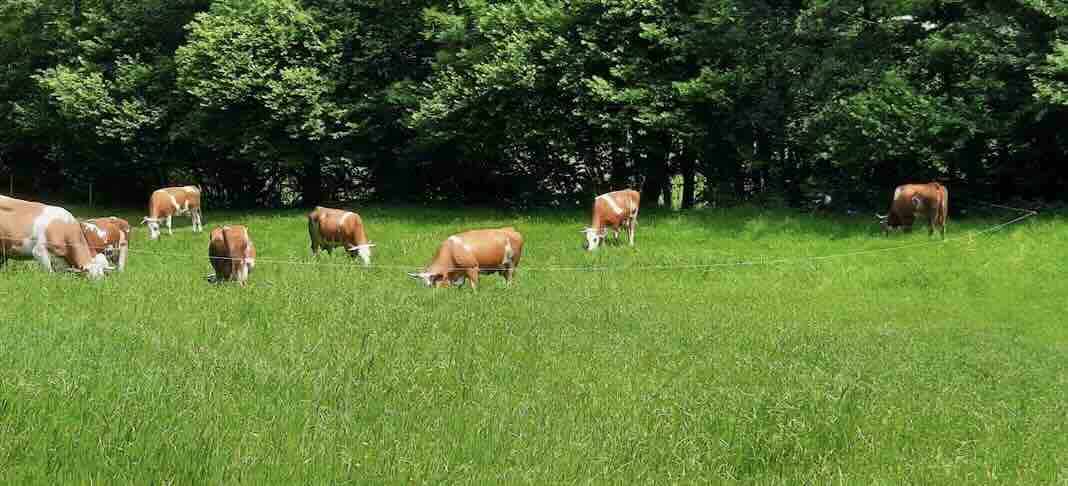}
	}
	\caption{In many \gls{ml} applications, such as monitoring livestock in remote regions, 
		direct access to raw \glspl{datapoint} is impractical. \gls{ml} methods often rely on external 
		databases curated by third parties, introducing potential vulnerabilities.}
	\label{fig:cow_herd}
\end{figure}

As a case in point, consider an \gls{ml} application for animal health-care 
based on monitoring livestock in remote regions. Direct access to raw \glspl{datapoint}, 
such as those depicted in Figure \ref{fig:cow_herd}, would require physically 
visiting distant pastures with specialized measurement equipment such as stomach sensors. 
Instead, developers typically rely on external databases assembled by 
researchers or veterinarians who collected the \gls{data} on-site.

This reliance on external \gls{data} is even more pronounced in \gls{fl} systems.  
One of the primary purposes of \gls{fl} is to leverage the information contained in 
the \glspl{localdataset} of many interconnected \glspl{device}, which form a \gls{flnetwork}.  
However, this raises a critical question:  
\emph{How can we be confident that every \gls{device} behaves as intended and faithfully 
	follows the agreed-upon \gls{fl} \gls{algorithm}?}

Except in the rare case where we have full control over every \gls{device} in 
the \gls{flnetwork}, it is essential to design \gls{fl} systems that are robust 
against potential \glspl{attack}. Here, an \gls{attack} refers to the intentional 
perturbation (or manipulation) of \gls{fl} system parts. 

This chapter is structured as follows: Section \ref{sec_a_simple_attack_model} 
discusses how such \glspl{attack} can be carried out by perturbing different components of an \gls{fl} system. 
Section \ref{sec_attack_types} distinguishes different \gls{attack} types according to their 
objectives. Section \ref{sec_design_choices_fl_attacks} provides some guidance on the 
design choices for \gls{gtvmin}-based methods to ensure robustness against attacks.

\subsection{A Simple Attack Model} 
\label{sec_a_simple_attack_model}

Consider an \gls{fl} system that implements one of the \gls{fl} \glspl{algorithm} discussed in Chapter \ref{lec_flalgorithms}. 
As discussed in Section \ref{sec_unified_form_algos}, these \glspl{algorithm} share a common form. 
Many widely-used \gls{fl} \glspl{algorithm} for parametric \glspl{localmodel} (with \glspl{modelparam} 
belonging to $\mathbb{R}^{\dimlocalmodel}$) compute and share the results (across the 
edges of the \gls{flnetwork}) of local updates 
\begin{equation} 
	\label{equ_update_attack_model}
\localparamsiter{\nodeidx}{\iteridx+1} = \argmin_{\localparams{\nodeidx} \in \mathbb{R}^{\dimlocalmodel}} 
\left[ \locallossfunc{\nodeidx}{\localparams{\nodeidx}} \!+\! \sum_{\nodeidx' \in \neighbourhood{\nodeidx}} \edgeweight_{\nodeidx,\nodeidx'} \gtvpenalty\big( \localparamsiter{\nodeidx'}{\iteridx} - \localparamsiter{\nodeidx}{\iteridx}\big) \right].
\end{equation}
	\begin{figure}[H]
	\begin{center}
		\begin{tikzpicture}
			\node[fill=black, circle, inner sep=1pt, label=left:{}] (d1) at (0.2,0.5) {};
			\node[fill=black, circle, inner sep=1pt, label=left:{}] (d2) at (-0.5,0) {};
			\node[fill=black, circle, inner sep=1pt, label=left:{}] (d3) at (0,-0.5) {};
			\node[above = 0.0cm of d2] (dataset) {$\localdataset{\nodeidx}$};
			\node[draw, ellipse, dashed, minimum width=2cm, minimum height=2cm, fit=(d1) (d2) (d3), inner sep=0pt] (ellipse) {};
			\node[] (update) at (3.5,0) {\eqref{equ_update_attack_model}};
			\node[right=0.5cm of update, text width=4cm, align=left,yshift=3pt] (out) 
			{$\localparamsiter{\nodeidx}{\iteridx+1}$};
			\begin{scope}
				\node[draw=gray, rounded corners, fit=(d1) (d2) (d3) (update) (out), inner sep=1.2em, xshift=-1cm, label=below:{\small device $\nodeidx$}] (devicebox) {};
			\end{scope}
			
			\draw[->,thick] (ellipse) -- (update) node[midway, below] {$\locallossfunc{\nodeidx}{\cdot}$};
			\draw[->,thick] (update) -- (out);
			
			\node[fill=black, above=1.2cm of d1, circle, inner sep=1.5pt, label=above:{$\localparamsiter{\nodeidx'}{\iteridx}$}] (np1)  {};
			\node[fill=black, above=1.2cm of out, circle, inner sep=1.5pt, label=above:{$\nodeidx'\in \neighbourhood{\nodeidx}$}] (np1_future)  {};
			\draw[->,thick] ($(out)+(-0.5,0)$) -- (np1_future);
			\draw[->, thick] (np1) -- (update);
		\end{tikzpicture}
	\end{center}
	\caption{A \gls{gtvmin}-based \gls{fl} system from the perspective of a specific \gls{device} $\nodeidx$. \label{fig_update_attack_model}}
\end{figure}  

During time-instant $\iteridx$, \gls{device} $\nodeidx$ solves \eqref{equ_update_attack_model}
in order to obtain new \glspl{modelparam} $\localparamsiter{\nodeidx}{\iteridx+1}$. Carefully 
note that update \eqref{equ_update_attack_model} involves the \glspl{modelparam} $\localparamsiter{\nodeidx'}{\iteridx}$ 
at \glspl{neighbor} $\nodeidx' \in \neighbourhood{\nodeidx'}$. In practice, these \glspl{modelparam} need 
to be communicated over some physical channel (e.g., a short-range wireless link) between \gls{device} 
$\nodeidx$ and \gls{device} $\nodeidx'$. Figure \ref{fig_update_attack_model} illustrates the information 
flow during the local update \eqref{equ_update_attack_model}. 

From the viewpoint of a specific \gls{device} $\nodeidx$, control is typically limited to the local \gls{lossfunc} $\locallossfunc{\nodeidx}{\localparams{\nodeidx}}$, which is often computed as the average \gls{loss} 
over the \gls{localdataset}.\footnote{This assumption may not always hold in practice—for instance, the \gls{fl} 
	application might not be granted full access to the operating system of device $\nodeidx$ (e.g., a 
	smartphone).} In contrast, the \glspl{modelparam} $\localparamsiter{\nodeidx'}{\iteridx}$ received 
	from neighbouring \glspl{device} may be unreliable: they can be intentionally perturbed (or poisoned). 
	In what follows, we describe two major classes of \glspl{attack} that exploit different parts of the \gls{fl} 
	system to manipulate the shared \glspl{modelparam} $\localparamsiter{\nodeidx'}{\iteridx}$ and 
	thereby influence the local update step \eqref{equ_update_attack_model}.

\subsubsection{Model Poisoning} 
If an attacker has control over some of the communication links within an \gls{fl} system, 
it can directly manipulate the \glspl{modelparam} shared between nodes. A \gls{model} 
poisoning \gls{attack} on the update \eqref{equ_update_attack_model} replaces the vector 
$\localparamsiter{\nodeidx'}{\iteridx}$, for some $\nodeidx' \in \neighbourhood{\nodeidx}$ 
with a perturbed vector $\widetilde{\weights}^{(\nodeidx',\iteridx)}$. We have already 
discussed the \gls{robustness} of the update \eqref{equ_update_attack_model}, for specific 
choices of $\gtvpenalty$, against perturbations in Section \ref{sec_robust_proximal_step}.

\subsubsection{Data Poisoning} 

Consider an attacker with access to the \gls{localdataset}s of a subset of \glspl{device} 
$\mathcal{W} \subset \nodes$ in the \gls{flnetwork}. By poisoning the \glspl{localdataset} 
at these compromised nodes, the attacker can manipulate the corresponding local 
updates \eqref{equ_update_attack_model}. Protecting a given \gls{device} $\nodeidx$ from 
such poisoning is non-trivial, especially when the attacker can exploit software vulnerabilities, 
such as those in smartphone operating systems \cite{Mohamed2017}.

The impact of the poisoned updates propagates from nodes $\nodeidx' \in \mathcal{W}$ through 
the edges of the \gls{flnetwork} during successive update steps. As a result, even nodes whose 
\glspl{localdataset} remain clean can eventually be affected -- provided they are connected to 
$\mathcal{W}$. In fact, if the \gls{flnetwork} $\graph$ is connected, the influence of poisoned 
updates can reach all nodes within a number of steps proportional to the \gls{graph}'s diameter.

Figure \ref{fig:chain_poison_propagation} illustrates this phenomenon in a chain-structured \gls{flnetwork} 
with three nodes $\nodeidx = 1, 2, 3$ connected by unit-weight edges $\edges = { \edge{1}{2}, \edge{2}{3} }$. 
The attacker poisons the \gls{localdataset} $\localdataset{1}$ at node $\nodeidx=1$ at time $\iteridx - 1$, 
resulting in a perturbed update at time $\iteridx$. This perturbation influences node $\nodeidx=2$ at time 
$\iteridx+1$, and subsequently node $\nodeidx=3$ at time $\iteridx+2$. The affected updates are marked 
by a red star (\textcolor{red}{$*$}) in Figure \ref{fig:chain_poison_propagation}.
\begin{figure}[h]
	\centering
\begin{tikzpicture}[scale=1.2, every node/.style={font=\small}, thick]
	
	\node at (1.0,3.2) {$\iteridx$};
	\node at (5.0,3.2) {$\iteridx{+}1$};
	\node at (9.0,3.2) {$\iteridx{+}2$};
	
	\node[draw, circle, fill=black, minimum size=5pt, inner sep=0pt] (N1t) at (1,2.5) {};
	\node[draw, circle, fill=black, minimum size=5pt, inner sep=0pt] (N2t) at (1,1.5) {};
	\node[draw, circle, fill=black, minimum size=5pt, inner sep=0pt] (N3t) at (1,0.5) {};
	
	\node[left=2pt of N1t] {$1$};
	\node[left=2pt of N2t] {$2$};
	\node[left=2pt of N3t] {$3$};
	
	\node[draw, circle, fill=black, minimum size=5pt, inner sep=0pt] (N1tp1) at (5,2.5) {};
	\node[draw, circle, fill=black, minimum size=5pt, inner sep=0pt] (N2tp1) at (5,1.5) {};
	\node[draw, circle, fill=black, minimum size=5pt, inner sep=0pt] (N3tp1) at (5,0.5) {};
	
	\node[draw, circle, fill=black, minimum size=5pt, inner sep=0pt] (N1tp2) at (9,2.5) {};
	\node[draw, circle, fill=black, minimum size=5pt, inner sep=0pt] (N2tp2) at (9,1.5) {};
	\node[draw, circle, fill=black, minimum size=5pt, inner sep=0pt] (N3tp2) at (9,0.5) {};
	
	\draw[->,dashed,thin] (N1t) -- (N1tp1);
	\draw[->,dashed,thin] (N1t) -- (N2tp1);
	\draw[->,dashed,thin] (N2t) -- (N2tp1);
	\draw[->,dashed,thin] (N2t) -- (N3tp1);
	\draw[->,dashed,thin] (N3t) -- (N3tp1);
	
	\draw[->,dashed,thin] (N1tp1) -- (N1tp2);
	\draw[->,dashed,thin] (N1tp1) -- (N2tp2);
	\draw[->,dashed,thin] (N2tp1) -- (N2tp2);
	\draw[->,dashed,thin] (N2tp1) -- (N3tp2);
	\draw[->,dashed,thin] (N3tp1) -- (N3tp2);
	
	\draw[-, thick] (N1t) -- (N2t);
	\draw[-, thick] (N2t) -- (N3t);
	
	\draw[-, thick] (N1tp1) -- (N2tp1);
	\draw[-, thick] (N2tp1) -- (N3tp1);
	
	\draw[-, thick] (N1tp2) -- (N2tp2);
	\draw[-, thick] (N2tp2) -- (N3tp2);
	
	\node[red] at (0.7,2.8) {*};  
	\node[red] at (4.7,2.8) {*};  
	\node[red] at (4.7,1.8) {*};  
	\node[red] at (8.7,2.8) {*};  
	\node[red] at (8.7,1.8) {*};  
	\node[red] at (8.7,0.8) {*};  
	
\end{tikzpicture}
	\caption{Propagation of the effect of a \gls{datapoisoning} \gls{attack} that perturbs the 
		update \eqref{equ_update_attack_model} of $\nodeidx=1$ during time $\iteridx$. 
	\label{fig:chain_poison_propagation}}
\end{figure}
 
\Gls{datapoisoning} can consist of adding the \gls{realization} of \gls{rv}s to the \gls{feature}s 
and \gls{label} of a \gls{datapoint}: We poison a \gls{datapoint} 
by replacing its \gls{feature}s $\featurevec$ and \gls{label} $\truelabel$ with $\widetilde{\featurevec} \defeq \featurevec + \Delta \featurevec$ 
and $\tilde{\truelabel} = \truelabel + \Delta \truelabel$. 

For \gls{fl} applications with \gls{localmodel}s being used for \gls{classification} of \gls{datapoint}s with 
a discrete \gls{label} (or category), we further distinguish between the following \gls{datapoisoning} 
strategies \cite{turner2019cleanlabel}: 
\begin{itemize} 
	\item {\bf Label Poisoning.} The attacker manipulates the \gls{label}s of \gls{datapoint}s in the \gls{trainset}. 
	\item {\bf Clean-Label \gls{attack}.} The attacker leaves the \gls{label}s untouched and 
	only manipulates the \gls{feature}s of \gls{datapoint}s in the \gls{trainset}. 
\end{itemize} 

The effect \gls{datapoisoning} is that the original local \gls{lossfunc}s $\locallossfunc{\nodeidx}{\cdot}$ 
in \gls{gtvmin} \eqref{equ_def_gtvmin_linreg} are replaced by perturbed local \gls{lossfunc}s 
$\perturbedlocallossfunc{\nodeidx}{\cdot}$. The degree of perturbation depends on the fraction 
of poisoned \gls{datapoint}s as well as the choice of the \gls{lossfunc} used to measure the 
\gls{prediction} error.

Different \gls{lossfunc}s provide varying levels of robustness against \gls{datapoisoning}. 
For example, using the \gls{abserr} yields increased robustness against perturbations 
of the \gls{label} values of a few \gls{datapoint}s, compared to the \gls{sqerrloss} 
(see Exercise \ref{prob_abserrlossrobust}). Another class of robust \gls{lossfunc}s is 
obtained by including a penalty term (as in \gls{regularization}). 

\subsection{\Gls{attack} Types} 
\label{sec_attack_types}

Based on their objective, we distinguish the following \glspl{attack} on \gls{fl} systems: 
\glspl{dosattack}, \gls{backdoor} \glspl{attack} and privacy (or \gls{modelinversion}) \glspl{attack} \cite{vassilev2024adversarial}.

\begin{itemize}
	\item \textbf{\Gls{dosattack}.} A \gls{dosattack} manipulates $\localparamsiter{\nodeidx'}{\iteridx}$ 
	in \eqref{equ_update_attack_model} to nudge the updates $\localparamsiter{\nodeidx}{\iteridx+1}$ 
	 towards \glspl{modelparam} $\overline{\weights}^{(\nodeidx)}$ with a large local \gls{loss}. 
	 In other words, the resulting \gls{hypothesis} $\bar{\hypothesis}^{(\nodeidx)}$ 
	 delivers poor \glspl{prediction} for the \glspl{datapoint} in the \gls{localdataset} $\localdataset{\nodeidx}$ 
	 (see Figure \ref{fig_dos_backdoor}) \cite{NIPS2017_f4b9ec30}.
	
	\item \textbf{\Gls{backdoor} \gls{attack}.} This \gls{attack} manipulates $\localparamsiter{\nodeidx'}{\iteridx}$ 
	in \eqref{equ_update_attack_model} to nudge the updates $\localparamsiter{\nodeidx}{\iteridx+1}$ 
	towards \glspl{modelparam} $\widetilde{\weights}^{(\nodeidx)}$ with a small \gls{loss} on the 
	\gls{localdataset} but highly irregular \glspl{prediction} for specific \glspl{featurevec}. 
	In other words, the \gls{hypothesis} $\tilde{\hypothesis}^{(\nodeidx)}$ ``behaves well'' 
	on $\localdataset{\nodeidx}$ but delivers pre-specified \glspl{prediction} on 
	a subset $\mathcal{K} \subseteq \featurespace$ of the \gls{featurespace}. We can 
	interpret the subset $\mathcal{K}$ as a backdoor which is opened by any \gls{datapoint} 
	with a \gls{featurevec} $\featurevec \in \mathcal{K}$ (see Figure \ref{fig_dos_backdoor}) \cite{pmlr-v108-bagdasaryan20a}.
	
	\item \textbf{Privacy (or \gls{modelinversion}) \gls{attack}.} This \gls{attack} manipulates 
	$\localparamsiter{\nodeidx'}{\iteridx}$ in \eqref{equ_update_attack_model} such that 
	the updates $\localparamsiter{\nodeidx}{\iteridx+1}$ maximally leak information 
	about \gls{sensattr}s of \gls{datapoint}s stored at \gls{device} $\nodeidx$. 
	One approach is to force another \gls{device} $\nodeidx'$ to learn a copy of \glspl{modelparam} $\localparams{\nodeidx}$ by 
	designing trivial local \glspl{lossfunc} and manipulating the structure of the \gls{flnetwork} 
	(see Exercise \ref{prob:whereisalice}). Once obtained, the copied \glspl{modelparam} can be 
	probed to reveal private information. A notable class of privacy \gls{attack}s is \gls{modelinversion}, 
	where an \gls{attack}er tries to reconstruct \gls{featurevec}s of \glspl{datapoint} \cite{Fredrikson2015}.  
\end{itemize}

\begin{figure}        
	\begin{tikzpicture}[scale = 1.3]
		\draw[->, very thick] (0,0.3) -- (6.5,0.3) node[right](x) {features $\featurevec$};       
		\draw[->, very thick] (0.3,0) -- (0.3,4.2) node[above] {label $\truelabel$};   

		\coordinate (l1)   at (0.8, 0.8*0.75+0.3);
		\coordinate (l2) at (0.9, 0.9*0.75+0.3);
		\coordinate (l3)   at (0.8,  4.3);
		\coordinate (l4)   at (0.9, 4.3);
		
		\draw[dashed, color=blue, very thick, scale =1, domain = -1:6, variable = \x]  plot ({\x},{\x*0.6+0.7}) node [right=0.2cm, font=\fontsize{15}{0}\selectfont] {$\hat{\hypothesis}^{(\nodeidx)}(\featurevec)$ (no \gls{attack})} ;   
		
		\draw[color=red, very thick, scale =1, domain = -1:6, variable = \x]  plot ({\x},{\x*0+3}) node [right=0.2cm, font=\fontsize{15}{0}\selectfont] {$\bar{\hypothesis}^{(\nodeidx)}(\featurevec)$ (\gls{dosattack})} ;   
		
		\draw[color=black, very thick, scale =1, domain = -0.5:0.8, variable = \x]  plot ({\x},{\x*0.75+0.3});       
		\draw[color=black, very thick, scale =1, domain = 0.9:6, variable = \x]  plot ({\x},{\x*0.75+0.3});   
		
		\coordinate (l5) at (5.8,4.5) ; 
		\node[above right=0.5 of l5, font=\fontsize{15}{0}\selectfont]{$\tilde{\hypothesis}^{(\nodeidx)}(\featurevec)$ (\gls{backdoor} \gls{attack})};
		
		\draw[color=black,very thick] (l1) -- (l3);  
		\draw[color=black,very thick] (l3) -- (l4);  
		\draw[color=black,very thick] (l2) -- (l4);  
		\foreach \l in {3,...,22}
		{
			\pgfmathsetseed{\l+2}
			\coordinate (mycenterpoint) at (${(.1*\l+rand}*(0.9,0.2)+{rnd*1.3+2.5+0.2*\l}*(0.2,0.5)+(1.4,0)$);
			\draw[fill=blue] (mycenterpoint) circle (0.13cm) node[] {};
		}
		\begin{scope}[shift={(2.9,2.4))},x={(1,1)},y={($(0,1)!1!90:(-1,1)$)}]
			\draw[red,fill=blue!90!white, opacity=0.3] (.5,0) ellipse (1.9 and 1.2);
		\end{scope}
		
		\coordinate (head) at ($(0.1,-0.1)+(l2)$);
		\coordinate (tail) at ($(0.5,-0.5)+(l2)$);
		\draw[<-, very thick, color=black] (head) -- (tail) node[right] {``backdoor"};  
		\node[above left=1 of x, font=\fontsize{15}{0}\selectfont]{local dataset $\localdataset{\nodeidx}$};
		
	\end{tikzpicture}
	\caption{A \gls{localdataset} $\localdataset{\nodeidx}$ along with three \gls{hypothesis} maps learned via 
		iterating \eqref{equ_update_attack_model} under three \gls{attack} scenarios. 
		\label{fig_dos_backdoor}}
\end{figure}

\clearpage
\subsection{Making \gls{fl} Robust Against \Glspl{attack}} 
\label{sec_design_choices_fl_attacks}

We next discuss how to make the update \eqref{equ_update_attack_model} more robust 
against the \glspl{attack} discussed in Section \ref{sec_attack_types}. Our focus will be on 
\gls{gtvmin}-based methods using the \gls{gtv} penalty $\gtvpenalty(\cdot) = \normgeneric{\cdot}{2}^{2}$. 
For this choice, \eqref{equ_update_attack_model} can be written as (see \eqref{equ_robust_update_for_squared_norm}) 
\begin{align}
	\label{equ_defence_update_perturb_sq_norm}
	\localparamsiter{\nodeidx}{\iteridx+1} & \in \argmin_{\localparams{\nodeidx} \in \mathbb{R}^{\dimlocalmodel}} \locallossfunc{\nodeidx}{\localparams{\nodeidx}} + \regparam \nodedegree{\nodeidx}
	\normgeneric{\localparams{\nodeidx}-\estlocalparams{\neighbourhood{\nodeidx}}}{2}^{2}, \nonumber \\  
	& \mbox{ with } 
	\estlocalparams{\neighbourhood{\nodeidx}}  \defeq (1/\nodedegree{\nodeidx}) \sum_{\nodeidx' \in \neighbourhood{\nodeidx}} \edgeweight_{\nodeidx,\nodeidx'} \localparamsiter{\nodeidx'}{\iteridx}.
\end{align} 
Here, we used the weighted \gls{nodedegree} 
$\nodedegree{\nodeidx} = \sum_{\nodeidx' \in \neighbourhood{\nodeidx}} \edgeweight_{\nodeidx,\nodeidx'}$ (see \eqref{equ_def_node_degree}). 

The update \eqref{equ_defence_update_perturb_sq_norm} can be attacked via manipulating the 
\glspl{modelparam} $ \localparamsiter{\nodeidx'}{\iteridx}$ and, in turn, their average $\estlocalparams{\neighbourhood{\nodeidx}}$. 
Consider an \gls{attack} that perturbs up to $\poisoningrate \cdot |\neighbourhood{\nodeidx}|$ of these 
\glspl{modelparam} (see Figure \ref{fig_perturbed_params}). It turns out that an effective defense 
against these perturbations is to replace the average by \cite{RobustTrmmedMean}
\begin{equation}
	\label{equ_def_trimmed_mean}
 (1/\nodedegree{\nodeidx})  \sum_{\nodeidx' \in \neighbourhood{\nodeidx}}  \edgeweight_{\nodeidx,\nodeidx'} \thresholdfct(\localparamsiter{\nodeidx'}{\iteridx}),   
\end{equation} 
with some generalized threshold (or clipping) function $\thresholdfct$. The literature on 
robust \gls{fl} has studied different constructions of $\thresholdfct$ \cite{FLTrust2021,RobustTrmmedMean}. 
Intuitively, the threshold function $\thresholdfct$ should not change clean \glspl{modelparam} 
$\localparamsiter{\nodeidx'}{\iteridx}$ but also limit the impact of perturbed $\localparamsiter{\nodeidx'}{\iteridx}$. 

For the special case of \gls{model} dimension, i.e., each \gls{localmodel} is 
parametrized by a single number $w \in \mathbb{R}$, one useful choice 
for $\thresholdfct$ in \eqref{equ_def_trimmed_mean} is 
\begin{equation} 
	\label{equ_def_clipping}
	\thresholdfct (w) = \begin{cases} \tau_{u} & \mbox{ for } w \geq \tau_{u} \\ 
		w & \mbox{ for } w \in [\tau_{l},\tau_{u}] \\ 
		a & \mbox{ for } w \leq \tau_{l}. \end{cases}
\end{equation} 
A natural choice for the thresholds $\tau_{l},\tau_{u}$ is to use order statistic of the values 
$\localparamsiter{\nodeidx'}{\iteridx}$, for $\nodeidx' \in \neighbourhood{\nodeidx}$. 
In particular, the upper threshold $\tau_{u}$ in \eqref{equ_def_clipping} is chosen such that 
it is exceeded by $\localparamsiter{\nodeidx'}{\iteridx}$ only for a small number of 
\glspl{neighbor} $\nodeidx' \in \neighbourhood{\nodeidx}$. The lower threshold $\tau_{l}$ 
in \eqref{equ_def_clipping} is chosen analogously (see Figure \ref{fig_perturbed_params}). 
The \gls{robustness} of using \eqref{equ_def_clipping} in the averaging step \eqref{equ_def_trimmed_mean} 
has been studied recently in \cite{RobustTrmmedMean}. 

Another important choice for the threshold function $\thresholdfct$ in 
\eqref{equ_def_trimmed_mean} is 
\begin{align} 
\label{equ_def_trimmed_mean_clipping}
		\thresholdfct(w) & = c \begin{cases} w & \mbox{ if } w \in \trimmedset \\ 
		      0 & \mbox{ otherwise,} \end{cases}  \nonumber \\ 
		      & \mbox{ with } 
			  c = \frac{| \neighbourhood{\nodeidx}|}{|\{ \nodeidx' \in \neighbourhood{\nodeidx}: \localparamsiter{\nodeidx'}{\iteridx} \in \trimmedset \}|}.
\end{align} 
Inserting \eqref{equ_def_trimmed_mean_clipping} into \eqref{equ_def_trimmed_mean} yields the trimmed 
mean \cite{Stigler1973}. Indeed, the effect of \eqref{equ_def_trimmed_mean_clipping} is that 
the average \eqref{equ_def_trimmed_mean} is computed over a subset (or trimmed version) 
$\trimmedset$ of $\localparamsiter{\nodeidx'}{\iteridx}$, for $\nodeidx' \in \neighbourhood{\nodeidx}$. 
Different constructions for the subset $\trimmedset$ in \eqref{equ_def_trimmed_mean_clipping} have 
been studied in the literature on robust \gls{fl} \cite{Auror2016,FLARE2022,Fang2024}. One such 
construction is based on the order statistic of $\localparamsiter{\nodeidx'}{\iteridx}$, for $\nodeidx' \in \neighbourhood{\nodeidx}$, 
by excluding the most extreme values \cite{pmlr-v80-yin18a}.

Note that \eqref{equ_def_trimmed_mean_clipping} is defined for scalar 
\glspl{modelparam} $\localparamsiter{\nodeidx'}{\iteridx}  \in \mathbb{R}$ (i.e., for \gls{localmodel}s with 
dimension $\dimlocalmodel=1$). We can generalize \eqref{equ_def_trimmed_mean_clipping} to 
higher dimensions $\dimlocalmodel>1$ by applying it separately to each entry 
$w^{(\nodeidx,\iteridx)}_{1},\ldots,w^{(\nodeidx,\iteridx)}_{\dimlocalmodel}$ of the 
\glspl{modelparam} $\localparamsiter{\nodeidx}{\iteridx}$. The robustness of \gls{gtvmin}-based 
methods using the averaging step \eqref{equ_def_trimmed_mean} has been studied in \cite{pmlr-v80-yin18a}. 

\begin{figure}
	\centering
	\begin{tikzpicture}[scale=0.7, y=0.5cm, x=0.8cm]
		\begin{scope}
			\foreach \x/\y in {
				1/2, 2/3, 3/1, 4/6, 5/8, 6/5, 7/8, 8/4, 9/7, 10/2
			} {
				\draw[dashed, gray] (\x, 0) -- (\x, \y);
				\filldraw[blue] (\x, \y) circle (2pt);
				\node[circle, inner sep=0pt] (ptA\x) at (\x, \y) {};
			}
			\node[above right=2pt and 0pt, blue] at (ptA7) {$\localparamsiter{\nodeidx'}{\iteridx}$};
			\node[above right=2pt and 0pt, blue] at (ptA3) {};
			\node at (5.5, -4) {(a) Original (``clean'') \glspl{modelparam}.};
		\end{scope}
		
		\begin{scope}[xshift=11cm]
			\foreach \x/\y in {
				1/2, 2/3, 4/6, 5/8, 6/5, 8/4, 9/7, 10/2
			} {
				\draw[dashed, gray] (\x, 0) -- (\x, \y);
				\filldraw[blue] (\x, \y) circle (2pt);
				\node[circle, inner sep=0pt] (ptB\x) at (\x, \y) {};
			}
			\foreach \x/\y in {
				3/-1, 7/11
			} {
				\draw[dashed, gray] (\x, 0) -- (\x, \y);
				\filldraw[red] (\x, \y) circle (2pt);
				\node[circle, inner sep=0pt] (ptB\x) at (\x, \y) {};
			}
			\node[above right=2pt and 2pt] at (ptB7) {$\localparamsiter{\nodeidx'}{\iteridx}$};
			\node[below right=2pt and 2pt] at (ptB3) {};
			\draw[dashed] (0.5, 0.5) -- (10.5, 0.5) node[right] {$\tau_{l}$};;
			\draw[dashed] (0.5, 9.5) -- (10.5, 9.5)  node[right] {$\tau_{u}$};;
			\node at (5.5, -4) {(b) Poisoned \glspl{modelparam}.};
		\end{scope}
	\end{tikzpicture}
	\caption{An \gls{attack} on \eqref{equ_update_attack_model} perturbs (adversarially) a fraction 
		$\poisoningrate$ of the received \glspl{modelparam} $\localparamsiter{\nodeidx'}{\iteridx}$. \label{fig_perturbed_params}}
\end{figure}

So far, our discussion focused on protecting the update \eqref{equ_defence_update_perturb_sq_norm} 
(which is the core step of \gls{gtvmin}-based methods that use the \gls{gtv} penalty $\gtvpenalty(\cdot)=\normgeneric{\cdot}{2}^{2}$) 
against \glspl{dosattack} and \gls{backdoor} \glspl{attack}. We now discuss how to protect 
\eqref{equ_defence_update_perturb_sq_norm} against privacy \glspl{attack}. 

For a fixed time $\iteridx$, we can ensure a prescribed level of \gls{diffpriv}  
by replacing the update \eqref{equ_defence_update_perturb_sq_norm} with a noisy version
\begin{equation}  
	\label{equ_defence_update_perturb_sq_norm_noisy}
\localparamsiter{\nodeidx}{\iteridx+1} + \sigma \cdot \mathbf{n}^{(\iteridx)} \mbox{, with scaled noise } \sigma \cdot \mathbf{n}^{(\iteridx)}.
\end{equation}  
This noisy update \eqref{equ_defence_update_perturb_sq_norm_noisy} is then shared with the 
\glspl{neighbor} $\nodeidx' \in \neighbourhood{\nodeidx}$. Any (or even each) of these \glspl{neighbor} 
could be involved in a privacy \gls{attack} that aims to learn a \gls{sensattr} of the \gls{localdataset} $\localdataset{\nodeidx}$. 

The noise term $\mathbf{n}^{(\iteridx)}$ in \eqref{equ_defence_update_perturb_sq_norm_noisy} 
is drawn independently for each time $\iteridx$ from a prescribed \gls{probdist}, such as the 
Laplace distribution or the normal distribution \cite{AlgoFoundDP}. A key challenge for 
implementing \eqref{equ_defence_update_perturb_sq_norm_noisy} is to find a useful 
choice for the noise strength $\sigma$. Increasing $\sigma$ results in stronger \gls{privprot} 
but typically degrades the accuracy of the trained \gls{localmodel}s \cite{Near2025DPGuidelines}. 
However, choosing $\sigma$ too small can result in insufficient \gls{privprot}.

The minimum value $\sigma$ required to ensure $(\varepsilon,\delta)$-\gls{diffpriv} (see Definition \ref{equ_def_dp}) 
with prescribed values $\varepsilon,\delta \geq 0$ depends on 
\begin{itemize} 
	\item how the shape of the local \gls{lossfunc} $\locallossfunc{\nodeidx}{\cdot}$ changes when \gls{datapoint}s are 
	added to (or removed from) the \gls{localdataset} $\localdataset{\nodeidx}$ (see \cite{DPERM}), 
	\item the value of the \gls{gtvmin} parameter $\regparam$, 
	\item the number of time instants $\iteridx$ during which the 
	update \eqref{equ_defence_update_perturb_sq_norm} is executed and the noisy result 
	\eqref{equ_defence_update_perturb_sq_norm_noisy} shared with the \glspl{neighbor} \cite{pmlr-v37-kairouz15,DworkBoostingDP}.
\end{itemize}

\clearpage
\subsection{Exercises} 

\noindent\refstepcounter{problem}\label{prob:gradientreconstruction}\textbf{\theproblem. \Gls{modelinversion} for \gls{linreg}.}  
Consider an \gls{erm}-based method for training a \gls{linmodel} using plain \gls{gd}.  
Assume that the \glspl{modelparam} are initialized to zero, $\weights^{(0)} = \mathbf{0} \in \mathbb{R}^{\dimlocalmodel}$, 
and that the \gls{trainerr} $\emperror(\weights)$ 
is the average \gls{sqerrloss} on a \gls{trainset}, 
$$ \dataset = \big\{\pair{\featurevec^{(1)}}{\truelabel^{(1)}}, \ldots,\pair{\featurevec^{(\samplesize)}}{\truelabel^{(\samplesize)}} \big\} .$$
Suppose an attacker can observe the sequence of \gls{gradient}s, $\nabla \emperror\big( \weights^{(\iteridx)} \big)$ 
computed during the first few iterations $\iteridx=0,1,\ldots$.  
To what extent is it possible, based solely on the observed \gls{gradient}s and the knowledge of zero initialization, to 
reconstruct the \gls{trainset}?

\noindent\refstepcounter{problem}\label{prob:dosattack}\textbf{\theproblem. \Gls{dosattack}.} 
Construct an \gls{flnetwork} of \gls{fmi} stations and store it as a \texttt{networkx.Graph()} object. Implement 
Algorithm \ref{alg_fed_gd} to learn, for each node $\nodeidx=1,\ldots,\nrnodes$, the \glspl{modelparam} of a 
\gls{linmodel}. Launch a \gls{dosattack} by poisoning the \gls{localdataset}s at increasingly 
many nodes $\nodeidx' \neq 1$. The goal of the attack is to increase the \gls{valerr} of the 
learned \glspl{modelparam} $\localparams{1}$ (at target node $\nodeidx=1$) by $20$ \%.

\noindent\refstepcounter{problem}\label{prob:backdoorattack}\textbf{\theproblem. A \gls{backdoor} \gls{attack}.} 
We now use a different collection of \gls{feature}s for a \gls{datapoint} (representing a temperature recording). 
In particular, we replace the numeric \gls{feature} representing the hour of the measurement 
with $24$ new features, stacked into the vector $\featurevec'=\big(\feature'_{1},\ldots,\feature'_{24}\big)^{T}$. 
These new \gls{feature}s are the one-hot encoding of the hour. For example, if the 
temperature recording has been taking during hour $0$ then $\feature'_{1}=1,\feature_{2}' =0,\ldots$. 
Implement \gls{backdoor} \gls{attack} using a specific hour, e.g., 03:00 - 04:00, as the key (or trigger).

\noindent\refstepcounter{problem}\label{prob_abserrlossrobust}\textbf{\theproblem. Robust \gls{loss}.} 
Consider a \gls{ml} application with \gls{datapoint}s characterized by a single numeric \gls{feature} $\feature\!\in\!\mathbb{R}$ 
and single numeric \gls{label} $\truelabel \in \mathbb{R}$. To predict the \gls{label} we train a \gls{linmodel} 
via \gls{erm} with two different choices for the \gls{lossfunc}. In particular, we learn a \gls{hypothesis} 
$\hypothesis^{(1)}$ via \gls{erm} with the \gls{sqerrloss} and another \gls{hypothesis} $\hypothesis^{(2)}$ 
by \gls{erm} with the \gls{abserr}. Try to find a \gls{trainset}, consisting of five \glspl{datapoint} such 
that $\pair{\feature^{(5)}}{\truelabel^{(5)}}$ is located above the curve $\hypothesis^{(2)}$ (in a \gls{scatterplot}). 
Verify that $\hypothesis^{(2)}$ does not change at all when re-training the \gls{linmodel} on a modified 
\gls{trainset} where the value $\truelabel^{(5)}$ is slightly perturbed. 


\newpage


\renewcommand{\thefigure}{\arabic{figure}}
\setcounter{figure}{0}


\section*{Glossary}
\addcontentsline{toc}{section}{Glossary}
The terminology and notation used throughout this book follow the \textit{Dictionary of
Applied Machine Learning}, which is openly available and kept up to date at
\url{https://dictionaryofml.org}. Earlier versions of this book reproduced the dictionary
as an appendix; it is now referenced rather than copied, so that readers always get the
current definitions.
%

\newpage
\bibliographystyle{IEEEtran}
\bibliography{FLBookLit.bib}

\end{document}